\documentclass[10pt,twocolumn,letterpaper]{article}

\usepackage[pagenumbers]{cvpr} 

\makeatletter
\@namedef{ver@everyshi.sty}{}
\makeatother

\usepackage{graphicx}
\usepackage{grffile}
\usepackage{amsmath}
\usepackage{amssymb}
\usepackage{booktabs}
\usepackage{tikz}
\usepackage{tikz-3dplot}
\usepackage{tikzscale}
\usepackage{times}
\usepackage{epsfig}
\usepackage{extarrows}
\usepackage{algorithm}
\usepackage{algpseudocode}

\usepackage{bm}
\usepackage{subcaption}
\usepackage{todonotes}
\usepackage{tcolorbox}
\usetikzlibrary{arrows,shapes,chains,matrix,positioning}
\usetikzlibrary{scopes,decorations,shadows,backgrounds,fit}
\usetikzlibrary{decorations.pathreplacing,calc,3d,quotes,decorations.text}
\usepackage{pgfmath}
\usepackage{threeparttable}

\usepackage{array}
\usepackage{mdwmath}
\usepackage{mdwtab}
\usepackage{eqparbox}
\usepackage{fixltx2e}
\usepackage{stfloats}
\usepackage{url}
\usepackage{diagbox}
\usepackage{verbatim}
\usepackage{multirow}
\usepackage{mathtools}
\usepackage{breqn}
\usepackage{appendix}

\usepackage[pagebackref,breaklinks,colorlinks]{hyperref}

\tikzstyle{block} = [rectangle, draw, fill=blue!20, text centered]

\usepackage[capitalize]{cleveref}
\crefname{section}{Sec.}{Secs.}
\Crefname{section}{Section}{Sections}
\Crefname{table}{Table}{Tables}
\crefname{table}{Tab.}{Tabs.}

\def\etal{{\it et al. }}

\newcommand{\versus}[1]{{\it vs.}}

\DeclareMathOperator{\median}{median}

\graphicspath{{figs/}}

\makeatletter
\def\thickhline{%
  \noalign{\ifnum0=`}\fi\hrule \@height \thickarrayrulewidth \futurelet
   \reserved@a\@xthickhline}
\def\@xthickhline{\ifx\reserved@a\thickhline
               \vskip\doublerulesep
               \vskip-\thickarrayrulewidth
             \fi
      \ifnum0=`{\fi}}
\makeatother

\newlength{\thickarrayrulewidth}
\newcommand{\ourmethod}{Diverse3DFace}

\def\cvprPaperID{7804} 
\def\confName{CVPR}
\def\confYear{2022}

\begin{document}

\title{Generating Diverse 3D Reconstructions from a Single Occluded Face Image}

\author{Rahul Dey \quad Vishnu Naresh Boddeti\\
Michigan State University, East Lansing, MI\\
{\tt\small {deyrahul,vishnu}@msu.edu}
}

\setlength{\textfloatsep}{8pt plus 2.0pt minus 4.0pt}
\setlength{\floatsep}{10pt plus 2.0pt minus 2.0pt}
\captionsetup[table]{skip=0.1cm} 
\setlength{\abovecaptionskip}{8pt}  
\setlength{\belowcaptionskip}{0pt}

\twocolumn[{
\renewcommand\twocolumn[1][]{#1}%
\maketitle
\begin{center}
    \vspace{-2em}
    \centering
    \captionsetup{font=small}
    \begin{minipage}{\textwidth}
    \begin{tikzpicture}
    \node (b1) {\includegraphics[width=0.09\linewidth]{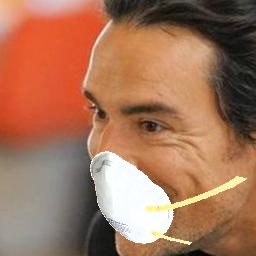}};
    \node[right of=b1, node distance=2.0cm] (b2) {\includegraphics[trim={400 80 400 100},clip,width=0.08\linewidth]{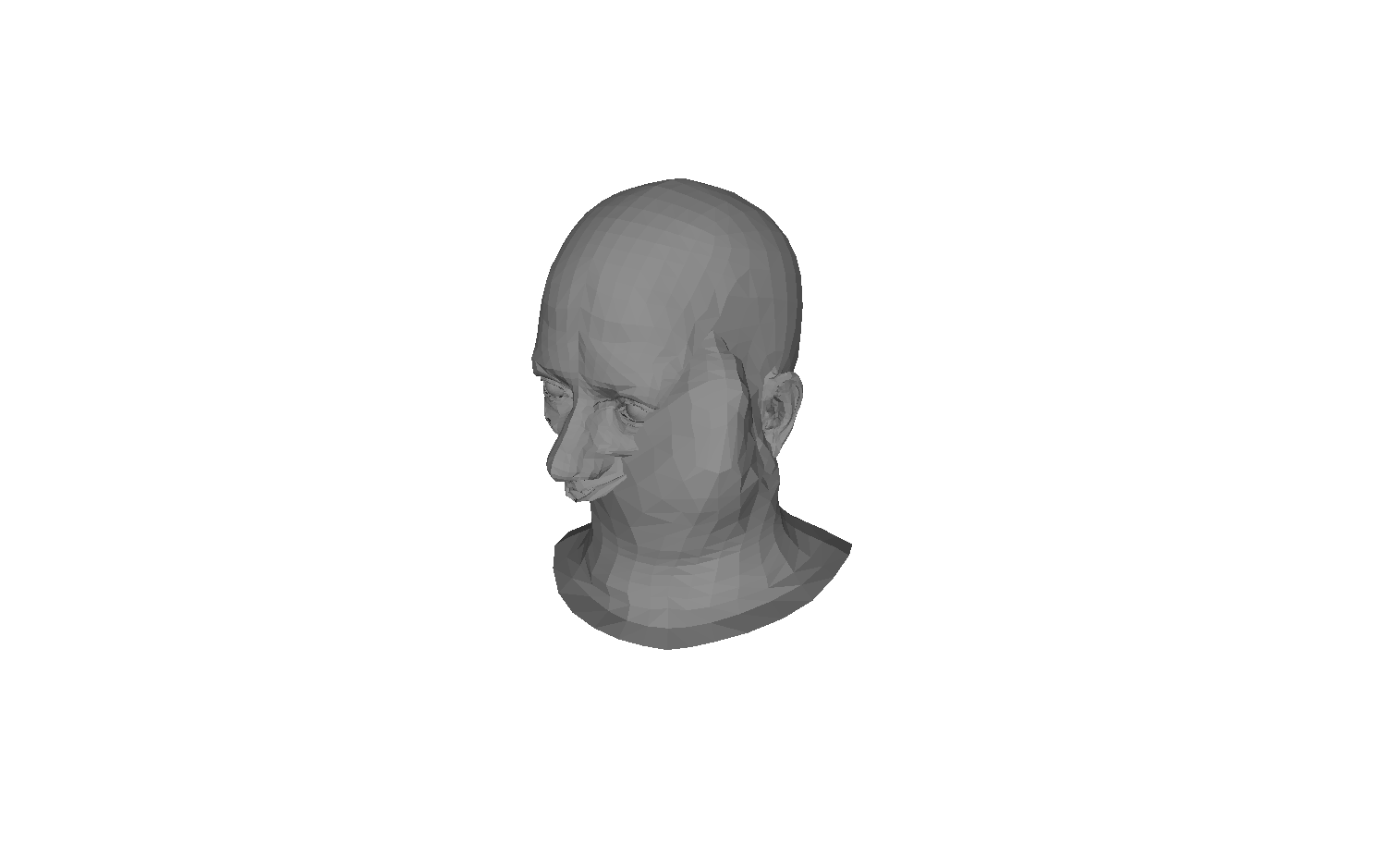}};
    \node[right of=b2, node distance=1.7cm] (b3) {\includegraphics[trim={400 80 400 100},clip,width=0.08\linewidth]{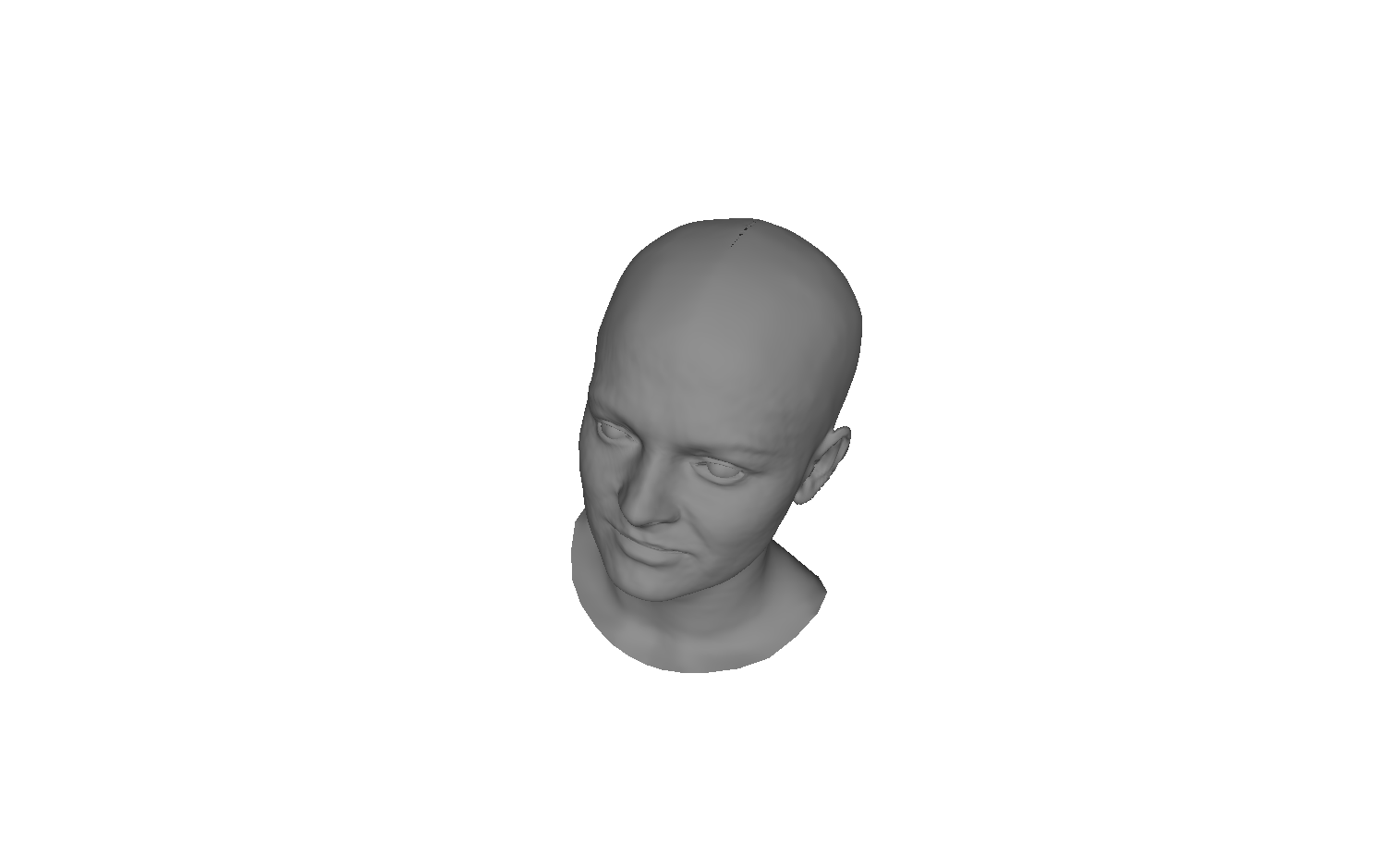}};
    \node[right of=b3, node distance=1.7cm] (b4) {\includegraphics[trim={400 80 400 100},clip,width=0.065\linewidth]{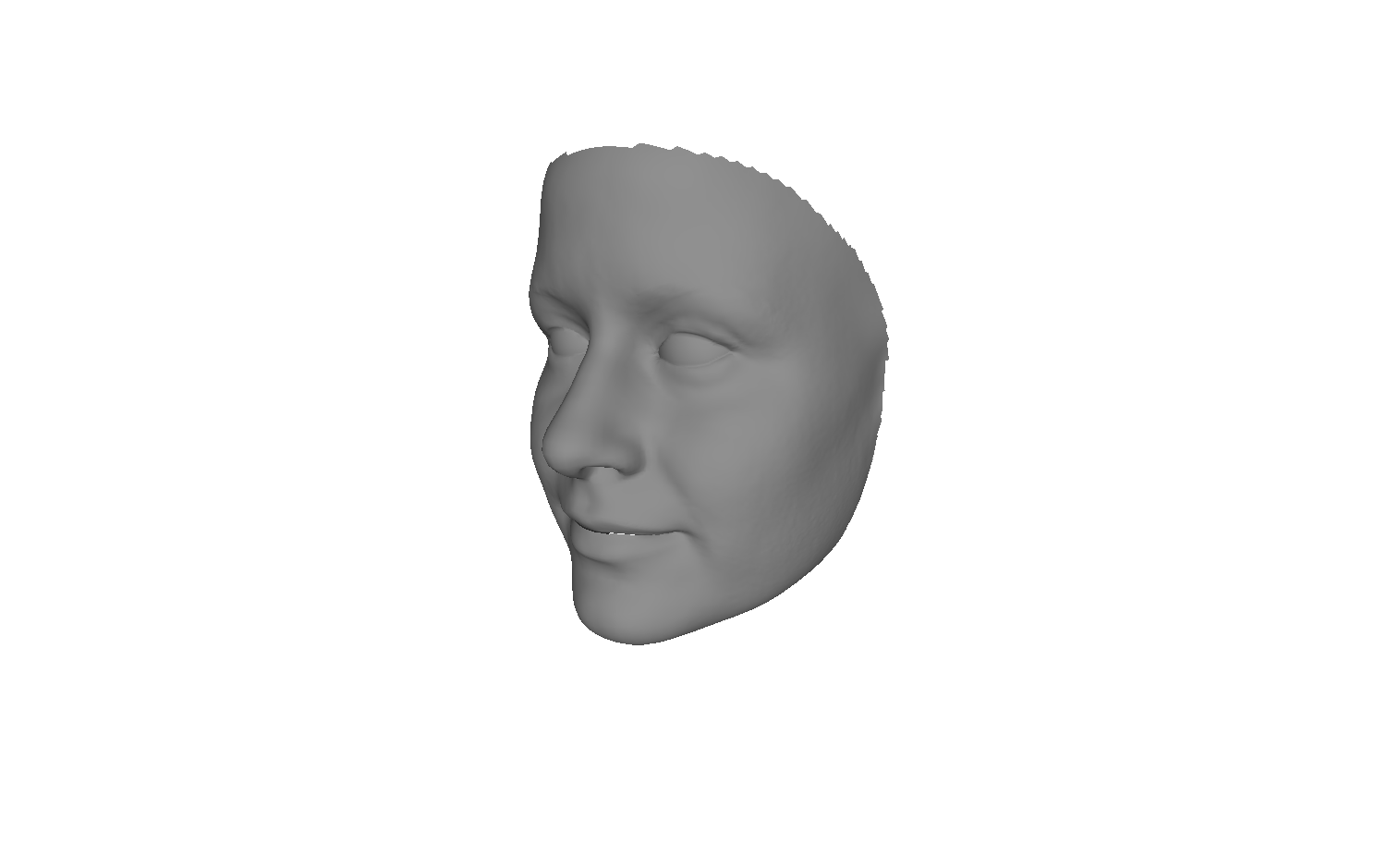}};
    \node[right of=b4, node distance=1.8cm] (b5) {\includegraphics[trim={400 80 400 100},clip,width=0.065\linewidth]{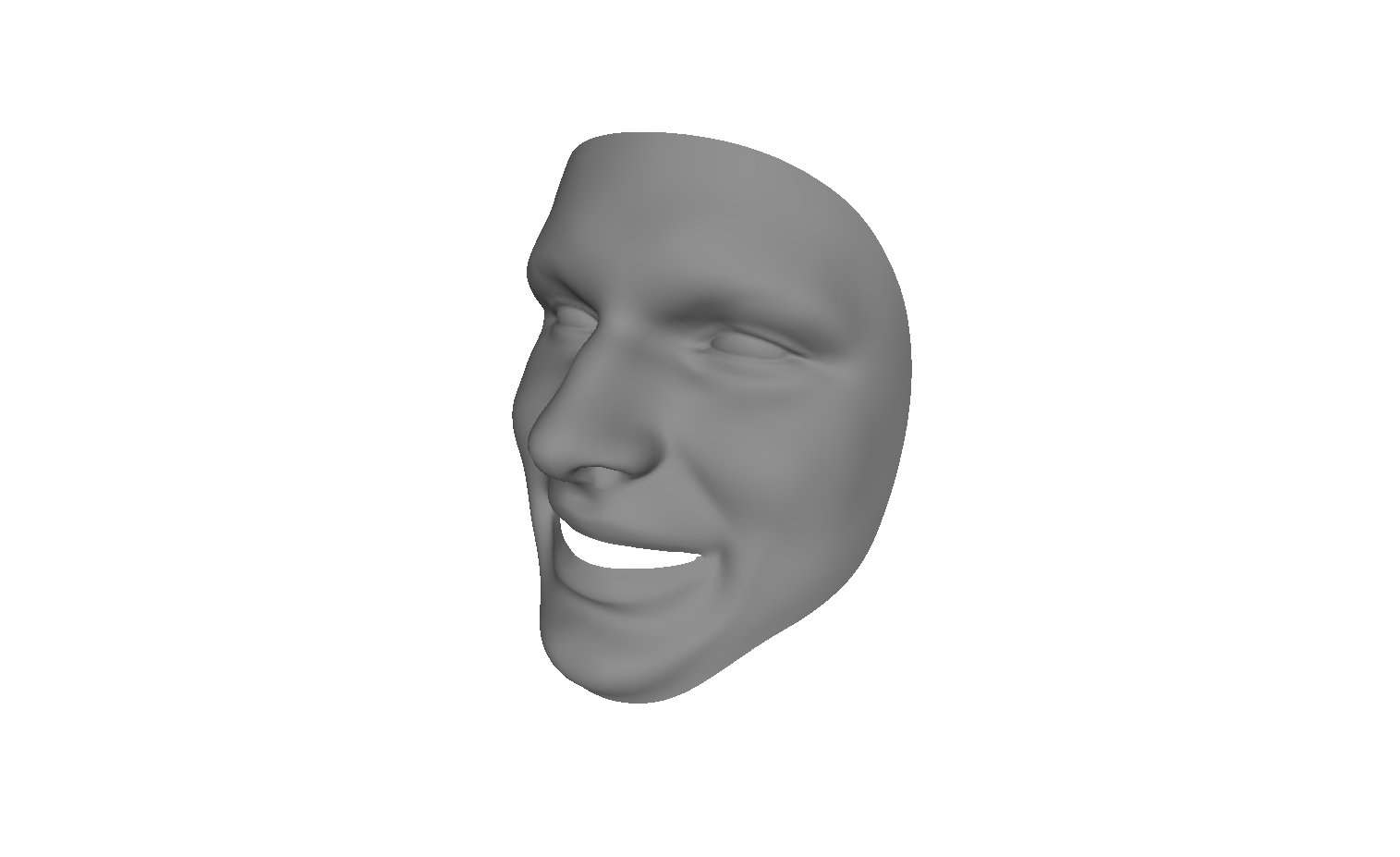}};
    \node[right of=b5, node distance=1.9cm] (b6) {\includegraphics[trim={400 80 400 100},clip,width=0.07\linewidth]{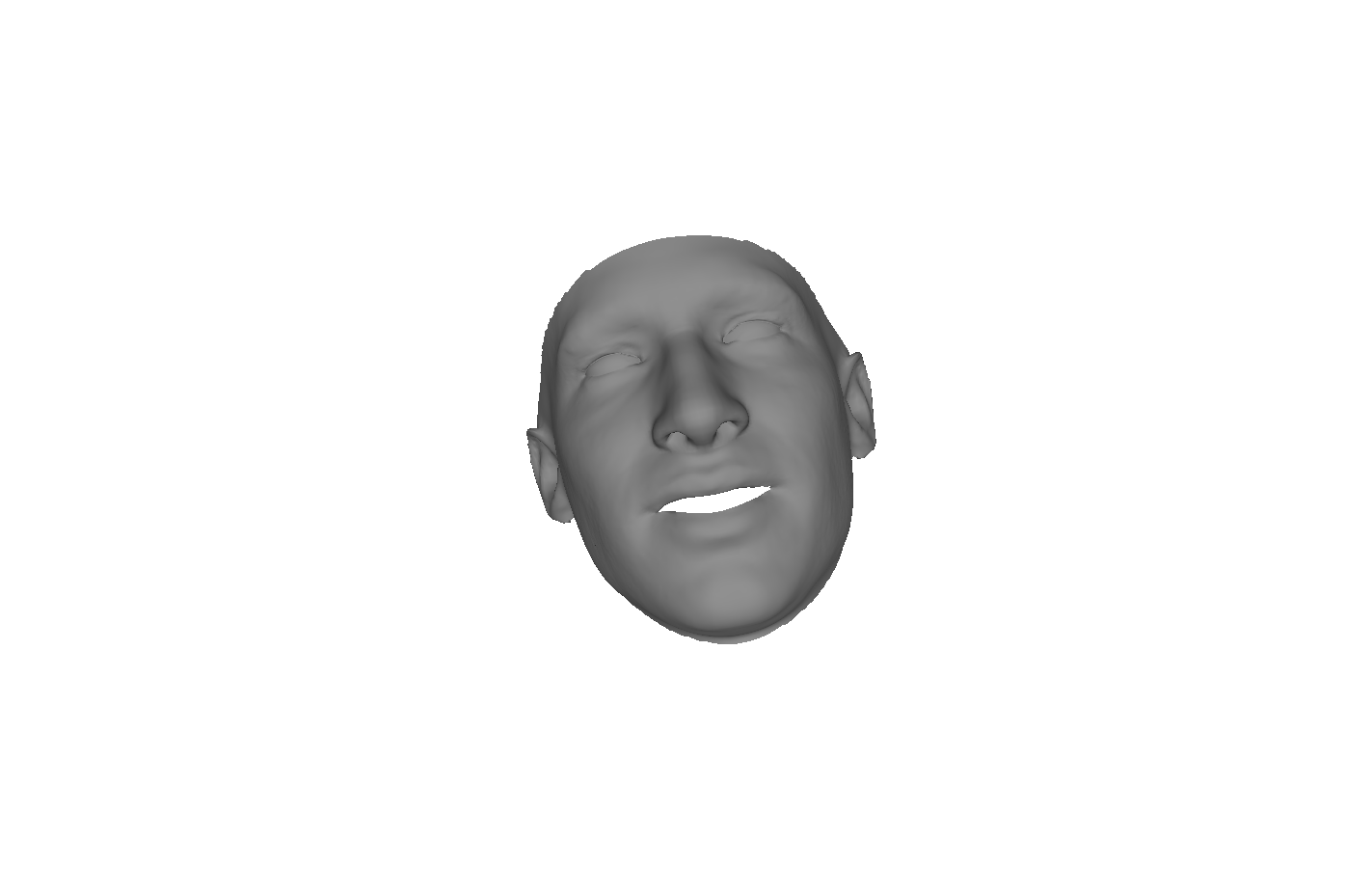}};
    \node[right of=b6, node distance=1.9cm] (b7) {\includegraphics[trim={400 80 400 100},clip,width=0.08\linewidth]{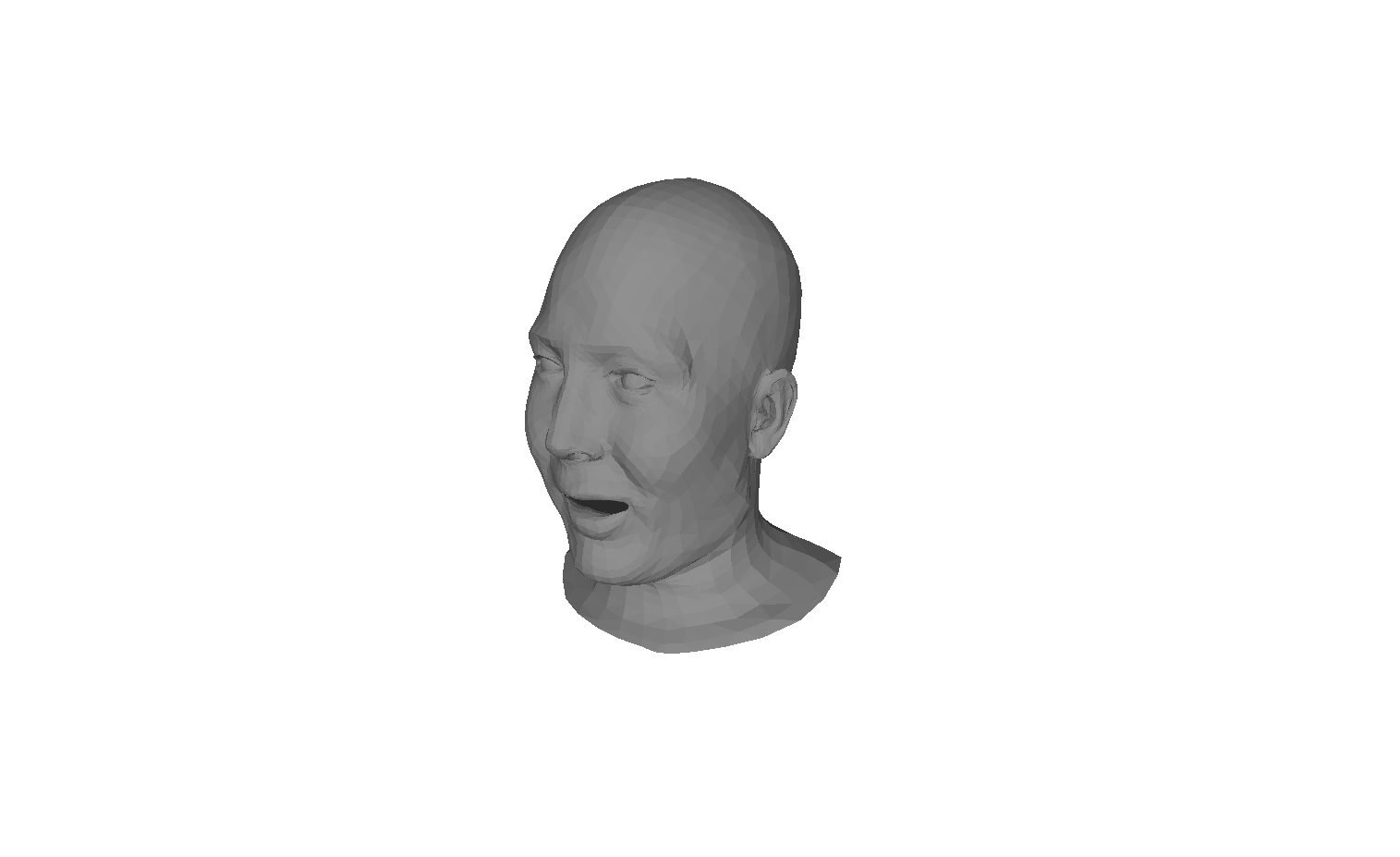}};
    \node[right of=b7, node distance=1.3cm] (b8) {\includegraphics[trim={400 80 400 100},clip,width=0.08\linewidth]{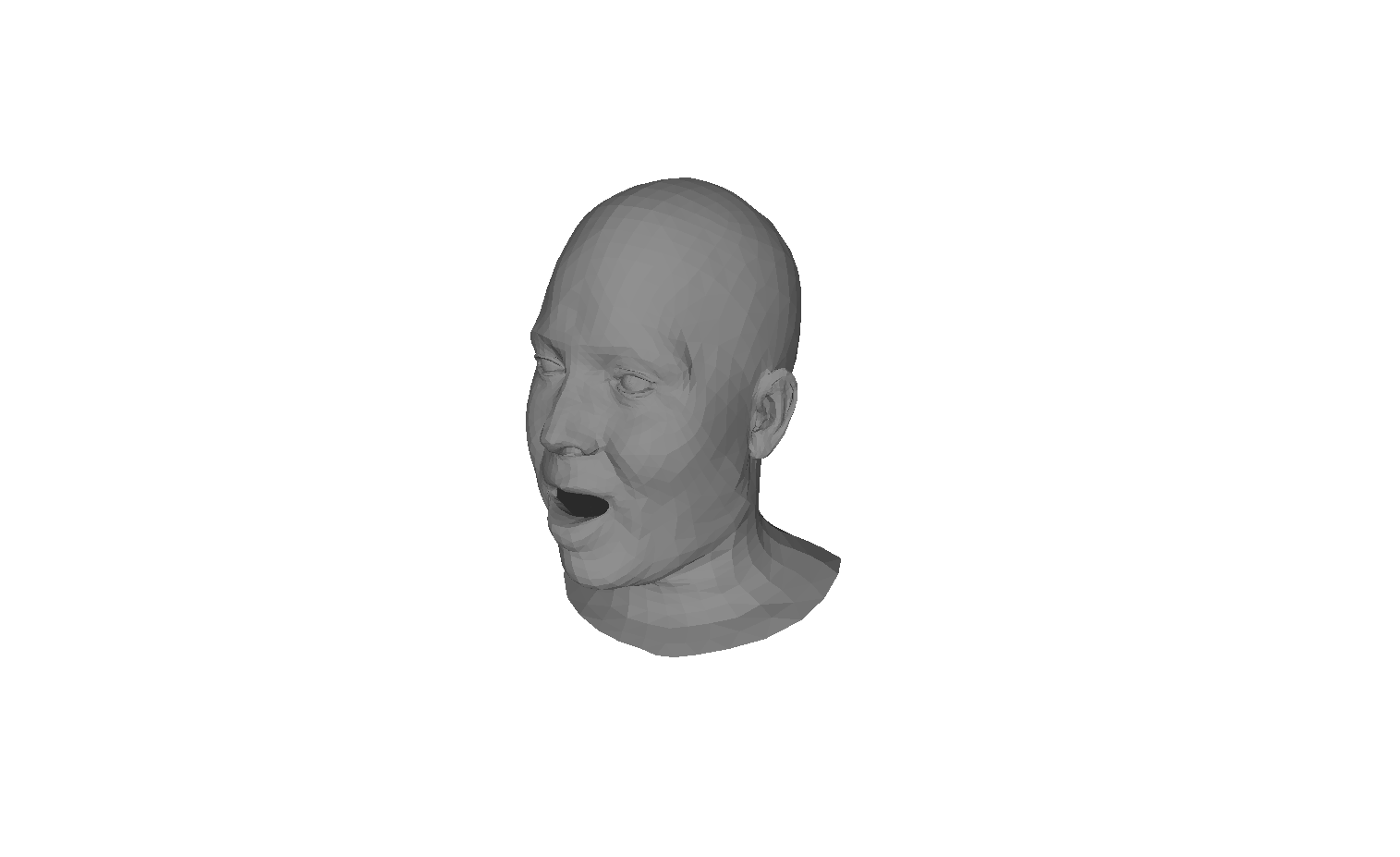}};
    \node[right of=b8, node distance=1.3cm] (b9) {\includegraphics[trim={400 80 400 100},clip,width=0.08\linewidth]{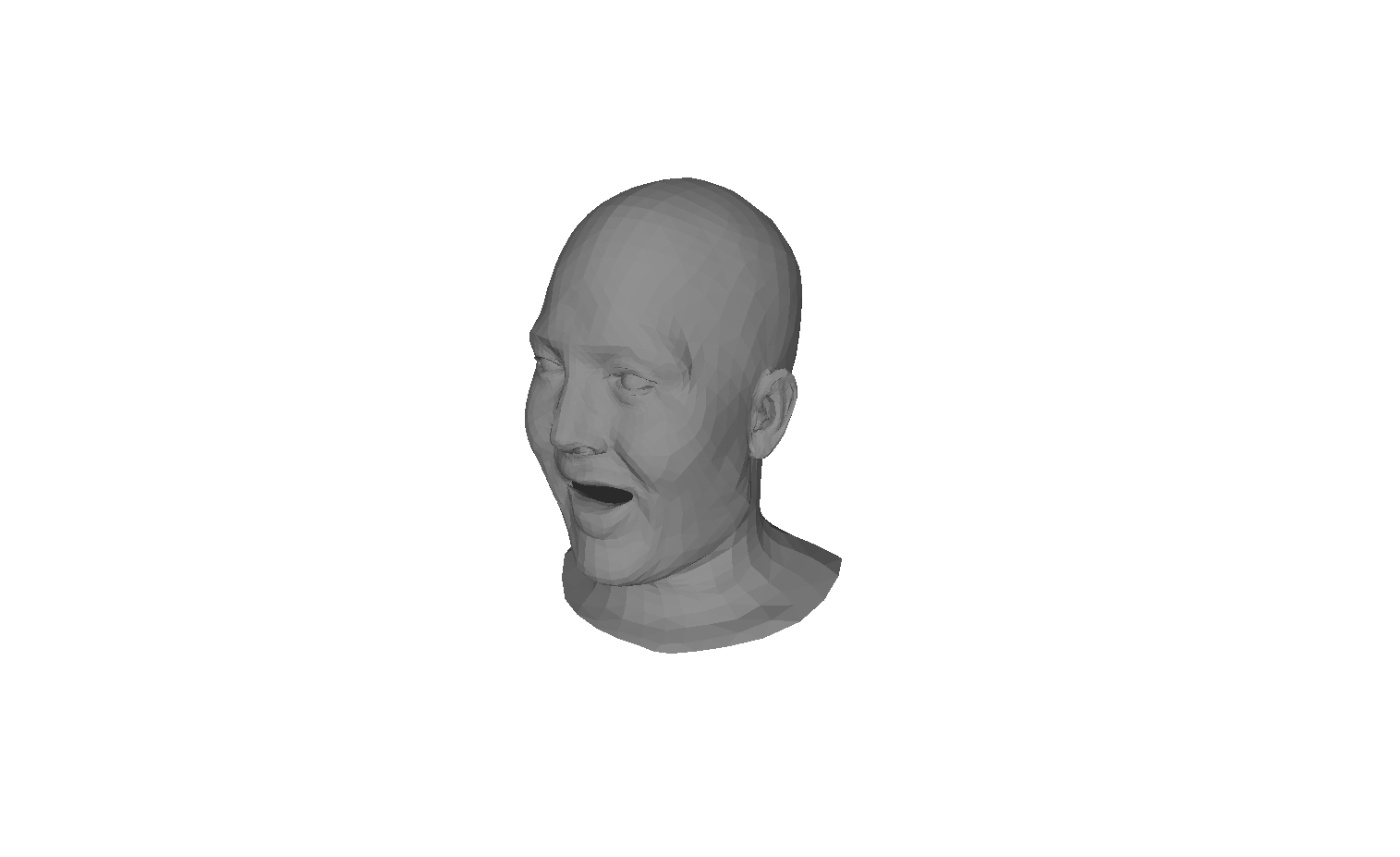}};
    \node[right of=b9, node distance=1.3cm] (b10) {\includegraphics[trim={400 80 400 100},clip,width=0.08\linewidth]{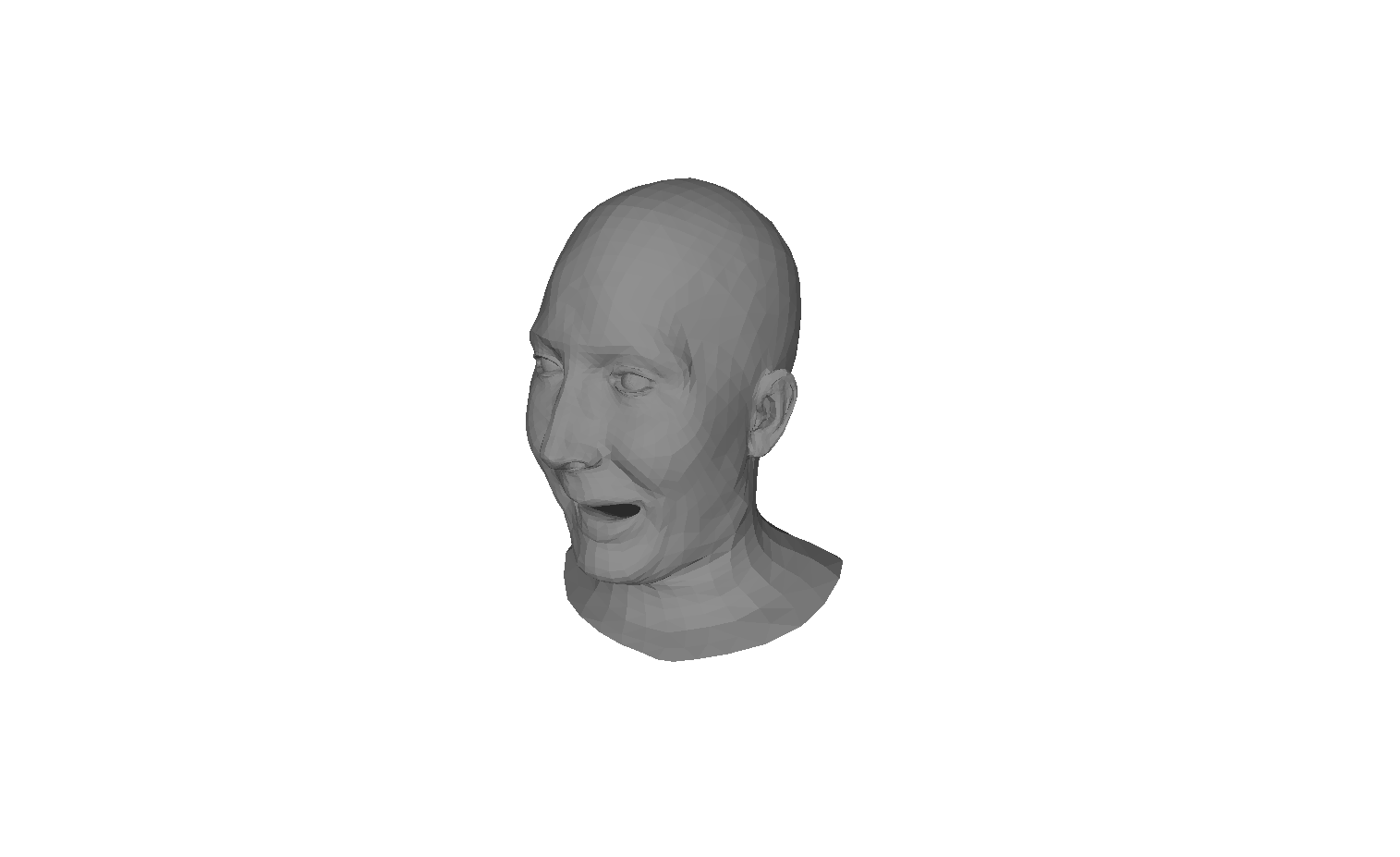}};
    \node[right of=b10, node distance=1.3cm] (b11) {\includegraphics[trim={400 80 400 100},clip,width=0.08\linewidth]{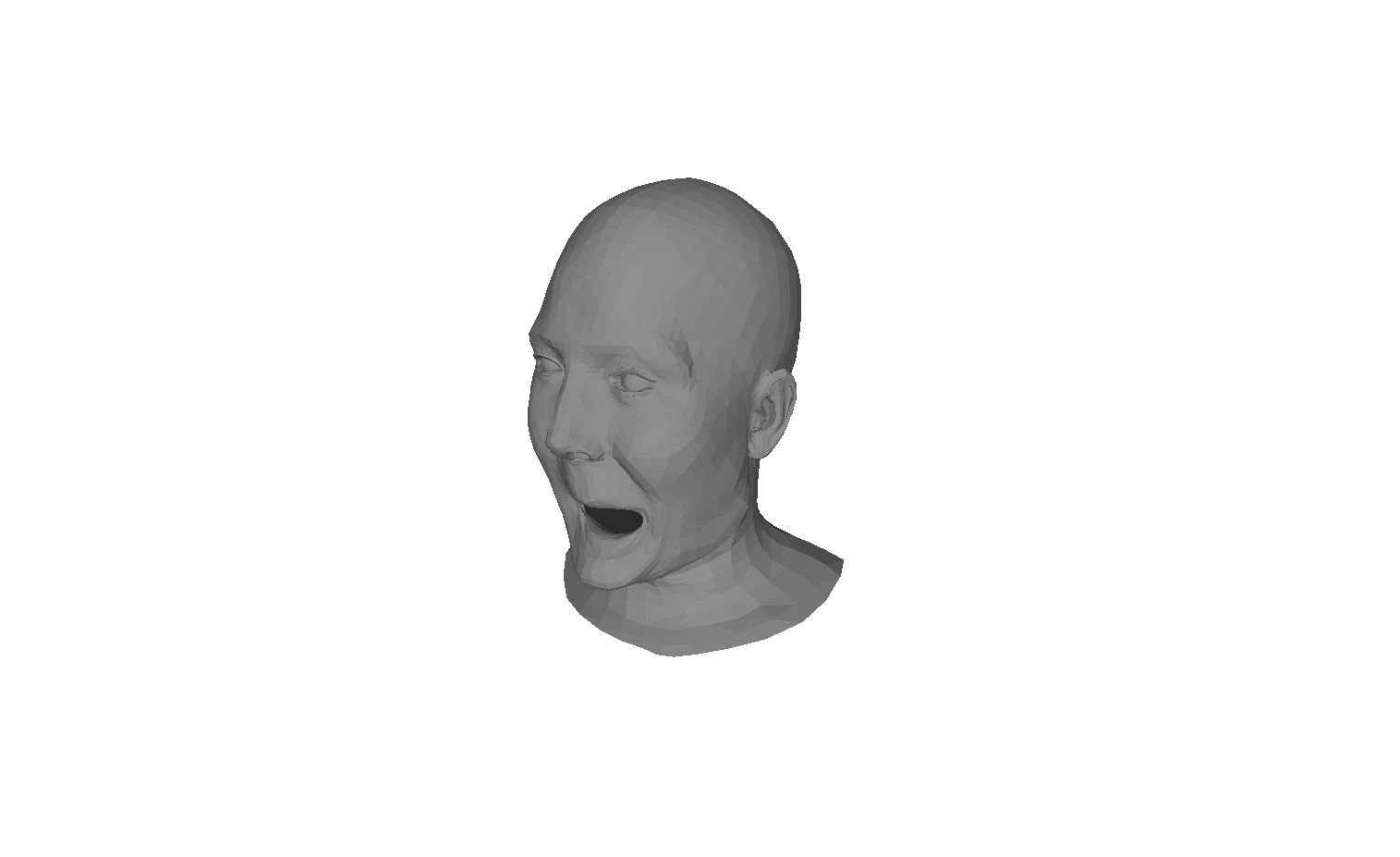}};
    
    \node[below of=b1, node distance=1.8cm] (d1) {\includegraphics[width=0.09\linewidth]{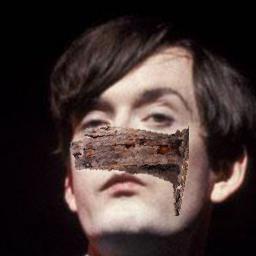}};
    \node[right of=d1, node distance=2.0cm] (d2) {\includegraphics[trim={400 80 400 100},clip,width=0.08\linewidth]{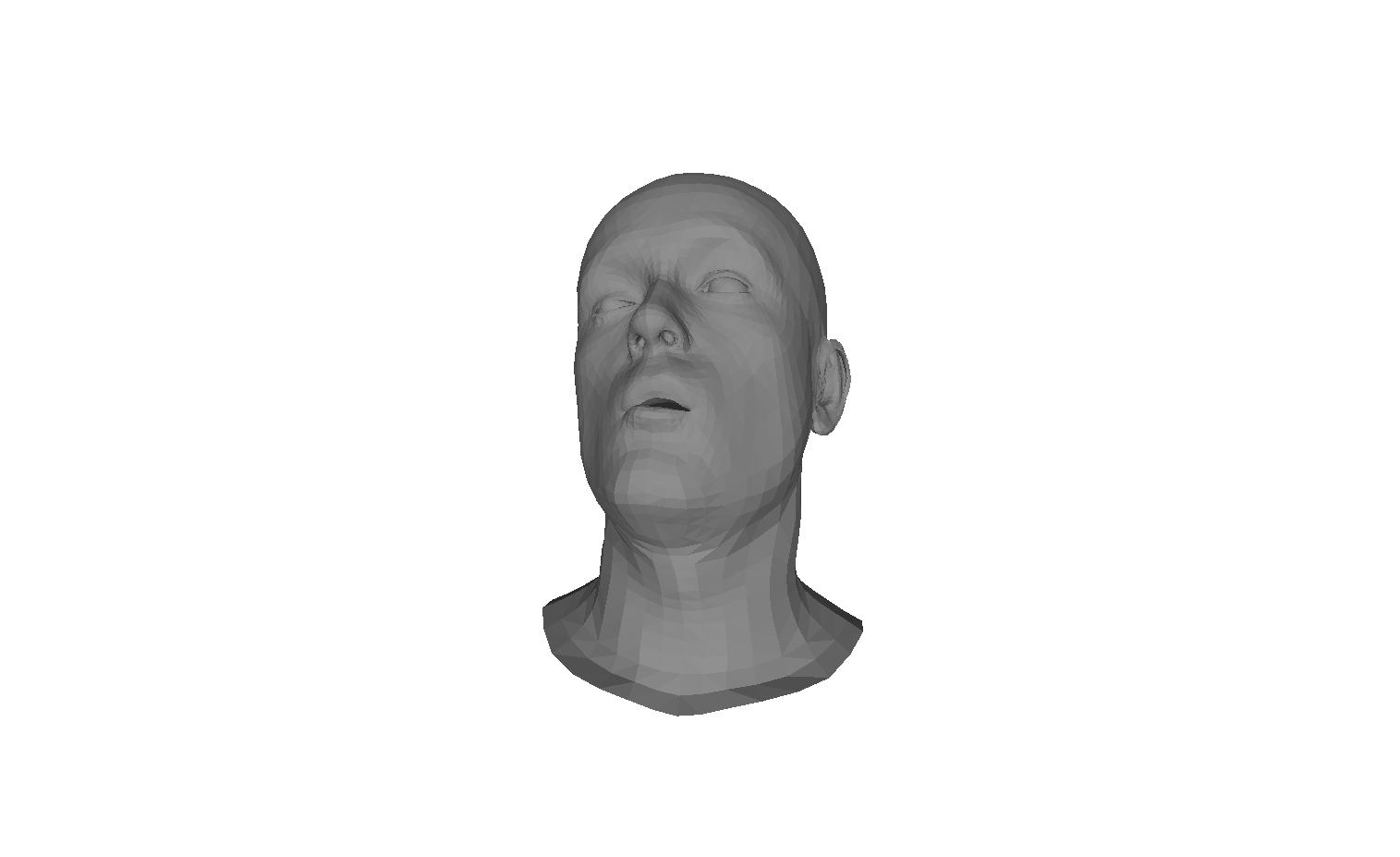}};
    \node[right of=d2, node distance=1.7cm] (d3) {\includegraphics[trim={400 80 400 100},clip,width=0.08\linewidth]{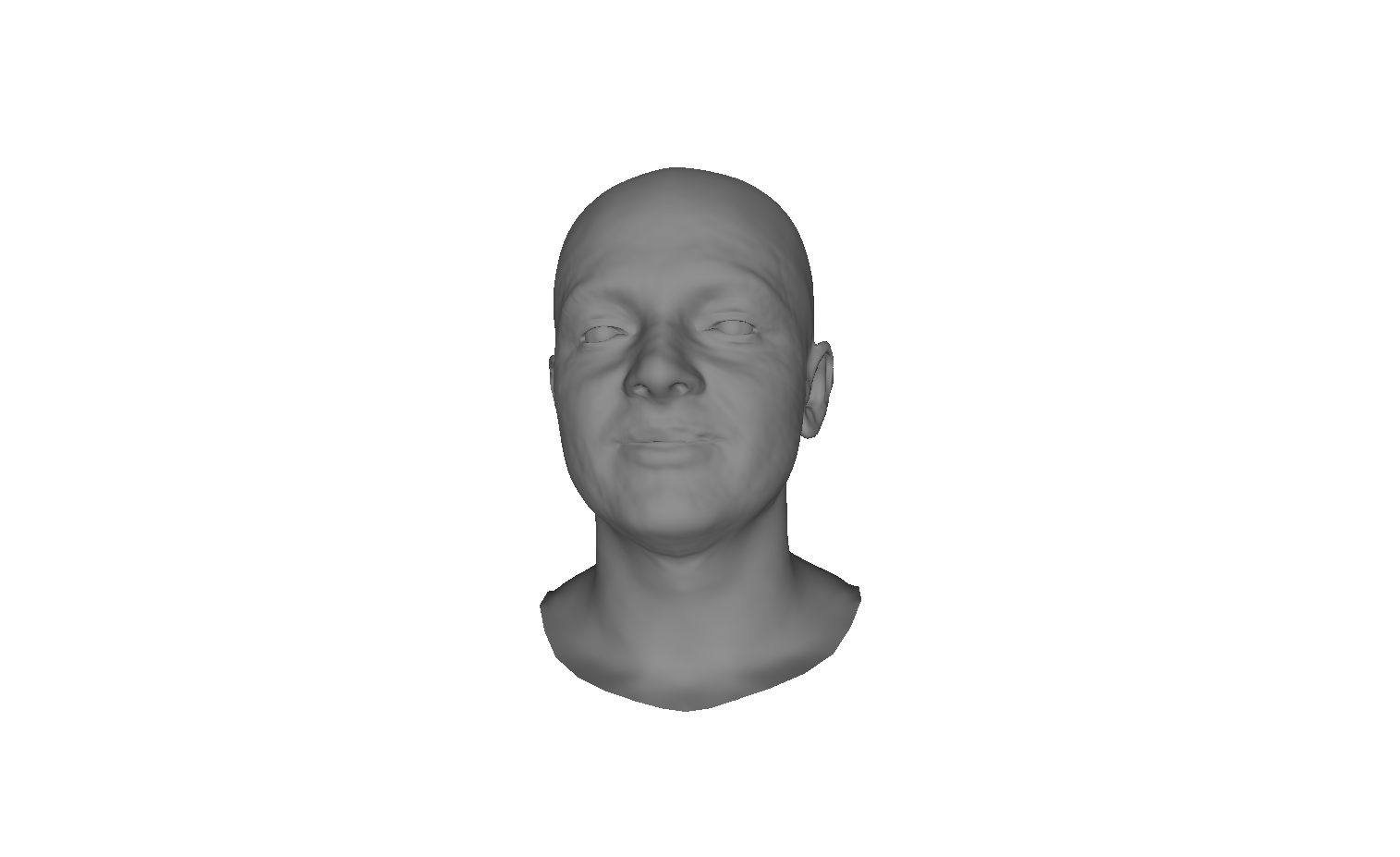}};
    \node[right of=d3, node distance=1.7cm] (d4) {\includegraphics[trim={400 80 400 100},clip,width=0.065\linewidth]{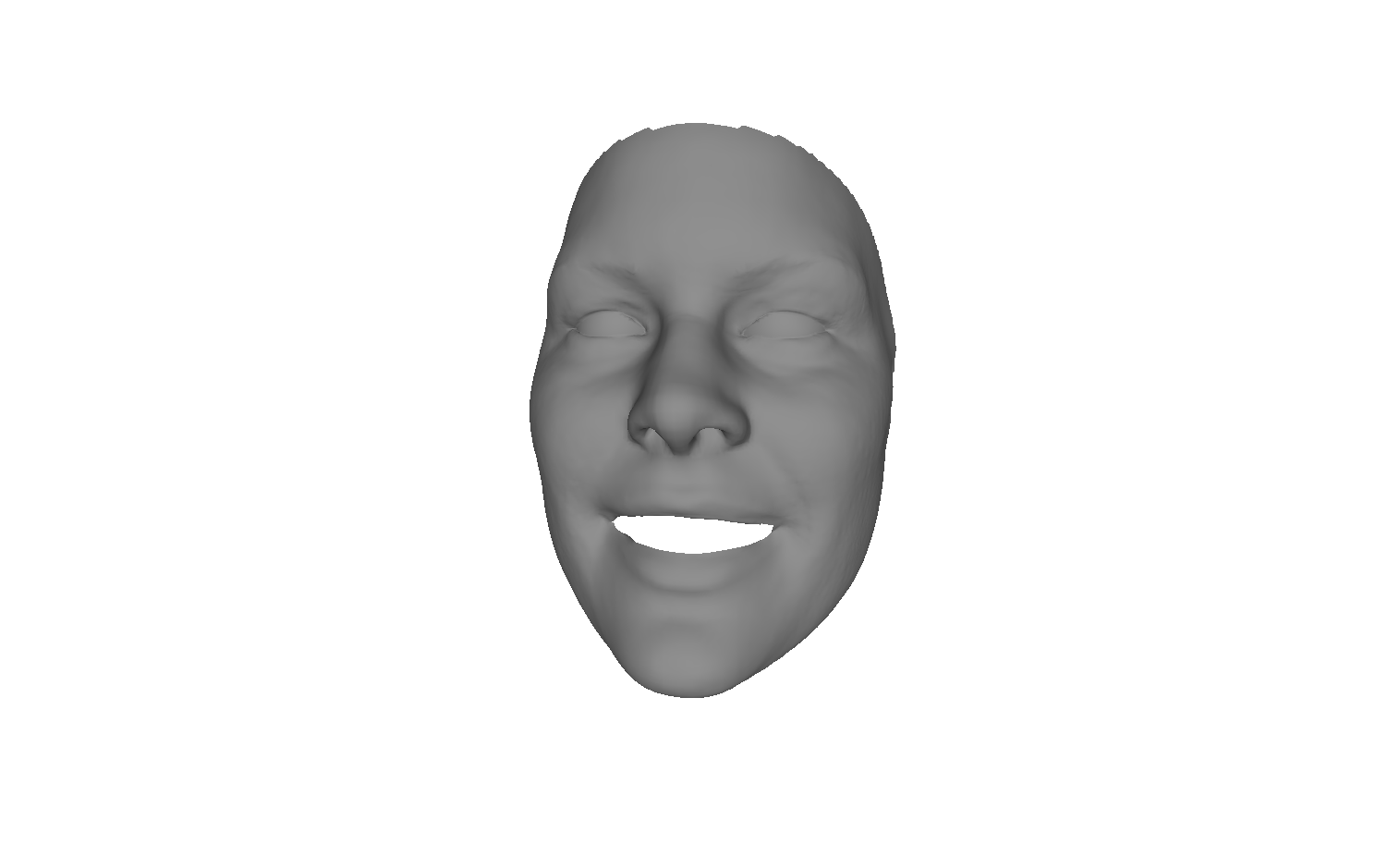}};
    \node[right of=d4, node distance=1.8cm] (d5) {\includegraphics[trim={400 80 400 100},clip,width=0.065\linewidth]{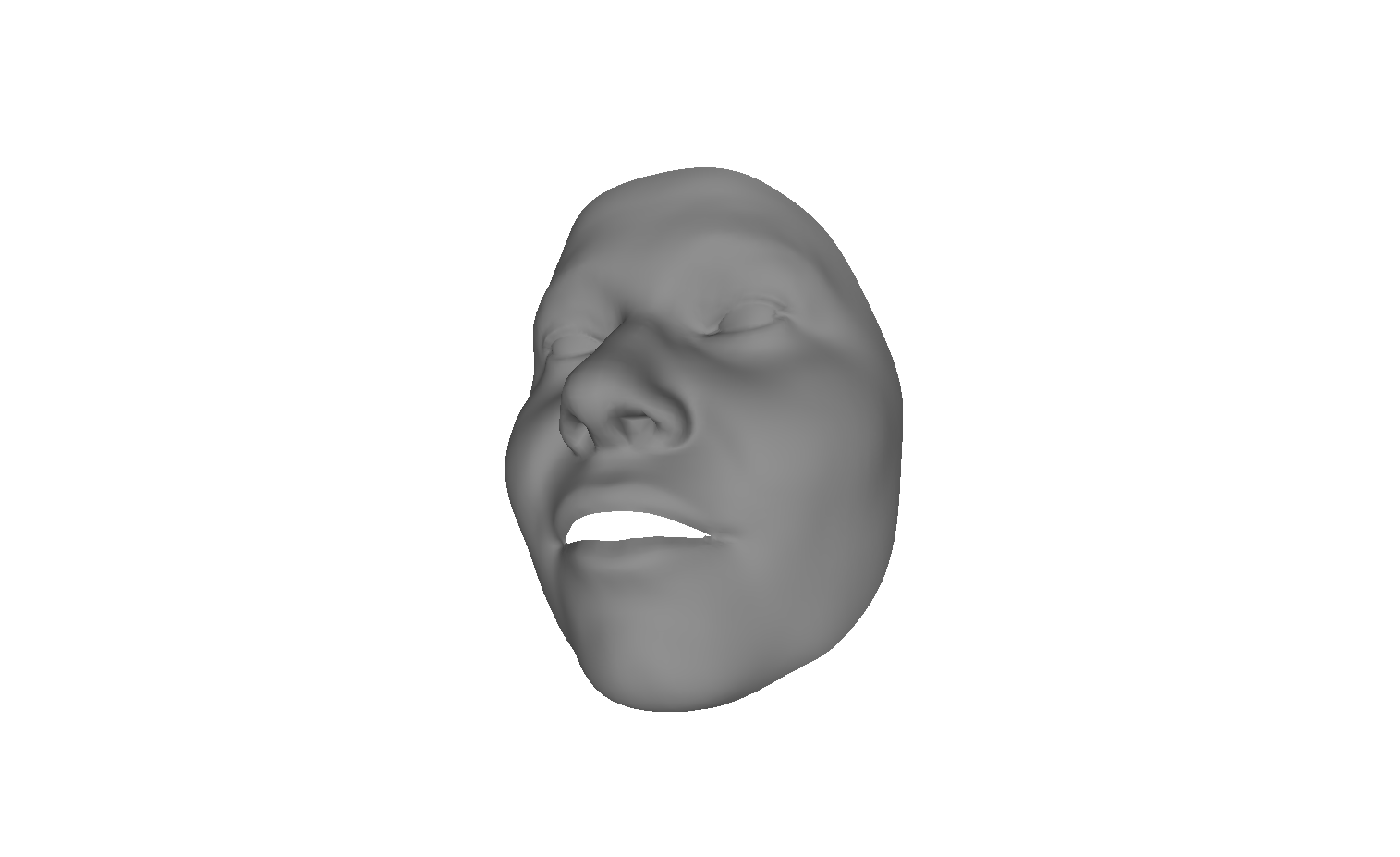}};
    \node[right of=d5, node distance=1.9cm] (d6) {\includegraphics[trim={400 80 400 100},clip,width=0.07\linewidth]{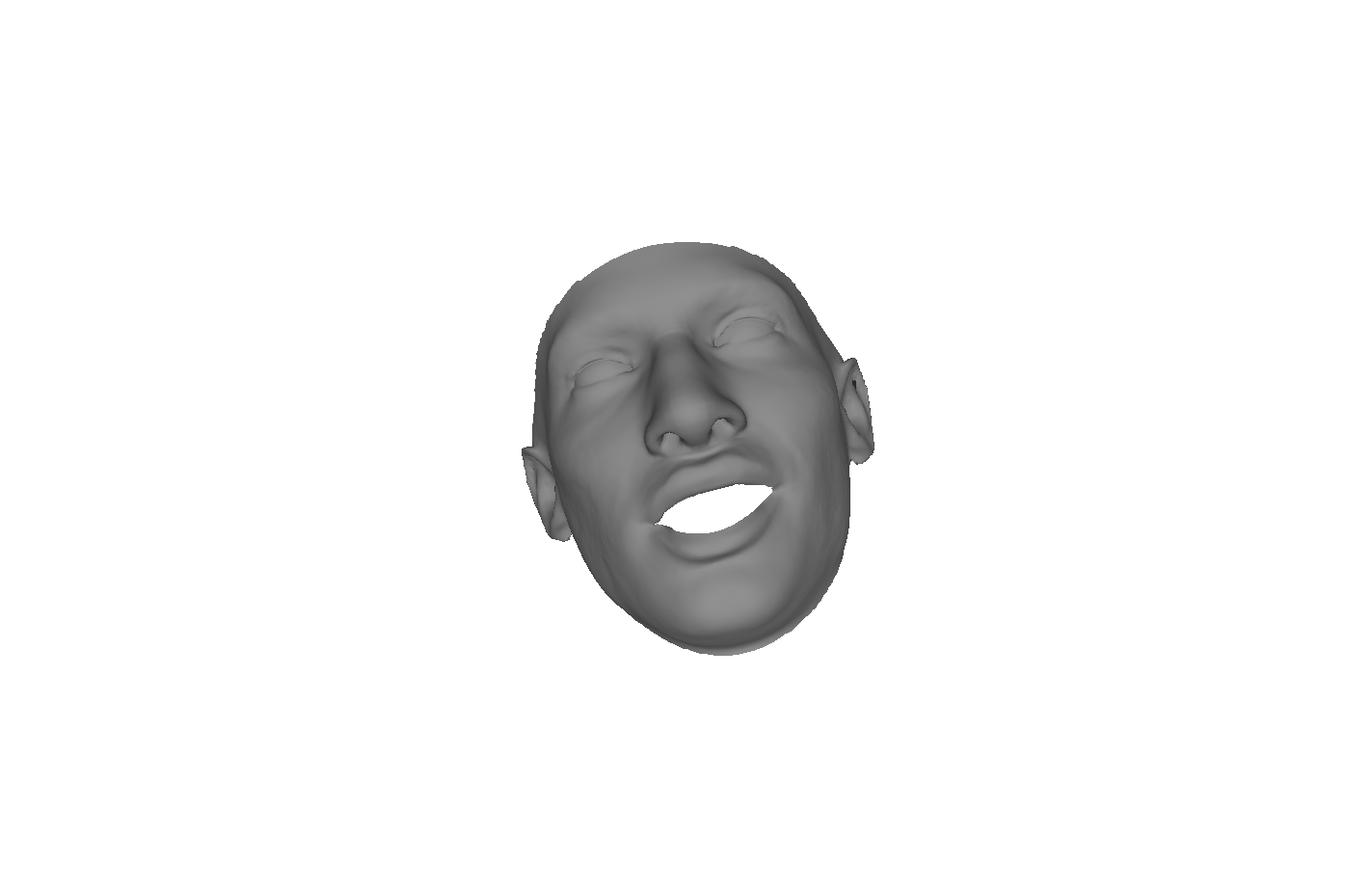}};
    \node[right of=d6, node distance=1.9cm] (d7) {\includegraphics[trim={400 80 400 100},clip,width=0.08\linewidth]{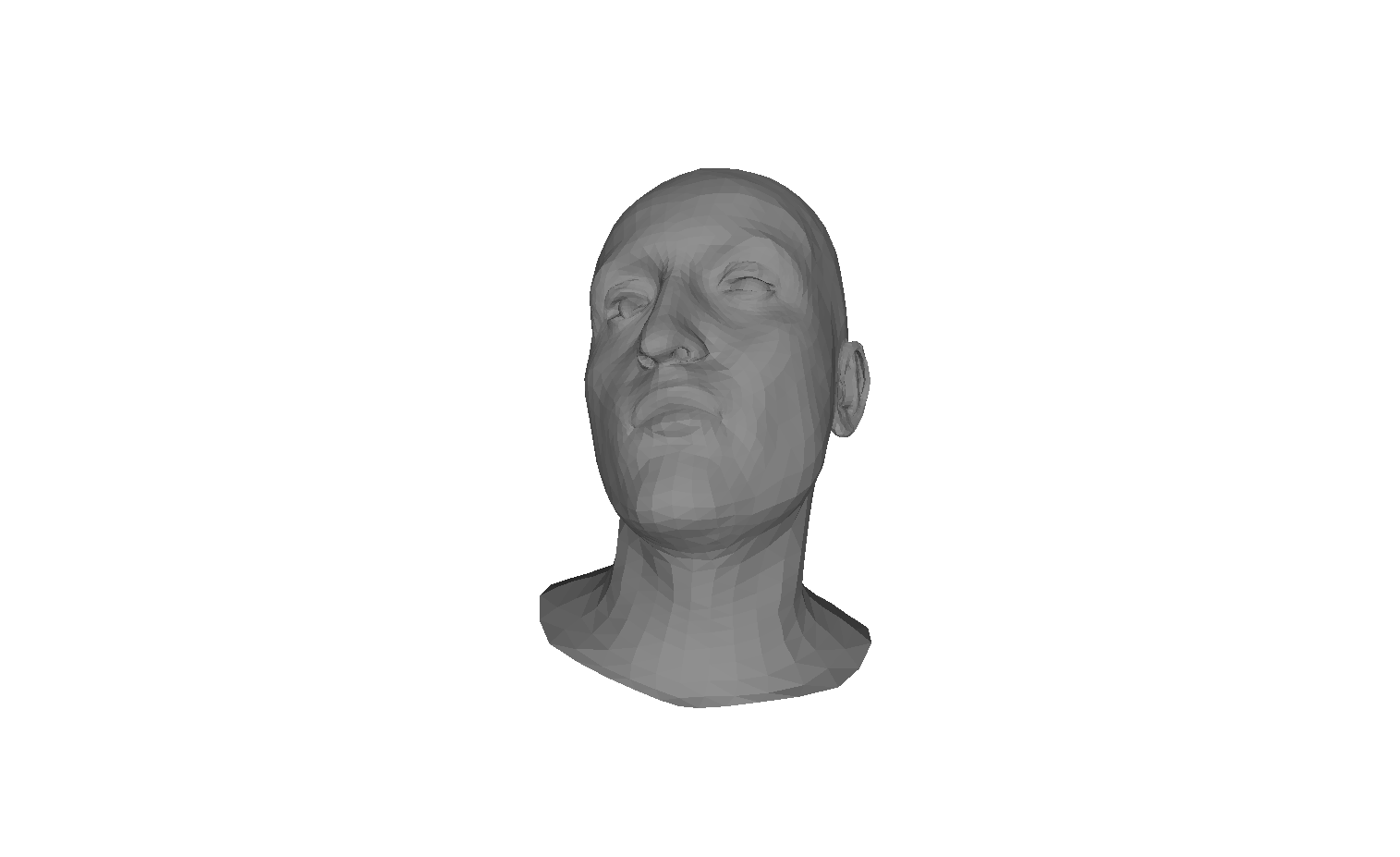}};
    \node[right of=d7, node distance=1.3cm] (d8) {\includegraphics[trim={400 80 400 100},clip,width=0.08\linewidth]{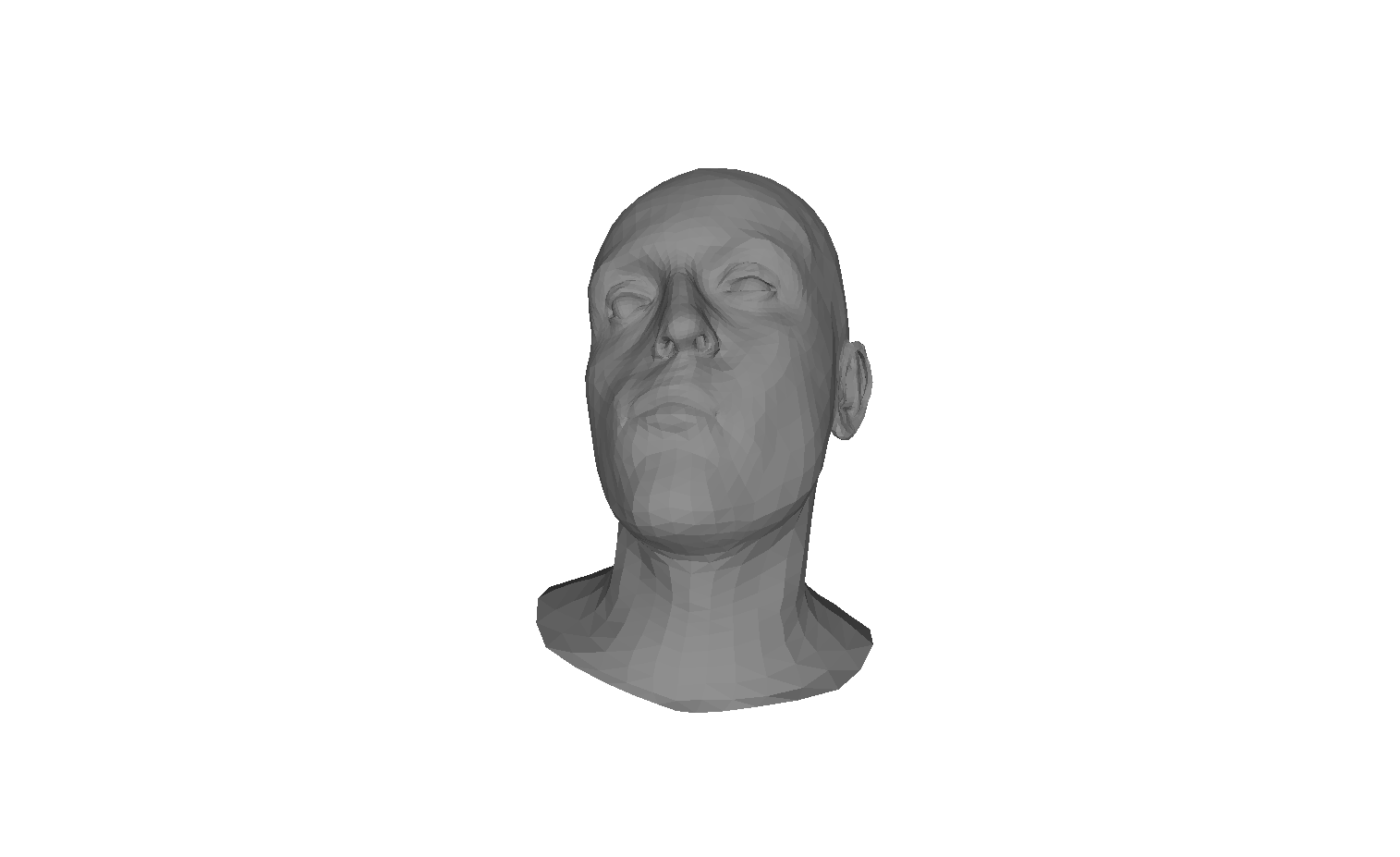}};
    \node[right of=d8, node distance=1.3cm] (d9) {\includegraphics[trim={400 80 400 100},clip,width=0.08\linewidth]{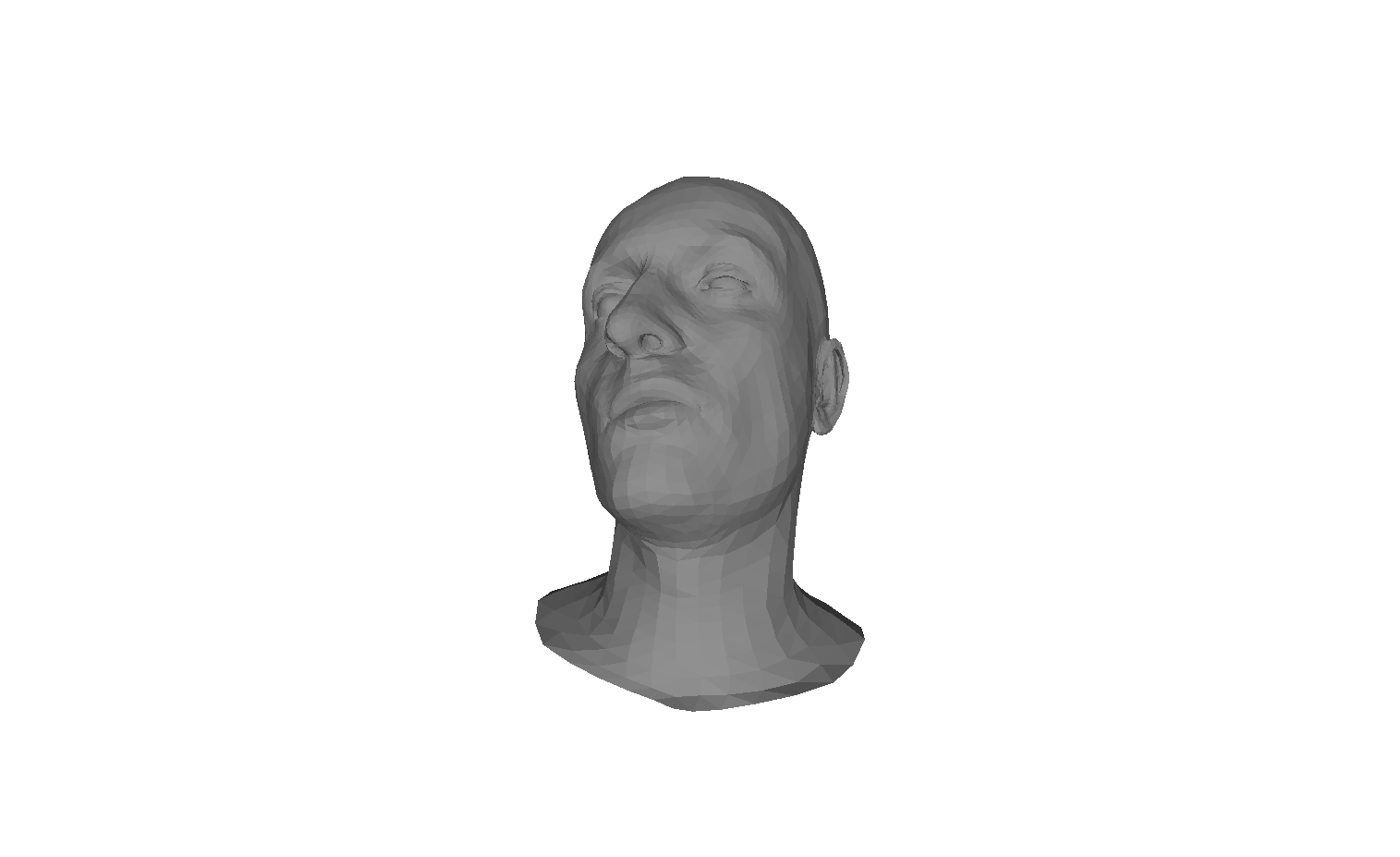}};
    \node[right of=d9, node distance=1.3cm] (d10) {\includegraphics[trim={400 80 400 100},clip,width=0.08\linewidth]{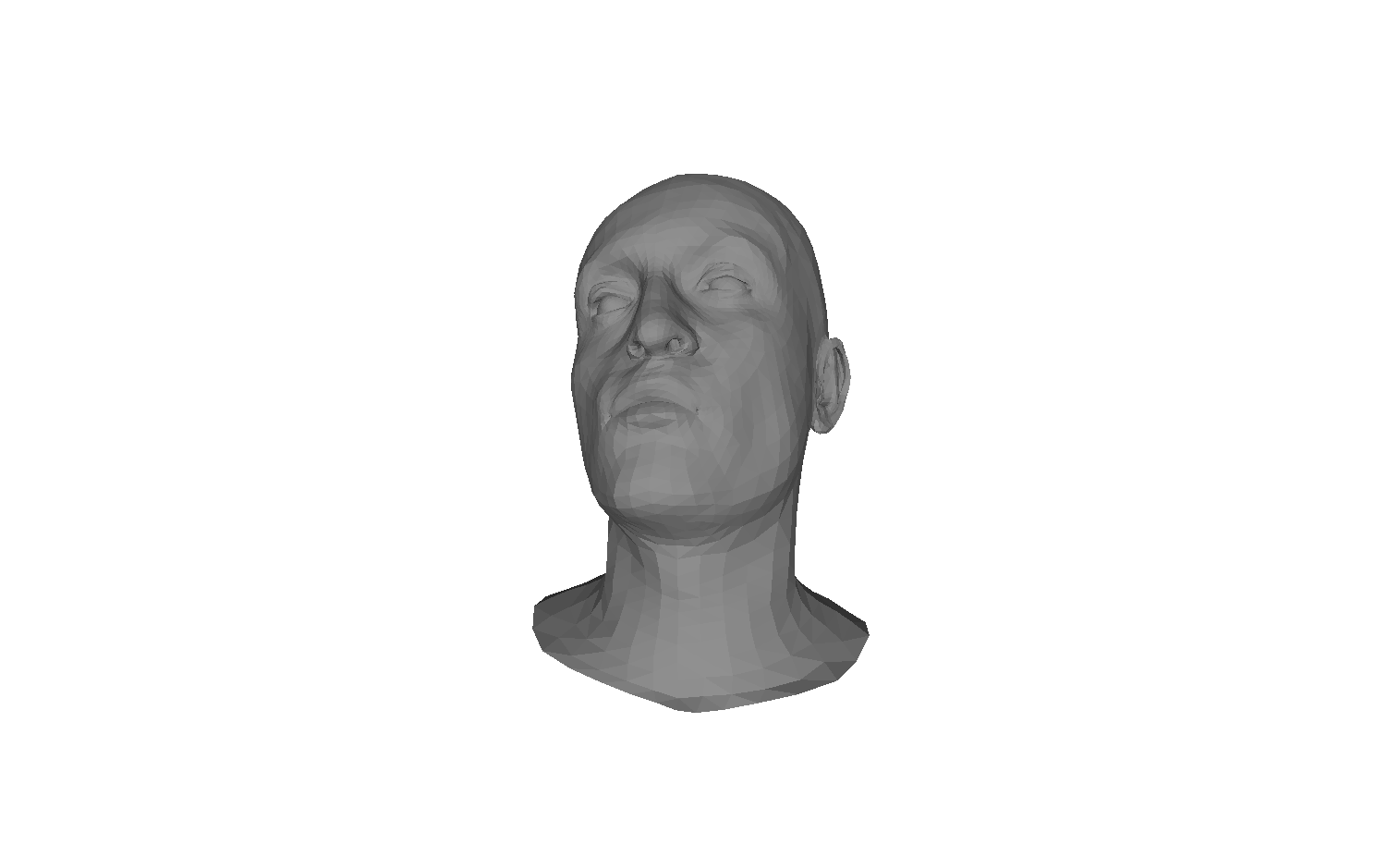}};
    \node[right of=d10, node distance=1.3cm] (d11) {\includegraphics[trim={400 80 400 100},clip,width=0.08\linewidth]{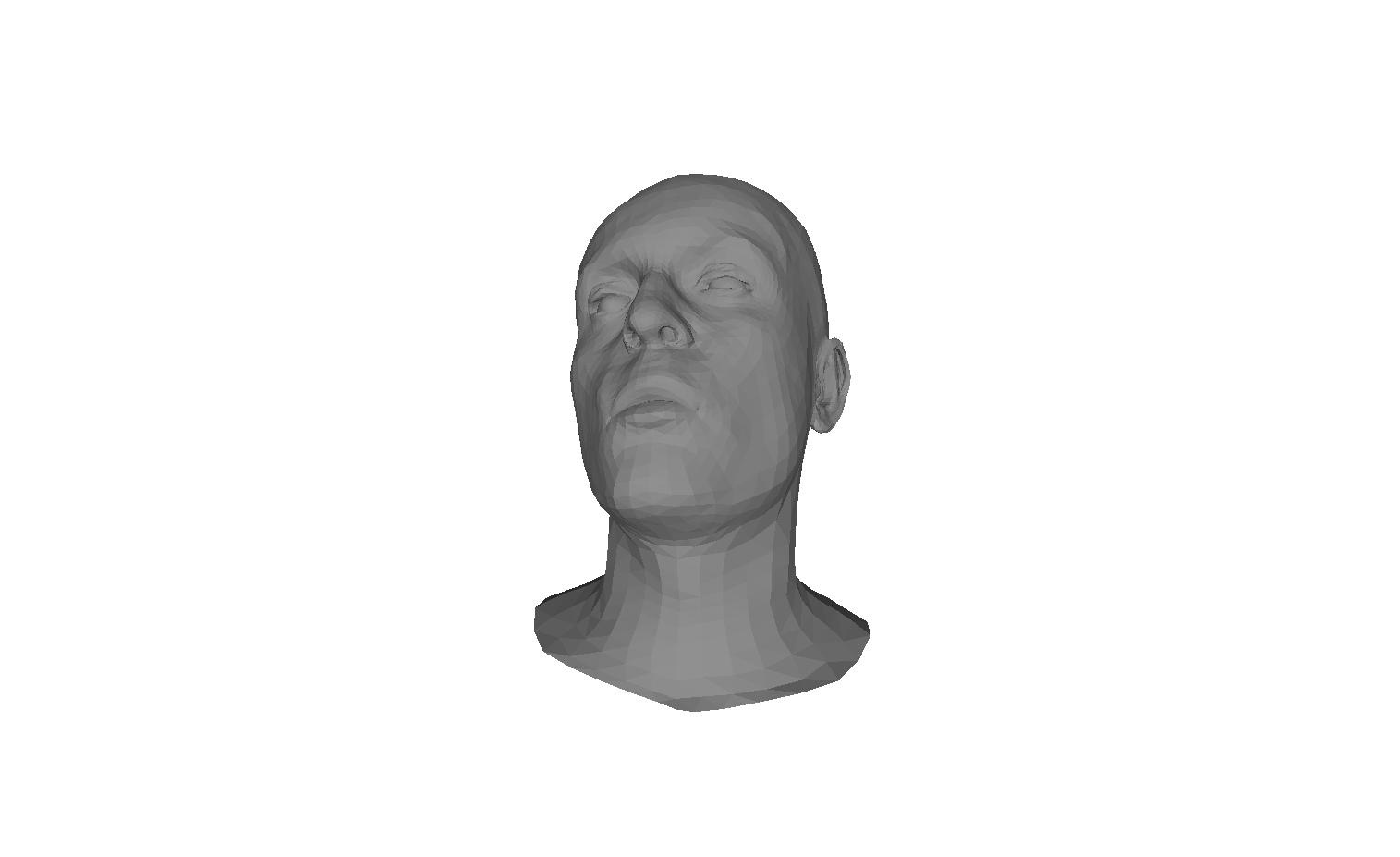}};
    
    \node[below of=d1, node distance=1.3cm] {\footnotesize Target Image};
    \node[below of=d2, node distance=1.3cm] {\footnotesize FLAME\cite{flame}};
    \node[below of=d3, node distance=1.3cm] {\footnotesize DECA\cite{deca}};
    \node[below of=d4, node distance=1.3cm, text width=1.9cm, align=center] {\footnotesize CFR-GAN\cite{occrobustwacv}};
    \node[below of=d5, node distance=1.3cm, text width=1.9cm, align=center] {\footnotesize Occ3DMM\cite{egger2018occlusion}};
    \node[below of=d6, node distance=1.3cm, text width=1.9cm, align=center] {\footnotesize Extreme3D\cite{tran2018extreme}};
    \node[below of=d9, node distance=1.3cm] {\footnotesize Reconstructions by \ourmethod{} (Ours)};
    
    \end{tikzpicture}
    \end{minipage}
    \vspace{-0.8em}
    \captionsetup{type=figure}\captionof{figure}{Diverse 3D reconstructions from a single occluded face image by \ourmethod{} \versus{} a singular solution by the baselines.\label{fig:teaser}}
\end{center}
}]
\thispagestyle{empty}

\begin{abstract}
	
Occlusions are a common occurrence in unconstrained face images. Single image 3D reconstruction from such face images often suffers from corruption due to the presence of occlusions. Furthermore, while a plurality of 3D reconstructions is plausible in the occluded regions, existing approaches are limited to generating only a single solution. To address both of these challenges, we present \ourmethod{}, which is specifically designed to simultaneously generate a diverse and realistic set of 3D reconstructions from a single occluded face image. It comprises three components; a global+local shape fitting process, a graph neural network-based mesh VAE, and a determinantal point process based diversity-promoting iterative optimization procedure. Quantitative and qualitative comparisons of 3D reconstruction on occluded faces show that \ourmethod{} can estimate 3D shapes that are consistent with the visible regions in the target image while exhibiting high, yet realistic, levels of diversity in the occluded regions. On face images occluded by masks, glasses, and other random objects, \ourmethod{} generates a distribution of 3D shapes having $\sim$50\% higher diversity on the occluded regions compared to the baselines. Moreover, our closest sample to the ground truth has $\sim$40\% lower MSE than the singular reconstructions by existing approaches. Code and data available at:  \href{https://github.com/human-analysis/diverse3dface}{https://github.com/human-analysis/diverse3dface}
\end{abstract}

\section{Introduction} \label{sec:introduction}

Single image-based 3D face reconstruction has improved significantly in recent years \cite{zollhofer2018state, egger20203d}. This includes advances in statistical models \cite{blanz1999morphable, bfm, flame, uhm} as well as neural network-based models \cite{mofa, tran2019learning, sfsnet2018sengupta, unsup3dcvpr2020, deca, ganfit, tran2019towards, wei20193d, tuan2017regressing}. However, facial occlusions remain a significant challenge to this task. In-the-wild face images often come with several forms of occlusions and unless dealt with explicitly, often lead to erroneous 3D reconstruction in terms of shape, expression, pose, \etc \cite{egger20203d, egger2018occlusion, tran2018extreme}.

3D reconstruction of partially occluded faces presents two main challenges. First, 3D reconstruction models need to selectively use features from the visible regions while ignoring those from the occluded parts. Failure to do so, either implicitly or explicitly, will lead to poor 3D reconstructions with an incorrect pose, expression, or both. Second, there could be a distribution of 3D reconstructions that are consistent with the visible parts in the image yet diverse on the occluded parts. Failure to account for all such modes limits the utility of 3D reconstruction models. Addressing these two challenges is the primary goal of this paper.

Existing 3D face reconstruction solutions, however, are ill-equipped to overcome both of these challenges simultaneously. From a \textbf{reconstruction perspective}, a majority of the approaches that reconstruct 3D faces from a single image restrict themselves to fully-visible face images. And, even those that explicitly account for facial occlusions~\cite{tran2018extreme,egger2018occlusion}, do so only in a holistic manner using a global model that implicitly uses features from the occluded regions as well. This form of global model-based fitting can introduce errors (see Fig.~\ref{fig:teaser}) in the pose and expression of the 3D reconstruction, especially when large portions of the face are occluded. From a \textbf{diversity perspective}, existing approaches are, by design, limited to only generating a single plausible 3D reconstruction. However, in many practical applications, for a single occluded face image, it is desirable to generate multiple reconstructions that are consistent on the visible parts of the face, while spanning a diverse yet realistic set of reconstructions on the occluded parts (see Fig.~\ref{fig:teaser}). While the concept of generating diverse solutions has been explored in other contexts such as image generation \cite{gdpp}, image completion \cite{picnet}, super-resolution \cite{explorablesuperres} and trajectory forecasting~\cite{yuan2019diverse}, they have not been explored for monocular 3D face reconstruction of occluded faces.

In this paper, we propose \ourmethod{} which is designed to simultaneously yield a diverse, yet plausible, set of 3D reconstructions from a single occluded face image. \ourmethod{} consists of three modules: a global + local shape fitting process, a graph neural network based variational autoencoder (Mesh-VAE), and a Determinantal Point Process (DPP)~\cite{dpp} based iterative optimization procedure. The global + local shape fitting process affords robustness against large occlusions by decoupling shape fitting on the visible regions from that of the occluded regions. The Mesh-VAE enables to learn a distribution over a compact latent space over the different factors of variation in the 3D shapes of faces. And, the DPP-based iterative optimization procedure enables us to sample from the latent space of the Mesh-VAE and optimize them to generate a diverse set of reconstructions spanning the different modes of the latent space. Our specific contributions in this paper are:

\noindent\textbf{--} We propose \ourmethod{}, a simple yet effective diversity promoting 3D face reconstruction approach that generates multiple plausible 3D reconstructions corresponding to a single occluded face image.

\noindent\textbf{--} For robustness to occlusions, we propose a global + local PCA model based shape fitting that disentangles the fitting on each facial component from the others. The models are learned from a dataset of FLAME \cite{flame} registered 3D meshes. During inference, the local perturbations on various facial components are added on top of a coarse global fit to generate the final detailed fitting.

\noindent\textbf{--} We employ a DPP~\cite{dpp} based diversity loss in the context of generating diverse 3D reconstructions of faces. We define the quality and similarity terms in the DPP kernel to maximize diversity while remaining in the space of realistic 3D head shapes.

\noindent\textbf{--} We conduct extensive qualitative and quantitative experiments to show the efficacy of the proposed approach in generating 3D reconstructions that are faithful to the visible face while simultaneously capturing multiple diverse modes on the occluded parts. The solution from \ourmethod{} that is closest to the ground truth is on average 30-50\% better than the unique solutions of the baselines \cite{deca, flame} in terms of per-vertex $\ell_2$-error.

\section{Related Work} \label{sec:related-work}

\vspace{5pt}
\noindent\textbf{Single Image 3D Face Reconstruction:} Blanz and Vetter \cite{blanz1999morphable} proposed the first 3DMM model of human faces. Since then, such models have grown to include complex pose and expression modalities in 3D faces \cite{bfm, gerig2018morphable}. Li \etal \cite{flame} proposed FLAME that models the full human head and allows non-linear control over joint poses to generate articulated expressive head instances. Many recent approaches adopted neural networks to model higher-order complexities in the shape and expression spaces \cite{mofa, sfsnet2018sengupta, neuralfaceediting, tran2019learning, tran2019towards, tuan2017regressing, ramon2021h3d, kim2018inversefacenet, ringnet, unsup3dcvpr2020, deca}. A few methods took a hybrid approach of fitting a non-linear neural network model to the target image to generate detailed 3D reconstructions \cite{ganfit, i3dmm}. More recently, advances in graph neural networks \cite{gcn, gat, chebconv, graphconv} have propagated using graph convolution operations to directly learn non-linear representation on a mesh surface while preserving the mesh topology \cite{coma, neural3dmm, meshconv}. Though these advances have significantly improved the modeling capabilities of 3D face reconstruction approaches, they are still limited when handling occlusions in face images. 

On the other hand, a few approaches are explicitly designed to handle occlusions \cite{tran2018extreme, egger2018occlusion, occrobustwacv}. Tran \etal \cite{tran2018extreme} trained a neural network to regress a robust foundation shape from a masked face image, over which a detailed bump map is added later. And, Egger \etal \cite{egger2018occlusion} simultaneously optimized an occlusion mask and the model parameters from an occluded image. However, these approaches rely on a global model to account for the entire face, including the occluded parts, which is sub-optimal as the lack of information from such parts needs to be countered using strong regularization. Moreover, they are limited to reconstructing a singular 3D solution without considering the plurality of solutions that can explain the occluded regions. In contrast, we address the dual problems of robustness and lack of uniqueness through a multistage approach that disentangles fitting on the visible regions from diversity modeling on the occluded ones.

\vspace{5pt}
\noindent\textbf{Diversity Promoting Generative Models:} Diversity promoting algorithms have been employed in several areas in computer vision where a distribution of outcomes is more desirable than a singular solution. Conditioning \cite{pix2pix, yang2019diversity} and regularization \cite{bicyclegan, madgan, suzuki2016joint, che2016mode, srivastava2017veegan} based techniques are useful to overcome mode-collapse and promote diversity in GANs \cite{gan}. As ill-posed problems, diversity promoting algorithms are also particularly useful for image completion and image super-resolution. Zheng \etal \cite{picnet} proposed a dual-pipeline C-VAE \cite{cvae} that maintains ground-truth fidelity in one path while allowing diversity on the other. Bahat \etal \cite{explorablesuperres} generated diverse super-resolution explanations by only enforcing consistency in the low-resolution space. Compared to image-based approaches that focus on diversity in the texture, 3D reconstruction requires modeling geometric diversity. As one of the most seminal works in this field, Kulesza and Taskar \cite{dpp} introduced the framework of Determinantal Point Processes (DPPs) to model diversity in machine learning tasks such as inference, sampling, marginalization, \etc. Yuan \etal \cite{yuan2019diverse, dlow} adopted DPP to sample multi-modal latent vectors for diverse human trajectory forecasting. Elfeki \etal \cite{gdpp} devised a DPP-based objective to train GANs and VAEs to emulate the diversity in real data. In this work, we adopt the idea of DPPs to generate diverse 3D reconstructions for an occluded face by discovering latent space representations that maximize plausible diversity on the occluded regions while remaining faithful to the visible parts.

\section{Background} \label{sec:background}

\noindent\textbf{Statistical Models of 3D Face Reconstruction}: Statistical 3D models such as BFM \cite{blanz1999morphable, bfm} and FLAME \cite{flame} allow for generating new face instances. These models often consist of a \textit{shape model} that explain geometric variations across identities, an \textit{expression model} that accounts for variations due to different facial expressions, and additionally a \textit{pose model} and an \textit{appearance model} to account for variations in pose and appearance, respectively. Specifically, FLAME \cite{flame} defines a 3D shape as:
\begin{align}
    S(\beta, \theta, \psi)=W(T(\beta, \theta, \psi), \mathbf{J}(\beta), \theta, \mathbf{\mathcal{W}}),
    \label{eqn:flame}
\end{align}
where the parameters $\beta, \theta, \psi$ represent the shape, pose and expression parameters, respectively; $\mathbf{J\in \mathbb{R}^{3K}}$ represents the locations of $K$ face joints around which $T(\beta, \theta, \psi)$ is rotated, and finally smoothed by the blend weights $\mathbf{\mathcal{W}}$. The un-aligned shape $T(\beta, \theta, \psi)$ is obtained by adding up the contributions of shape, expression and pose variations on top of a template shape $\mathbf{\bar{T}}$:
\begin{small}
\begin{align}
    \label{eqn:flame_shape}
    T(\beta, \theta, \psi)=\mathbf{\bar{T}} + B_S(\beta; \mathcal{S}) + B_P(\theta; \mathcal{P}) + B_E(\psi; \mathcal{E})
\end{align}
\end{small}

The shape and expression variations are modeled by linear blendshapes $B_S(\beta; \mathcal{S})=\mathcal{S}\beta$ and $B_E(\psi; \mathcal{E})=\mathcal{E}\psi$, where $\mathcal{S}\in \mathbb{R}^{3N\times |\beta|}$ and $\mathcal{E} \in \mathbb{R}^{3N\times |\psi|}$ are orthonormal shape and expression bases, respectively, learned using PCA and $N$ is the number of vertices. The pose blendshape function is defined as $B_P(\theta; \mathcal{P})= \left( R(\theta) - R(\theta*) \right) \mathcal{P}$, where $R(\theta)$ comprises of rotation matrices around the $K$ joints and $\mathcal{P} \in \mathbb{R}^{3N \times 9K}$ are the pose blendshapes describing the vertex offsets from the rest pose activated by $R$.

\vspace{3pt}
\noindent\textbf{Determinantal Point Processes:} Determinantal Point Processes (DPPs) originated in quantum physics to model the negative correlations between the quantum states of fermions \cite{macchi1975coincidence}. DPPs were first introduced in machine learning by Kulesza and Taskar \cite{dpp} as a probabilistic model of repulsion between points. A point process over a ground set $\mathcal{Y}$ describes the probability of all its $2^{\mathcal{Y}}$ subsets. A point process is determinantal when the probability of choosing a random subset $Y \subseteq \mathcal{Y}$ is given by the determinant of the sub-kernel matrix $\mathbf{L}_Y$ indexed by the elements of Y, \ie, $P(Y \subseteq \mathcal{Y}) = det(\mathbf{L}_Y)$. Given a data matrix $B\in \mathbb{R}^{D\times N}$, we can compute the kernel as the Gram matrix $\mathbf{L}=B^TB$. In this case, the determinant of the sub-kernel matrix $det(\mathbf{L}_Y)$ is related to the volume spanned by the elements of $B$. Thus, conceptually, DPP assigns a higher probability to a subset whose elements tend to be orthogonal (diverse) to each other, thus spanning a larger volume.

\section{Approach} \label{sec:approach}

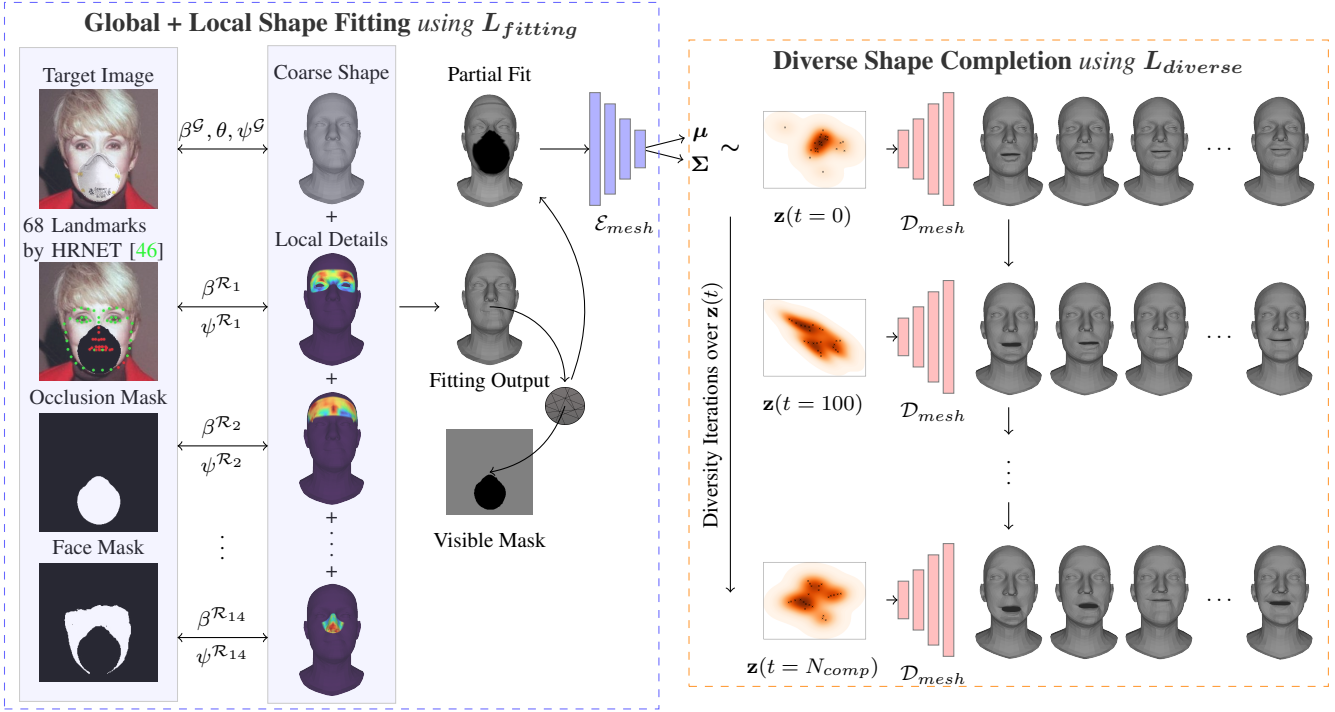
\begin{figure*}
    \centering
    \tikzstyle{enc_layer} = [block, opacity=.5, rotate=90, minimum height=4pt, inner sep=0, fill=blue!60]
    \tikzstyle{dec_layer} = [block, opacity=.5, rotate=90, minimum height=4pt, inner sep=0, fill=red!50]
    \footnotesize
    \begin{tikzpicture}
    \node (input) {\includegraphics[width=.09\linewidth]{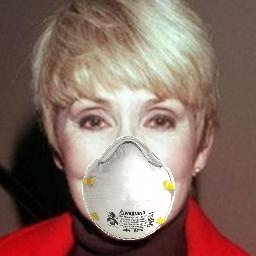}};
    \node [above of=input, node distance=0.95cm] {Target Image};
    \node [below of=input, node distance=2.3cm] (lmk_image) {\includegraphics[width=.09\linewidth]{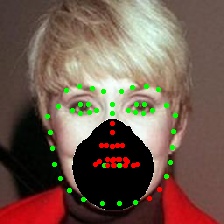}};
    \node [above of=lmk_image, node distance=1.1cm, text width=2cm] {68 Landmarks by HRNET \cite{hrnet}};
    \node [below of=lmk_image, node distance=2.0cm] (occ_mask) {\includegraphics[width=.09\linewidth]{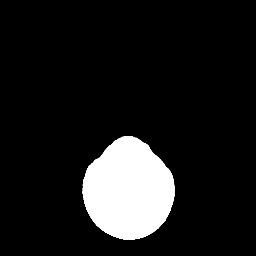}};
    \node [above of=occ_mask, node distance=1.0cm] {Occlusion Mask};
    \node [below of=occ_mask, node distance=2.0cm] (img_mask) {\includegraphics[width=.09\linewidth]{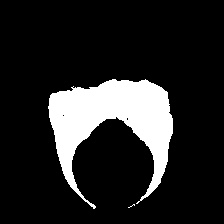}};
    \node [above of=img_mask, node distance=1.0cm] {Face Mask};
    \node [block,minimum width=2.1cm,minimum height=8.7cm,below of=input,node distance=3cm,opacity=0.2] {}; 
    
    \node [right of=input, node distance=3.1cm] (coarse_shape) {\includegraphics[trim={400 80 400 100},clip,width=.065\linewidth]{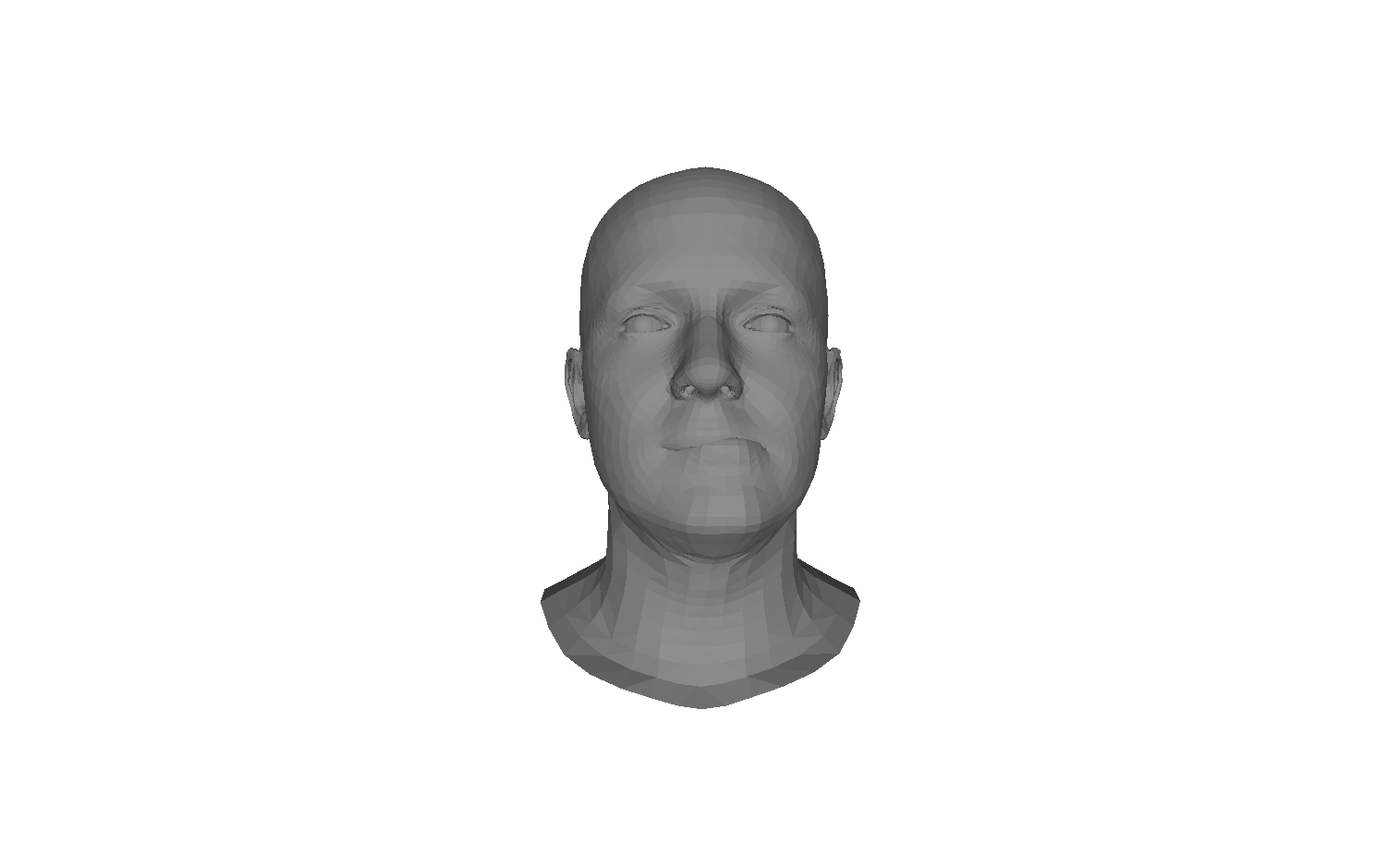}};
    \node [above of=coarse_shape, node distance=1.0cm] {Coarse Shape};
    \node [below of=coarse_shape, node distance=0.9cm] {+};

    \node [below of=coarse_shape, node distance=2.1cm] (fine_eye_region) {\includegraphics[trim={400 100 400 100},clip,width=0.065\linewidth]{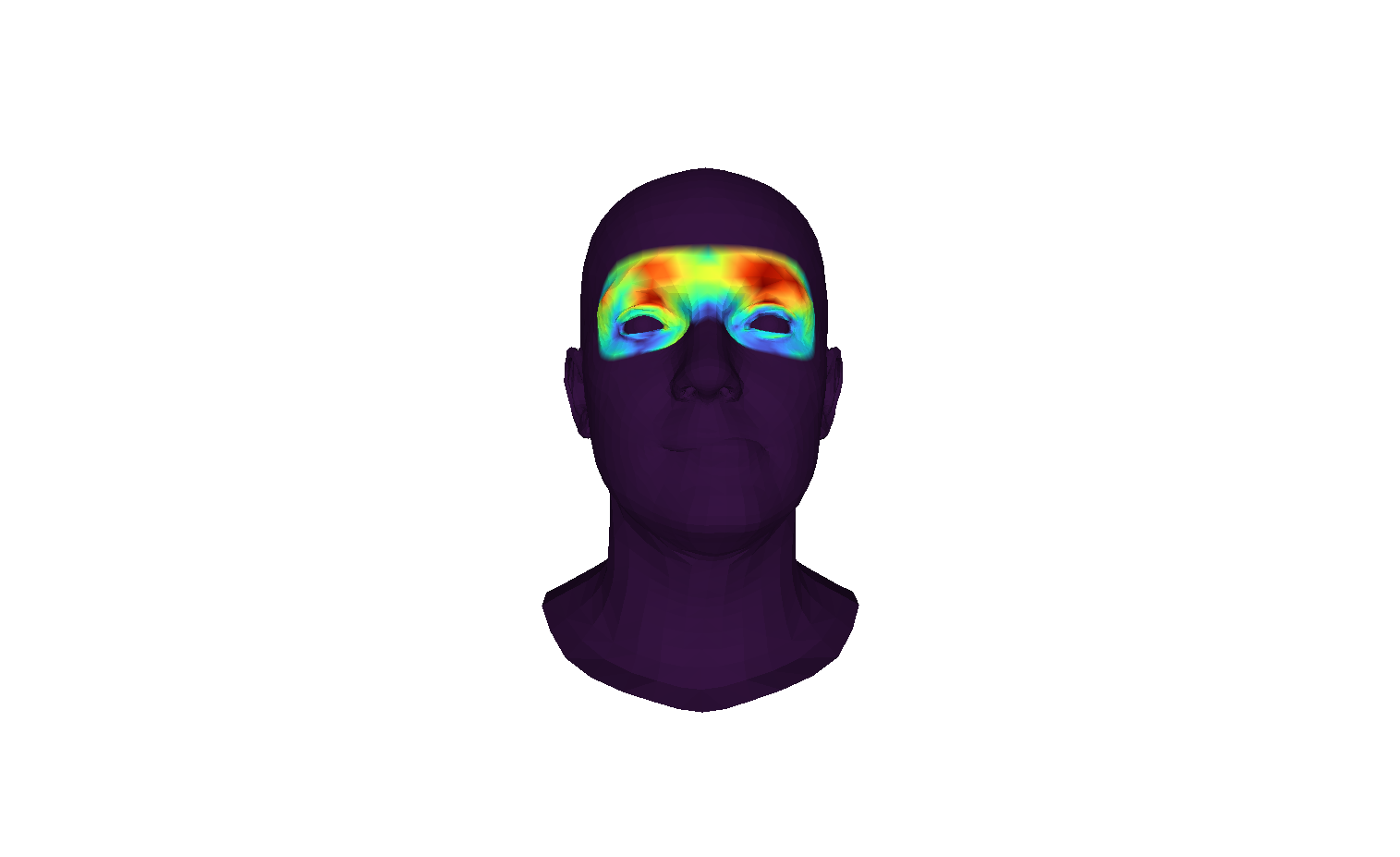}};
    \node [below of=fine_eye_region, node distance=0.95cm] {+};
    \node [below of=fine_eye_region, node distance=1.85cm] (fine_forehead) {\includegraphics[trim={400 100 400 100},clip,width=0.065\linewidth]{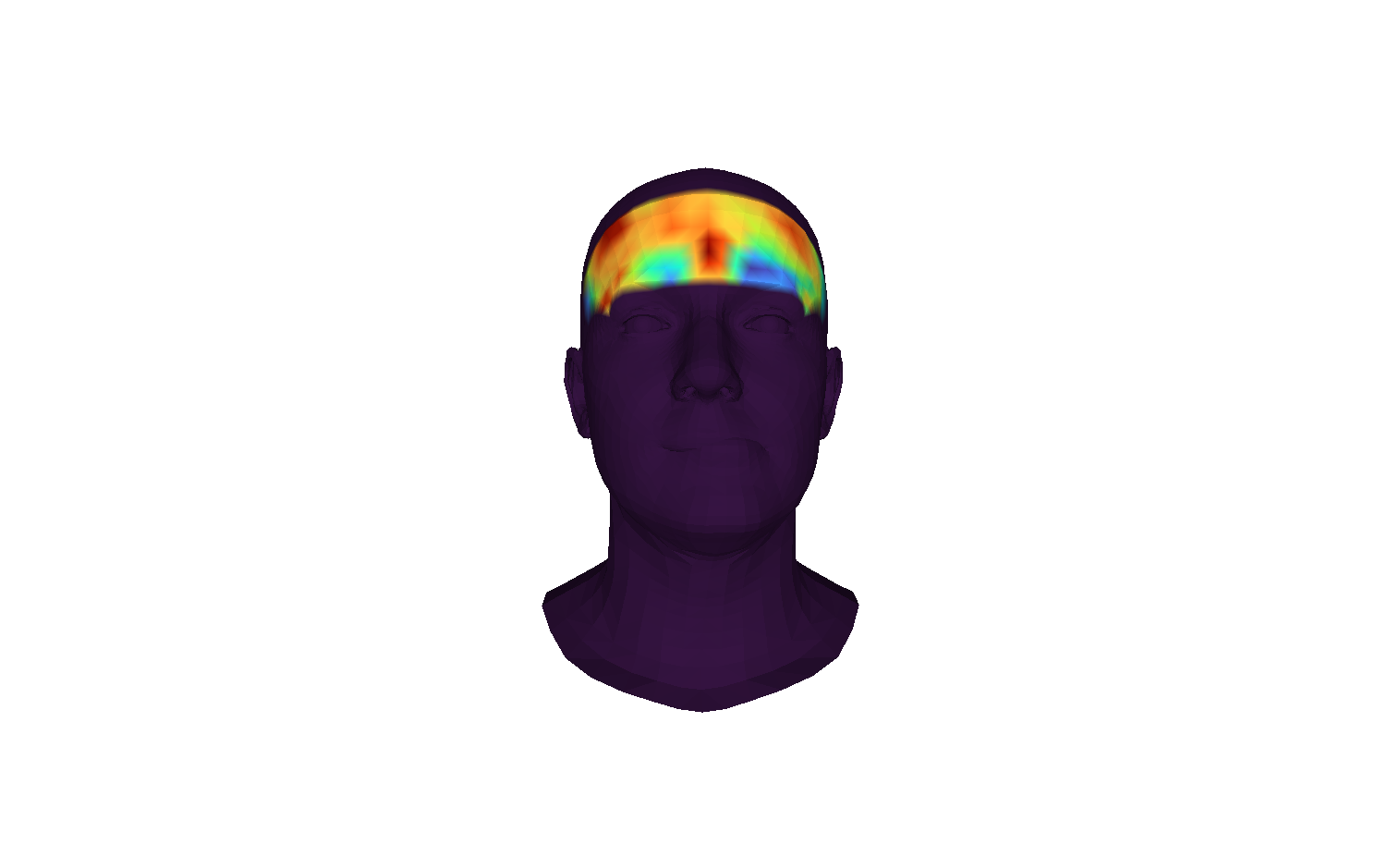}};
    \node [below of=fine_forehead, node distance=0.95cm] {+};
    \node [below of=fine_forehead, node distance=1.2cm] (fine_dots) {$\vdots$};
    \node [below of=fine_forehead, node distance=1.67cm] {+};
    \node [below of=fine_forehead, node distance=2.55cm] (fine_scalp) {\includegraphics[trim={400 100 400 100},clip,width=0.065\linewidth]{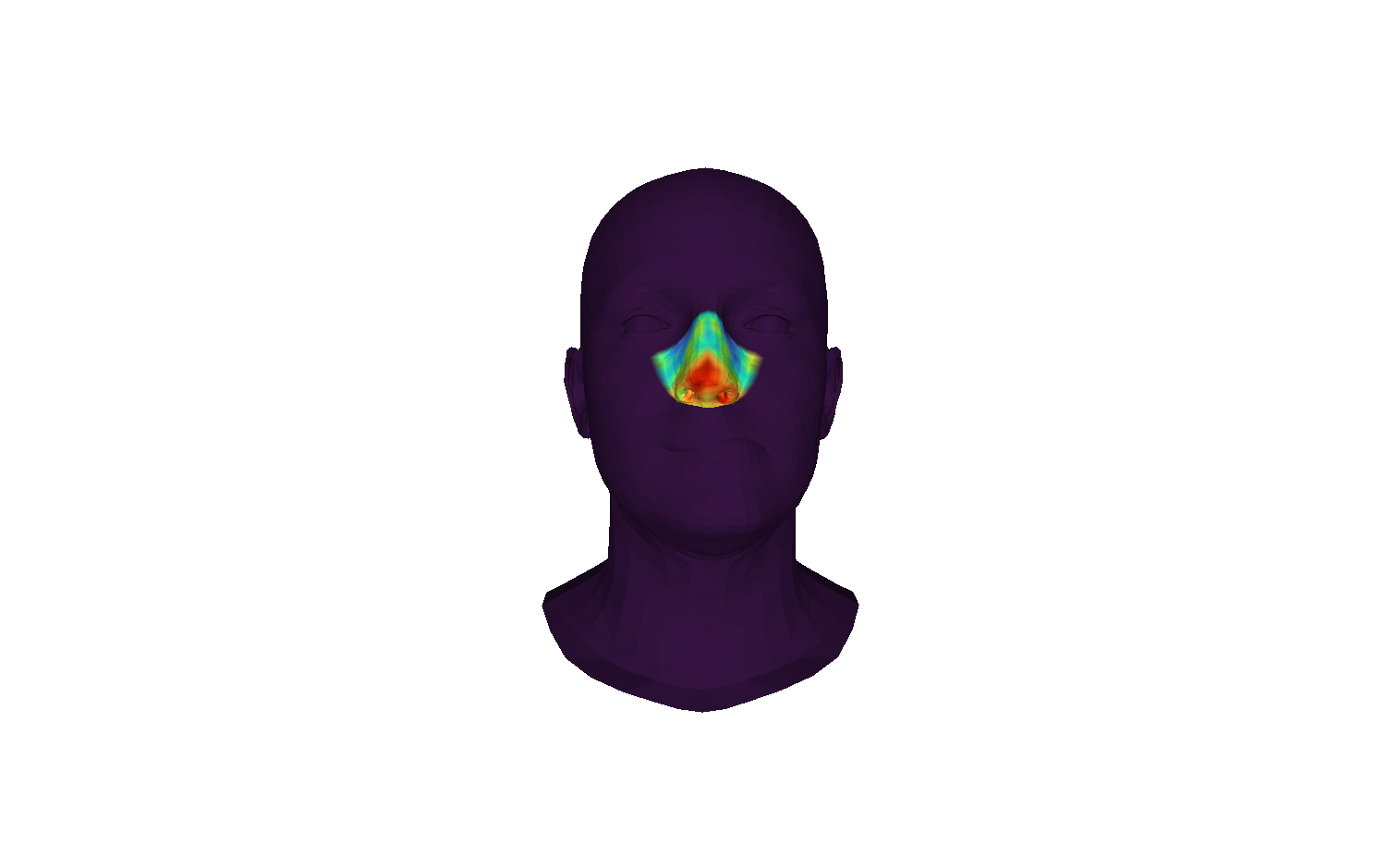}};
    \node [above of=fine_eye_region, node distance=0.9cm] {Local Details};
    \node [block,minimum width=1.7cm,minimum height=8.75cm,below of=fine_eye_region,node distance=0.9cm,opacity=0.2] {}; 
    \coordinate [left of=fine_eye_region, node distance=0.85cm] (fine_eye_region_left);
    \draw[<->] (fine_eye_region_left) -- node [midway,above] {$\beta^{\mathcal{R}_1}$} node [midway,below] {$\psi^{\mathcal{R}_1}$} ++ (-1.2,0);
    \coordinate [left of=fine_forehead, node distance=0.85cm] (fine_forehead_left);
    \draw[<->] (fine_forehead_left) -- node [midway,above] (local_r2) {$\beta^{\mathcal{R}_2}$} node [midway,below] {$\psi^{\mathcal{R}_2}$} ++ (-1.2,0);
    \coordinate [left of=fine_scalp, node distance=0.85cm] (fine_scalp_left);
    \draw[<->] (fine_scalp_left) -- node [midway,above] {$\beta^{\mathcal{R}_{14}}$} node [midway,below] {$\psi^{\mathcal{R}_{14}}$} ++ (-1.2,0);
    \node[below of=local_r2, node distance=1.5cm] {$\vdots$};
    \coordinate [left of=coarse_shape, node distance=0.85cm] (coarse_left);
    \draw[<->] (coarse_left) -- node [midway,above] {$\beta^{\mathcal{G}}, \theta, \psi^{\mathcal{G}}$} ++ (-1.2,0);

    \node [right of=fine_eye_region, node distance=2.1cm] (fine_shape) {\includegraphics[trim={400 100 400 100},clip,width=0.065\linewidth]{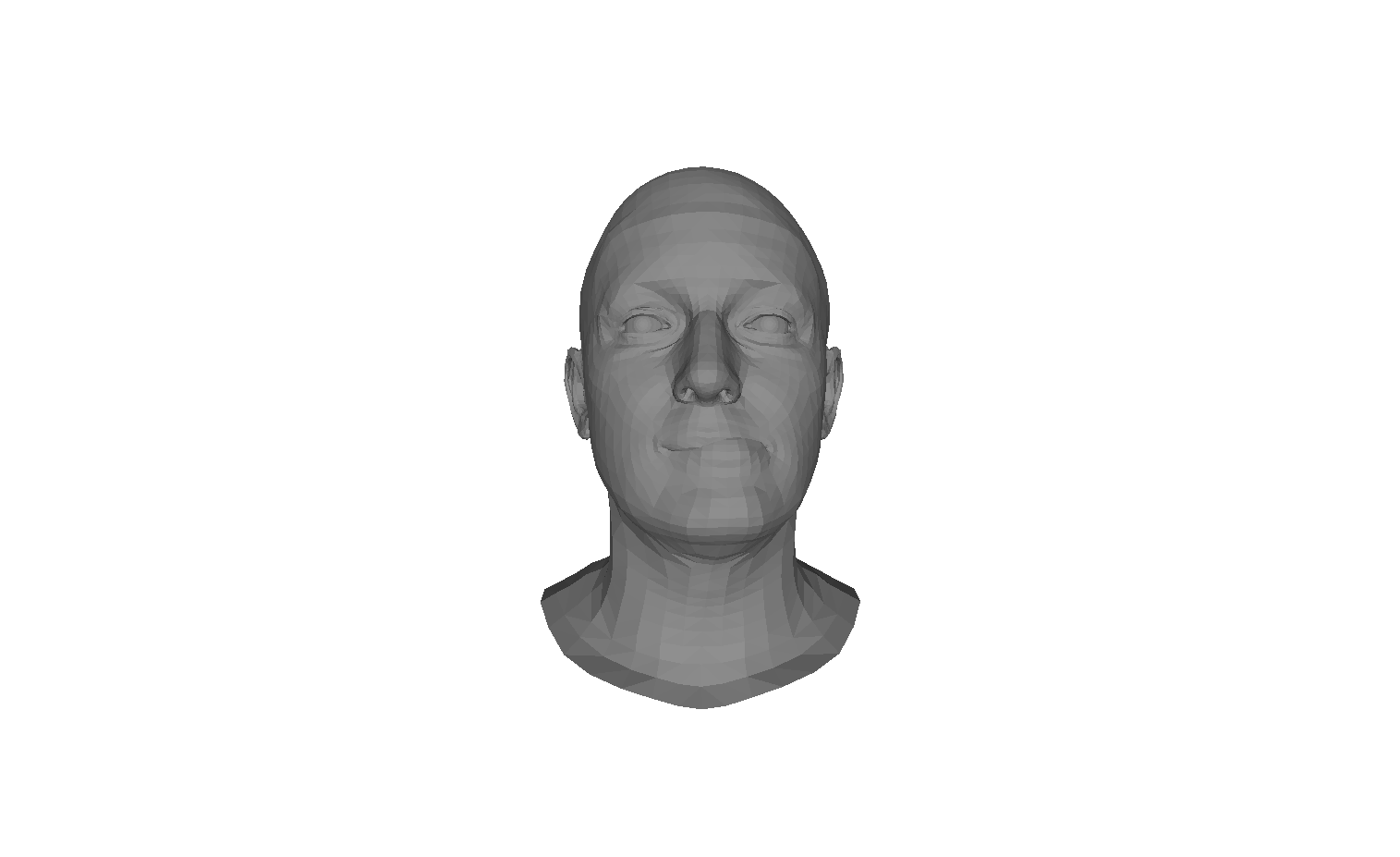}};
    \draw[<-] (fine_shape) -- ++(-1.2,0);
    \node [below of=fine_shape, node distance=2.2cm] (vis_mask) {\includegraphics[width=0.065\linewidth]{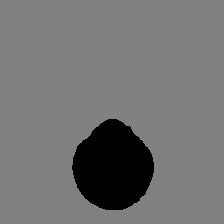}};
    \node [above of=fine_shape, node distance=2.1cm] (partial_fit) {\includegraphics[trim={400 100 400 100},clip,width=0.065\linewidth]{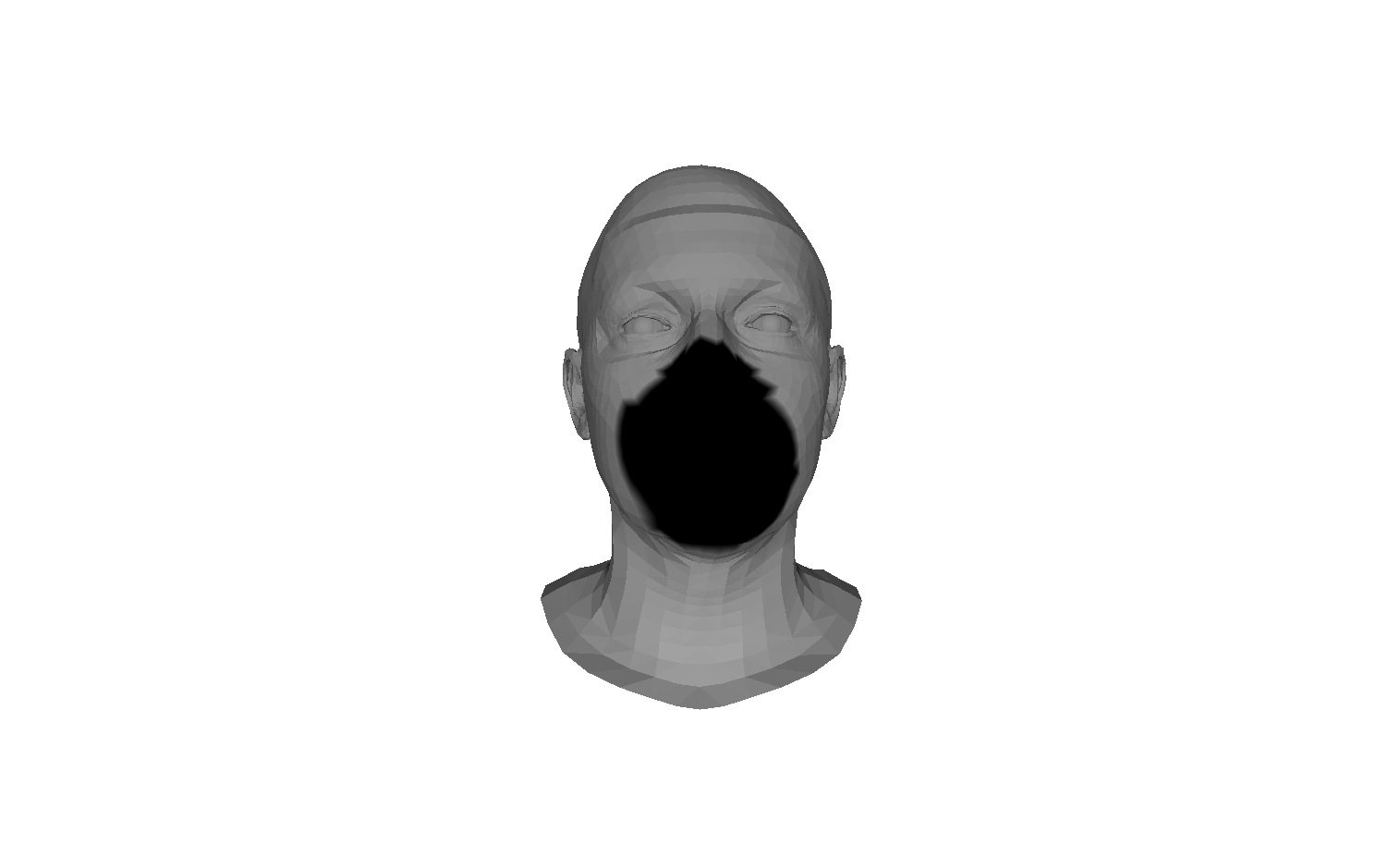}};
    \coordinate (mid_fine_occmask) at ($(fine_shape)!0.6!(vis_mask)$);
    \node [right of=mid_fine_occmask, node distance=1cm] (grid) {\includegraphics[width=.03\linewidth]{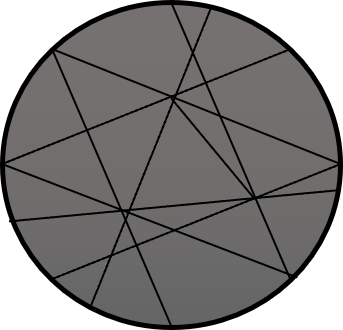}};
    \draw[->] (fine_shape) arc (90:0:1);
    \draw[bend left, ->] (grid.center) to (vis_mask.center);
    \draw[bend right, ->] (grid) to (partial_fit);
    \node [below of=vis_mask, node distance=0.9cm] {Visible Mask};
    \node [below of=fine_shape, node distance=1cm] {Fitting Output};
    \node [above of=partial_fit, node distance=1cm] {Partial Fit};
    
    \node[right of=partial_fit, node distance=1.4cm, enc_layer, minimum width=1.5cm] (enc1) {};
    \node[right of=enc1, node distance=0.2cm, enc_layer, minimum width=1.2cm] (enc2) {};
    \node[right of=enc2, node distance=0.2cm, enc_layer, minimum width=0.8cm] (enc3) {};
    \node[right of=enc3, node distance=0.2cm, enc_layer, minimum width=0.5cm] (enc4) {};
    \coordinate[right of=enc4, node distance=0.8cm] (enc_output_center);
    \node [above of=enc_output_center, node distance=0.2cm] (mu) {$\boldsymbol{\mu}$};
    \node [below of=enc_output_center, node distance=0.2cm] (sigma) {$\boldsymbol{\Sigma}$};
    \draw[->] (partial_fit) -- (enc1);
    \draw[->] (enc4) -- (mu);
    \draw[->] (enc4) -- (sigma);
    \node[below of=enc3, node distance=1cm] {$\mathcal{E}_{mesh}$};
    
    \node [block,dashed,minimum width=8.7cm,minimum height=9.4cm,above of=fine_forehead,node distance=1.2cm,fill opacity=0,draw=blue!60,label={[yshift=-0.05cm,anchor=north,text opacity=0.8]north:\normalsize{\textbf{Global + Local Shape Fitting} \textit{using} $\boldsymbol{L_{fitting}}$}}] {}; 
    
    \node [right of=enc_output_center, node distance=1.5cm] (tsne_z1) {\includegraphics[width=.1\linewidth]{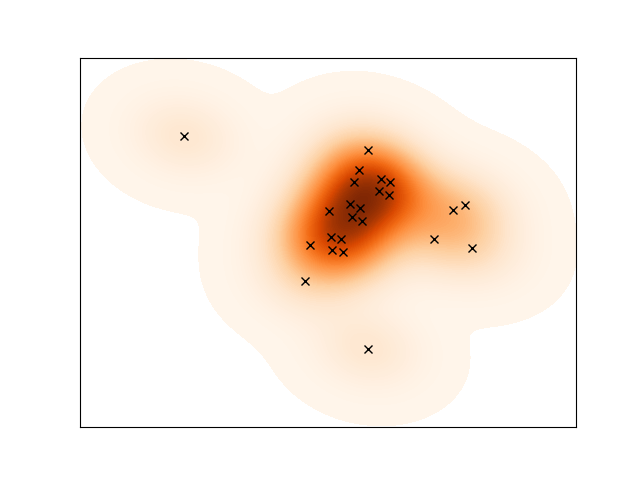}};
    \node [below of=tsne_z1, node distance=0.9cm] (z0_label) {$\mathbf{z}(t=0)$};
    \node [right of=enc_output_center, node distance=0.4cm] {$\boldsymbol{\sim}$};
    \node[right of=tsne_z1, node distance=1.2cm, dec_layer, minimum width=0.5cm] (dec1) {};
    \node[right of=dec1, node distance=0.2cm, dec_layer, minimum width=0.8cm] (dec2) {};
    \node[right of=dec2, node distance=0.2cm, dec_layer, minimum width=1.2cm] (dec3) {};
    \node[right of=dec3, node distance=0.2cm, dec_layer, minimum width=1.5cm] (dec4) {};
    \node[right of=dec4, node distance=0.8cm] (iter0_1) {\includegraphics[trim={400 100 400 100},clip,width=0.065\linewidth]{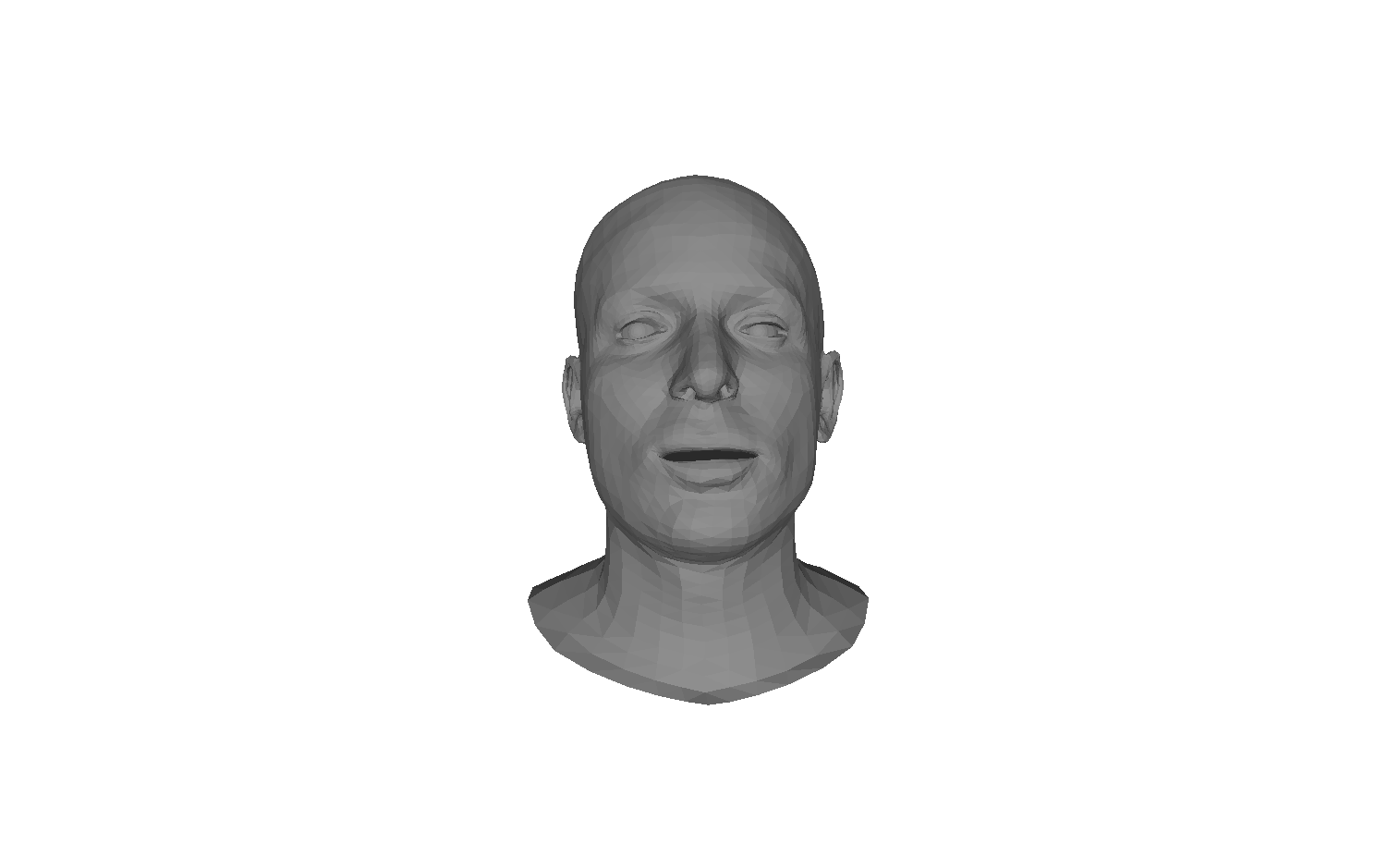}};
    \node[right of=iter0_1, node distance=1cm] (iter0_2) {\includegraphics[trim={400 100 400 100},clip,width=0.065\linewidth]{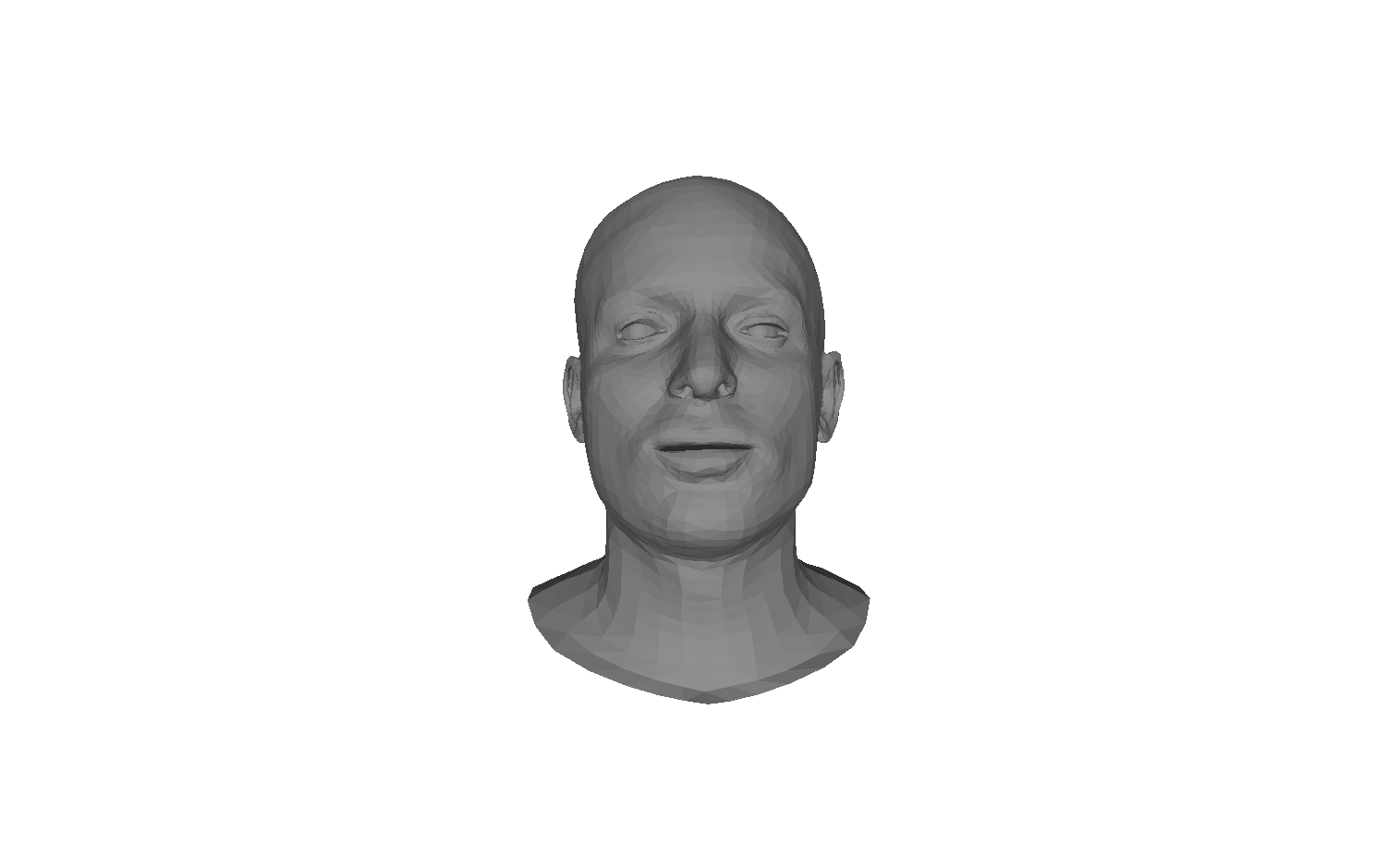}};
    \node[right of=iter0_2, node distance=1cm] (iter0_3) {\includegraphics[trim={400 100 400 100},clip,width=0.065\linewidth]{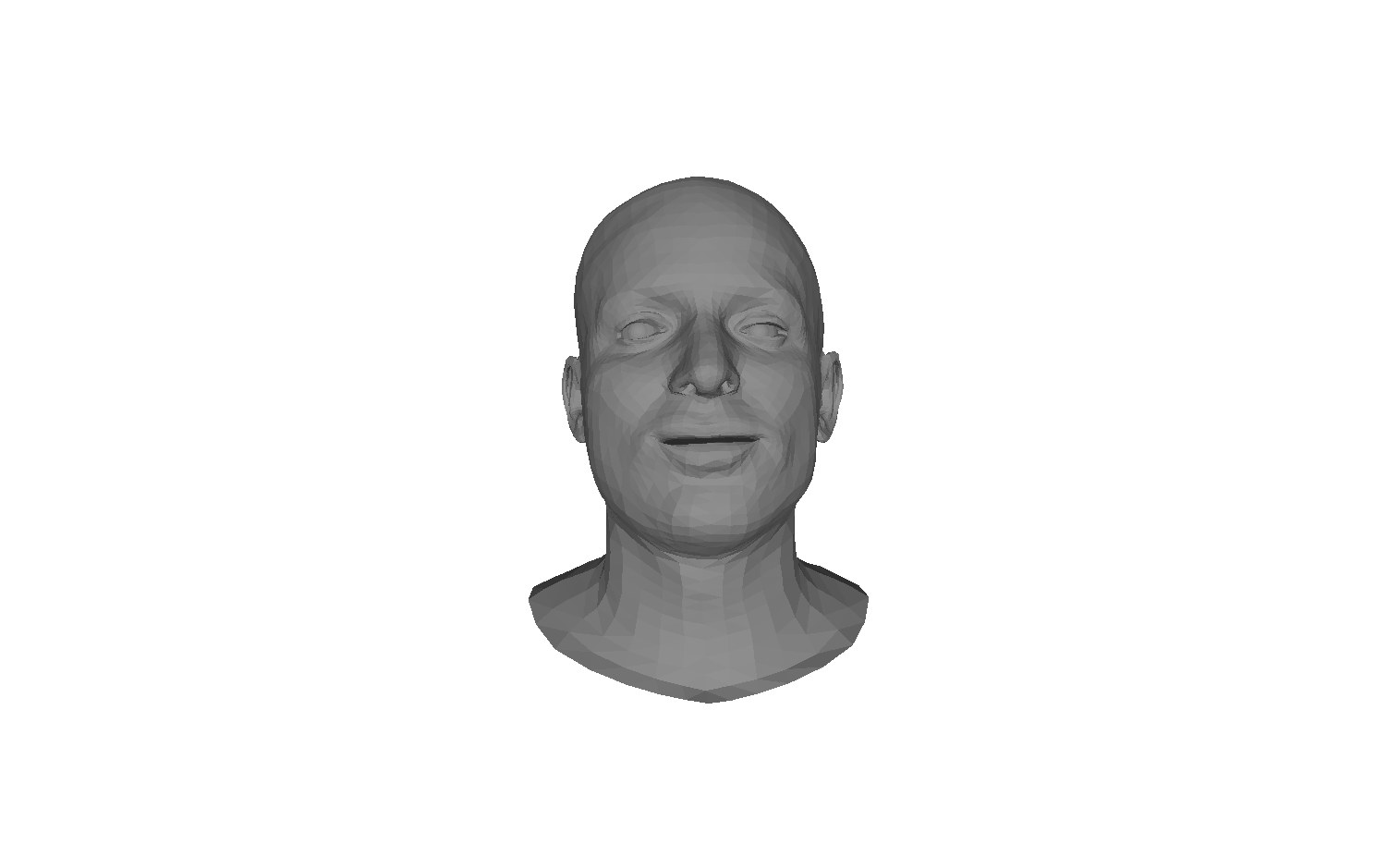}};
    \node[right of=iter0_3, node distance=0.8cm] (iter0_dots) {$\hdots$};
    \node[right of=iter0_dots, node distance=0.8cm] (iter0_4) {\includegraphics[trim={400 100 400 100},clip,width=0.065\linewidth]{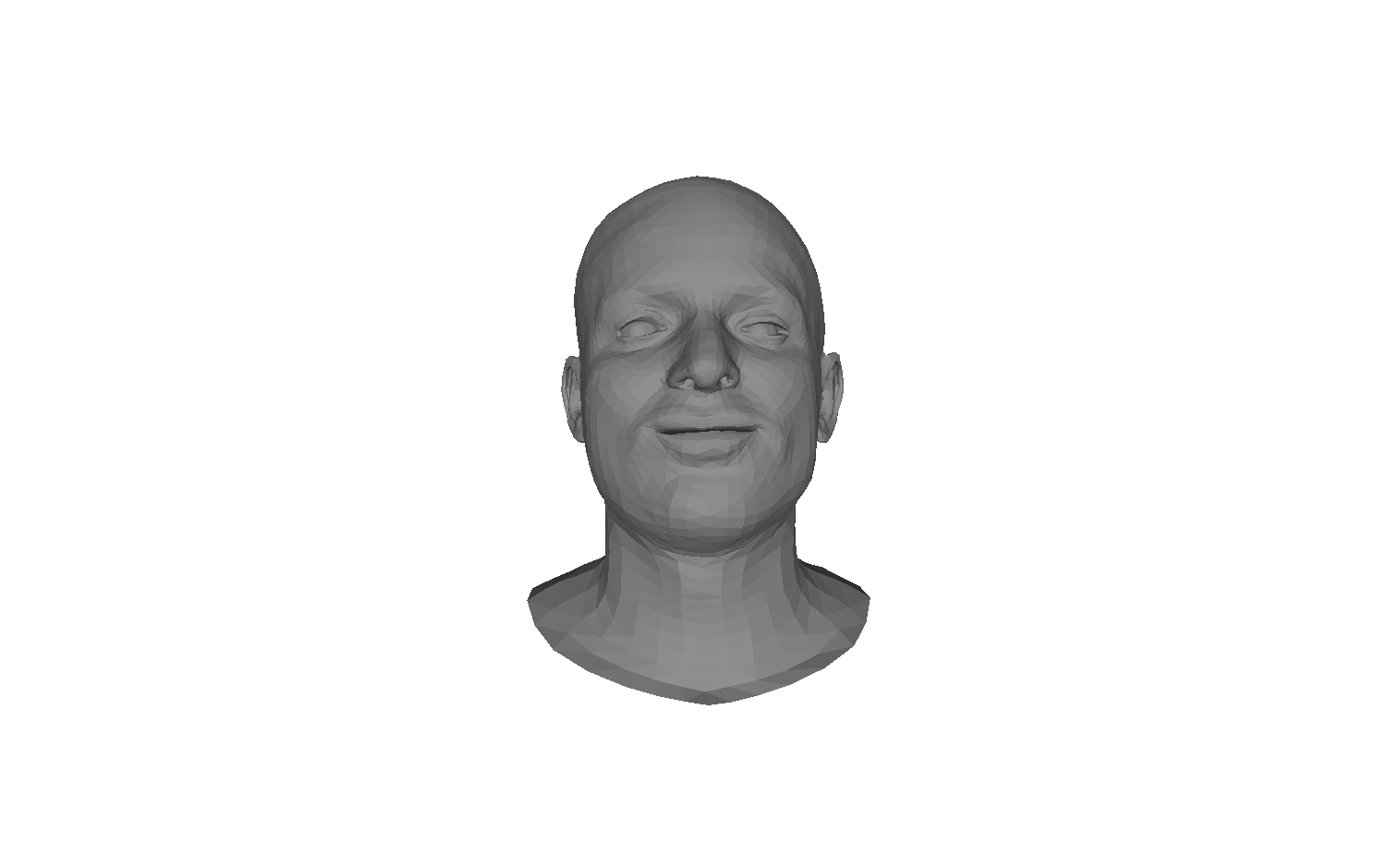}};
    \node[below of=dec3, node distance=1cm] {$\mathcal{D}_{mesh}$};
    \draw[->] (tsne_z1) -- (dec1);
    
    \node [below of=tsne_z1, node distance=2.5cm] (tsne_z2) {\includegraphics[width=.1\linewidth]{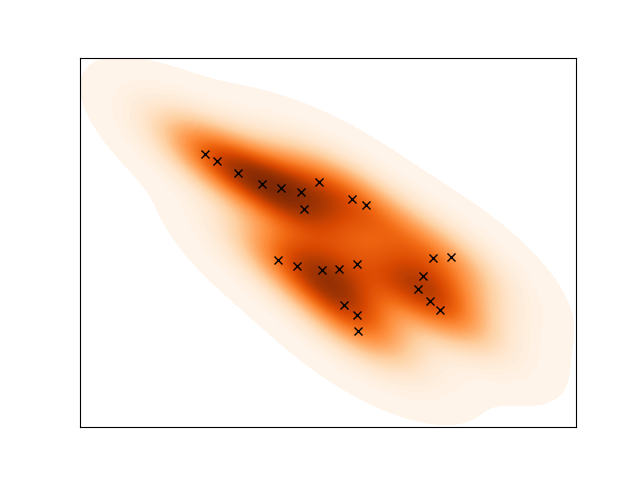}};
    \node [below of=tsne_z2, node distance=0.9cm] {$\mathbf{z}(t=100)$};
    \node[right of=tsne_z2, node distance=1.2cm, dec_layer, minimum width=0.5cm] (dec1) {};
    \node[right of=dec1, node distance=0.2cm, dec_layer, minimum width=0.8cm] (dec2) {};
    \node[right of=dec2, node distance=0.2cm, dec_layer, minimum width=1.2cm] (dec3) {};
    \node[right of=dec3, node distance=0.2cm, dec_layer, minimum width=1.5cm] (dec4) {};
    \node[right of=dec4, node distance=0.8cm] (iter1_1) {\includegraphics[trim={400 100 400 100},clip,width=0.065\linewidth]{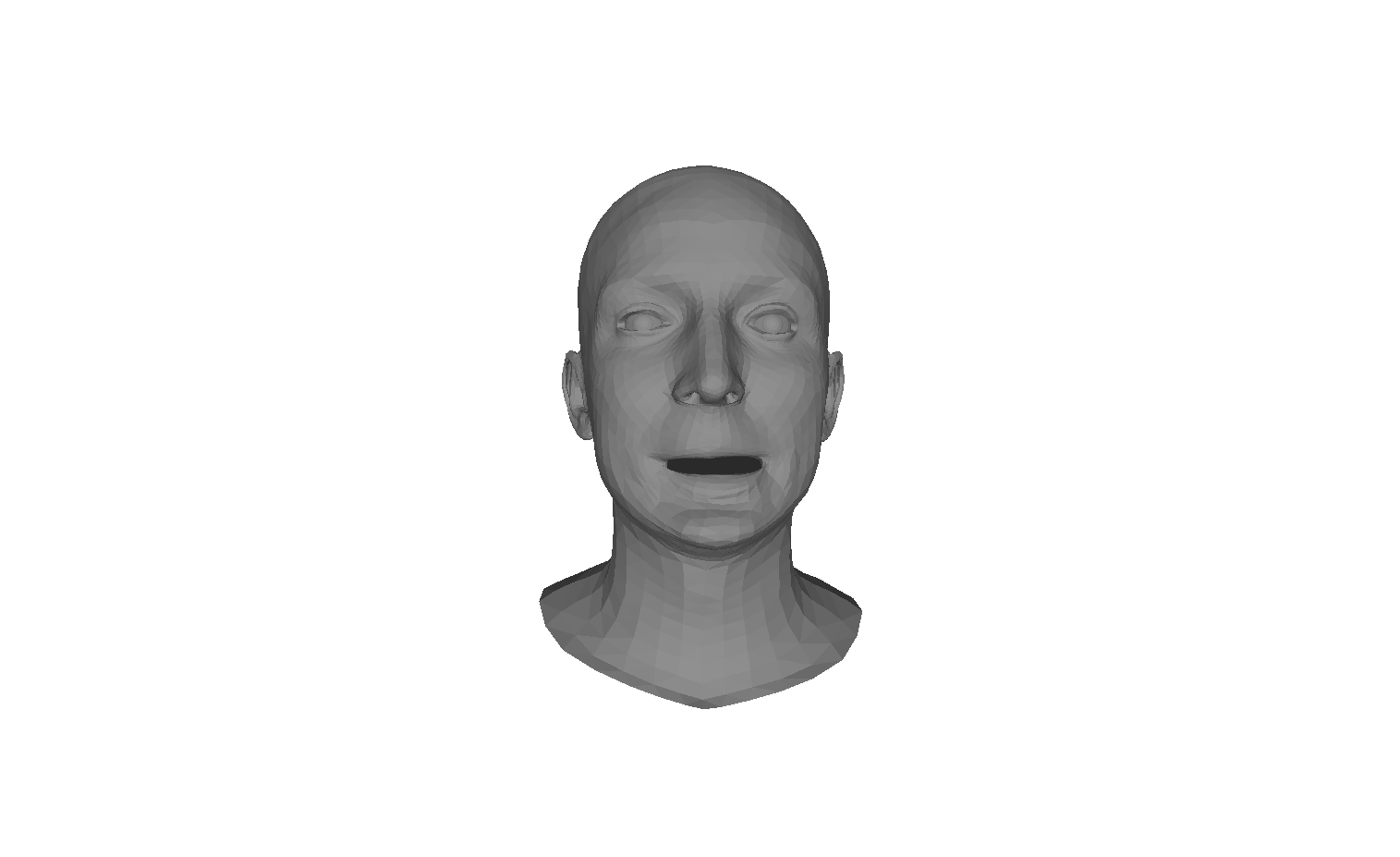}};
    \node[right of=iter1_1, node distance=1cm] (iter1_2) {\includegraphics[trim={400 100 400 100},clip,width=0.065\linewidth]{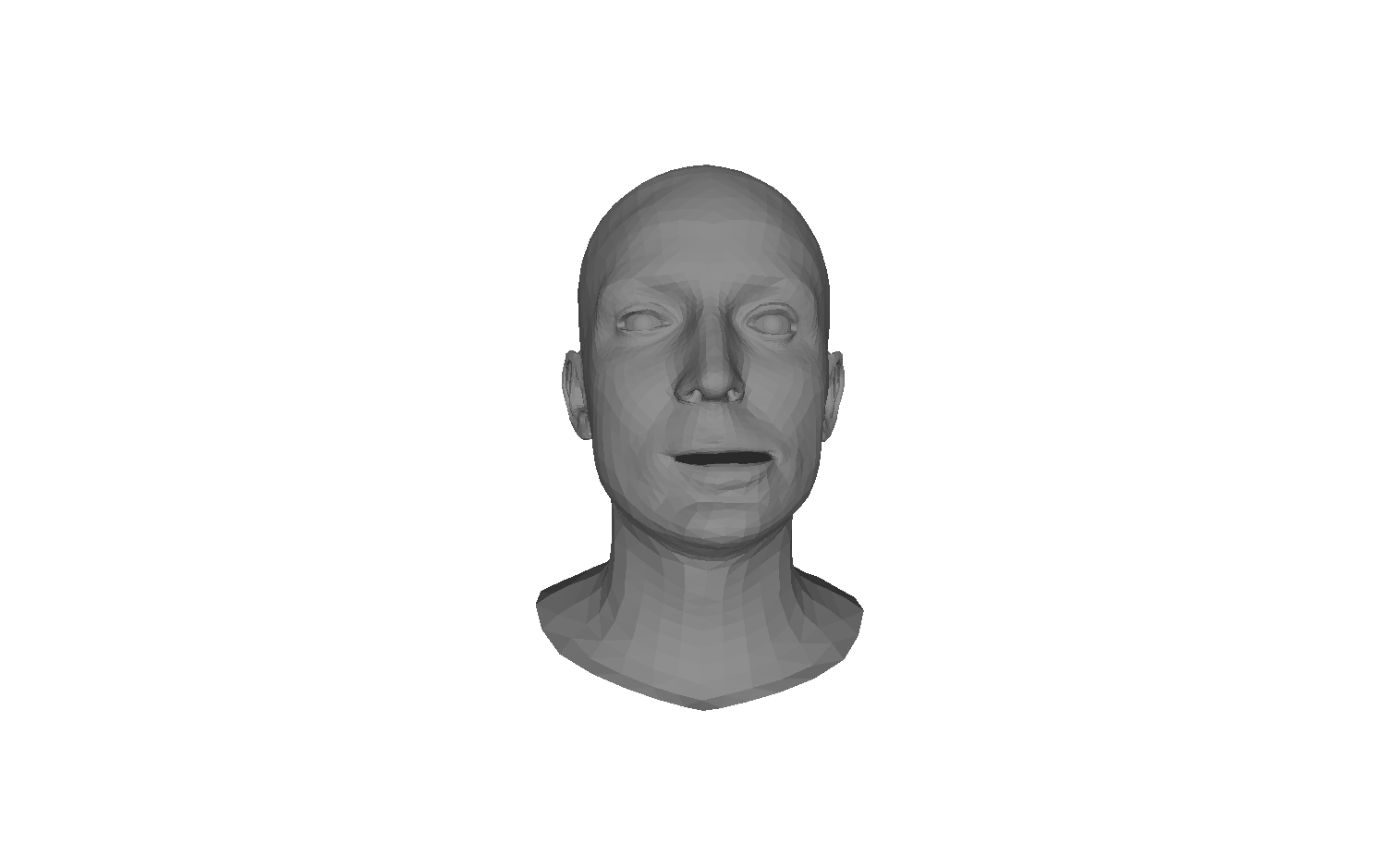}};
    \node[right of=iter1_2, node distance=1cm] (iter1_3) {\includegraphics[trim={400 100 400 100},clip,width=0.065\linewidth]{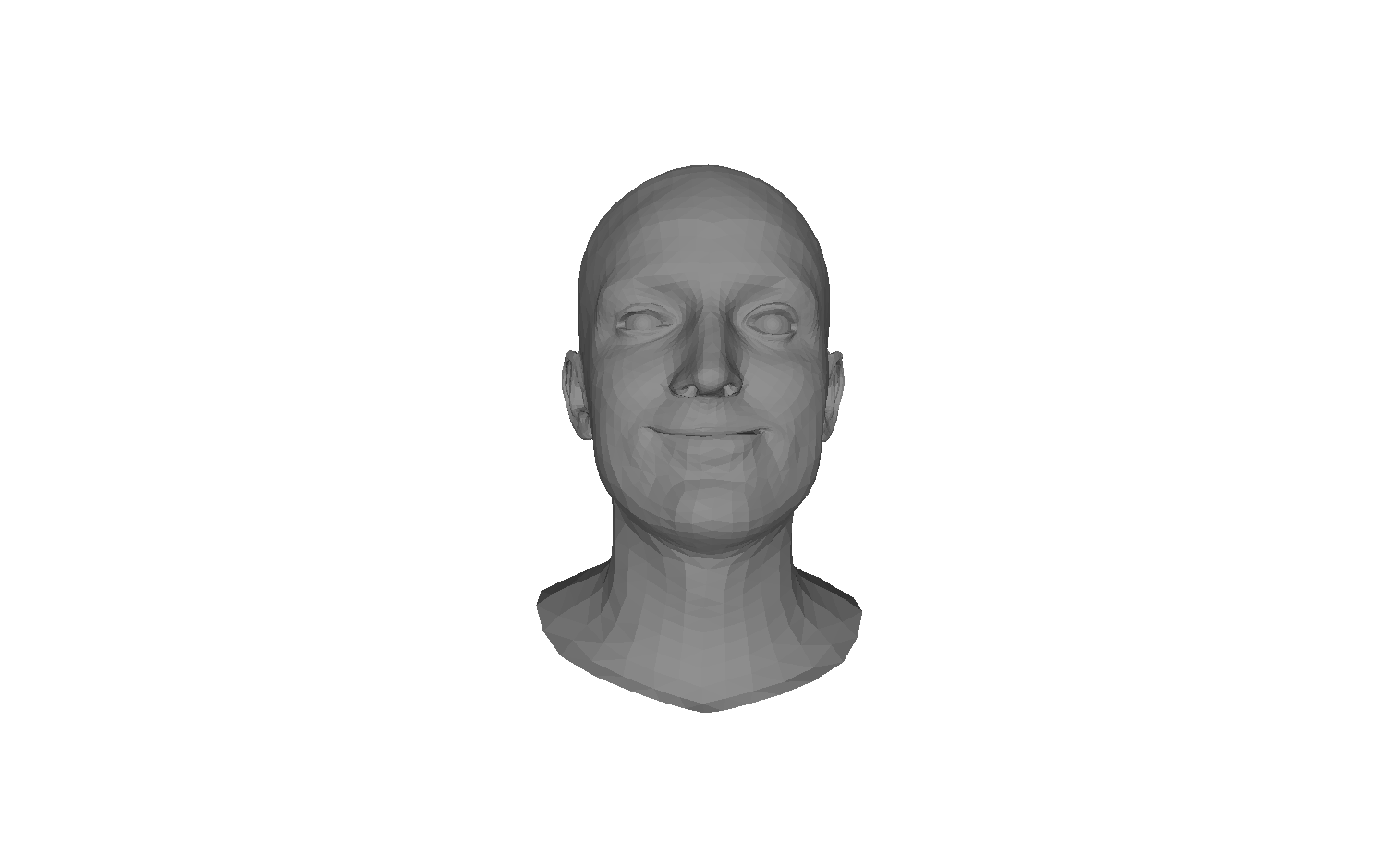}};
    \node[right of=iter1_3, node distance=0.8cm] (iter1_dots) {$\hdots$};
    \node[right of=iter1_dots, node distance=0.8cm] (iter1_4) {\includegraphics[trim={400 100 400 100},clip,width=0.065\linewidth]{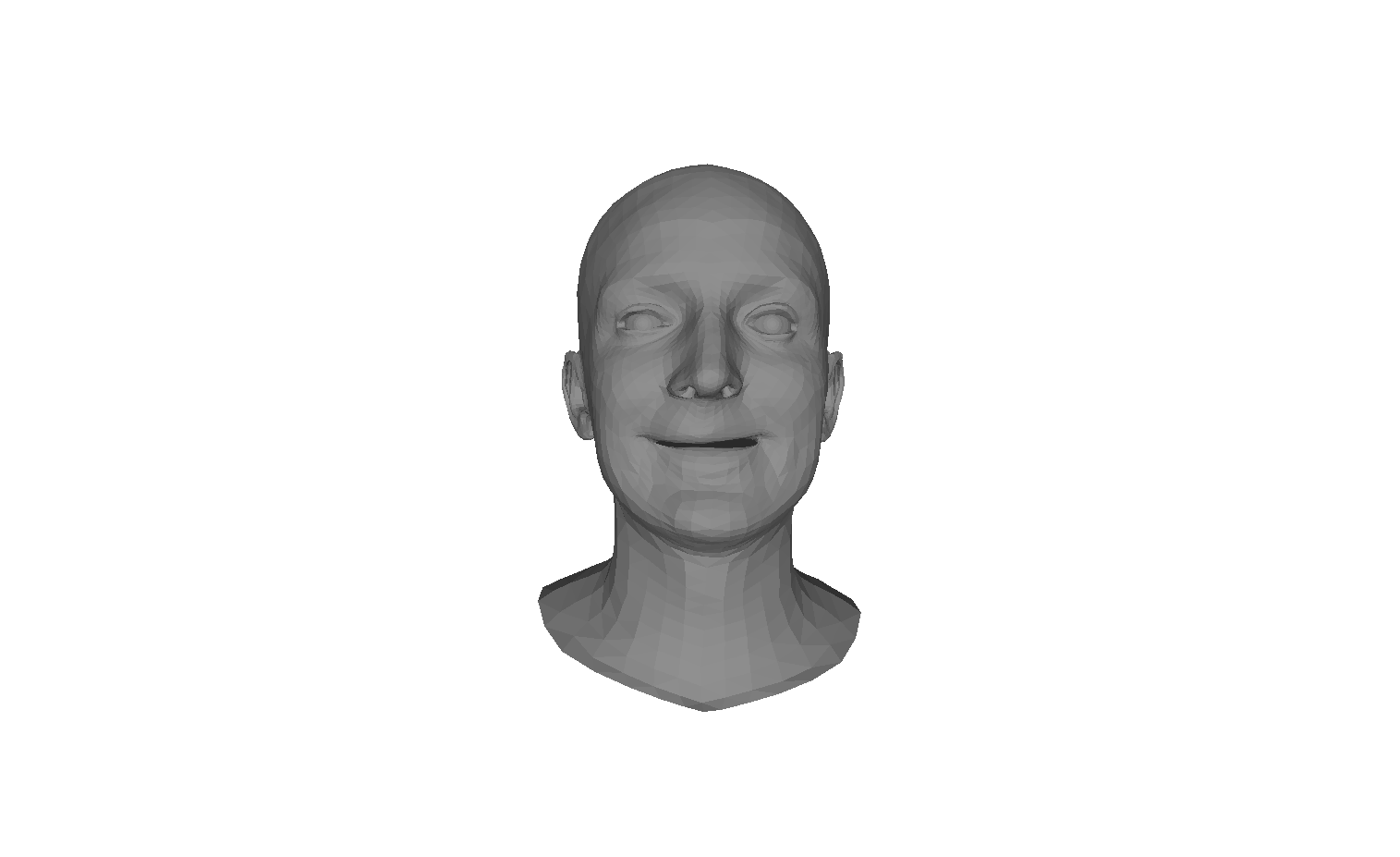}};
    \node[below of=dec3, node distance=1cm] {$\mathcal{D}_{mesh}$};
    \draw[->] (iter0_1) -- (iter1_1);
    \draw[->] (tsne_z2) -- (dec1);
    
    \node [below of=tsne_z2, node distance=3.5cm] (tsne_z3) {\includegraphics[width=.1\linewidth]{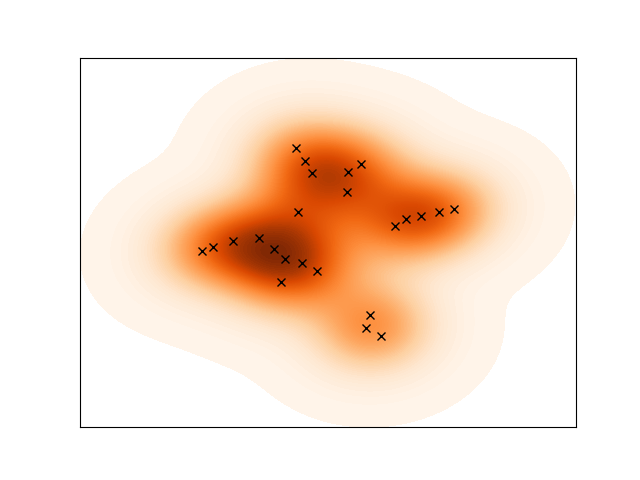}};
    \node [below of=tsne_z3, node distance=0.9cm] {$\mathbf{z}(t=N_{comp})$};
    \node[right of=tsne_z3, node distance=1.2cm, dec_layer, minimum width=0.5cm] (dec1) {};
    \node[right of=dec1, node distance=0.2cm, dec_layer, minimum width=0.8cm] (dec2) {};
    \node[right of=dec2, node distance=0.2cm, dec_layer, minimum width=1.2cm] (dec3) {};
    \node[right of=dec3, node distance=0.2cm, dec_layer, minimum width=1.5cm] (dec4) {};
    \node[right of=dec4, node distance=0.8cm] (iter2_1) {\includegraphics[trim={400 100 400 100},clip,width=0.065\linewidth]{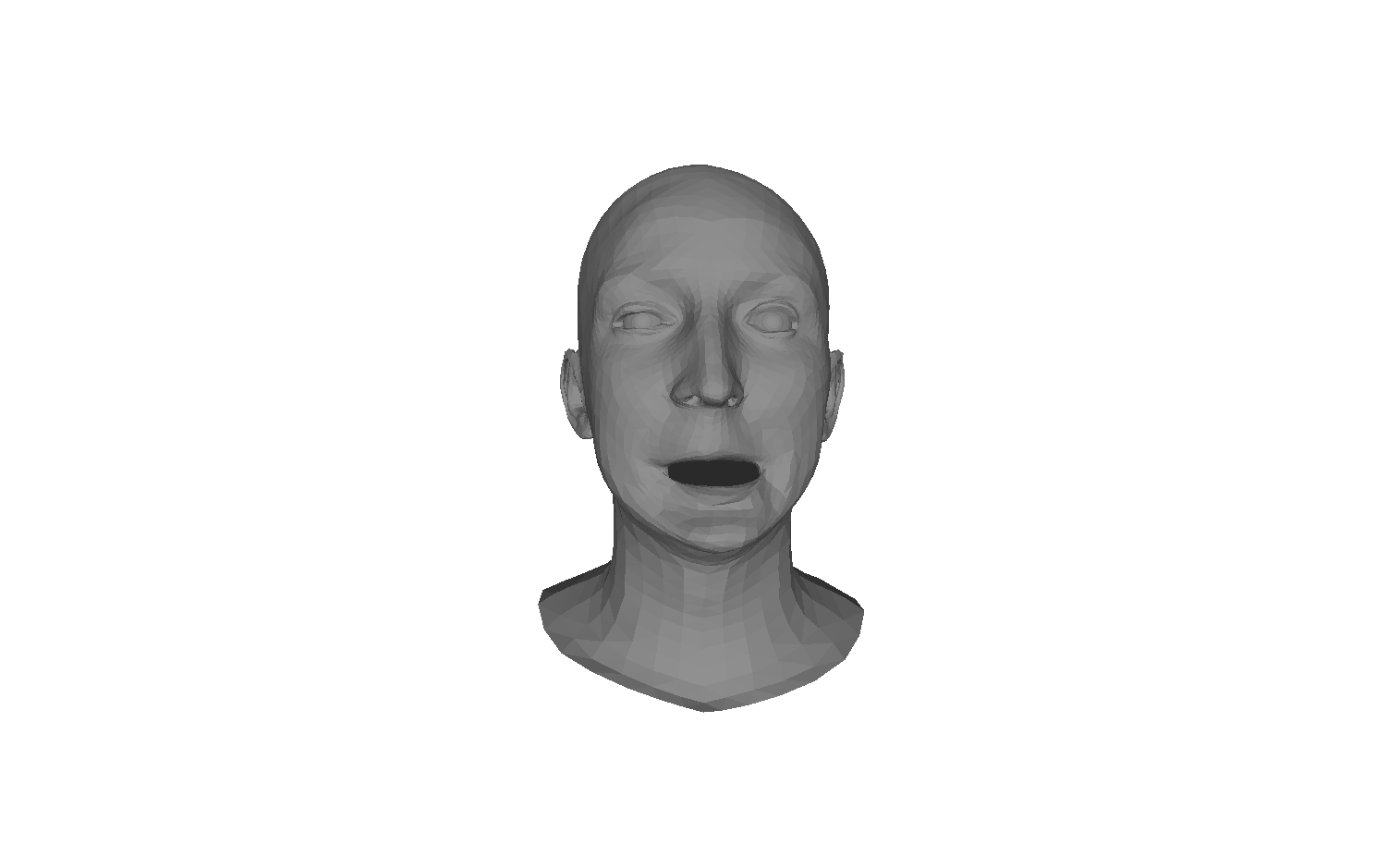}};
    \node[right of=iter2_1, node distance=1cm] (iter2_2) {\includegraphics[trim={400 100 400 100},clip,width=0.065\linewidth]{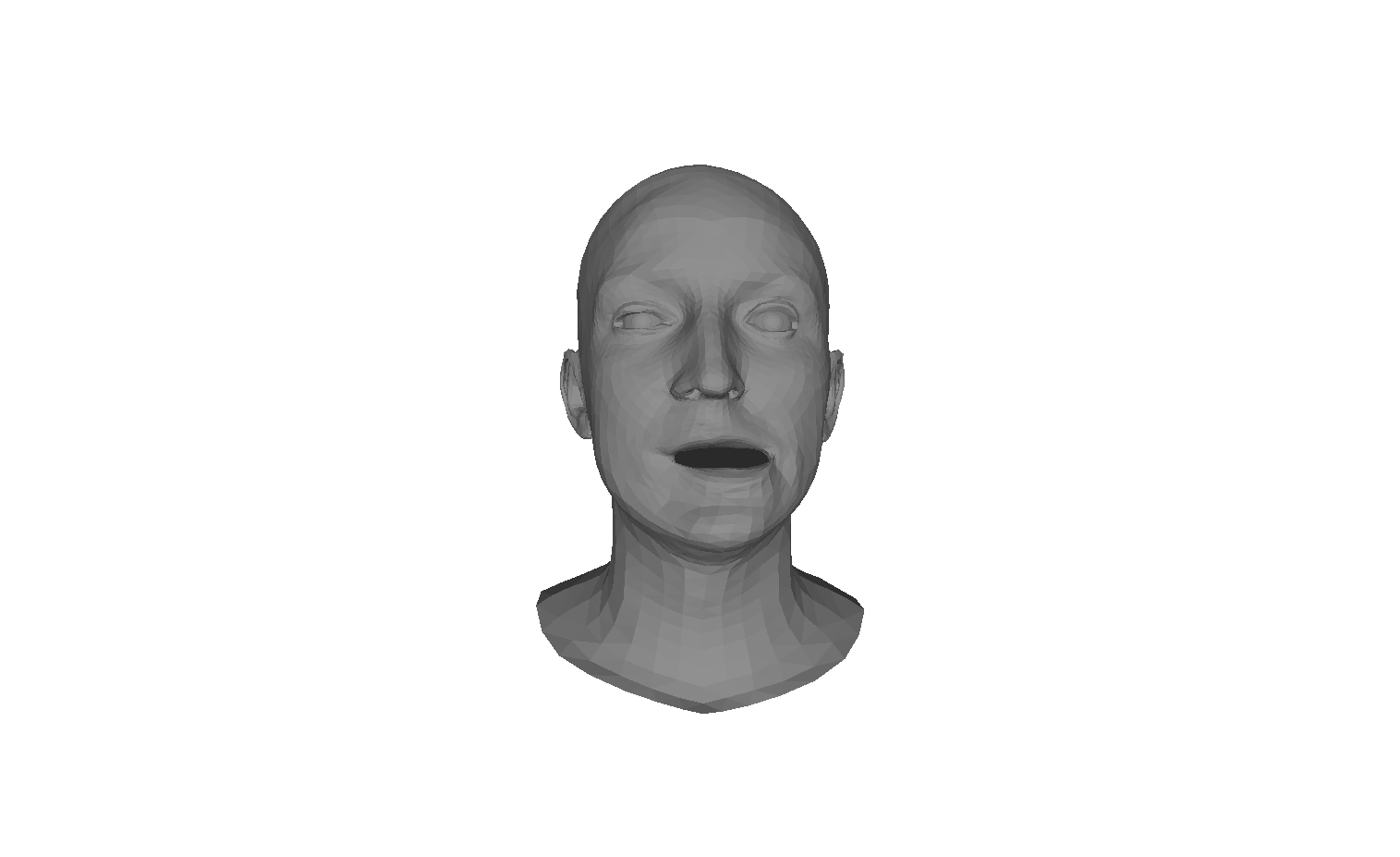}};
    \node[right of=iter2_2, node distance=1cm] (iter2_3) {\includegraphics[trim={400 100 400 100},clip,width=0.065\linewidth]{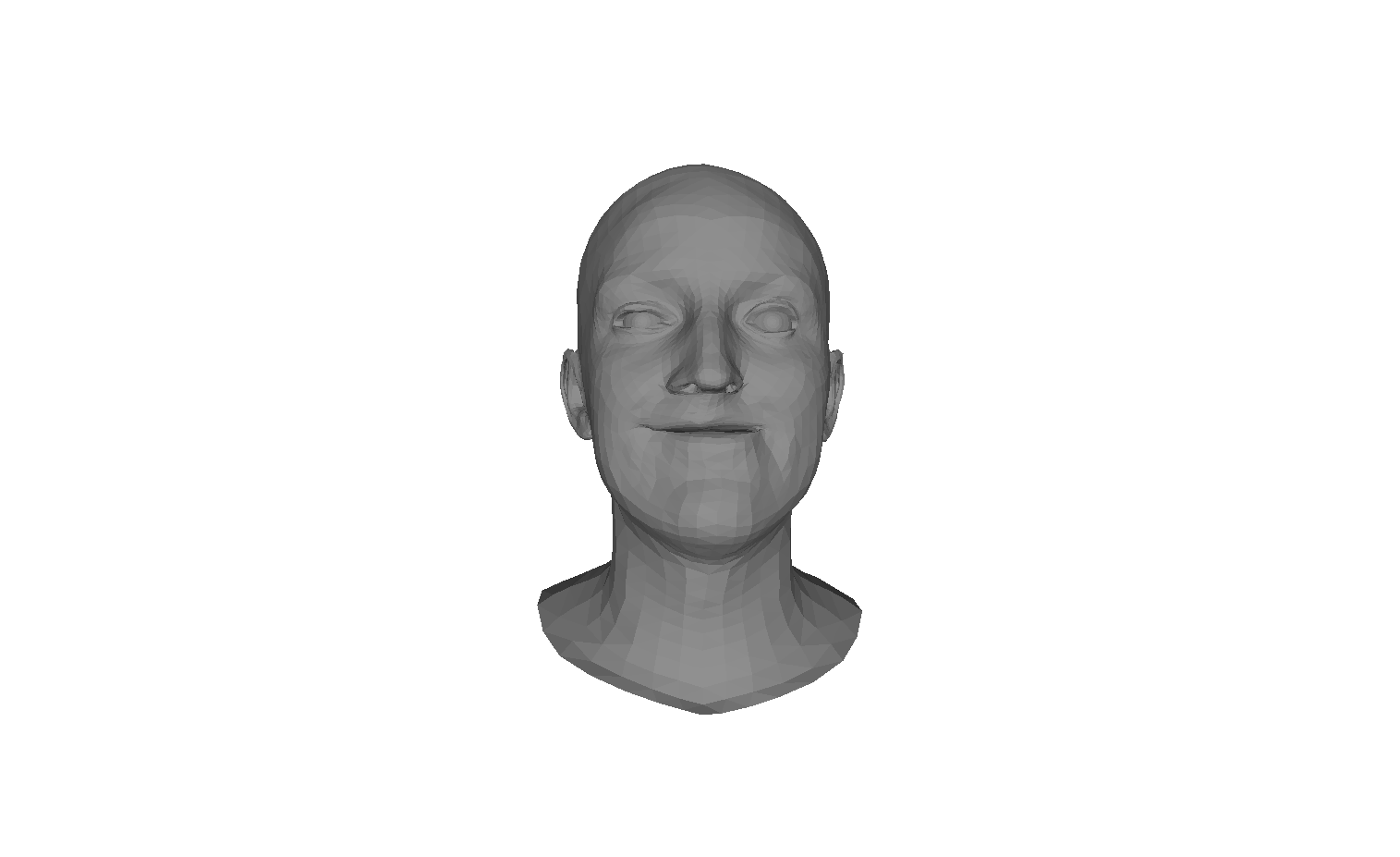}};
    \node[right of=iter2_3, node distance=0.8cm] (iter2_dots) {$\hdots$};
    \node[right of=iter2_dots, node distance=0.8cm] (iter2_4) {\includegraphics[trim={400 100 400 100},clip,width=0.065\linewidth]{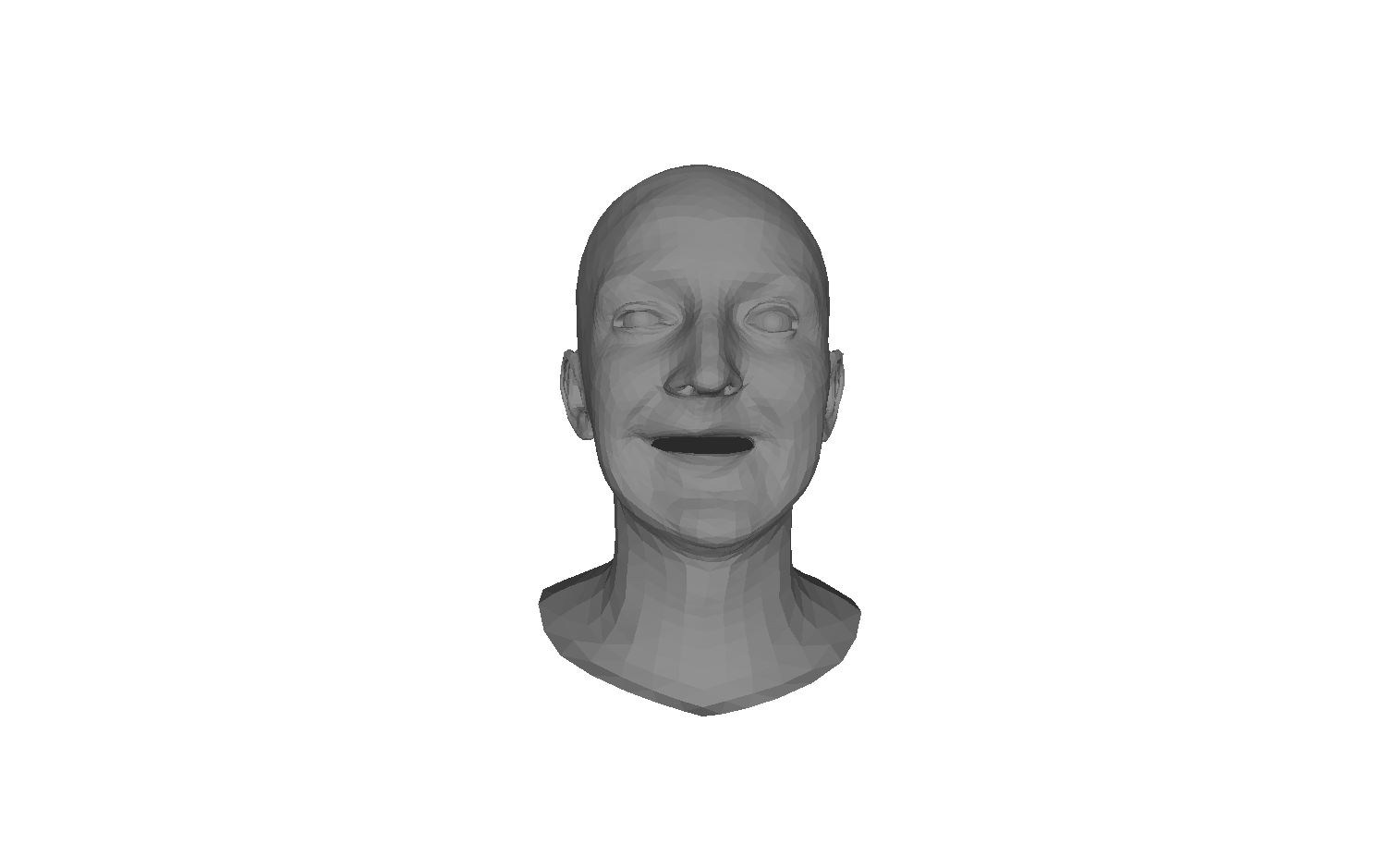}};
    \node[below of=dec3, node distance=1cm] {$\mathcal{D}_{mesh}$};
    \draw[->] (iter1_1) -- ++ (0,-1.3);
    \node[below of=iter1_1, node distance=1.7cm] {$\vdots$};
    \draw[<-] (iter2_1) -- ++ (0,1.3);
    \draw[->] (tsne_z3) -- (dec1);
    
    \coordinate[left of=z0_label, node distance=1.1cm] (z0_label_left);
    \draw[->] (z0_label_left) -- node [midway, above, rotate=90] {Diversity Iterations over $\mathbf{z}(t)$} ++(0,-5);
    
    \node [block,dashed,minimum width=8.5cm,minimum height=8.6cm,below of=iter1_1,node distance=0.35cm,fill opacity=0,draw=orange!80,label={[yshift=-0.05cm,anchor=north,text opacity=0.8]north:\normalsize{\textbf{Diverse Shape Completion} \textit{using} $\boldsymbol{L_{diverse}}$}}] {}; 
    \end{tikzpicture}
    \caption{\textbf{Overview:} As input, we need the target image, the occlusion mask, facial landmarks, and optionally a face mask. We use the HRNET model \cite{hrnet} to obtain both the landmark locations and their confidence values, which we use to estimate the occlusion labels. Given these input, we first fit our proposed \textit{global + local blendshape model} to obtain the coarse and local fittings as outlined in ~\cref{alg:fitting}, which we then add together to obtain the final fitting. We re-project the fitted shape onto the visible mask to obtain a partial fit, zeroed out on the occluded regions. We map the partial fit onto a latent space using the \textit{Mesh-VAE} encoder $\mathcal{E}_{mesh}$ and sample $N$ latent vectors $\mathbf{z}$. We then iteratively optimize the $\mathbf{z}$'s to capture diverse modes with respect to the occluded regions while remaining consistent with the visible regions as outlined in ~\cref{alg:diversity} to obtain the final set of 3D reconstructions.}
    \label{fig:overview}
\end{figure*}

Reconstructing diverse 3D shapes in a single stage, using only a global model, is sub-optimal due to multiple reasons, as we show in our experiments (\cref{subsec:quantitative}). First, fitting a global model to a few visible sub-regions requires striking a careful trade-off between robustness and local fidelity which is challenging to achieve. Second, diversification of the occluded regions will inadvertently affect the quality of fitting on the visible regions, and vice-versa. Given these observations, we propose a three-step approach to generate diverse, yet realistic 3D reconstructions from an occluded face image. In step 1, we use an ensemble of disentangled global+local shape models to perform robust 3D reconstruction w.r.t the visible parts of the face. In step 2, we employ a VAE to map the partial fit to a latent space from which multiple reconstructions can be drawn. Finally, in step 3 we iteratively optimize the latent embeddings to promote realistic geometric diversity on the occluded face regions while maintaining fidelity to the visible ones. We now describe our complete algorithm along with its different components.

\subsection{Global + Local Shape Model}
A robust partial 3D reconstruction that accurately fits the visible parts of the face is a prerequisite for generating diverse solutions. Existing approaches of occlusion-robust 3D reconstruction typically employ a global model to fit or regress based on the visible regions \cite{egger2018occlusion, tran2018extreme}. Because of the \textit{global} nature of such models, errors in occlusion segmentation affect the quality of 3D reconstruction \cite{saito2016real}, even on the visible parts (see ~\cref{fig:flame_vs_component}). Typically, strong regularization is employed to mitigate such effects. However, while heavier regularization leads to more robustness against occlusions, it comes at the cost of sub-optimal fitting. This observation, along with the successful application of localized deformation components in computer graphics \cite{neumann2013sparse, eyeshaveit}, motivated us to adopt an ensemble of global + local models as an effective approach to generate robust 3D reconstructions w.r.t the visible parts. Note that, in this stage of our solution, we are not concerned about the reconstruction quality in the occluded regions. We now describe the details of our proposed global+local 3D head model.

Our global+local shape model is based on the FLAME mesh topology \cite{flame}. We use the FLAME registered D3DFACS \cite{d3dfacs} and CoMA \cite{coma} datasets to compute the local PCA models. The FLAME \cite{flame} model comes with vertex masks corresponding to 14 parts on the human head. We trained individual PCA models corresponding to each of these parts to account for local variations. To do so, we first take FLAME-registered meshes and fit the full FLAME model \cite{flame} to these by optimizing the following fitting loss:
\begin{align}
    L_{fit} = \min_{\beta, \theta, \psi} ||S^{gt} - \Tilde{S}(\beta,\theta,\psi)||,
\end{align}

Here $\Tilde{S}(\beta,\theta,\psi)$ is obtained using ~\cref{eqn:flame,eqn:flame_shape}. We then \textit{unpose} both the ground-truth and the fitted shapes by removing the variations due to pose $\theta$ as described in \cite{flame} and obtain $S^{gt}_0$ and $\Tilde{S}(\beta,0,\psi)$, respectively. The full FLAME model consists of $|\beta|=300$ shapes and $|\psi|=100$ expression bases to account for complete global variations. From this, we retain the top $N_S$ shape and $N_E$ expression bases (based on eigenvalues) and discard the rest to compute shape residuals $\Tilde{S}^{res}=S^{gt}_0 - \Tilde{S}^{coarse}$, where
\begin{align}
    \Tilde{S}^{coarse} = \bar{\mathbf{T}} + \sum_{n=1}^{N_S} \beta_n \mathcal{S}_n + \sum_{n=1}^{N_E} \psi_n \mathcal{E}_n
\end{align}

We then compute the region-wise shape and expression PCA models $(\mathcal{S}^{\mathcal{R}_i}, \mathcal{E}^{\mathcal{R}_i})$ using the region-wise residuals $M_{\mathcal{R}_i}\odot \Tilde{S}^{res}$ (here $M_{\mathcal{R}_i}$ is the vertex-mask for the $i$-th region). For computing the shape bases, we set $N_S=10$ and $N_E=100$ (removing all expression variations); while for the expression bases, we set $N_E=10$ and $N_S=300$ (removing all identity variations). The global + local model can then be represented as,
\begin{small}
\begin{align}
    \label{eqn:global_local_model}
    T(\beta^{\mathcal{G}}, \beta^{\mathcal{R}}, \theta, \psi^{\mathcal{G}}, \psi^{\mathcal{R}})= T_{\mathcal{G}}(\beta^{\mathcal{G}}, \theta, \psi^{\mathcal{G}}) + T_{\mathcal{R}}(\beta^{\mathcal{R}}, \psi^{\mathcal{R}}),
\end{align}
\end{small}
\noindent where $T_{\mathcal{G}}(\beta^{\mathcal{G}}, \theta, \psi^{\mathcal{G}})$ is the coarse global shape given by the top $N_S$ shape and $N_E$ expression global bases, along with the pose blendshapes $\mathcal{P}$ (~\cref{eqn:flame_shape}); and $T_{\mathcal{R}}(\beta^{\mathcal{R}}, \psi^{\mathcal{R}})$ represent the local variations and is given by,
\begin{small}
\begin{align}
    \label{eqn:region_wise_shape}
    \tiny
    T_{\mathcal{R}}(\beta^{\mathcal{R}}, \psi^{\mathcal{R}}) = \sum_{\mathcal{R}_i} \left( \sum_{n=1}^{|\beta^{\mathcal{R}_i}|} \beta^{\mathcal{R}_i}_n \mathcal{S}^{\mathcal{R}_i}_n + \sum_{n=1}^{|\psi^{\mathcal{R}_i}|} \psi^{\mathcal{R}_i}_n \mathcal{E}^{\mathcal{R}_i}_n \right)
\end{align}
\end{small}

\subsection{Shape Completion using Mesh-VAE}
We use the global+local model to fit robust 3D reconstruction on the visible parts of the occluded face. But this does not ensure robust and consistent reconstruction on the occluded parts since the local PCA models have noisy (occluded) or no data to fit to. To address this drawback and to enable the generation of a distribution of plausible 3D reconstructions rather than a singular solution, which is one of our primary goals, we adopt a mesh-based VAE (dubbed \textit{Mesh-VAE}) as our shape completion model.

We assume that human head meshes can be mapped onto a continuous and regularized low-dimensional latent space $\mathcal{Z}$. Then, given a partial 3D mesh $\mathbf{S}_m$, the Mesh-VAE learns the conditional likelihood of mesh completions $\mathbf{S}_c$ and the corresponding latent embeddings $\mathbf{z}$:
\begin{align}
    p(\mathbf{S}_c, \mathbf{z} | \mathbf{S}_m) = p(\mathbf{z} | \mathbf{S}_m) p(\mathbf{S}_c | \mathbf{z}, \mathbf{S}_m),
\end{align}

\subsection{DPP Driven Shape Diversification}
Even though the Mesh-VAE can sample multiple shape completions from $p(\mathbf{S}_c | \mathbf{z}, \mathbf{S}_m)$, in practice, the generated samples from a VAE are not guaranteed to cover all the modes \cite{yuan2019diverse} (see~\cref{subsec:quantitative}). To enforce diversity, we formulate a DPP on shape completions and develop a diversity loss to optimize their latent embeddings.

We adopt the quality-diversity based formulation of the DPP kernel $\mathbf{L}$ \cite{dpp}, which seeks to balance the quality of samples with their diversity. Specifically, for elements $i, j$ in a set, its kernel entry is given by $L_{i,j}=q_iS_{i,j}q_j$, where $q_i$ denotes the quality of element $i$, and $S_{i,j}$ represents the similarity between $i$ and $j$. Maximizing the determinant of such a kernel matrix implies maximizing the quality of each sample while minimizing the similarity between distinct samples. For two shape completions $\mathbf{S}_c^i$ and $\mathbf{S}_c^j$, we define the similarity as
\begin{align}
    S_{i,j} = \exp \left(- \frac{k}{\median_{i,j}( dist_{i,j} )} dist_{i,j} \right),
\end{align}
where $dist_{i,j} = ||\mathbf{S}_c^i - \mathbf{S}_c^j||_2$ is the $\ell_2$ distance between the $i$-th and $j$-th shape completions and $k$ is a scaling factor. To ensure that the completed samples look realistic, we relate the quality of a sample with the probability of its latent embedding $\mathbf{z}_i$ lying within 3$\sigma$ of the prior $\mathcal{N}(\boldsymbol{0, I})$ as:
\begin{align}
    q_i = \exp(- \max(0, \mathbf{z}_i^T\mathbf{z}_i - 3\sqrt{d})),
\end{align}
where $d$ is the dimensionality of $\mathbf{z}_i$. For numerical stability \cite{yuan2019diverse}, we adopt expected cardinality of $\mathbf{L}$ as the DPP loss:
\begin{align}
    \label{eqn:dpp}
    L_{dpp} = - tr \left( \mathbf{I} - (\mathbf{L} + \mathbf{I})^{-1} \right)
\end{align}

\subsection{Inference}
Given an occluded face image $\mathbf{I}_m$, our goal is to generate a distribution of plausible 3D reconstructions $\mathbf{S}_c^1, ..., \mathbf{S}_c^M$. We do this in three steps which we describe below:

\noindent \textbf{Step 1} \textit{Partial Shape Fitting:} In this stage, we first fit our global + local PCA model on the visible parts of the face image $\mathbf{I}_m$ to obtain a partial reconstruction $\mathbf{S}_m$. We employ the following fitting loss:
\begin{align}
    \label{eqn:fitting}
    L_{fitting} = \lambda_1^f L_{lmk} + \lambda_2^f L_{pho} + \lambda_3^f L_{reg},
\end{align}
where $L_{lmk}$ is the landmark loss, $L_{pho}$ is the photometric loss and $L_{reg}$ applies $\ell_2$-regularization over the model parameters. We use an off-the-shelf landmark detector HRNET \cite{hrnet} to detect 68 landmarks on the face along with their confidence values. We mark those landmarks as visible whose confidence exceeds a threshold $\tau$ (set to 0.2) and apply the landmark loss on those points. To add local details, we apply a photometric loss between the input image and a rendered image $\mathbf{I}_{ren}=\mathcal{R}(\mathbf{S}_m, B_{tex}(\gamma, \mathcal{T}), c)$, where $B_{tex}(\gamma, \mathcal{T})$ is the estimated texture and $c$ the estimated camera parameters. We restrict the photometric loss to the visible face region using the face mask $M_f$ and the occlusion mask $M_o$:
\begin{align}
    \label{eqn:photometric}
    L_{pho} = ||(\mathbf{I}_m - \mathbf{I}_{ren})\odot M_f \odot (\mathbf{1}-M_o)||_1
\end{align}

\noindent \textbf{Step 2} We use the encoder to map the partial fit $\mathbf{S}_m$ to a latent distribution from which we sample the latent embeddings $\mathbf{z} \sim \mathcal{N}(\boldsymbol{\mu}, diag(\boldsymbol{\sigma}^2)),$ where $\boldsymbol{\mu, \sigma} = \mathcal{E}_{mesh}(\mathbf{S}_m)$.

\noindent \textbf{Step 3} \textit{Diversity Promoting Shape Completion:} In this stage, we perform a diversity promoting iterative shape completion routine, which forces the latent embeddings towards diverse modes w.r.t the occluded regions while remaining faithful to the visible regions. At each iteration, we obtain a distribution of shape completions using the decoder $\mathbf{S}_c^j=\mathcal{D}_{mesh}(\mathbf{z}_j), \forall j=1...M$, and update the $\mathbf{z}$'s to minimize a diversity loss:
\begin{align}
    \label{eqn:diversity}
    L_{diversity} = \lambda_1 L_S + \lambda_2 L_{pho} + \lambda_3 L_{dpp}
\end{align}
Here $L_S$ is the shape consistency loss defined as the $\ell_1$-norm between the $\mathbf{S}_c^j$'s and $\mathbf{S}_m$ applied on the visible vertices, $L_{pho}$ is the photometric loss (\cref{eqn:photometric}) and $L_{dpp}$ is the DPP loss (\cref{eqn:dpp}). The loss coefficients are set to have similar magnitude for all the loss components.

We outline the full steps for partial shape fitting and diversification in ~\cref{alg:fitting} and ~\cref{alg:diversity}, respectively.

\begin{algorithm}
\caption{Shape Fitting on the Visible Face Regions}\label{alg:fitting}
\textbf{Input:} Image $\mathbf{I}_m$, Occlusion mask $M_o$, Face mask $\mathbf{M}_f$, Global models $\mathcal{S}, \mathcal{E}, \mathcal{P}$, Local models $\mathcal{S}^{\mathcal{R}_i}$, $\mathcal{E}^{\mathcal{R}_i}$ for $i=1\text{ to }14$, Texture model $\mathcal{T}$, Landmarks detector $\mathcal{H}$\\
\textbf{Parameters:} $\beta, \theta, \psi, \gamma, c, \beta^{\mathcal{R}_i}, \psi^{\mathcal{R}_i}$ for $i=1\text{ to }14$\\
\textbf{Hyperparameters:} $\tau=0.1, n_{iter}, \lambda_1^f, \lambda_2^f, \lambda_3^f, \eta$\\
\textbf{Output:} Partially fitted shape $\mathbf{S}_m$
\begin{algorithmic}
\State Detect landmarks from image  $\mathbf{L}_I, \mathbf{L}_{conf} \gets \mathcal{H}(\mathbf{I}_m)$
\State Set $\mathbf{L}_{valid} \gets 1 \text{ when } \mathbf{L}_{conf} > \tau$ else 0
\For{$j=1$ to $n_{iter}$}
\State Obtain $\mathbf{S}_m$ using ~\cref{eqn:global_local_model,eqn:flame_shape,eqn:flame,eqn:region_wise_shape}
\State Select 68 landmarks from shape $\mathbf{L}_S \gets M_{lmk}(\mathbf{S})$
\State Obtain rendered image $\mathbf{I}_{ren} \gets \mathcal{R}(\mathbf{S}, B_{tex}(\gamma, \mathcal{T}), c)$
\State $L_{lmk}^f \gets ||(\mathbf{L}_S - \mathbf{L}_I) \odot \mathbf{L}_{valid}||_1$
\State $L_{pho}^f \gets ||(\mathbf{I}_m - \mathbf{I}_{ren})\odot\mathbf{M}_f \odot (\mathbf{1}-M_o)||_1$
\State $L_{reg}^f \gets \ell_2$ regularization loss over all parameters
\State $L_{fitting} = \lambda_1^f L_{lmk}^f + \lambda_2^f L_{pho}^f + \lambda_3^f L_{reg}^f$
\State Update $p \gets p - \eta \nabla_p L_{fitting}$ for $p \in \beta, \theta, \psi, \gamma, c,  \beta^{\mathcal{R}_i}, \psi^{\mathcal{R}_i}$ for $i=1\text{ to }14$
\EndFor
\end{algorithmic}
\end{algorithm}

\begin{algorithm}
\caption{Diverse Shape Completions}\label{alg:diversity}
\textbf{Input:} Mesh-VAE Encoder $\mathcal{E}_{mesh}$ and Decoder $\mathcal{D}_{mesh}$; From~\cref{alg:fitting}: $\mathbf{I}_m, M_o, \mathbf{M}_f, \mathbf{L}_I, \mathbf{L}_{valid}, \theta, \gamma, c, \mathcal{T}$\\
\textbf{Hyperparameters:} $n_{comp}, \lambda_1, \lambda_2, \lambda_3, \eta$\\
\textbf{Output:} $M$ Shape completions $\{\mathbf{S}_{c}^{j=1:M}\}$
\begin{algorithmic}
\State Sample the vertex mask $M_o^v$ by projecting $\mathbf{S}$ onto $M_o$
\State Obtain latent parameters $\boldsymbol{\mu, \sigma} \gets \mathcal{E}_{mesh}(\mathbf{S}_m \odot M_o^v)$
\State Sample $M$ latent vectors $\mathbf{z}_1,...,\mathbf{z}_M \sim \mathcal{N}(\boldsymbol{\mu, \sigma^2I})$
\For{$i=1$ to $n_{comp}$}
\State Obtain $\mathbf{S}_c^j \gets \mathcal{D}_{mesh}(\mathbf{z}_j)$ for $j=1...M$
\State Obtain $\mathbf{I}_{ren,j} \gets \mathcal{R}(\mathbf{S}_c^j, B_{tex}(\gamma, \mathcal{T}), c)$ for $j=1...M$
\State $L_S \gets \sum_{j=1}^M ||(\mathbf{S}_c^j - \mathbf{S}_m) \odot (\mathbf{1} -M_o^v)||_1$
\State $L_{pho} \gets \sum_{j=1}^M ||(\mathbf{I}_m - \mathbf{I}_{ren,j})\odot\mathbf{M}_f \odot (\mathbf{1}-M_o)||_1$
\State $L_{dpp} \gets \mathcal{L}_{dpp}(\mathbf{S}_c^{j=1:M} \odot M_o^v)$ using ~\cref{eqn:dpp}
\State $L_{diversity} = \lambda_1 L_{S} + \lambda_2 L_{pho} + \lambda_3 L_{dpp}$
\State Update $\mathbf{z}_j \gets \mathbf{z}_j - \eta \nabla_{\mathbf{z}_j} L_{diversity}$ for $j=1\text{ to }M$
\EndFor
\end{algorithmic}
\end{algorithm}

\section{Experimental Evaluation} \label{sec:experiments}

\noindent\textbf{Datasets:} We use the FLAME \cite{flame} registered head meshes from the CoMA \cite{coma} and D3DFACS \cite{d3dfacs} datasets for training the Mesh-VAE, as well as for evaluating the proposed approach. Note that, other than the Mesh-VAE, our approach does not involve training any other modules. We split the two datasets into 80:10:10 train:val:test splits based on subject ID. We train the Mesh-VAE model using the combined training splits from the two datasets. During training, we augment the meshes with occlusion masks of random (contiguous) shapes at random locations. To evaluate our approach, we use the test split of the CoMA dataset \cite{coma} consisting of subjects that were excluded from training. Furthermore, we conduct a qualitative evaluation on the un-annotated images from the CelebA dataset \cite{celeba}. For both datasets, the test images are artificially augmented with occlusions such as masks, glasses, and other random objects.

\noindent\textbf{Implementation:} We implement the Mesh-VAE as a fully convolutional graph neural network (GNN) based upon the MeshConv architecture presented in \cite{meshconv}. MeshConv \cite{meshconv} uses spatially varying convolution kernels to account for the irregularity of local mesh structures and was shown to outperform fixed kernel-based GNN approaches \cite{gcn, chebconv, graphconv, gat, coma, neural3dmm} on reconstruction tasks. To train Mesh-VAE as a shape completion model, we augment the training meshes with random continuous masks covering 25-40\% of the vertices. However, in practice, directly training the Mesh-VAE for inpainting is very challenging, especially with large degrees of occlusions. We adopt a curriculum learning \cite{bengio2009curriculum} approach to overcome this challenge and progressively introduce larger occlusions during the training process, i.e., we start with easier shape completion tasks and progressively increase its difficulty. We use a combination of $\ell_1$-reconstruction, $\ell_1$-Laplacian, and the KL-divergence losses to train the network. Note that we do not use partial shape completions fitted to occluded face images using either the FLAME \cite{flame} or our global+local model to train the Mesh-VAE, and instead use ground truth meshes to avoid any bias towards either shape model.

\noindent\textbf{Baselines:} To evaluate the efficacy of \ourmethod{} in terms of diversity and robustness to occlusions, we compare against baselines such as FLAME \cite{flame}, DECA \cite{deca}, CFR-GAN \cite{occrobustwacv}, Occ3DMM \cite{egger2018occlusion} and Extreme3D \cite{tran2018extreme} using publicly available implementations or pretrained models (wherever applicable). Due to the difficulty and unreliability in obtaining dense correspondence between FLAME and other mesh topologies, we perform a quantitative comparison only against methods based on the FLAME \cite{flame} topology. In other cases, we report qualitative comparisons based on face images with various occlusions patterns.

\noindent\textbf{Metrics:} The goal of this paper is to generate diverse yet realistic 3D reconstructions of occluded face images. Such an approach should have three desired qualities: 1) the reconstructed shapes should fit as accurately as possible to the visible regions, 2) the occluded regions should be diverse from each other, and 3) at least one of the reconstructed shapes should be very similar to the ground truth shape. There is no prior work on diverse 3D reconstruction, and as such, there are no established metrics. So we define the following three metrics to evaluate the aforementioned qualities: (1) \textbf{Closest Sample Error (CSE)}: the per-vertex $\ell_2$-error between the ground-truth shape and the closest reconstructed shape (lower is better), (2) \textbf{Average Self Distance-Visible (ASD-V)}: the per-vertex $\ell_2$-distance on the visible regions between a 3D completion and its closest neighbor, averaged across all the samples (lower is better), and (3) \textbf{Average Self Distance-Occluded (ASD-O)}: ASD on occluded regions (higher is better). These metrics are inspired by those defined for diverse trajectory forecasting~\cite{yuan2019diverse}.

\begin{table}[!h]
    \centering
    \resizebox{\columnwidth}{!}{
    \begin{tabular}{c|ccc}
        \hline
        \textbf{Occlusion} & DECA \cite{deca} & FLAME \cite{flame} & Global+Local (Ours)\\
        \hline
        Glasses & 57.83 & 47.89 & \textbf{39.98}\\
        Face-mask & 61.18 & 30.37 & \textbf{30.11}\\
        Random & 70.34 & 47.56 & \textbf{38.27}\\
        \hline
        Overall & 62.91 & 41.24 & \textbf{35.85}\\
        \hline
    \end{tabular}
    }
    \caption{Comparison of 3D reconstruction accuracy evaluated in terms of mean shape error (MSE) $\times 10^{-3}$.}
    \label{tab:fitting}
\end{table}

\begin{table*}
\centering
\setlength{\tabcolsep}{3pt}
\newcolumntype{C}[1]{>{\centering\let\newline\\\arraybackslash\hspace{0pt}}m{#1}}
\resizebox{\textwidth}{!}{
\begin{tabular}{c|C{1.3cm}C{1.6cm}C{1.7cm}|C{1.3cm}C{1.6cm}C{1.7cm}|C{1.3cm}C{1.6cm}C{1.7cm}|C{1.3cm}C{1.6cm}C{1.7cm}|ccc}
    \hline
    \textbf{Occlusion} & \multicolumn{3}{c|}{FLAME+DPP} & \multicolumn{3}{c|}{Global+Local+DPP} & \multicolumn{3}{c|}{Global+Local+VAE} & \multicolumn{3}{c|}{FLAME+VAE+DPP} & \multicolumn{3}{c}{Global+Local+VAE+DPP (Ours)}\\
    \textbf{Type} & \textbf{CSE} $(\downarrow)$ & \textbf{ASD-V} $(\downarrow)$ & \textbf{ASD-O} $(\downarrow)$ & \textbf{CSE} $(\downarrow)$ & \textbf{ASD-V} $(\downarrow)$ & \textbf{ASD-O} $(\uparrow)$ & \textbf{CSE} $(\downarrow)$ & \textbf{ASD-V} $(\downarrow)$ & \textbf{ASD-O} $(\uparrow)$ & \textbf{CSE} $(\downarrow)$ & \textbf{ASD-V} $(\downarrow)$ & \textbf{ASD-O} $(\uparrow)$ & \textbf{CSE} $(\downarrow)$ & \textbf{ASD-V} $(\downarrow)$ & \textbf{ASD-O} $(\uparrow)$\\
    \hline
    Glasses & 41.26 & 3.83 & 3.26 & 38.17 & 2.25 & 3.11 & \textbf{32.88} & 1.01 & 1.38 & 42.58 & 0.63 & 4.43 & 36.30 & \textbf{0.61} & \textbf{4.50}\\
    Face-mask & 28.14 & 3.07 & 4.58 & 28.06 & 2.30 & 3.57 & \textbf{25.95} & 0.89 & 1.79 & 27.97 & \textbf{0.61} & 7.88 & 27.58 & 0.85 & \textbf{7.89}\\
    Random & 43.12 & 3.61 & 4.06 & 38.85 & 2.59 & 3.51 & \textbf{36.58} & 0.97 & 1.61 & 43.00 & 0.78 & 5.44 & 39.11 & \textbf{0.72} & \textbf{5.62}\\
    \hline
    Overall & 36.81 & 3.61 & 4.06 & 34.55 & 2.35 & 3.39 & \textbf{31.18} & 0.95 & 1.59 & 37.45 & 0.77 & 5.92 & 33.71 & \textbf{0.73} & \textbf{6.05}\\
    \hline
\end{tabular}}
\caption{Evaluation of diverse reconstructions by the baselines \versus{} \ourmethod{} in terms of CSE, ASD-V and ASD-O (in order of $10^{-3}$).\label{tab:diversity}}
\vspace{-0.6cm}
\end{table*}

\begin{figure*}
    \centering
    \begin{tikzpicture}
    \footnotesize
    \node (a1) {\includegraphics[width=0.09\linewidth]{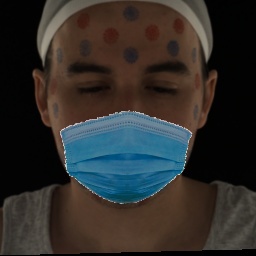}};
    \node[right of=a1, node distance=1.7cm] (a2) {\includegraphics[trim={400 80 400 100},clip,width=0.08\linewidth]{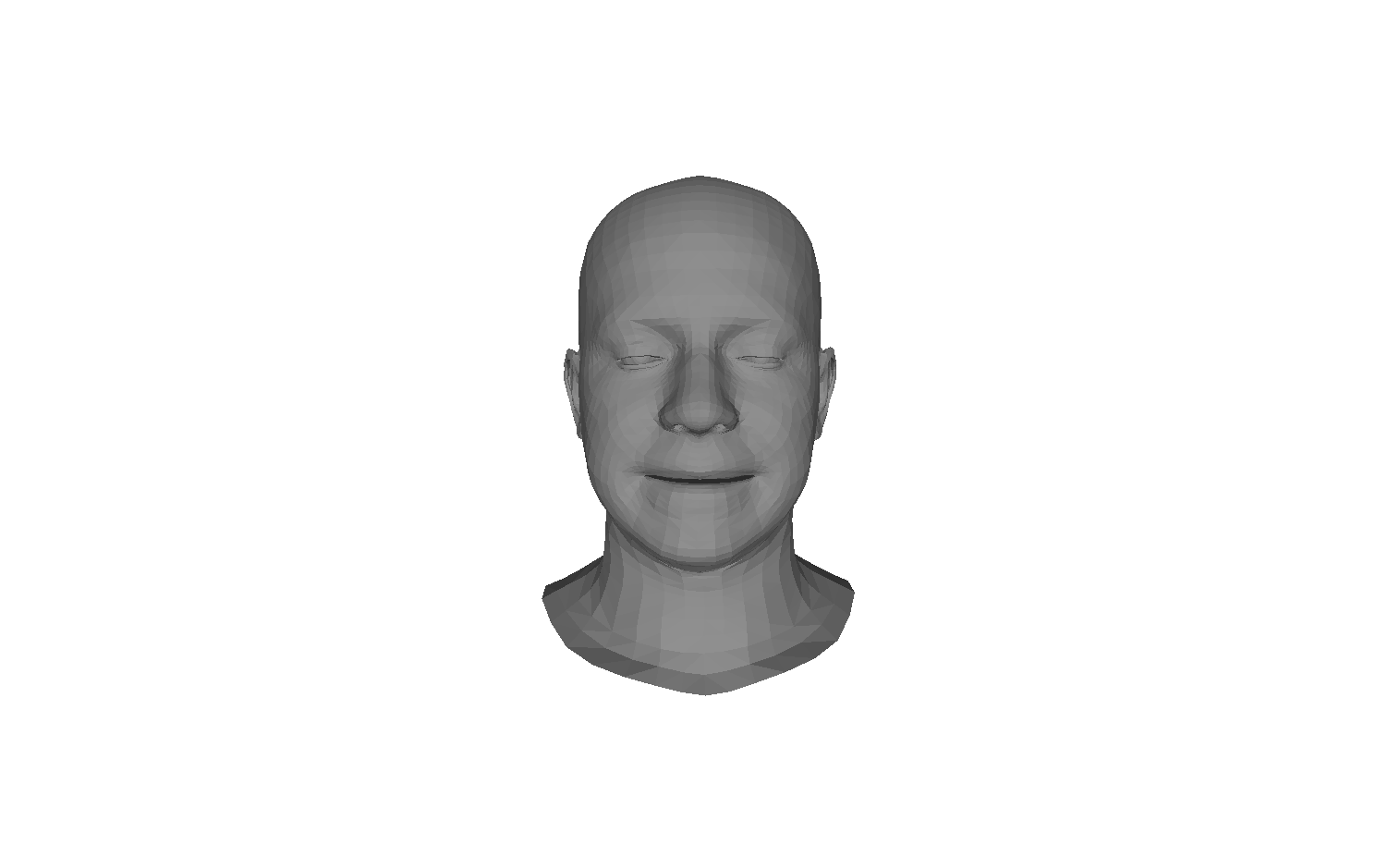}};
    \node[right of=a2, node distance=1.5cm] (a3) {\includegraphics[trim={400 80 400 100},clip,width=0.08\linewidth]{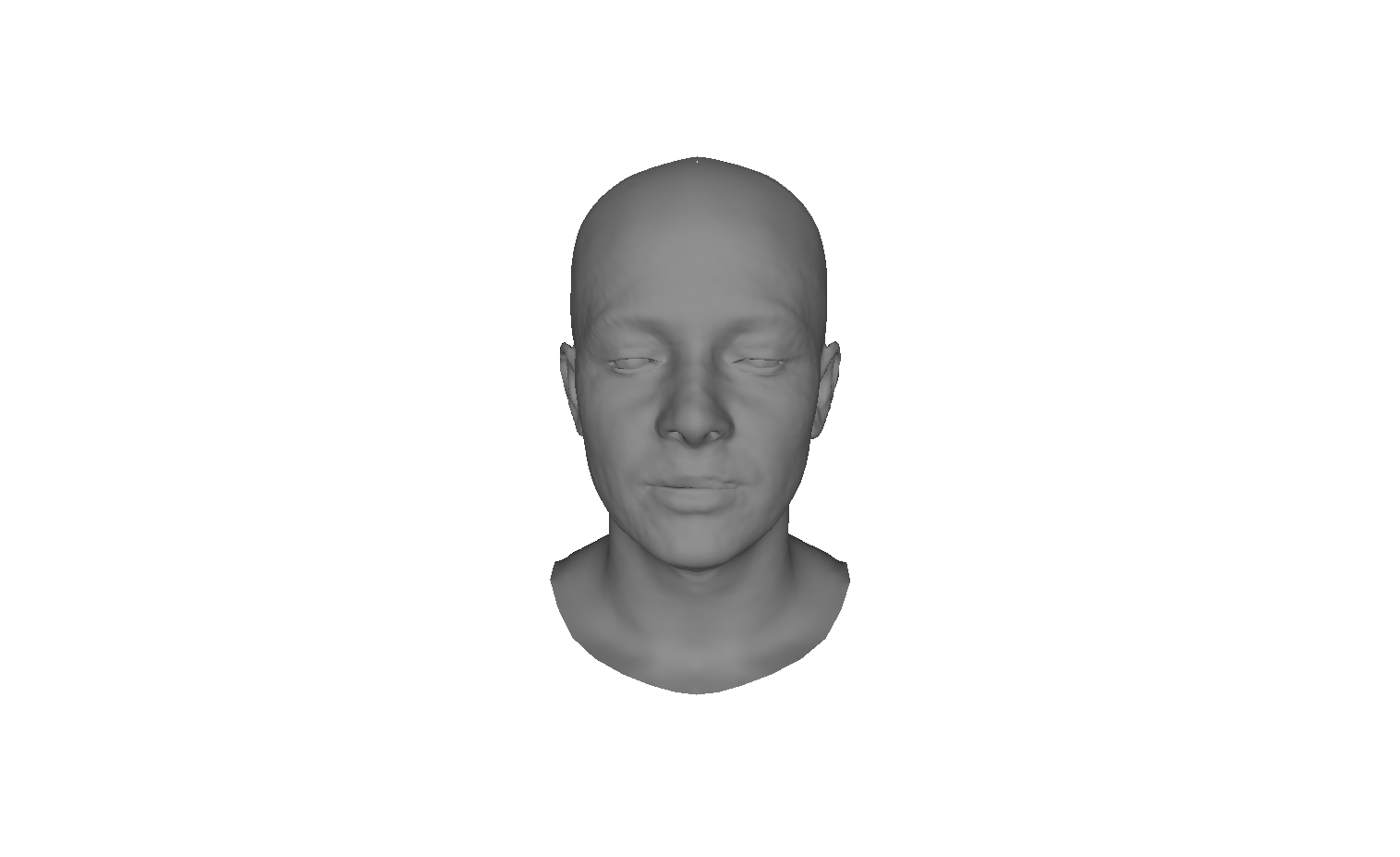}};
    \node[right of=a3, node distance=1.6cm] (a4) {\includegraphics[trim={400 80 400 100},clip,width=0.075\linewidth]{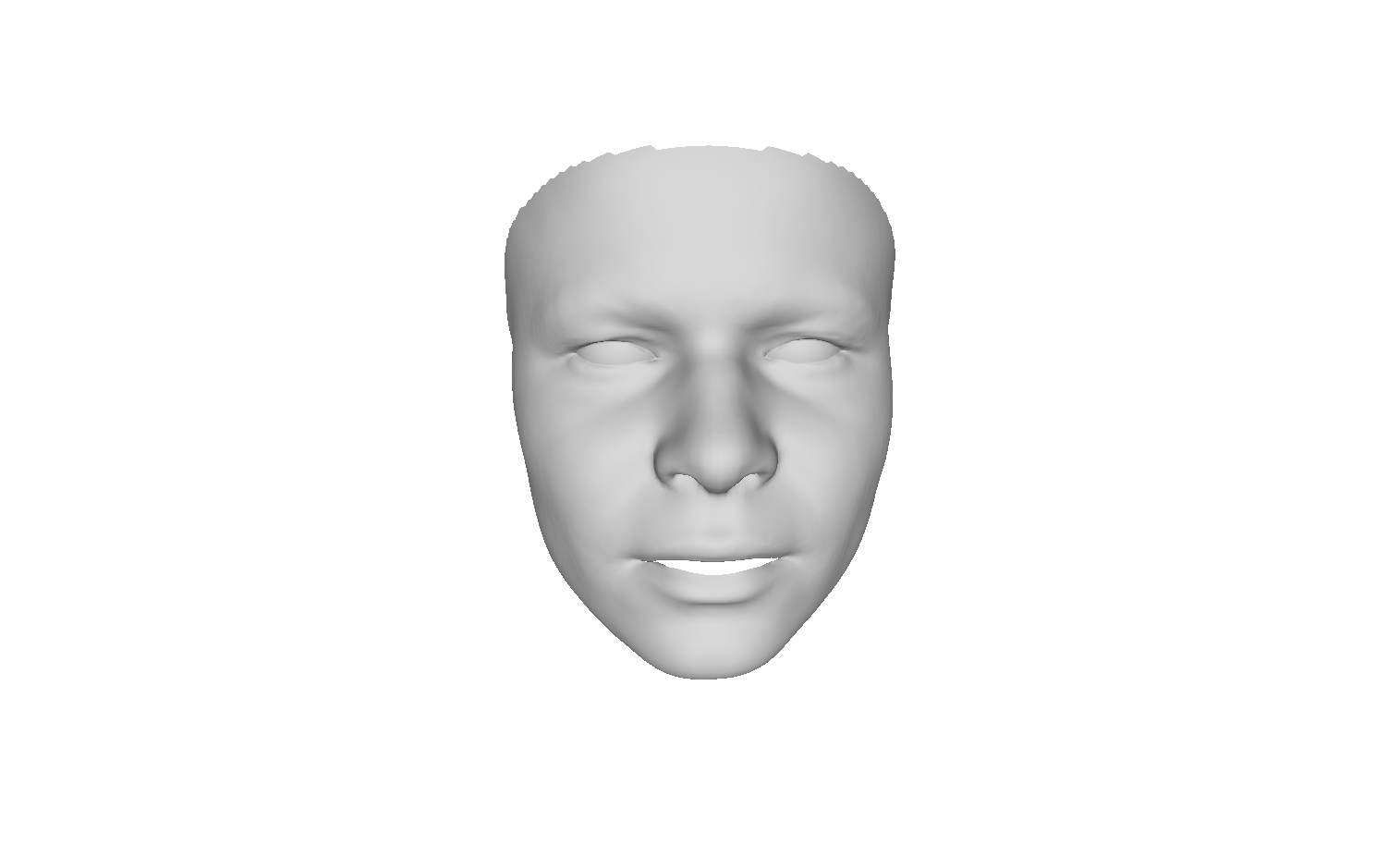}};
    \node[right of=a4, node distance=1.8cm] (a5) {\includegraphics[trim={400 80 400 100},clip,width=0.07\linewidth]{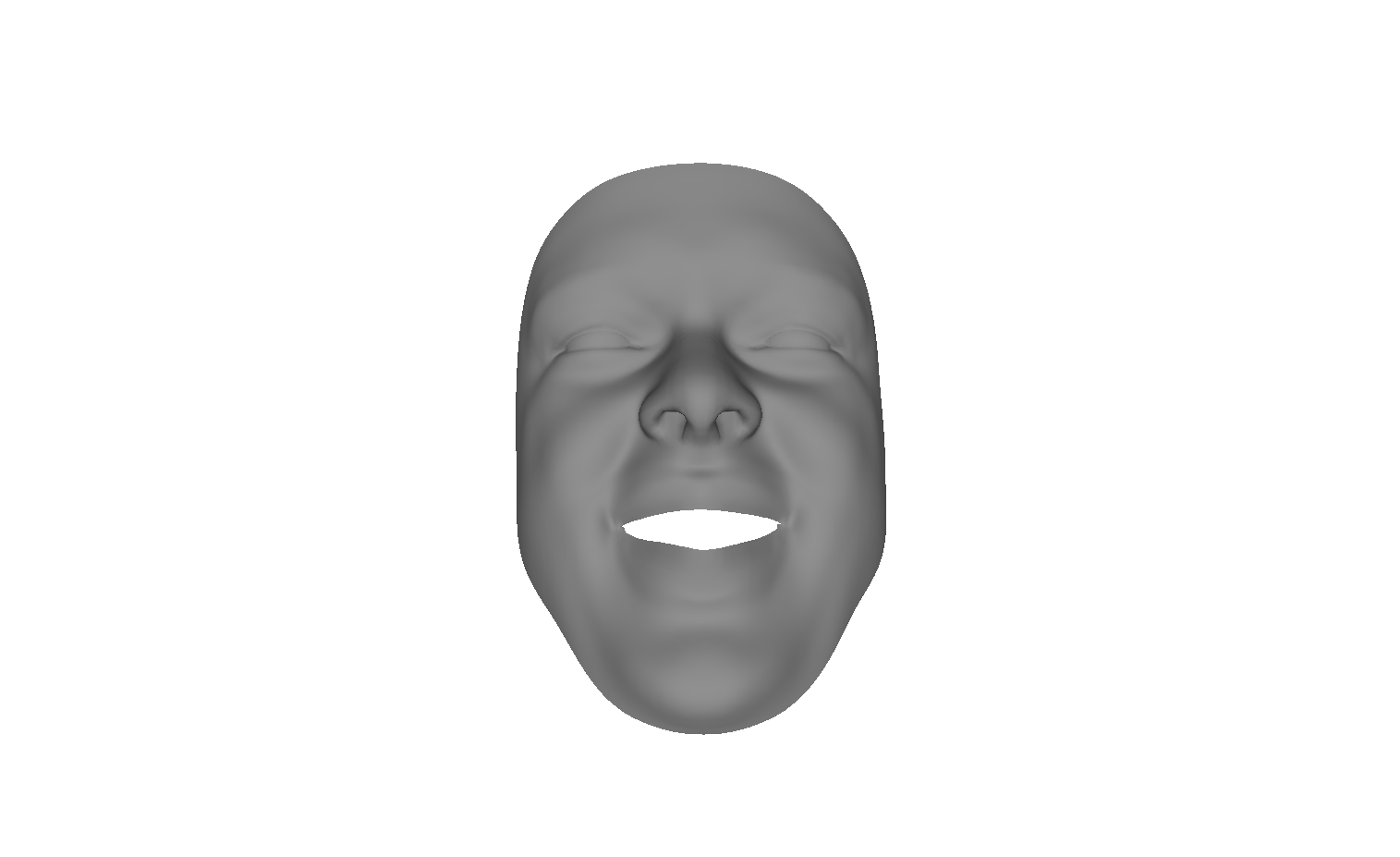}};
    \node[right of=a5, node distance=1.9cm] (a6) {\includegraphics[trim={400 80 400 100},clip,width=0.085\linewidth]{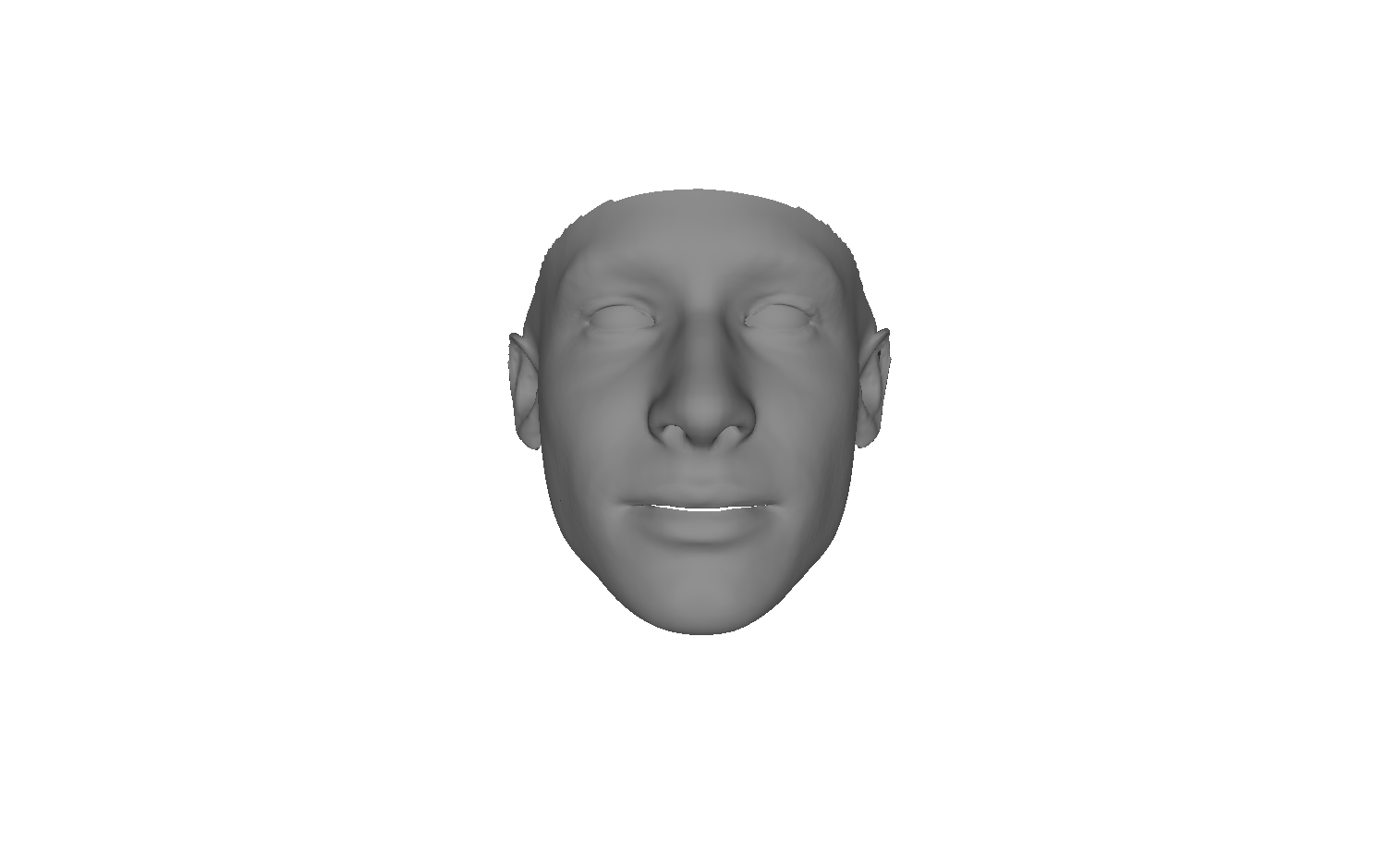}};
    
    \node[right of=a6, node distance=1.8cm] (a7) {\includegraphics[trim={400 80 400 100},clip,width=0.08\linewidth]{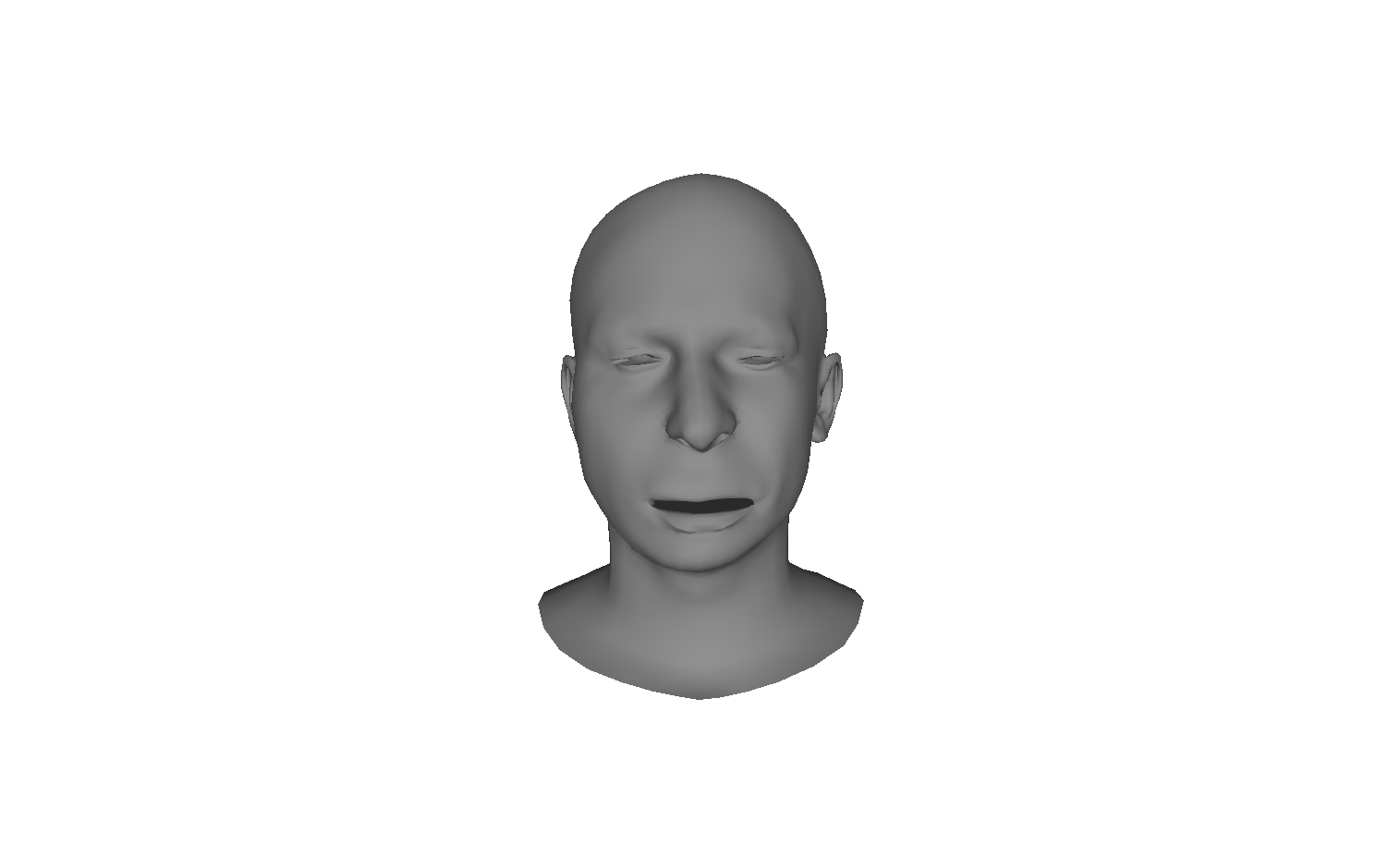}};
    \coordinate[right of=a7, node distance=1.3cm] (a8);
    \node[above of=a8, node distance=0.43cm] (a81) {\includegraphics[trim={400 240 380 300},clip,width=0.09\linewidth]{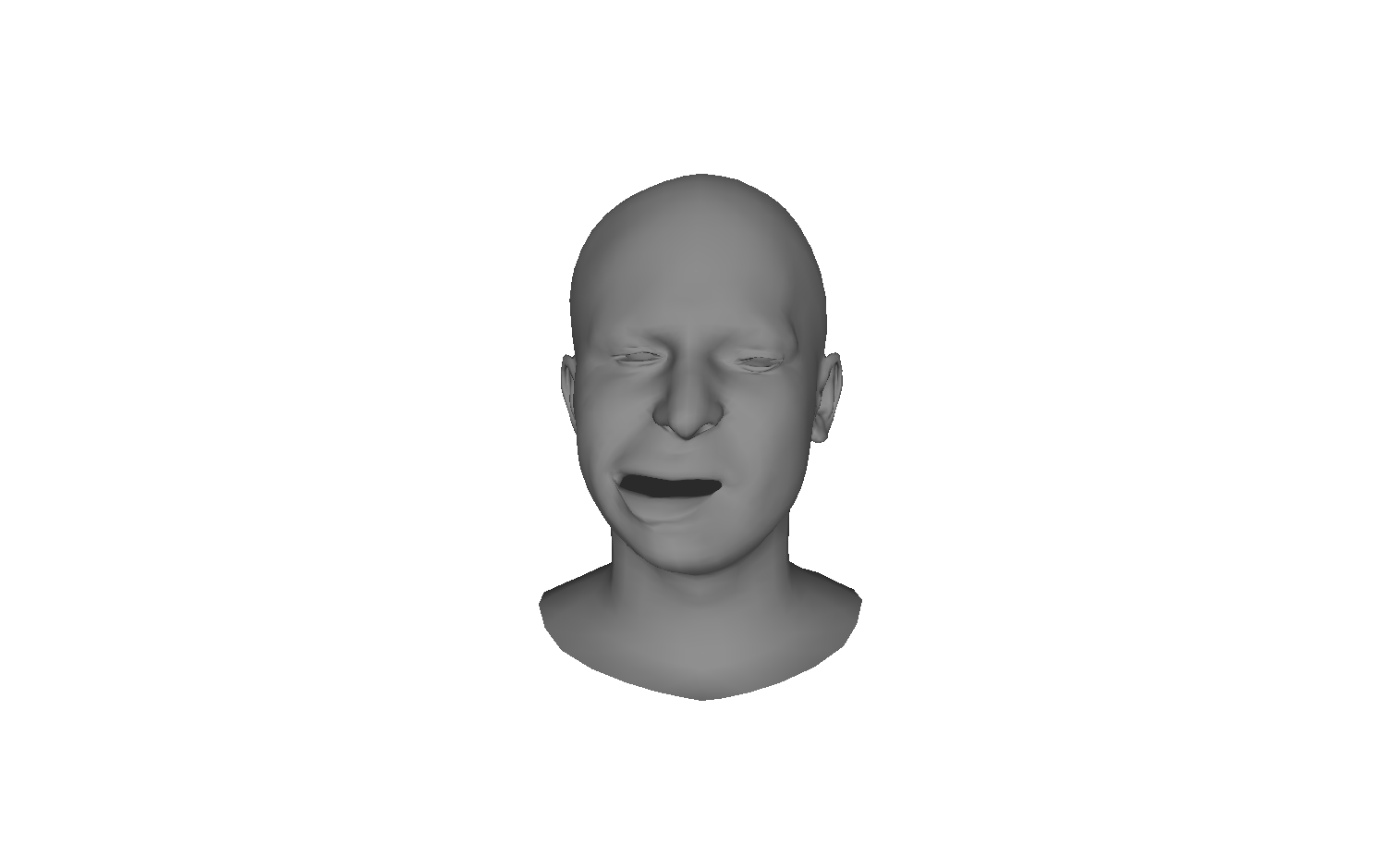}};
    \node[below of=a8, node distance=0.43cm] (a82) {\includegraphics[trim={400 240 380 300},clip,width=0.09\linewidth]{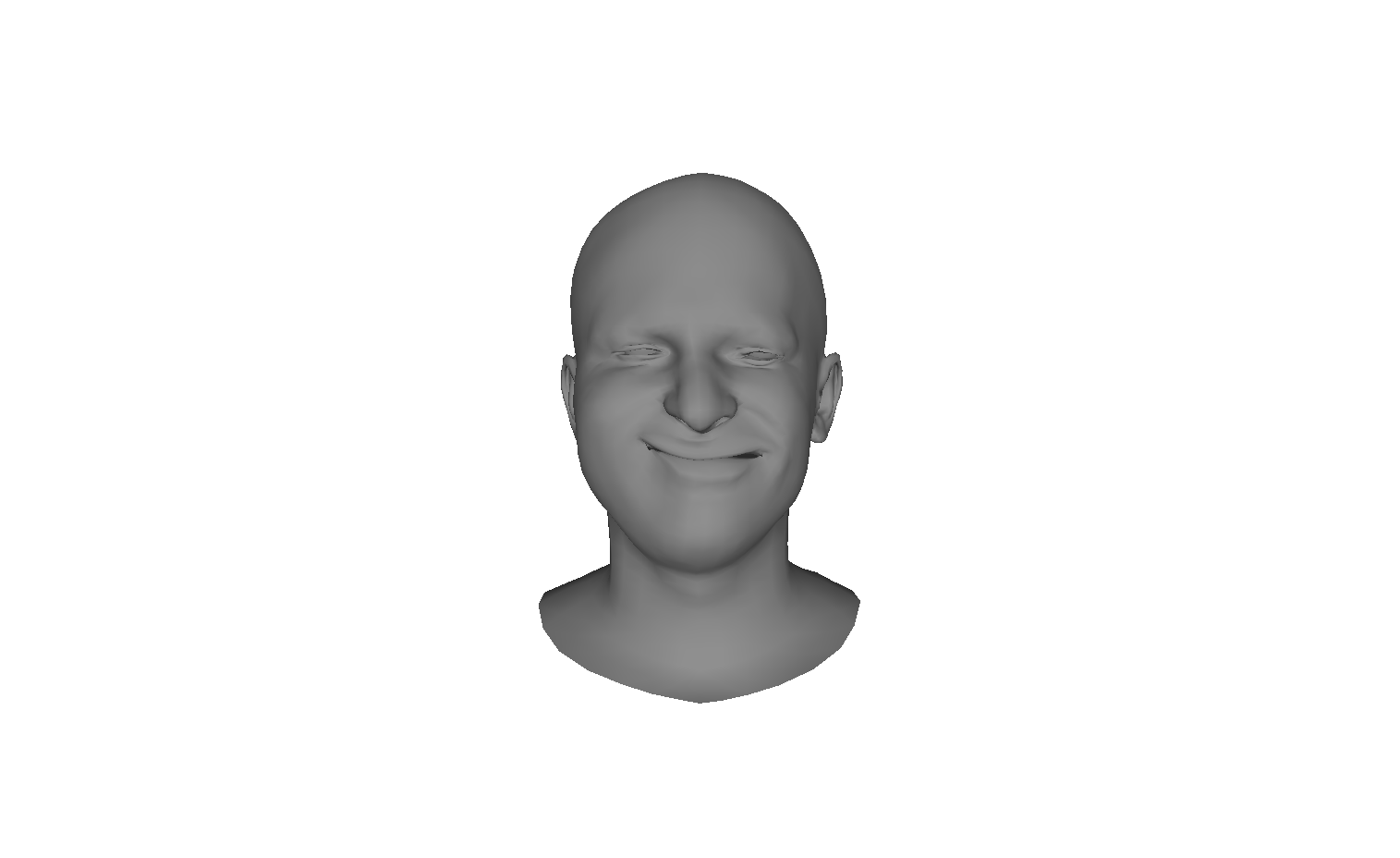}};
   \coordinate[right of=a8, node distance=1.3cm] (a9);
    \node[above of=a9, node distance=0.43cm] (a91) {\includegraphics[trim={400 240 380 300},clip,width=0.09\linewidth]{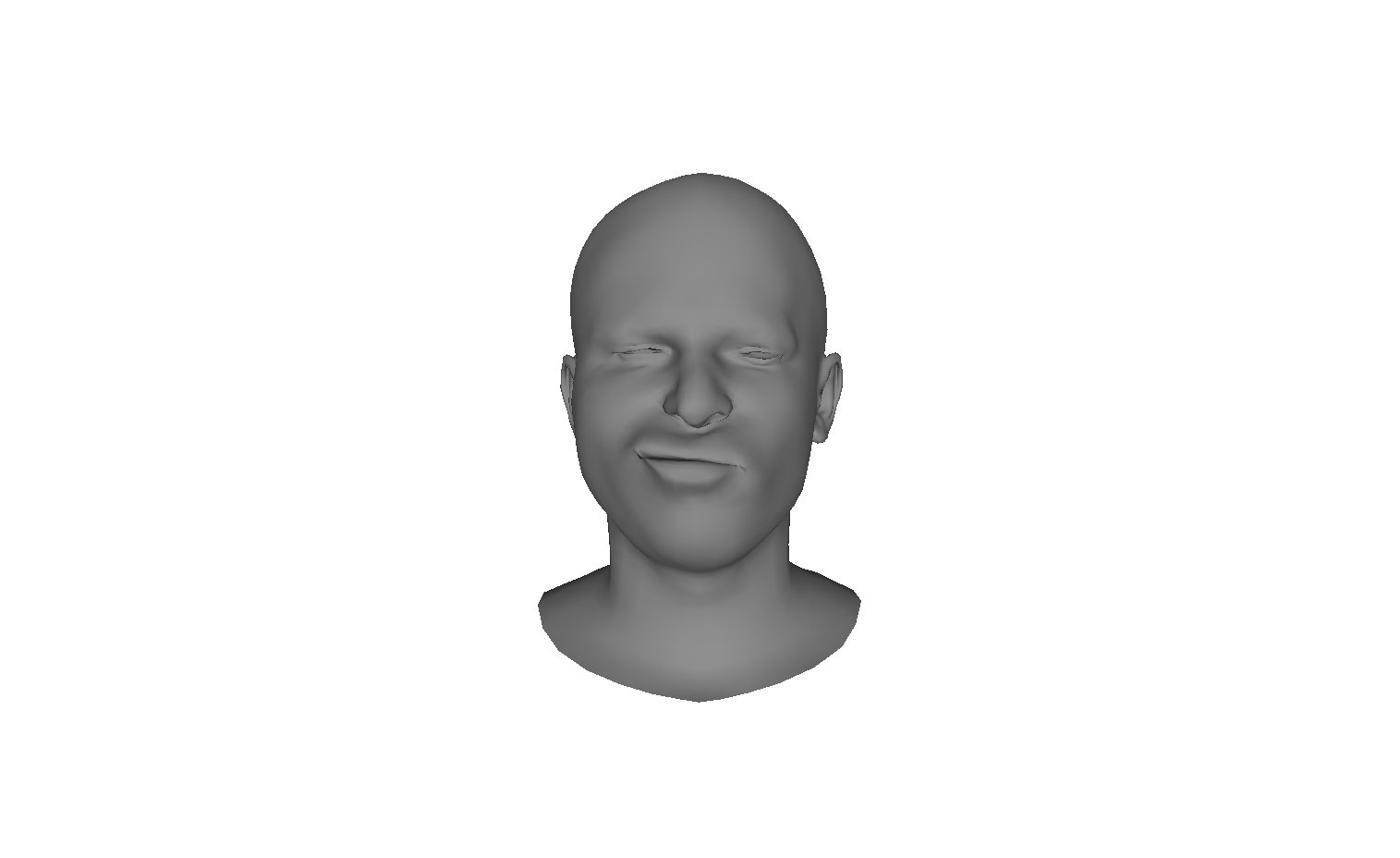}};
    \node[below of=a9, node distance=0.43cm] (a92) {\includegraphics[trim={400 240 380 300},clip,width=0.09\linewidth]{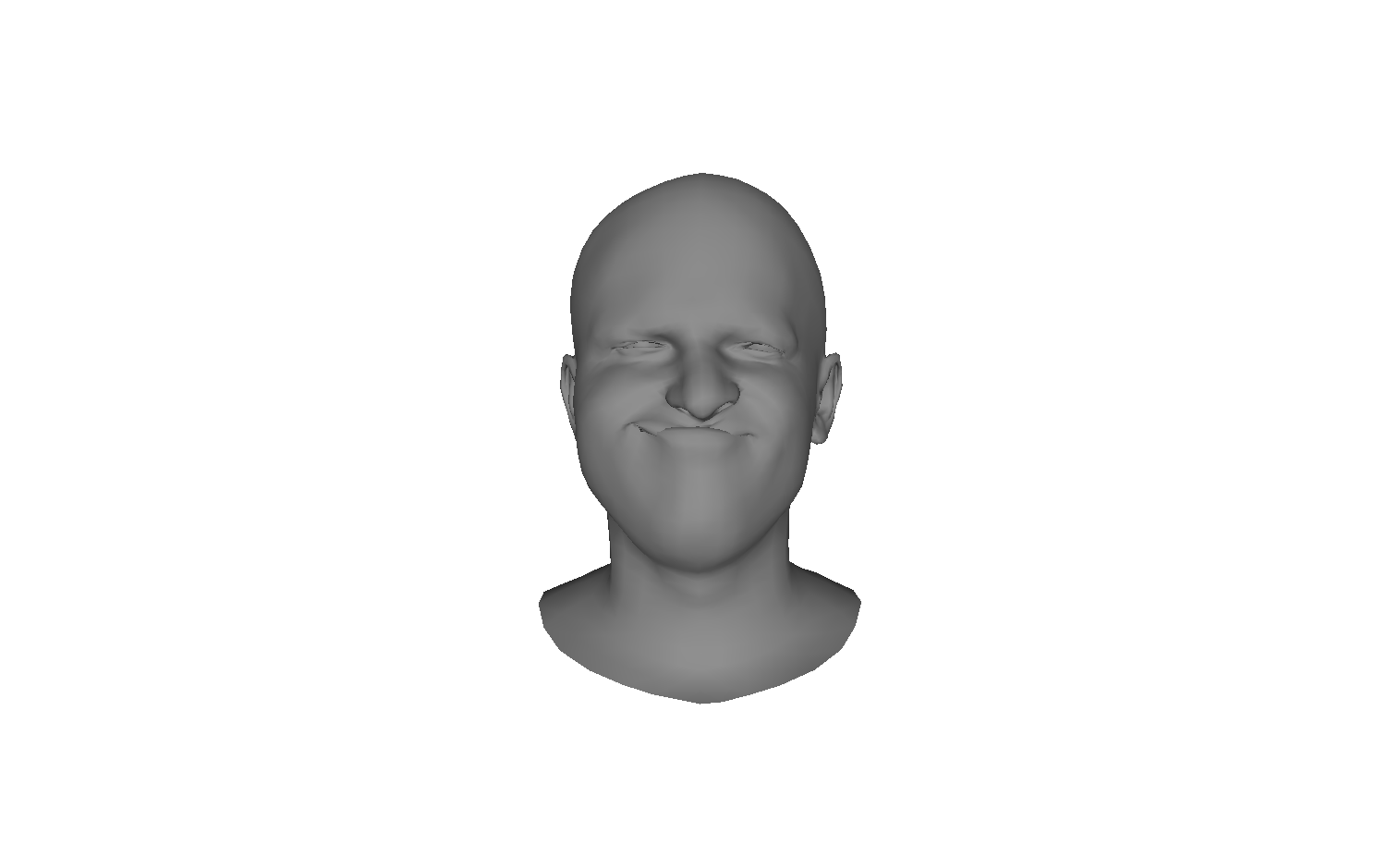}};
    \coordinate[right of=a9, node distance=1.3cm] (a10);
    \node[above of=a10, node distance=0.43cm] (a101) {\includegraphics[trim={400 240 380 300},clip,width=0.09\linewidth]{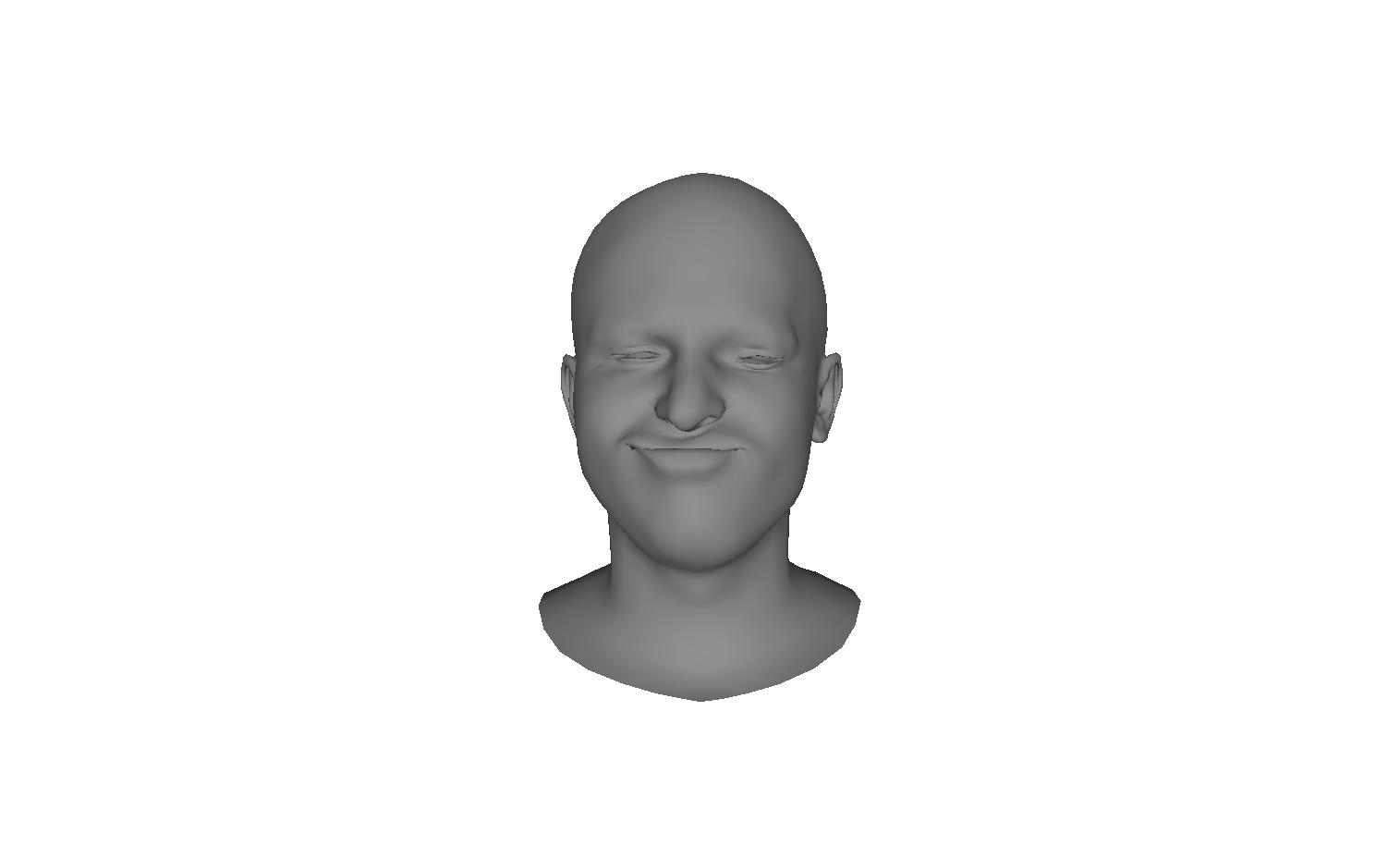}};
    \node[below of=a10, node distance=0.43cm] (a102) {\includegraphics[trim={400 240 380 300},clip,width=0.09\linewidth]{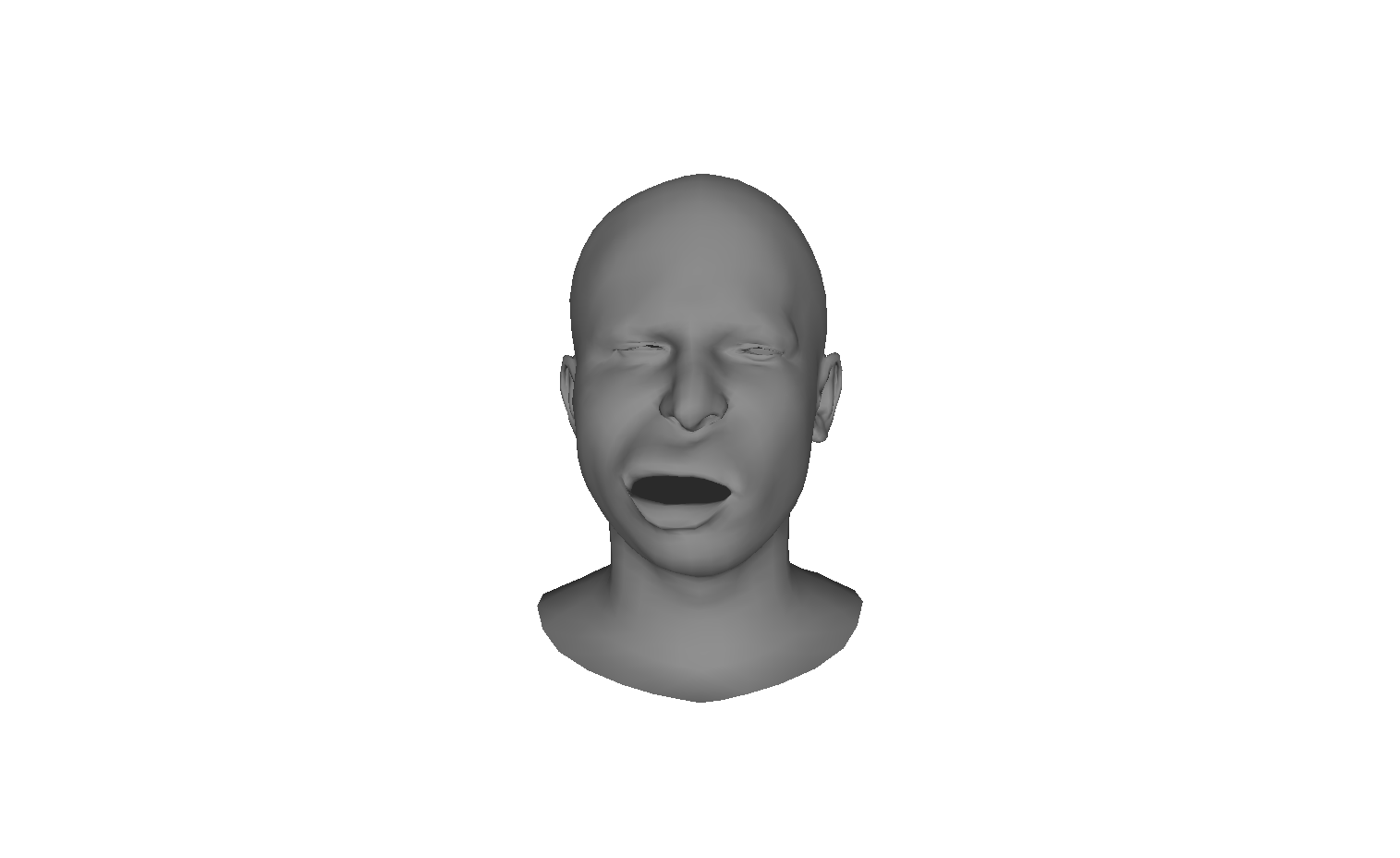}};
    \node[right of=a10, node distance=1.7cm] (a11) {\includegraphics[trim={400 80 400 100},clip,width=0.08\linewidth]{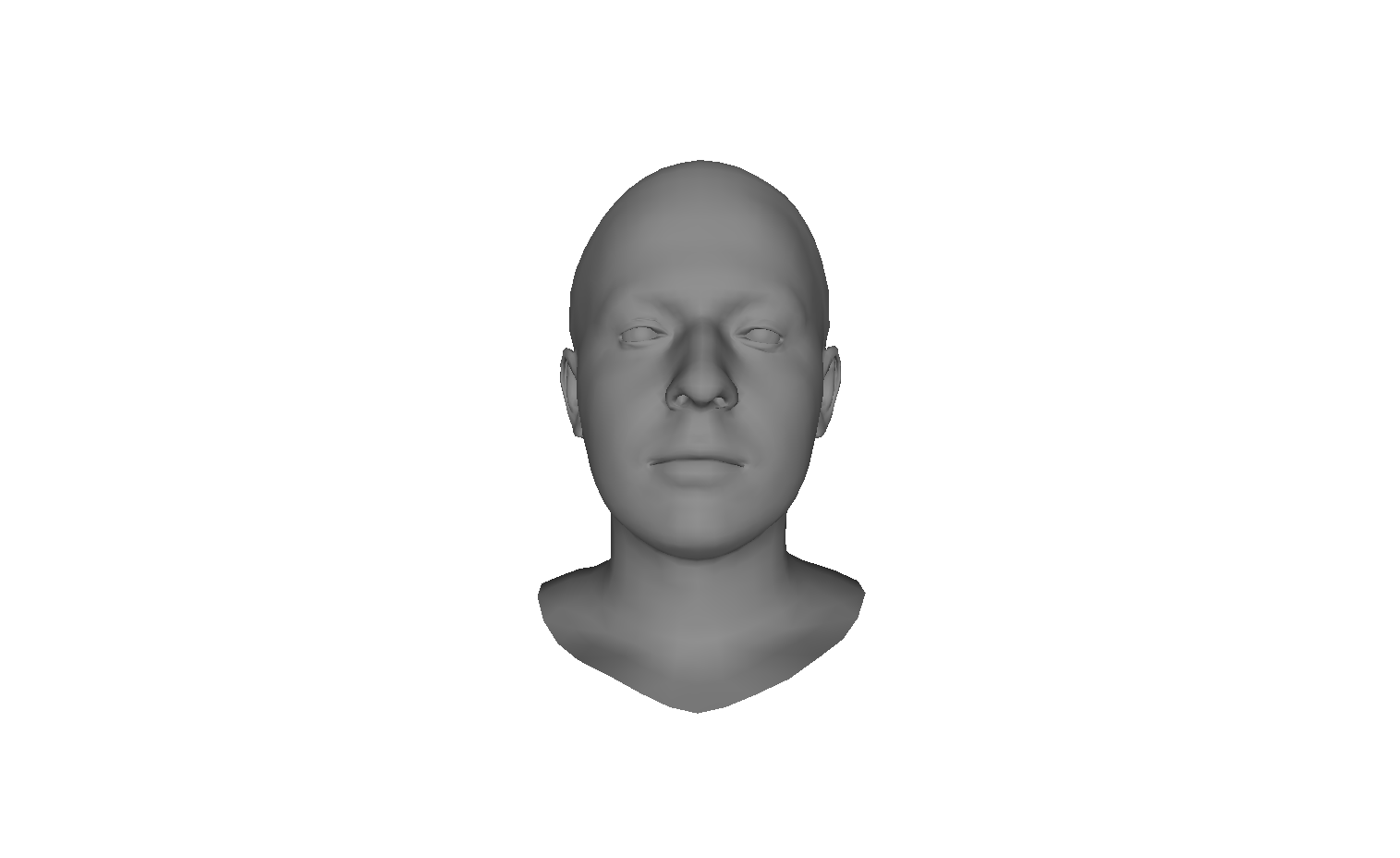}};
    
    \node[below of=a1, node distance=2.0cm] (b1) {\includegraphics[width=0.09\linewidth]{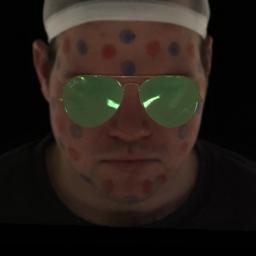}};
    \node[right of=b1, node distance=1.7cm] (b2) {\includegraphics[trim={400 80 400 100},clip,width=0.08\linewidth]{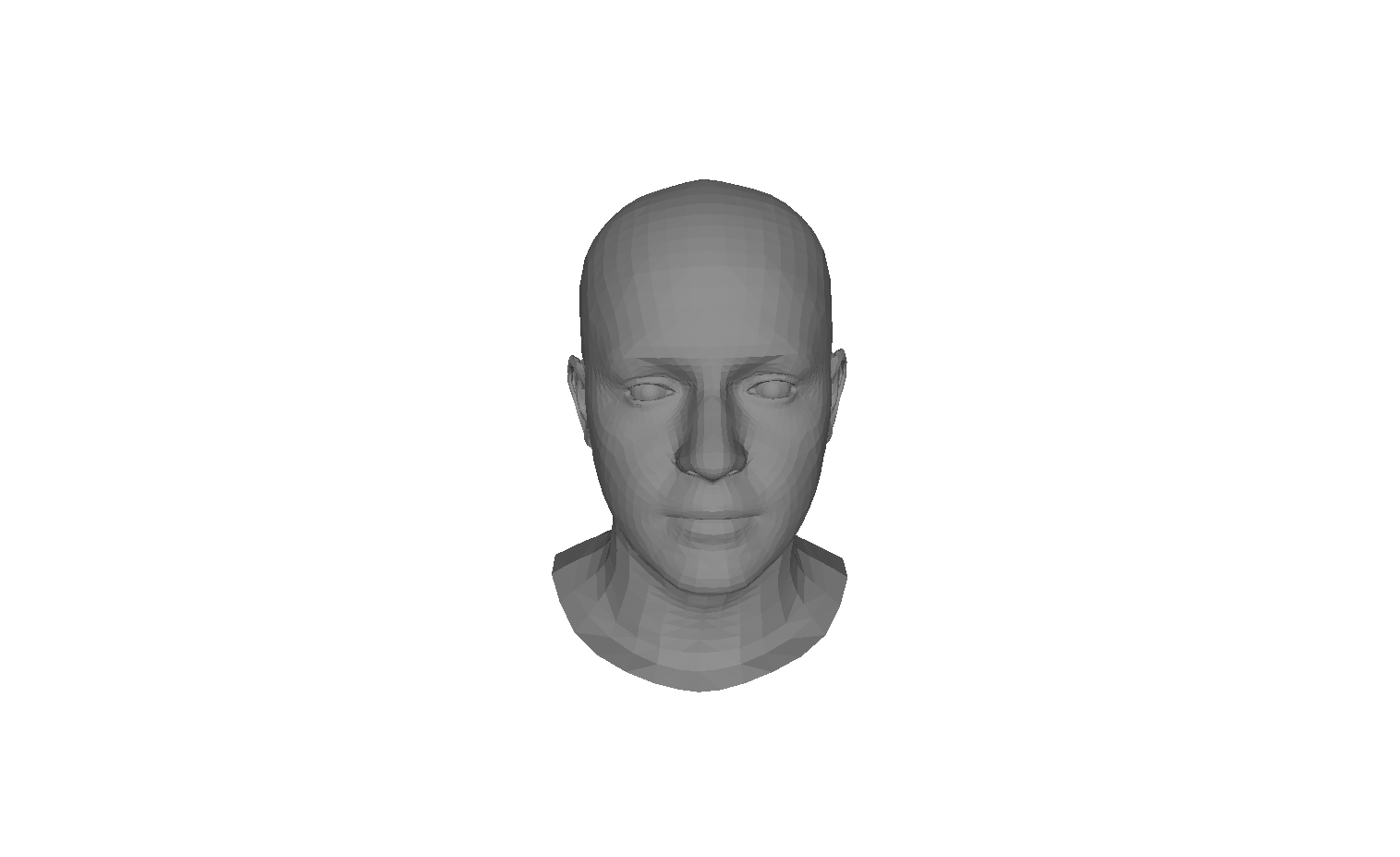}};
    \node[right of=b2, node distance=1.5cm] (b3) {\includegraphics[trim={400 80 400 100},clip,width=0.08\linewidth]{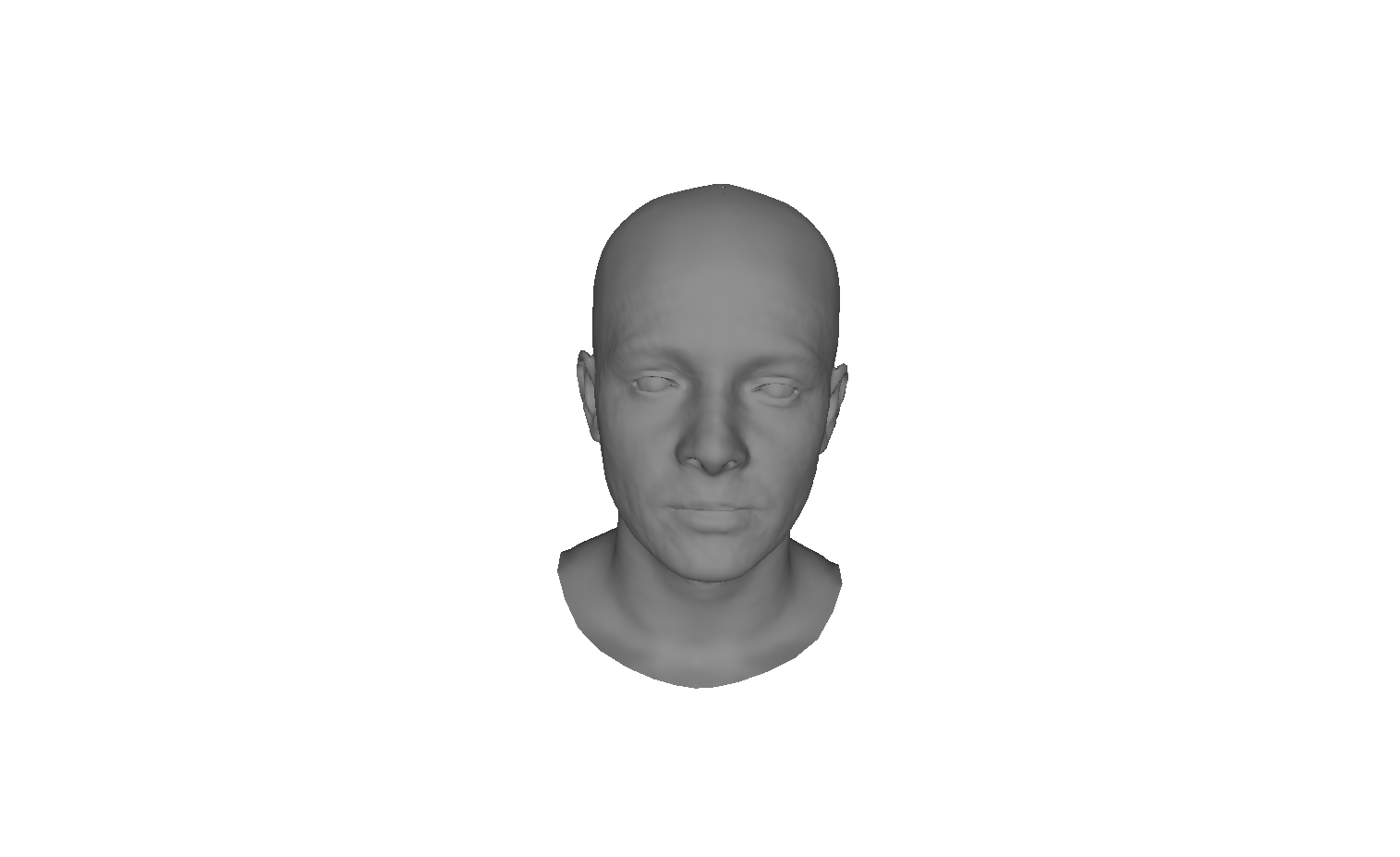}};
    \node[right of=b3, node distance=1.6cm] (b4) {\includegraphics[trim={400 80 400 100},clip,width=0.075\linewidth]{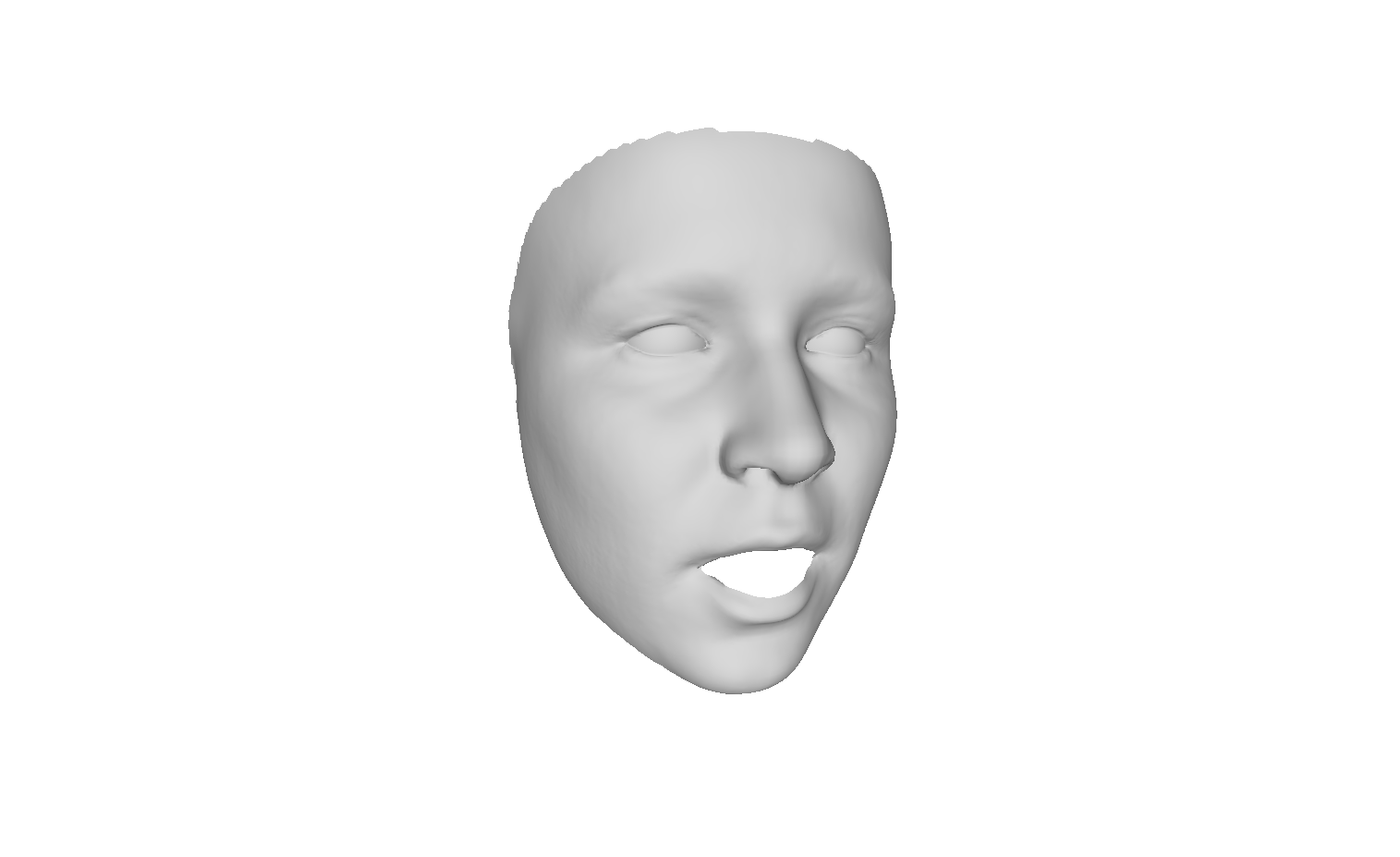}};
    \node[right of=b4, node distance=1.8cm] (b5) {\includegraphics[trim={400 80 400 100},clip,width=0.07\linewidth]{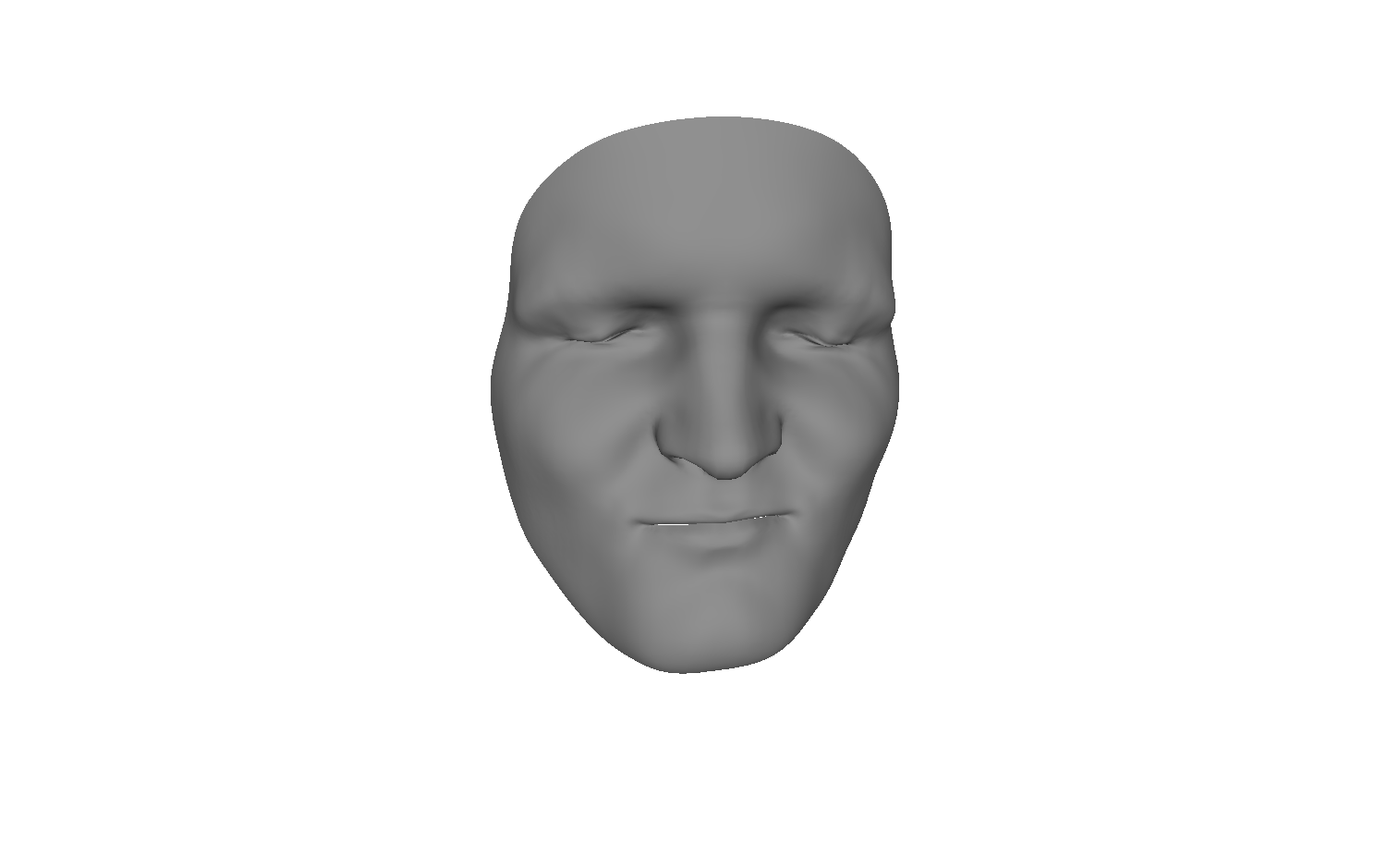}};
    \node[right of=b5, node distance=1.9cm] (b6) {\includegraphics[trim={400 80 400 100},clip,width=0.085\linewidth]{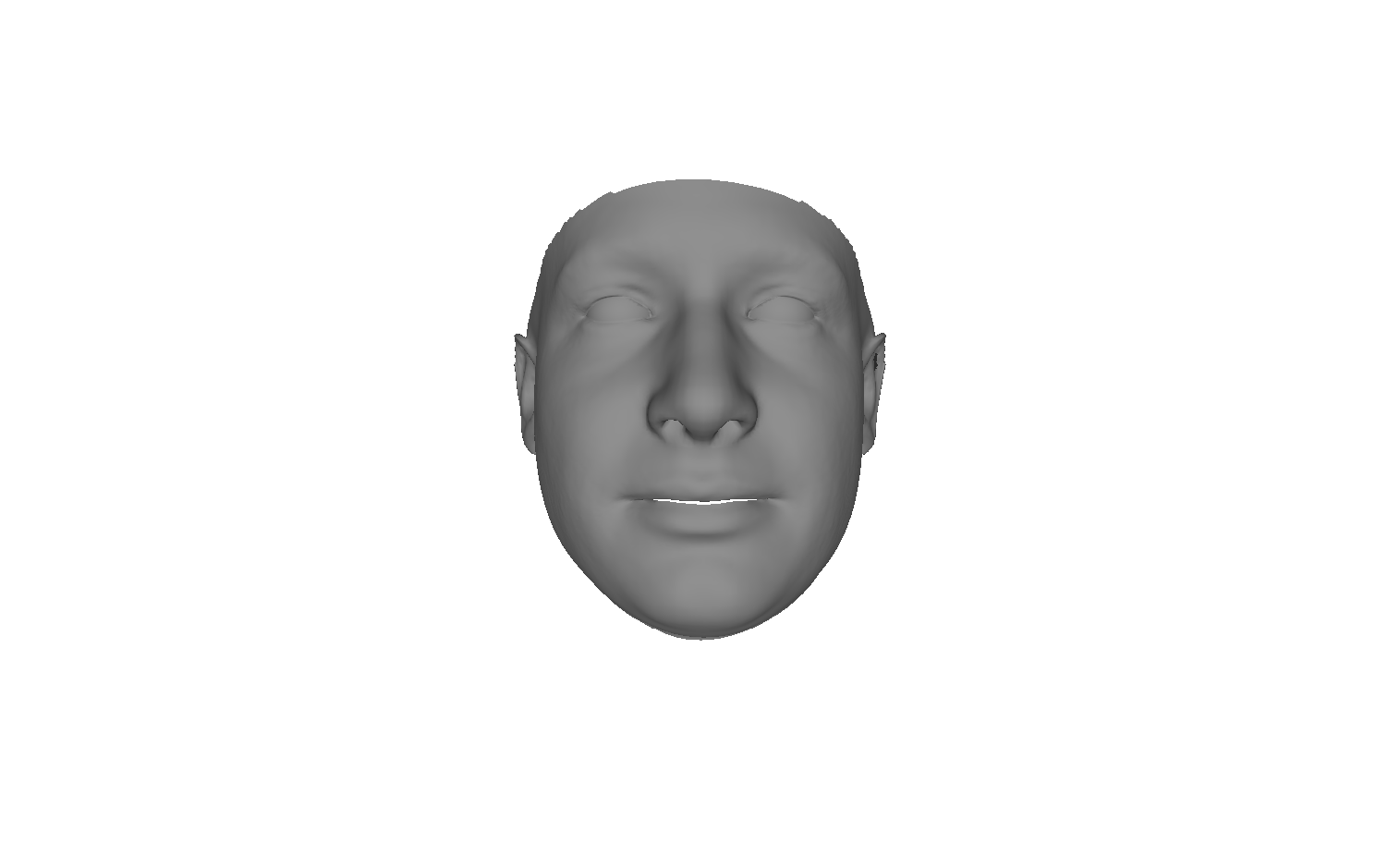}};
    
    \node[right of=b6, node distance=1.8cm] (b7) {\includegraphics[trim={400 80 400 100},clip,width=0.08\linewidth]{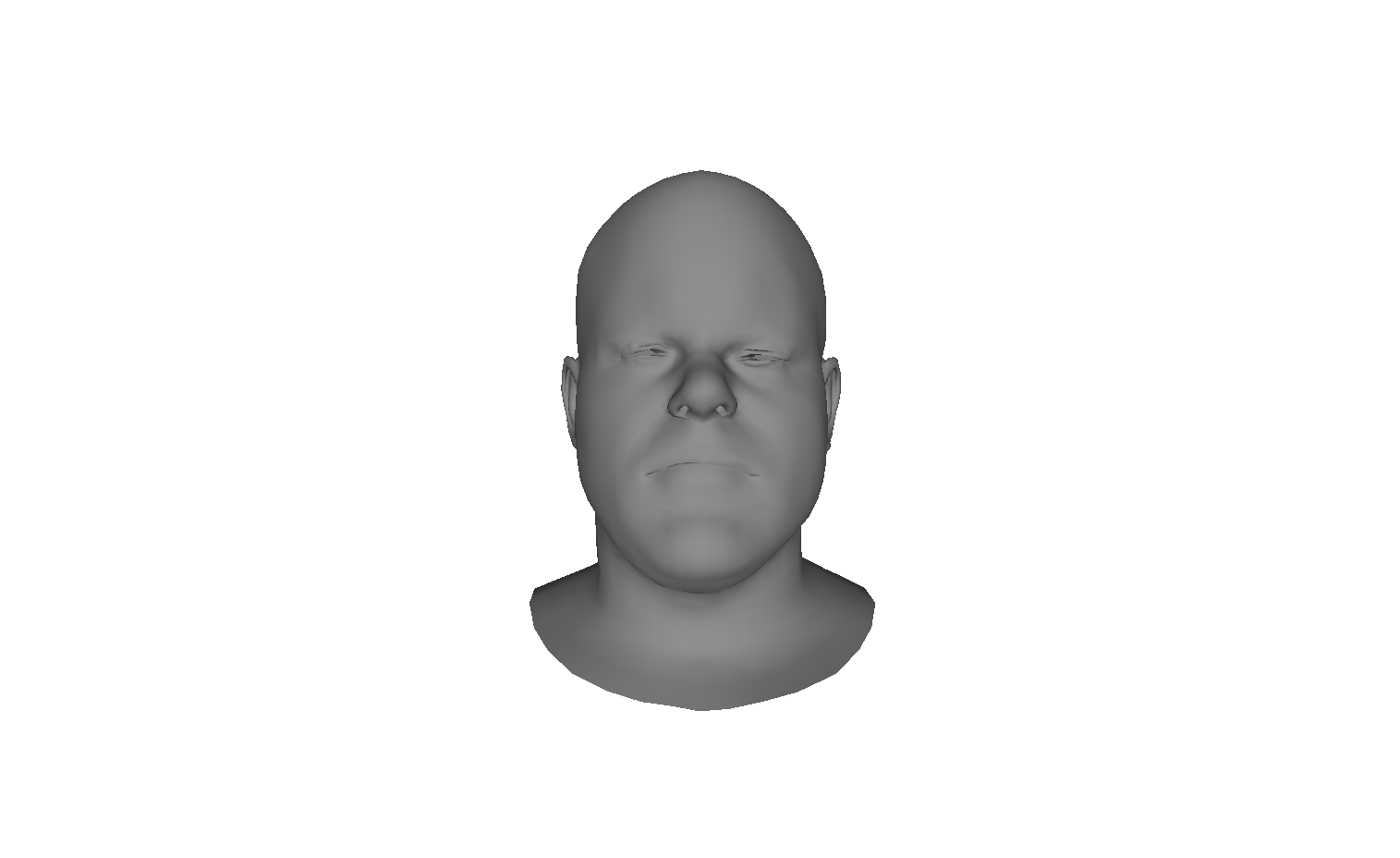}};
    \coordinate[right of=b7, node distance=1.3cm] (b8);
    \node[above of=b8, node distance=0.43cm] (b81) {\includegraphics[trim={400 340 380 200},clip,width=0.09\linewidth]{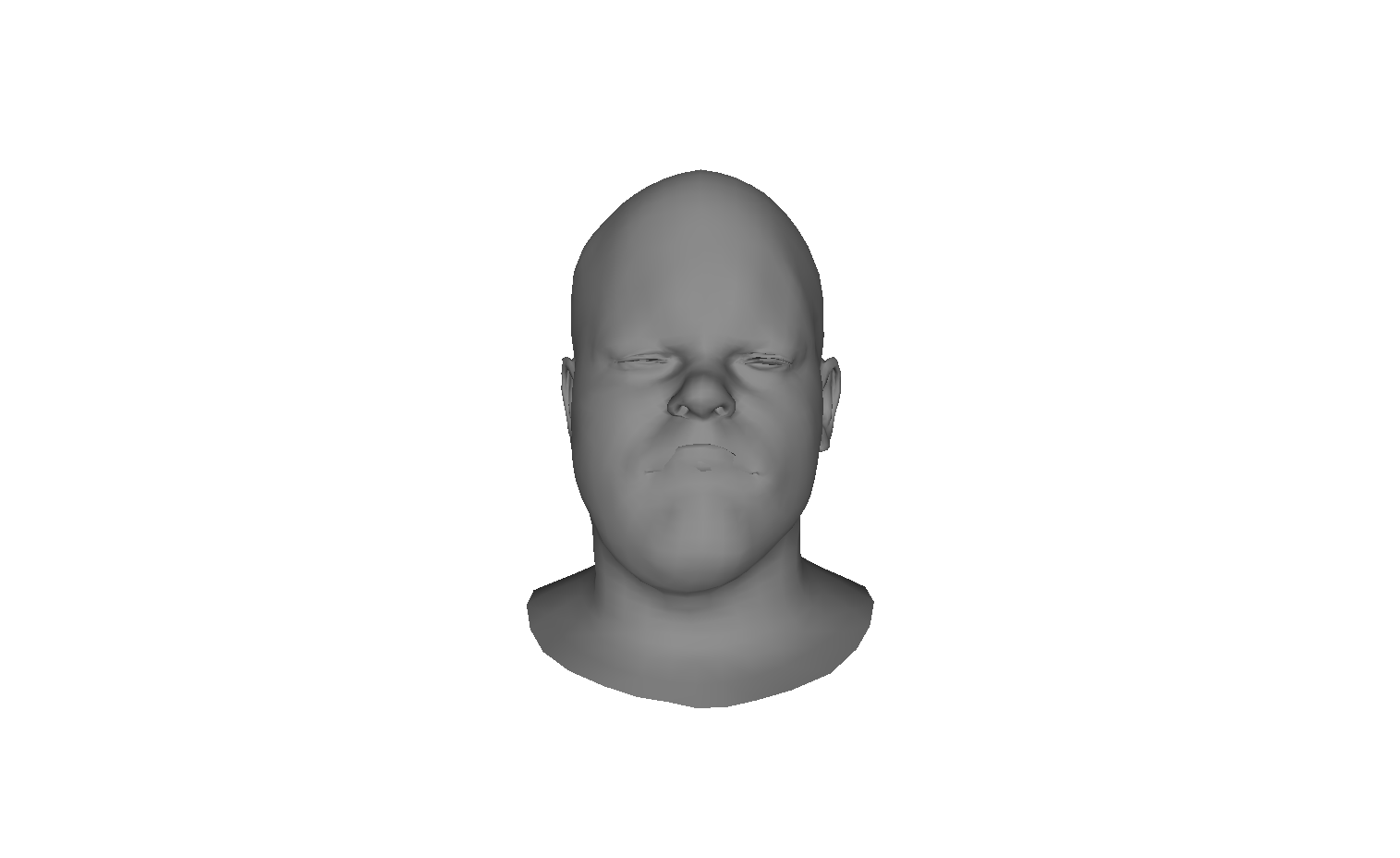}};
    \node[below of=b8, node distance=0.43cm] (b82) {\includegraphics[trim={400 340 380 200},clip,width=0.09\linewidth]{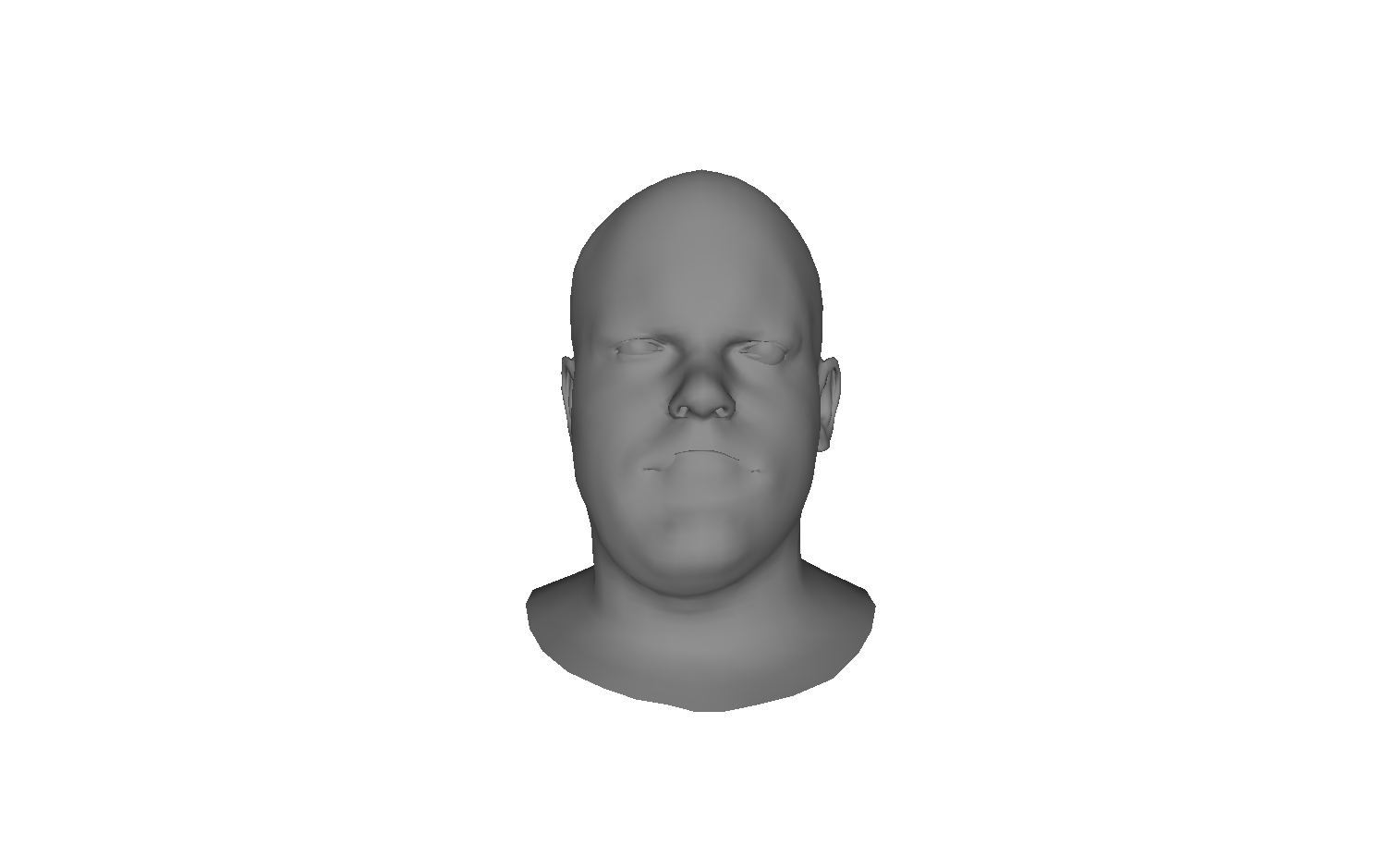}};
   \coordinate[right of=b8, node distance=1.3cm] (b9);
    \node[above of=b9, node distance=0.43cm] (b91) {\includegraphics[trim={400 340 380 200},clip,width=0.09\linewidth]{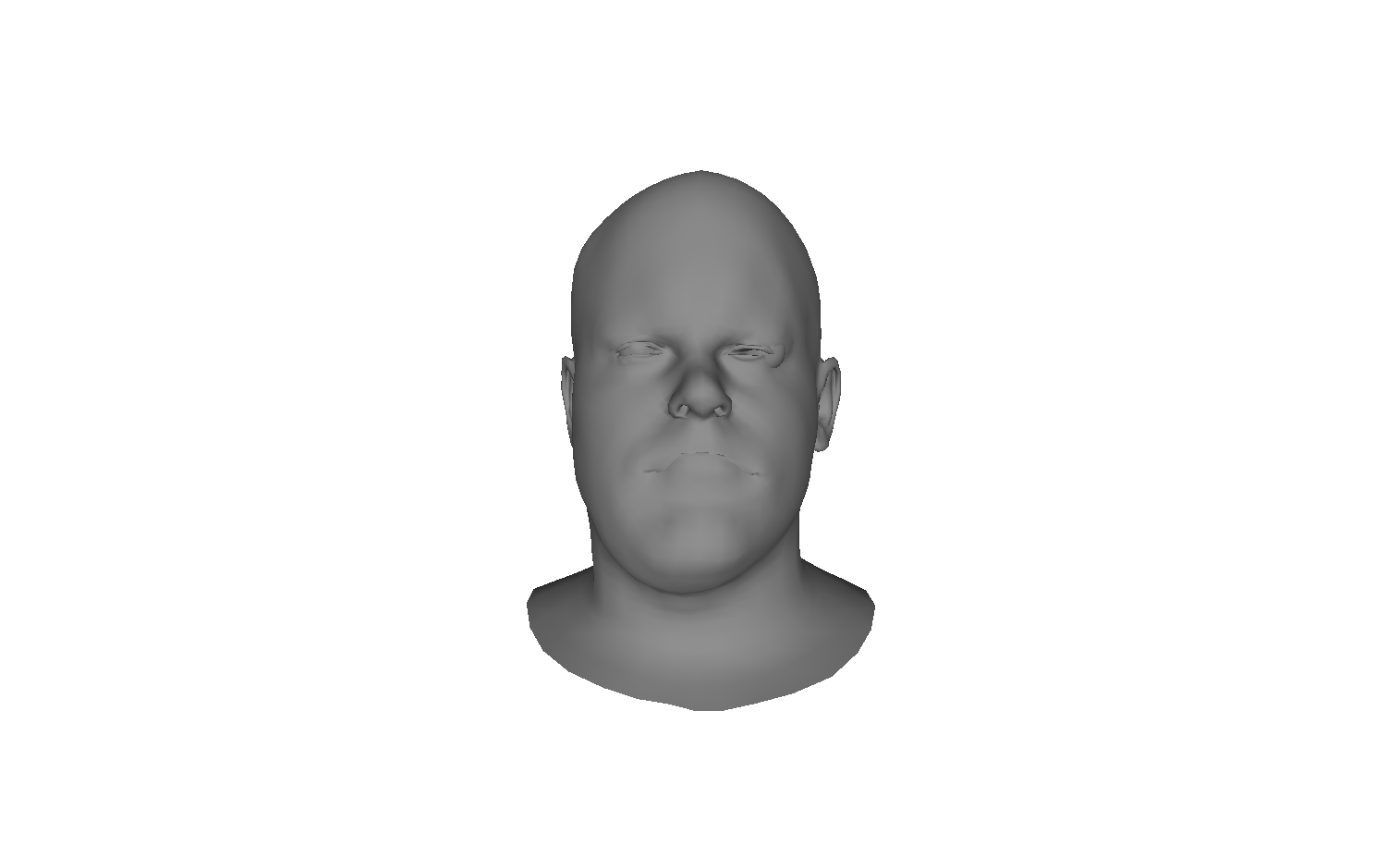}};
    \node[below of=b9, node distance=0.43cm] (b92) {\includegraphics[trim={400 340 380 200},clip,width=0.09\linewidth]{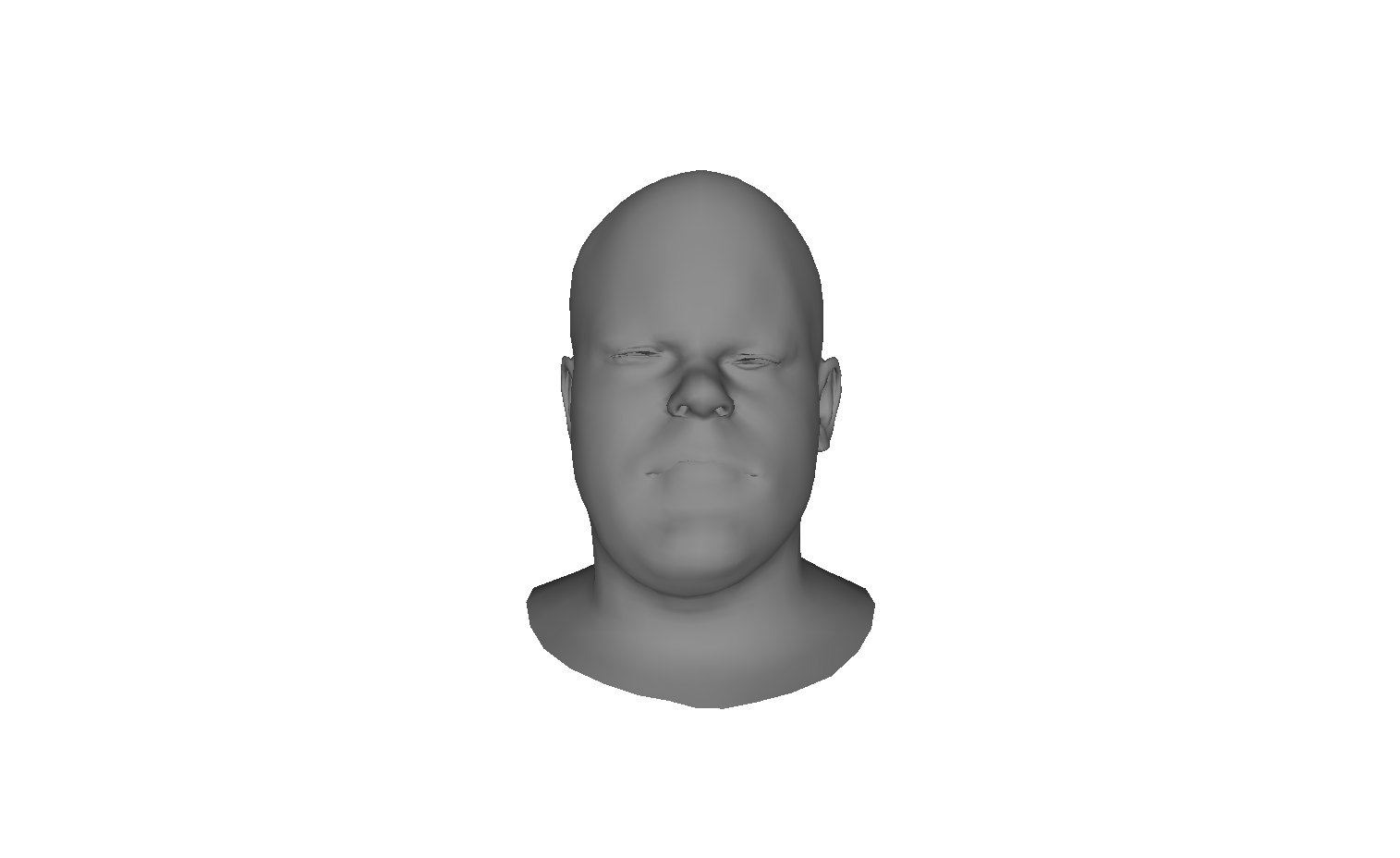}};
    \coordinate[right of=b9, node distance=1.3cm] (b10);
    \node[above of=b10, node distance=0.43cm] (b101) {\includegraphics[trim={400 340 380 200},clip,width=0.09\linewidth]{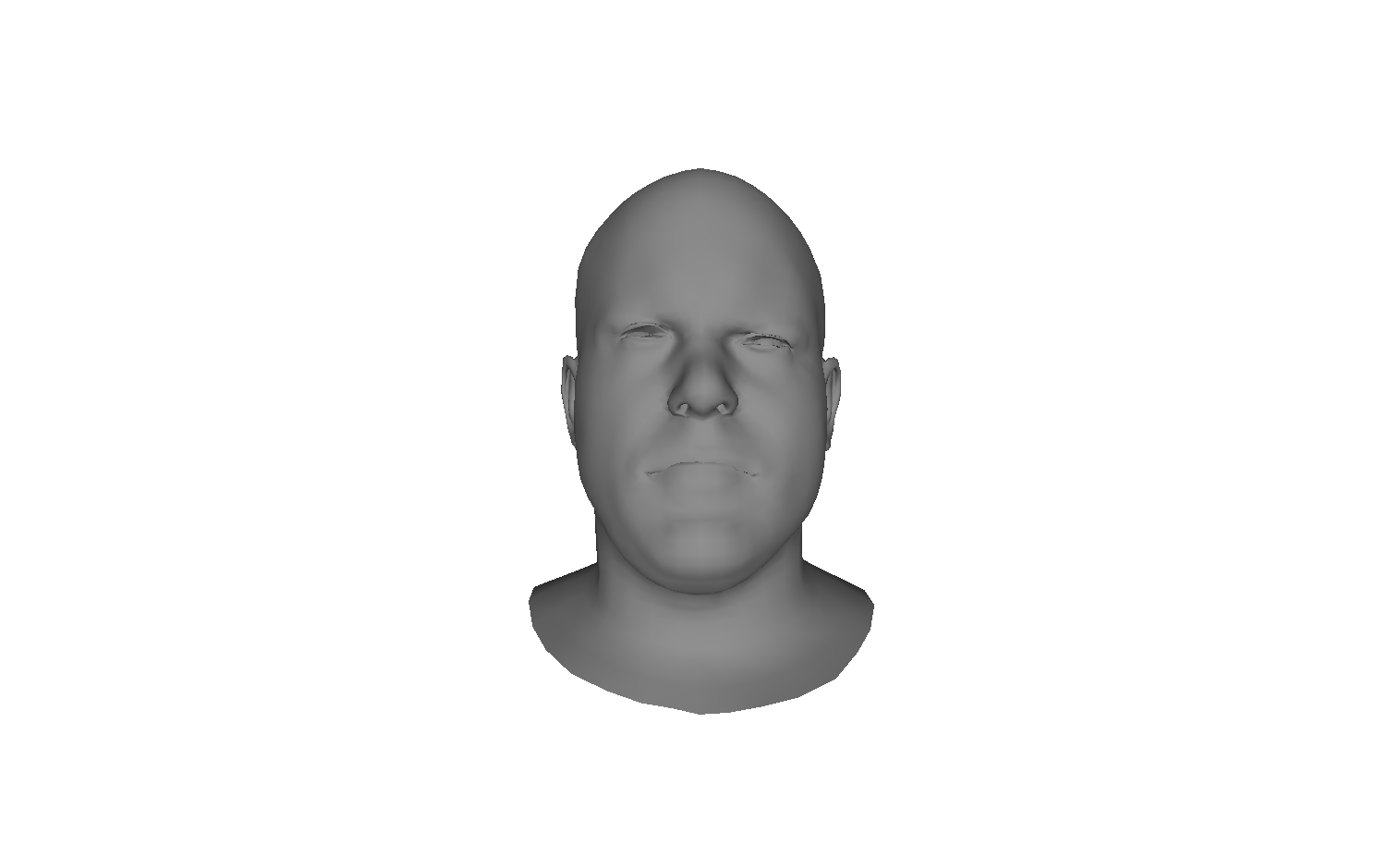}};
    \node[below of=b10, node distance=0.43cm] (b102) {\includegraphics[trim={400 340 380 200},clip,width=0.09\linewidth]{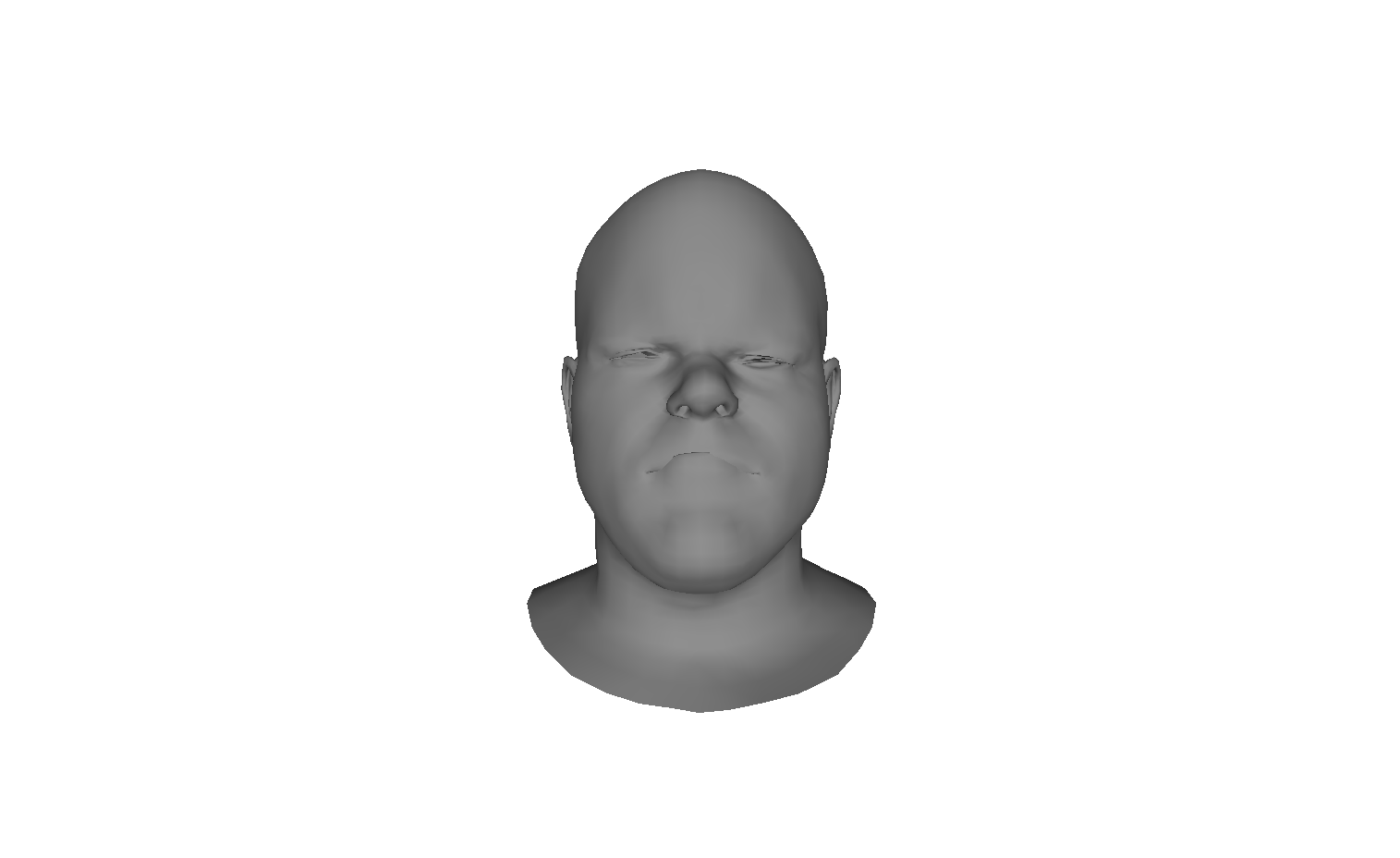}};
    \node[right of=b10, node distance=1.7cm] (b11) {\includegraphics[trim={400 80 400 100},clip,width=0.08\linewidth]{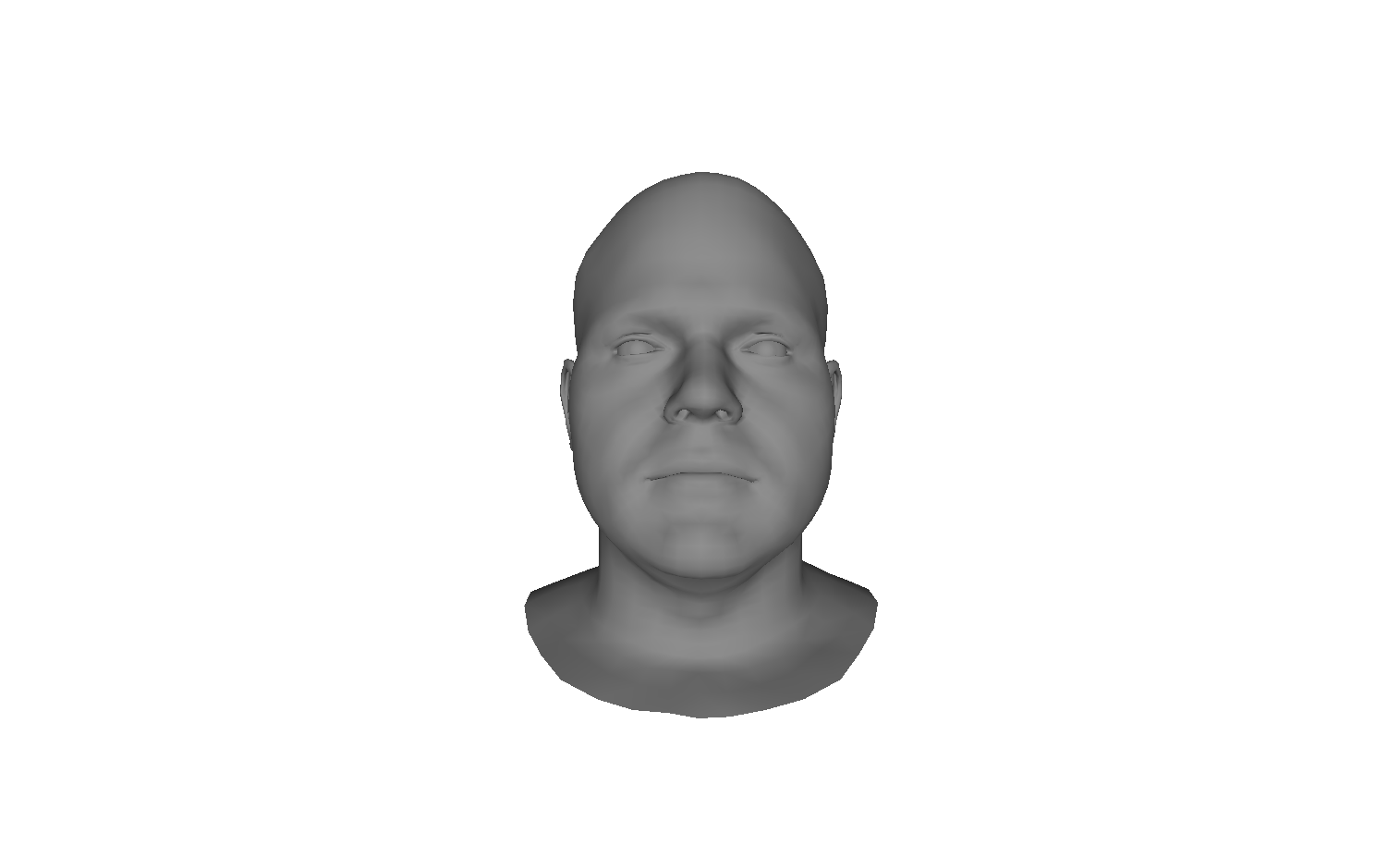}};
    
    \node[below of=b1, node distance=2.0cm] (c1) {\includegraphics[width=0.09\linewidth]{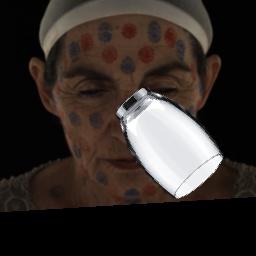}};
    \node[right of=c1, node distance=1.7cm] (c2) {\includegraphics[trim={400 80 400 100},clip,width=0.08\linewidth]{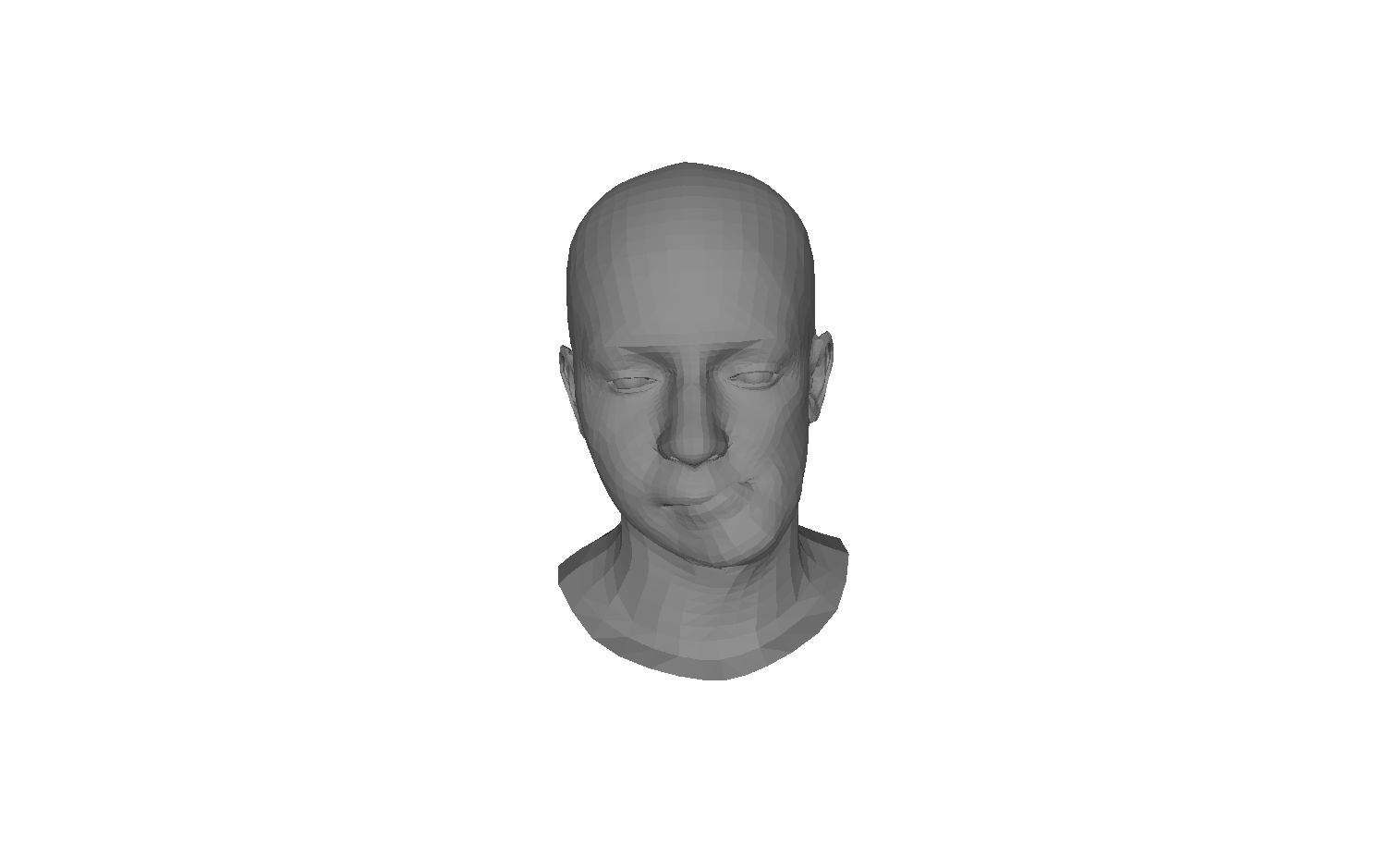}};
    \node[right of=c2, node distance=1.5cm] (c3) {\includegraphics[trim={400 80 400 100},clip,width=0.08\linewidth]{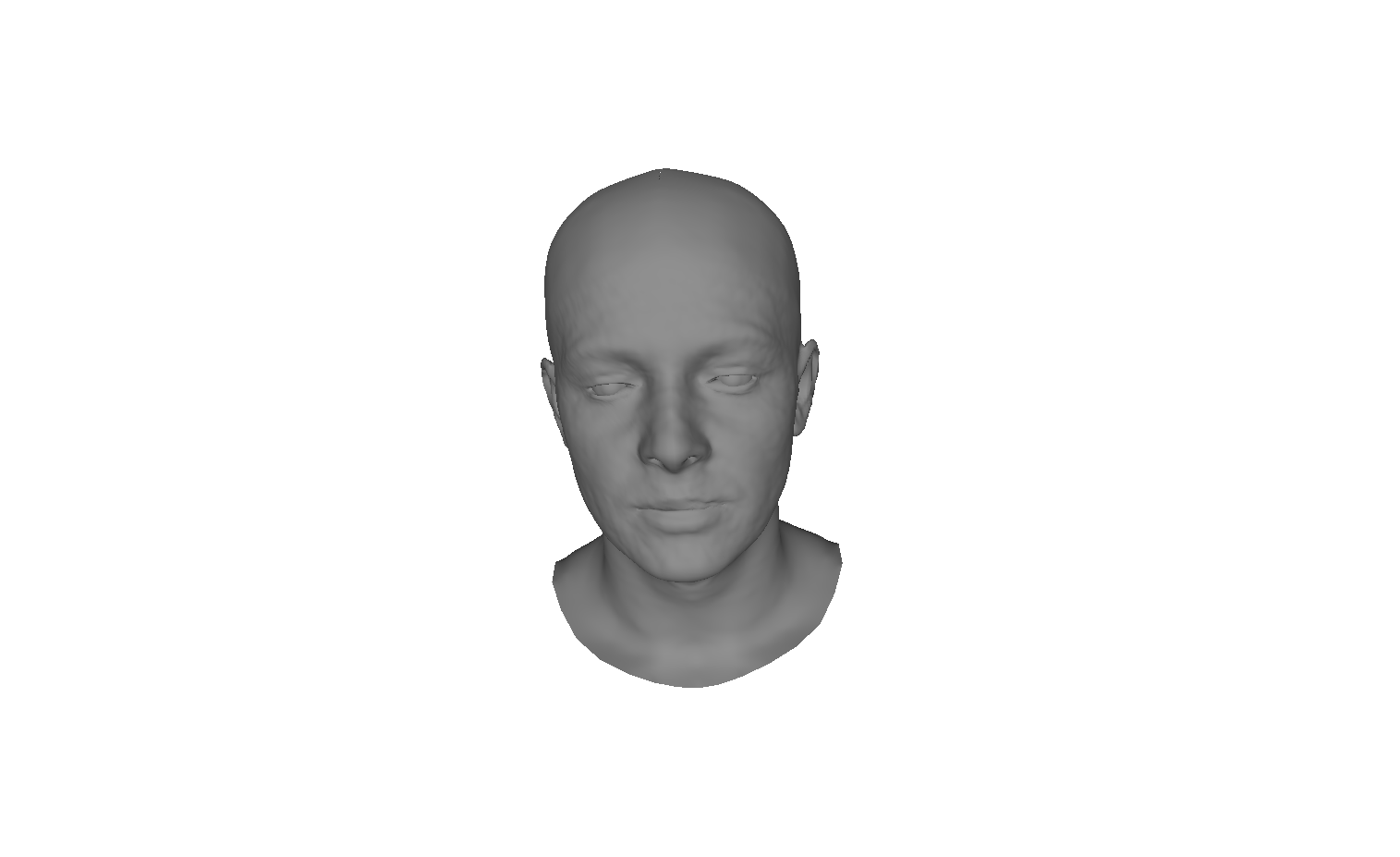}};
    \node[right of=c3, node distance=1.6cm] (c4) {\includegraphics[trim={400 80 400 100},clip,width=0.075\linewidth]{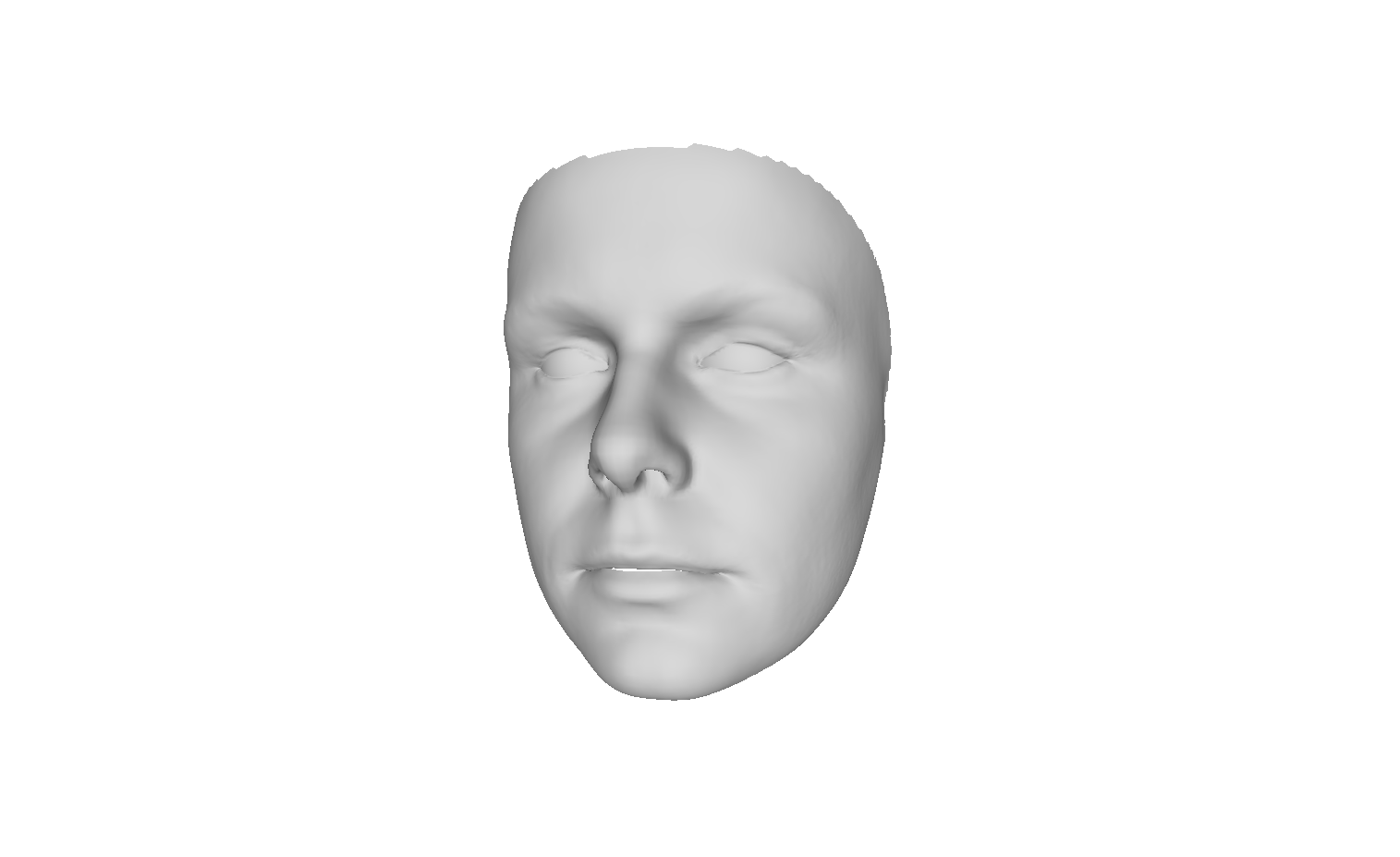}};
    \node[right of=c4, node distance=1.8cm] (c5) {\includegraphics[trim={400 80 400 100},clip,width=0.07\linewidth]{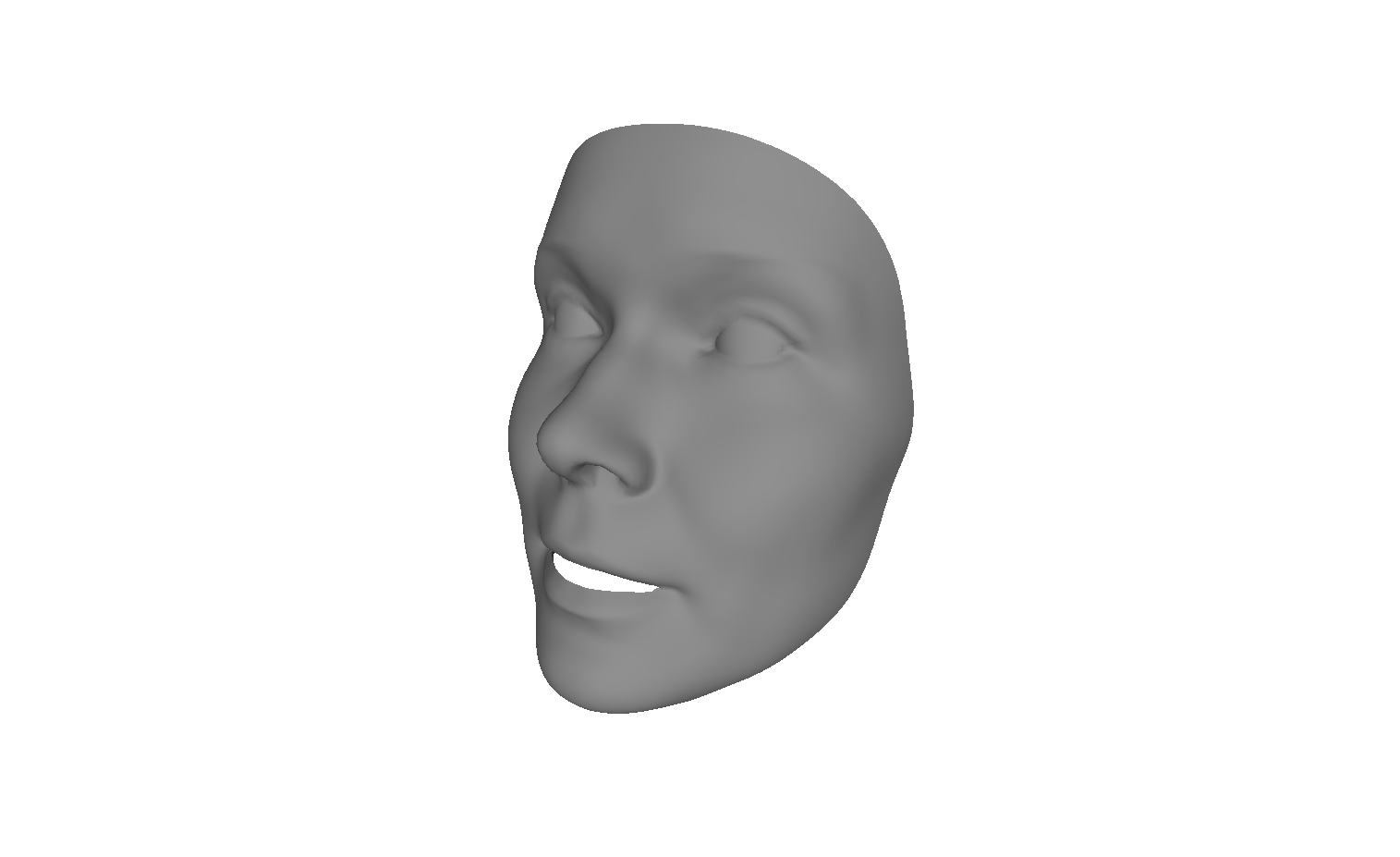}};
    \node[right of=c5, node distance=1.9cm] (c6) {\includegraphics[trim={400 80 400 100},clip,width=0.085\linewidth]{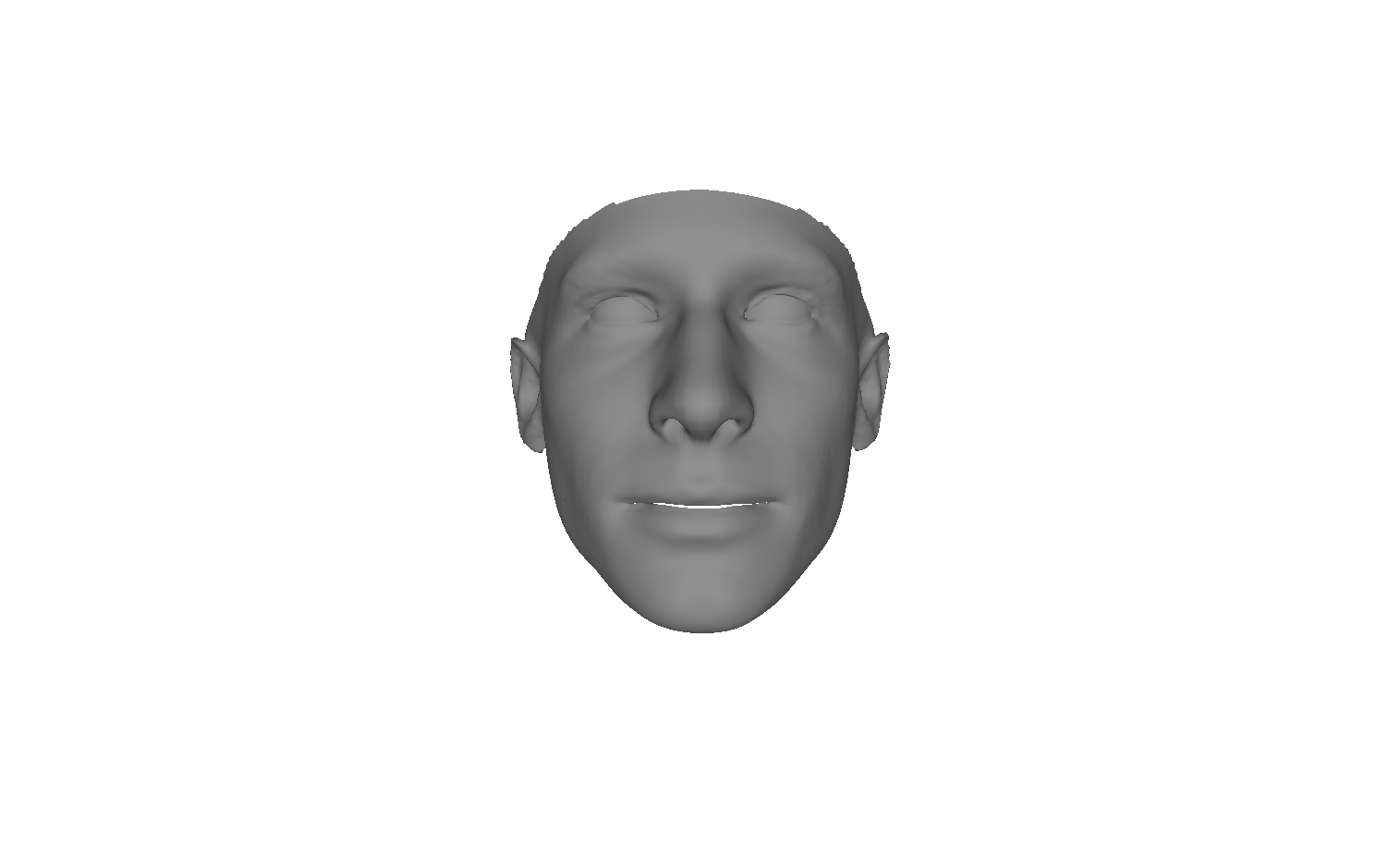}};
    
    \node[right of=c6, node distance=1.8cm] (c7) {\includegraphics[trim={400 80 400 100},clip,width=0.08\linewidth]{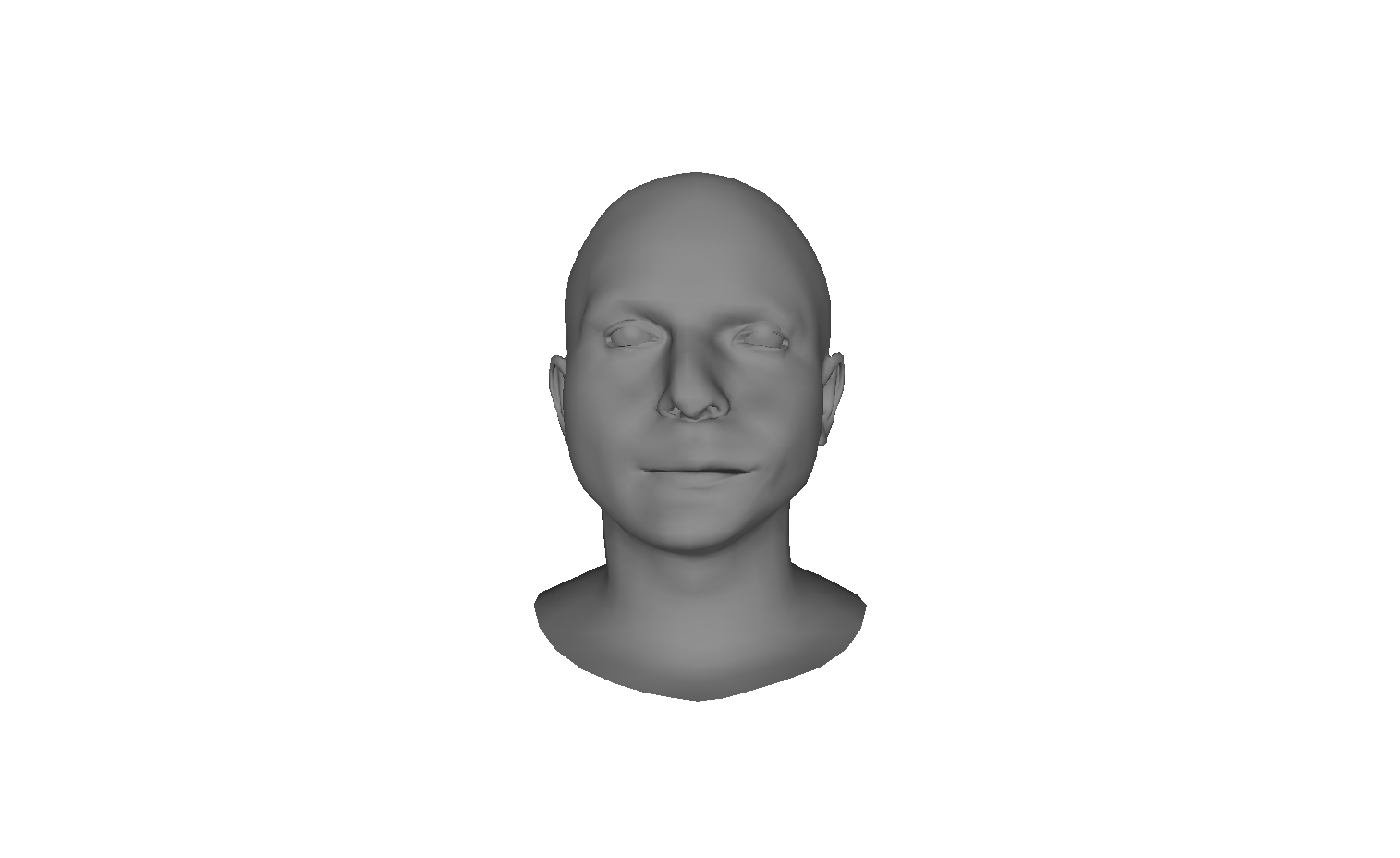}};
    \coordinate[right of=c7, node distance=1.3cm] (c8);
    \node[above of=c8, node distance=0.43cm] (c81) {\includegraphics[trim={400 240 380 300},clip,width=0.09\linewidth]{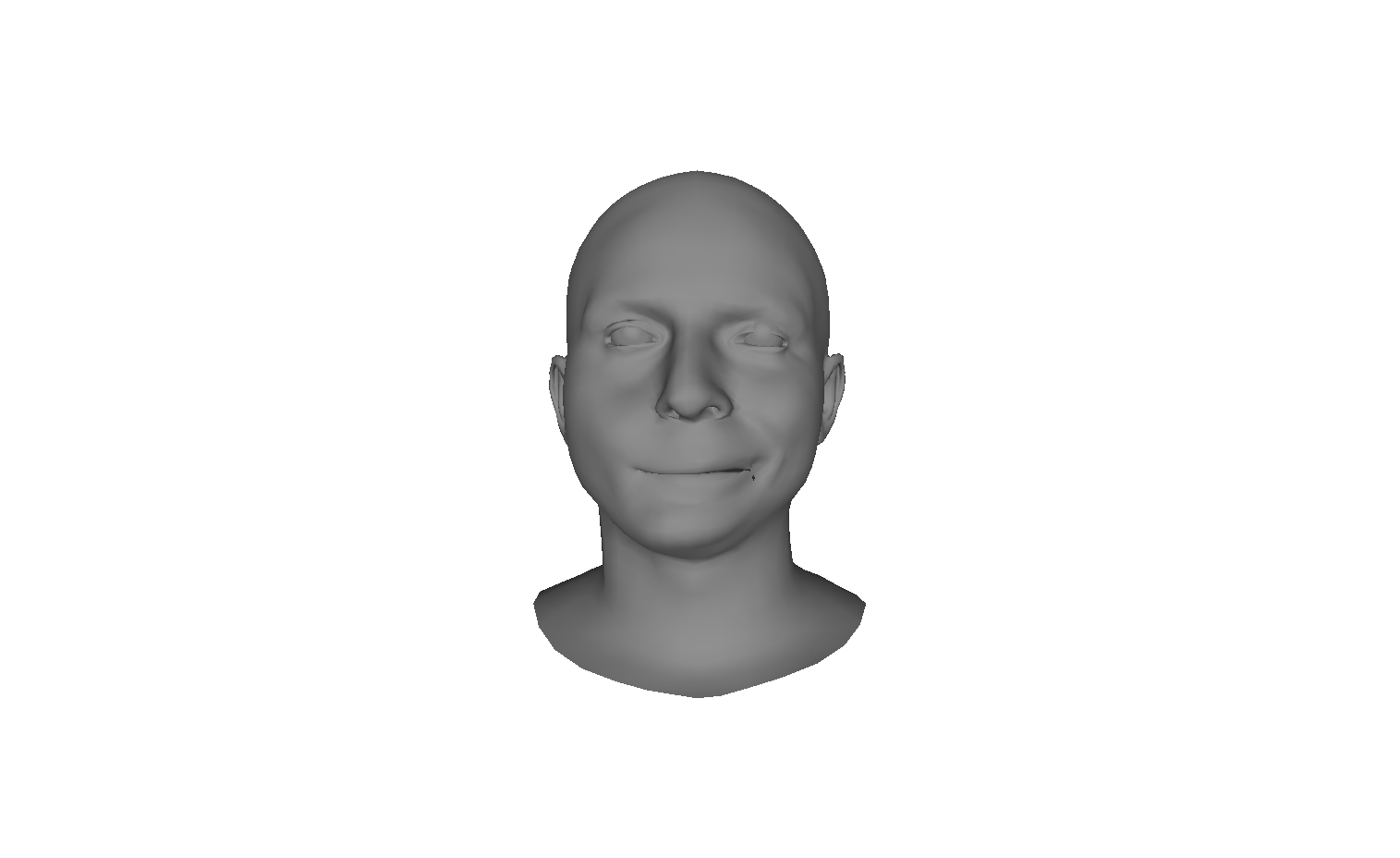}};
    \node[below of=c8, node distance=0.43cm] (c82) {\includegraphics[trim={400 240 380 300},clip,width=0.09\linewidth]{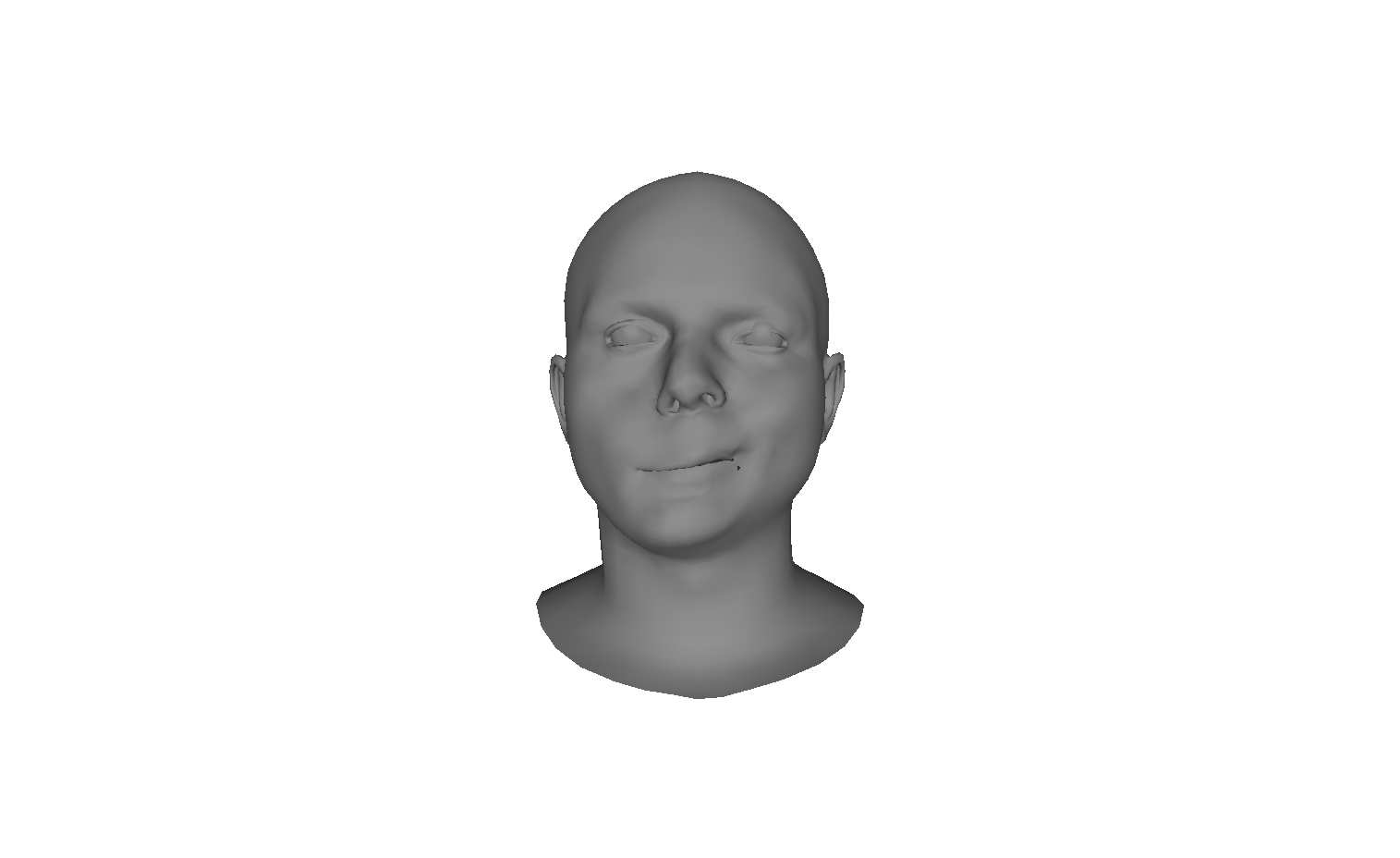}};
   \coordinate[right of=c8, node distance=1.3cm] (c9);
    \node[above of=c9, node distance=0.43cm] (c91) {\includegraphics[trim={400 240 380 300},clip,width=0.09\linewidth]{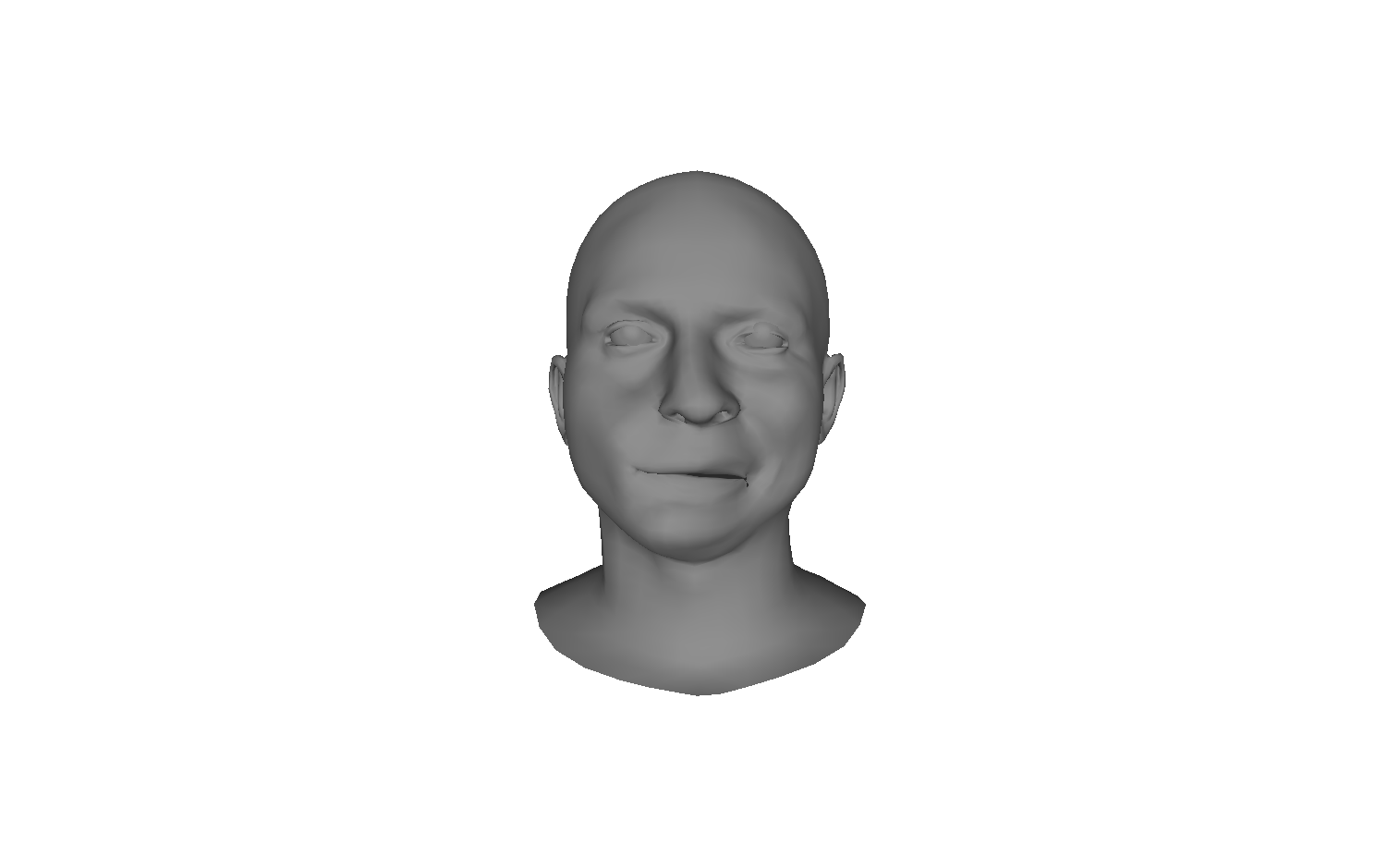}};
    \node[below of=c9, node distance=0.43cm] (c92) {\includegraphics[trim={400 240 380 300},clip,width=0.09\linewidth]{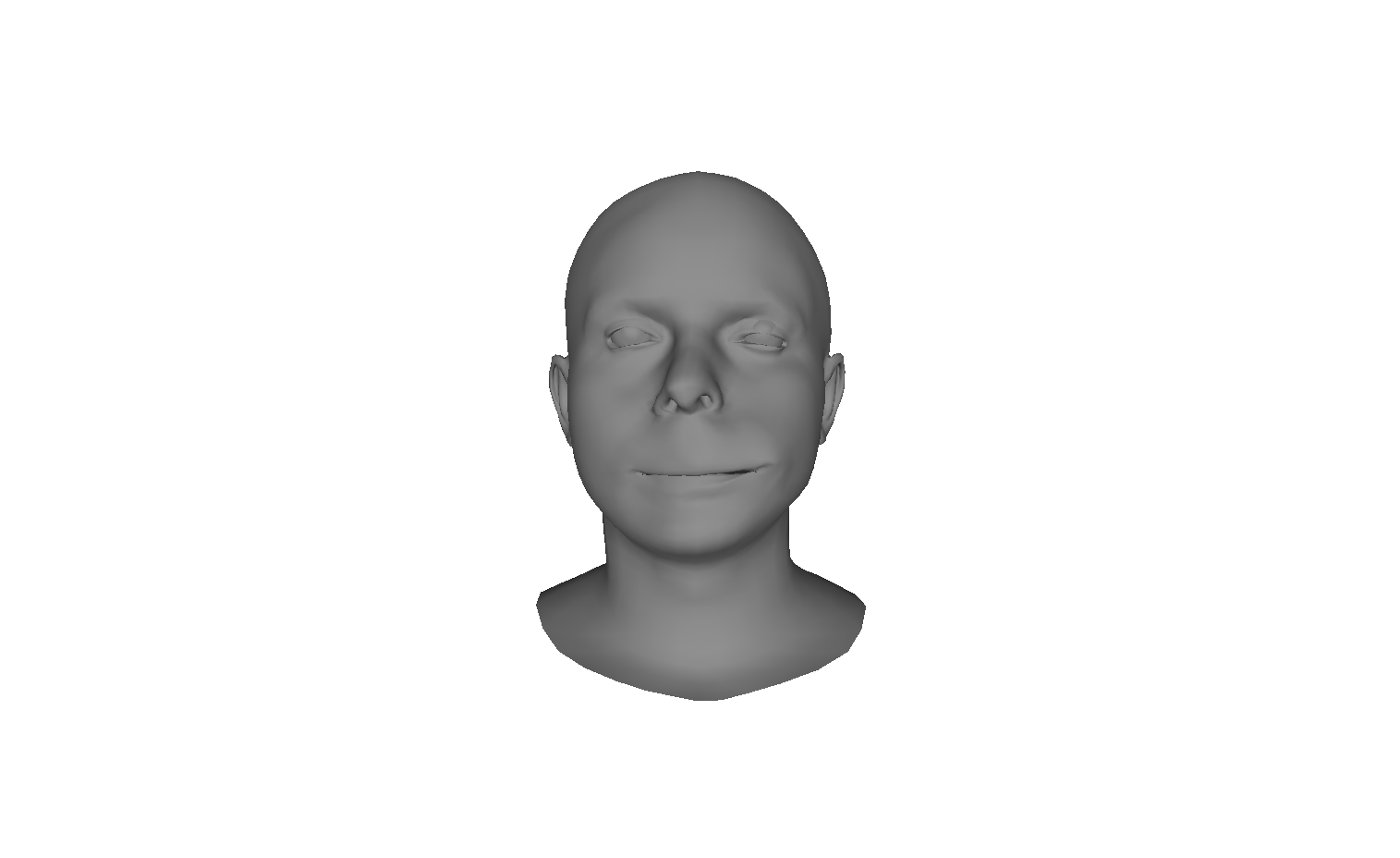}};
    \coordinate[right of=c9, node distance=1.3cm] (c10);
    \node[above of=c10, node distance=0.43cm] (c101) {\includegraphics[trim={400 240 380 300},clip,width=0.09\linewidth]{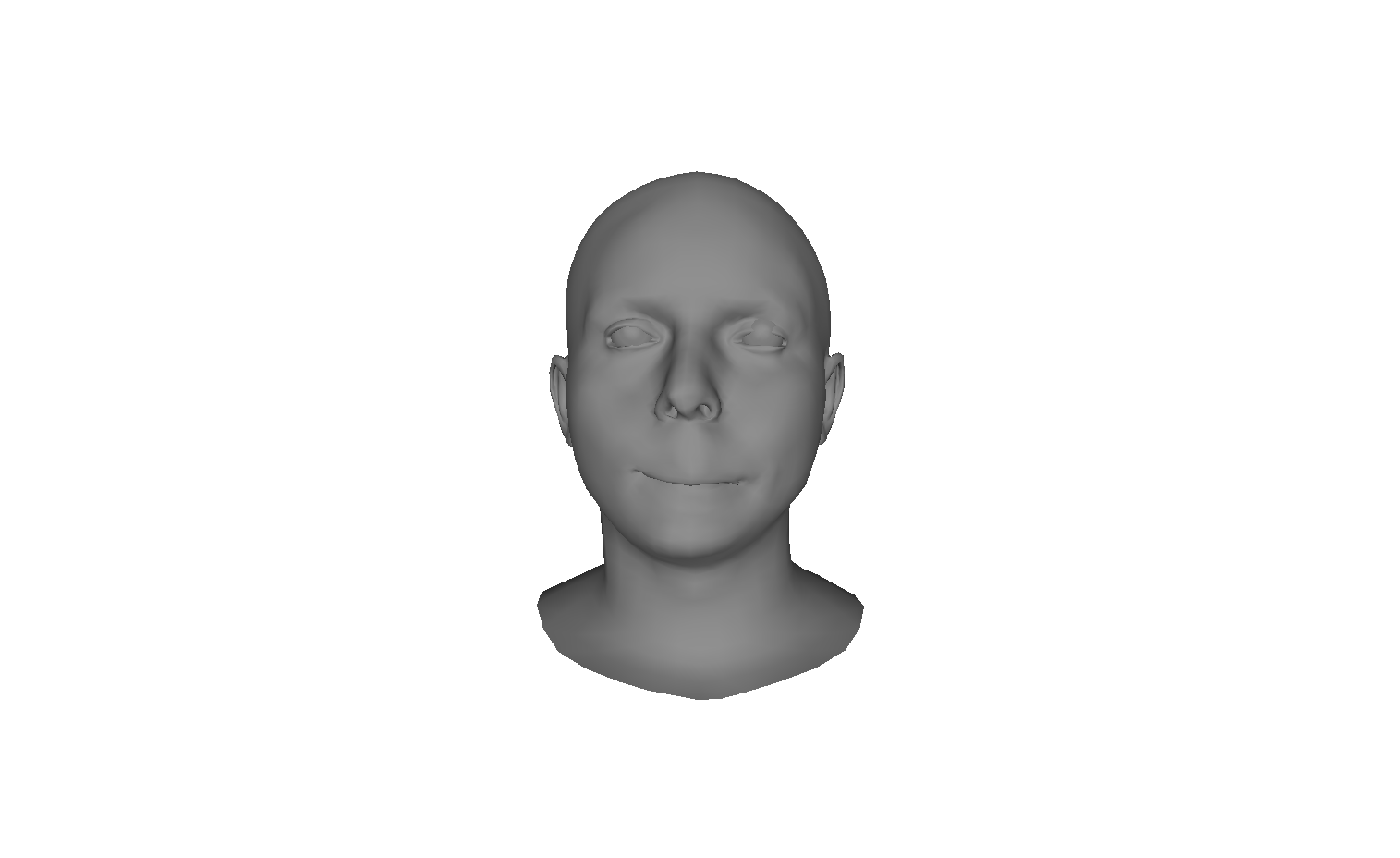}};
    \node[below of=c10, node distance=0.43cm] (c102) {\includegraphics[trim={400 240 380 300},clip,width=0.09\linewidth]{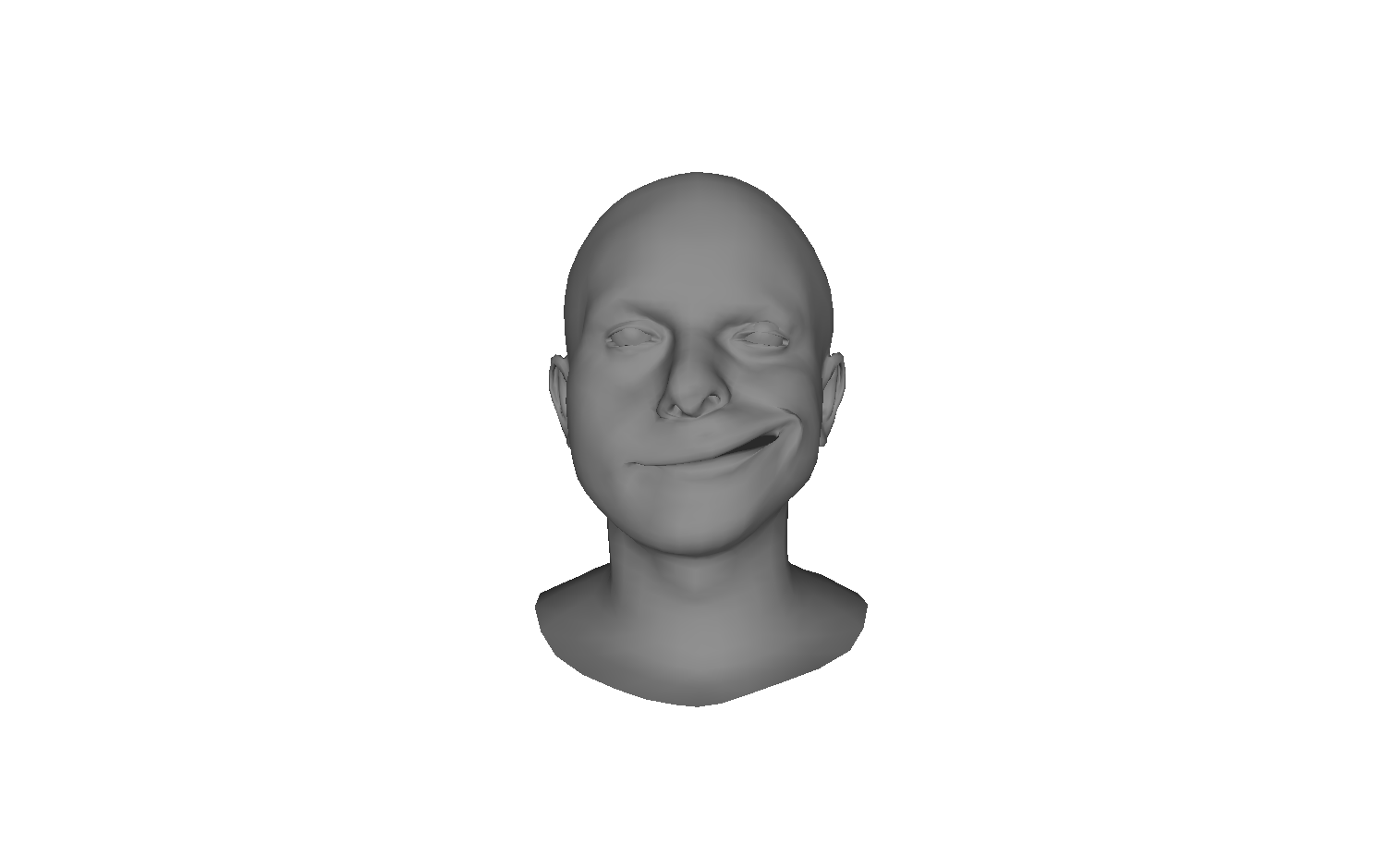}};
    \node[right of=c10, node distance=1.7cm] (c11) {\includegraphics[trim={400 80 400 100},clip,width=0.08\linewidth]{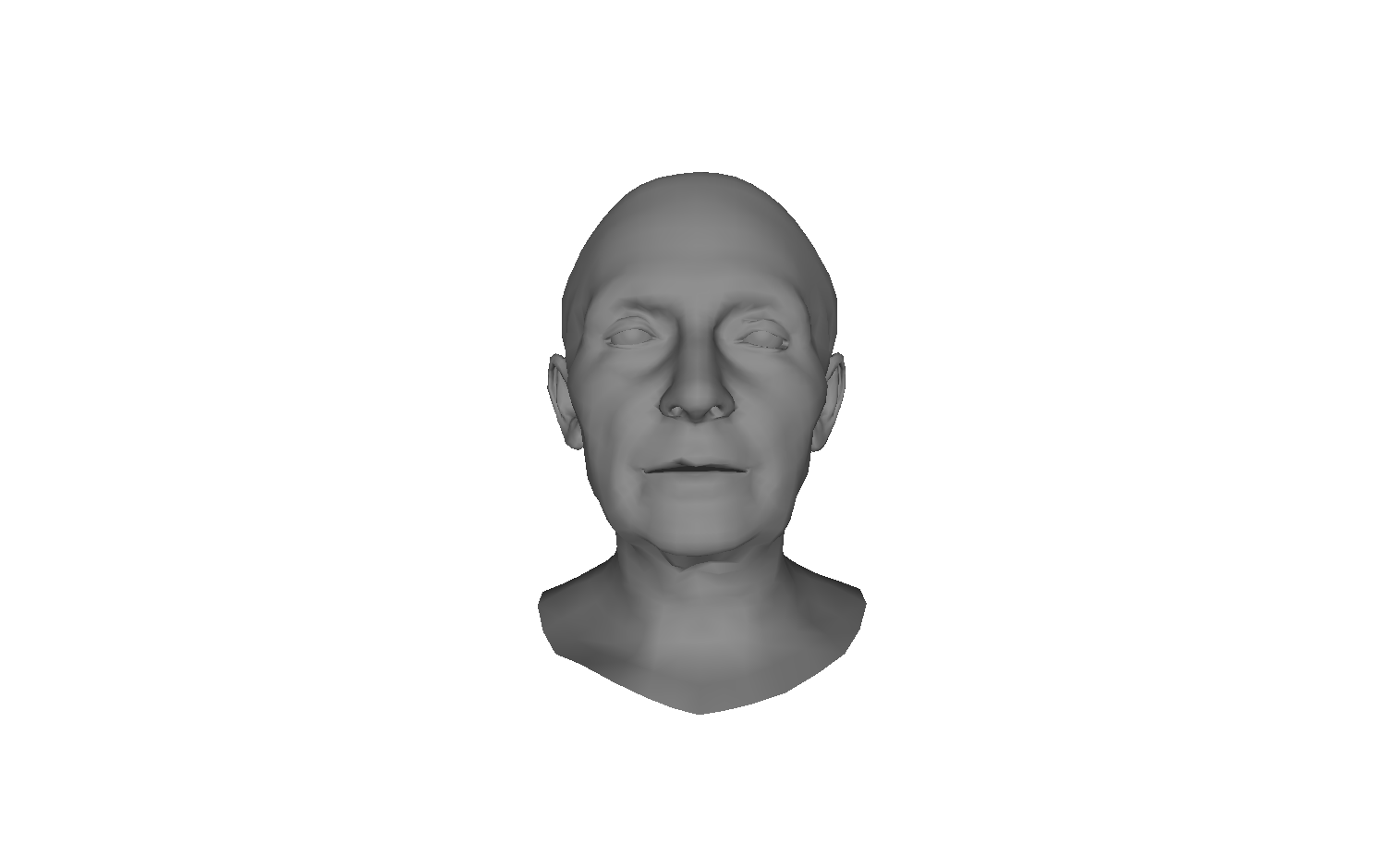}};
    
    \node[below of=c1, node distance=1.3cm] {Target Image};
    \node[below of=c2, node distance=1.3cm] {FLAME \cite{flame}};
    \node[below of=c3, node distance=1.3cm] {DECA \cite{deca}};
    \node[below of=c4, node distance=1.3cm, text width=1.8cm, align=center] {CFR-GAN \cite{occrobustwacv}};
    \node[below of=c5, node distance=1.3cm, text width=1.8cm, align=center] {Occ3DMM \cite{egger2018occlusion}};
    \node[below of=c6, node distance=1.3cm, text width=1.9cm, align=center] {Extreme3D \cite{tran2018extreme}};
    \coordinate (c89) at ($(c8)!0.5!(c9)$);
    \node[below of=c89, node distance=1.3cm] {Reconstructions by \ourmethod{} (Ours)};
    \node[below of=c11, node distance=1.3cm] {Ground truth};
    \end{tikzpicture}
    \caption{\textbf{Qualitative evaluation on the CoMA dataset \cite{coma}}: Reconstructed singular 3D meshes from the target image by the baselines \versus{} the diverse reconstructions (one full shape followed by six partial zoomed-in variations) from \ourmethod{}.}
    \label{fig:coma}
\end{figure*}

\subsection{Quantitative Results}\label{subsec:quantitative}
~\cref{tab:fitting} reports the 3D reconstruction accuracy in terms of mean shape error (MSE) on artificially occluded test images from the CoMA dataset \cite{coma} for different approaches using the FLAME \cite{flame} topology. Across all occlusion types, our proposed global+local model reports the lowest MSE values. The large gap between FLAME (fitting) \cite{flame}, DECA \cite{deca} and our approach demonstrates the necessity of region-specific model fitting for occlusion robustness.

Due to the lack of existing diverse 3D reconstruction approaches, we formulate four baselines to evaluate the diversity performance of \ourmethod{}: 1) fitting FLAME on the visible parts plus DPP loss on the occluded parts (FLAME+DPP), 2) replace FLAME in (1) with our global+local model (Global+Local+DPP), 3) fitting global+local model followed by shape completions by the Mesh-VAE as per the learned distribution $p(\bm{S}_c, \bm{z}|\bm{S}_m)$ (Global+Local+VAE), and 4) replacing the global+local model with FLAME\cite{flame} in \ourmethod{} (FLAME+VAE+DPP). We report the quantitative metrics in ~\cref{tab:diversity}. Across all occlusion types, FLAME+DPP and Global+Local+DPP report much higher \textit{CSE} and \textit{ASD-V,} and lower \textit{ASD-O} than \ourmethod{}. Though Global+Local+VAE obtains lower \textit{CSE} than \ourmethod{}, it does so at the cost of reduced diversity in terms of \textit{ASD-O}. FLAME+VAE+DPP reports better diversity metrics but at the cost of higher \textit{CSE} errors. On the other hand, \ourmethod{} reports the lowest \textit{ASD-V}, the highest \textit{ASD-O}, and the second-lowest \textit{CSE}, satisfying the three desired qualities mentioned earlier. These observations confirm our hypothesis that explicitly accounting for occlusions and optimizing for diversity can lead to 3D reconstructions that are both more accurate (on the visible regions) and more geometrically diverse (on the occluded regions). Among the different occlusion types, we report the highest \textit{ASD-O} for face-masks. These results are consistent with the fact that human faces have higher variability in the mouth and nose regions, which our approach is able to learn and reproduce.

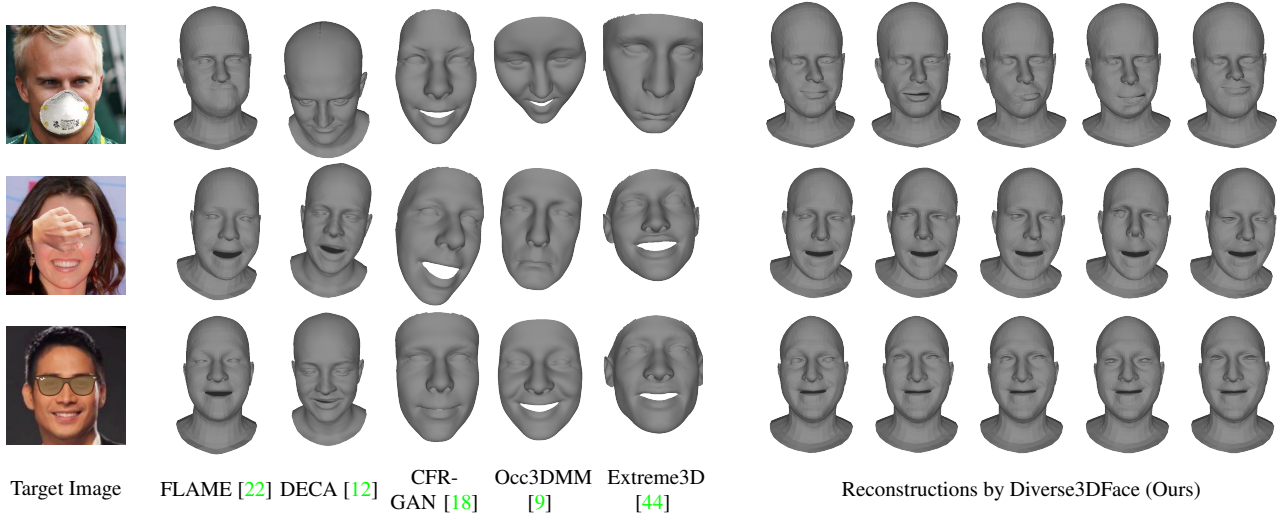
\begin{figure*}
    \centering
    \begin{tikzpicture}
    \footnotesize
    \node (a1) {\includegraphics[width=0.09\linewidth]{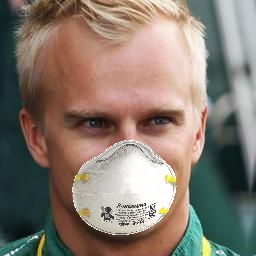}};
    \node[right of=a1, node distance=2.0cm] (a2) {\includegraphics[trim={400 80 400 100},clip,width=0.08\linewidth]{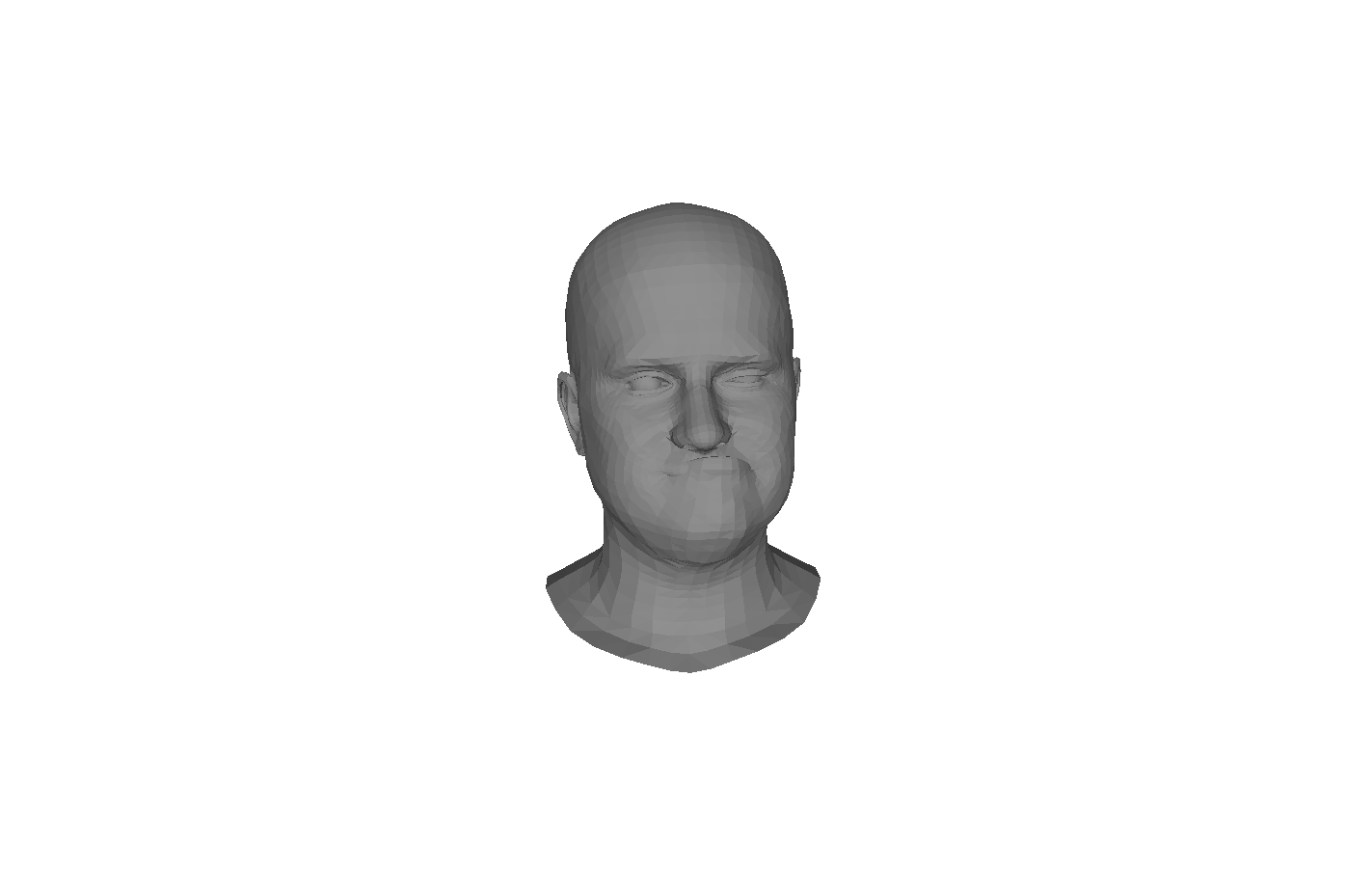}};
    \node[right of=a2, node distance=1.5cm] (a3) {\includegraphics[trim={400 80 400 100},clip,width=0.08\linewidth]{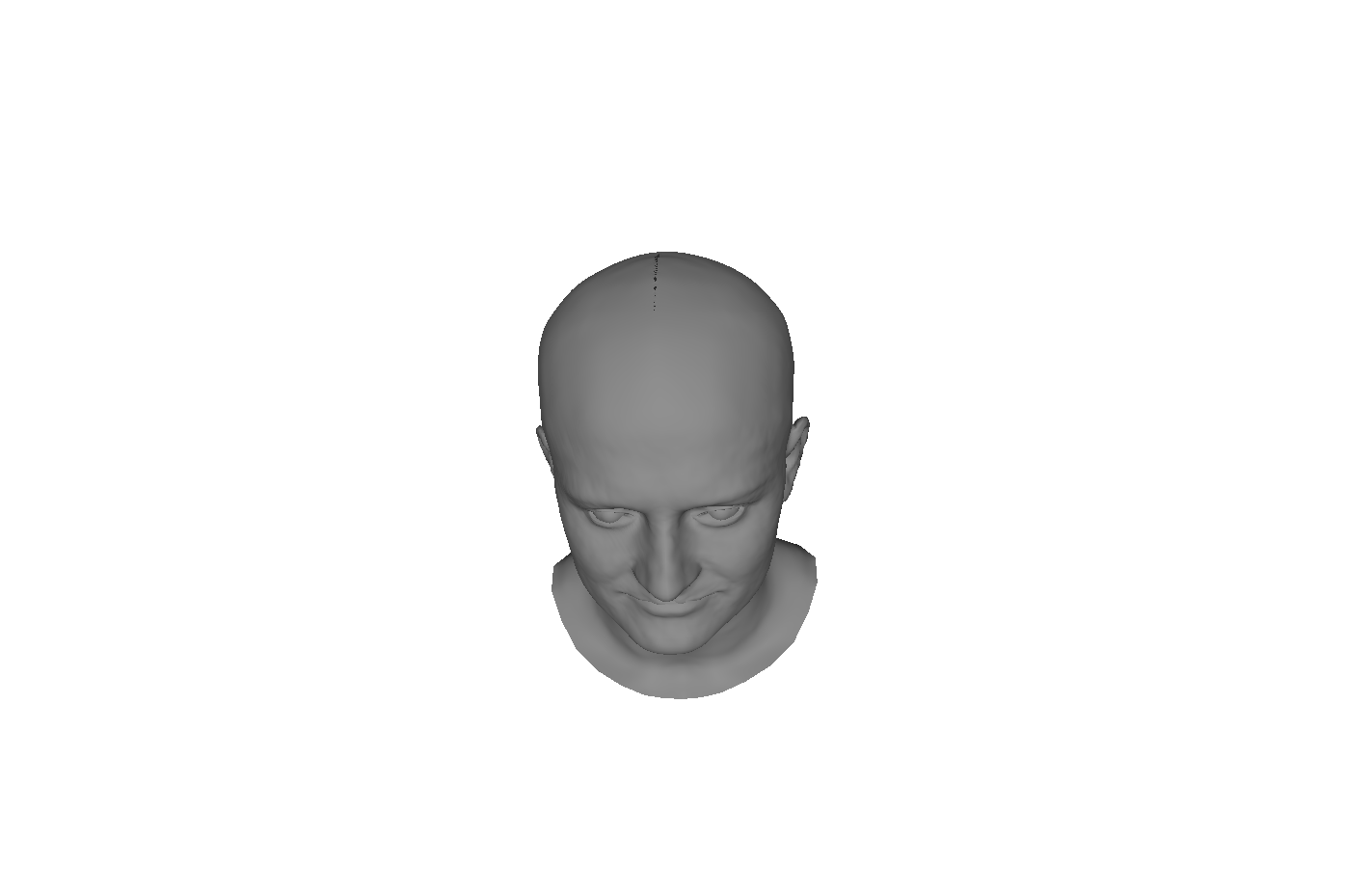}};
    \node[right of=a3, node distance=1.4cm] (a4) {\includegraphics[trim={400 80 400 100},clip,width=0.075\linewidth]{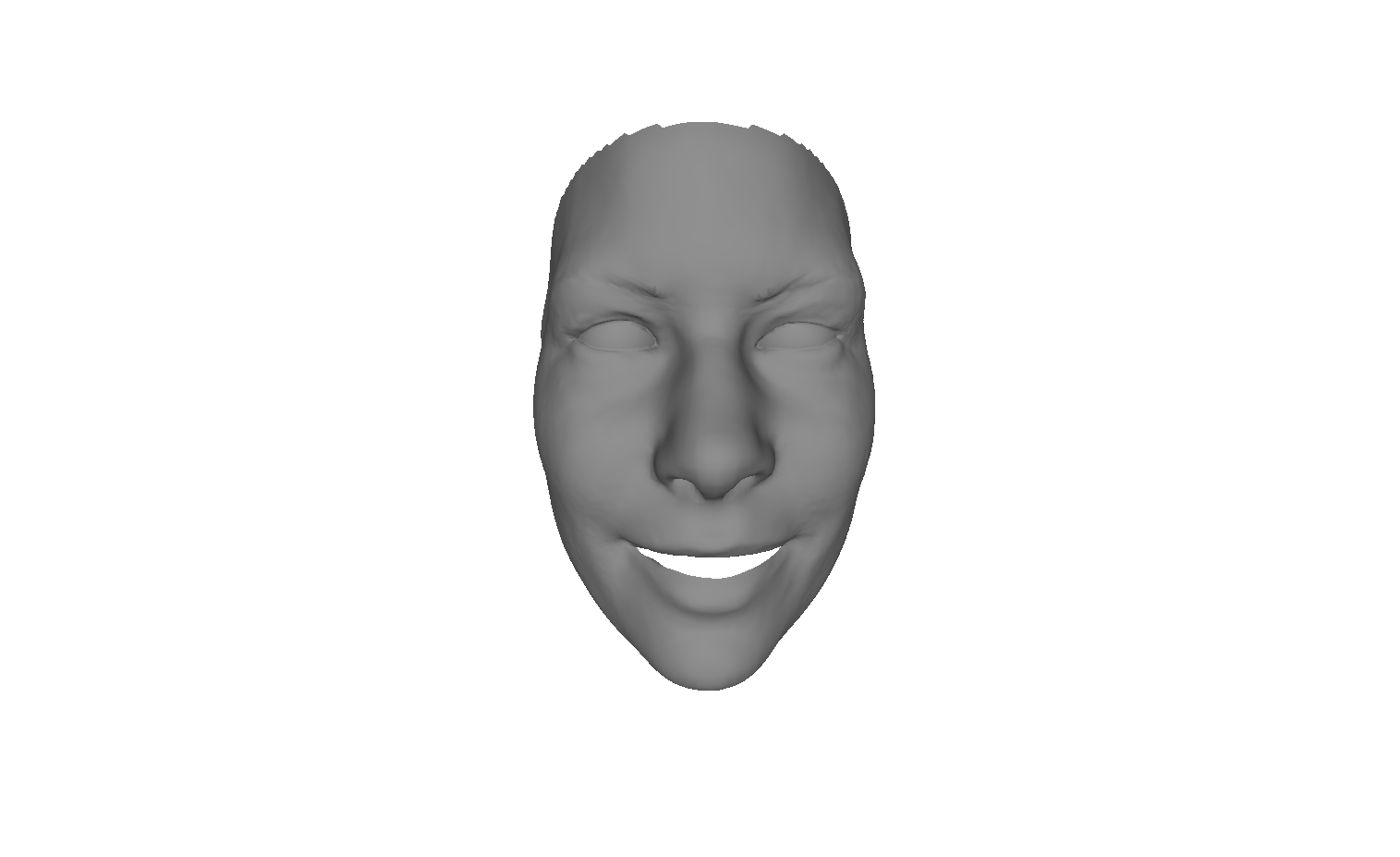}};
    \node[right of=a4, node distance=1.4cm] (a5) {\includegraphics[trim={400 80 400 100},clip,width=0.07\linewidth]{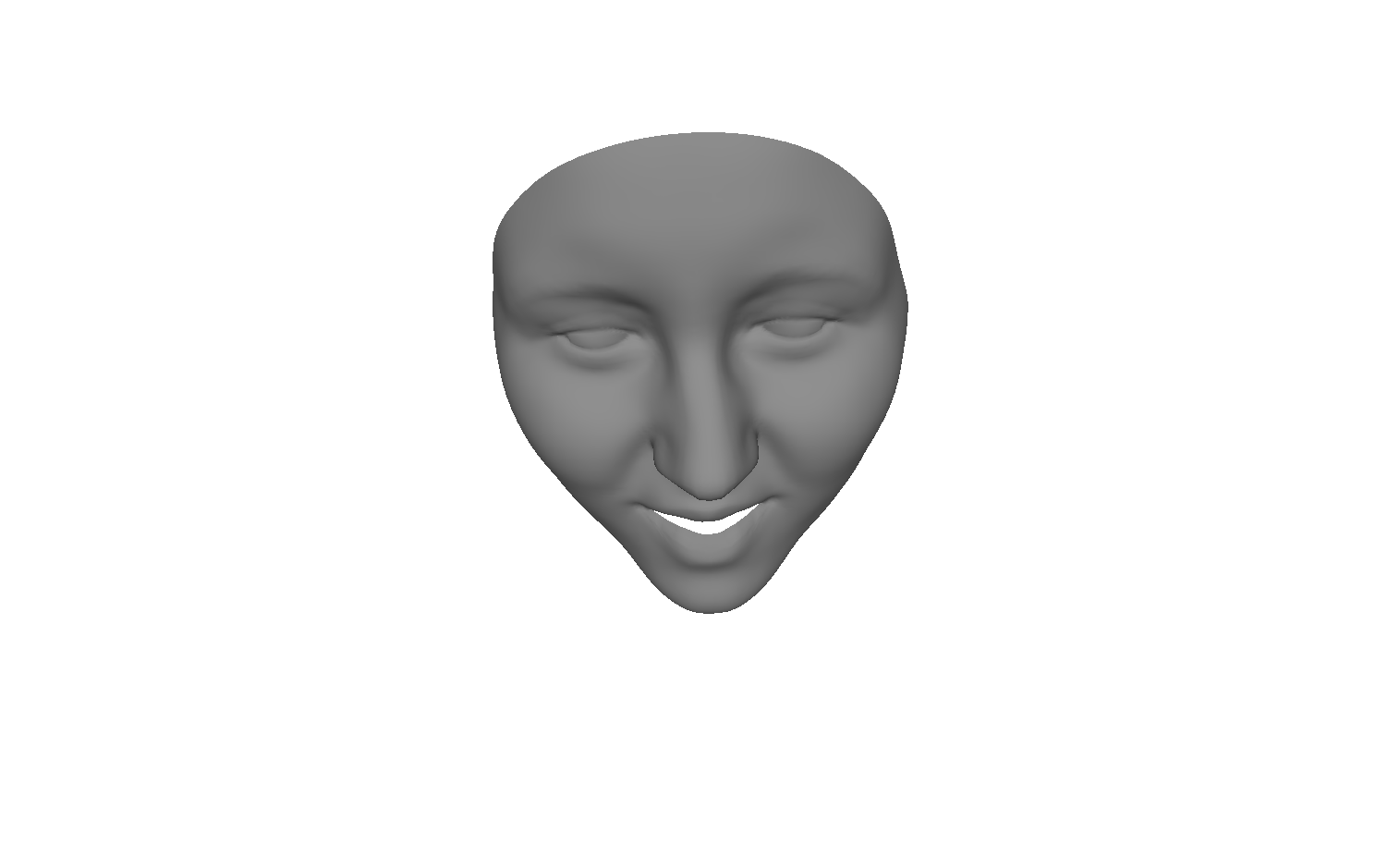}};
    \node[right of=a5, node distance=1.5cm] (a6) {\includegraphics[trim={400 80 400 100},clip,width=0.075\linewidth]{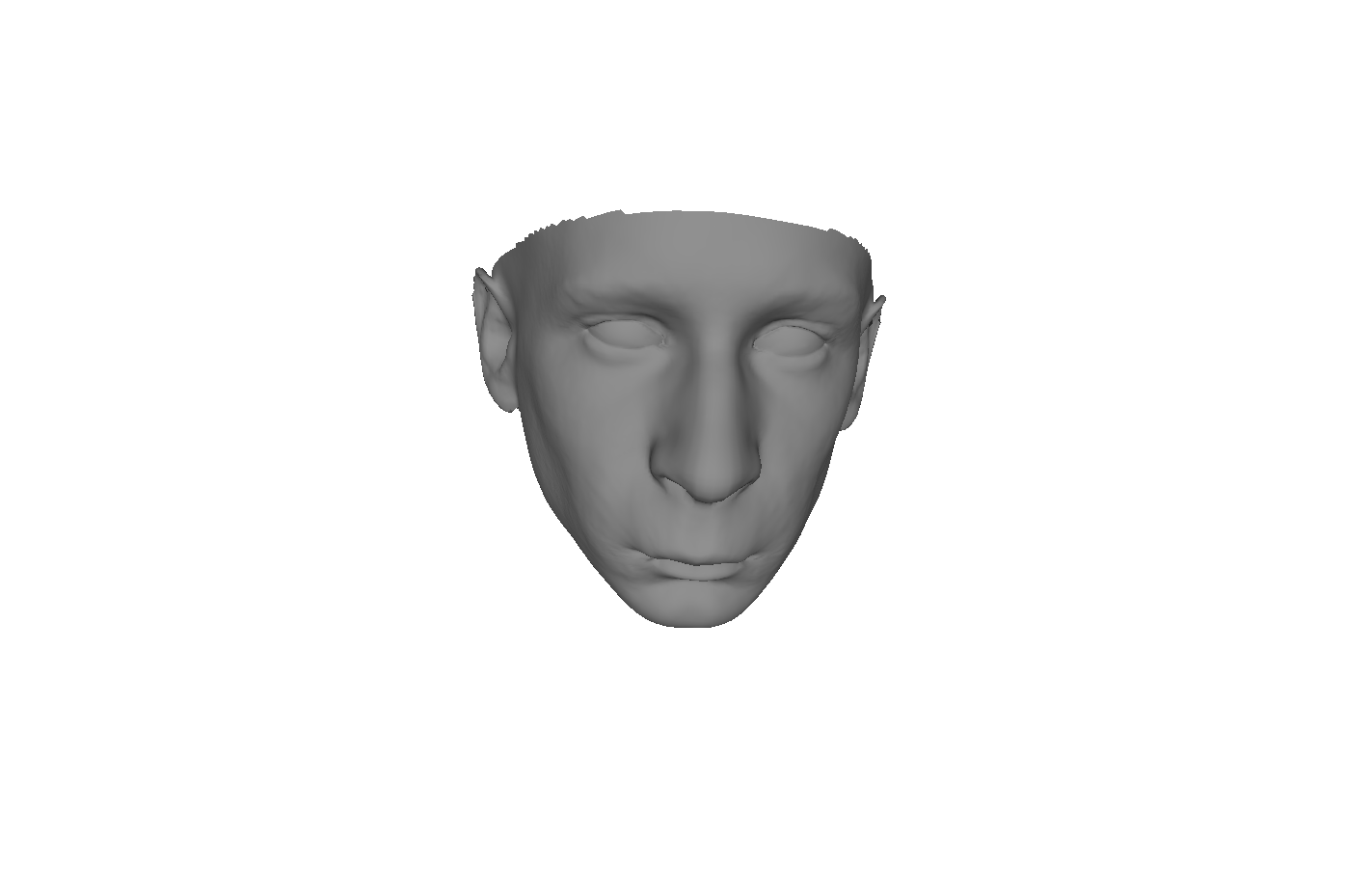}};
    \node[right of=a6, node distance=2.1cm] (a7) {\includegraphics[trim={400 80 400 100},clip,width=0.08\linewidth]{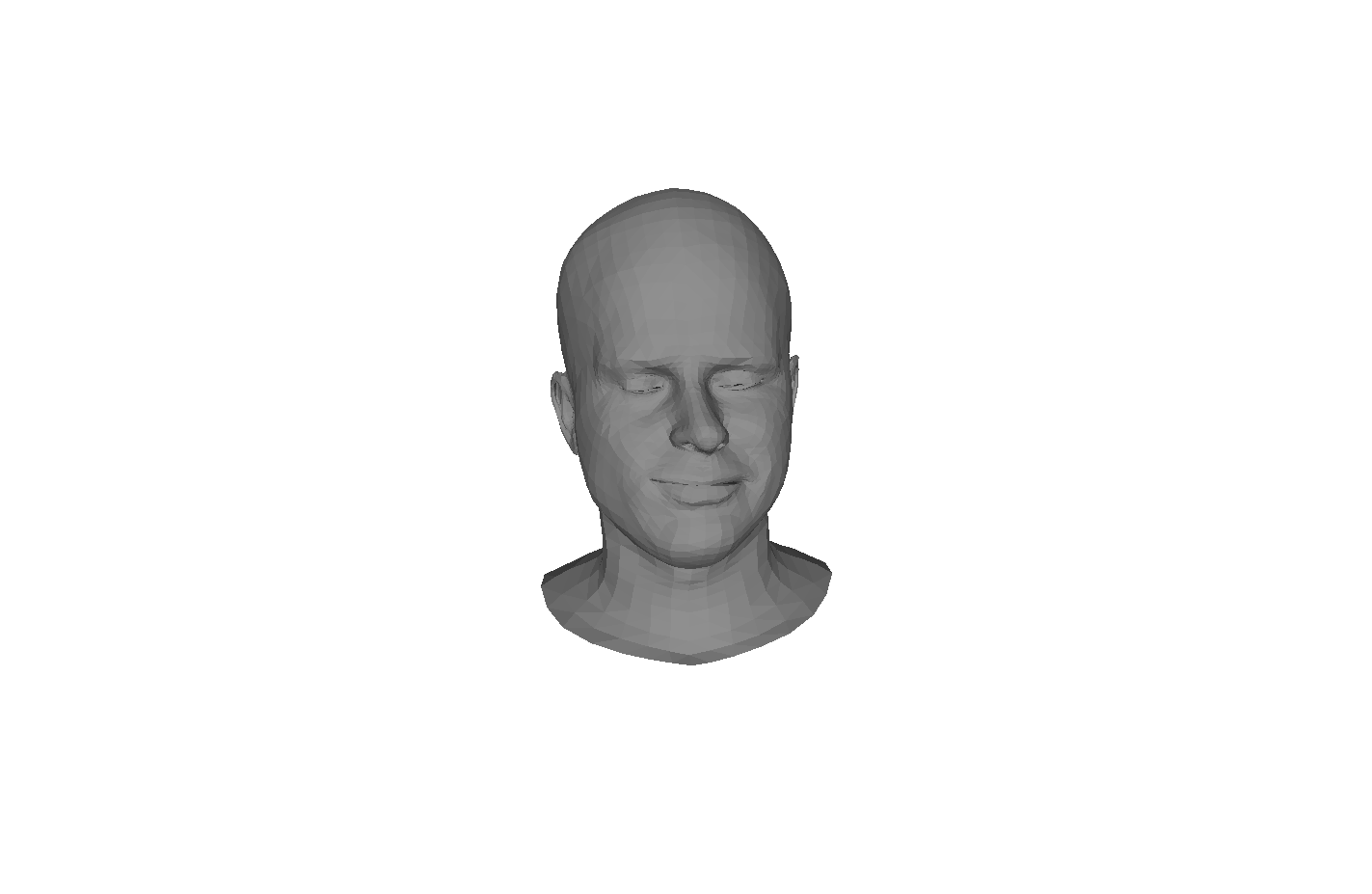}};
    \node[right of=a7, node distance=1.4cm] (a8) {\includegraphics[trim={400 80 400 100},clip,width=0.08\linewidth]{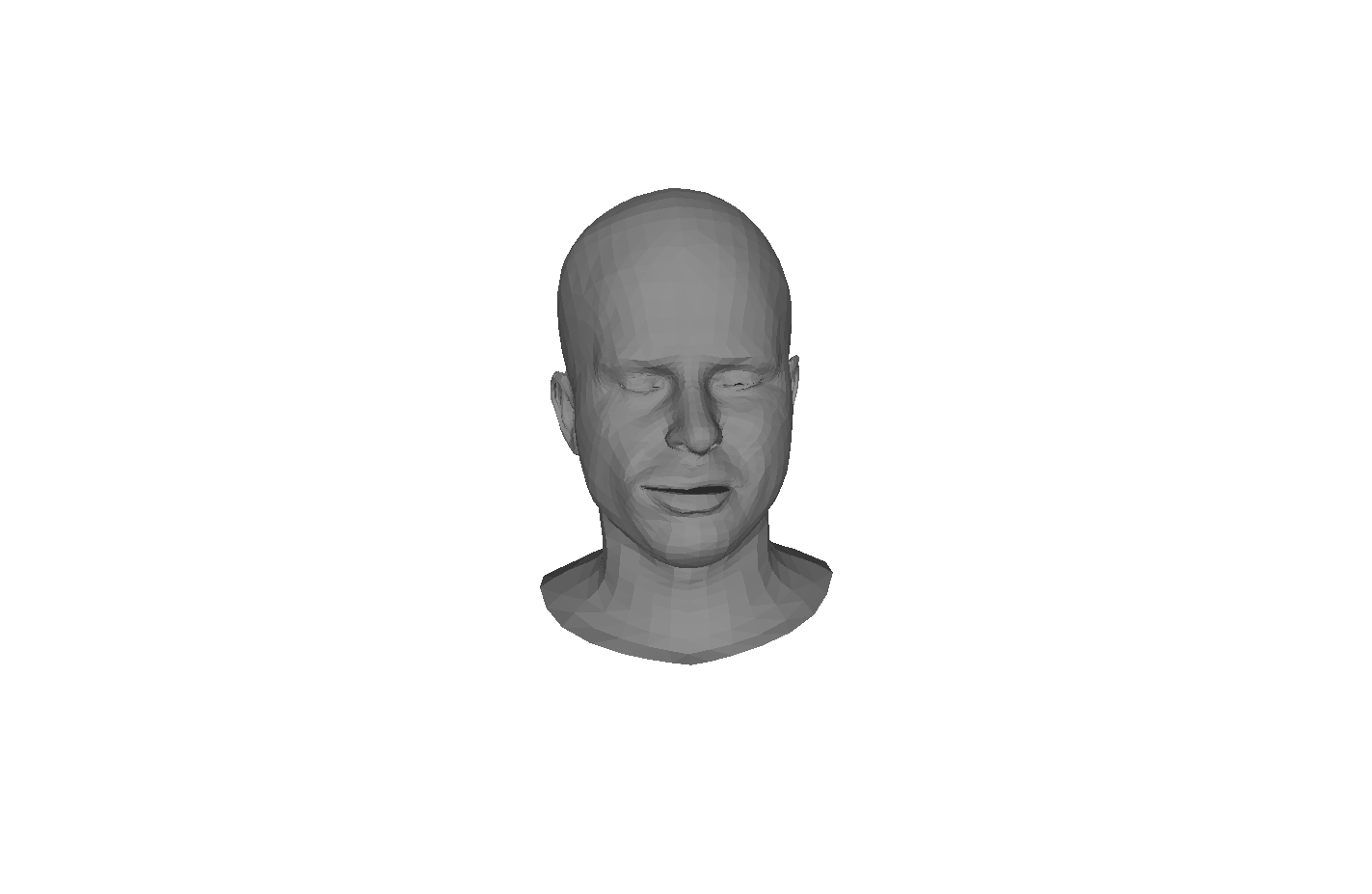}};
    \node[right of=a8, node distance=1.4cm] (a9) {\includegraphics[trim={400 80 400 100},clip,width=0.08\linewidth]{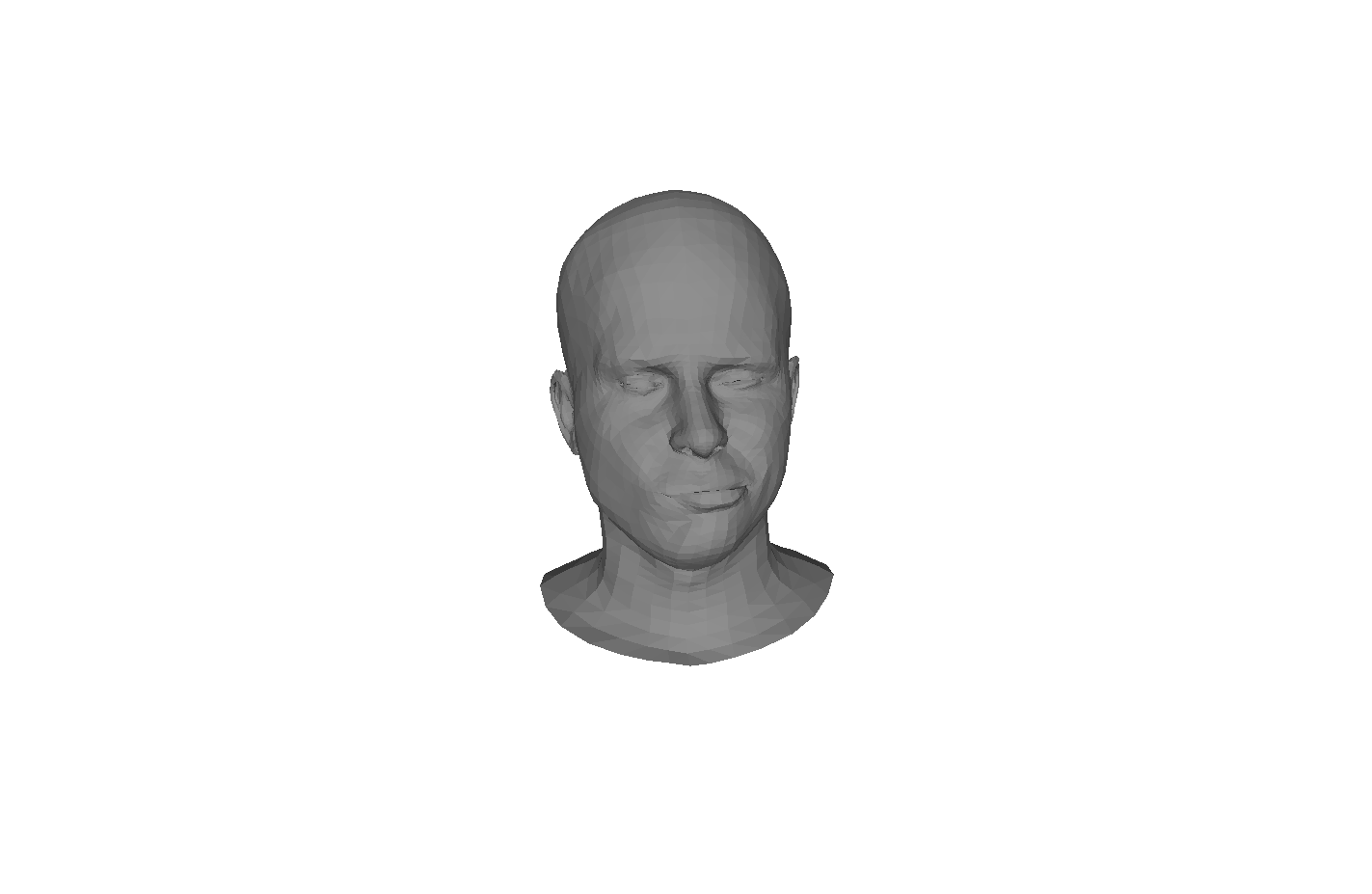}};
    \node[right of=a9, node distance=1.4cm] (a10) {\includegraphics[trim={400 80 400 100},clip,width=0.08\linewidth]{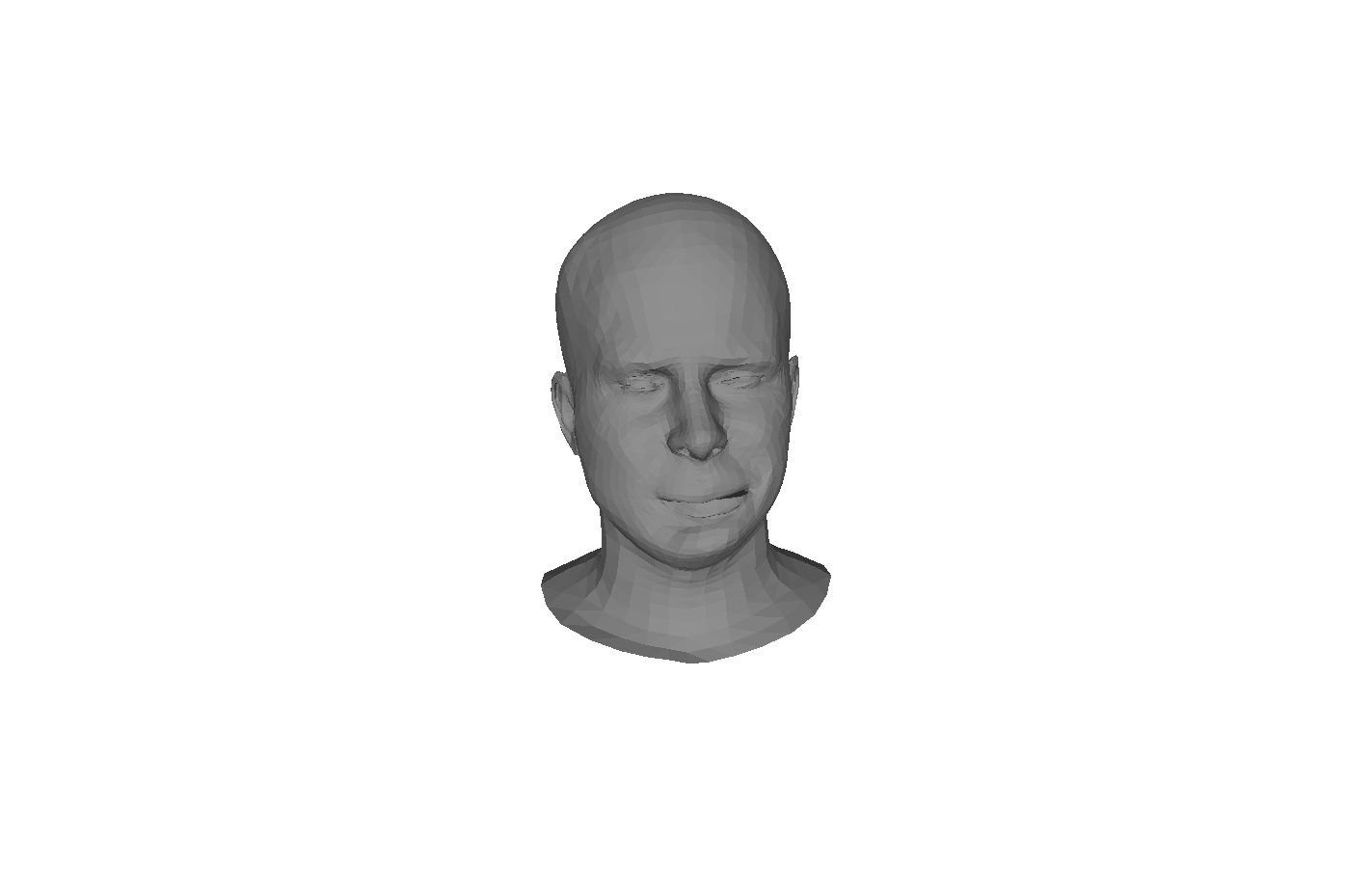}};
    \node[right of=a10, node distance=1.4cm] (a11) {\includegraphics[trim={400 80 400 100},clip,width=0.08\linewidth]{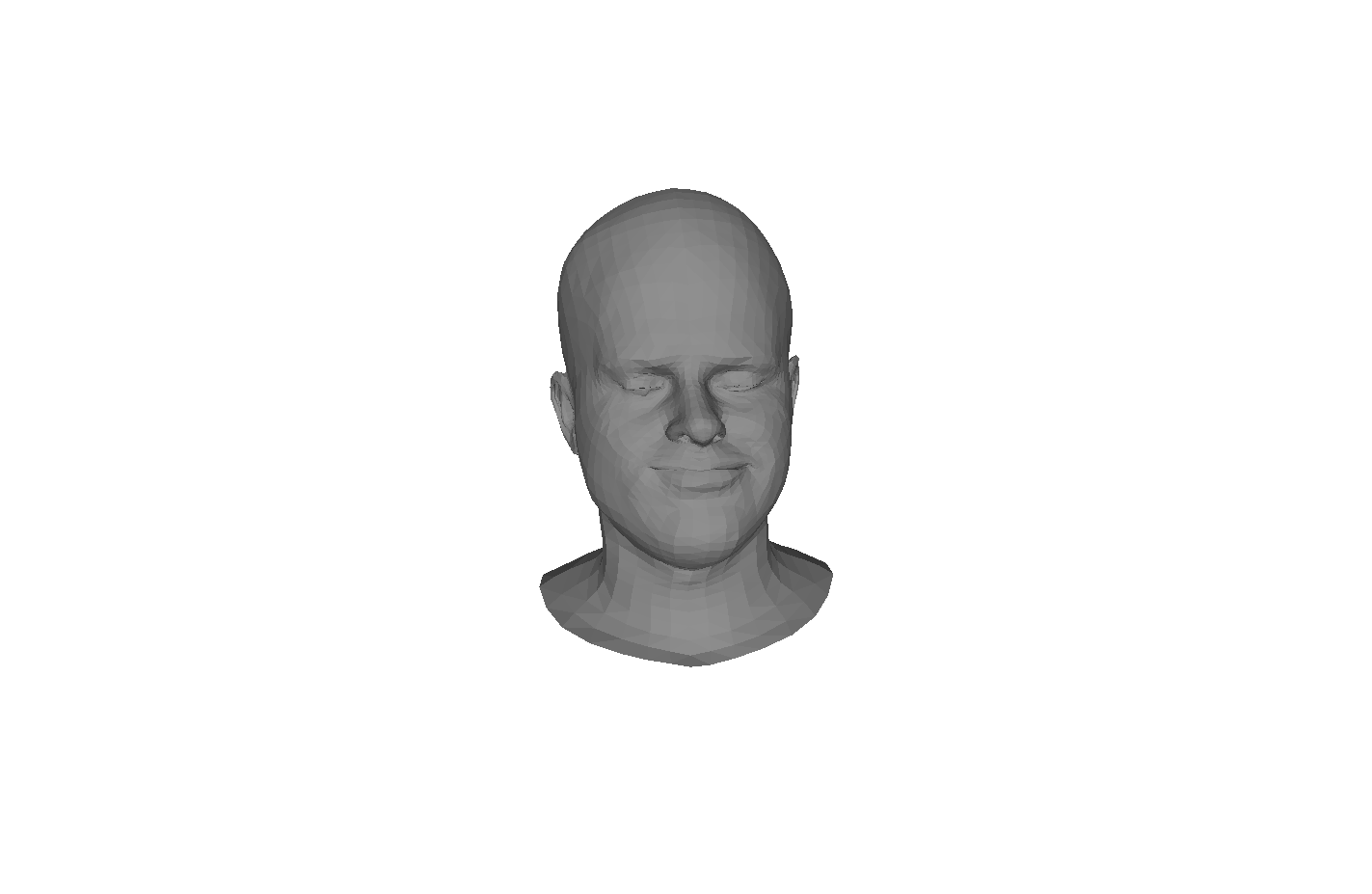}};
        
    \node[below of=a1, node distance=2.0cm] (c1) {\includegraphics[width=0.09\linewidth]{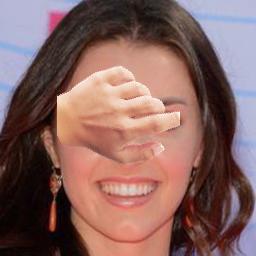}};
    \node[right of=c1, node distance=2.0cm] (c2) {\includegraphics[trim={400 80 400 100},clip,width=0.08\linewidth]{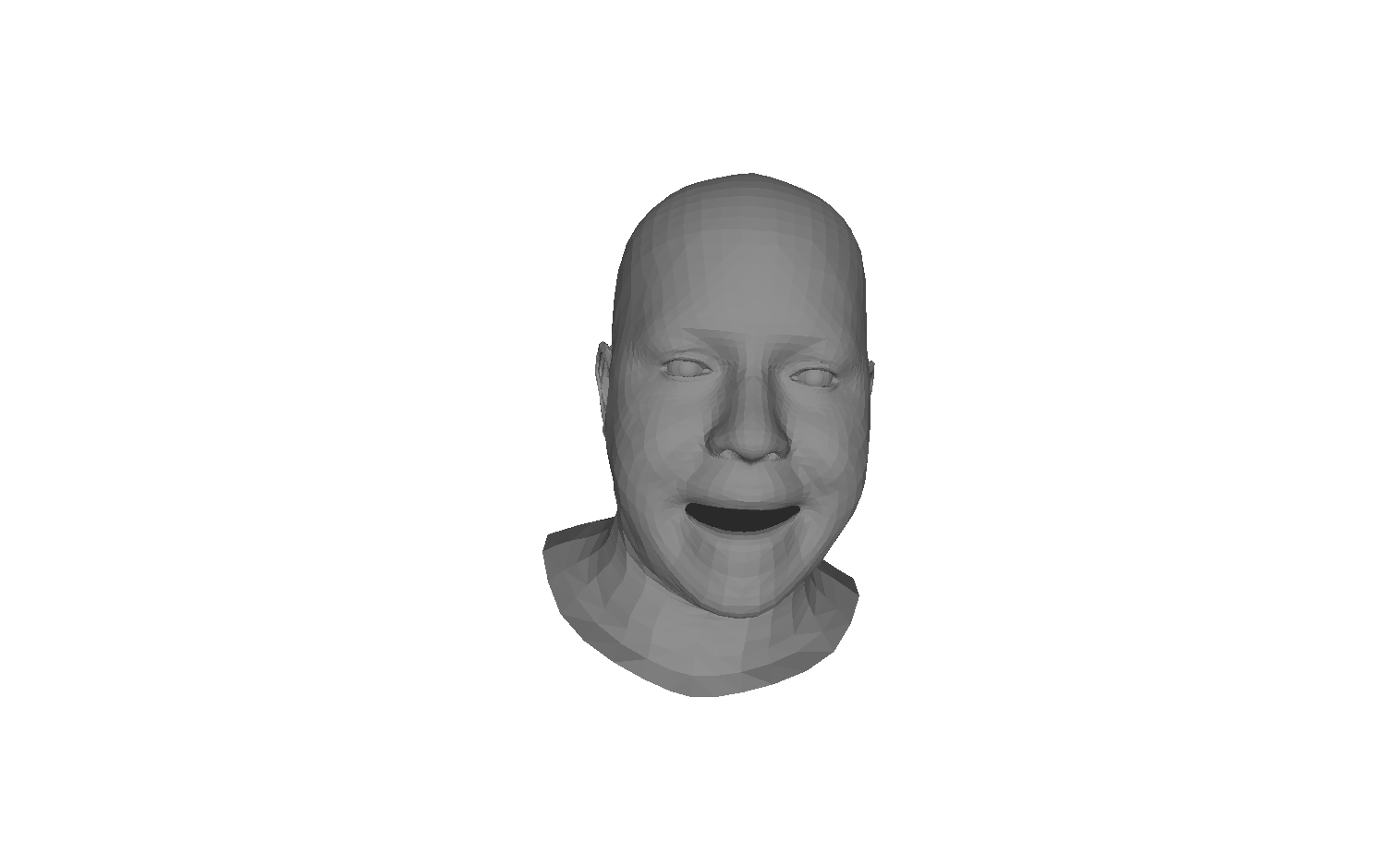}};
    \node[right of=c2, node distance=1.5cm] (c3) {\includegraphics[trim={400 80 400 100},clip,width=0.08\linewidth]{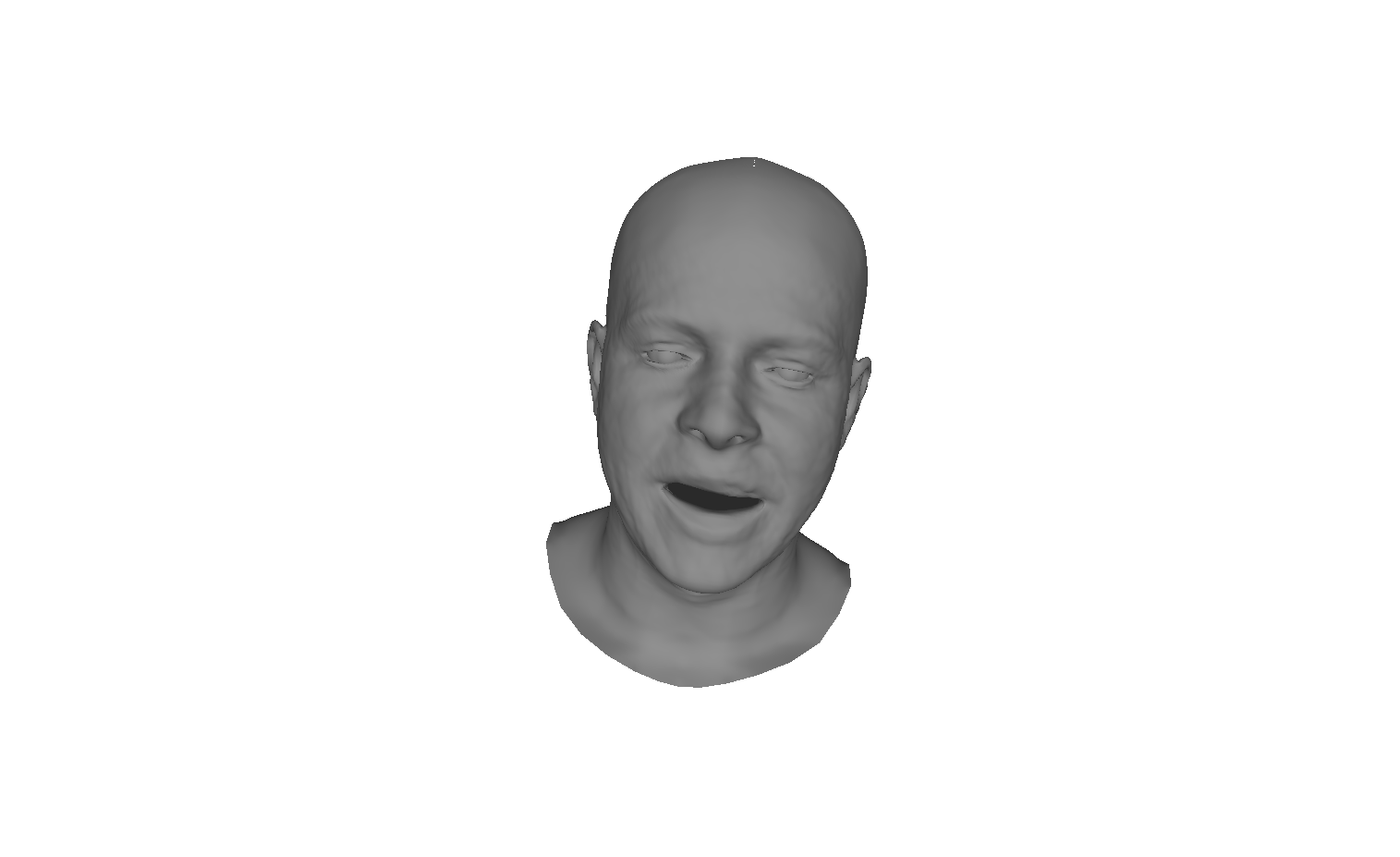}};
    \node[right of=c3, node distance=1.4cm] (c4) {\includegraphics[trim={400 80 400 100},clip,width=0.075\linewidth]{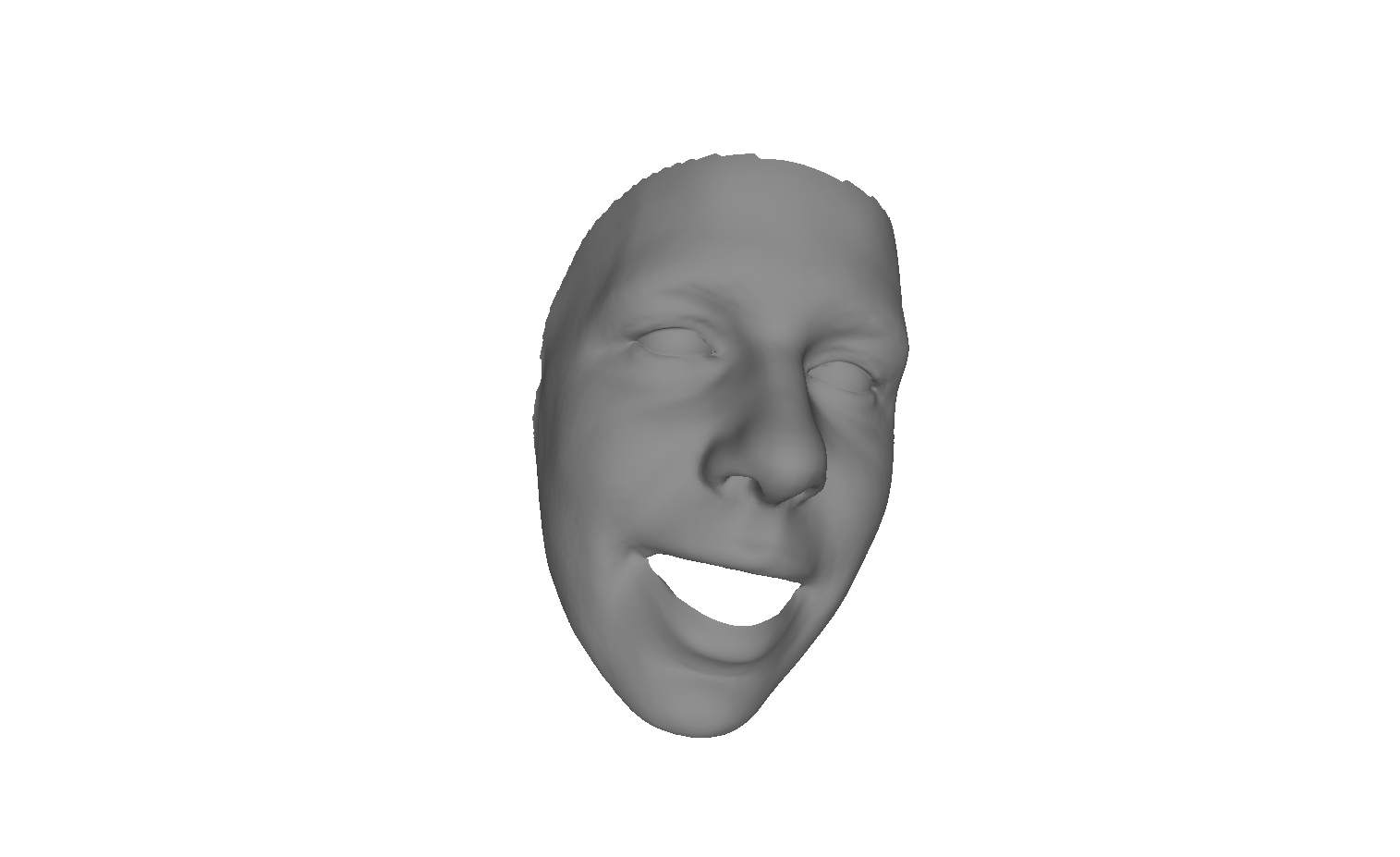}};
    \node[right of=c4, node distance=1.4cm] (c5) {\includegraphics[trim={400 80 400 100},clip,width=0.07\linewidth]{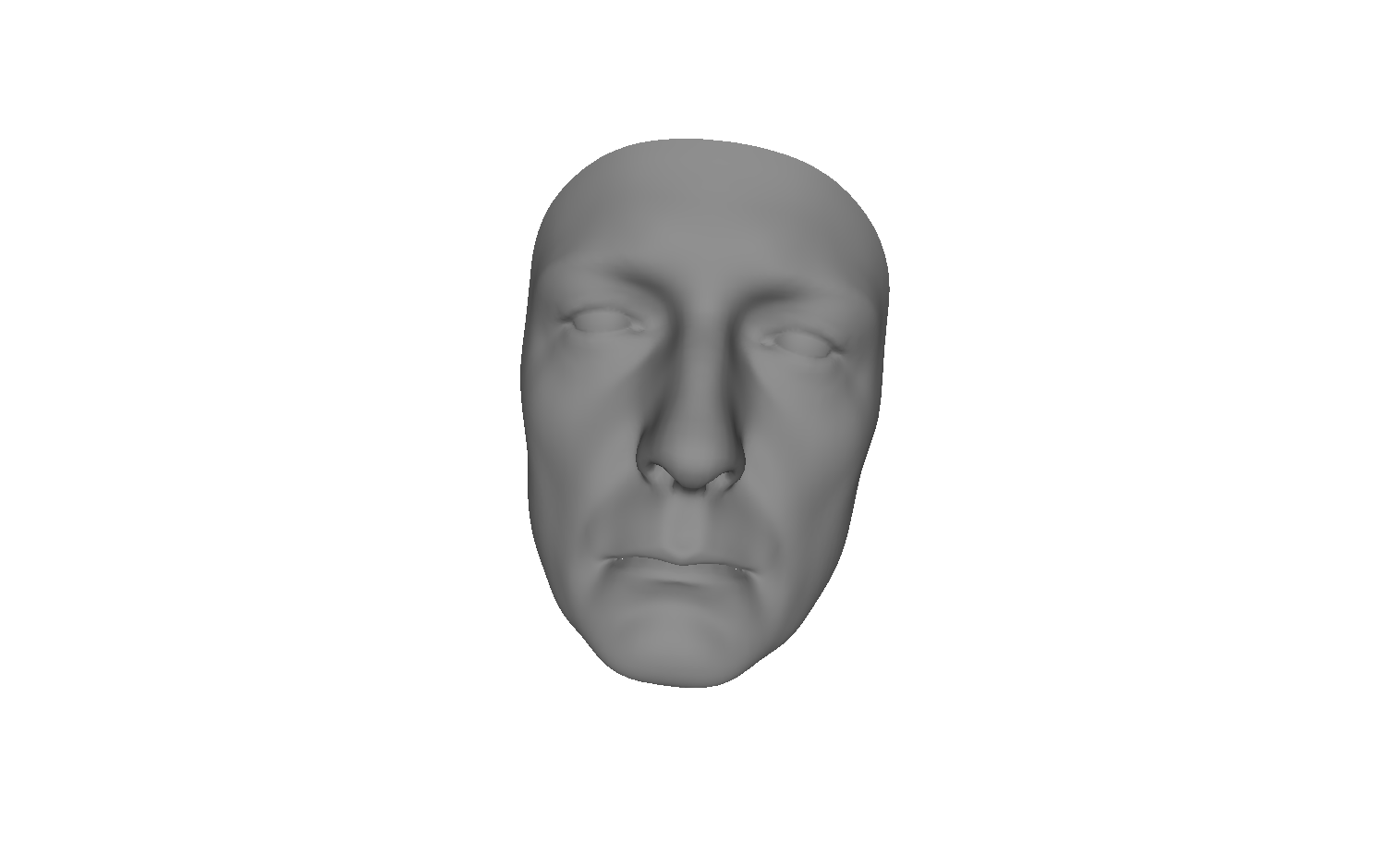}};
    \node[right of=c5, node distance=1.5cm] (c6) {\includegraphics[trim={400 80 400 100},clip,width=0.075\linewidth]{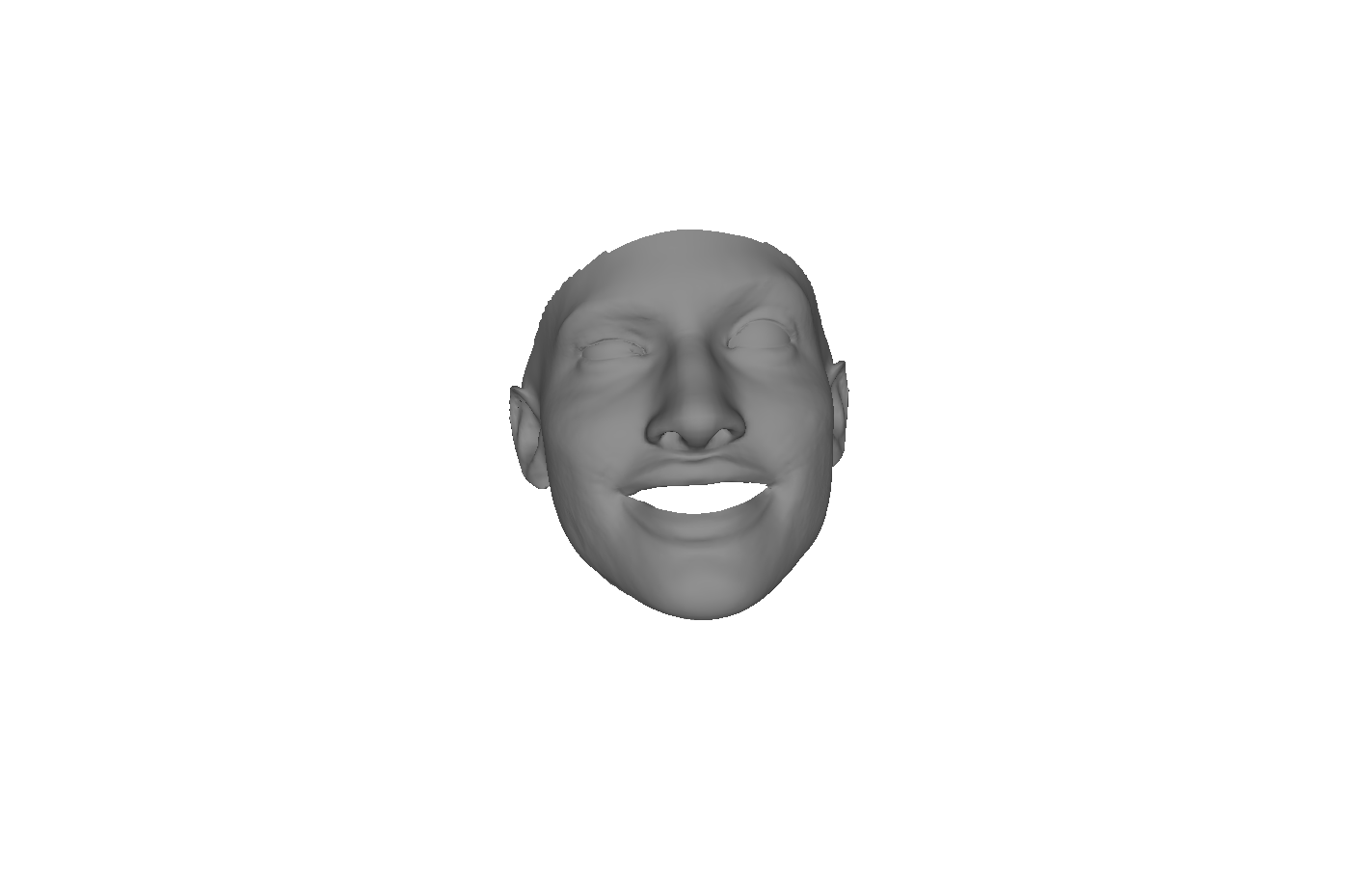}};
    \node[right of=c6, node distance=2.1cm] (c7) {\includegraphics[trim={400 80 400 100},clip,width=0.08\linewidth]{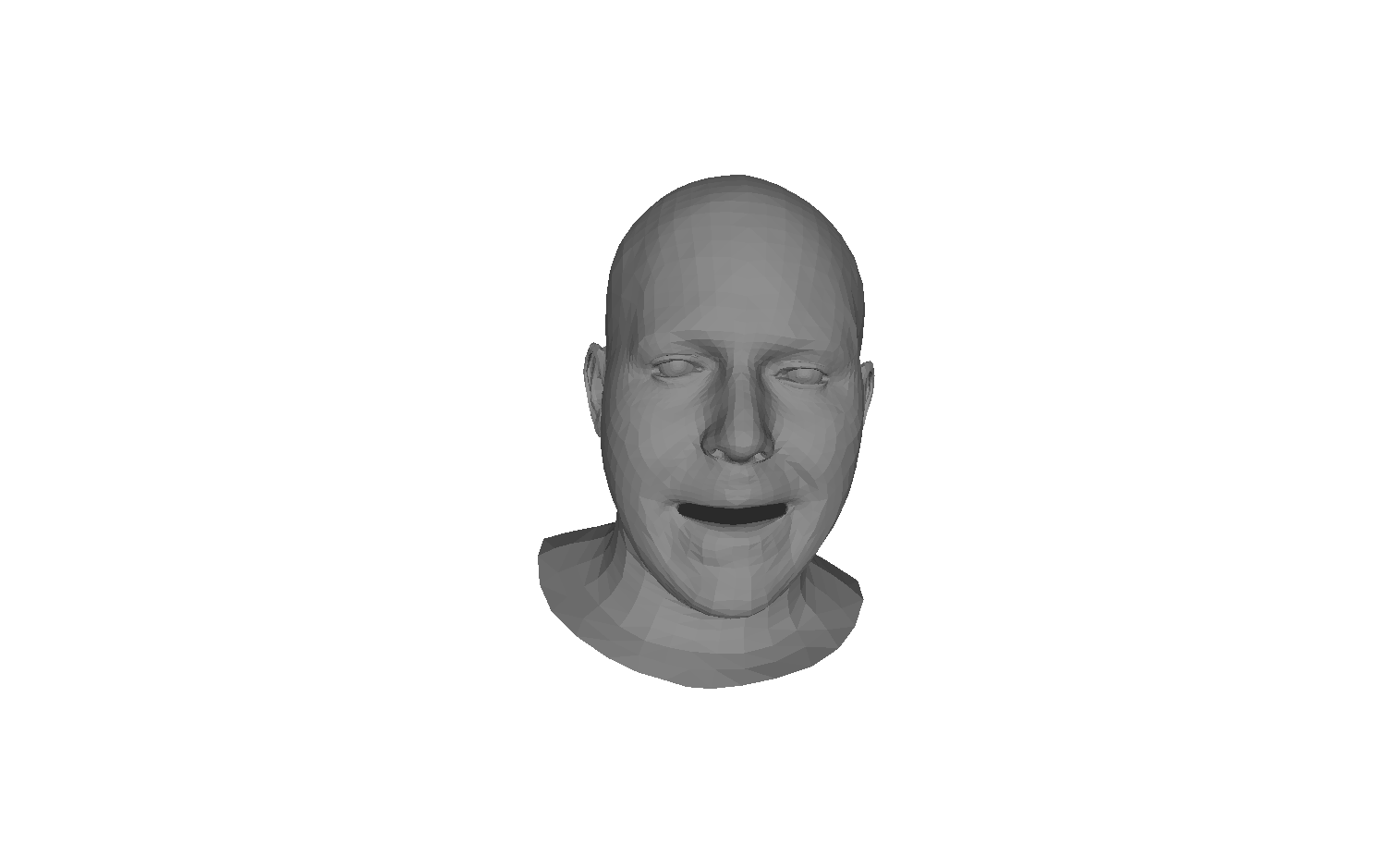}};
    \node[right of=c7, node distance=1.4cm] (c8) {\includegraphics[trim={400 80 400 100},clip,width=0.08\linewidth]{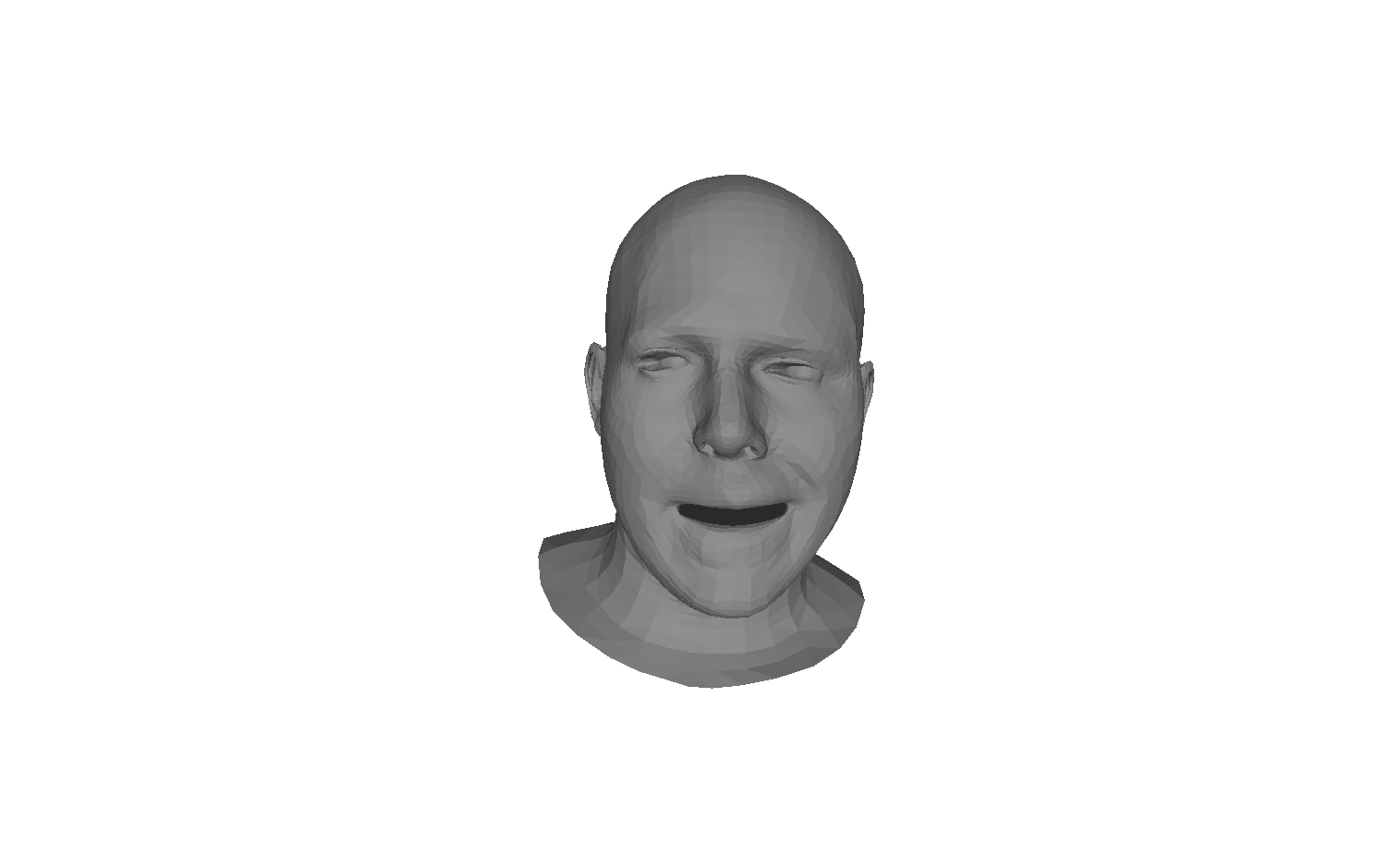}};
    \node[right of=c8, node distance=1.4cm] (c9) {\includegraphics[trim={400 80 400 100},clip,width=0.08\linewidth]{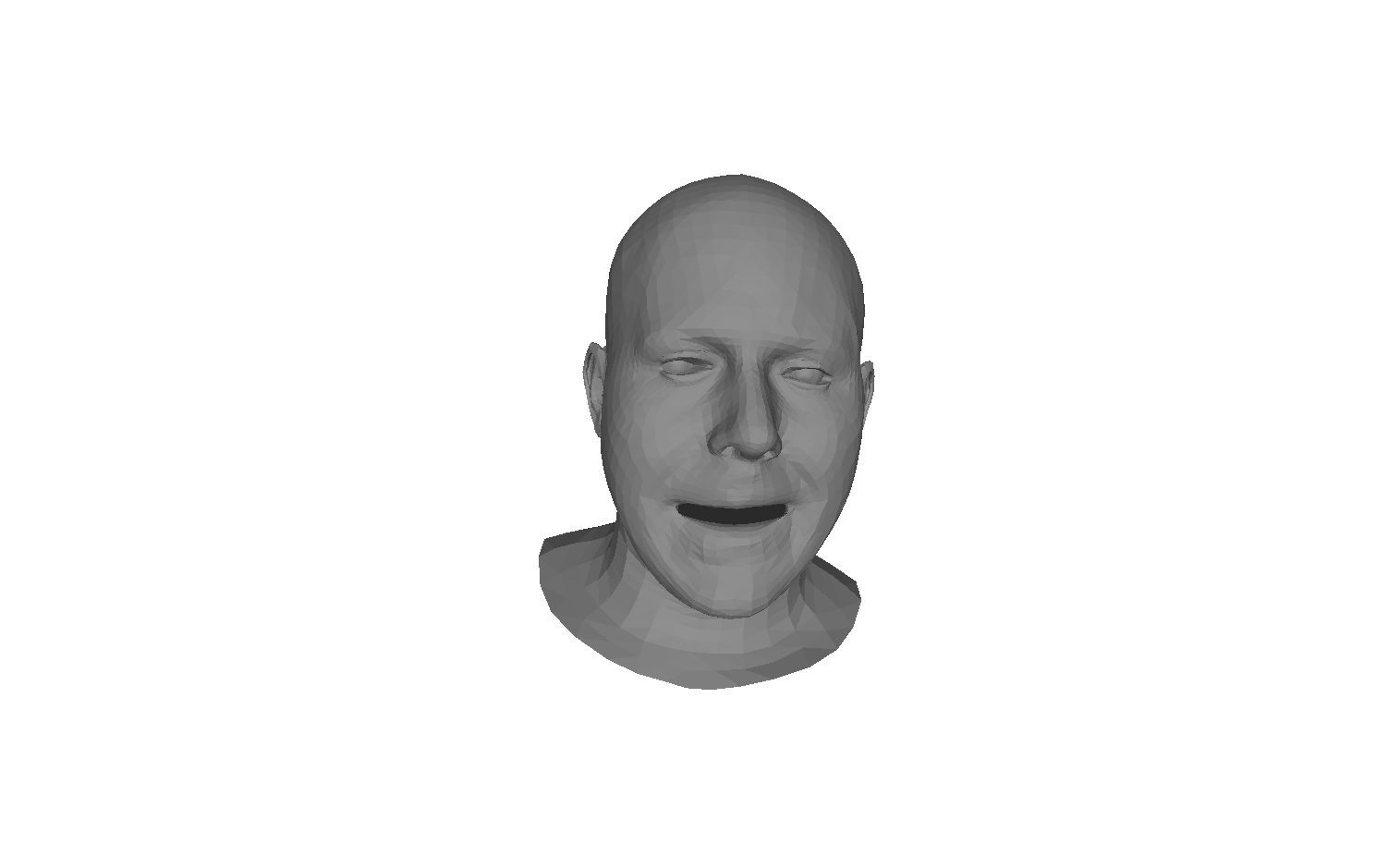}};
    \node[right of=c9, node distance=1.4cm] (c10) {\includegraphics[trim={400 80 400 100},clip,width=0.08\linewidth]{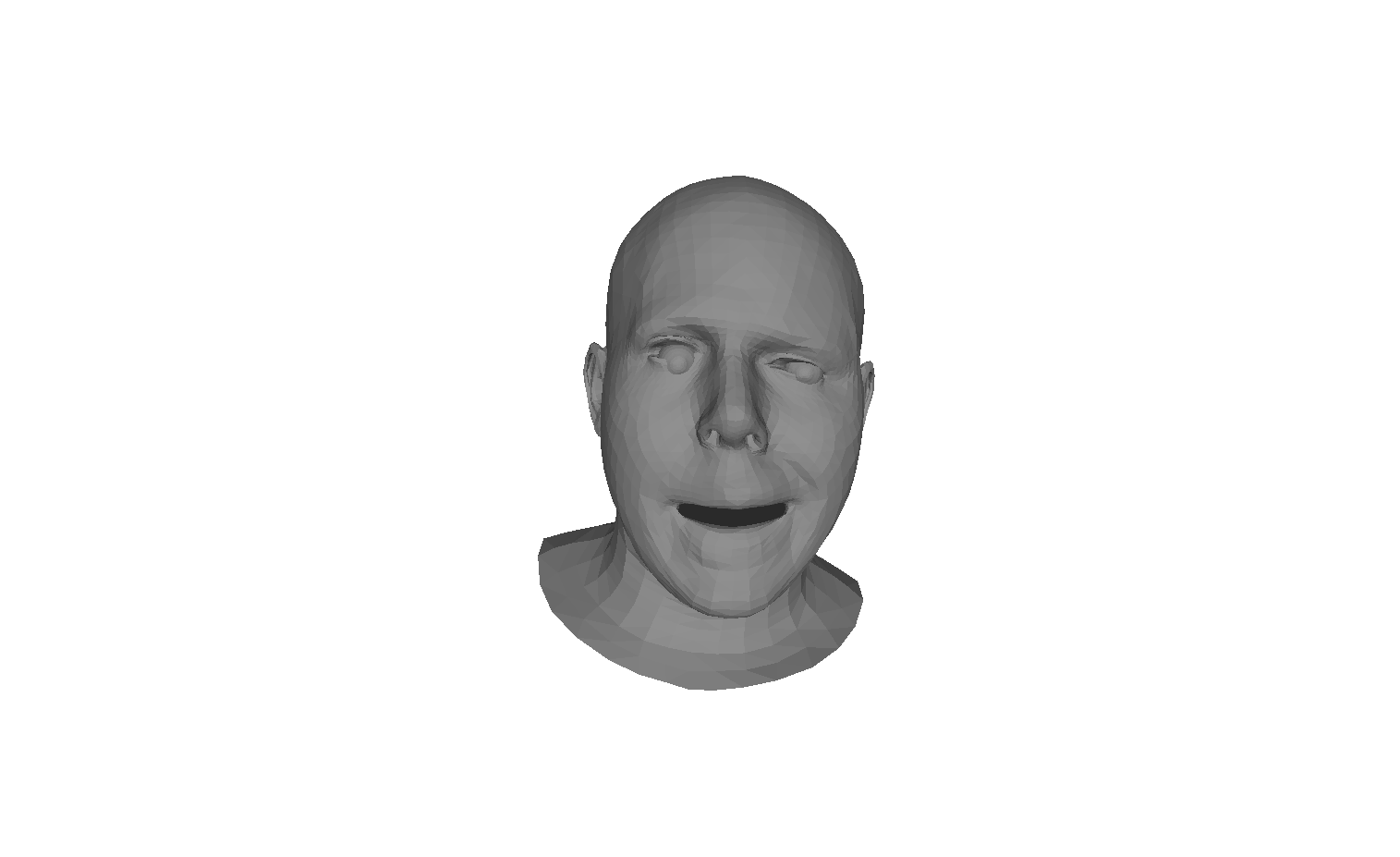}};
    \node[right of=c10, node distance=1.4cm] (c11) {\includegraphics[trim={400 80 400 100},clip,width=0.08\linewidth]{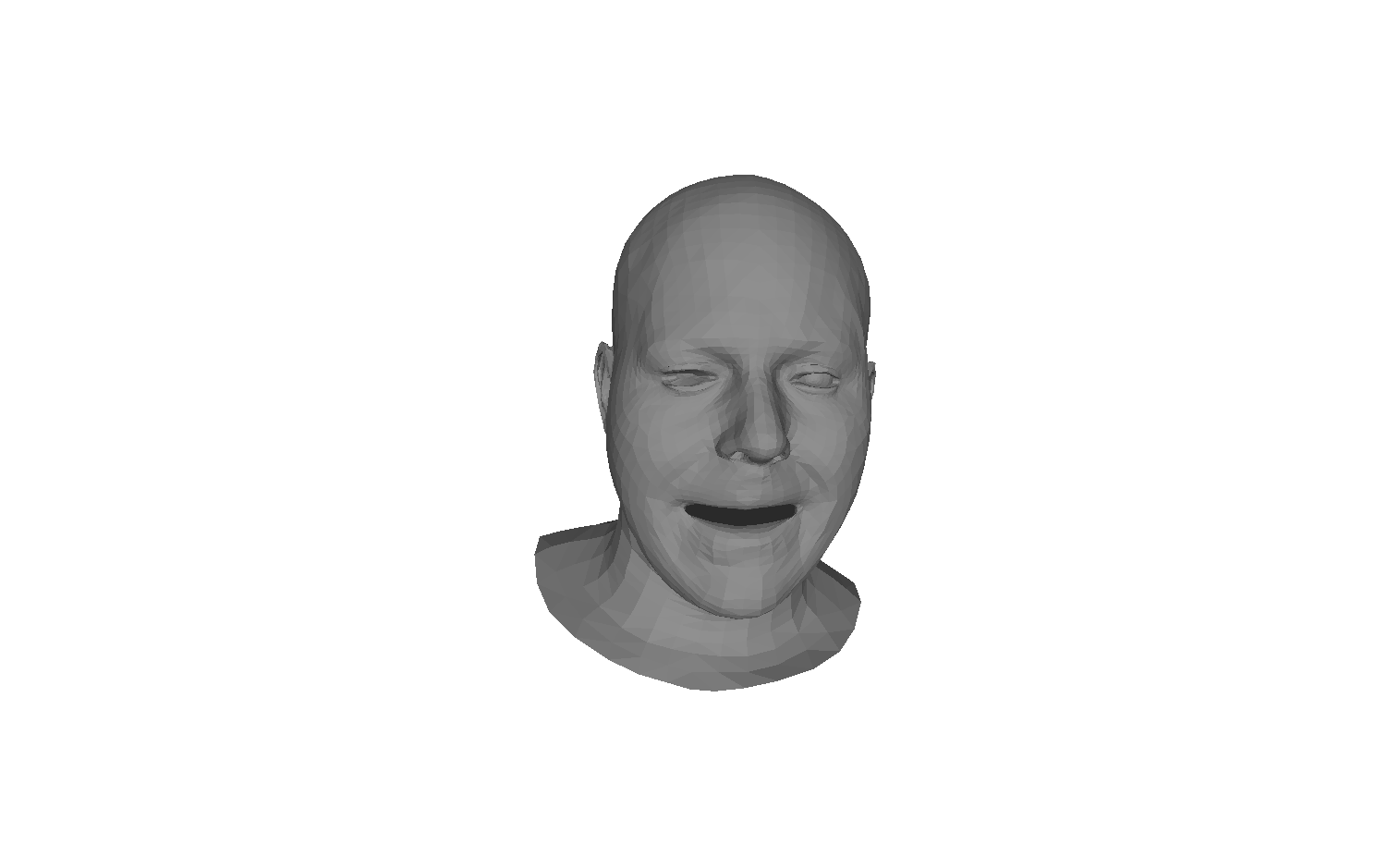}};
        
    \node[below of=c1, node distance=2.0cm] (e1) {\includegraphics[width=0.09\linewidth]{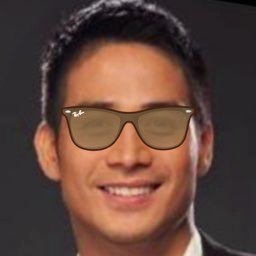}};
    \node[right of=e1, node distance=2.0cm] (e2) {\includegraphics[trim={400 80 400 100},clip,width=0.08\linewidth]{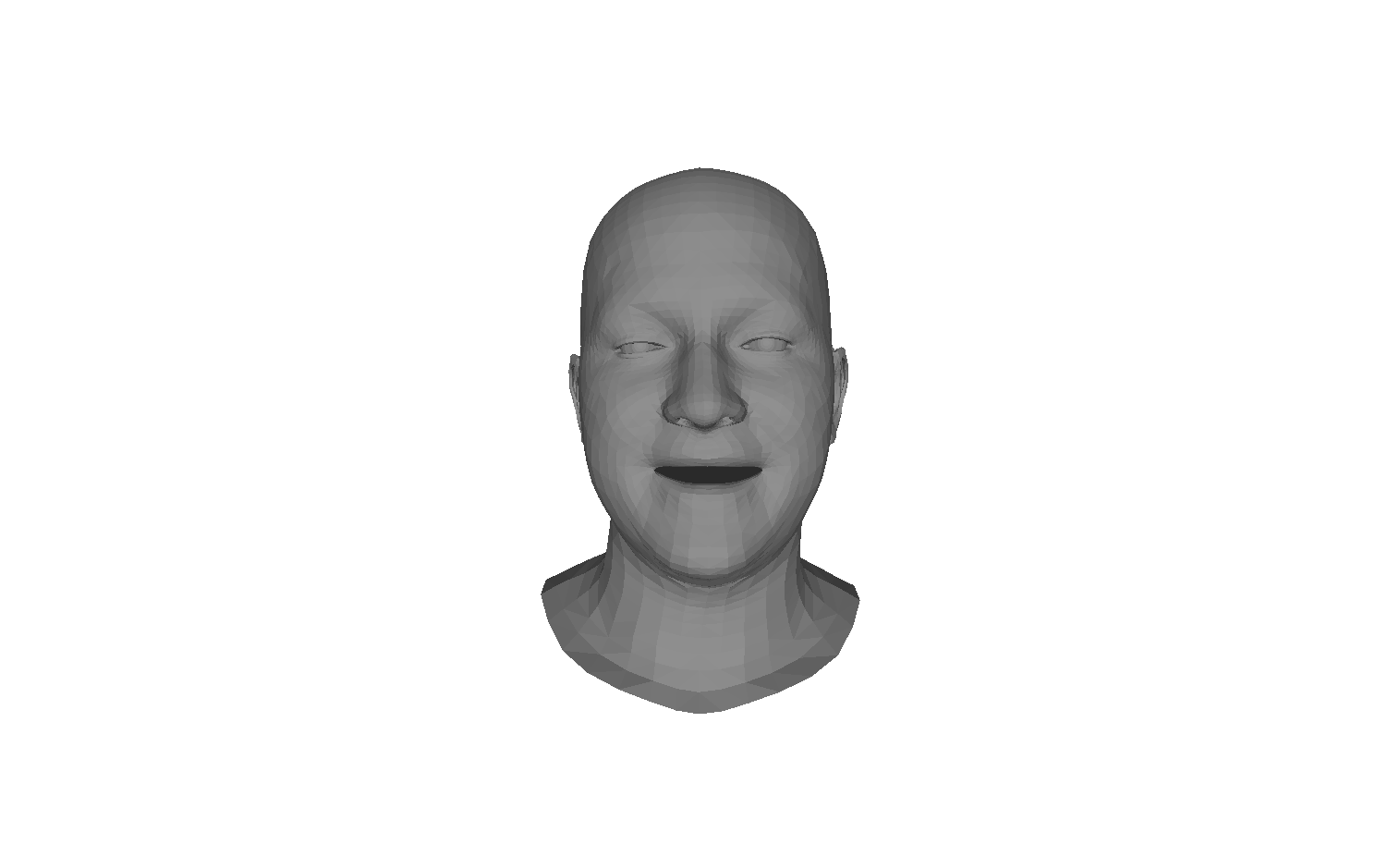}};
    \node[right of=e2, node distance=1.5cm] (e3) {\includegraphics[trim={400 80 400 100},clip,width=0.08\linewidth]{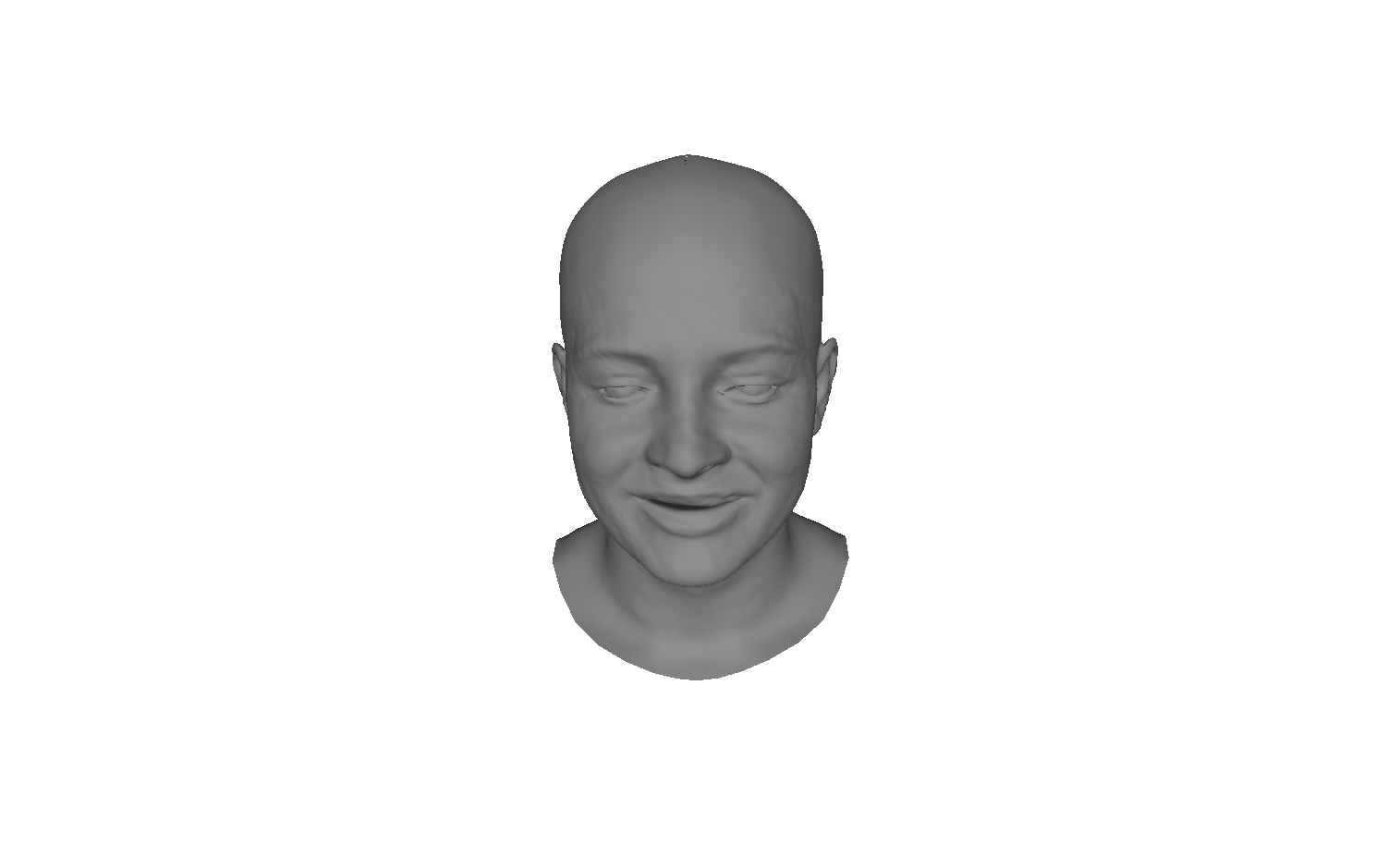}};
    \node[right of=e3, node distance=1.4cm] (e4) {\includegraphics[trim={400 80 400 100},clip,width=0.075\linewidth]{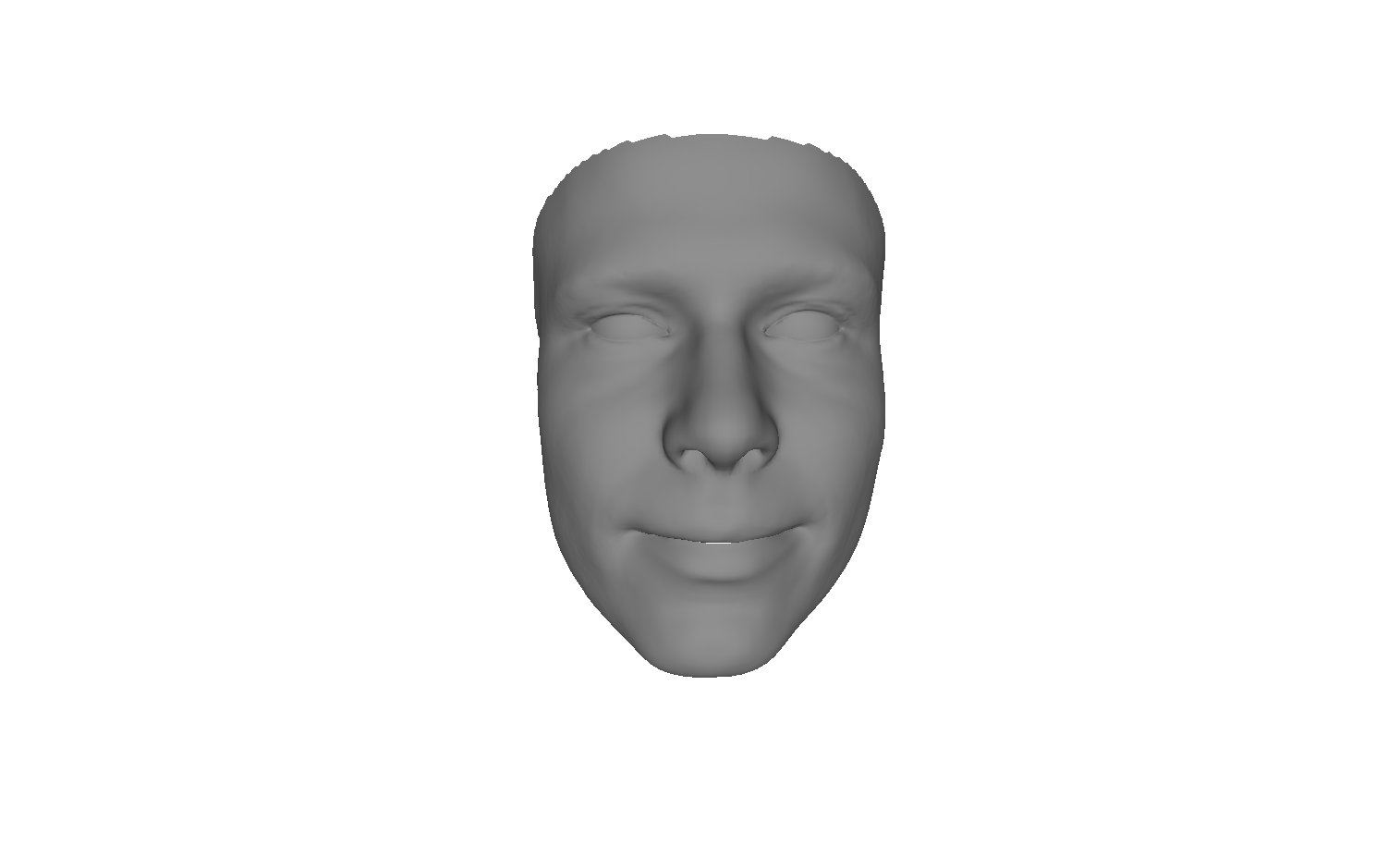}};
    \node[right of=e4, node distance=1.4cm] (e5) {\includegraphics[trim={400 80 400 100},clip,width=0.07\linewidth]{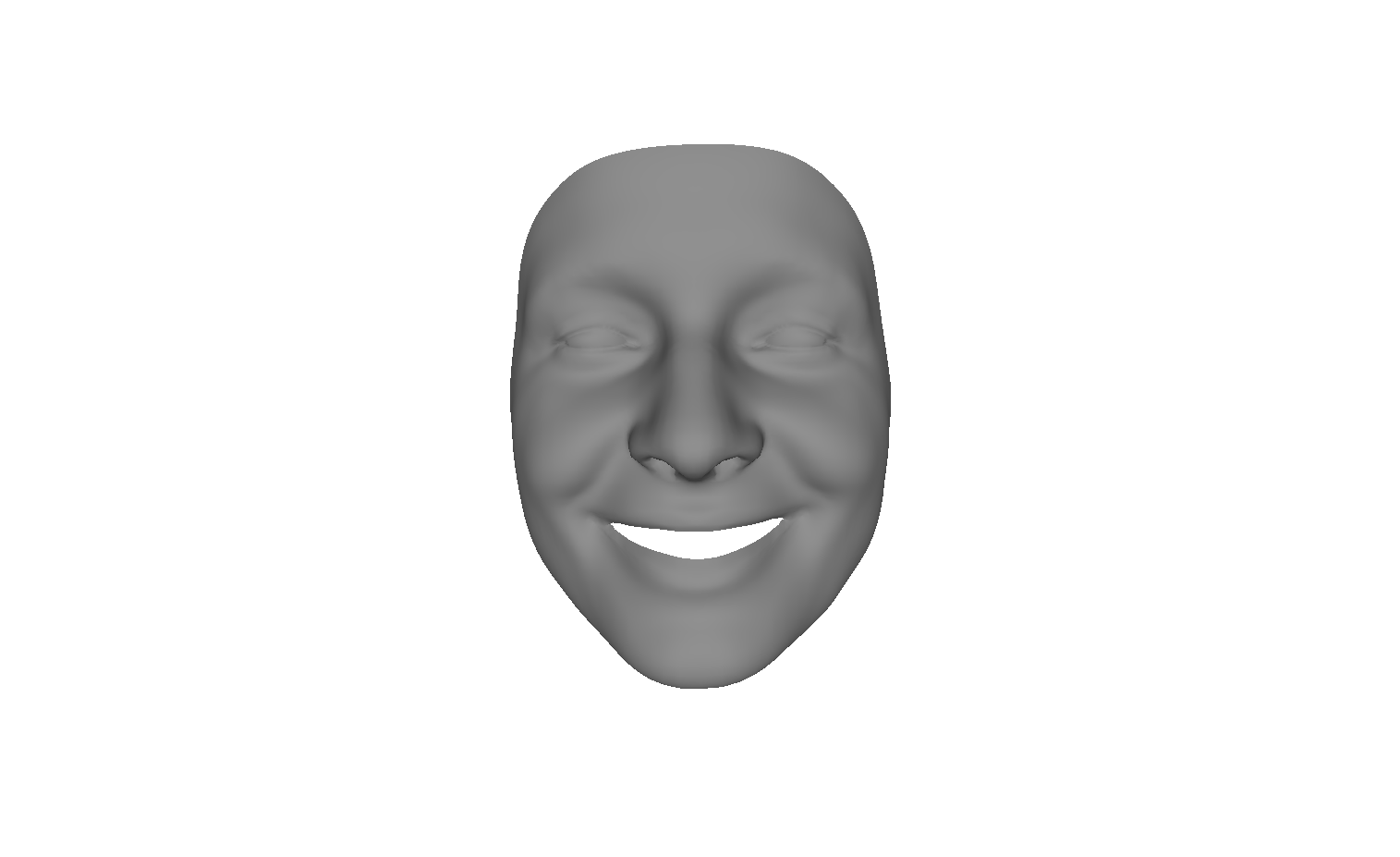}};
    \node[right of=e5, node distance=1.5cm] (e6) {\includegraphics[trim={400 80 400 100},clip,width=0.075\linewidth]{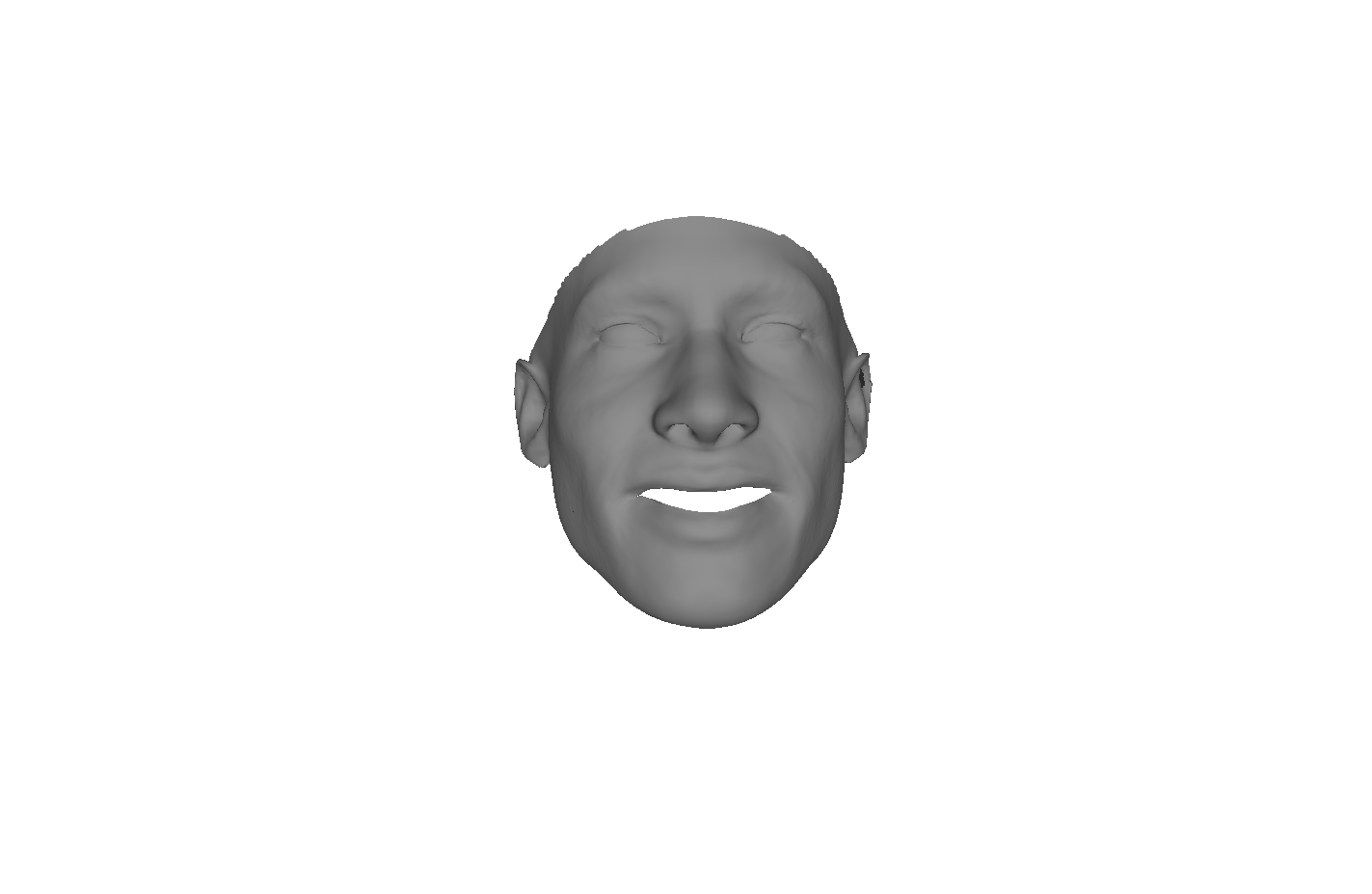}};
    \node[right of=e6, node distance=2.1cm] (e7) {\includegraphics[trim={400 80 400 100},clip,width=0.08\linewidth]{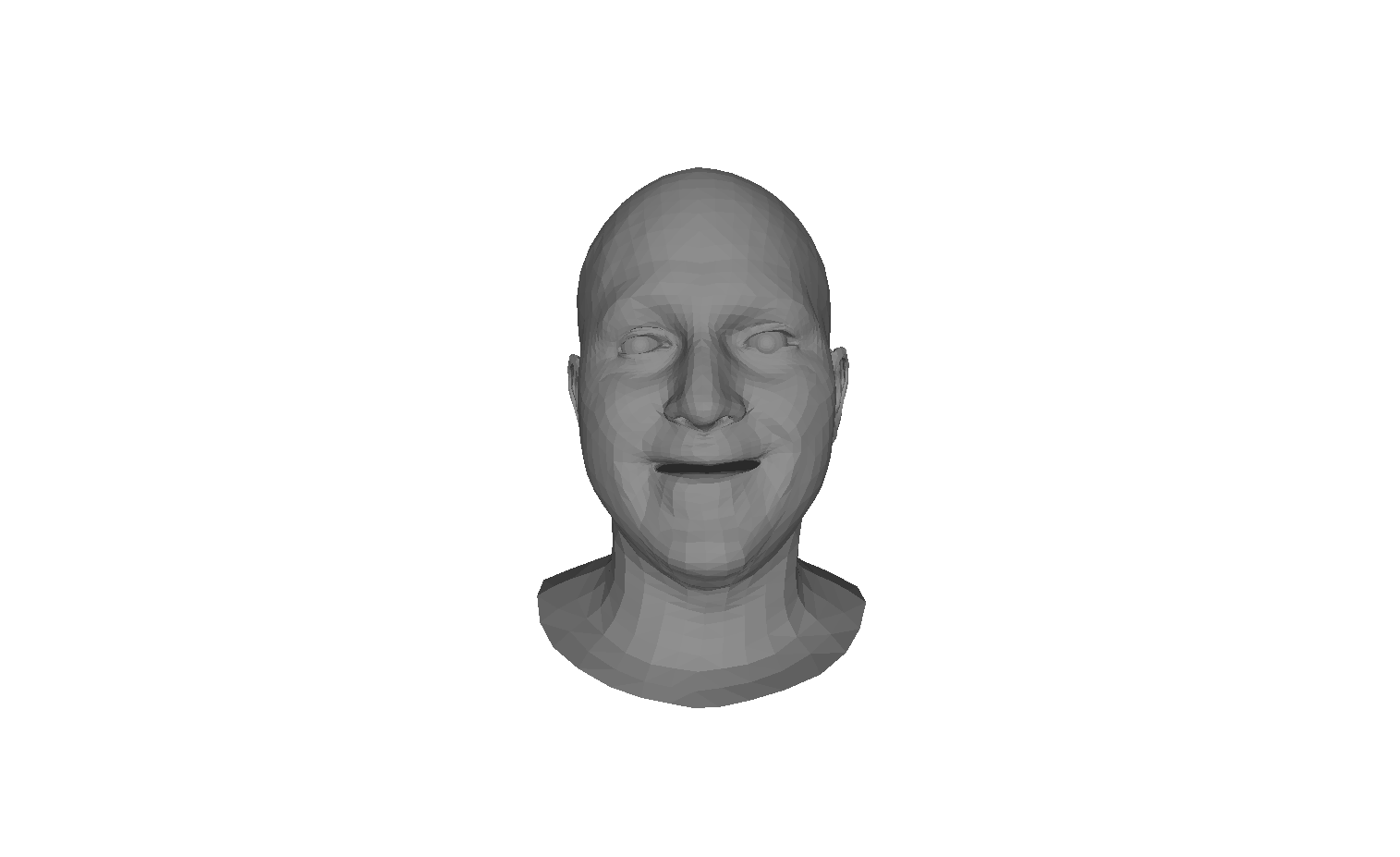}};
    \node[right of=e7, node distance=1.4cm] (e8) {\includegraphics[trim={400 80 400 100},clip,width=0.08\linewidth]{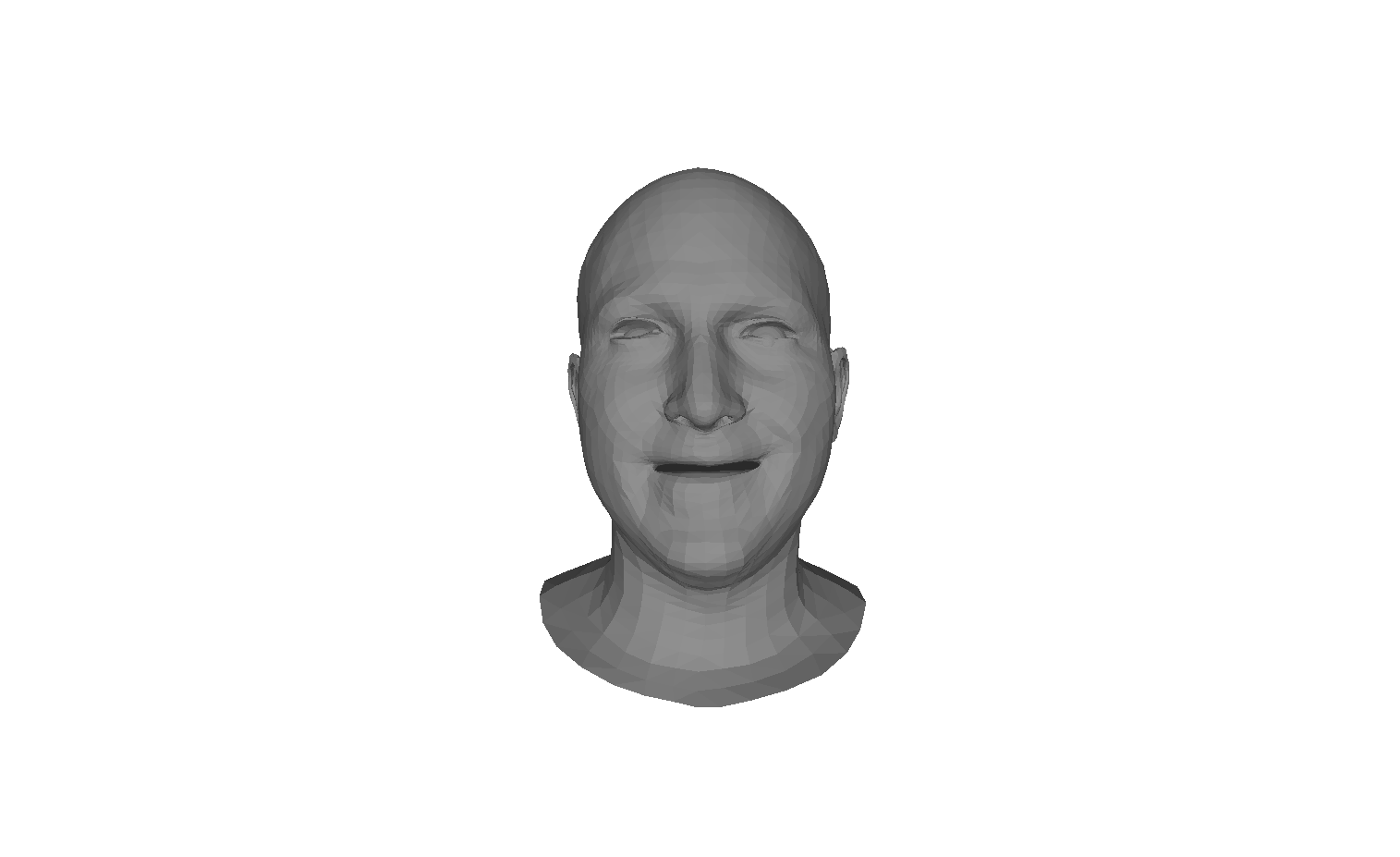}};
    \node[right of=e8, node distance=1.4cm] (e9) {\includegraphics[trim={400 80 400 100},clip,width=0.08\linewidth]{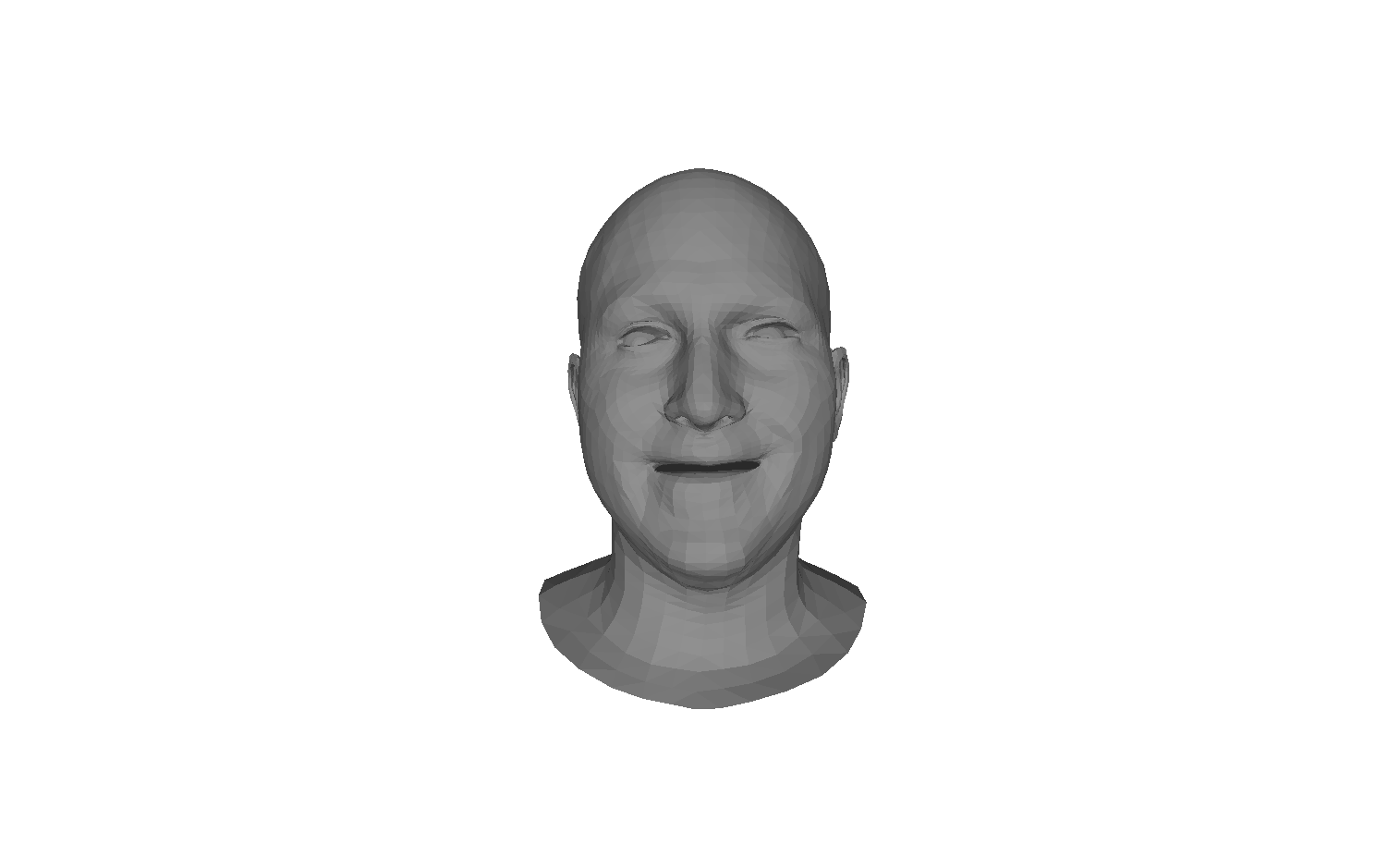}};
    \node[right of=e9, node distance=1.4cm] (e10) {\includegraphics[trim={400 80 400 100},clip,width=0.08\linewidth]{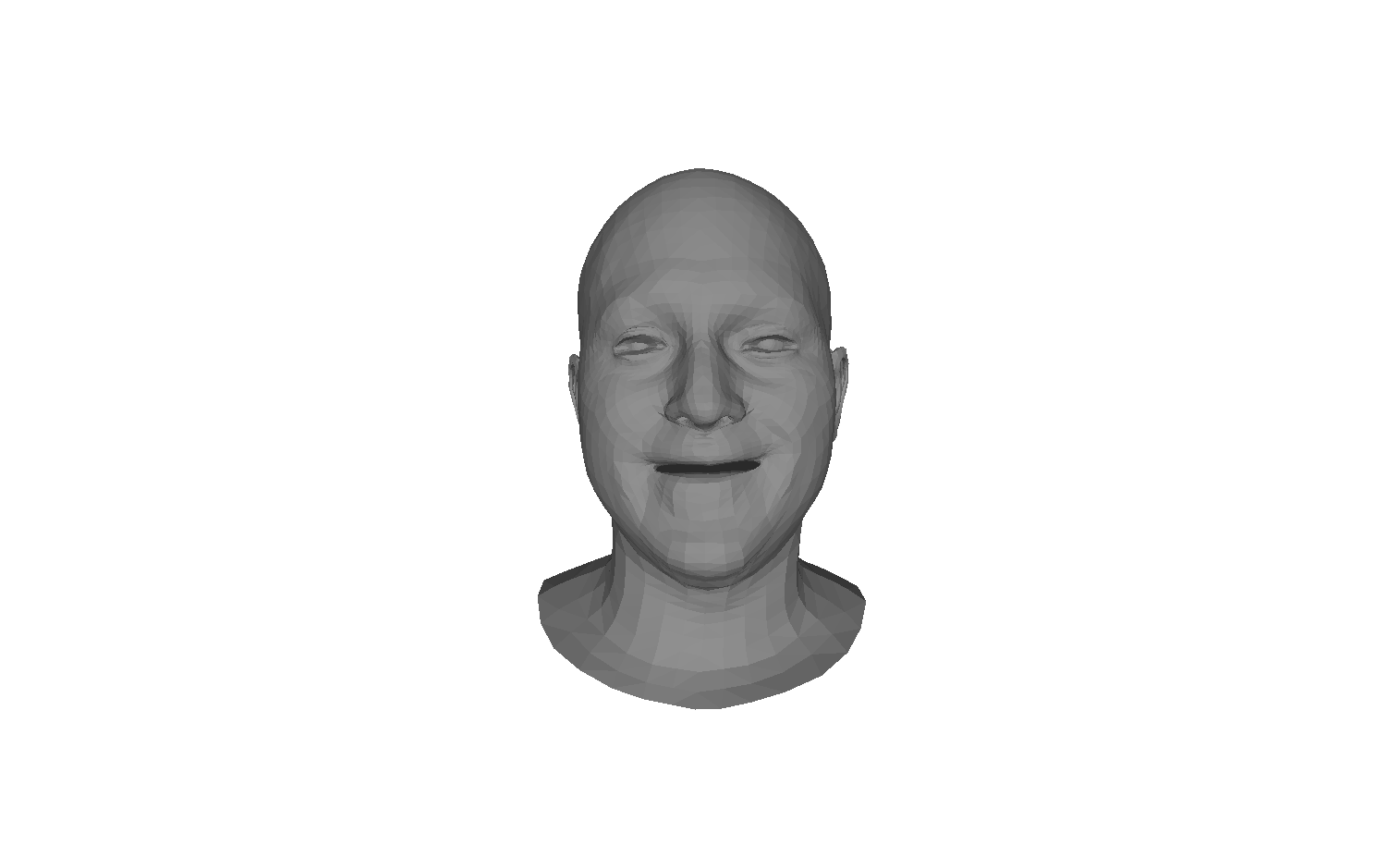}};
    \node[right of=e10, node distance=1.4cm] (e11) {\includegraphics[trim={400 80 400 100},clip,width=0.08\linewidth]{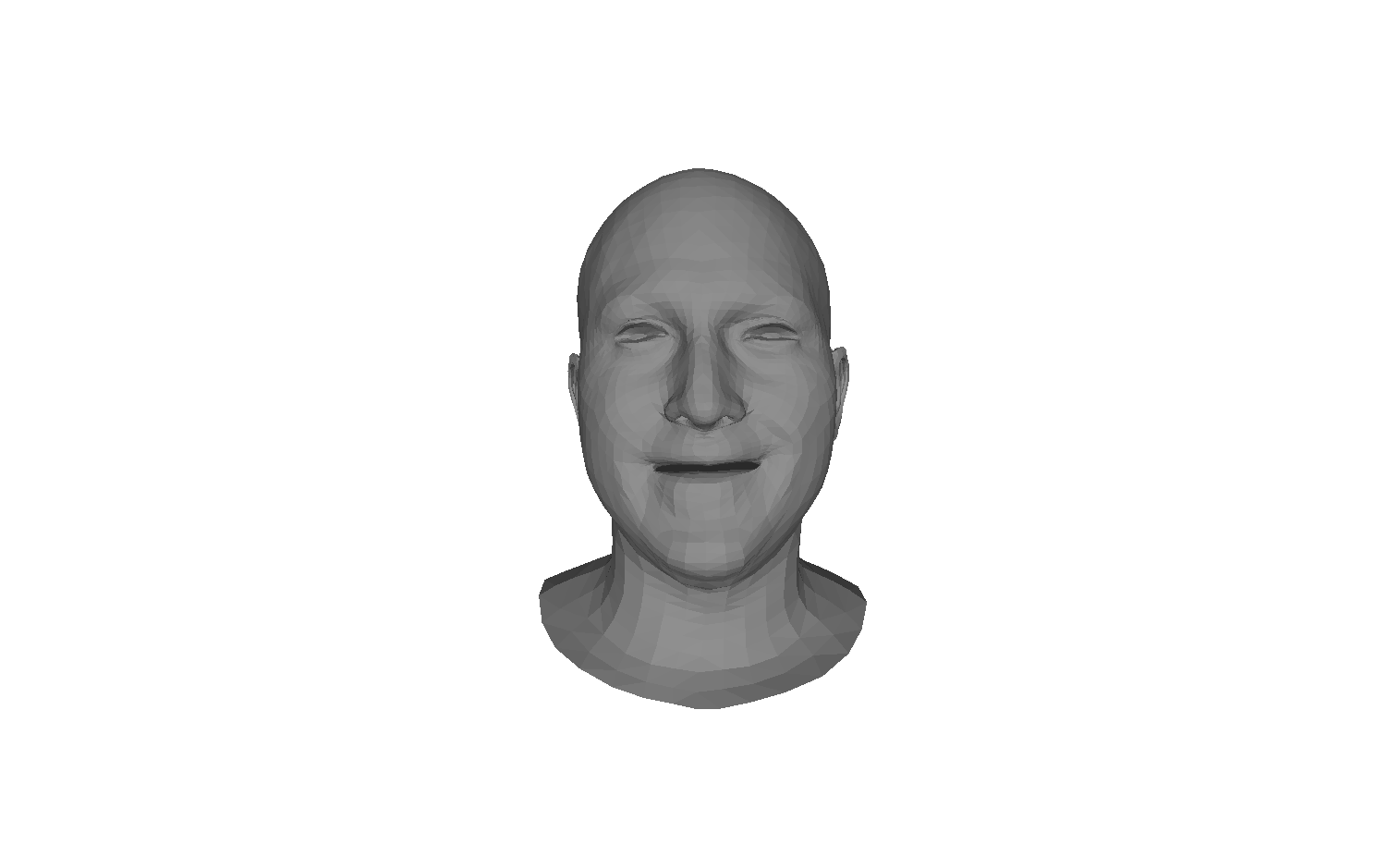}};
    
    \node[below of=e1, node distance=1.4cm] {Target Image};
    \node[below of=e2, node distance=1.4cm] {FLAME \cite{flame}};
    \node[below of=e3, node distance=1.4cm] {DECA \cite{deca}};
    \node[below of=e4, node distance=1.4cm, text width=1.2cm, align=center] {CFR-GAN \cite{occrobustwacv}};
    \node[below of=e5, node distance=1.4cm, text width=1.2cm, align=center] {Occ3DMM \cite{egger2018occlusion}};
    \node[below of=e6, node distance=1.4cm, text width=1.2cm, align=center] {Extreme3D \cite{tran2018extreme}};
    \node[below of=e9, node distance=1.4cm] {Reconstructions by \ourmethod{} (Ours)};
    \end{tikzpicture}
    \caption{\textbf{Qualitative evaluation on the CelebA dataset \cite{celeba}}: Reconstructed singular 3D meshes from the target image by the baselines \versus{} the diverse reconstructions from \ourmethod{}.}
    \label{fig:celeba}
\end{figure*}

\begin{figure}
    \centering
    \begin{tikzpicture}
    \node (a1) {\includegraphics[width=0.22\columnwidth]{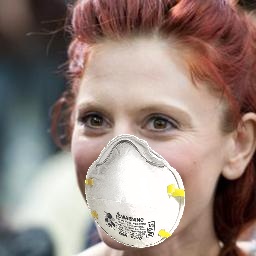}};
    \node[below of=a1, node distance=2.0cm] (a2) {\includegraphics[trim={400 80 400 100},clip,width=0.18\linewidth]{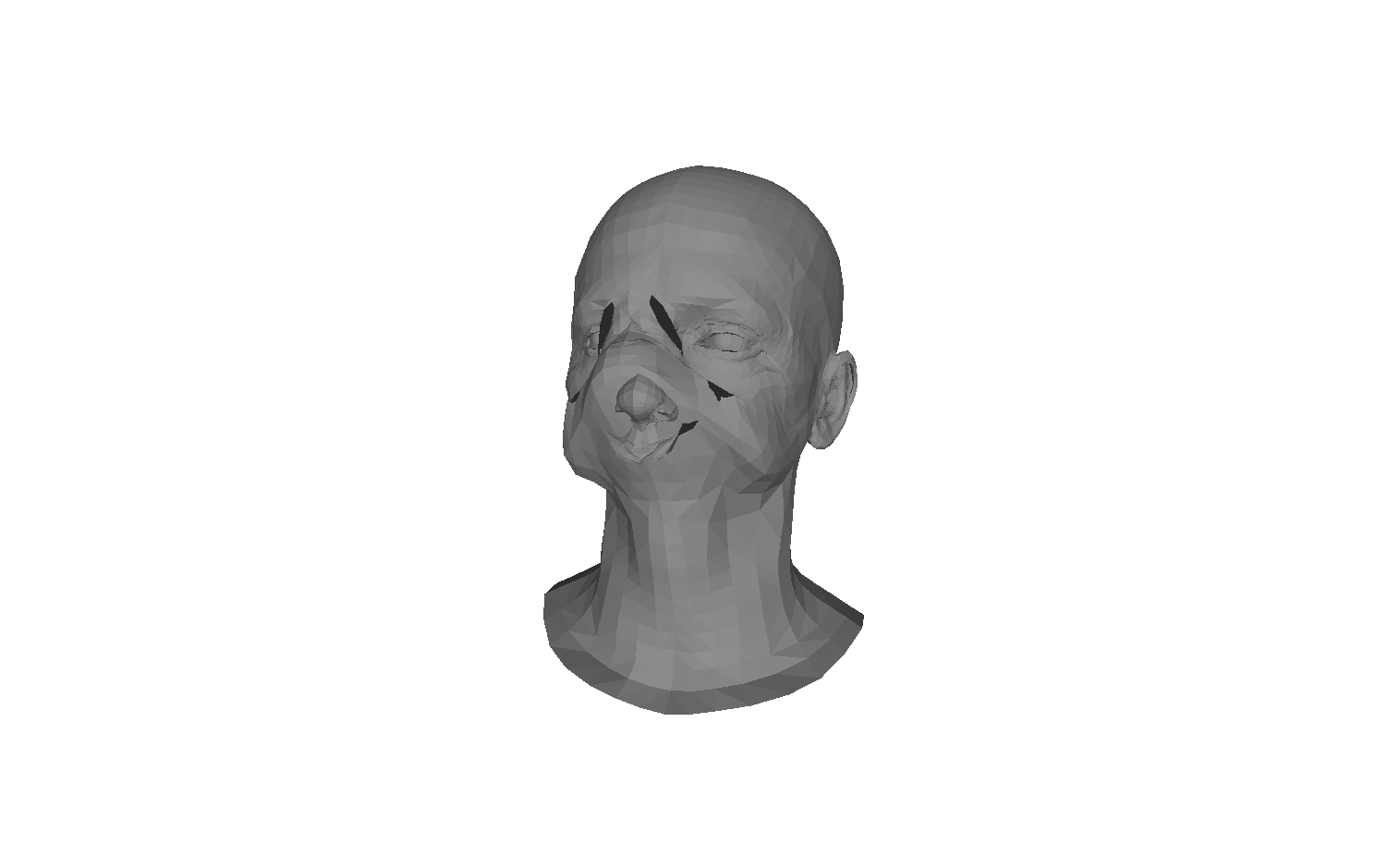}};
    \node[below of=a2, node distance=2.0cm] (a3) {\includegraphics[trim={400 80 400 100},clip,width=0.18\linewidth]{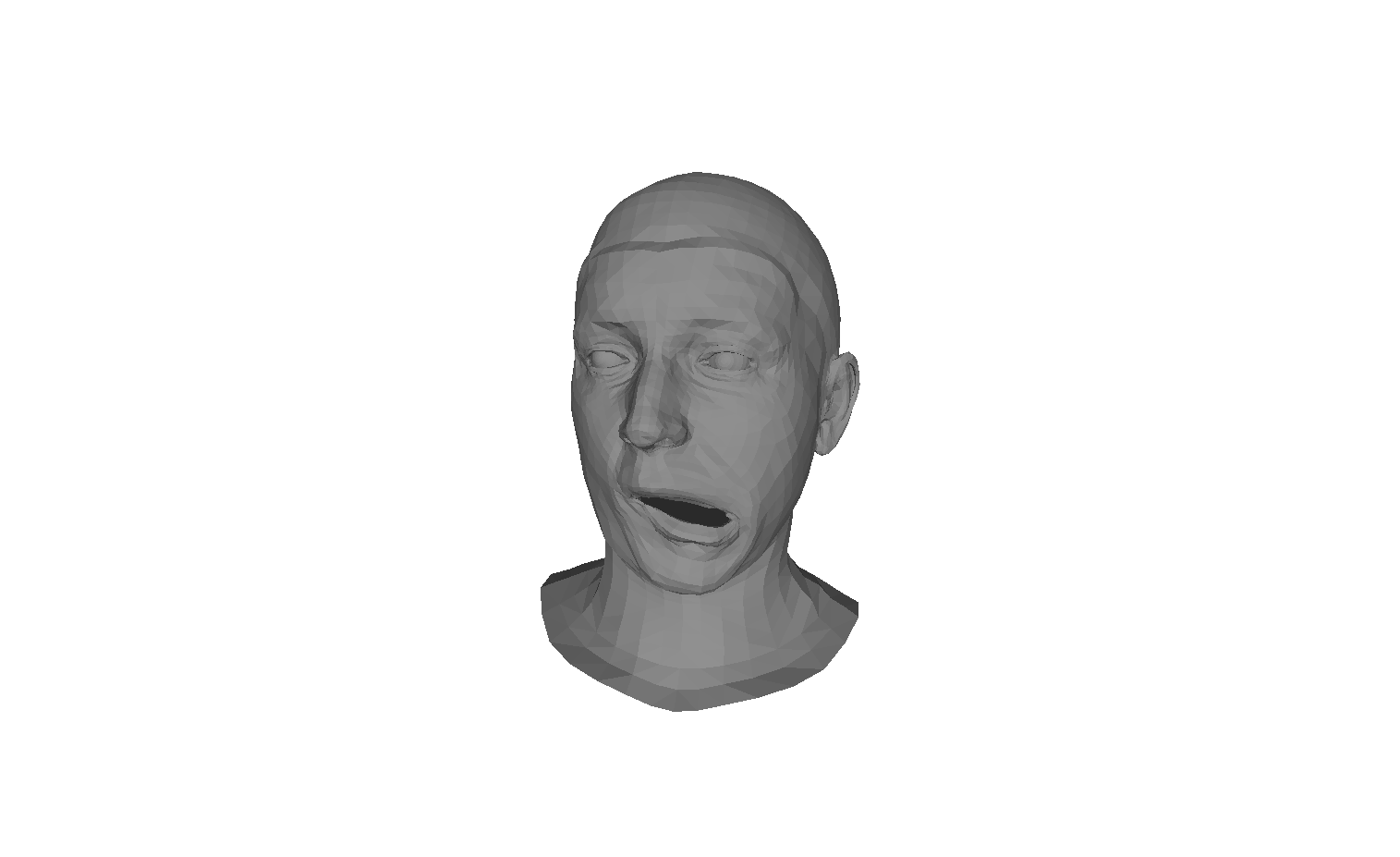}};
    
    \node[right of=a1, node distance=2.0cm] (b1) {\includegraphics[width=0.22\columnwidth]{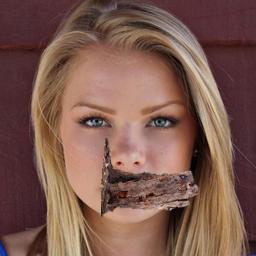}};
    \node[below of=b1, node distance=2.0cm] (b2) {\includegraphics[trim={400 80 400 100},clip,width=0.18\linewidth]{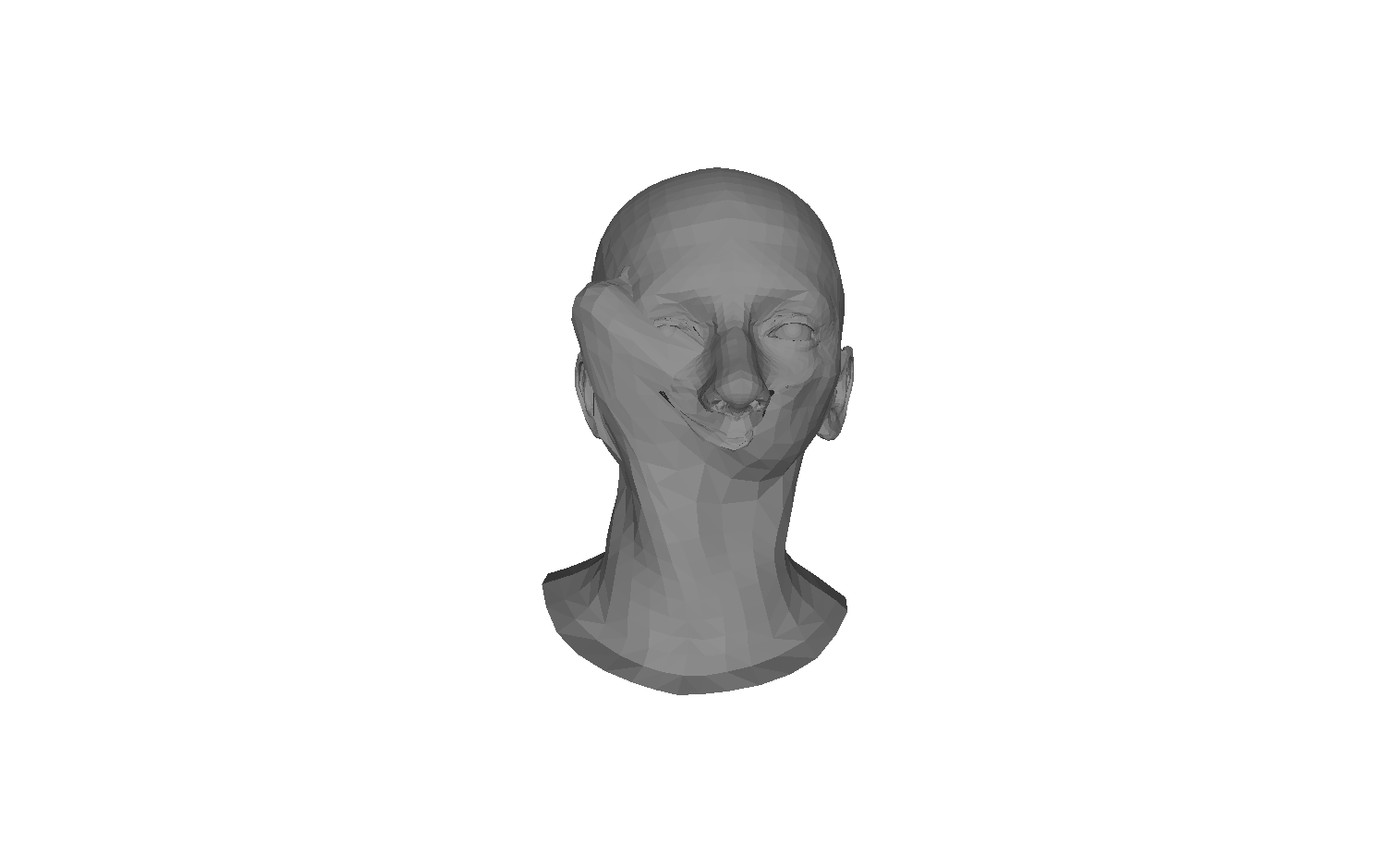}};
    \node[below of=b2, node distance=2.0cm] (b3) {\includegraphics[trim={400 80 400 100},clip,width=0.18\linewidth]{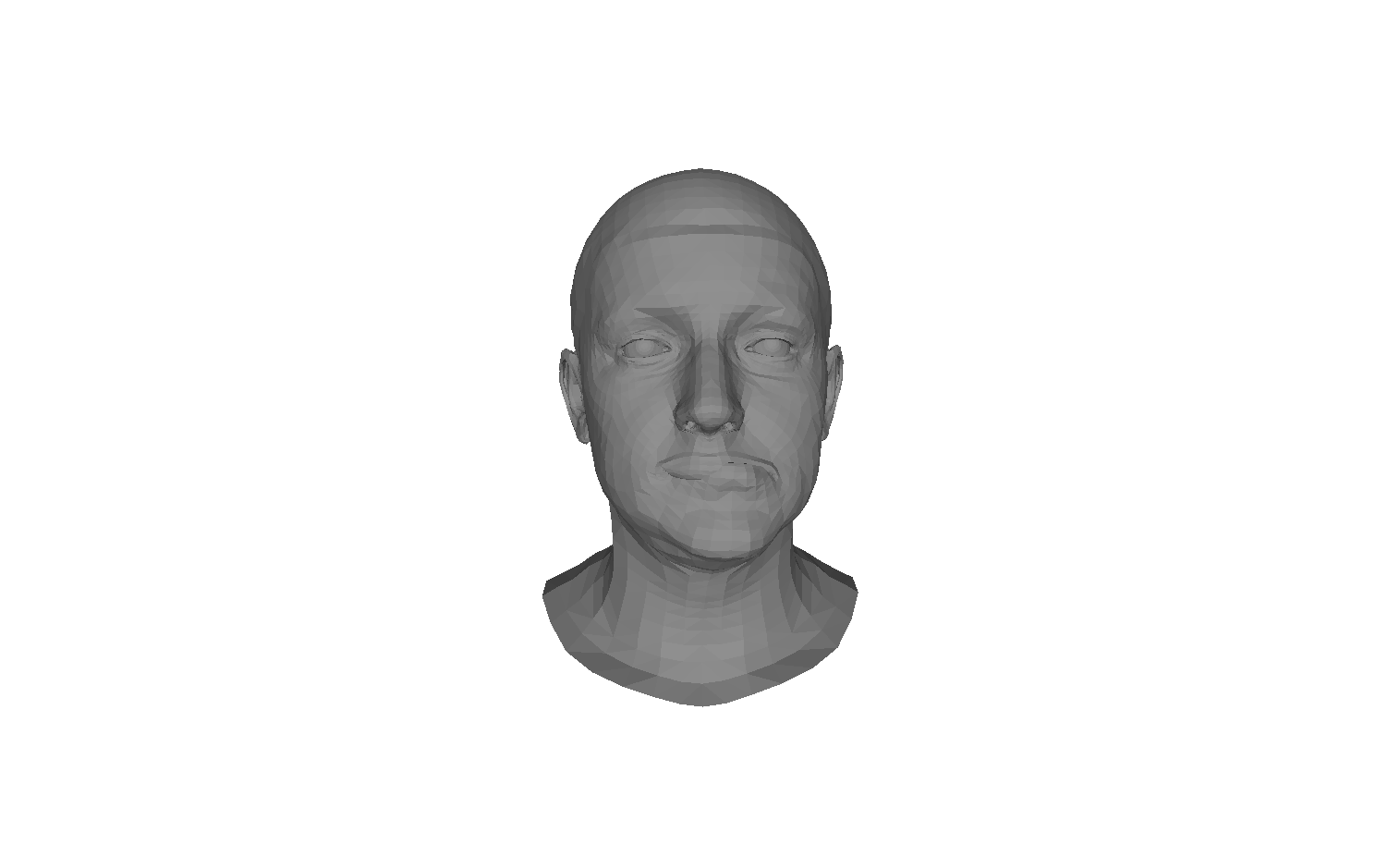}};
    
    \node[right of=b1, node distance=2.0cm] (c1) {\includegraphics[width=0.22\columnwidth]{figs/celeba/masked/000041/000041_N95.jpg}};
    \node[below of=c1, node distance=2.0cm] (c2) {\includegraphics[trim={400 80 400 100},clip,width=0.18\linewidth]{figs/celeba/masked/000041/flame.png}};
    \node[below of=c2, node distance=2.0cm] (c3) {\includegraphics[trim={400 80 400 100},clip,width=0.18\linewidth]{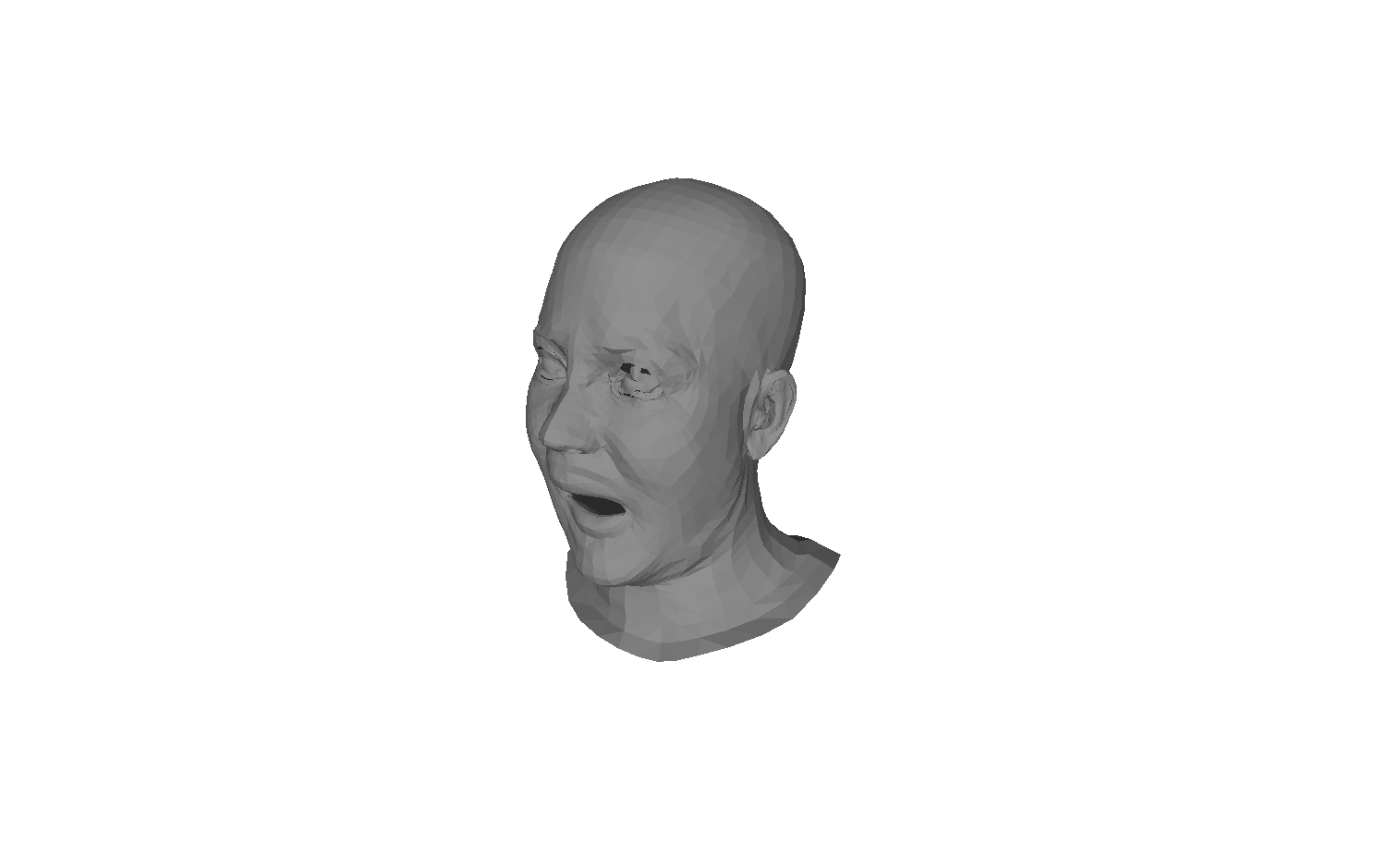}};
    
    \node[right of=c1, node distance=2.0cm] (d1) {\includegraphics[width=0.22\columnwidth]{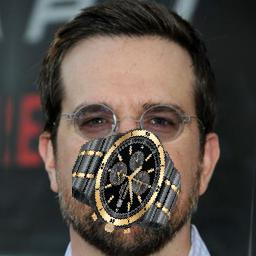}};
    \node[below of=d1, node distance=2.0cm] (d2) {\includegraphics[trim={400 80 400 100},clip,width=0.18\linewidth]{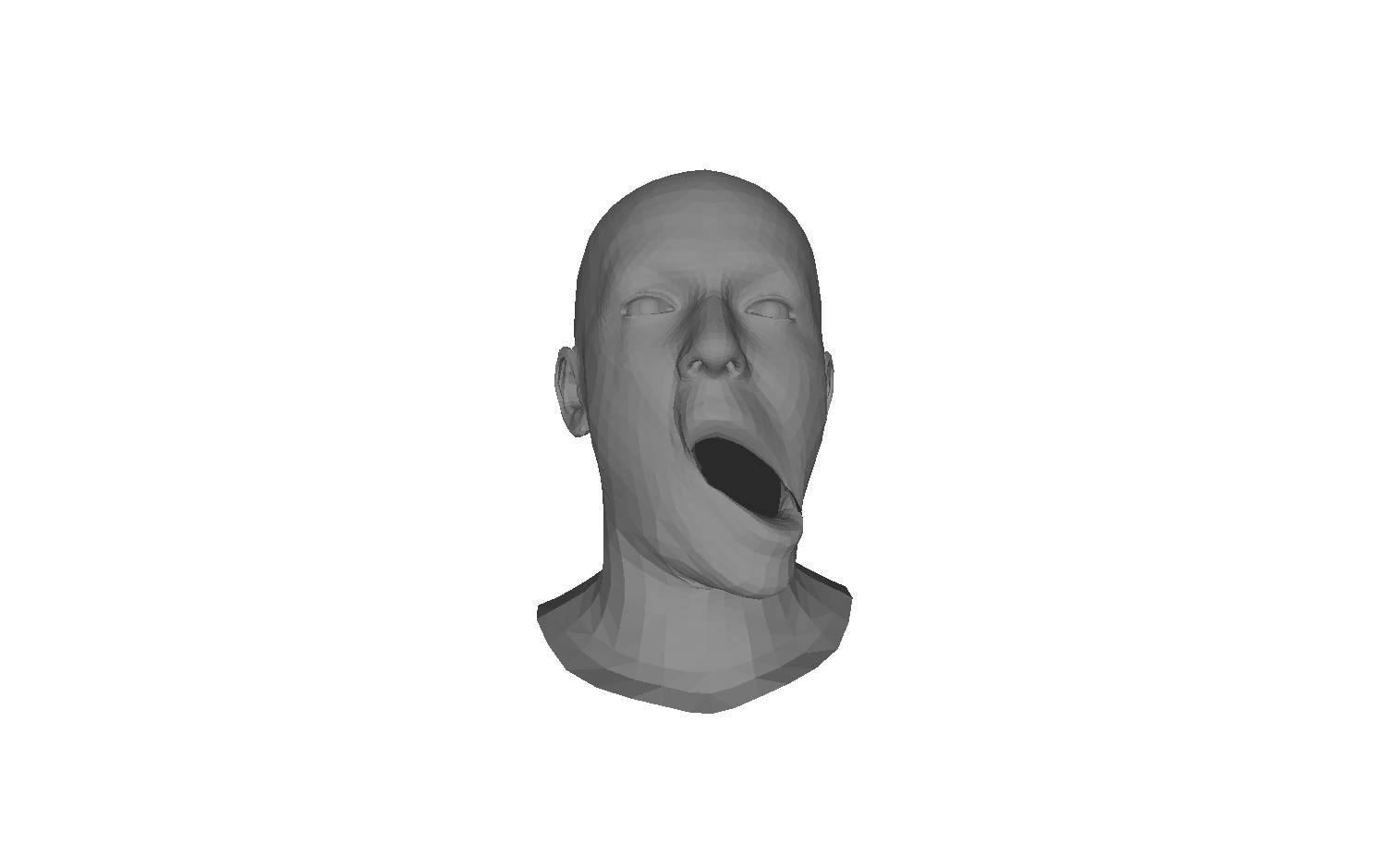}};
    \node[below of=d2, node distance=2.0cm] (d3) {\includegraphics[trim={400 80 400 100},clip,width=0.18\linewidth]{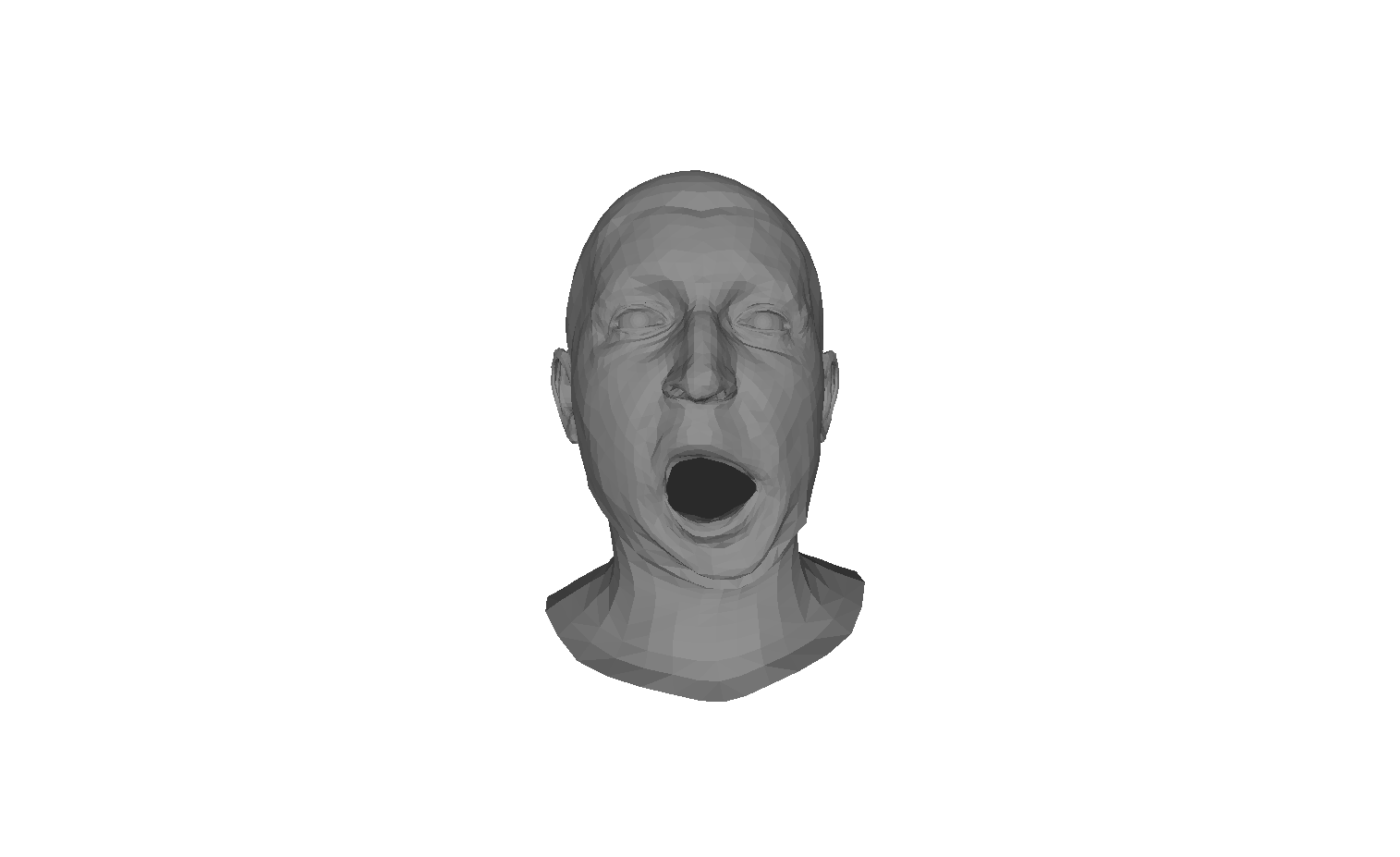}};
       
    \end{tikzpicture}
    \caption{FLAME \cite{flame} based fitting (middle row) \versus{} our Global+Local fitting (last row) on occluded face images (top row).}
    \label{fig:flame_vs_component}
\end{figure}

\subsection{Qualitative Results}
\cref{fig:coma} shows qualitative results of 3D reconstruction on the artificially occluded CoMA \cite{coma} images. All the baselines can only generate a single 3D reconstruction w.r.t the target image. We observe that the reconstructions generated by \ourmethod{} look diverse yet plausible and visually more faithful to the ground truth in the visible regions. In comparison, FLAME-based fitting \cite{flame}, and DECA \cite{deca} do not explicitly handle occlusions and generate soft and erroneous shapes. CFR-GAN \cite{occrobustwacv} and Occ-3DMM \cite{egger2018occlusion} get the pose wrong in multiple instances. Extreme3D \cite{tran2018extreme} generates visually better reconstructions of the visible parts of the face but gets the expression wrong in the second row. In ~\cref{fig:celeba}, we show further visual comparisons on the occlusion-augmented images from the CelebA \cite{celeba} dataset. Note that we do not have ground truth scans for these images. However, visual results suggest that the baselines, by being holistic models, do not explicitly exclude features from the occluded regions and often get incorrect poses and expressions on these images. Meanwhile, the reconstructions from \ourmethod{} look diverse on the occluded regions yet consistent w.r.t to the visible parts of the face.

\vspace{10pt}
\noindent \textbf{FLAME vs Global+Local PCA Model: } In addition to the quantitative comparison done in ~\cref{tab:fitting}, we qualitatively compare the occlusion robustness of the global FLAME \cite{flame} model \versus{} our global+local model. In ~\cref{fig:flame_vs_component}, we show some failure cases of the FLAME \cite{flame} based fitting on severely occluded images. Notice the severe deformations on the FLAME \cite{flame} fitted outputs, especially around the mouth. In contrast, the fittings by our global+local models look more faithful and detailed with respect to the visible parts. These observations further support our claim that a global+local model-based fitting performs better than a global-model based fitting on occluded face images.

\section{Conclusion} \label{sec:conclusion}
We proposed \ourmethod{}, an approach to reconstruct diverse yet plausible 3D reconstructions corresponding to a single occluded face image. Our approach was motivated by the fact that, in the presence of occlusions, a distribution of plausible 3D reconstructions is more desirable than a single unique solution. We proposed a three-step solution that first fits a robust partial shape using an ensemble of global+local PCA models, maps it to a latent space, and iteratively optimizes the embeddings to promote diversity in the occluded parts while retaining fidelity with respect to the visible parts of the face. Experimental evaluation across multiple occlusion types and datasets show the efficacy of \ourmethod{}, both in terms of robustness and diversity, compared to multiple baselines. To our knowledge, this is the first approach that generates a distribution of diverse 3D reconstructions of a single occluded face image.

A limitation of the proposed approach is its dependence on the robustness of the global+local fitting in the first step for further diverse completions. Although such a locally disentangled fitting demonstrably performs better than a global model fitting, it may still be affected in cases where the initial landmark or face-mask estimates are wrong.

\appendix
\appendixpage
\section{Further Experiments}\label{sec:experiments}

\begin{figure}
    \centering
    \includegraphics[width=\columnwidth]{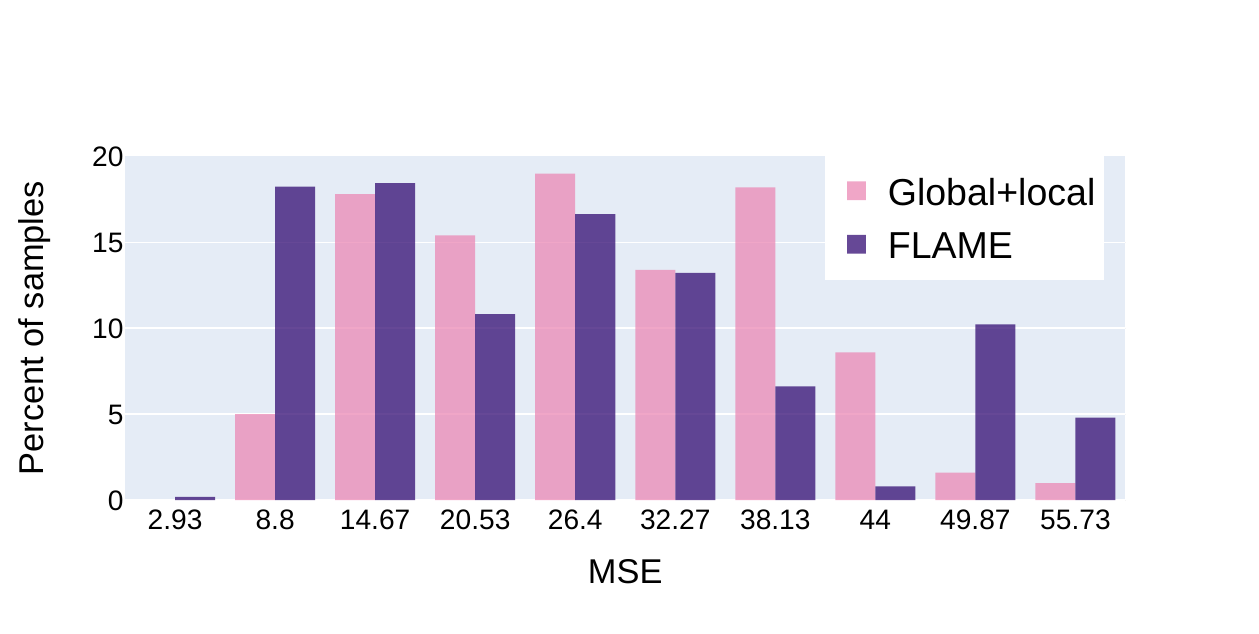}
    \caption{Histogram of MSE for shape fitting on occluded face images by FLAME \cite{flame} and our Global+local model.}
    \label{fig:histogram}
\end{figure}

\begin{table*}
    \centering
    \setlength{\tabcolsep}{2pt}
    \resizebox{\textwidth}{!}{
    \begin{tabular}{c|ccc|ccc|ccc|ccc}
        \hline
        \textbf{Occlusion} & \multicolumn{3}{c|}{FLAME+DPP} & \multicolumn{3}{c|}{Global+Local+DPP} & \multicolumn{3}{c|}{Gloal+Local+VAE} & \multicolumn{3}{c}{\ourmethod{} (Ours)}\\
        \textbf{Type} & \textbf{ASD-V} $(\downarrow)$ & \textbf{ASD-O} $(\uparrow)$ & $\mathbf{\frac{ASD-O}{ASD-V}} (\uparrow)$ & \textbf{ASD-V} $(\downarrow)$ & \textbf{ASD-O} $(\uparrow)$ & $\mathbf{\frac{ASD-O}{ASD-V}} (\uparrow)$ & \textbf{ASD-V} $(\downarrow)$ & \textbf{ASD-O} $(\uparrow)$ & $\mathbf{\frac{ASD-O}{ASD-V}} (\uparrow)$ & \textbf{ASD-V} $(\downarrow)$ & \textbf{ASD-O} $(\uparrow)$ & $\frac{ASD-O}{ASD-V} (\uparrow)$\\
        \hline
        Glasses & 3.44 & 2.98 & 0.866 & 2.15 & 2.99 & 1.391 & 0.81 & 1.17 & 1.444 & \textbf{0.68} & \textbf{3.56} & \textbf{5.235}\\
        Face-mask & 3.45 & 4.93 & 1.429 & 2.85 & 3.99 & 1.400 & \textbf{0.75} & 1.62 & 2.160 & 1.03 & \textbf{7.47} & \textbf{7.252}\\
        Random & 4.12 & 4.23 & 1.027 & 3.17 & 3.84 & 1.211 & \textbf{0.79} & 1.29 & 1.633 & 0.83 & \textbf{4.30} & \textbf{5.181}\\
        \hline
        Overall & 3.86 & 4.44 & 1.150 & 3.03 & 3.88 & 1.281 & \textbf{0.78} & 1.41 & 1.808 & 0.90 & \textbf{5.41} & \textbf{6.011}\\
        \hline
    \end{tabular}
    }
    \caption{Quantitative evaluation of the diversity in 3D reconstruction of occluded faces from the CelebA dataset \cite{celeba} between the baselines \versus{} \ourmethod{} in terms of the ASD-V and ASD-O metrics (in order of $10^{-3}$) and the ratio between them.}
    \label{tab:diversity}
\end{table*}

\begin{table*}
    \centering
    \begin{subtable}[t]{0.48\textwidth}
    \centering
    \resizebox{0.98\textwidth}{!}{
    \begin{tabular}{|l|c|c|c|c|c|}
        \hline
         \backslashbox{$\bm{k}$}{$\bm{n_{\sigma}}$} & \textbf{1} & \textbf{2} & \textbf{3} & \textbf{4} & \textbf{5} \\
         \hline
         \textbf{0.1} & 0.53 & 0.81 & 0.93 & 1.40 & 1.88\\
         \hline
         \textbf{0.25} & 0.69 & 0.95 & 1.18 & 1.61 & 1.98\\
         \hline
         \textbf{0.5} & 0.86 & 1.02 & 1.30 & 1.94 & 2.14\\
         \hline
         \textbf{1} & 0.81 & 1.05 & 1.23 & 1.92 & 2.03\\
         \hline
         \textbf{2} & 0.79 & 0.98 & 1.06 & 1.57 & 1.98\\
         \hline
    \end{tabular}}
    \caption{ASD-V $(\downarrow)$\label{tab:hyper_asdv}}
    \end{subtable} \hfill
    \begin{subtable}[t]{0.5\textwidth}
    \centering
    \resizebox{0.98\textwidth}{!}{
    \begin{tabular}{|l|c|c|c|c|c|}
        \hline
         \backslashbox{$\bm{k}$}{$\bm{n_{\sigma}}$} & \textbf{1} & \textbf{2} & \textbf{3} & \textbf{4} & \textbf{5} \\
         \hline
         \textbf{0.1} & 3.63 & 4.92 & 5.62 & 7.17 & 8.64\\
         \hline
         \textbf{0.25} & 4.13 & 6.37 & 7.65 & 8.18 & 10.73\\
         \hline
         \textbf{0.5} & 5.98 & 8.25 & 9.16 & 11.19 & 14.53\\
         \hline
         \textbf{1} & 5.18 & 7.89 & 8.84 & 10.72 & 12.96\\
         \hline
         \textbf{2} & 4.42 & 6.68 & 7.40 & 9.78 & 12.21\\
         \hline
    \end{tabular}}
    \caption{ASD-O $(\uparrow)$\label{tab:hyper_asdo}}
    \end{subtable}
    \caption{Effect of the hyperparameters $k$ and $n_{\sigma}$ on the diversity metrics \textit{ASD-V} and \textit{ASD-O} on the CoMA dataset \cite{coma}.}
    \label{tab:diversity_hparams}
\end{table*}

\begin{figure*}
    \centering
    \begin{tikzpicture}
    \node (a1) {\includegraphics[width=0.11\linewidth]{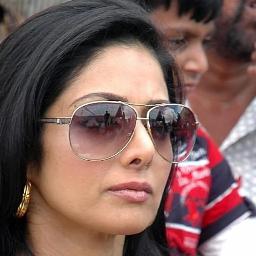}};
    \node[right of=a1, node distance=2.5cm] (a2) {\includegraphics[trim={400 80 400 100},clip,width=0.09\linewidth]{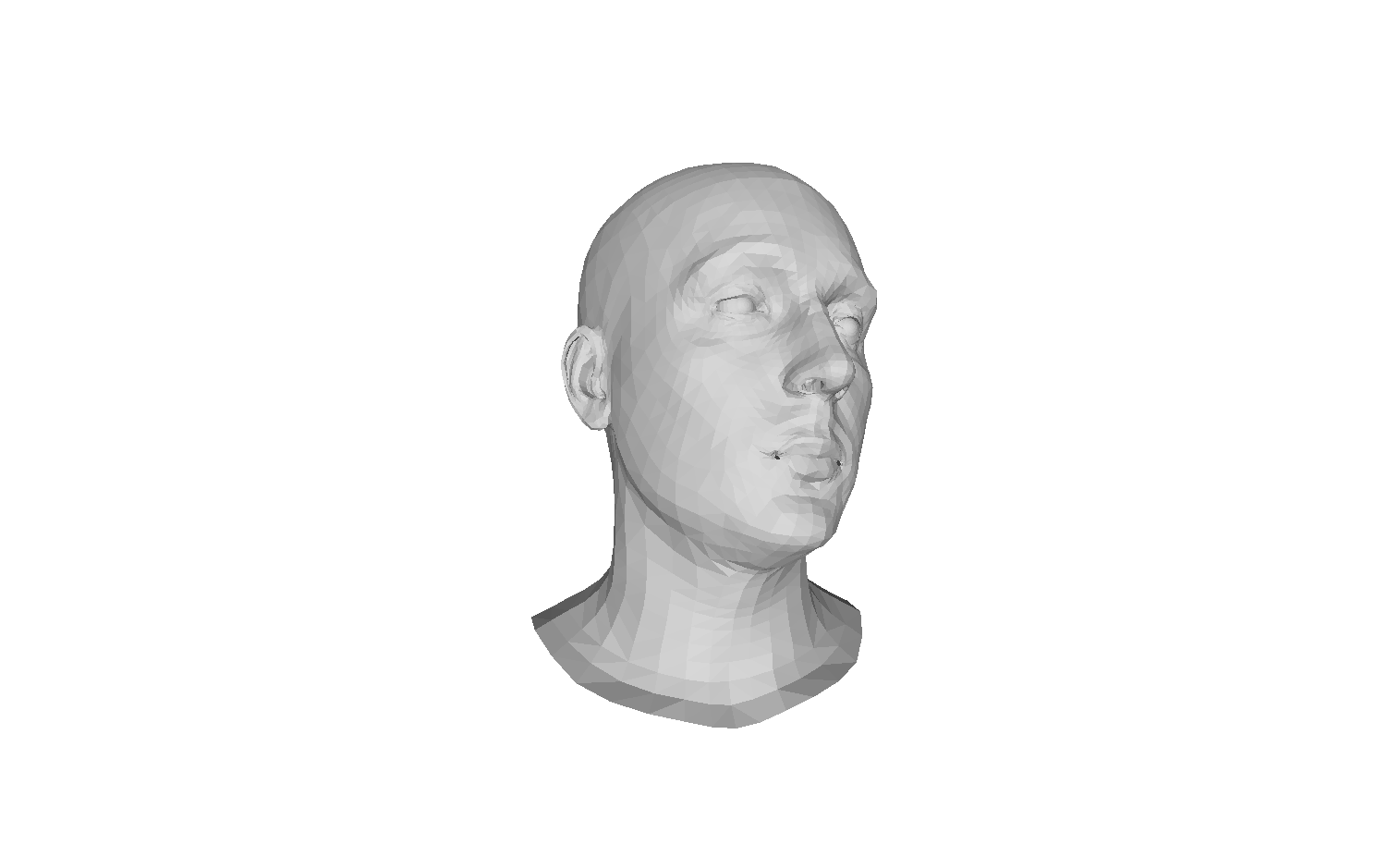}};
    \node[right of=a2, node distance=2.3cm] (a3) {\includegraphics[trim={400 80 400 100},clip,width=0.09\linewidth]{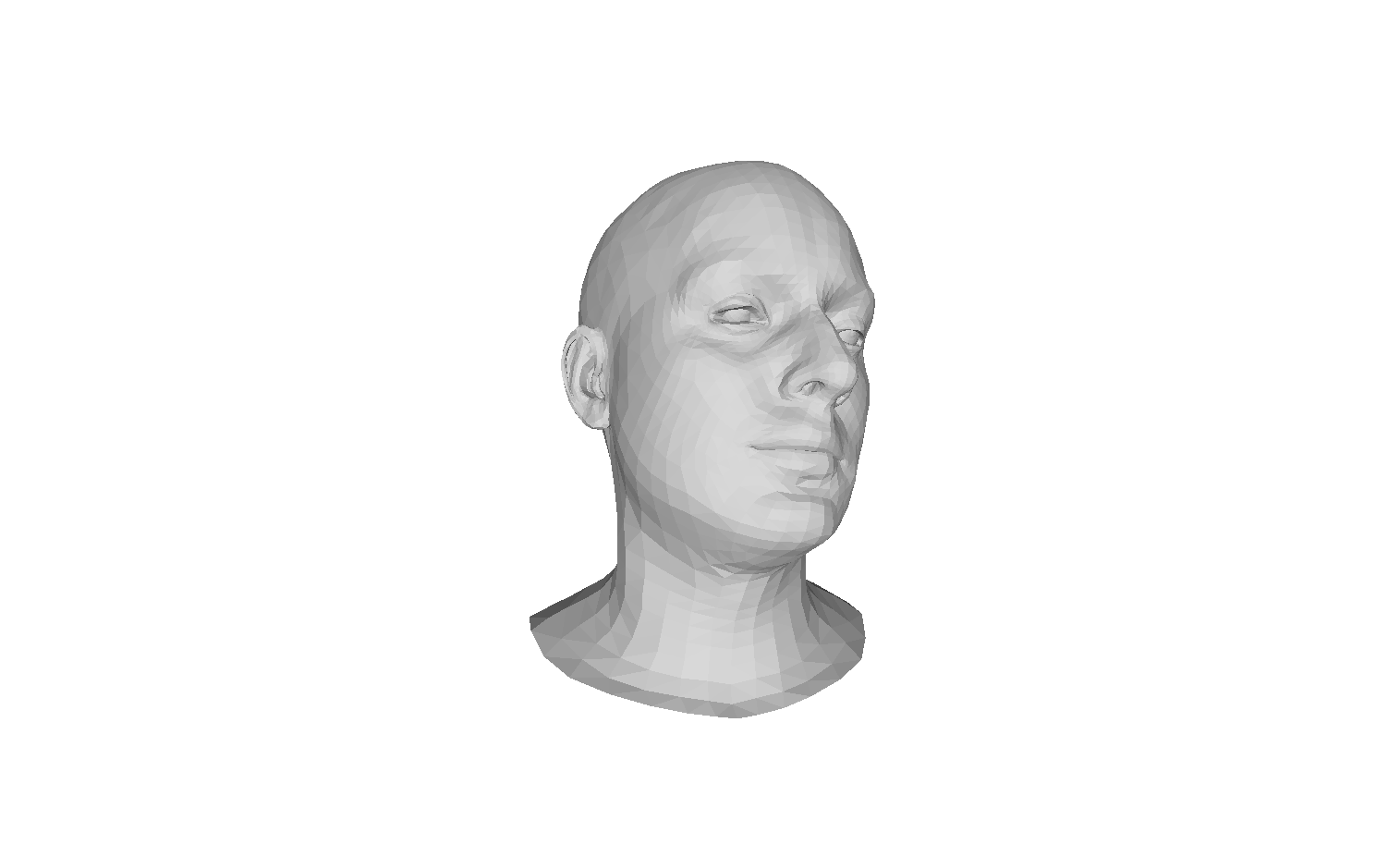}};
    \node[right of=a3, node distance=1.6cm] (a4) {\includegraphics[trim={400 80 400 100},clip,width=0.09\linewidth]{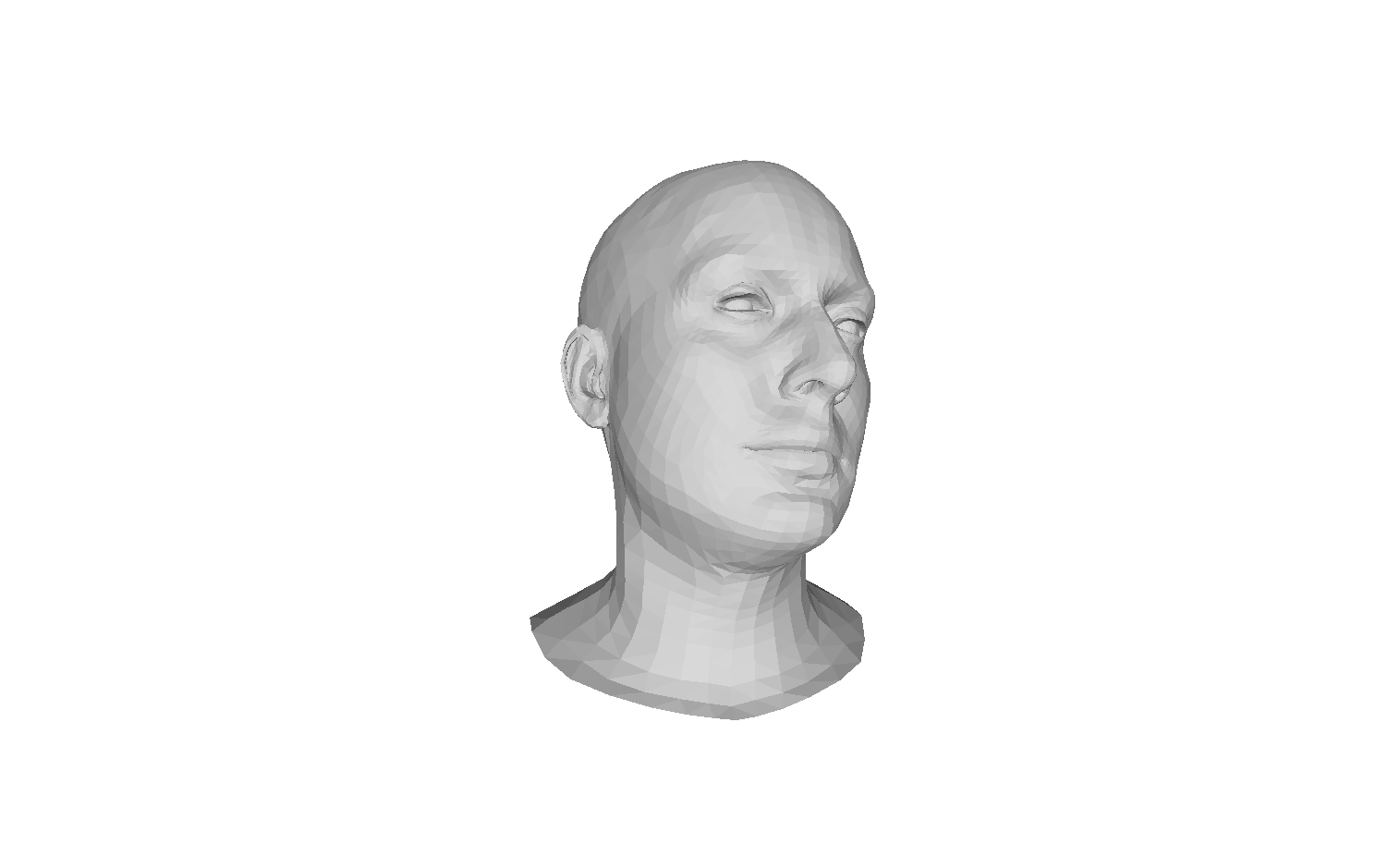}};
    \node[right of=a4, node distance=1.6cm] (a5) {\includegraphics[trim={400 80 400 100},clip,width=0.09\linewidth]{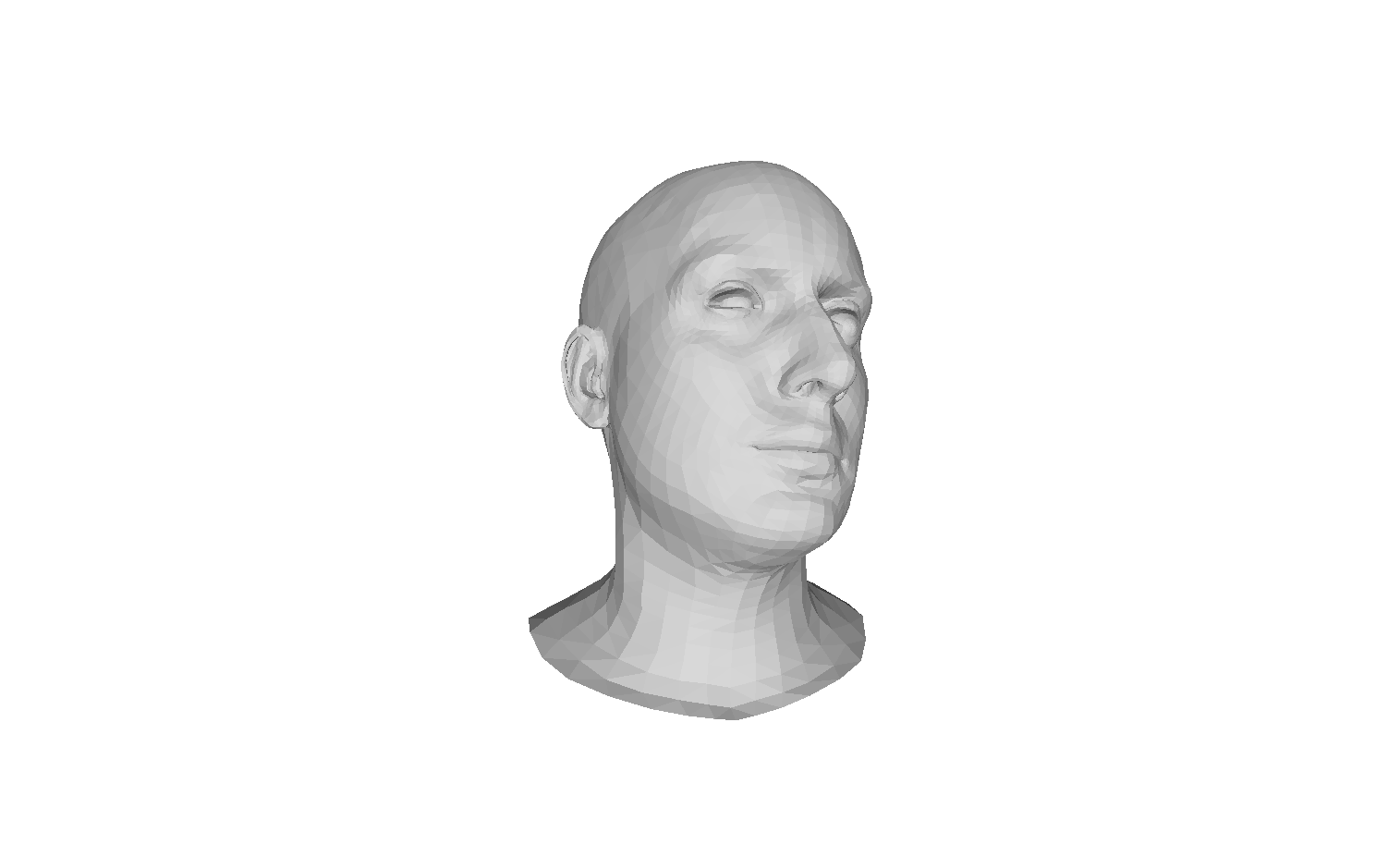}};
    \node[right of=a5, node distance=1.6cm] (a6) {\includegraphics[trim={400 80 400 100},clip,width=0.09\linewidth]{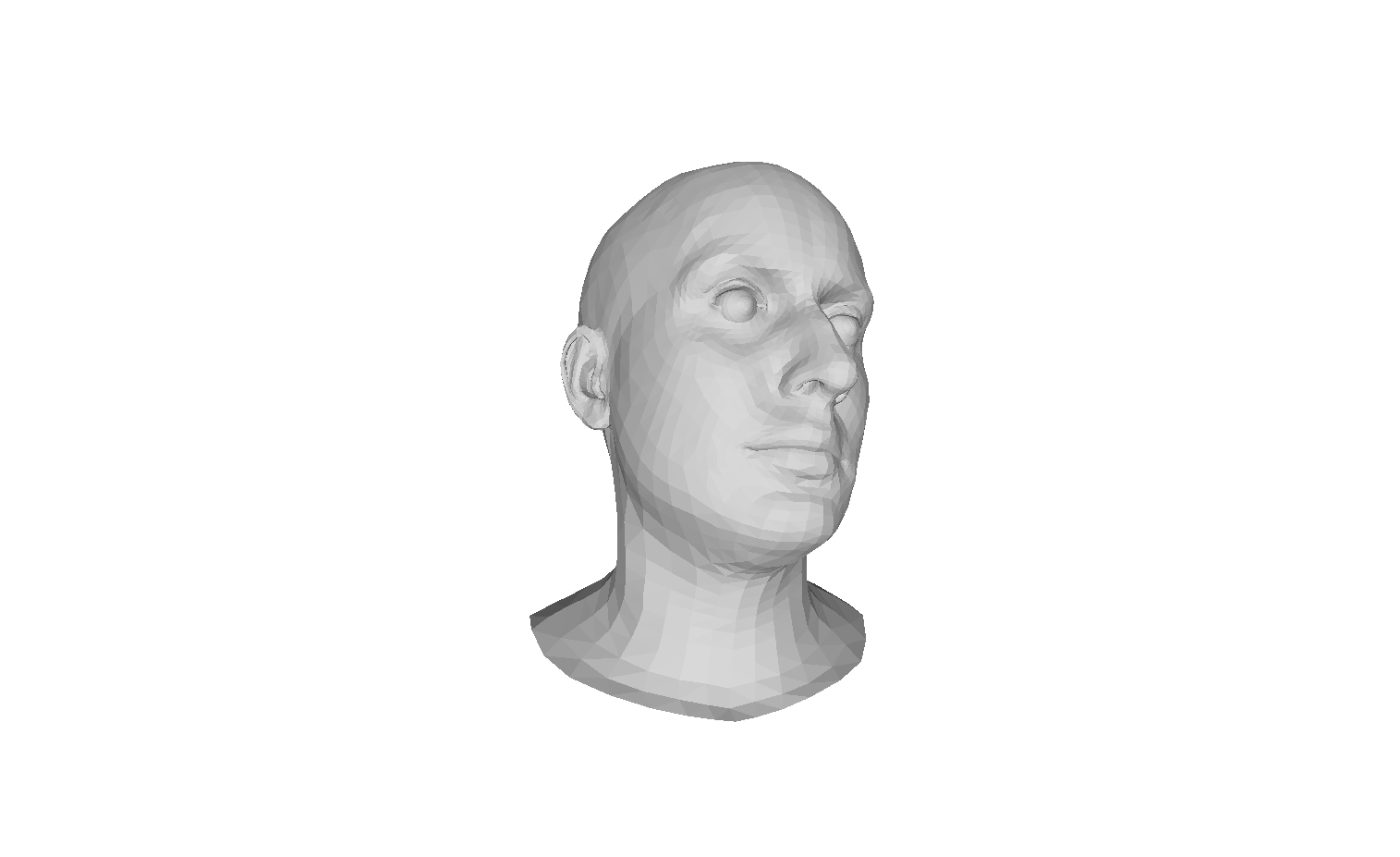}};
    \node[right of=a6, node distance=1.6cm] (a7) {\includegraphics[trim={400 80 400 100},clip,width=0.09\linewidth]{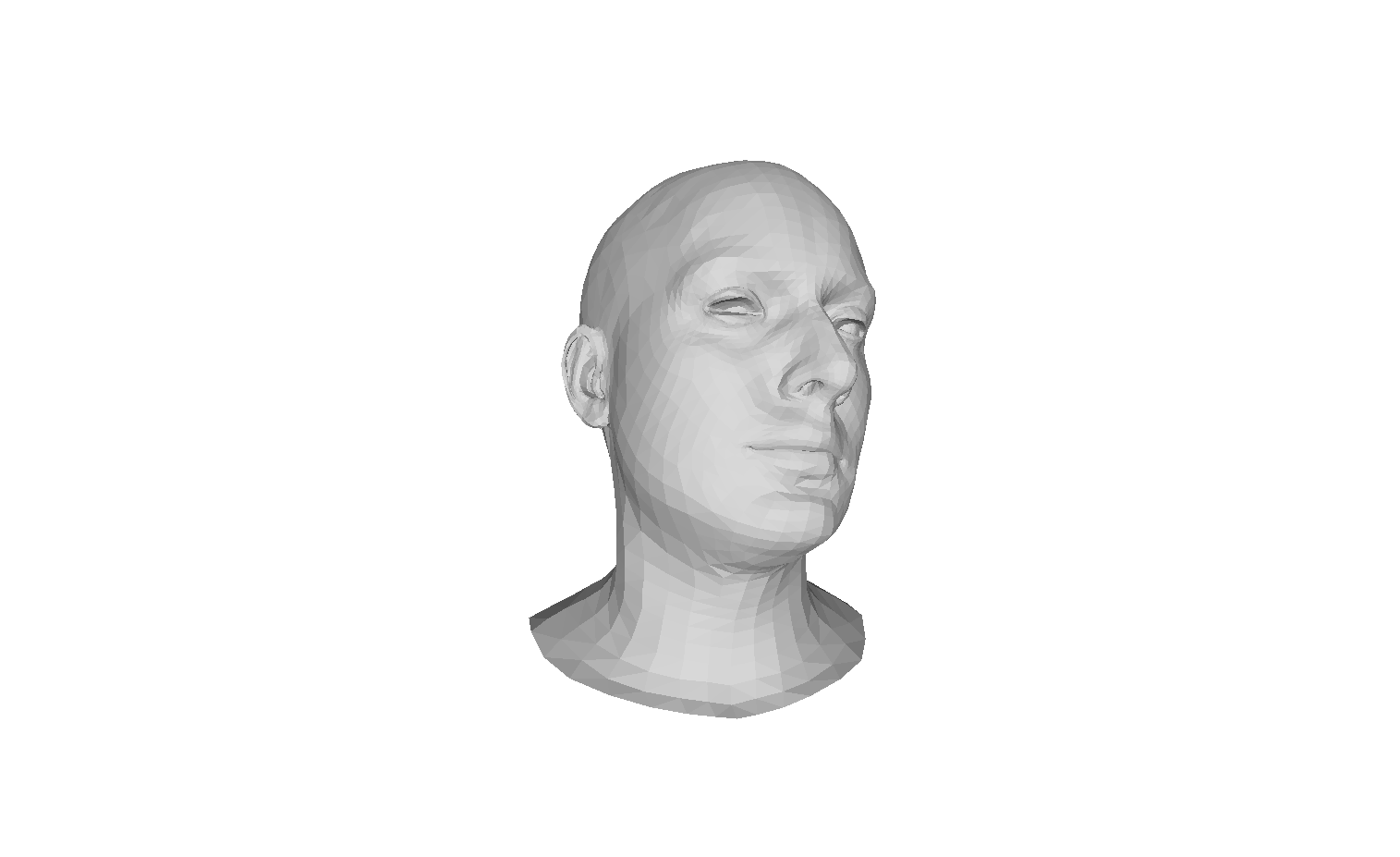}};
    \node[right of=a7, node distance=1.6cm] (a8) {\includegraphics[trim={400 80 400 100},clip,width=0.09\linewidth]{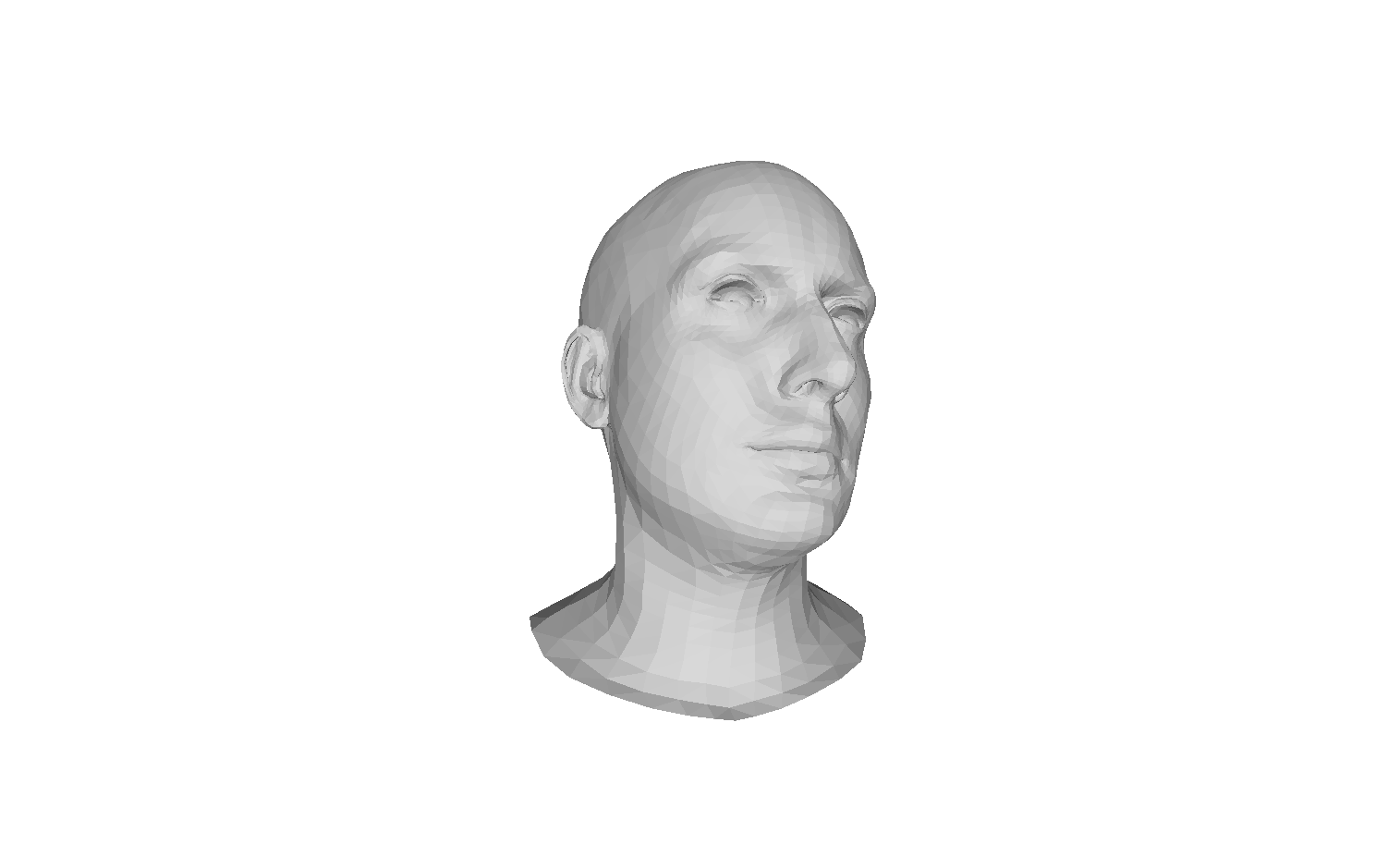}};
    \node[right of=a8, node distance=1.6cm] (a9) {\includegraphics[trim={400 80 400 100},clip,width=0.09\linewidth]{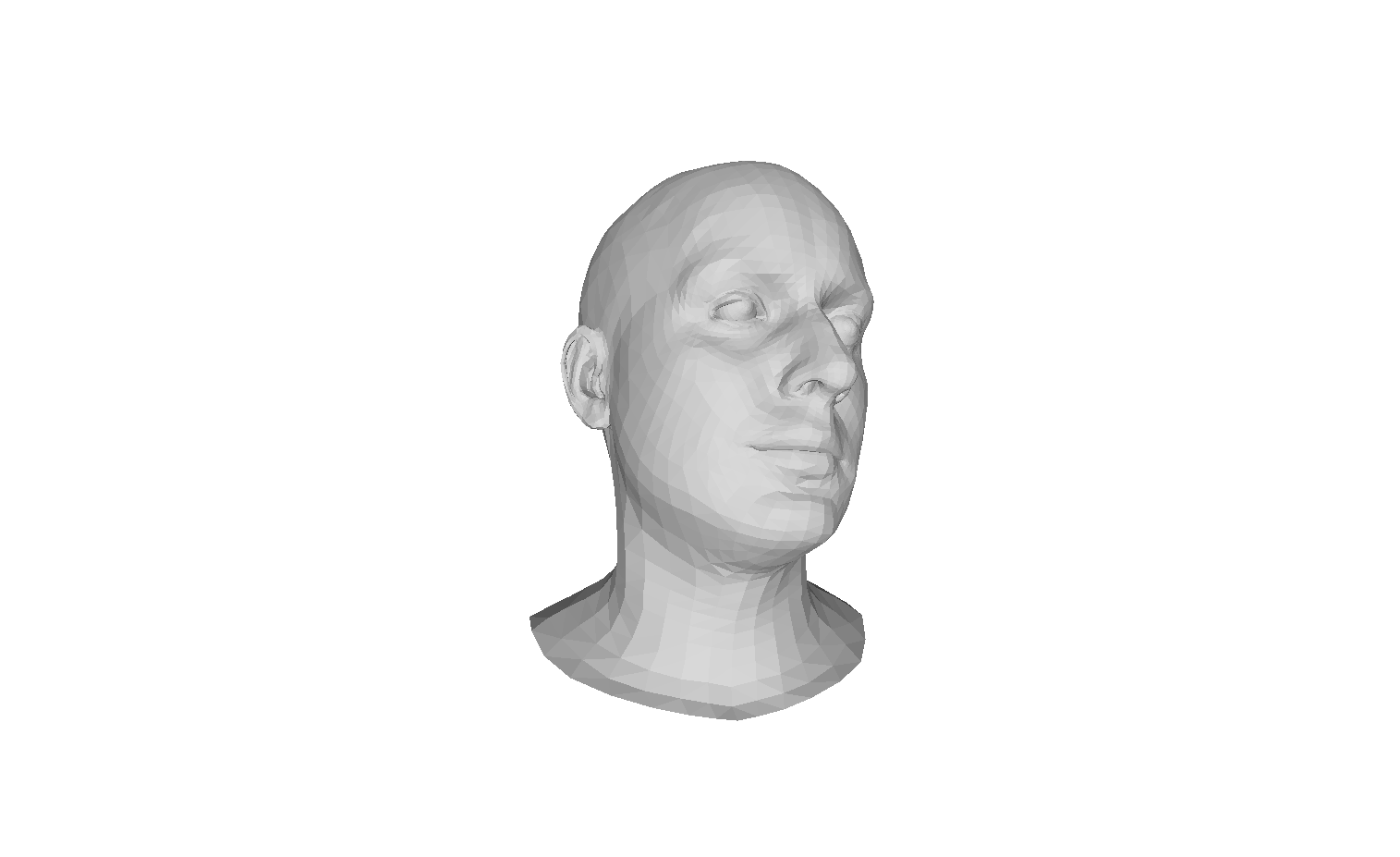}};
    
    \node[below of=a1, node distance=2.1cm] (b1) {\includegraphics[width=0.11\linewidth]{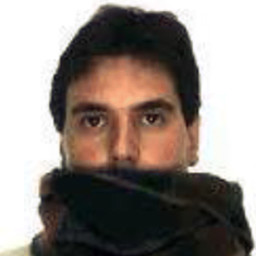}};
    \node[right of=b1, node distance=2.5cm] (b2) {\includegraphics[trim={400 80 400 100},clip,width=0.09\linewidth]{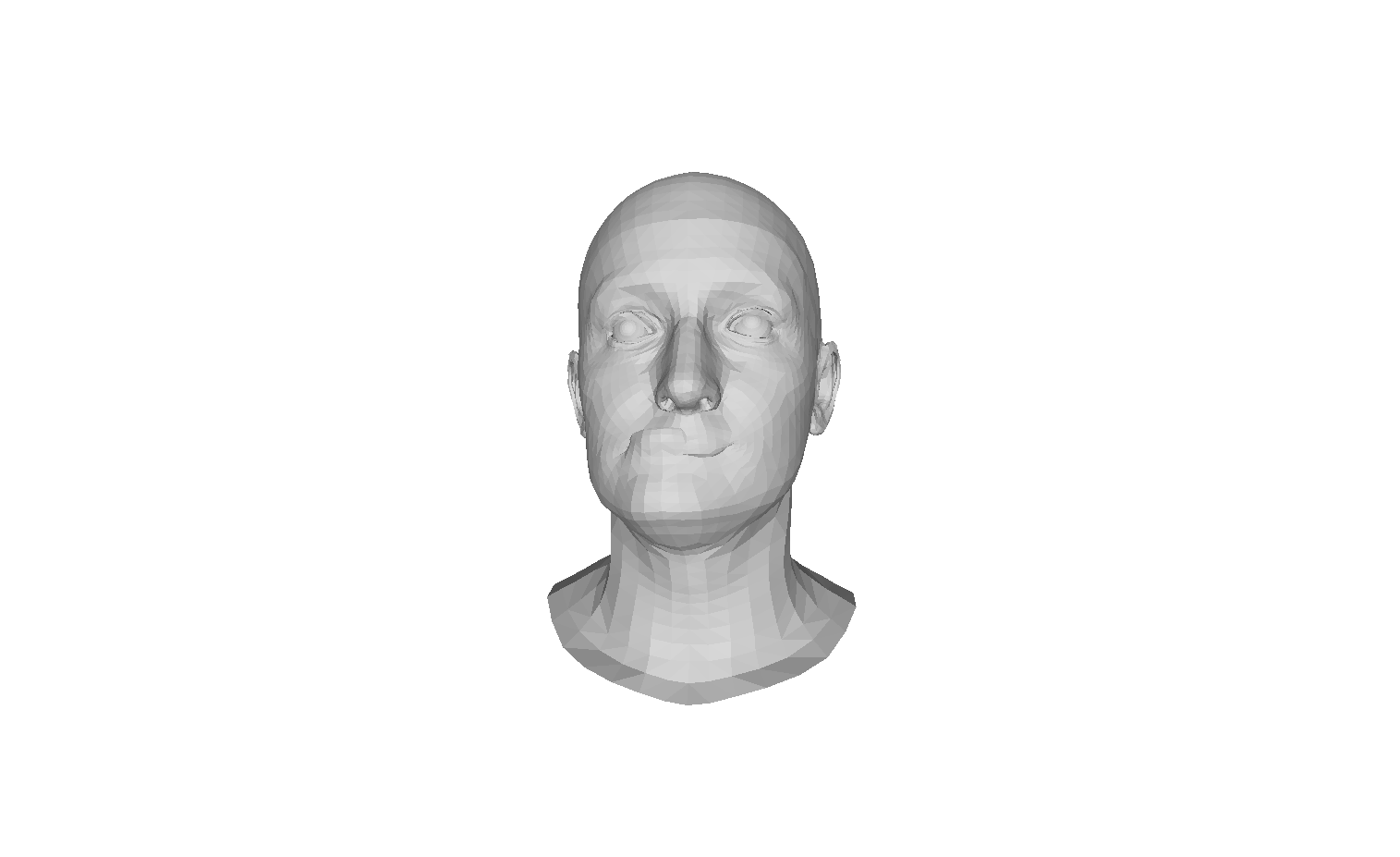}};
    \node[right of=b2, node distance=2.3cm] (b3) {\includegraphics[trim={400 80 400 100},clip,width=0.09\linewidth]{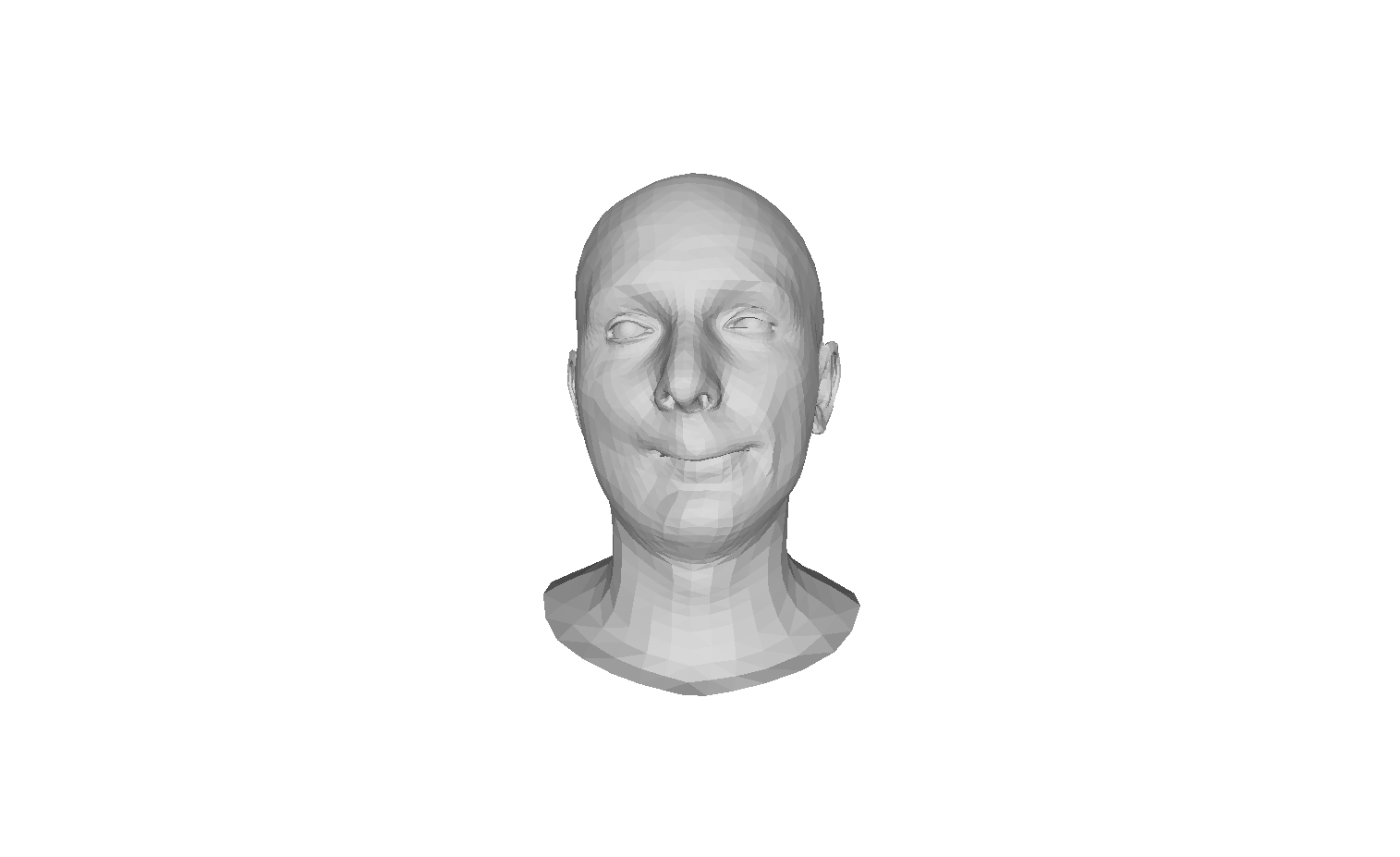}};
    \node[right of=b3, node distance=1.6cm] (b4) {\includegraphics[trim={400 80 400 100},clip,width=0.09\linewidth]{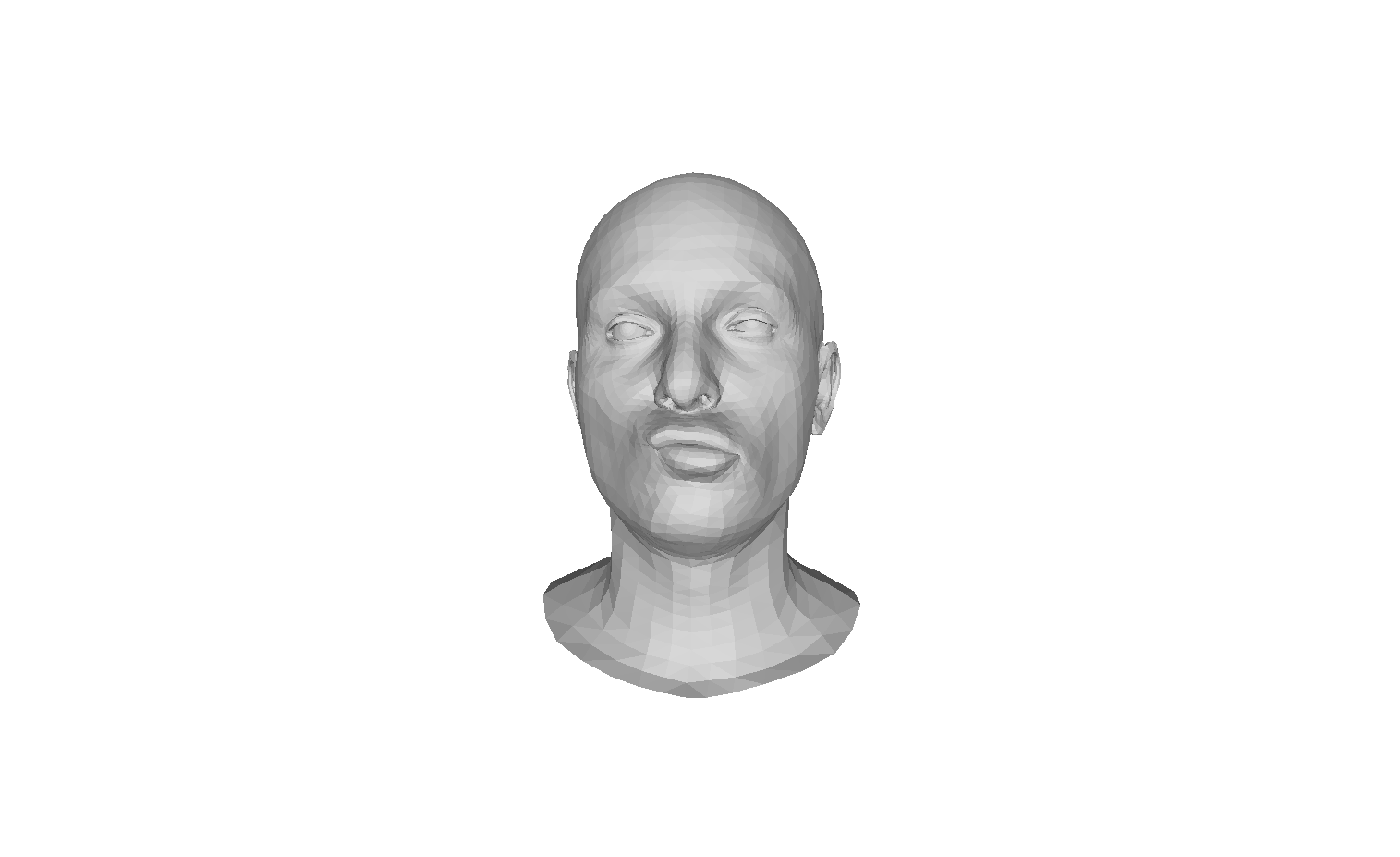}};
    \node[right of=b4, node distance=1.6cm] (b5) {\includegraphics[trim={400 80 400 100},clip,width=0.09\linewidth]{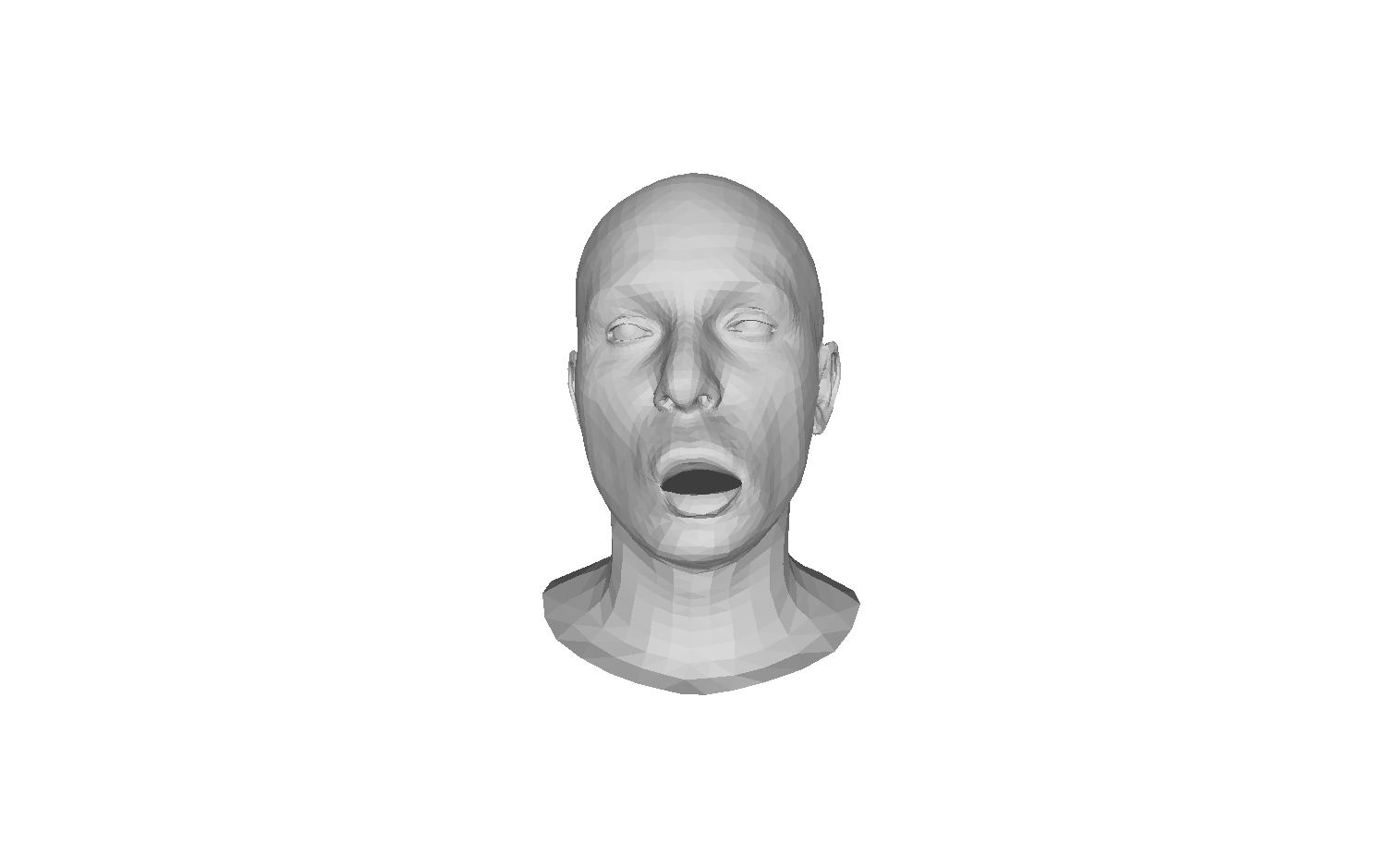}};
    \node[right of=b5, node distance=1.6cm] (b6) {\includegraphics[trim={400 80 400 100},clip,width=0.09\linewidth]{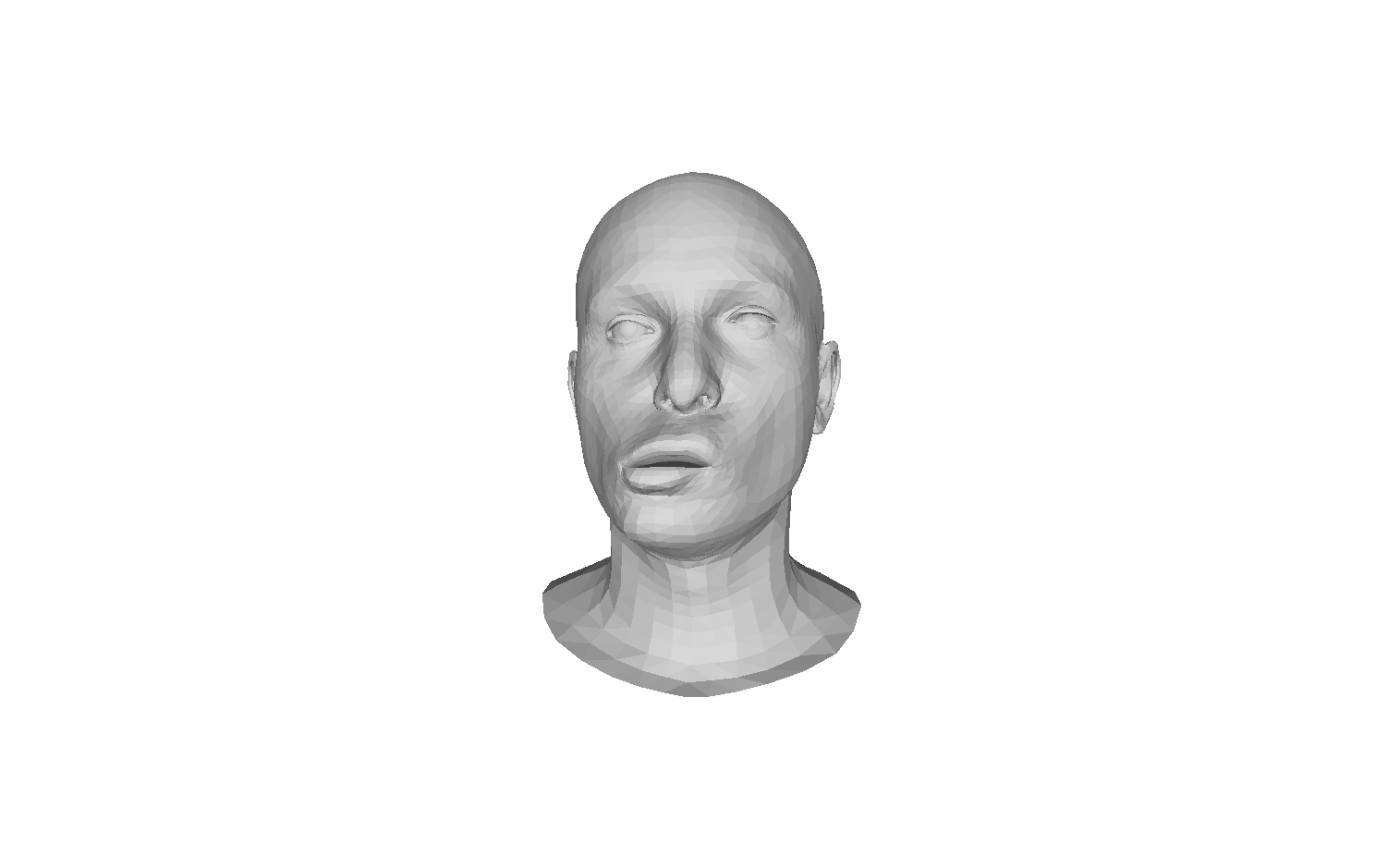}};
    \node[right of=b6, node distance=1.6cm] (b7) {\includegraphics[trim={400 80 400 100},clip,width=0.09\linewidth]{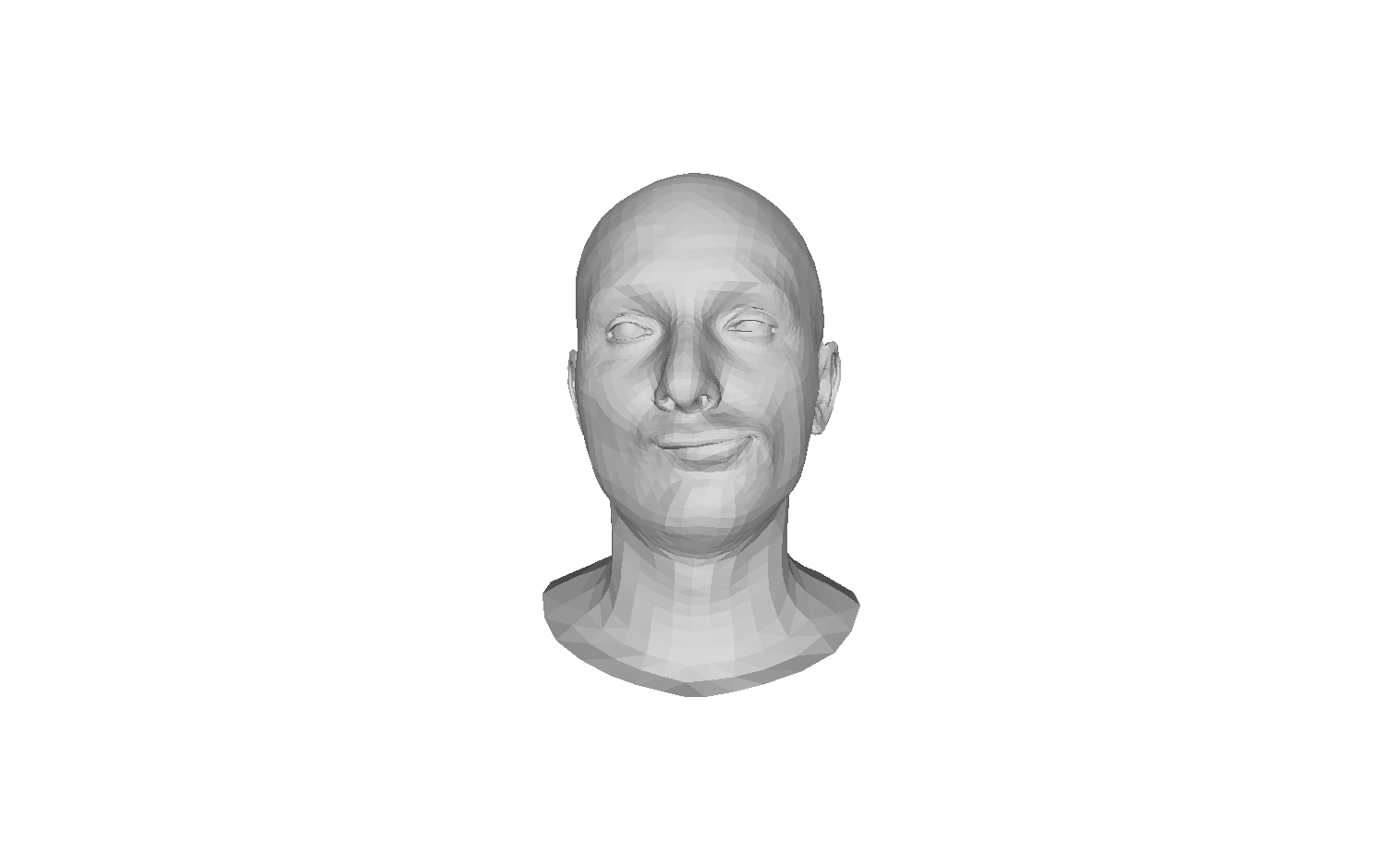}};
    \node[right of=b7, node distance=1.6cm] (b8) {\includegraphics[trim={400 80 400 100},clip,width=0.09\linewidth]{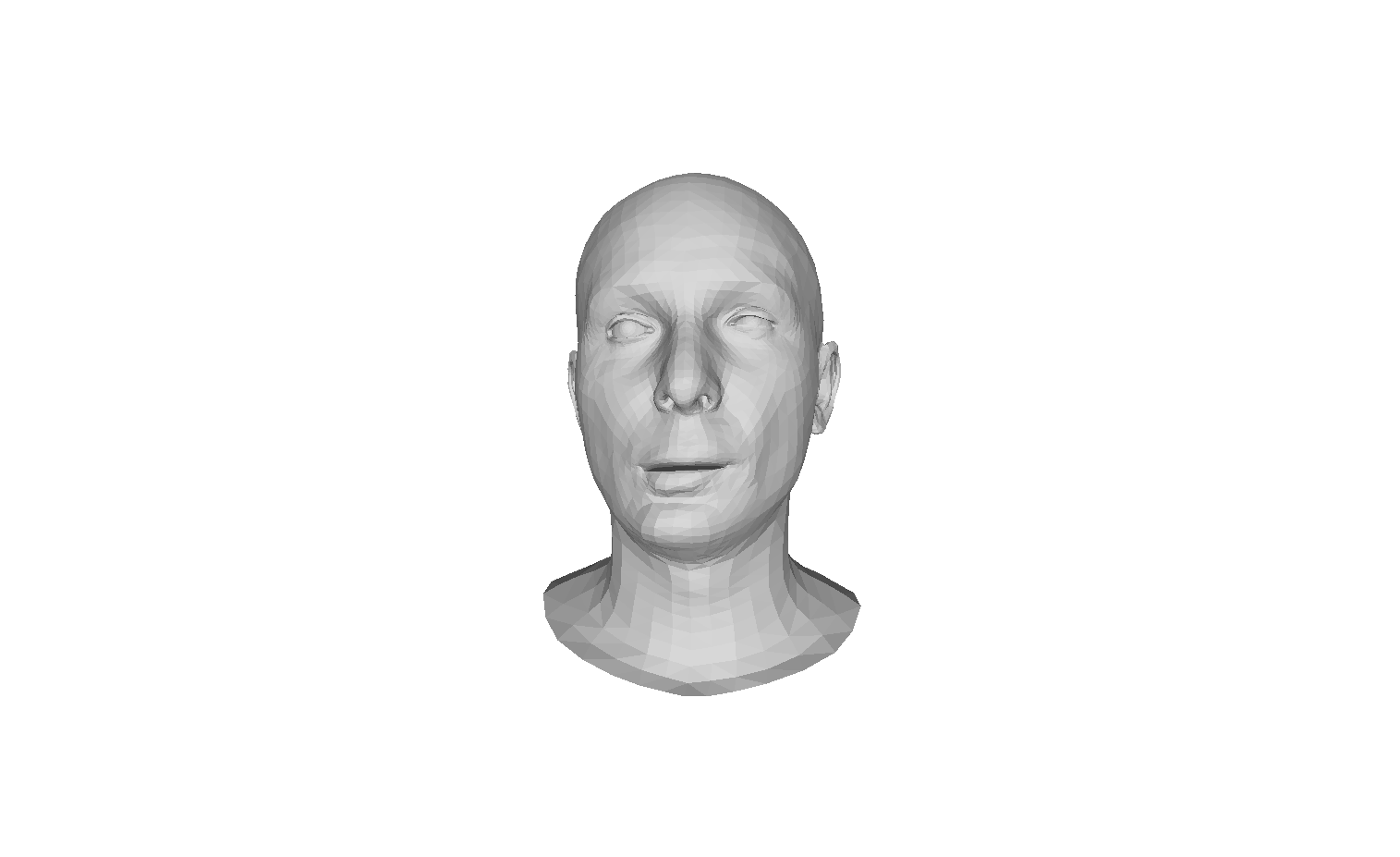}};
    \node[right of=b8, node distance=1.6cm] (b9) {\includegraphics[trim={400 80 400 100},clip,width=0.09\linewidth]{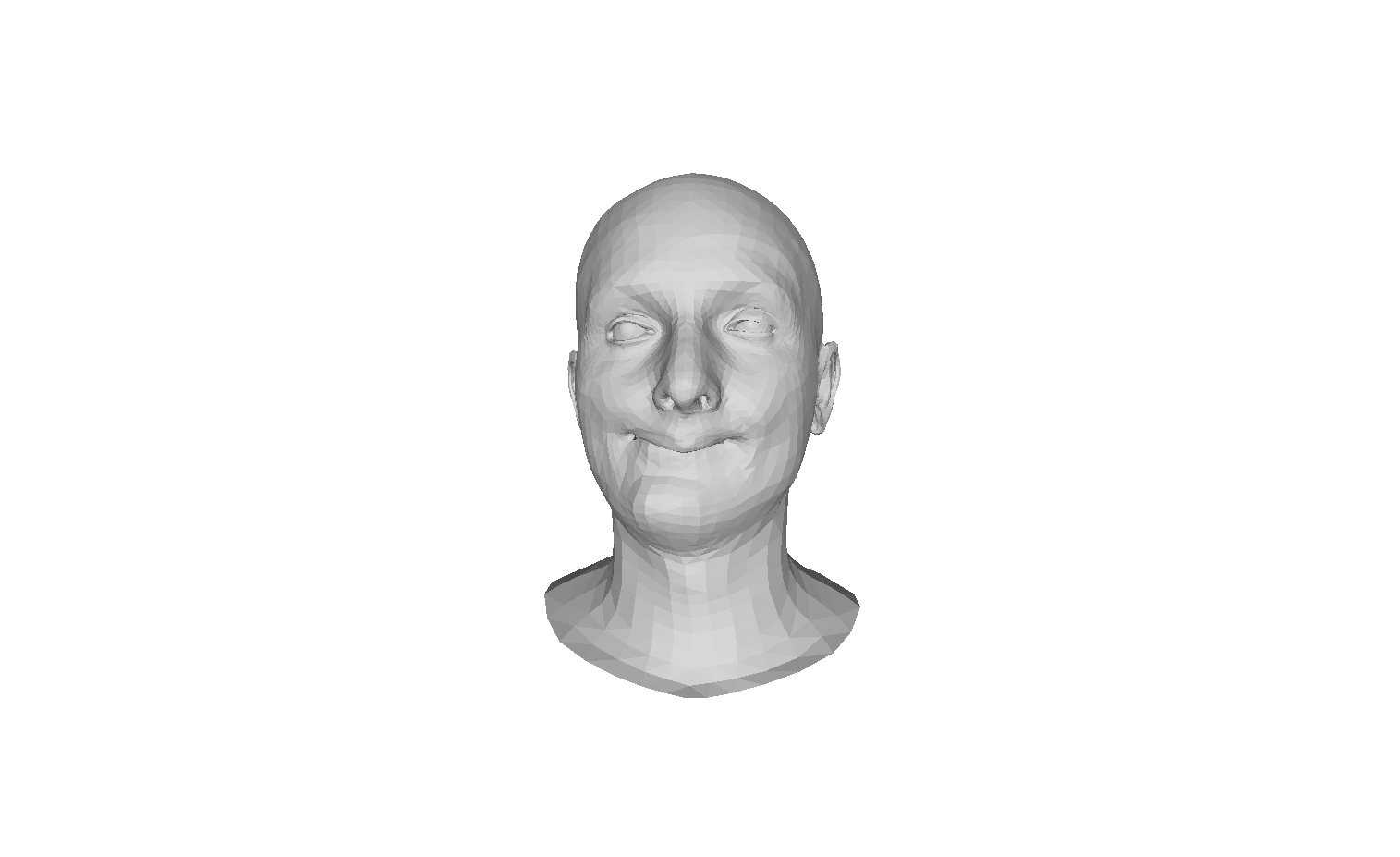}};
    
    \node[below of=b1, node distance=2.1cm] (c1) {\includegraphics[width=0.11\linewidth]{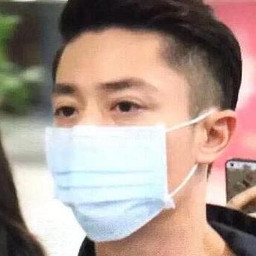}};
    \node[right of=c1, node distance=2.5cm] (c2) {\includegraphics[trim={400 80 400 100},clip,width=0.09\linewidth]{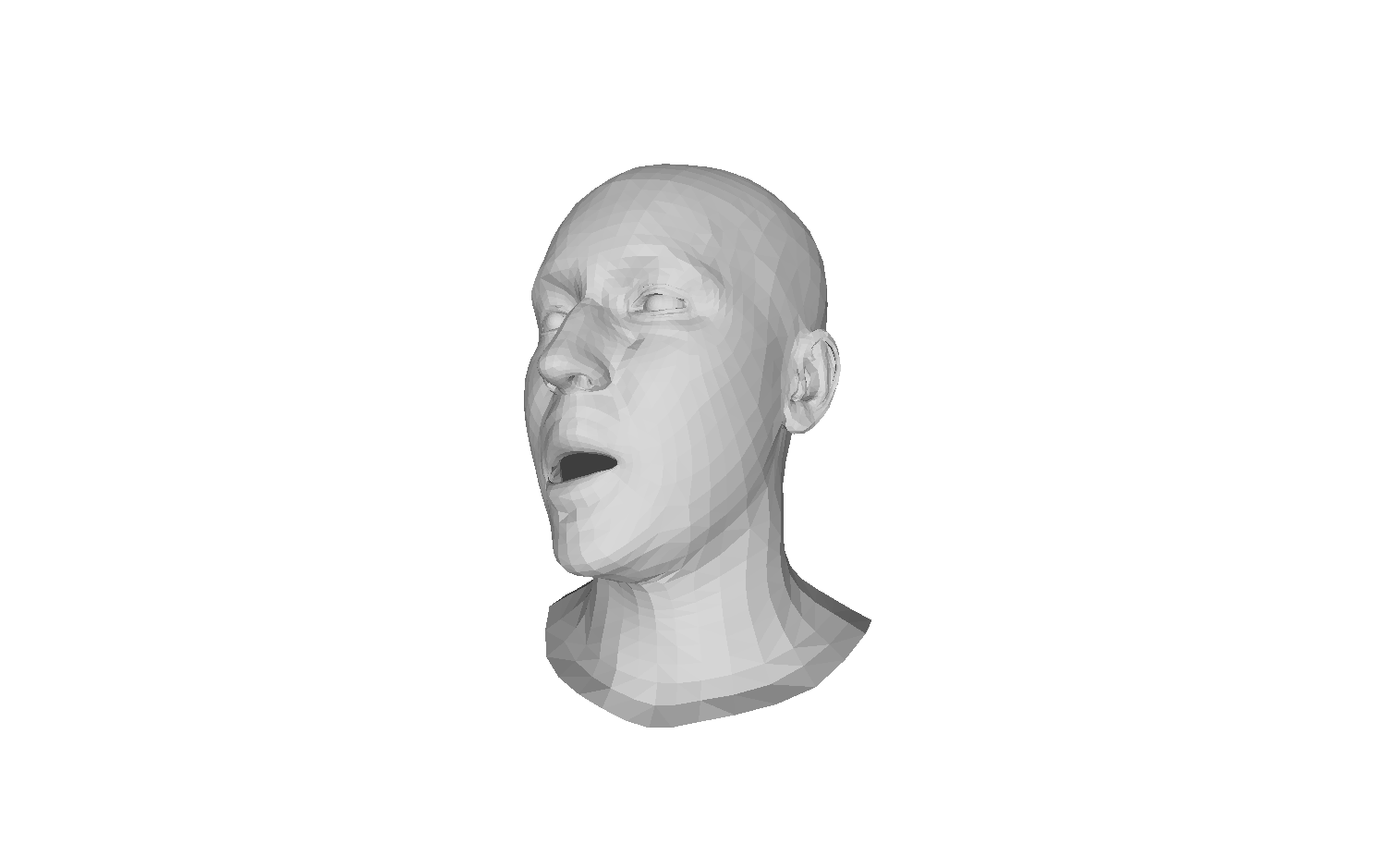}};
    \node[right of=c2, node distance=2.3cm] (c3) {\includegraphics[trim={400 80 400 100},clip,width=0.09\linewidth]{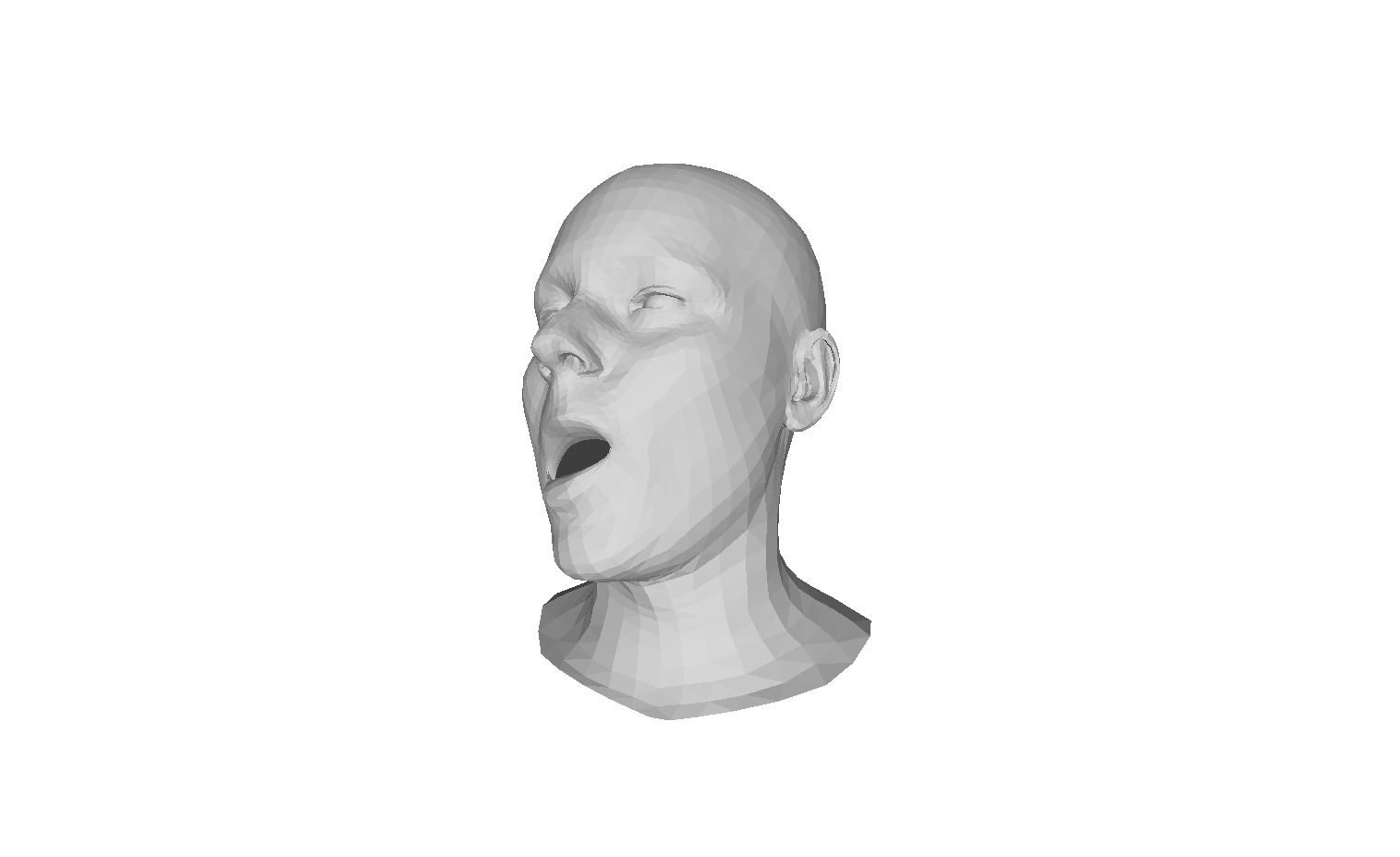}};
    \node[right of=c3, node distance=1.6cm] (c4) {\includegraphics[trim={400 80 400 100},clip,width=0.09\linewidth]{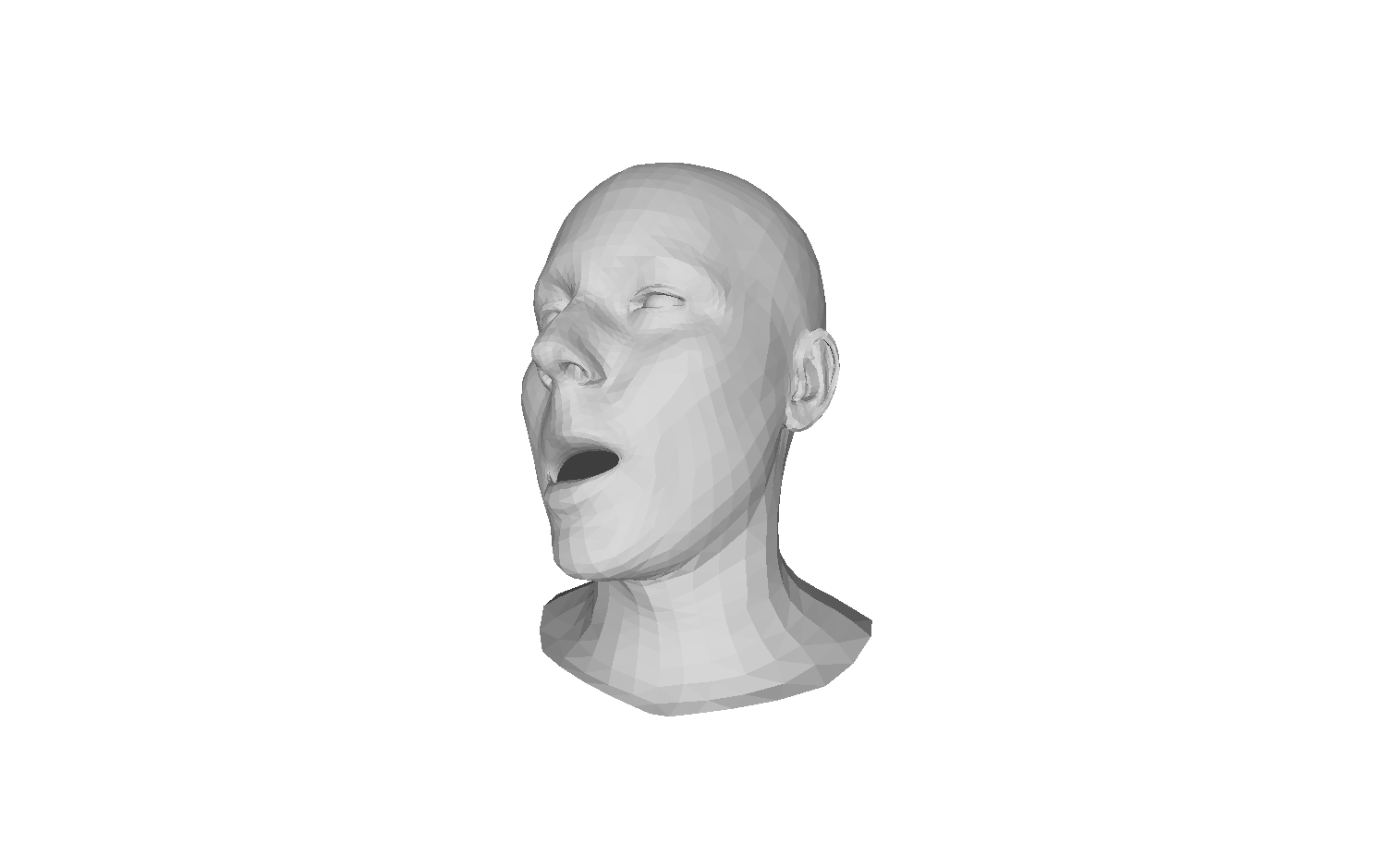}};
    \node[right of=c4, node distance=1.6cm] (c5) {\includegraphics[trim={400 80 400 100},clip,width=0.09\linewidth]{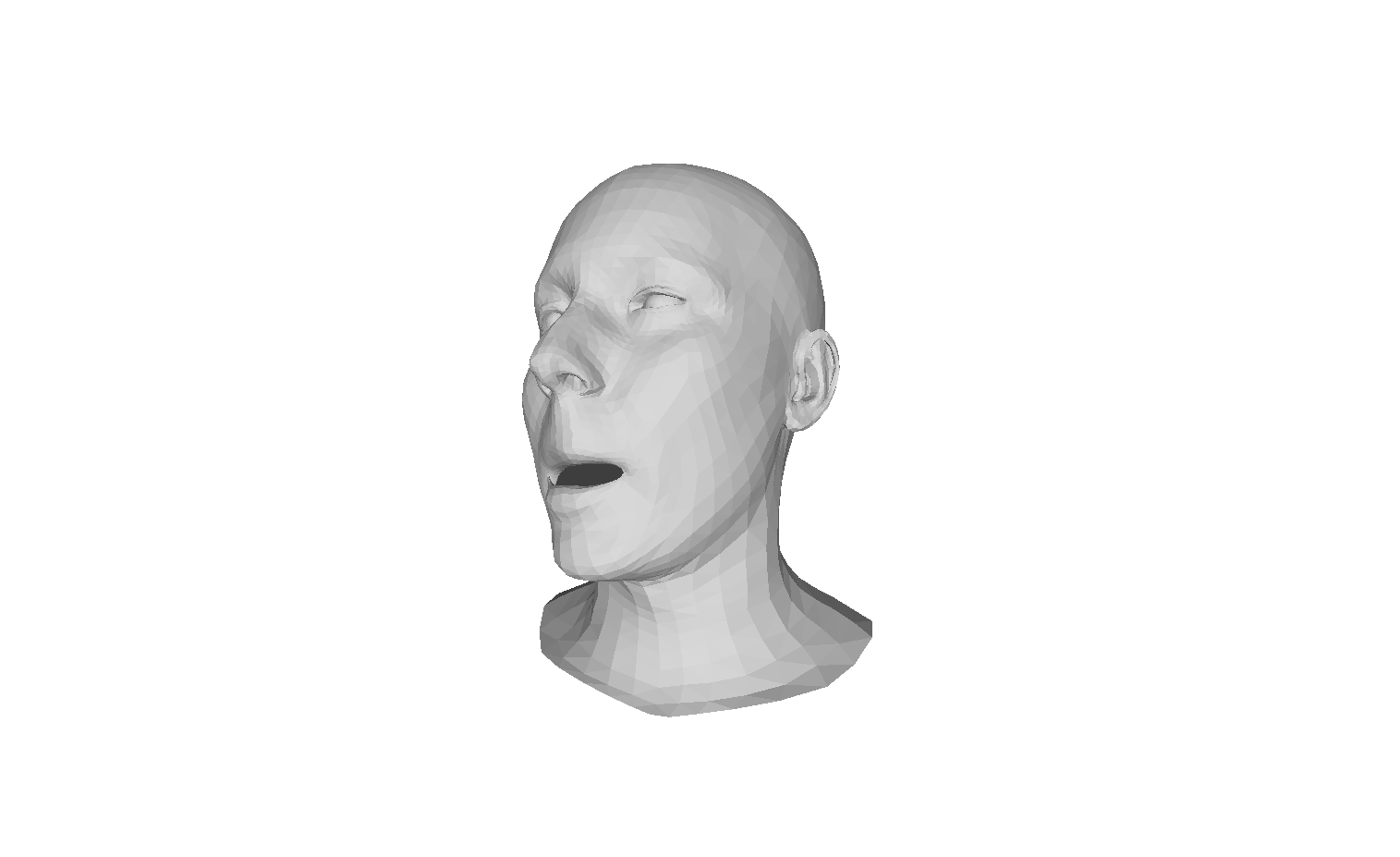}};
    \node[right of=c5, node distance=1.6cm] (c6) {\includegraphics[trim={400 80 400 100},clip,width=0.09\linewidth]{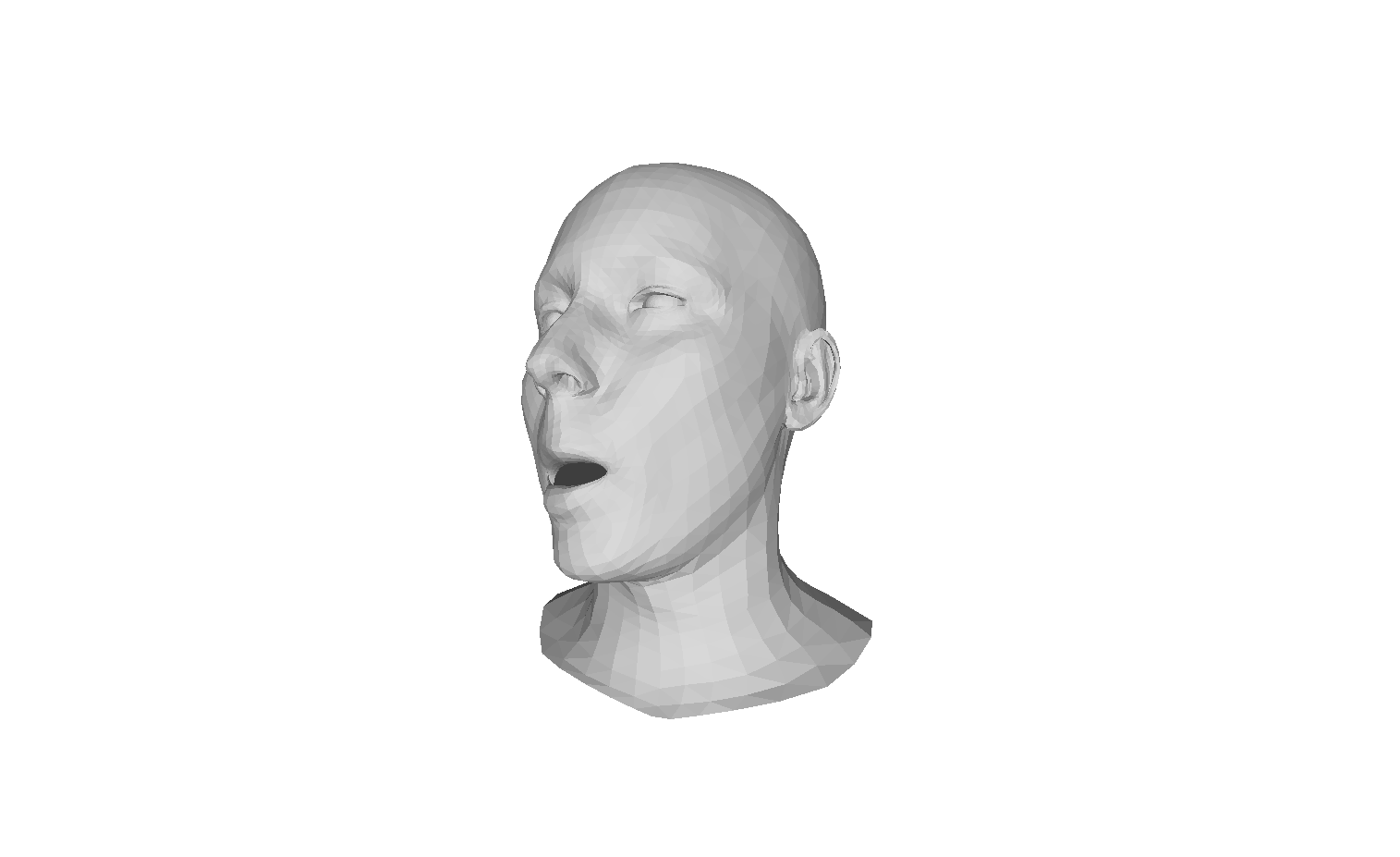}};
    \node[right of=c6, node distance=1.6cm] (c7) {\includegraphics[trim={400 80 400 100},clip,width=0.09\linewidth]{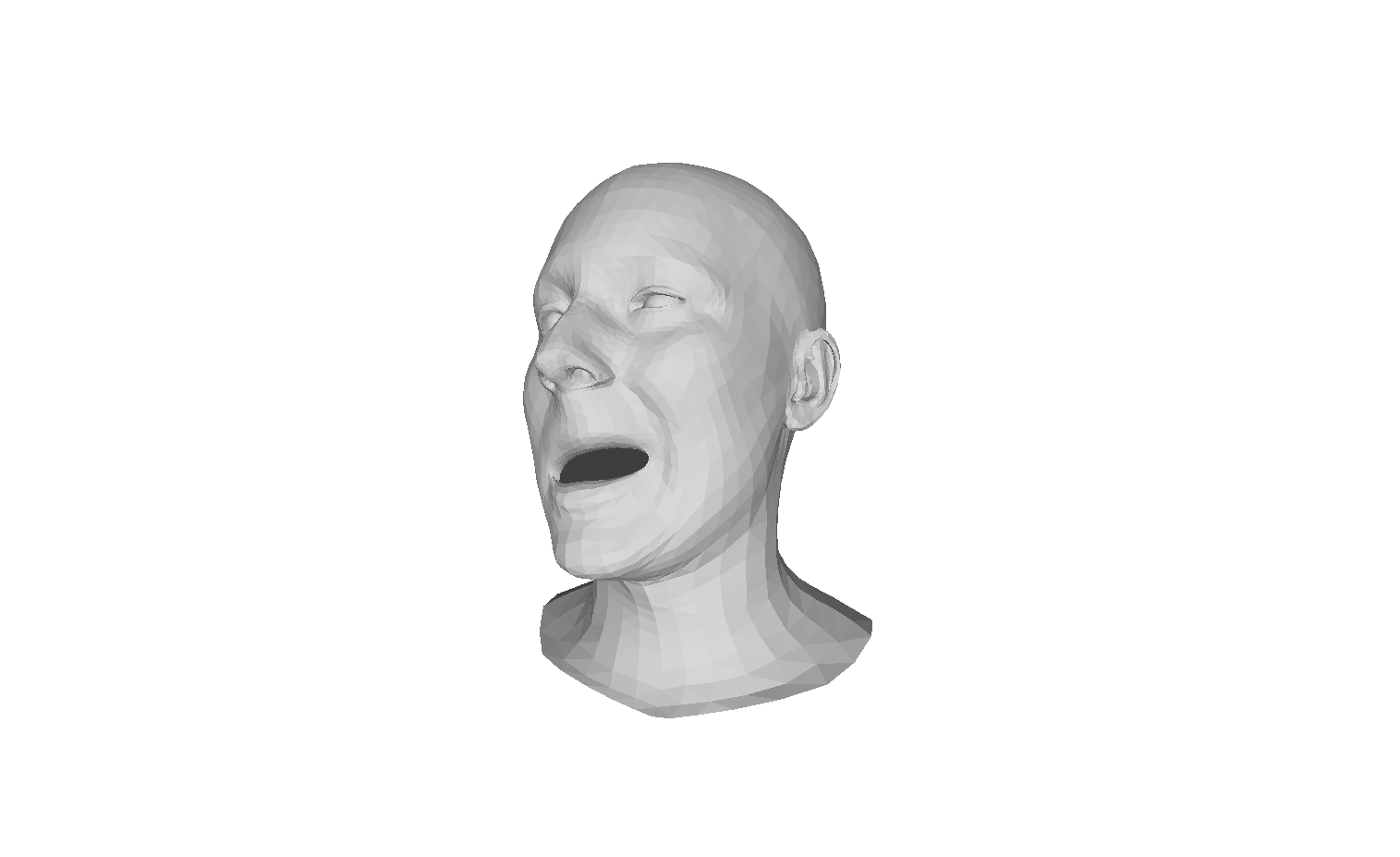}};
    \node[right of=c7, node distance=1.6cm] (c8) {\includegraphics[trim={400 80 400 100},clip,width=0.09\linewidth]{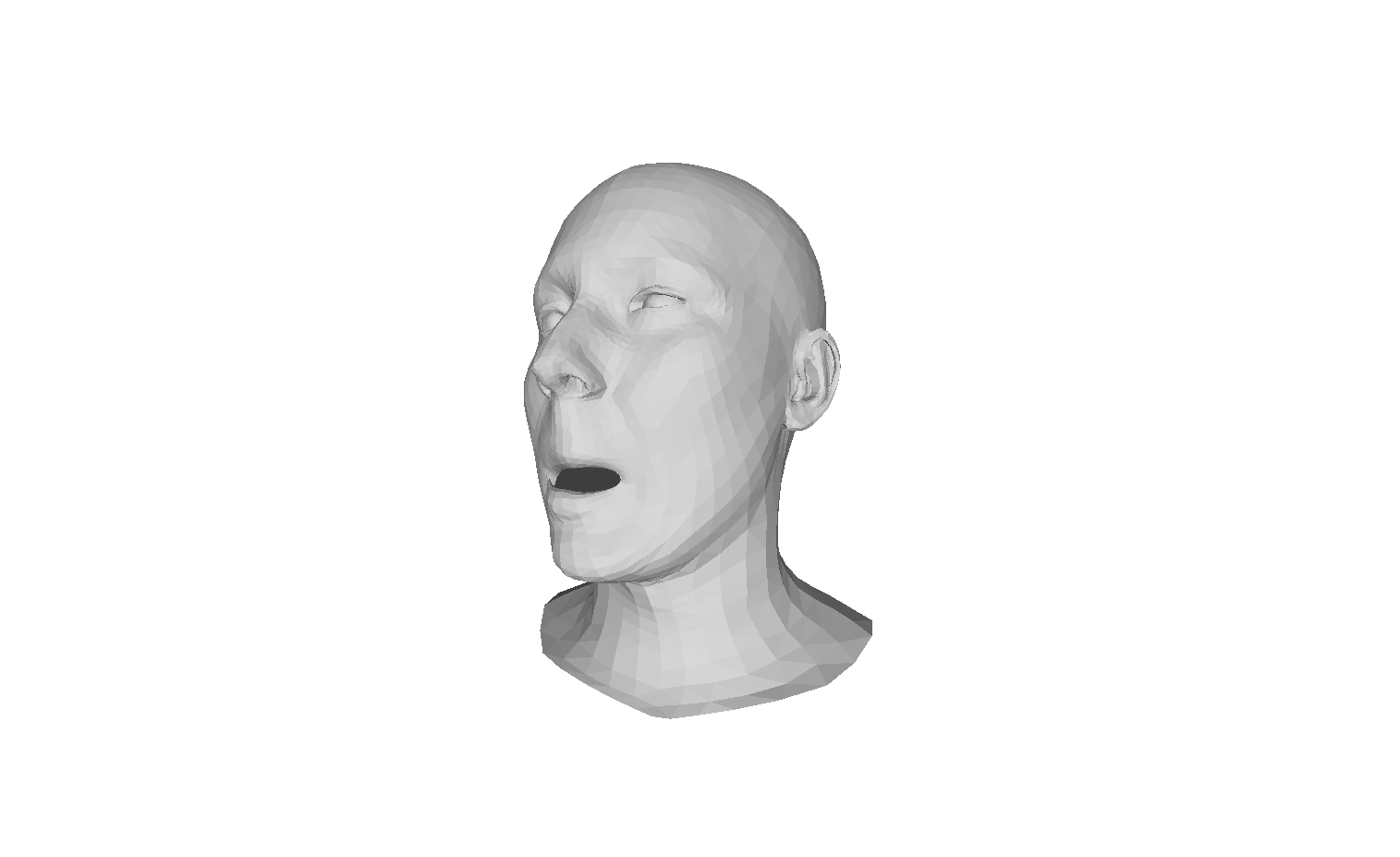}};
    \node[right of=c8, node distance=1.6cm] (c9) {\includegraphics[trim={400 80 400 100},clip,width=0.09\linewidth]{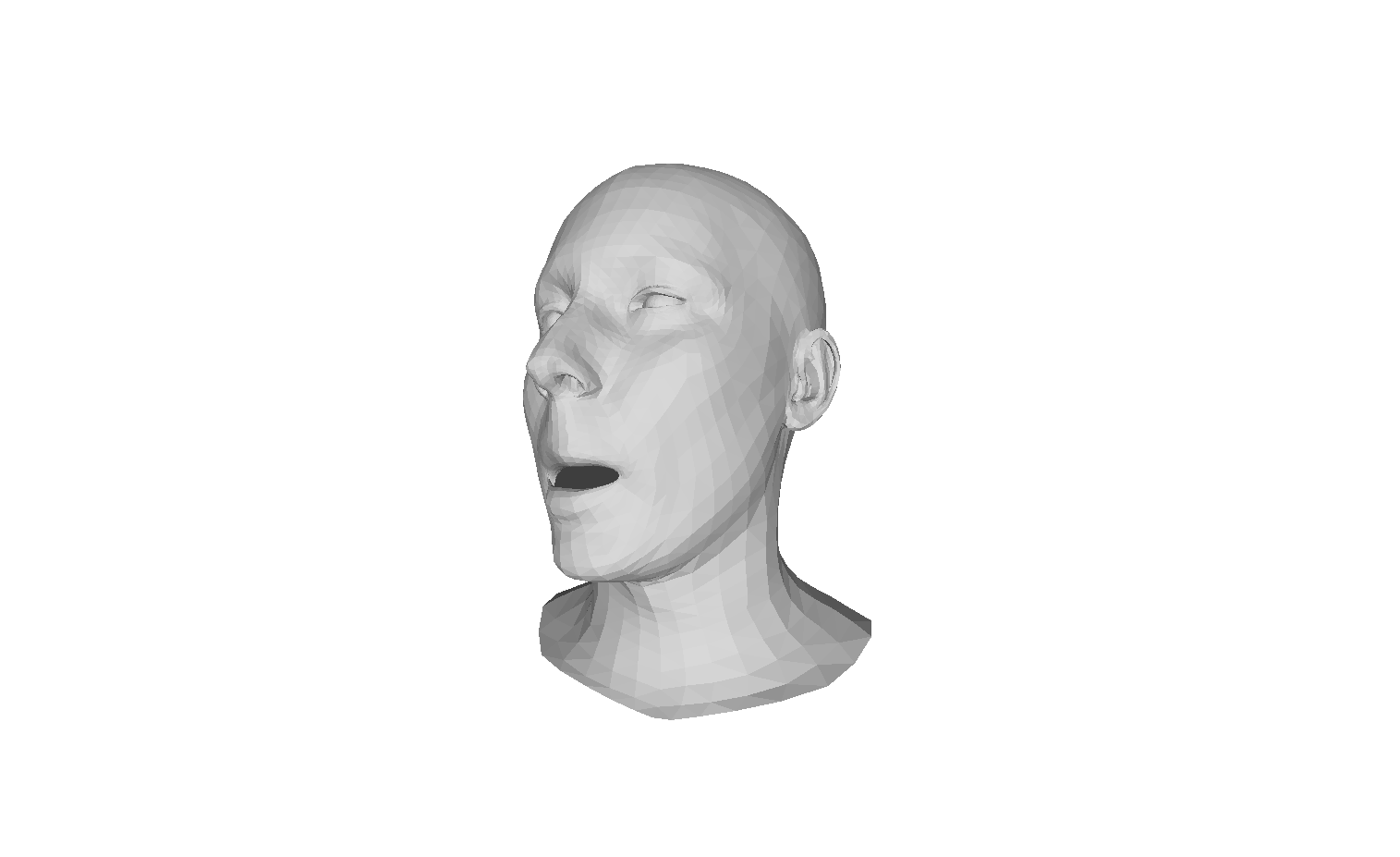}};
    
    \node[below of=c1, node distance=2.1cm] (d1) {\includegraphics[width=0.11\linewidth]{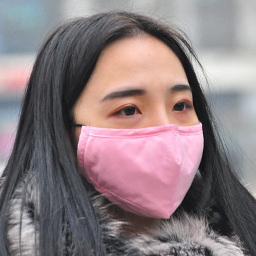}};
    \node[right of=d1, node distance=2.5cm] (d2) {\includegraphics[trim={400 80 400 100},clip,width=0.09\linewidth]{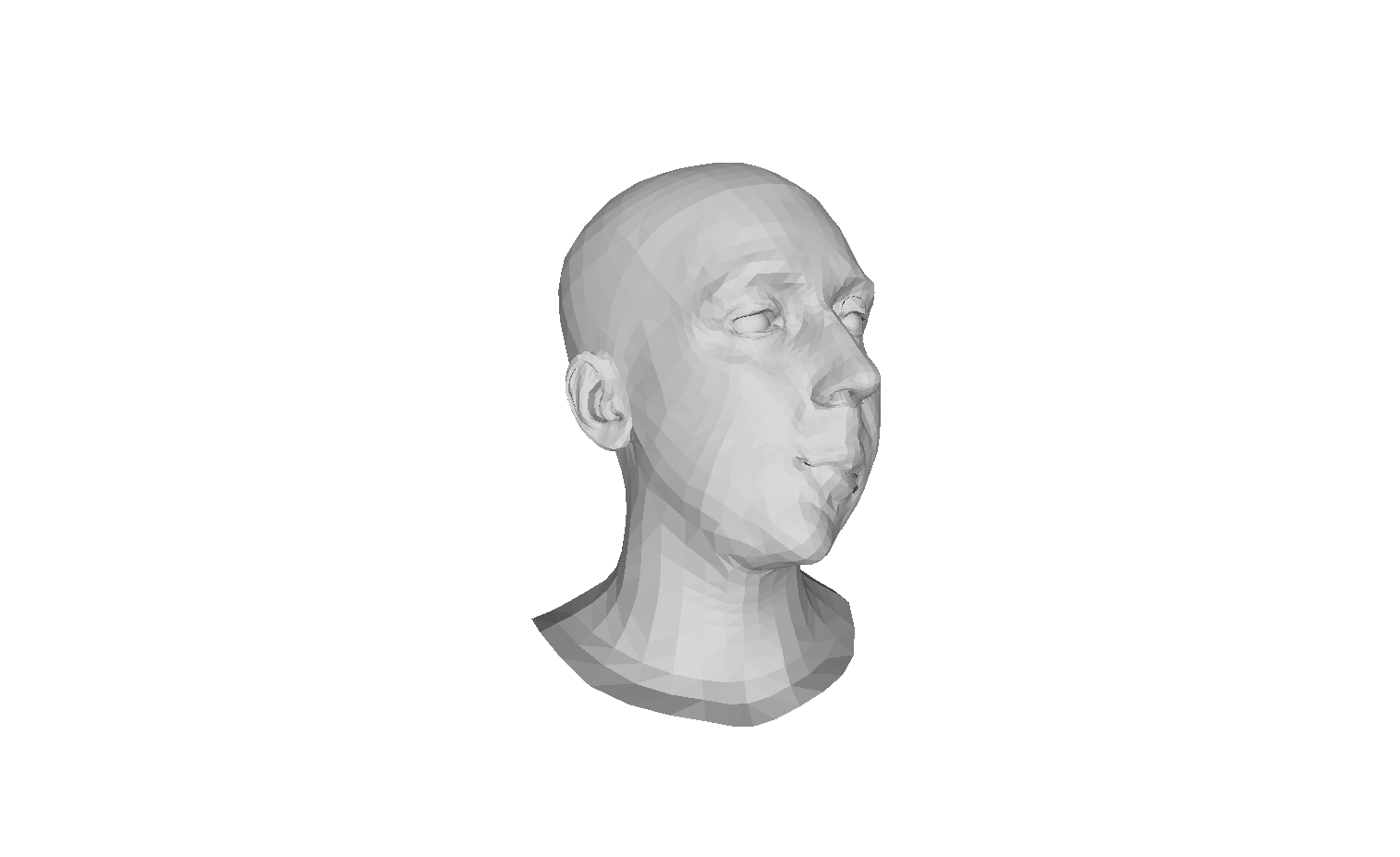}};
    \node[right of=d2, node distance=2.3cm] (d3) {\includegraphics[trim={400 80 400 100},clip,width=0.09\linewidth]{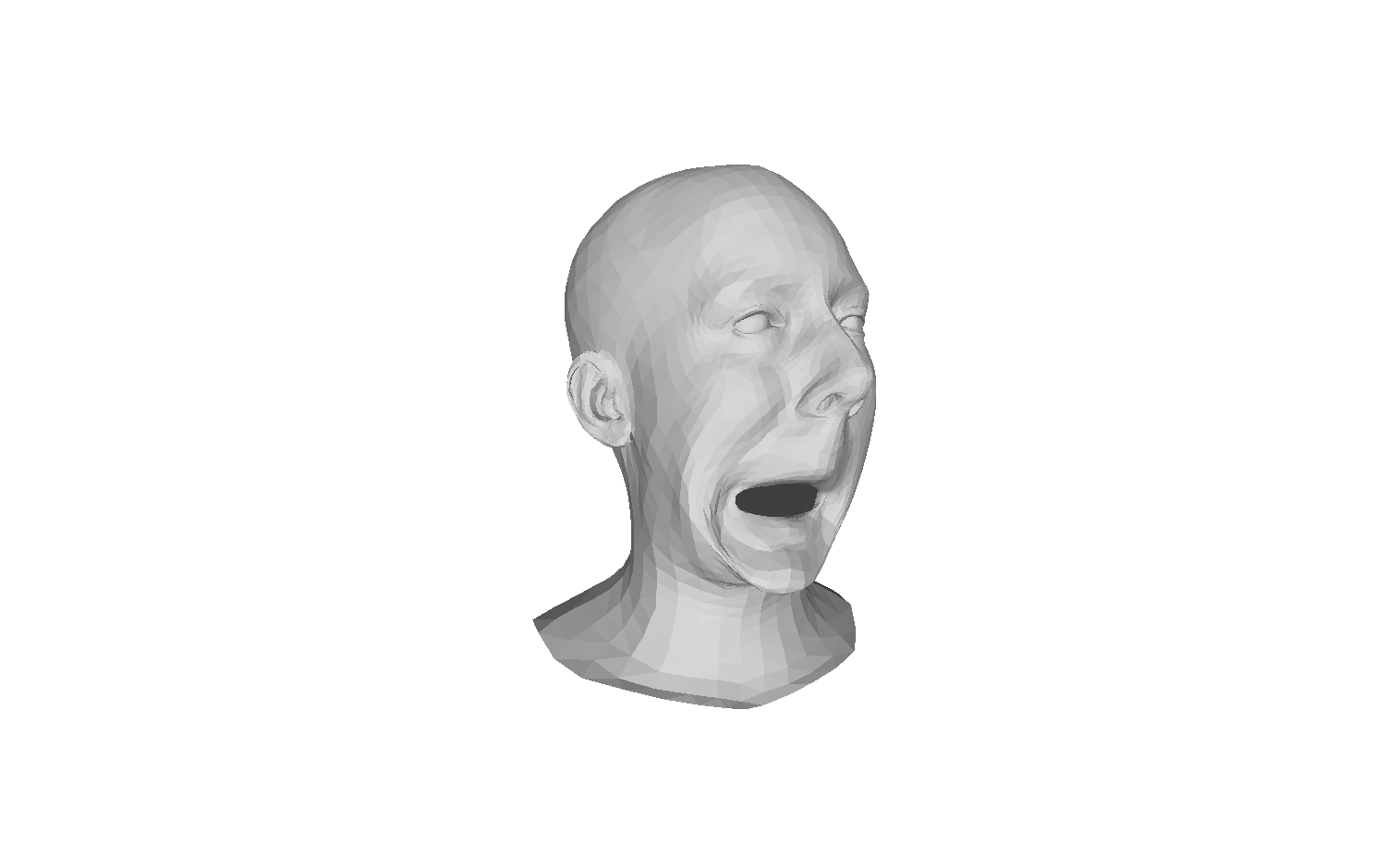}};
    \node[right of=d3, node distance=1.6cm] (d4) {\includegraphics[trim={400 80 400 100},clip,width=0.09\linewidth]{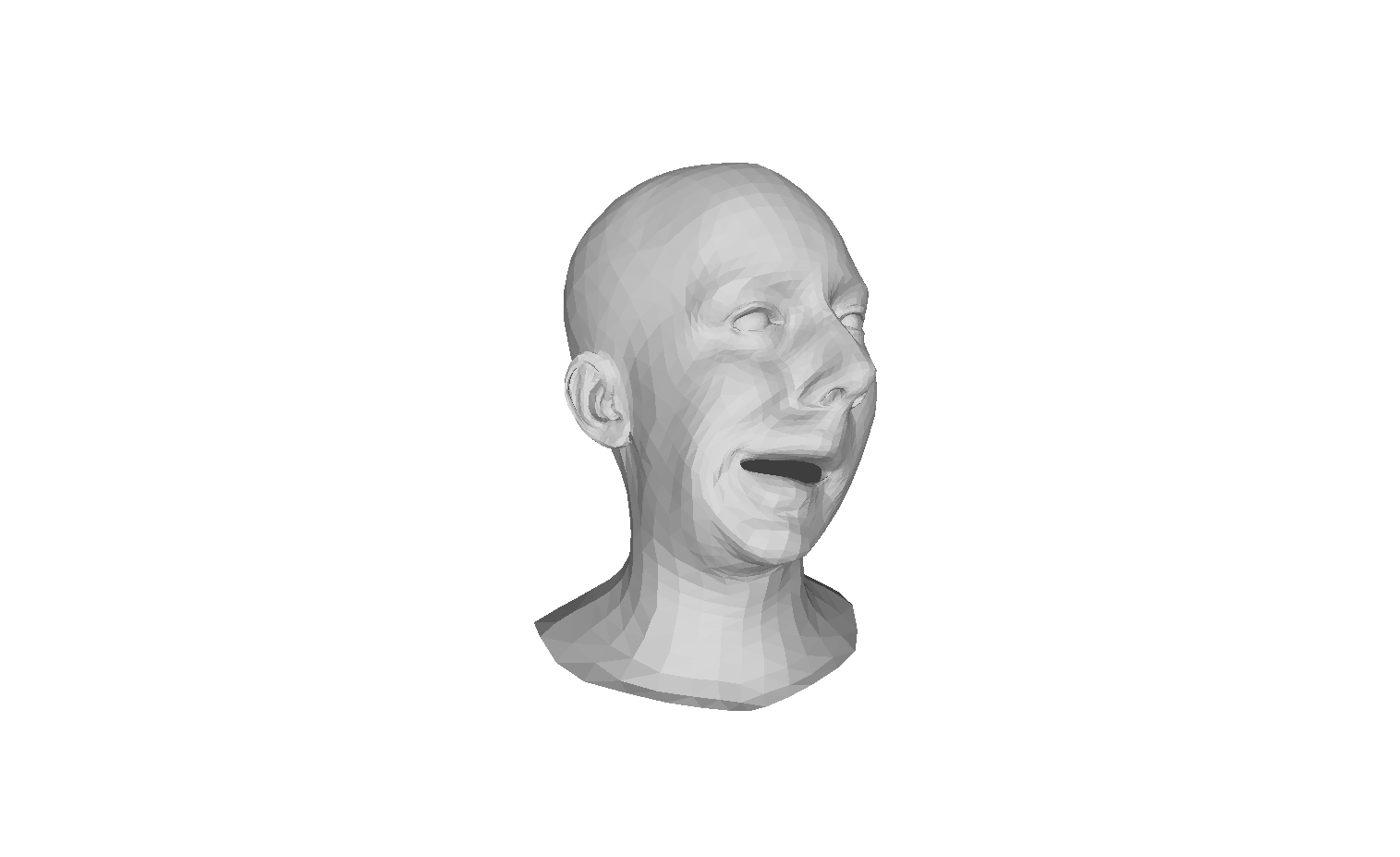}};
    \node[right of=d4, node distance=1.6cm] (d5) {\includegraphics[trim={400 80 400 100},clip,width=0.09\linewidth]{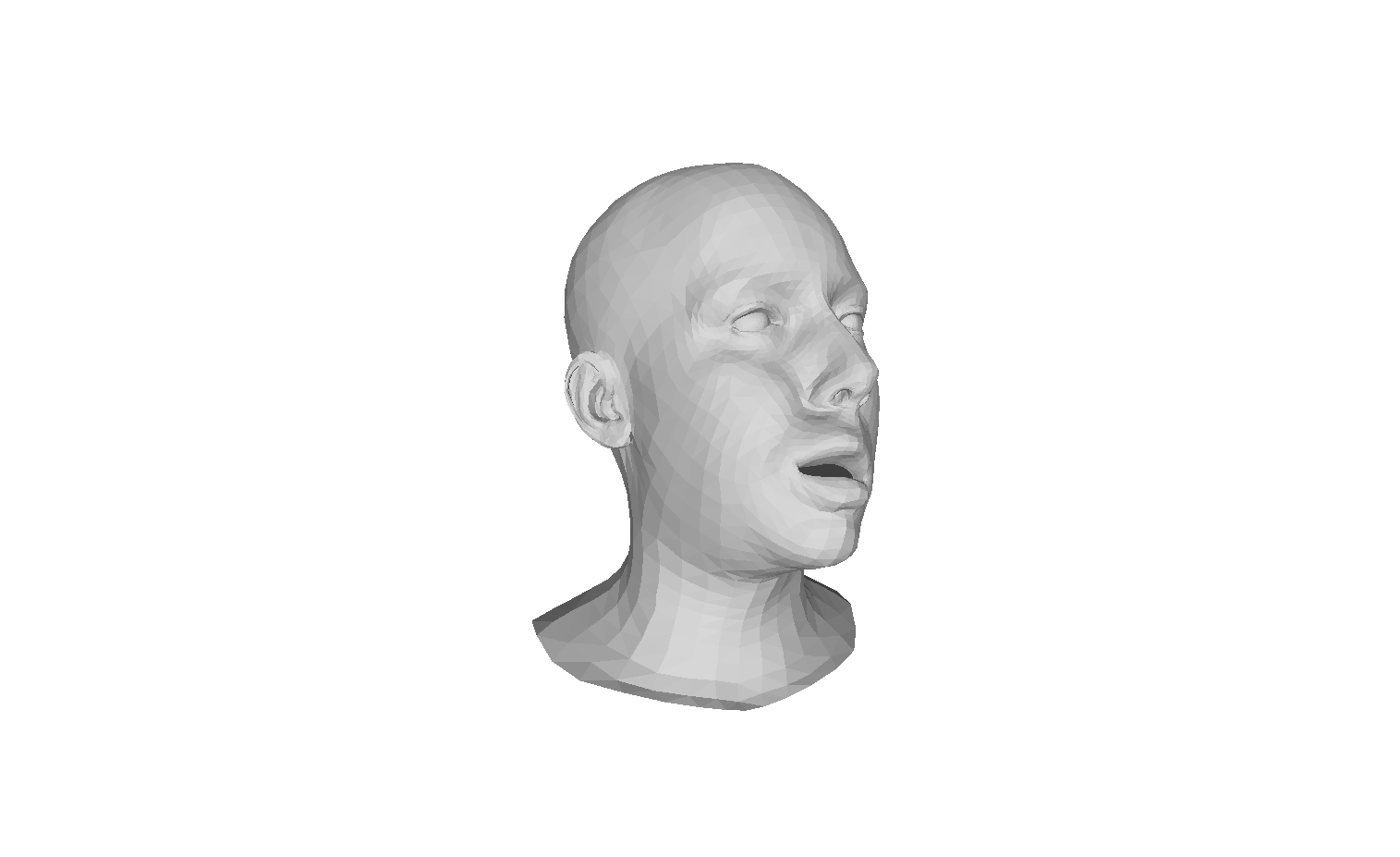}};
    \node[right of=d5, node distance=1.6cm] (d6) {\includegraphics[trim={400 80 400 100},clip,width=0.09\linewidth]{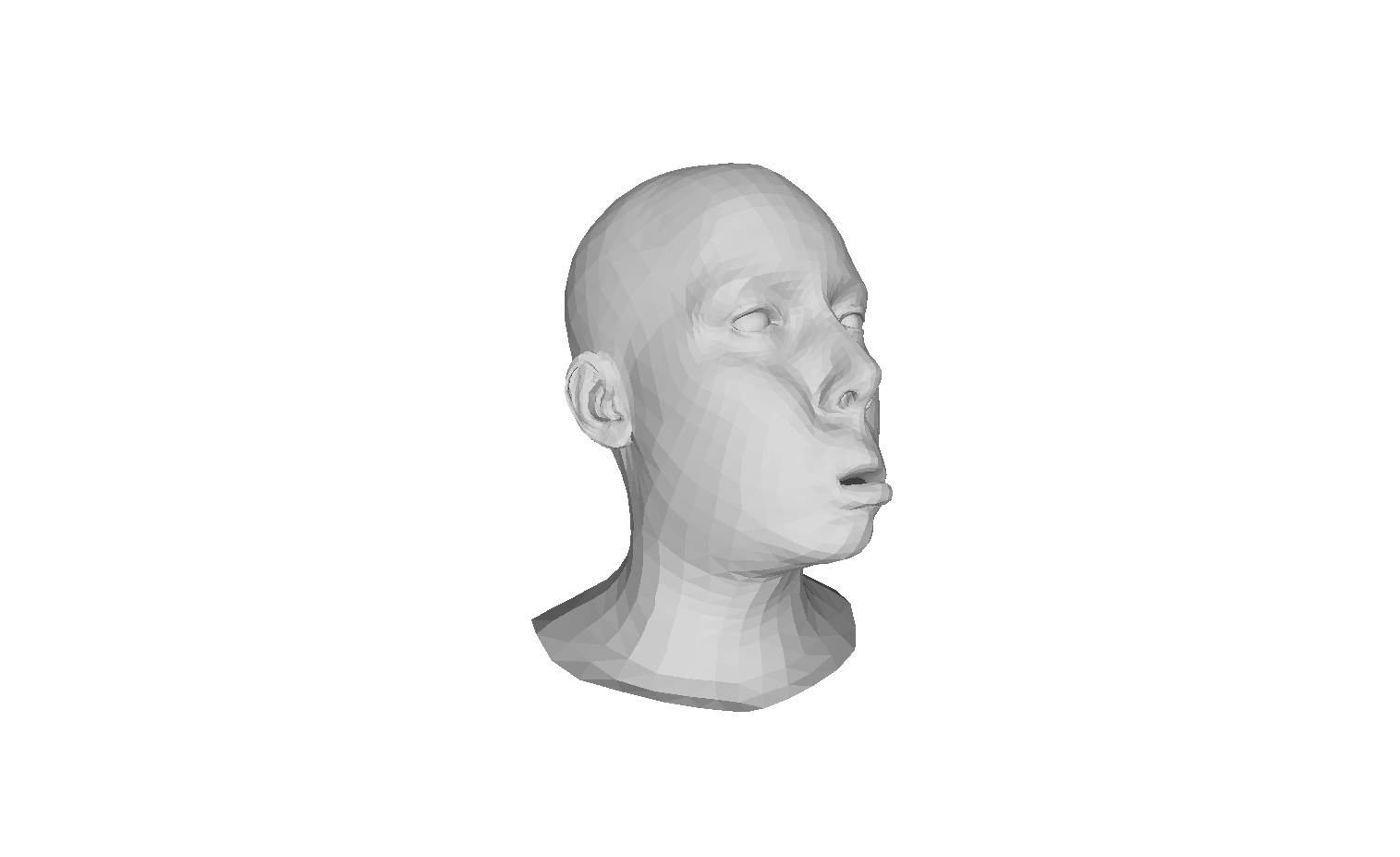}};
    \node[right of=d6, node distance=1.6cm] (d7) {\includegraphics[trim={400 80 400 100},clip,width=0.09\linewidth]{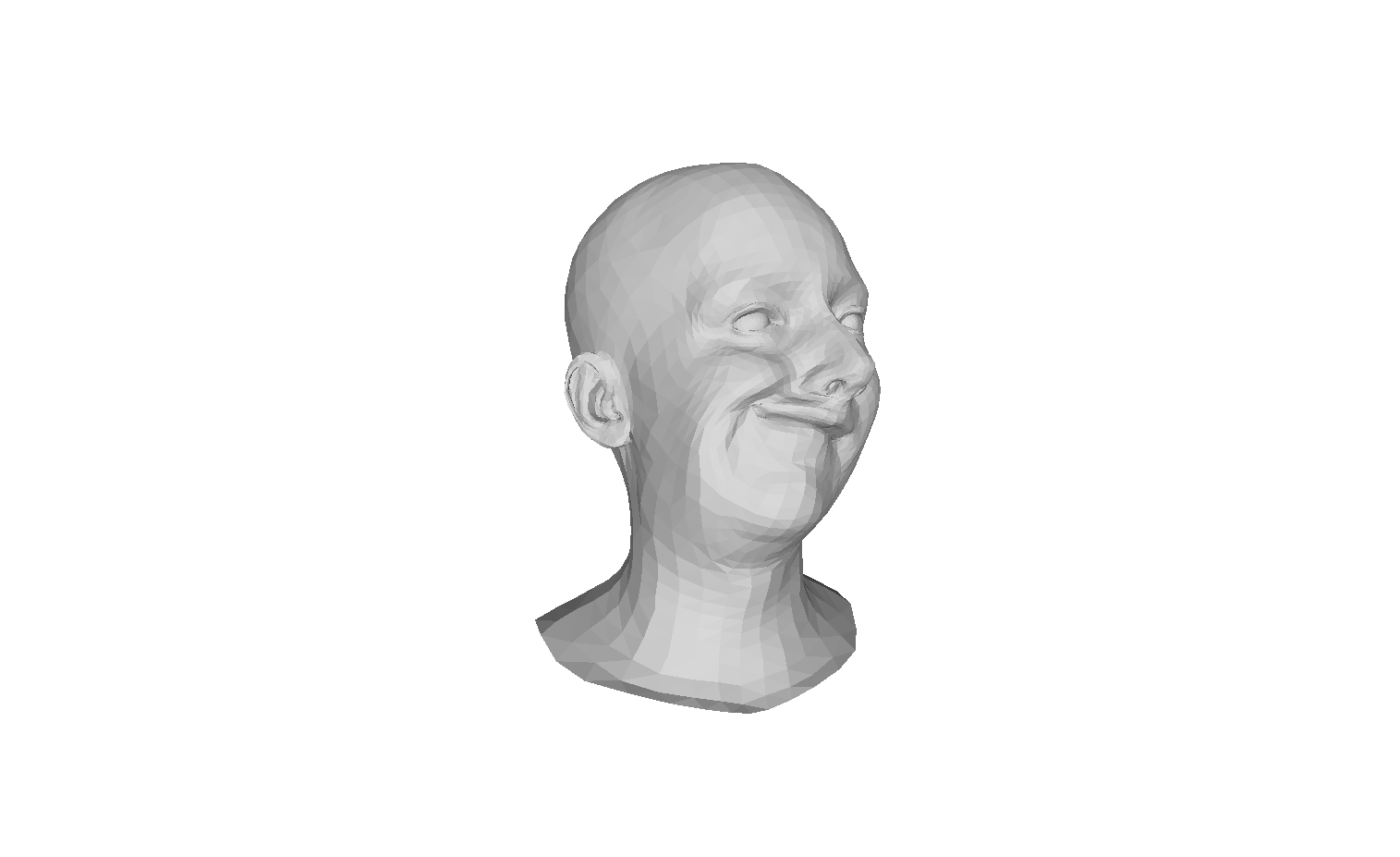}};
    \node[right of=d7, node distance=1.6cm] (d8) {\includegraphics[trim={400 80 400 100},clip,width=0.09\linewidth]{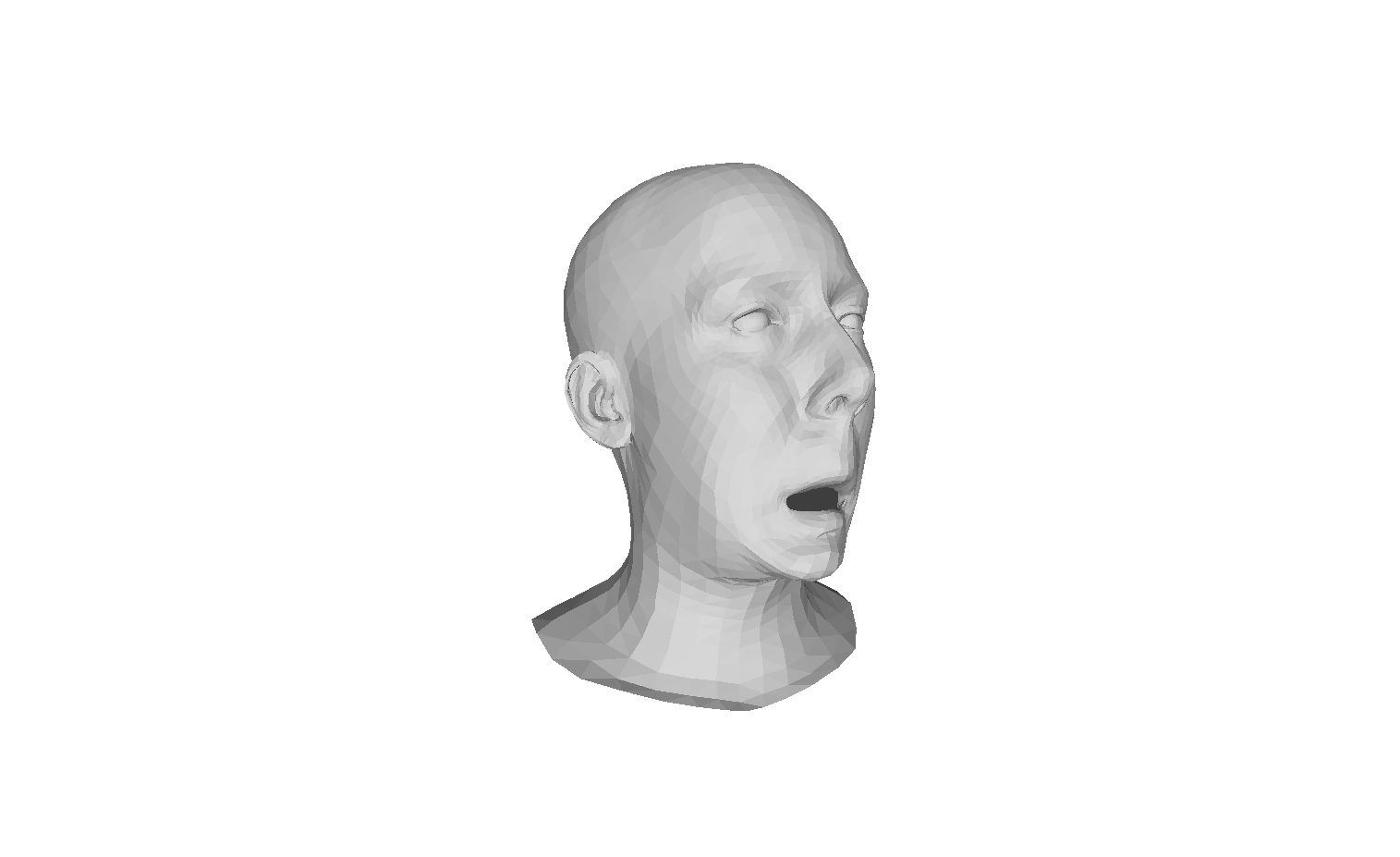}};
    \node[right of=d8, node distance=1.6cm] (d9) {\includegraphics[trim={400 80 400 100},clip,width=0.09\linewidth]{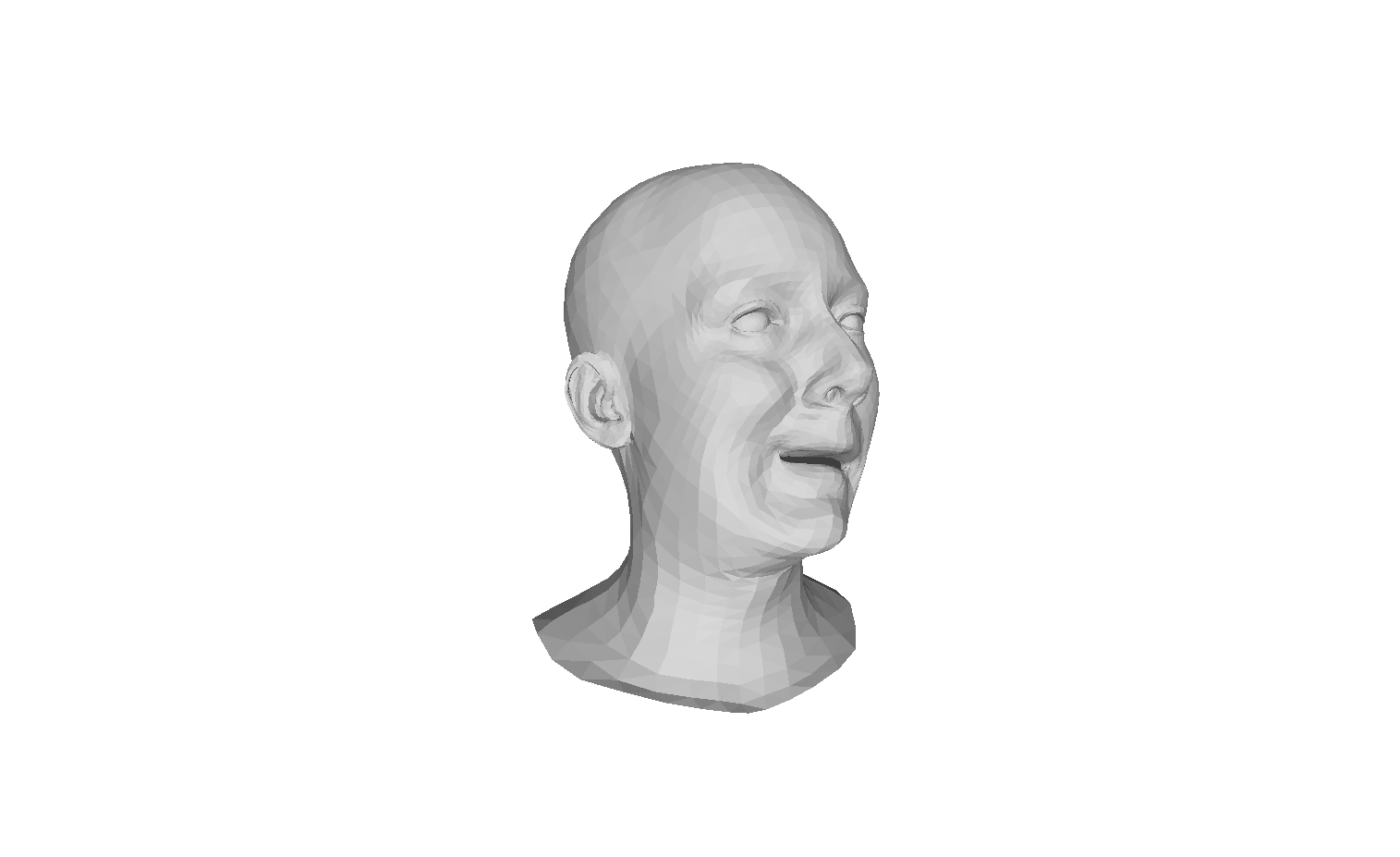}};
    
    \node[below of=d1, node distance=1.8cm] {Target Image};
    \node[below of=d2, node distance=1.8cm, text width=3cm, align=center] {Fitting by Global-local model};
    \node[below of=d6, node distance=1.8cm] {3D Reconstructions by \ourmethod{}};
    \end{tikzpicture}
    \caption{Set of 3D reconstructions by \ourmethod{} on real-world occluded face images.}
    \label{fig:realocc}
\end{figure*}
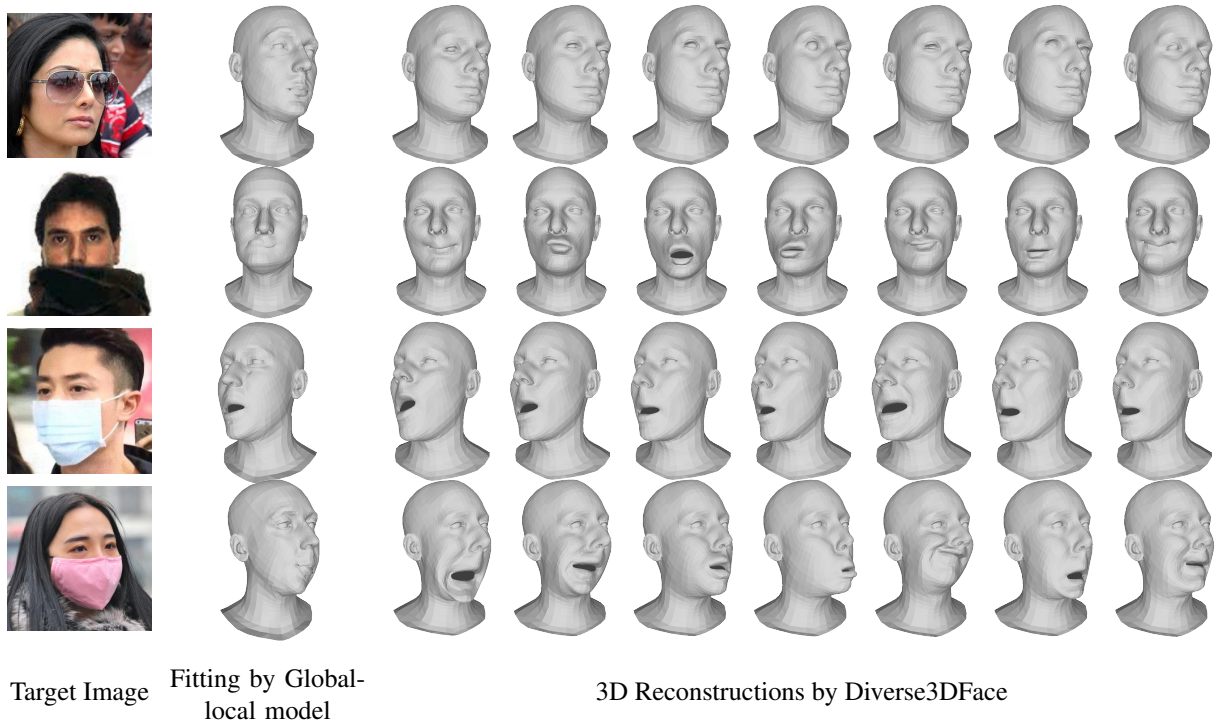

\begin{figure*}
    \centering
    \begin{tikzpicture}
    \node (a1) {\includegraphics[width=0.11\linewidth]{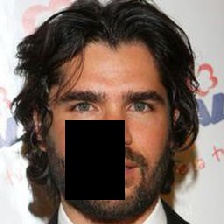}};
    \node[right of=a1, node distance=2.5cm] (a2) {\includegraphics[trim={400 80 400 100},clip,width=0.09\linewidth]{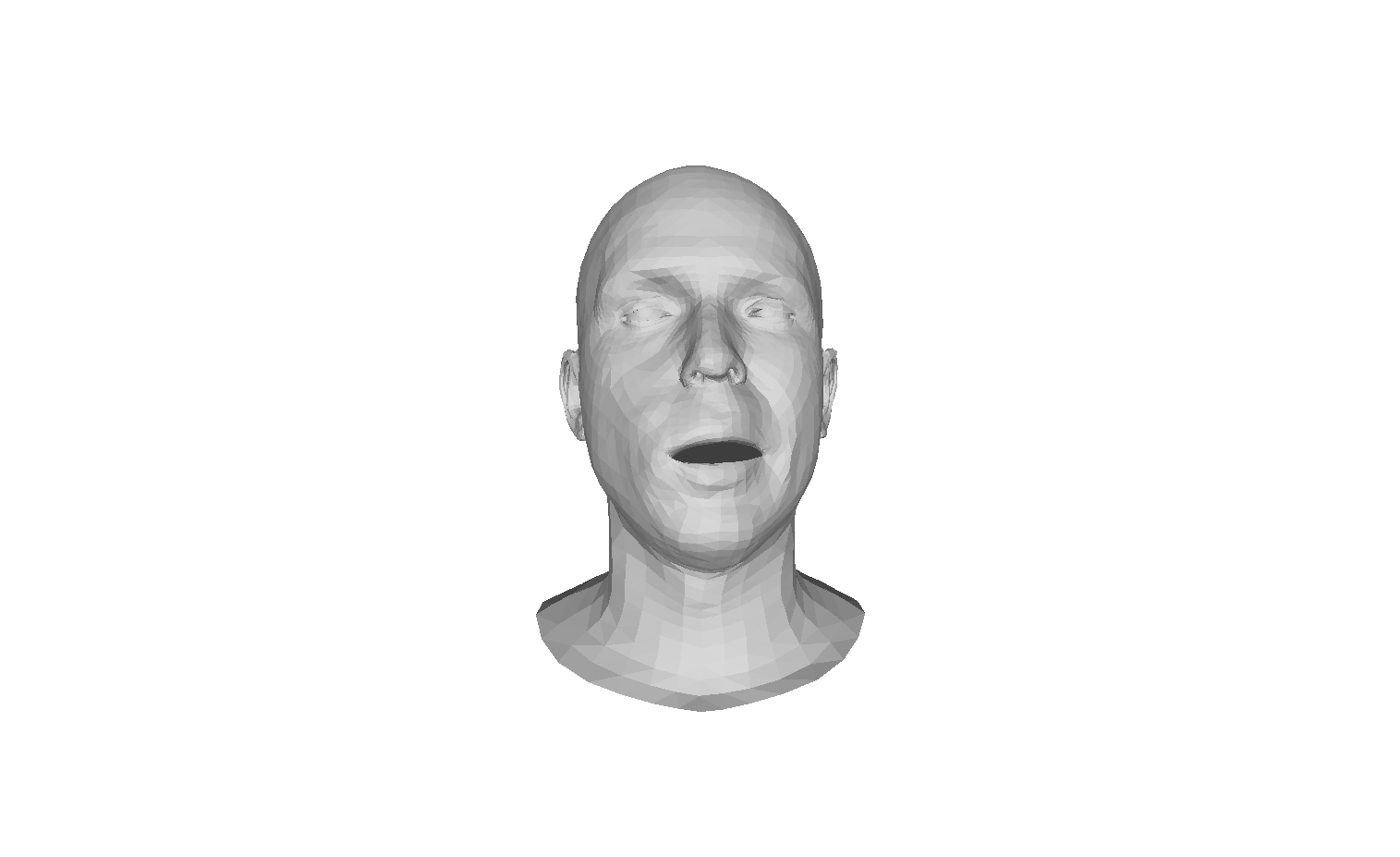}};
    \node[right of=a2, node distance=1.8cm] (a3) {\includegraphics[trim={400 80 400 100},clip,width=0.09\linewidth]{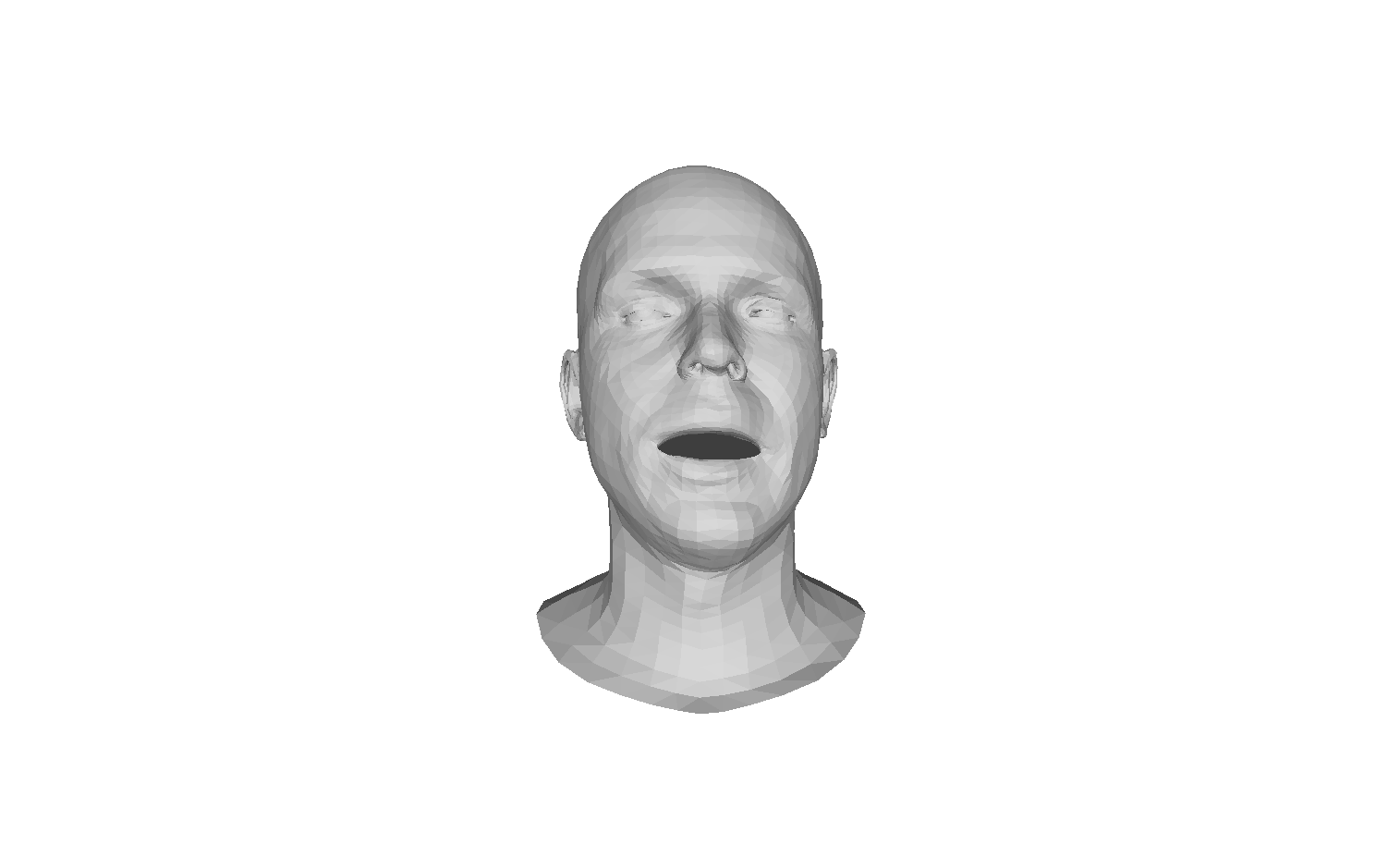}};
    \node[right of=a3, node distance=1.8cm] (a4) {\includegraphics[trim={400 80 400 100},clip,width=0.09\linewidth]{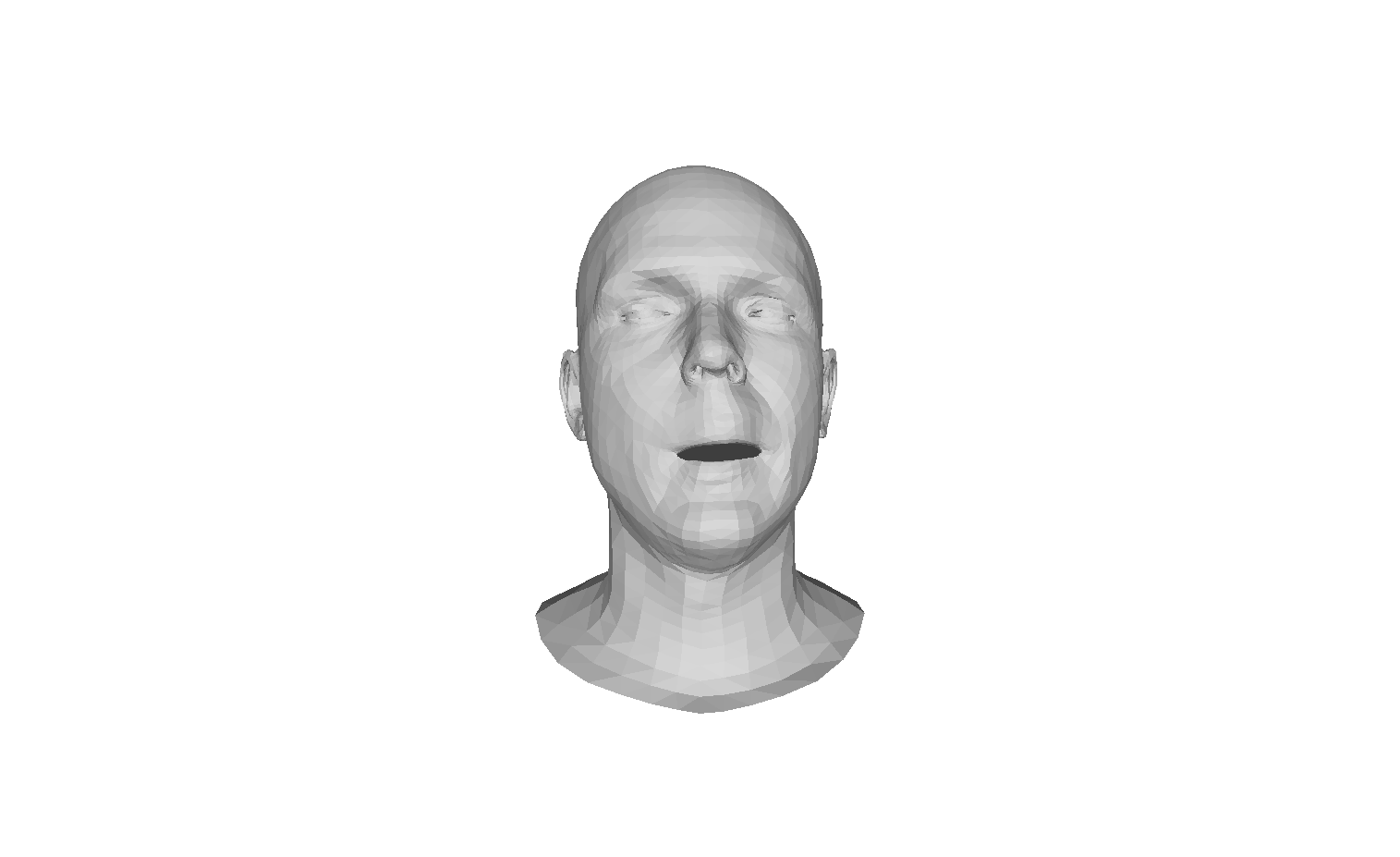}};
    \node[right of=a4, node distance=1.8cm] (a5) {\includegraphics[trim={400 80 400 100},clip,width=0.09\linewidth]{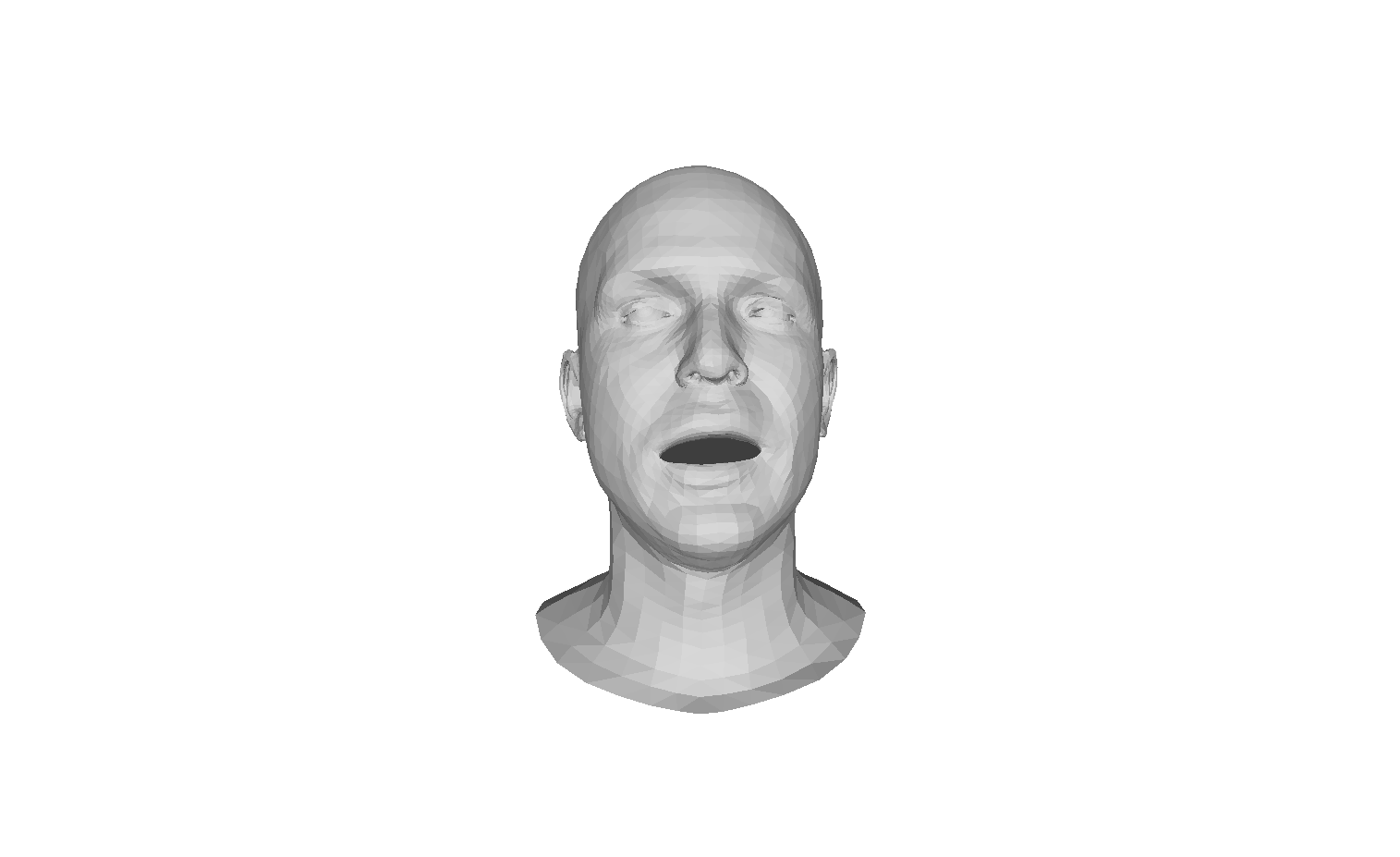}};
    \node[right of=a5, node distance=1.8cm] (a6) {\includegraphics[trim={400 80 400 100},clip,width=0.09\linewidth]{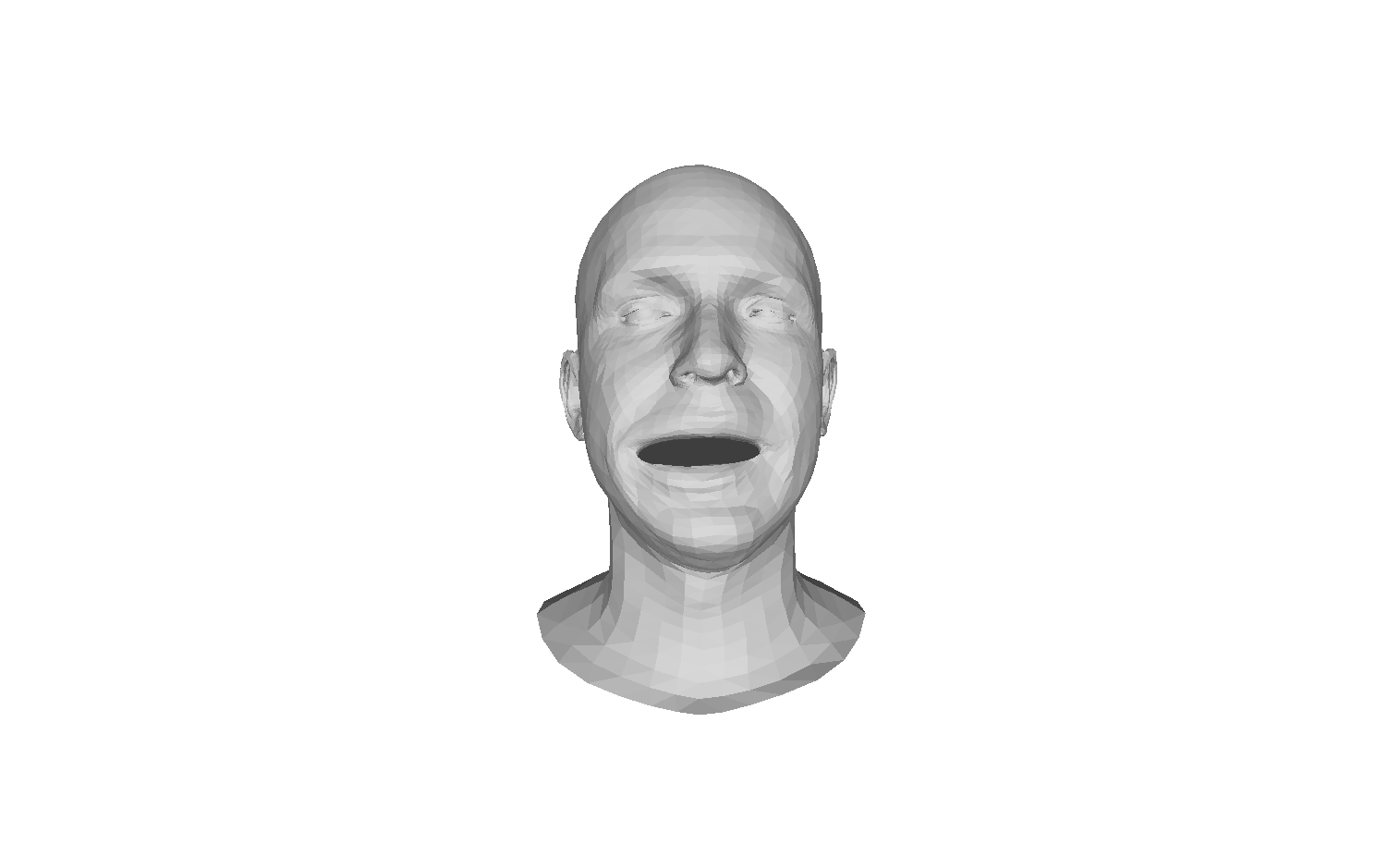}};
    \node[right of=a6, node distance=1.8cm] (a7) {\includegraphics[trim={400 80 400 100},clip,width=0.09\linewidth]{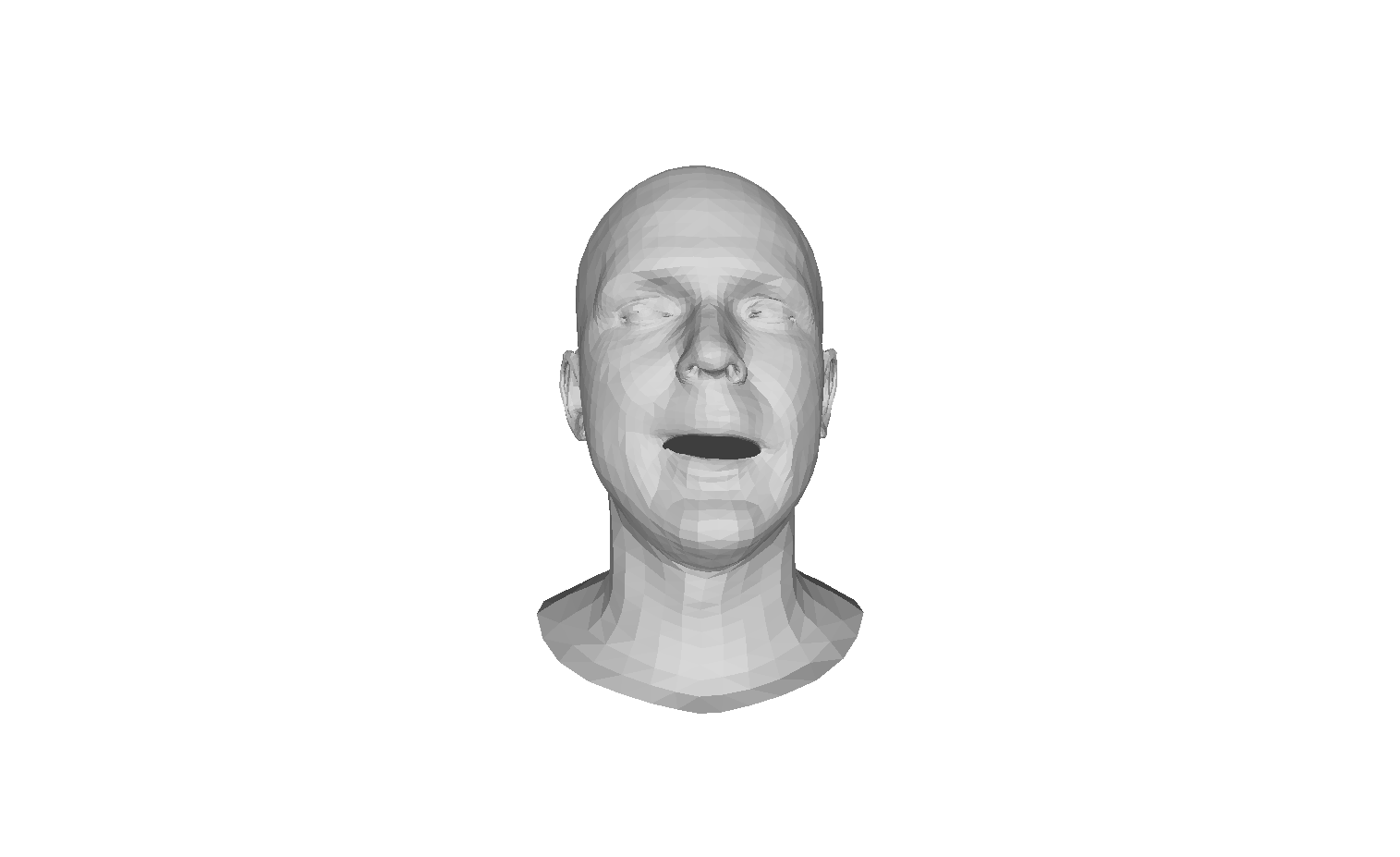}};
    \node[right of=a7, node distance=1.8cm] (a8) {\includegraphics[trim={400 80 400 100},clip,width=0.09\linewidth]{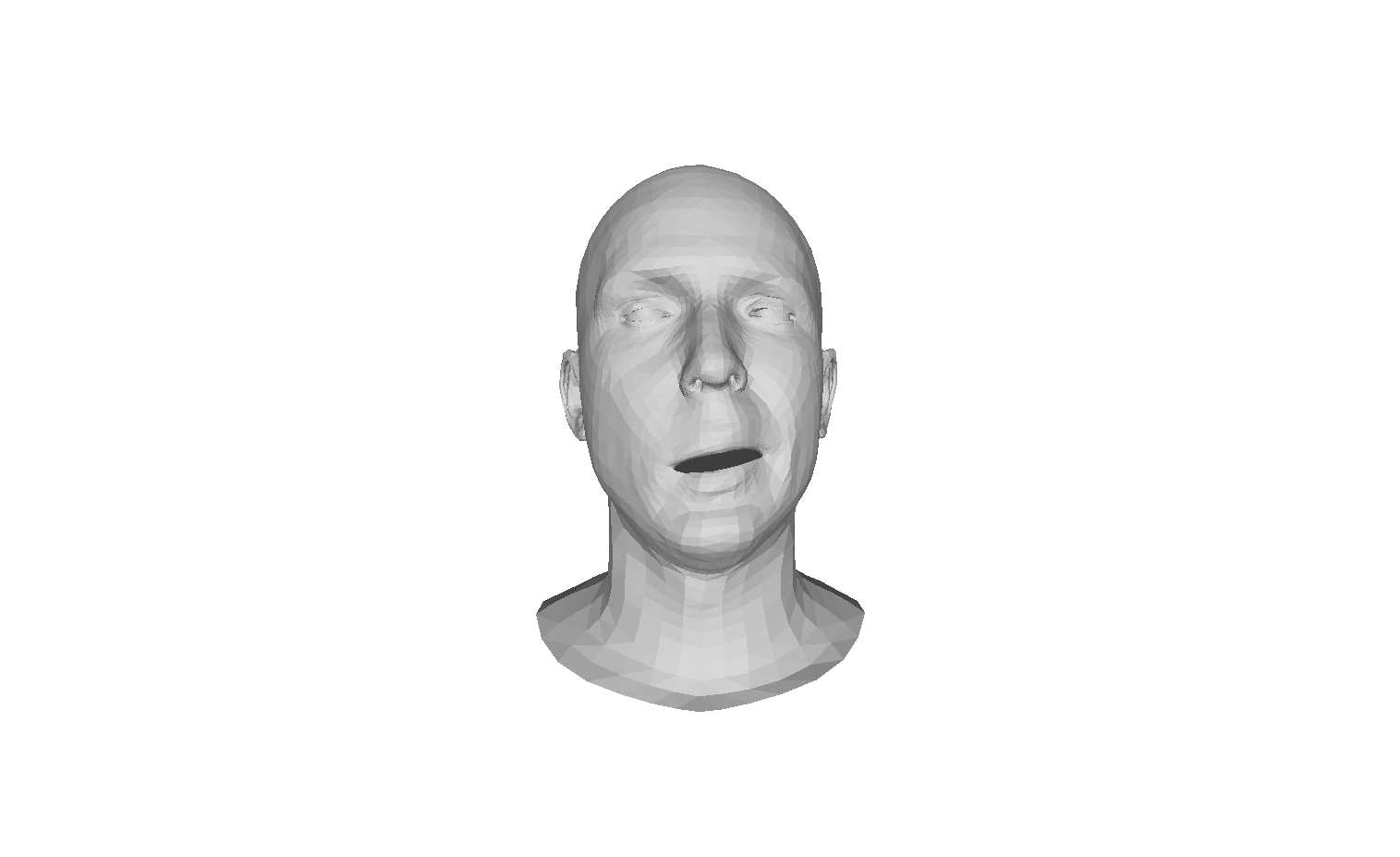}};
    \node[right of=a8, node distance=1.8cm] (a9) {\includegraphics[trim={400 80 400 100},clip,width=0.09\linewidth]{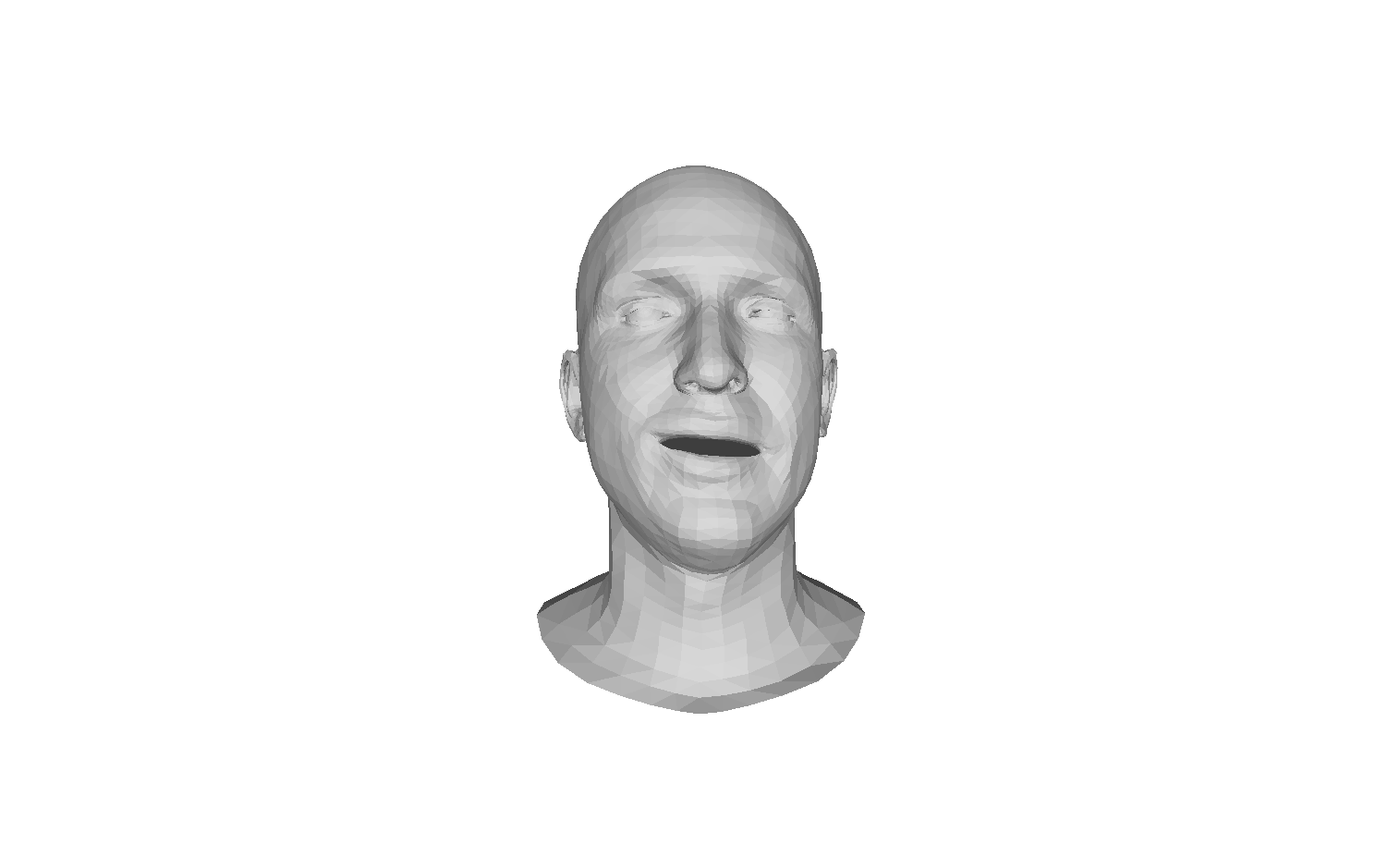}};
    
    \node[below of=a1, node distance=2.1cm] (b1) {\includegraphics[width=0.11\linewidth]{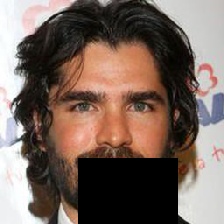}};
    \node[right of=b1, node distance=2.5cm] (b2) {\includegraphics[trim={400 80 400 100},clip,width=0.09\linewidth]{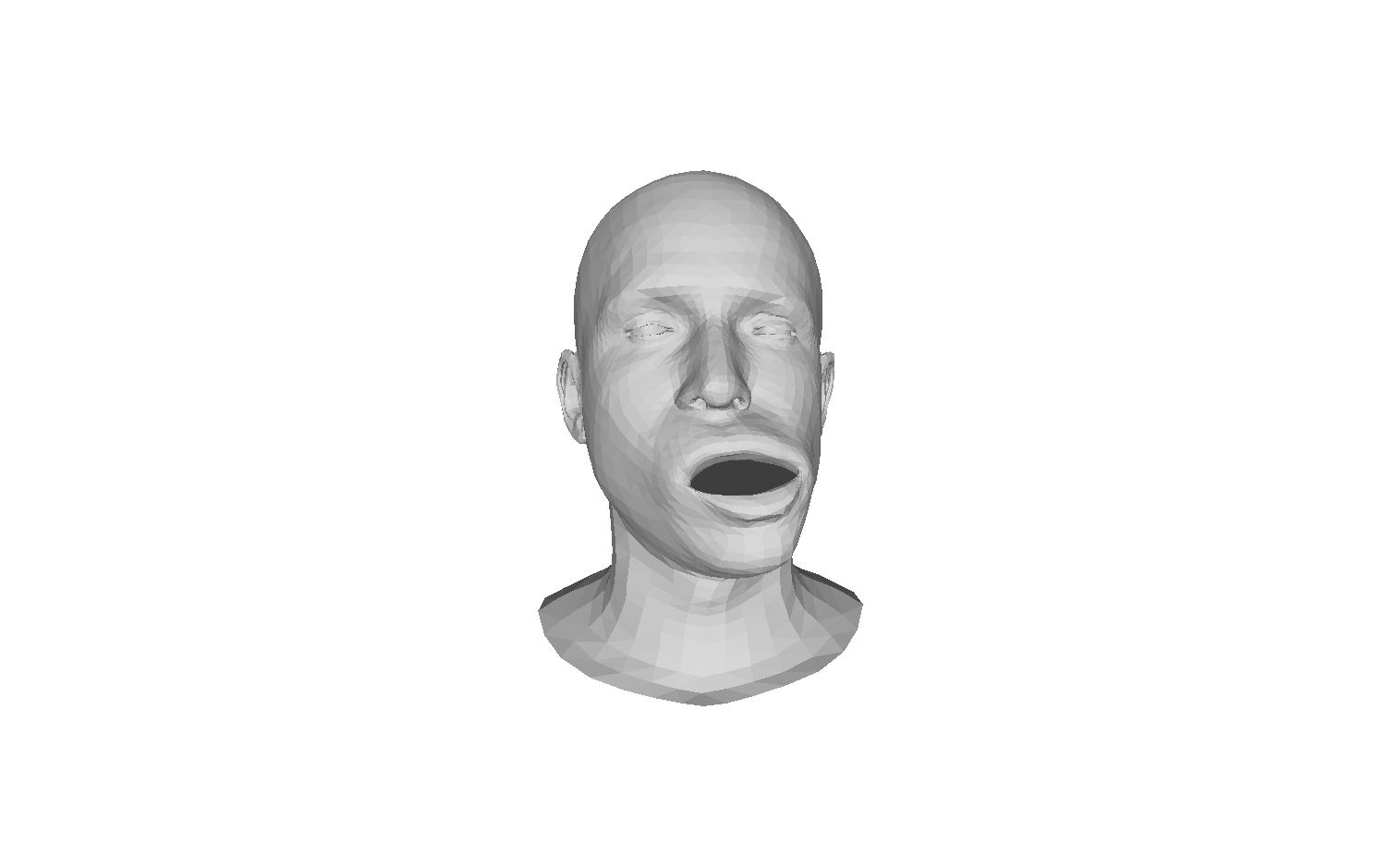}};
    \node[right of=b2, node distance=1.8cm] (b3) {\includegraphics[trim={400 80 400 100},clip,width=0.09\linewidth]{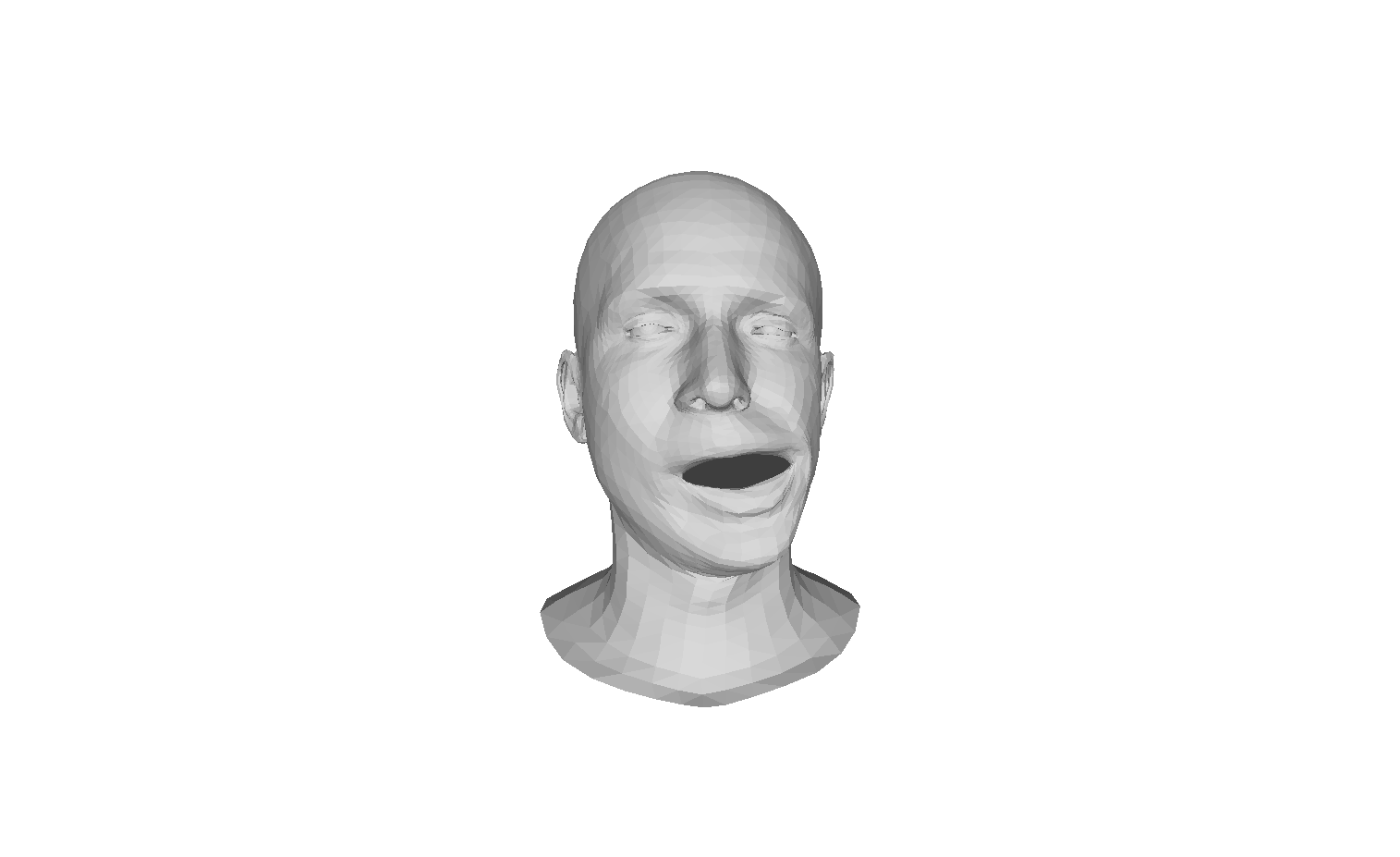}};
    \node[right of=b3, node distance=1.8cm] (b4) {\includegraphics[trim={400 80 400 100},clip,width=0.09\linewidth]{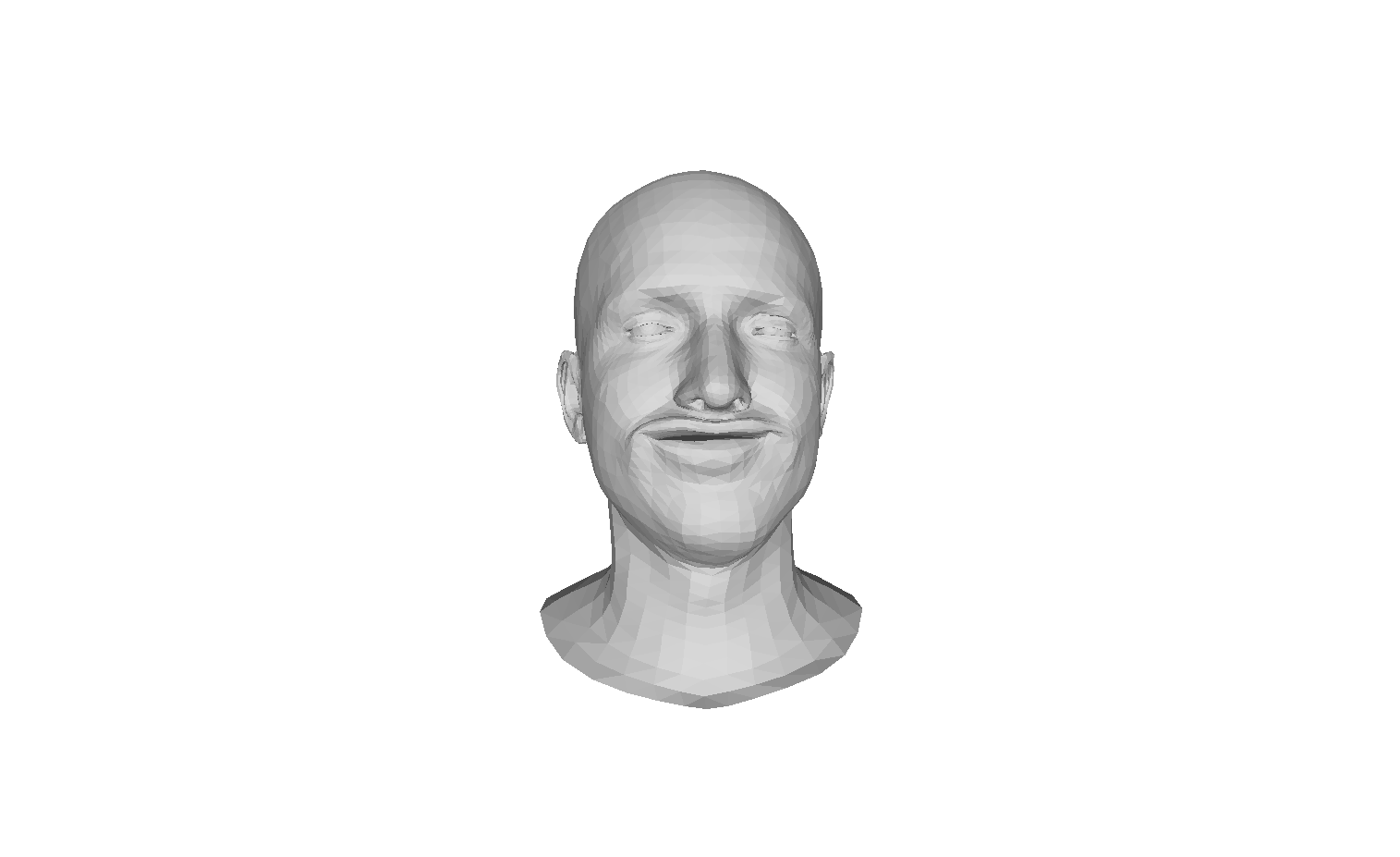}};
    \node[right of=b4, node distance=1.8cm] (b5) {\includegraphics[trim={400 80 400 100},clip,width=0.09\linewidth]{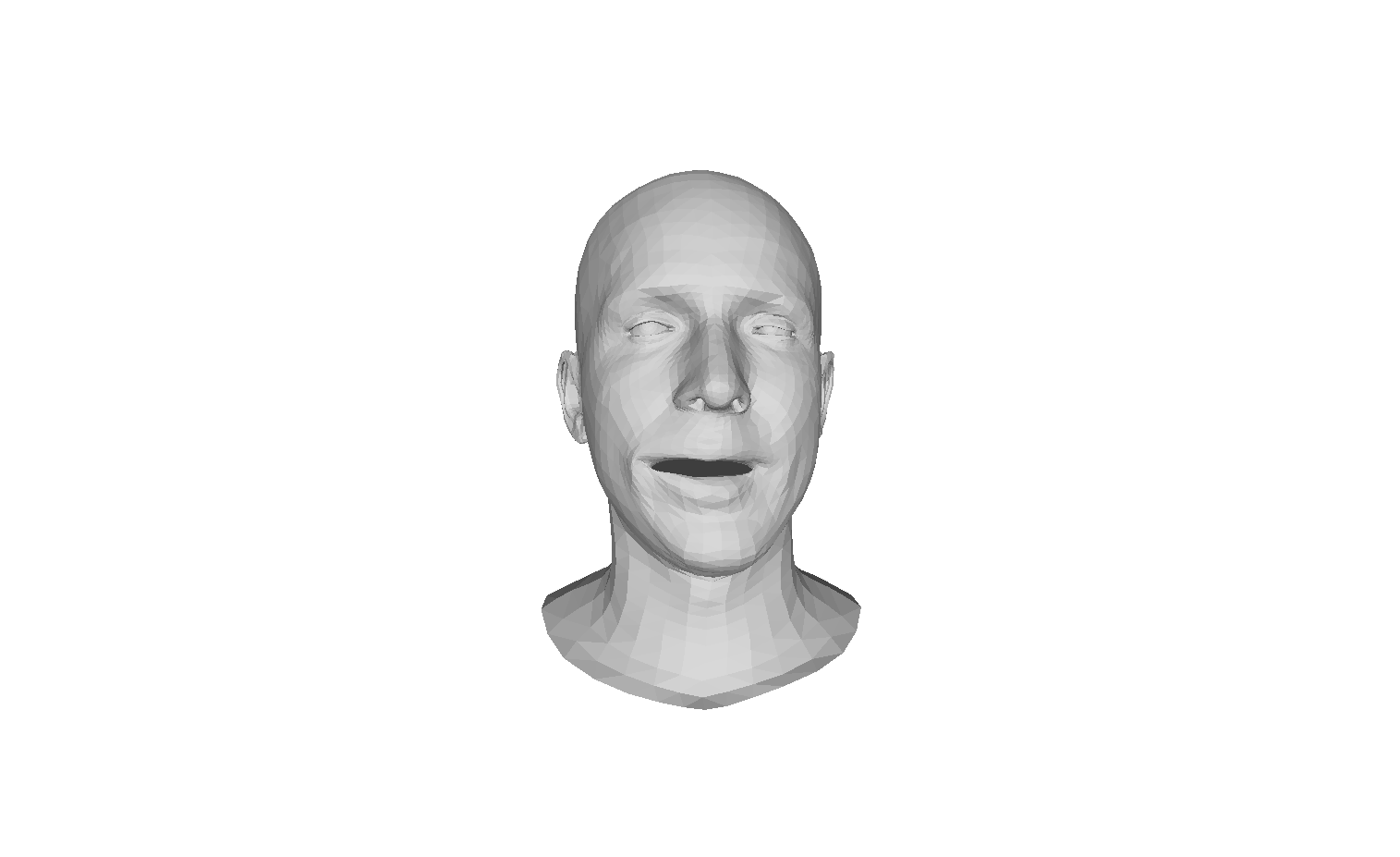}};
    \node[right of=b5, node distance=1.8cm] (b6) {\includegraphics[trim={400 80 400 100},clip,width=0.09\linewidth]{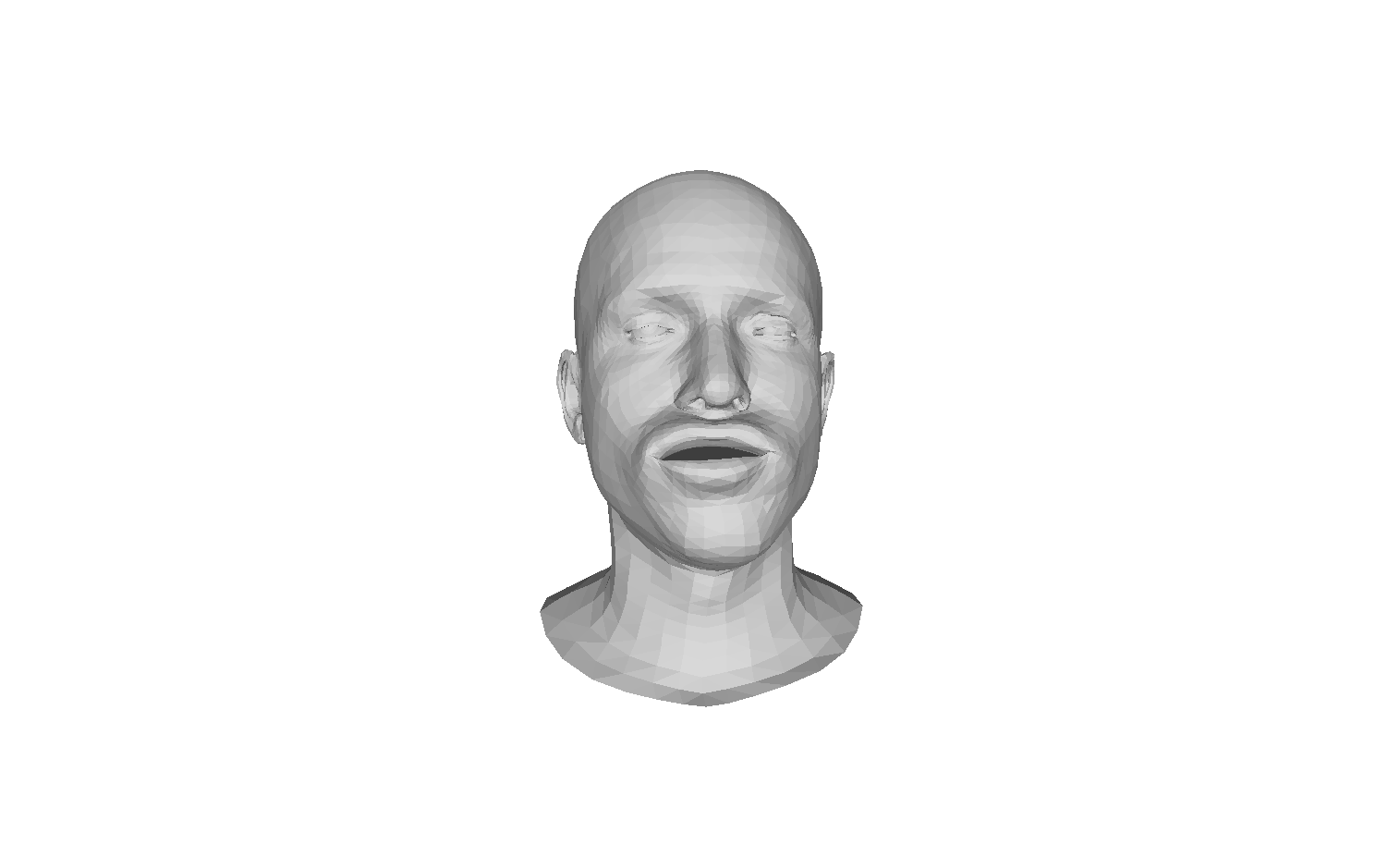}};
    \node[right of=b6, node distance=1.8cm] (b7) {\includegraphics[trim={400 80 400 100},clip,width=0.09\linewidth]{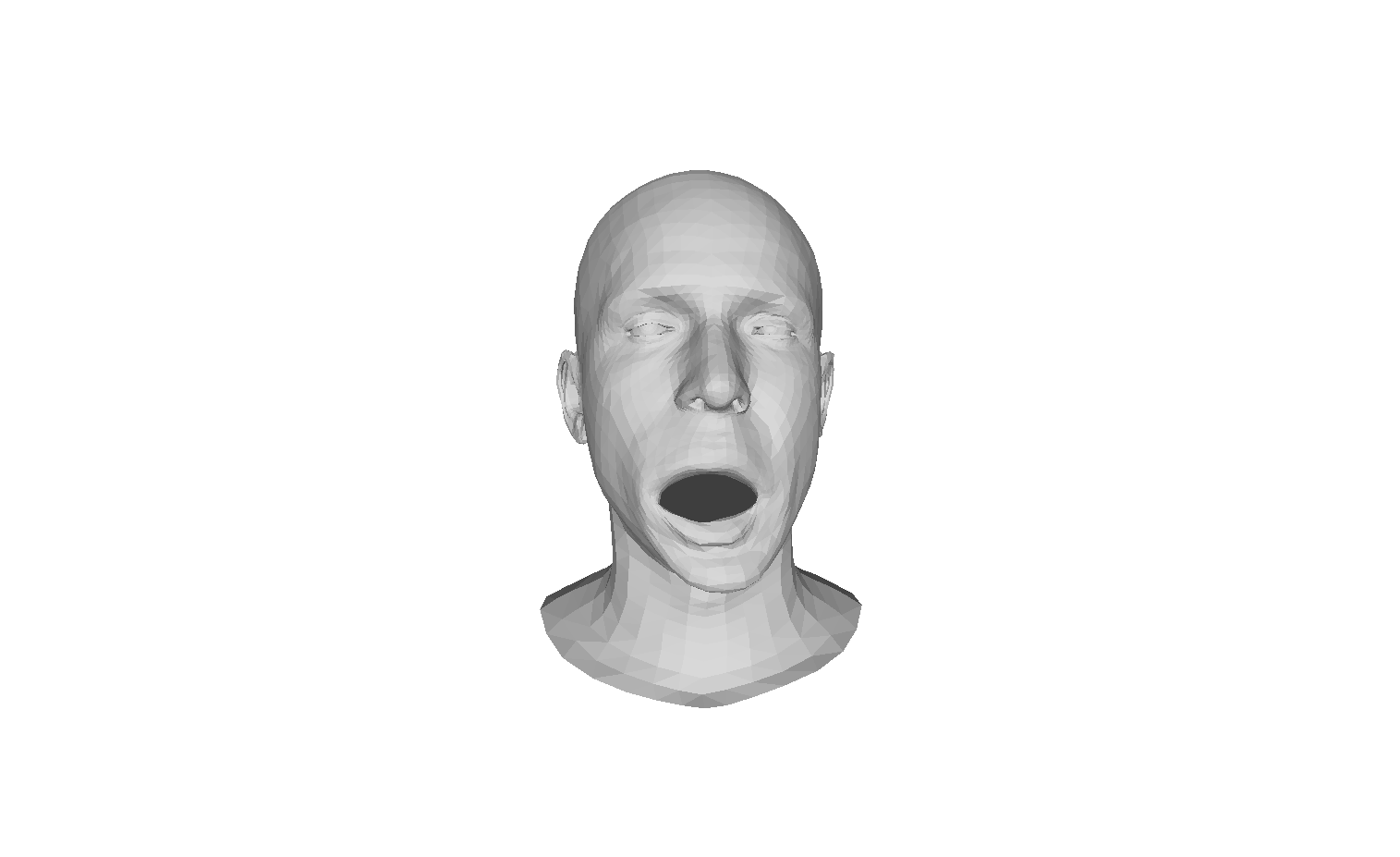}};
    \node[right of=b7, node distance=1.8cm] (b8) {\includegraphics[trim={400 80 400 100},clip,width=0.09\linewidth]{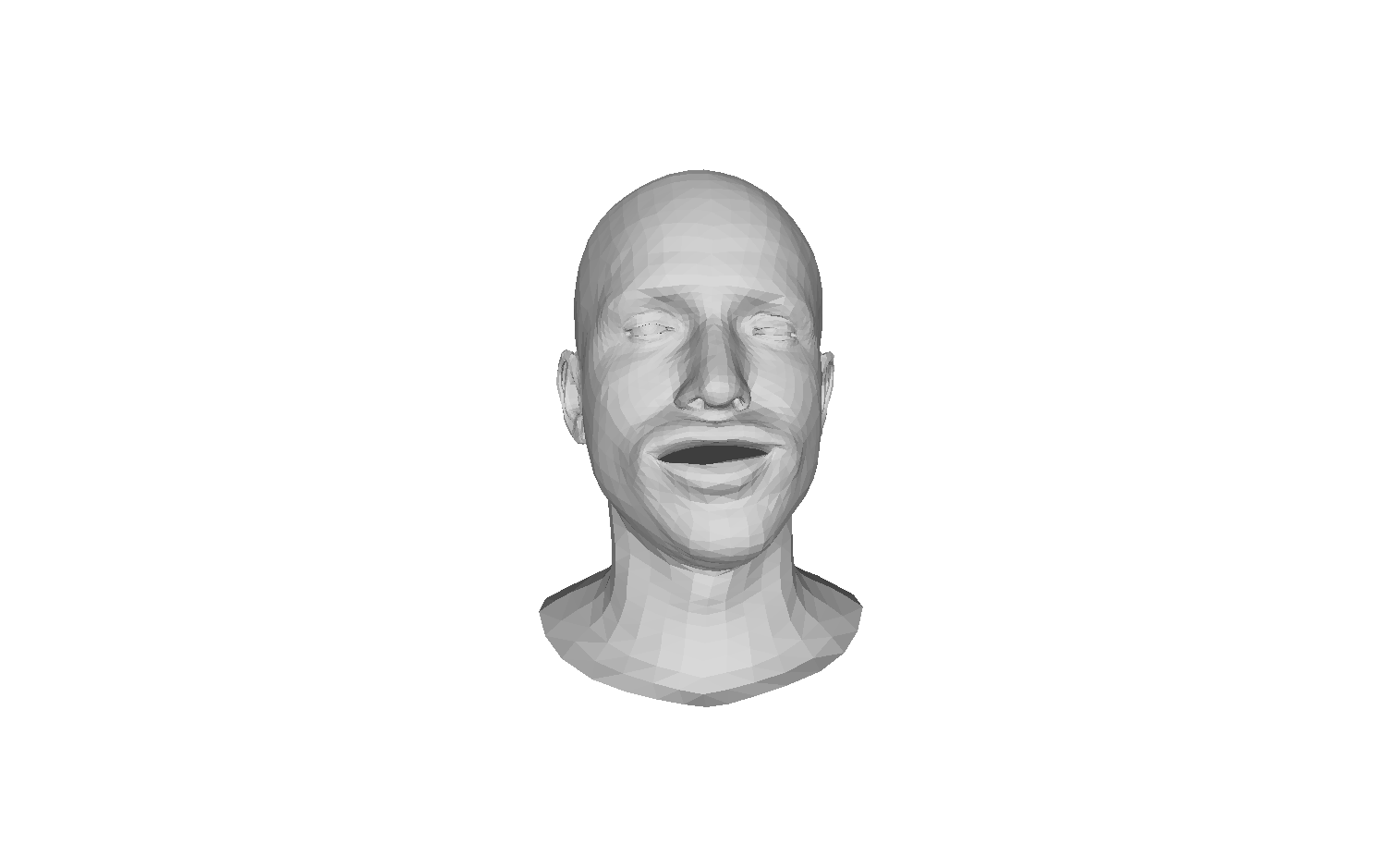}};
    \node[right of=b8, node distance=1.8cm] (b9) {\includegraphics[trim={400 80 400 100},clip,width=0.09\linewidth]{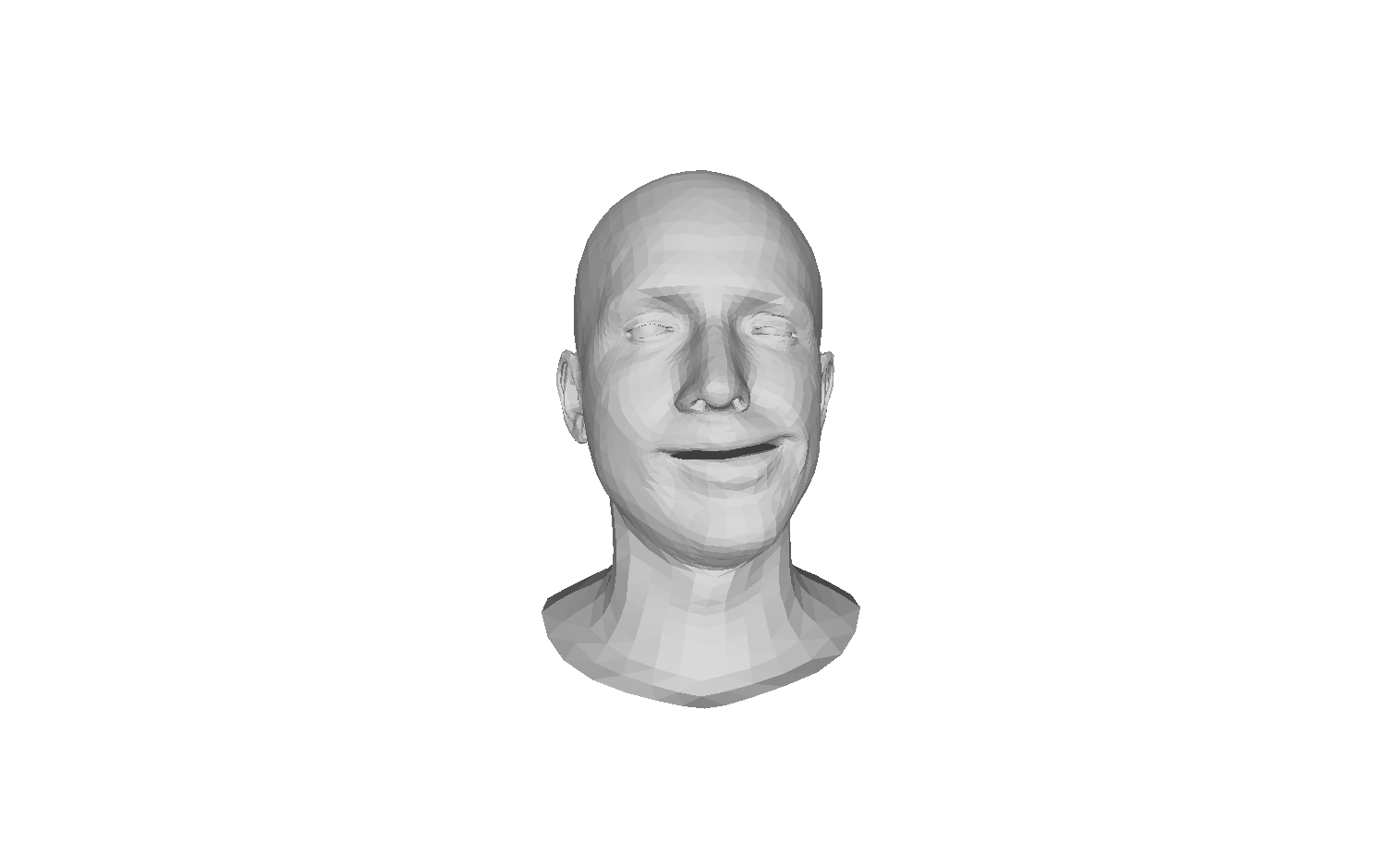}};
    
    \node[below of=b1, node distance=2.1cm] (c1) {\includegraphics[width=0.11\linewidth]{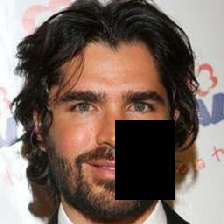}};
    \node[right of=c1, node distance=2.5cm] (c2) {\includegraphics[trim={400 80 400 100},clip,width=0.09\linewidth]{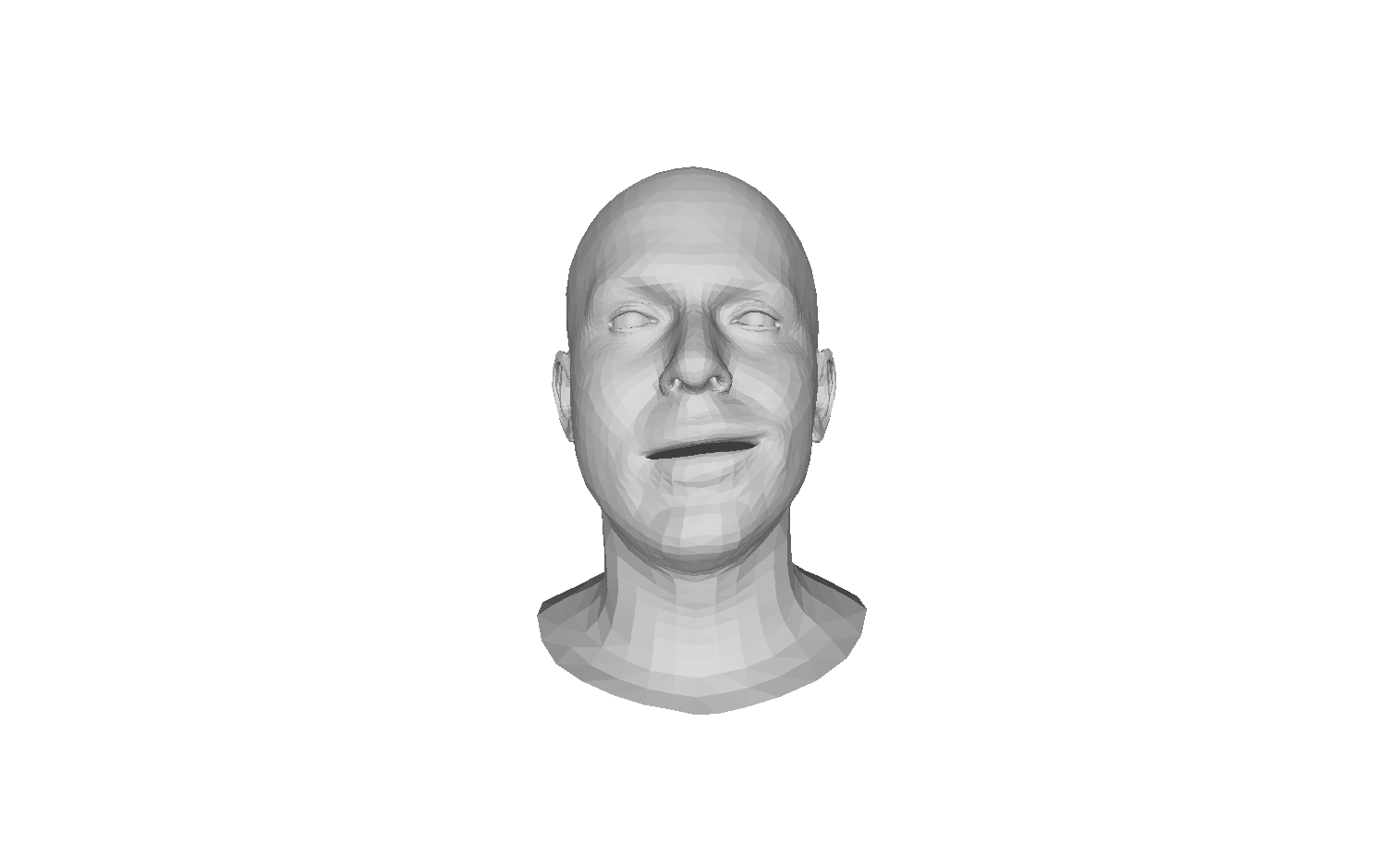}};
    \node[right of=c2, node distance=1.8cm] (c3) {\includegraphics[trim={400 80 400 100},clip,width=0.09\linewidth]{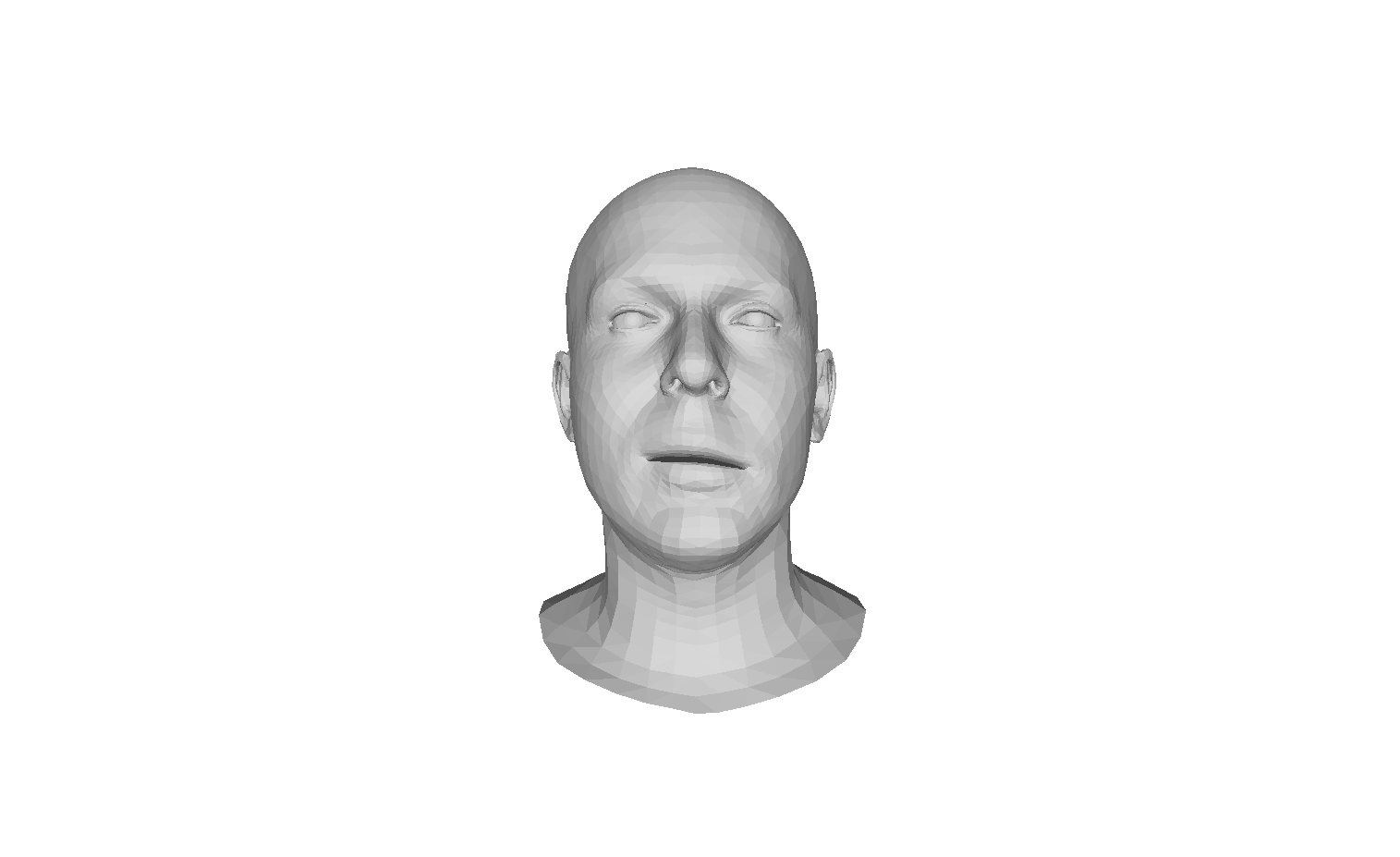}};
    \node[right of=c3, node distance=1.8cm] (c4) {\includegraphics[trim={400 80 400 100},clip,width=0.09\linewidth]{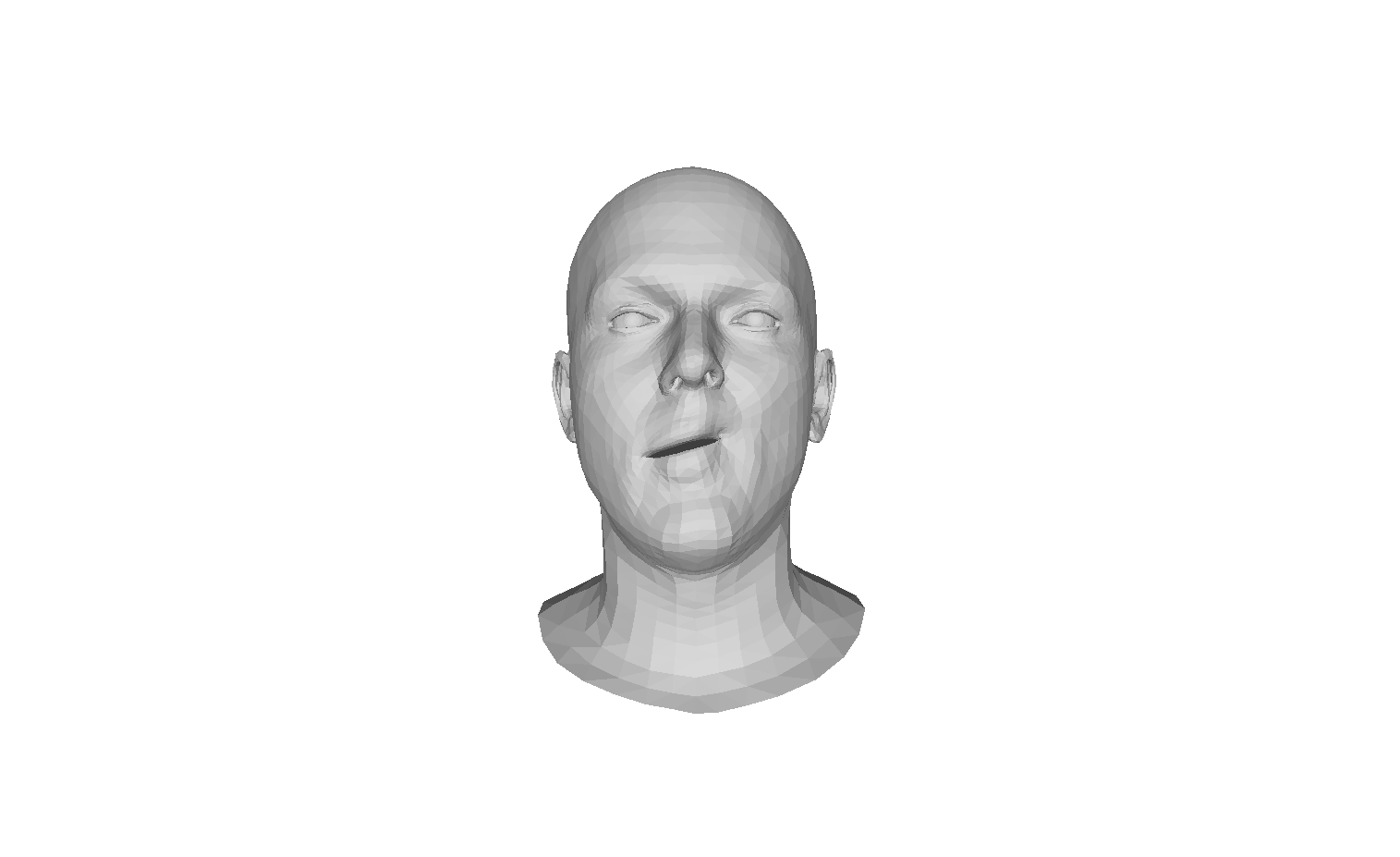}};
    \node[right of=c4, node distance=1.8cm] (c5) {\includegraphics[trim={400 80 400 100},clip,width=0.09\linewidth]{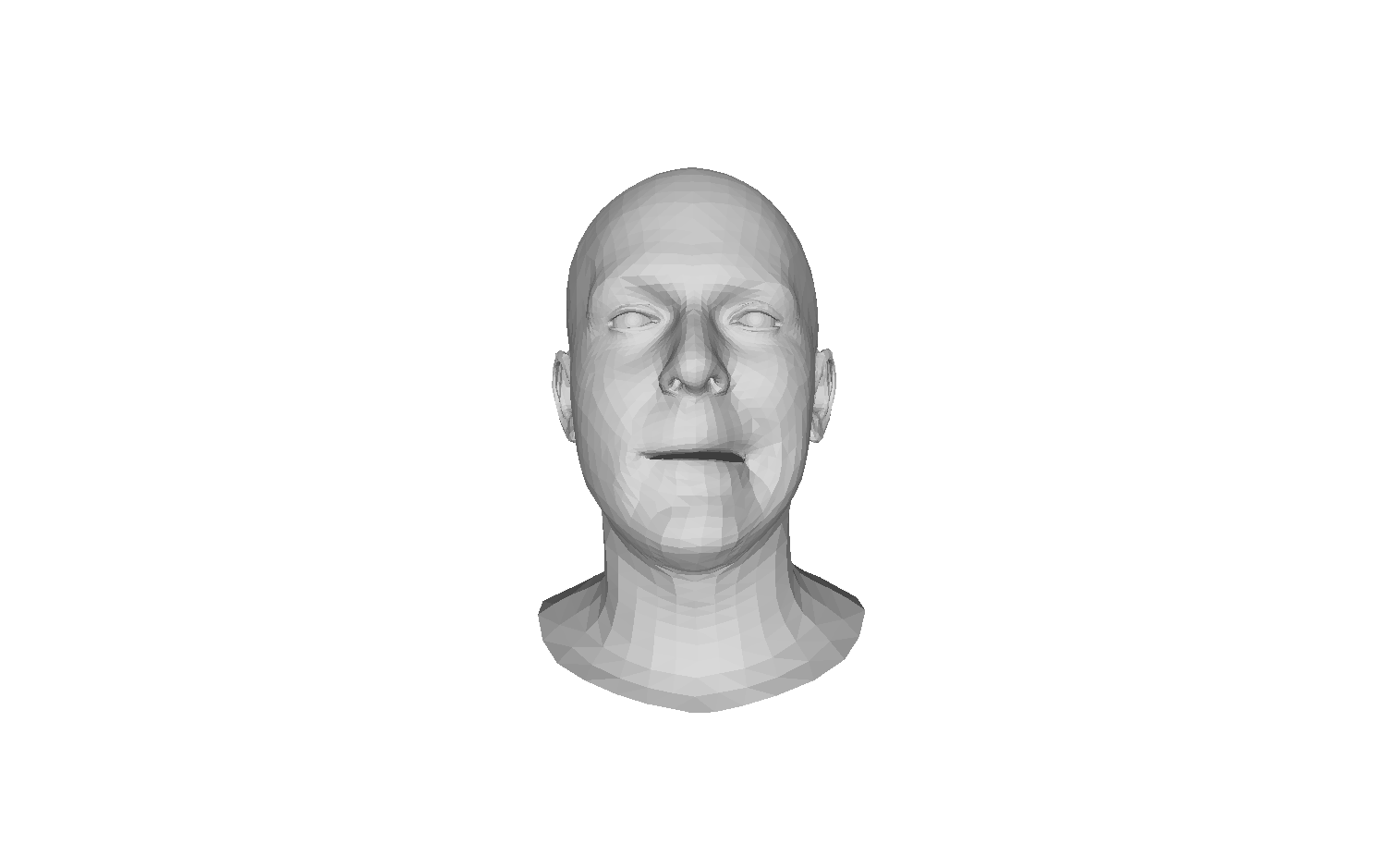}};
    \node[right of=c5, node distance=1.8cm] (c6) {\includegraphics[trim={400 80 400 100},clip,width=0.09\linewidth]{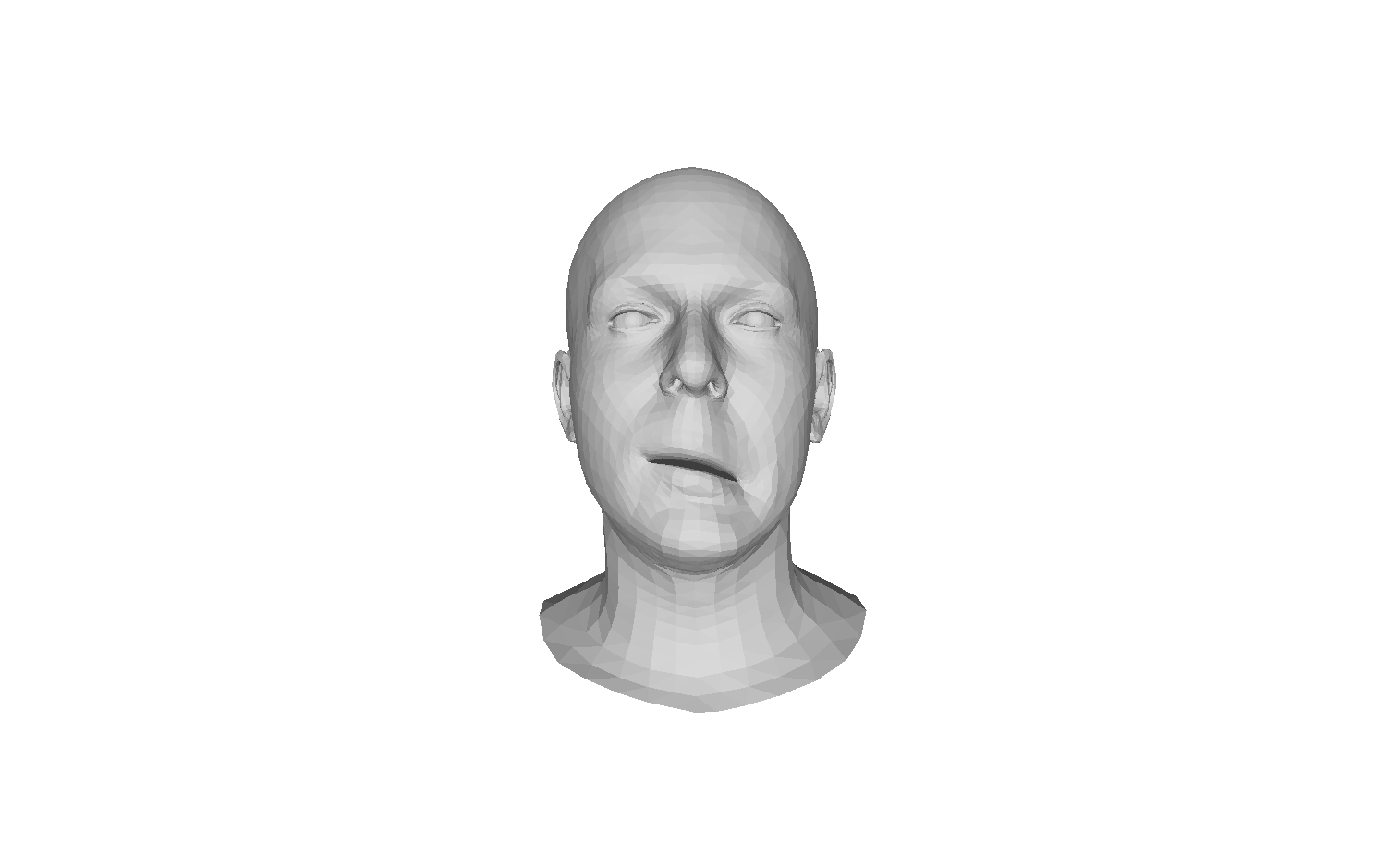}};
    \node[right of=c6, node distance=1.8cm] (c7) {\includegraphics[trim={400 80 400 100},clip,width=0.09\linewidth]{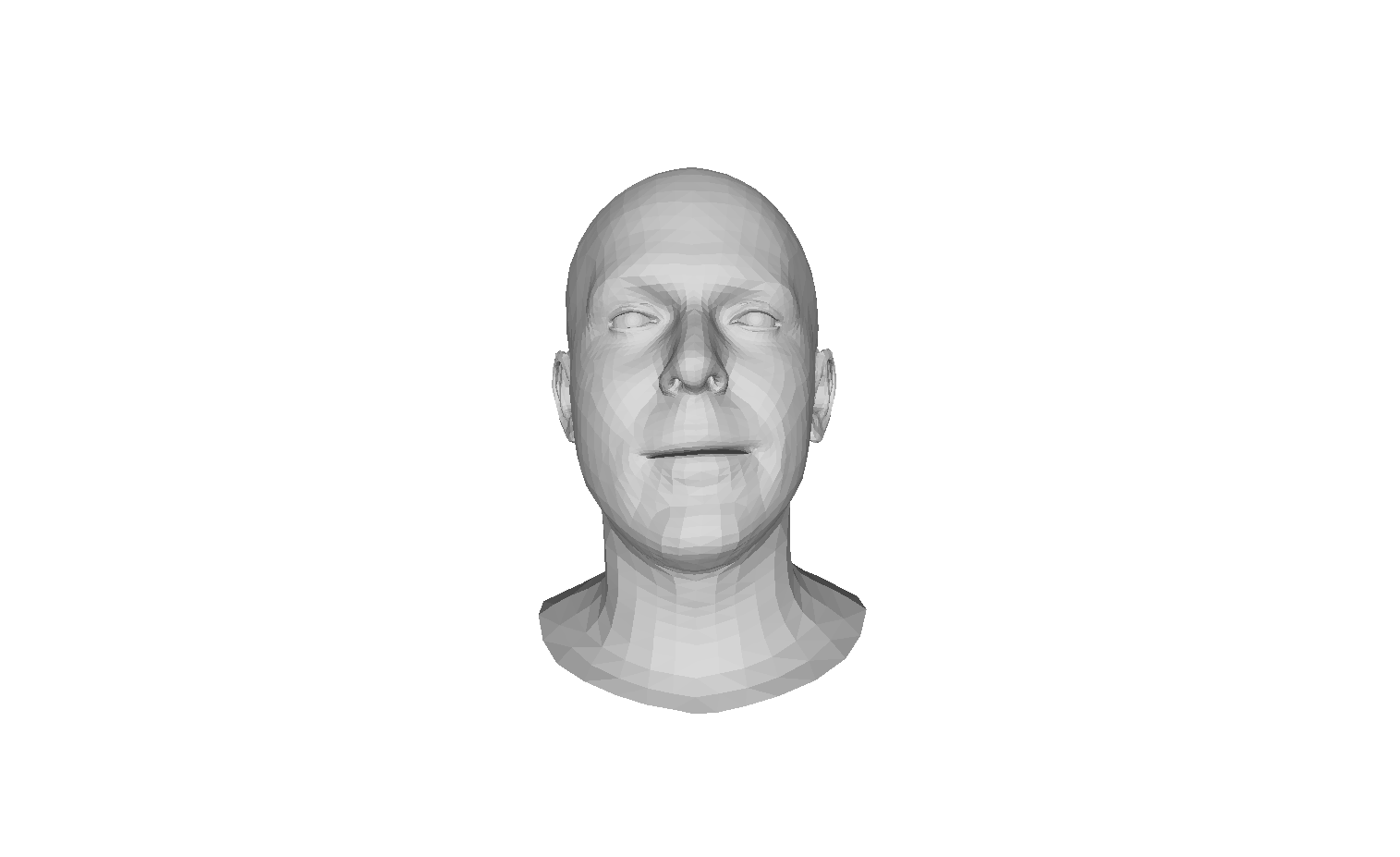}};
    \node[right of=c7, node distance=1.8cm] (c8) {\includegraphics[trim={400 80 400 100},clip,width=0.09\linewidth]{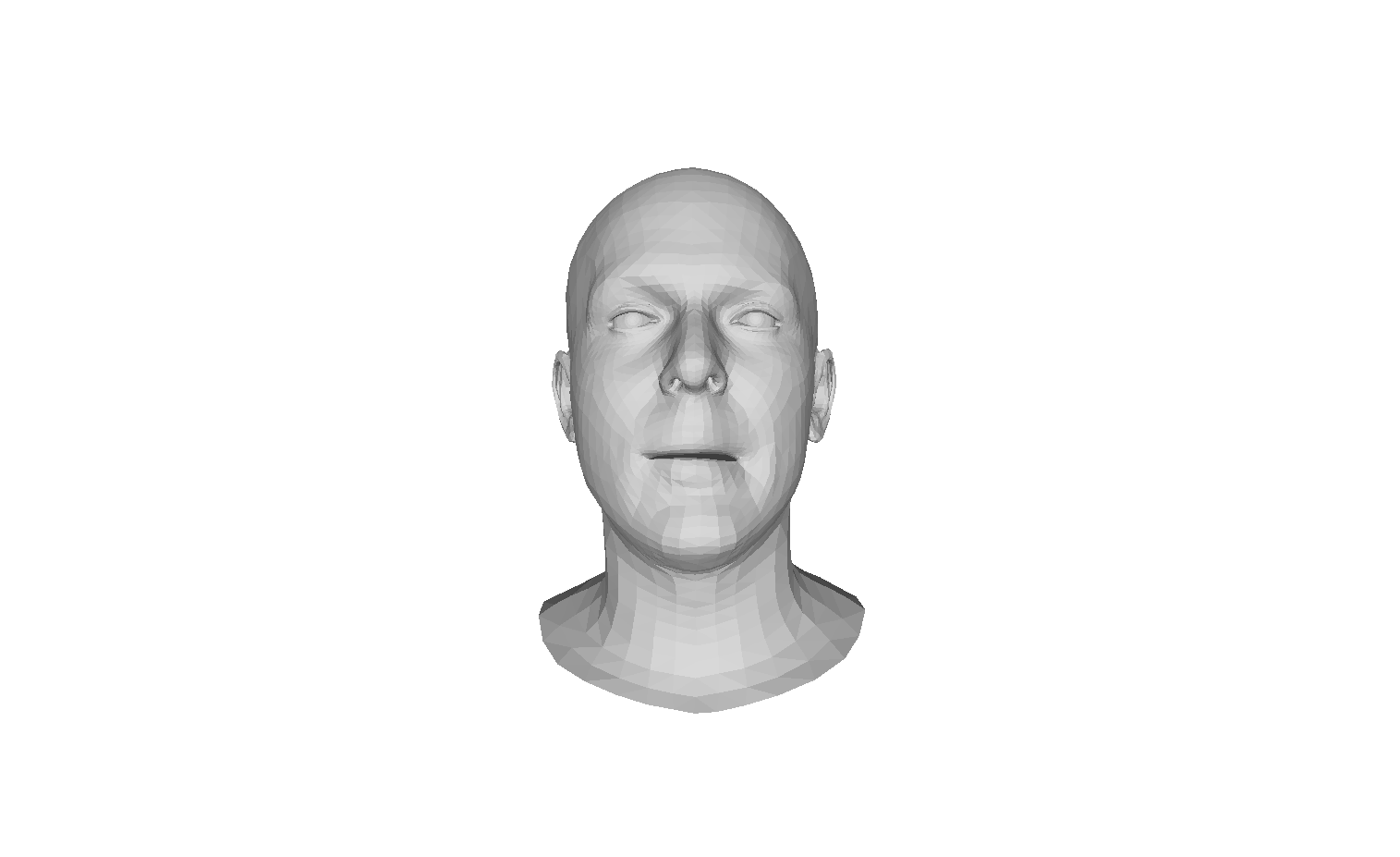}};
    \node[right of=c8, node distance=1.8cm] (c9) {\includegraphics[trim={400 80 400 100},clip,width=0.09\linewidth]{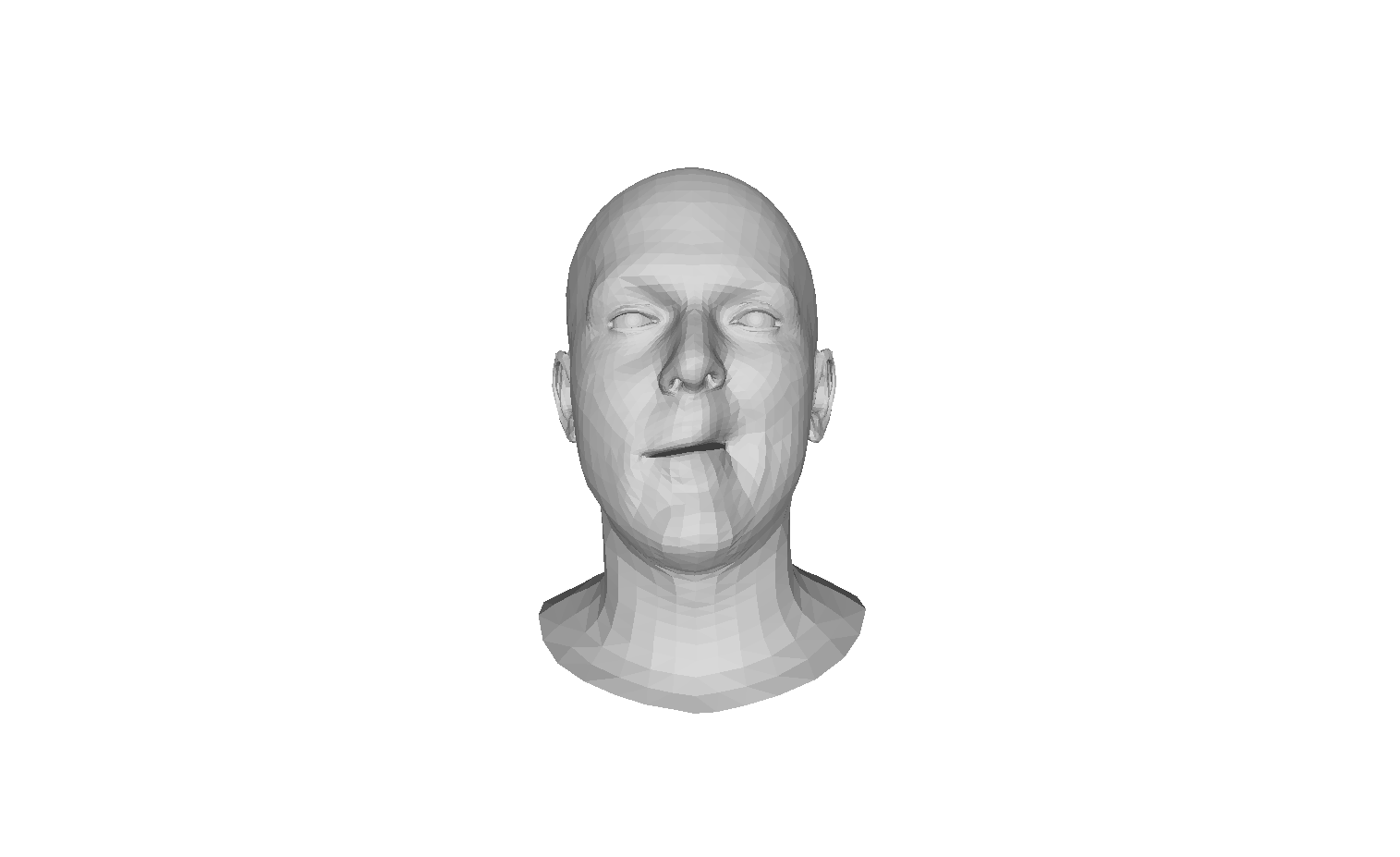}};
    
    \node[below of=c1, node distance=2.1cm] (d1) {\includegraphics[width=0.11\linewidth]{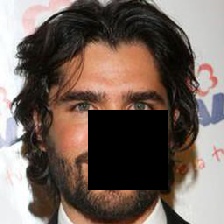}};
    \node[right of=d1, node distance=2.5cm] (d2) {\includegraphics[trim={400 80 400 100},clip,width=0.09\linewidth]{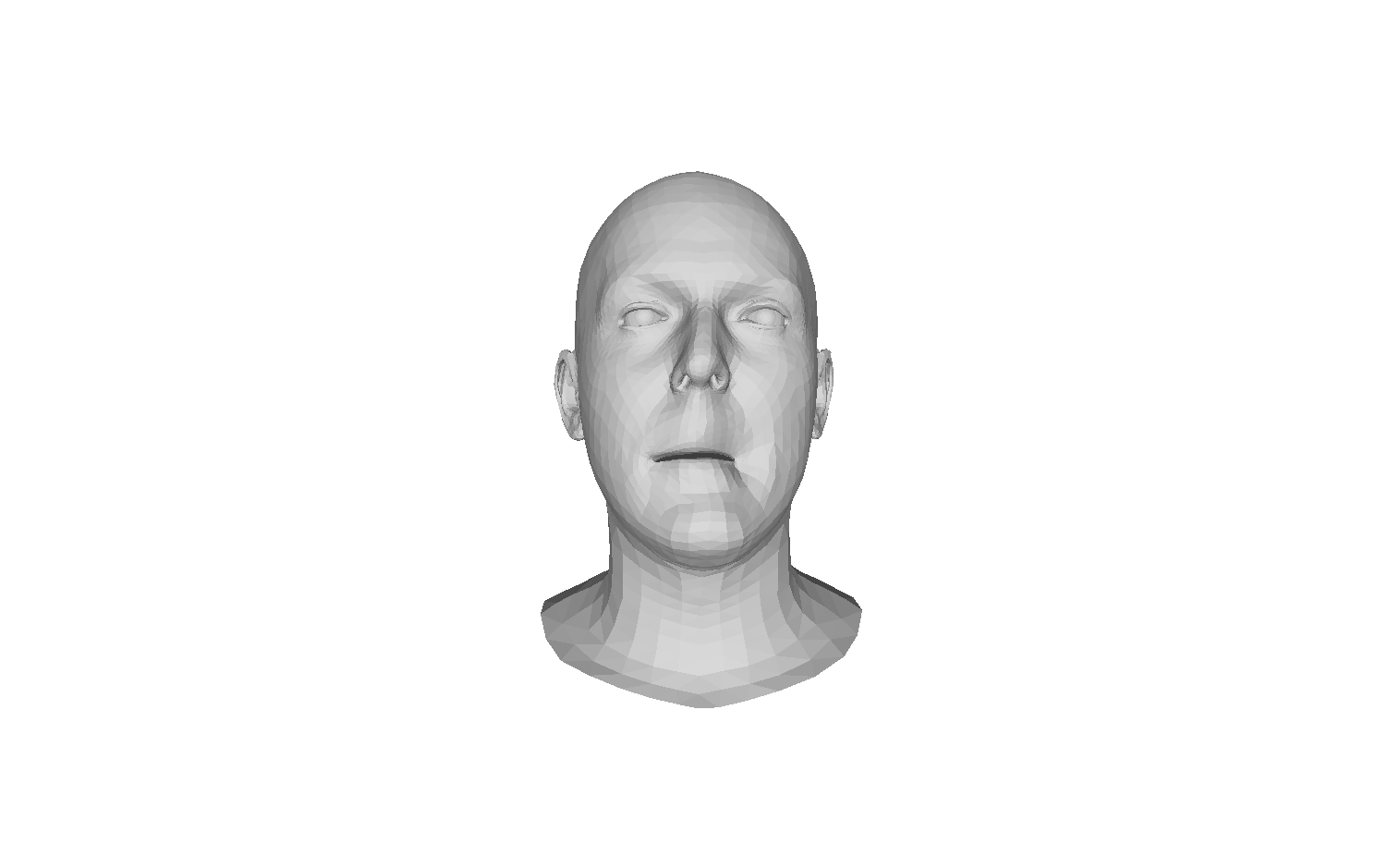}};
    \node[right of=d2, node distance=1.8cm] (d3) {\includegraphics[trim={400 80 400 100},clip,width=0.09\linewidth]{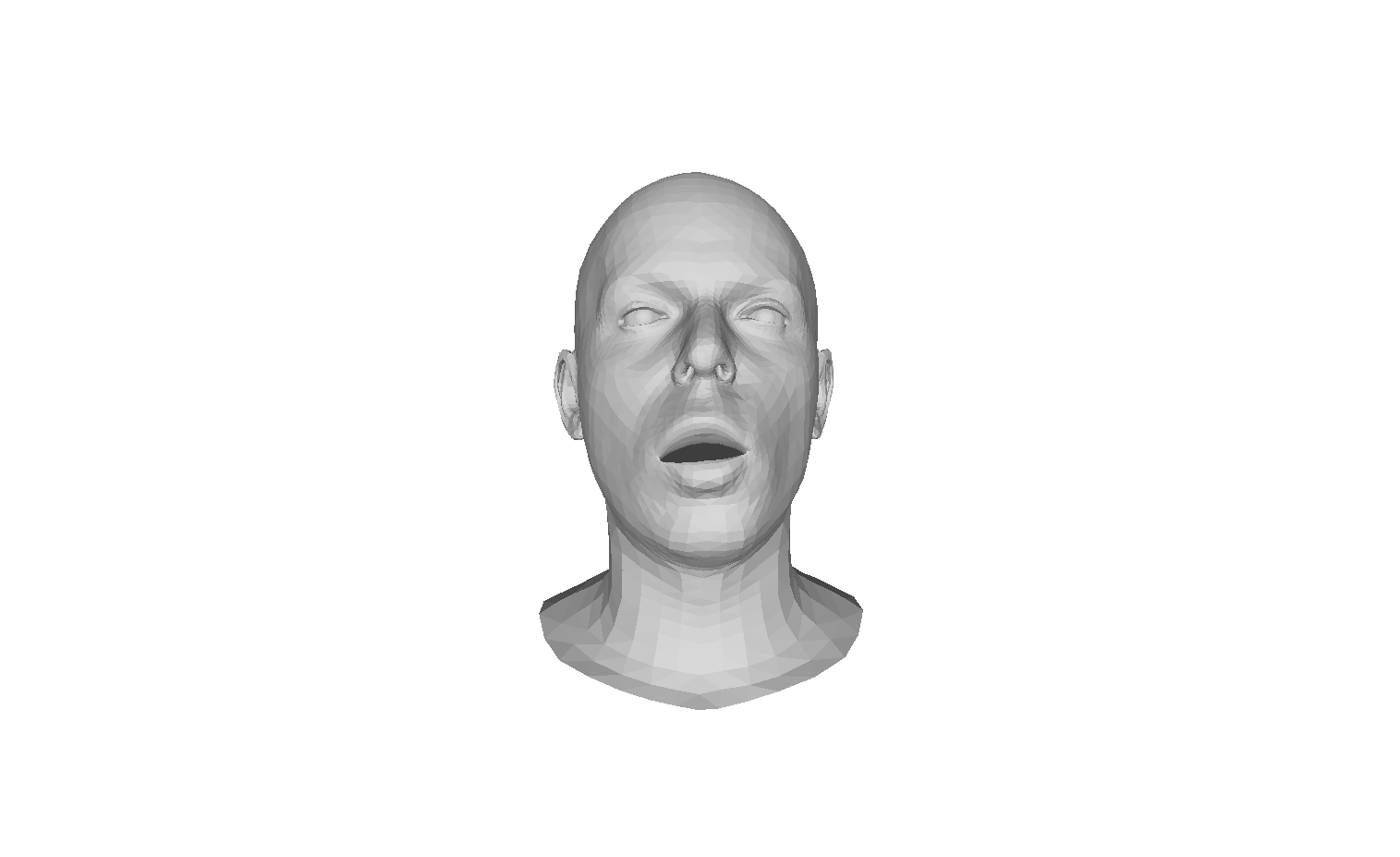}};
    \node[right of=d3, node distance=1.8cm] (d4) {\includegraphics[trim={400 80 400 100},clip,width=0.09\linewidth]{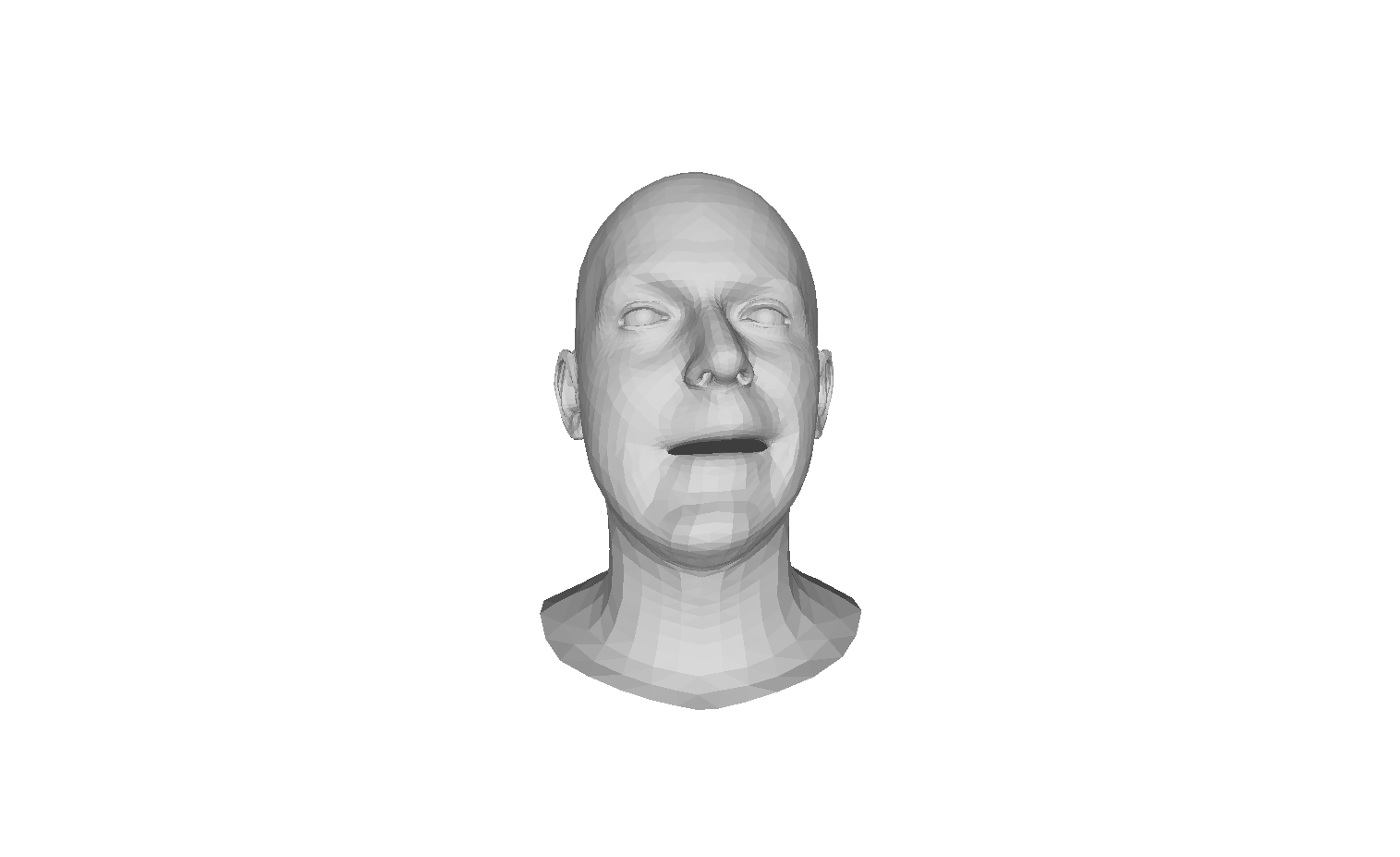}};
    \node[right of=d4, node distance=1.8cm] (d5) {\includegraphics[trim={400 80 400 100},clip,width=0.09\linewidth]{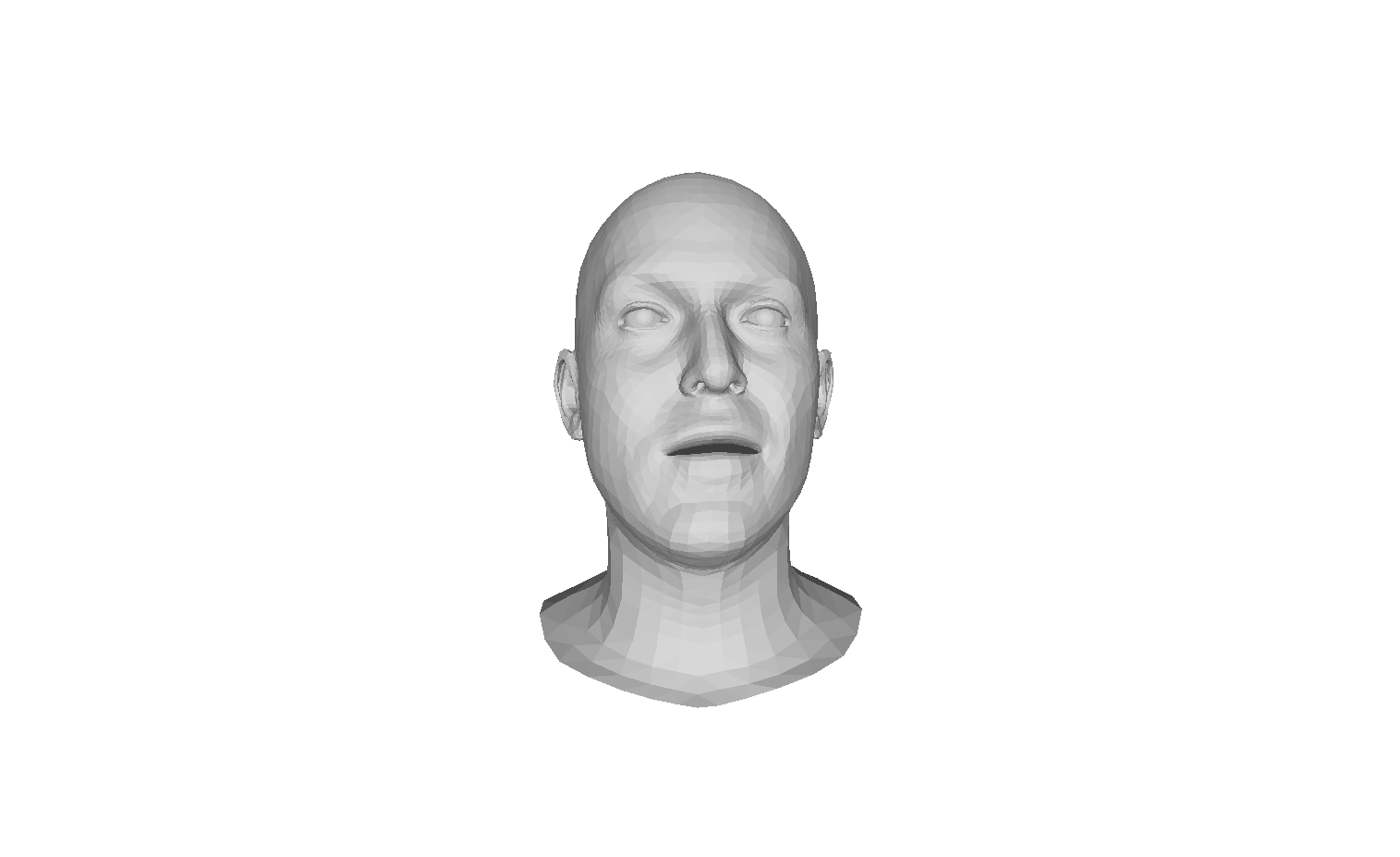}};
    \node[right of=d5, node distance=1.8cm] (d6) {\includegraphics[trim={400 80 400 100},clip,width=0.09\linewidth]{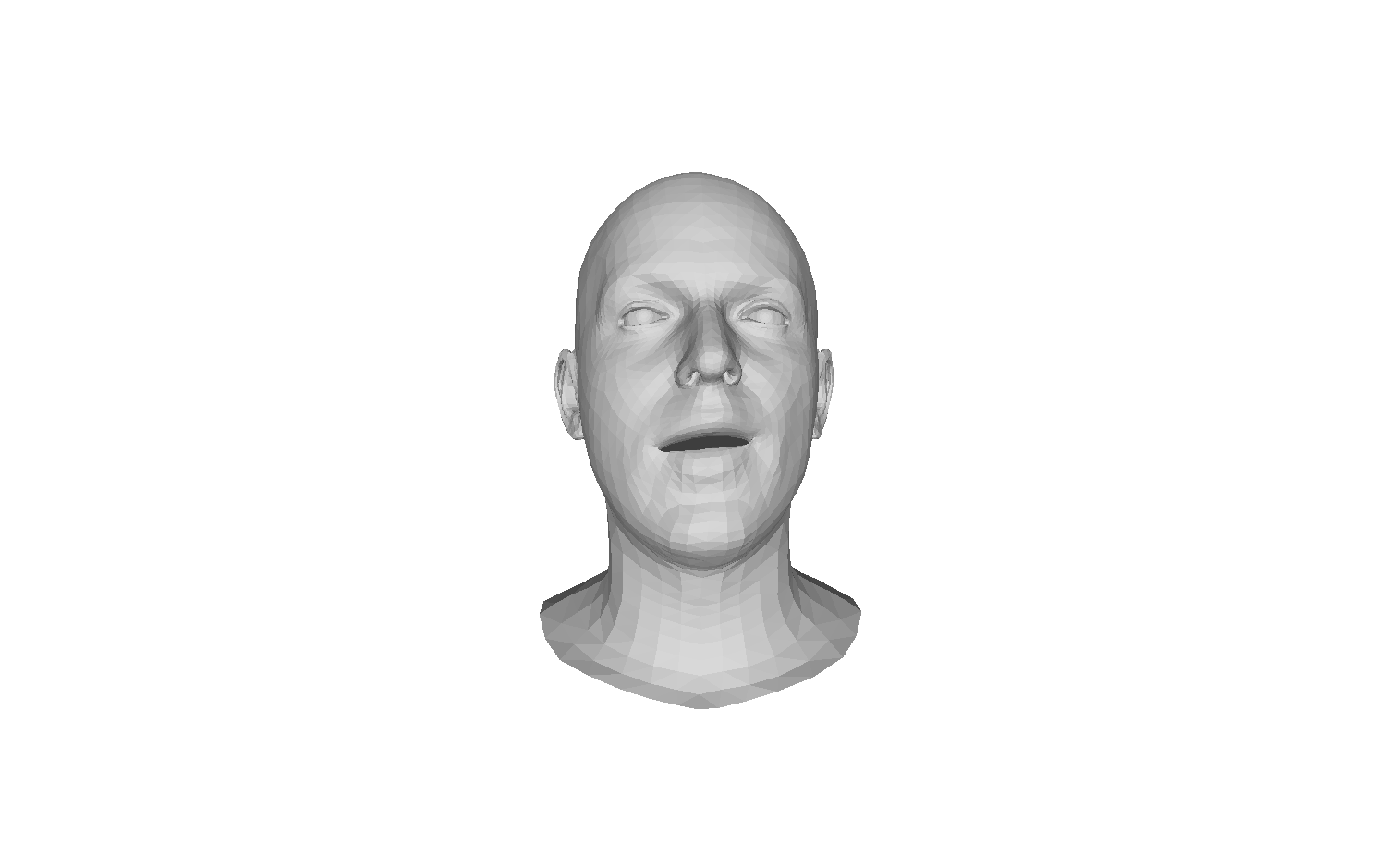}};
    \node[right of=d6, node distance=1.8cm] (d7) {\includegraphics[trim={400 80 400 100},clip,width=0.09\linewidth]{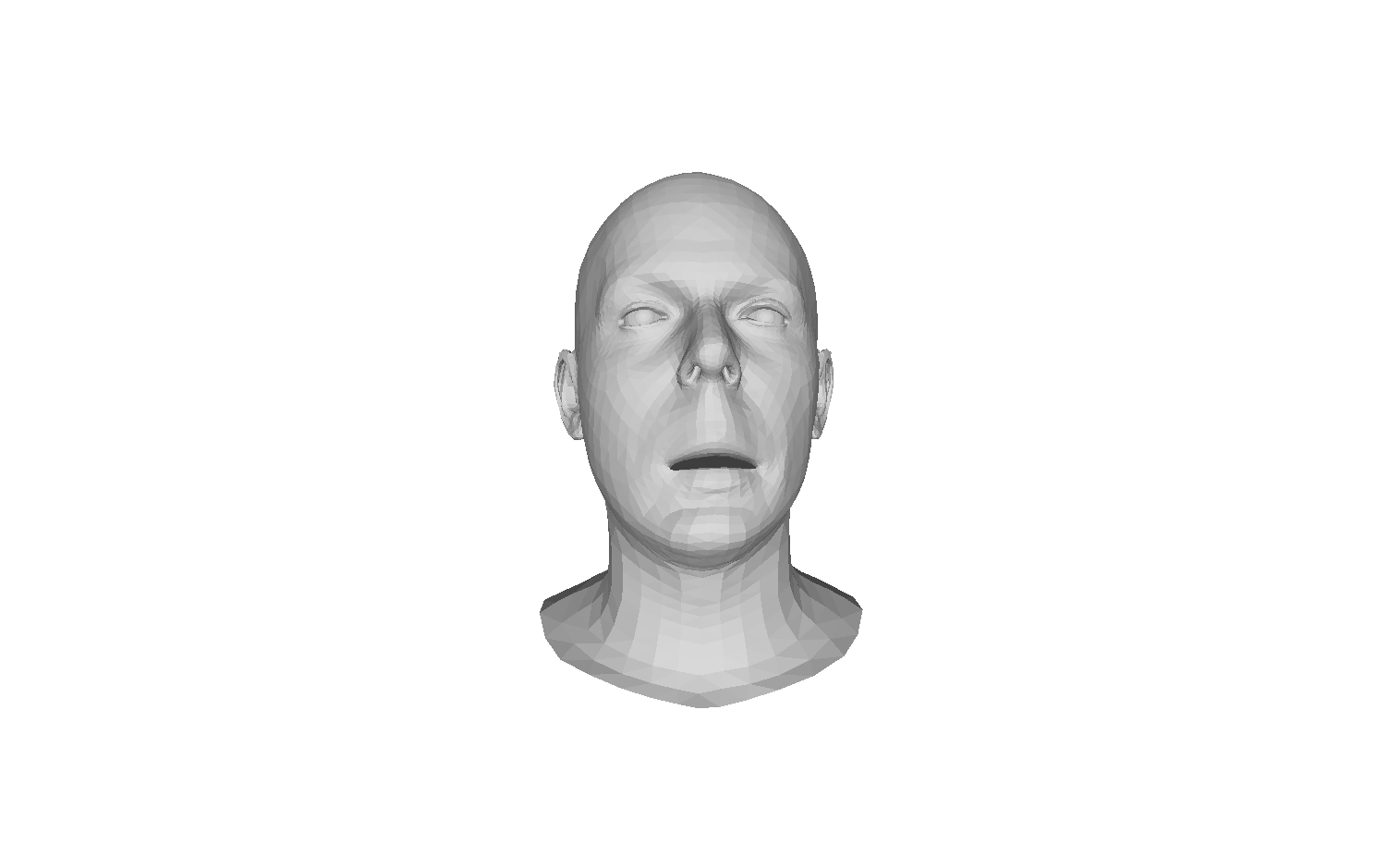}};
    \node[right of=d7, node distance=1.8cm] (d8) {\includegraphics[trim={400 80 400 100},clip,width=0.09\linewidth]{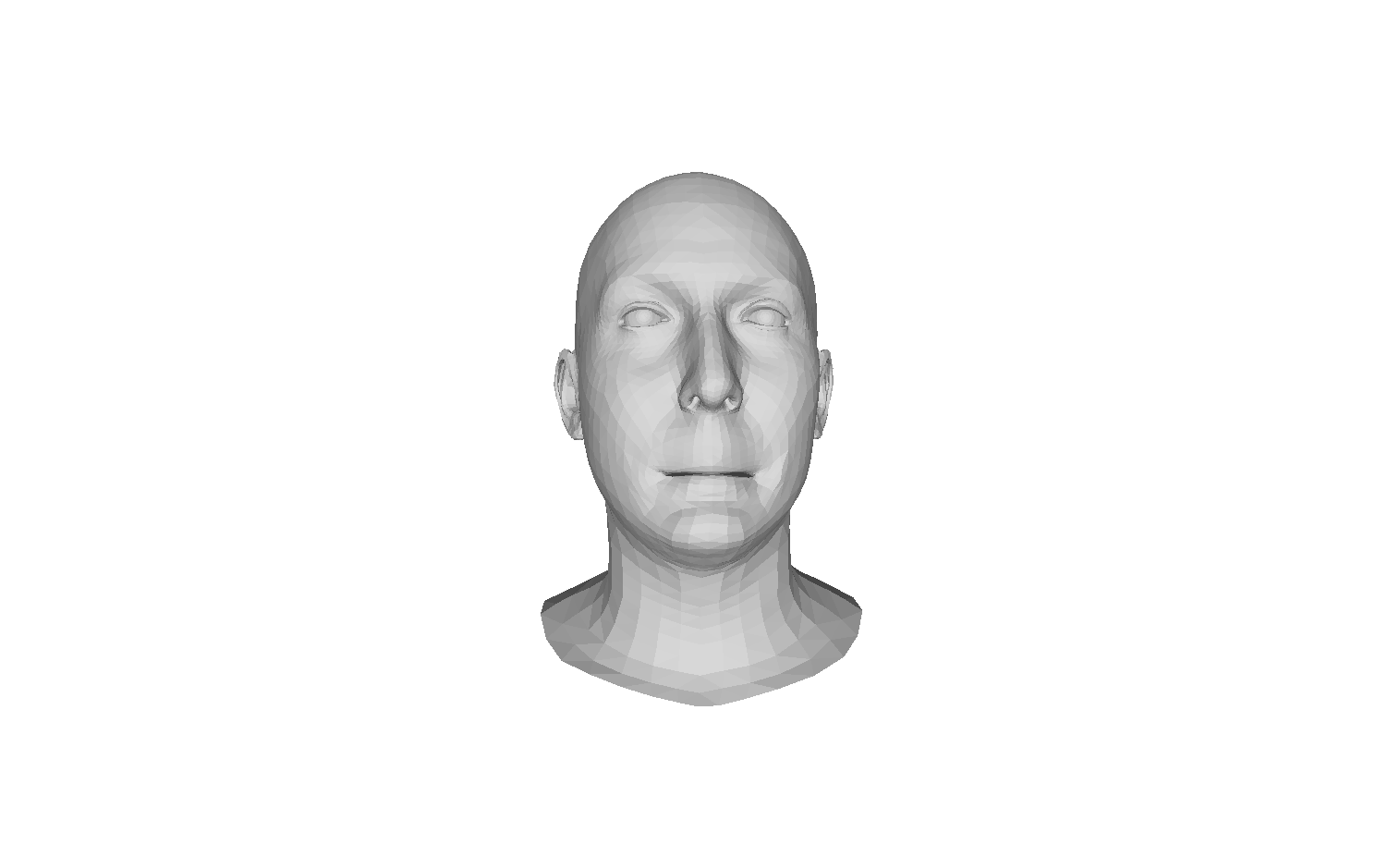}};
    \node[right of=d8, node distance=1.8cm] (d9) {\includegraphics[trim={400 80 400 100},clip,width=0.09\linewidth]{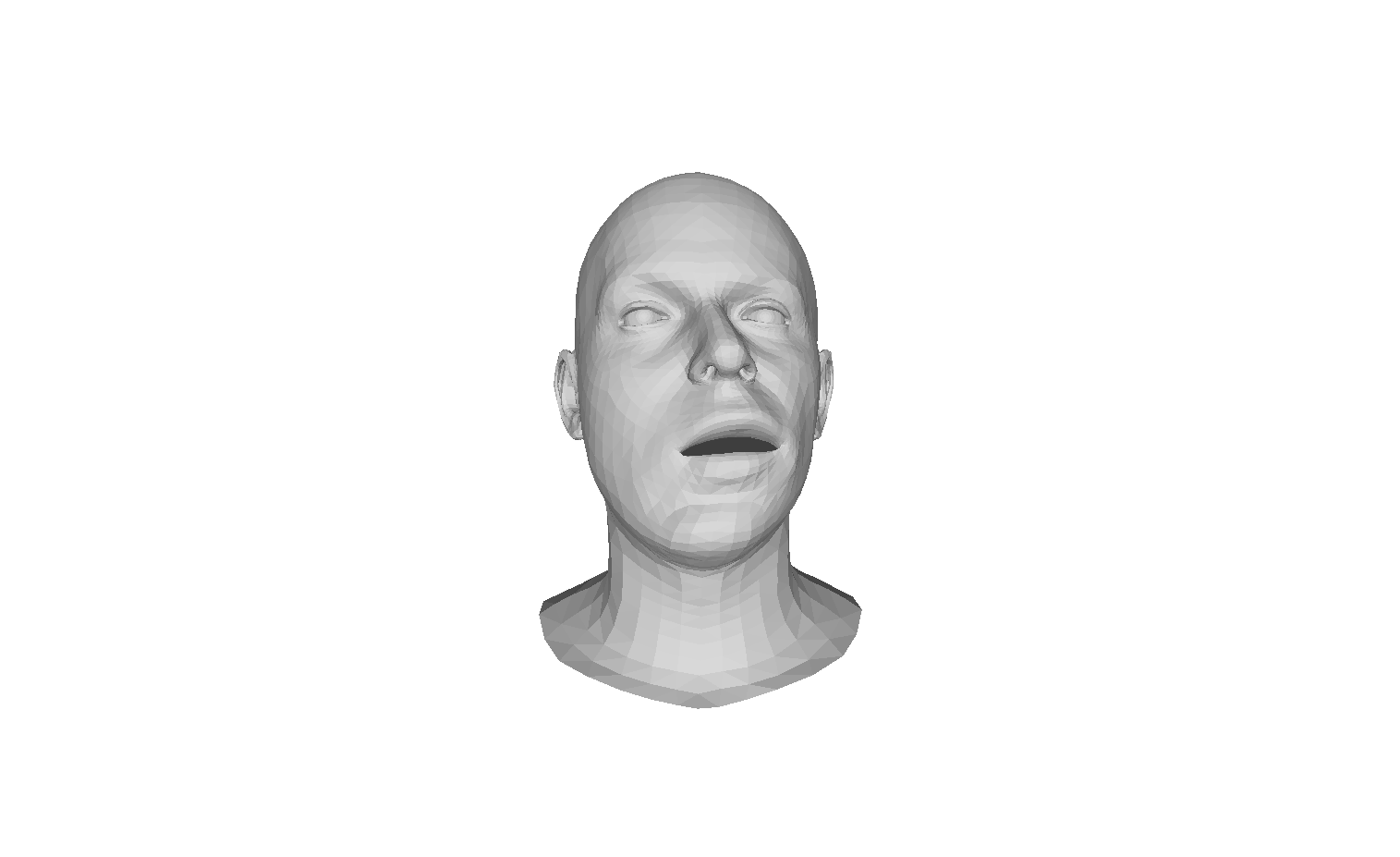}};
    
    \node[below of=d1, node distance=2.1cm] (e1) {\includegraphics[width=0.11\linewidth]{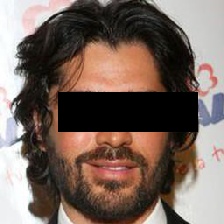}};
    \node[right of=e1, node distance=2.5cm] (e2) {\includegraphics[trim={400 80 400 100},clip,width=0.09\linewidth]{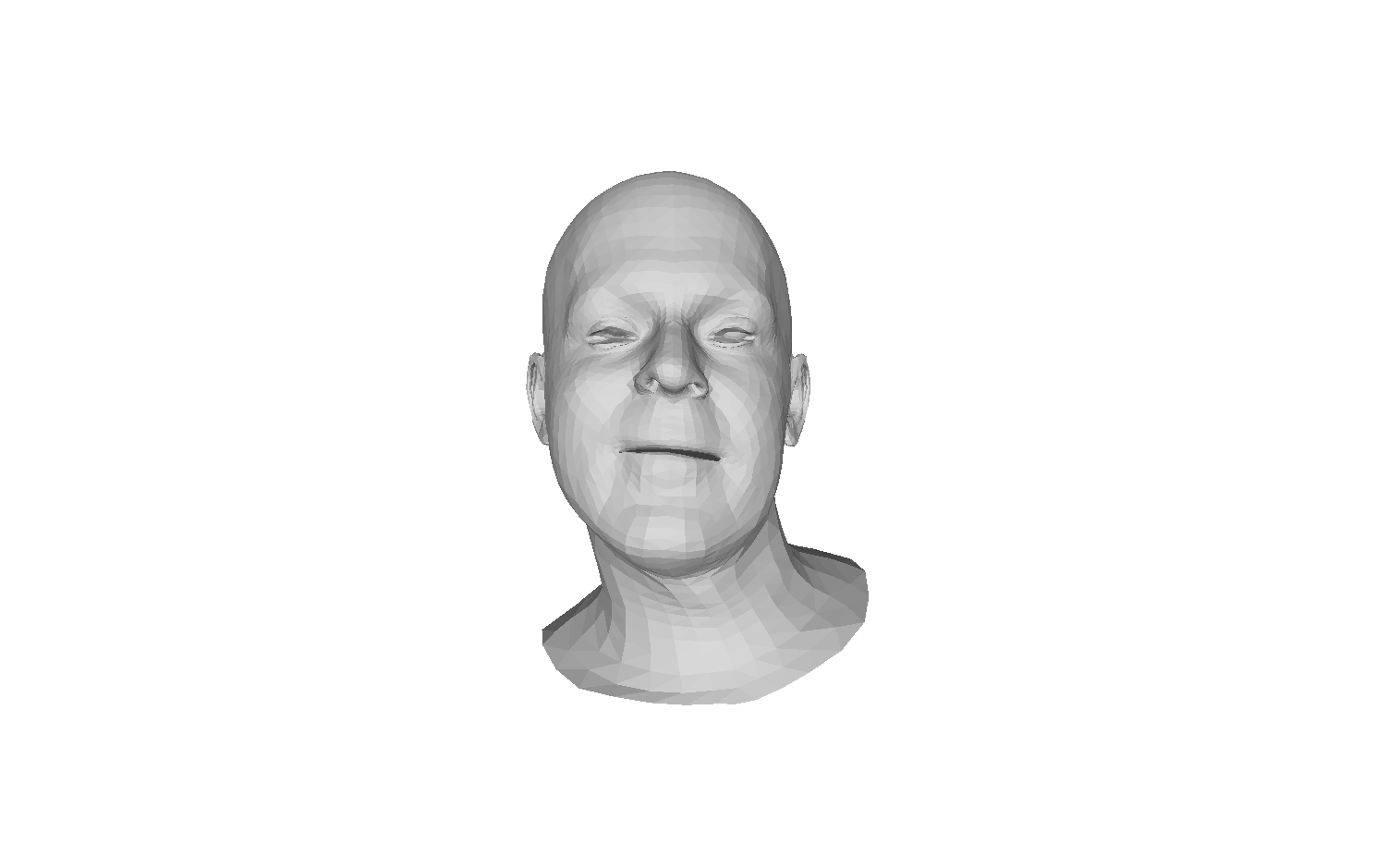}};
    \node[right of=e2, node distance=1.8cm] (e3) {\includegraphics[trim={400 80 400 100},clip,width=0.09\linewidth]{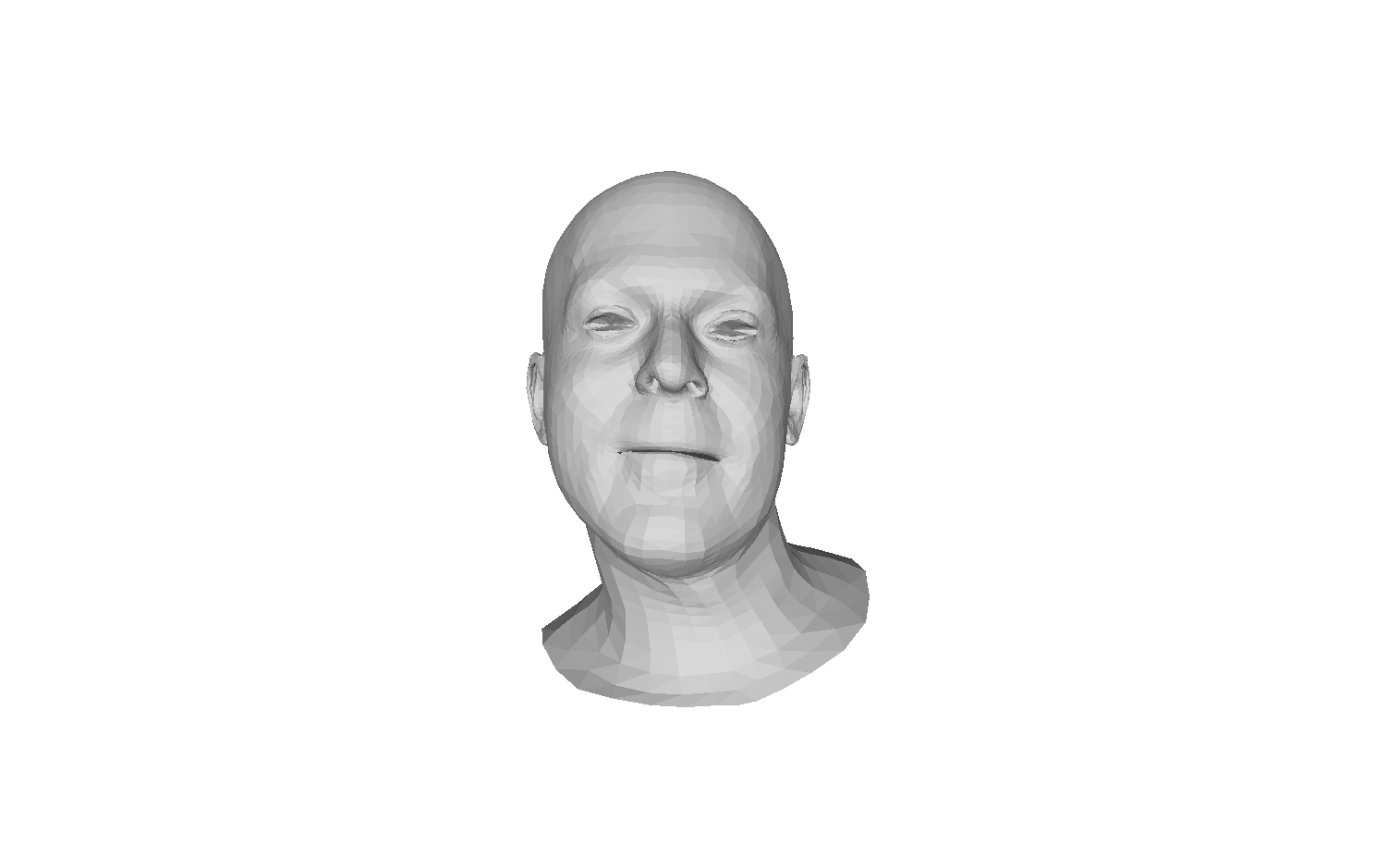}};
    \node[right of=e3, node distance=1.8cm] (e4) {\includegraphics[trim={400 80 400 100},clip,width=0.09\linewidth]{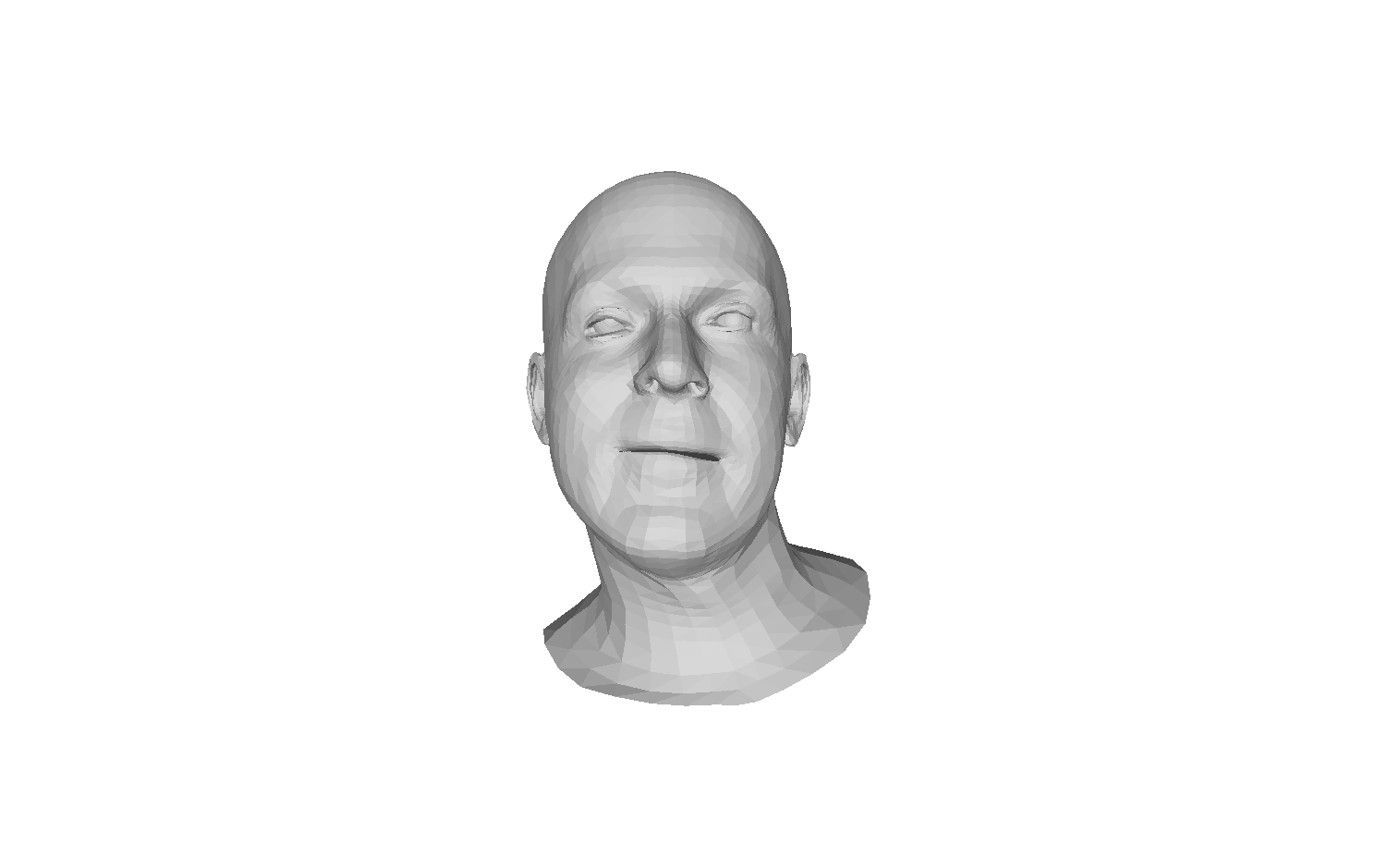}};
    \node[right of=e4, node distance=1.8cm] (e5) {\includegraphics[trim={400 80 400 100},clip,width=0.09\linewidth]{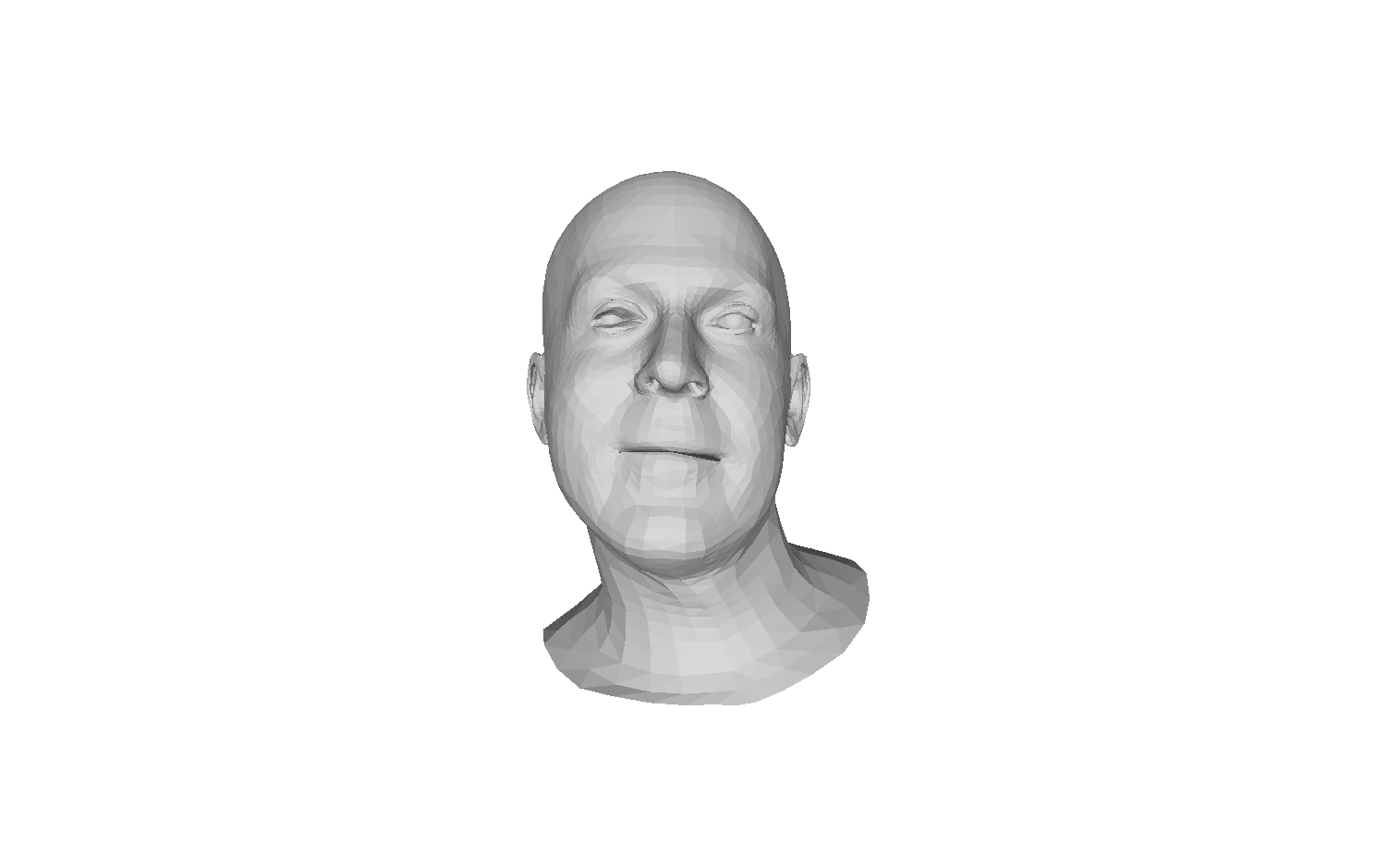}};
    \node[right of=e5, node distance=1.8cm] (e6) {\includegraphics[trim={400 80 400 100},clip,width=0.09\linewidth]{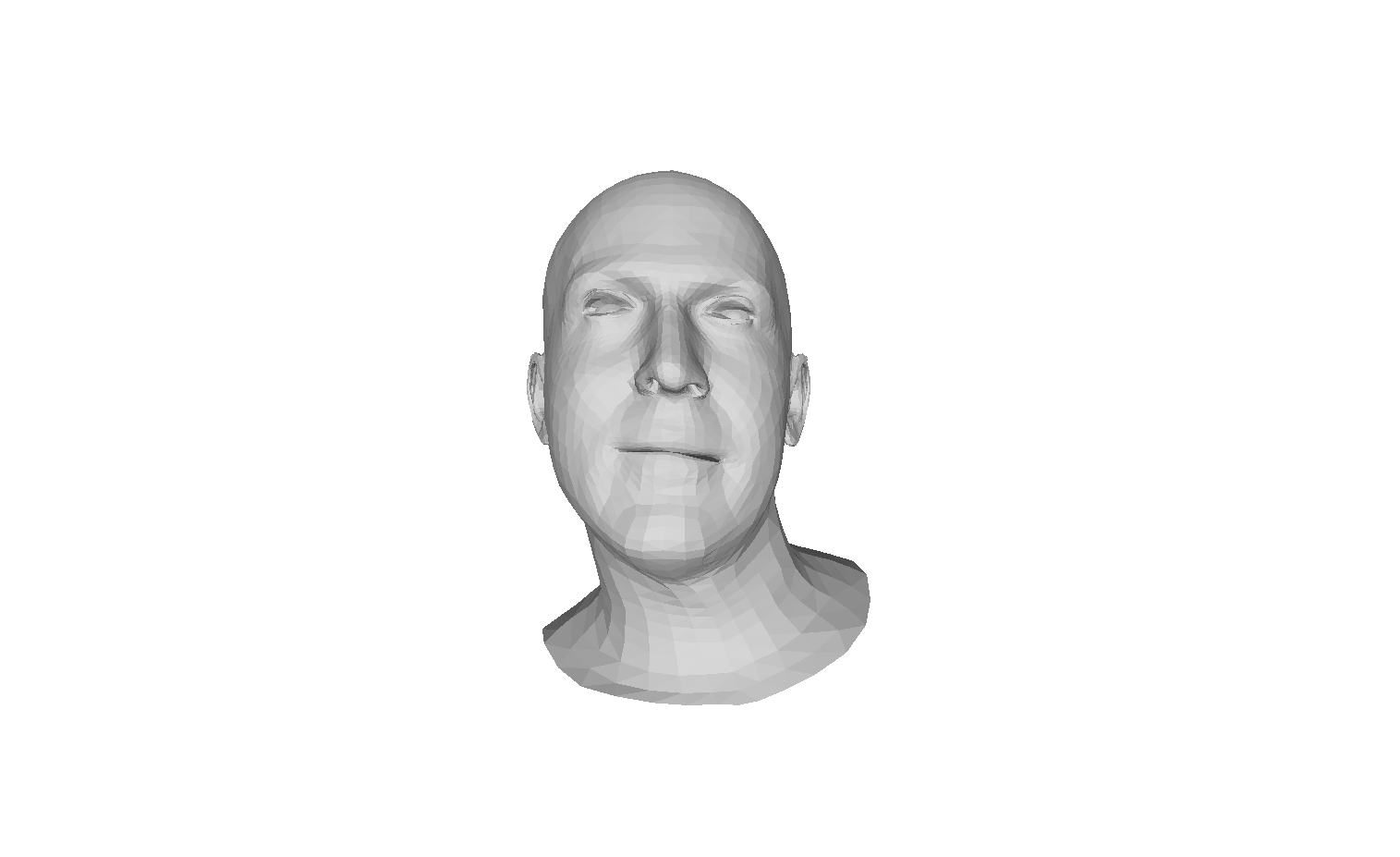}};
    \node[right of=e6, node distance=1.8cm] (e7) {\includegraphics[trim={400 80 400 100},clip,width=0.09\linewidth]{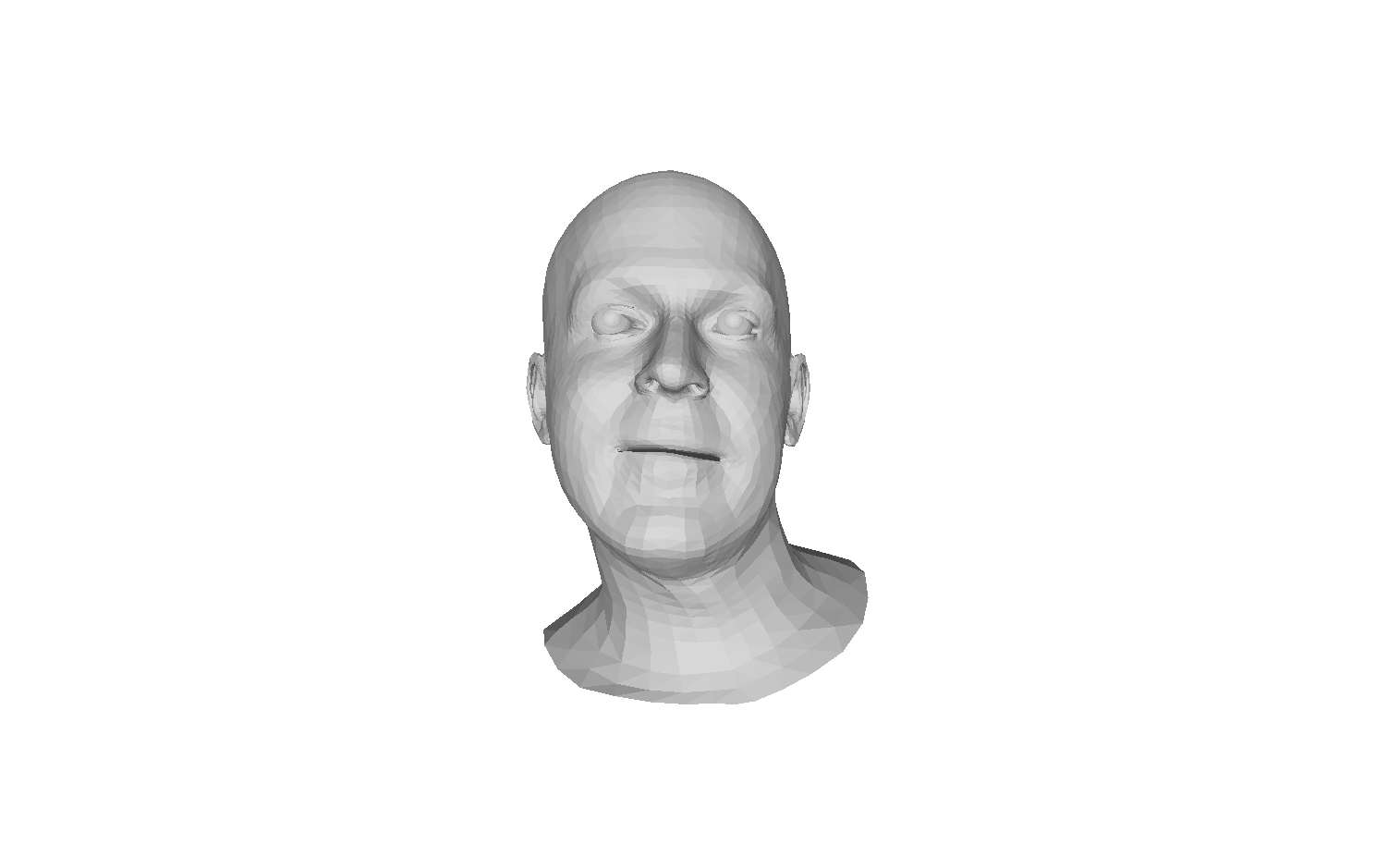}};
    \node[right of=e7, node distance=1.8cm] (e8) {\includegraphics[trim={400 80 400 100},clip,width=0.09\linewidth]{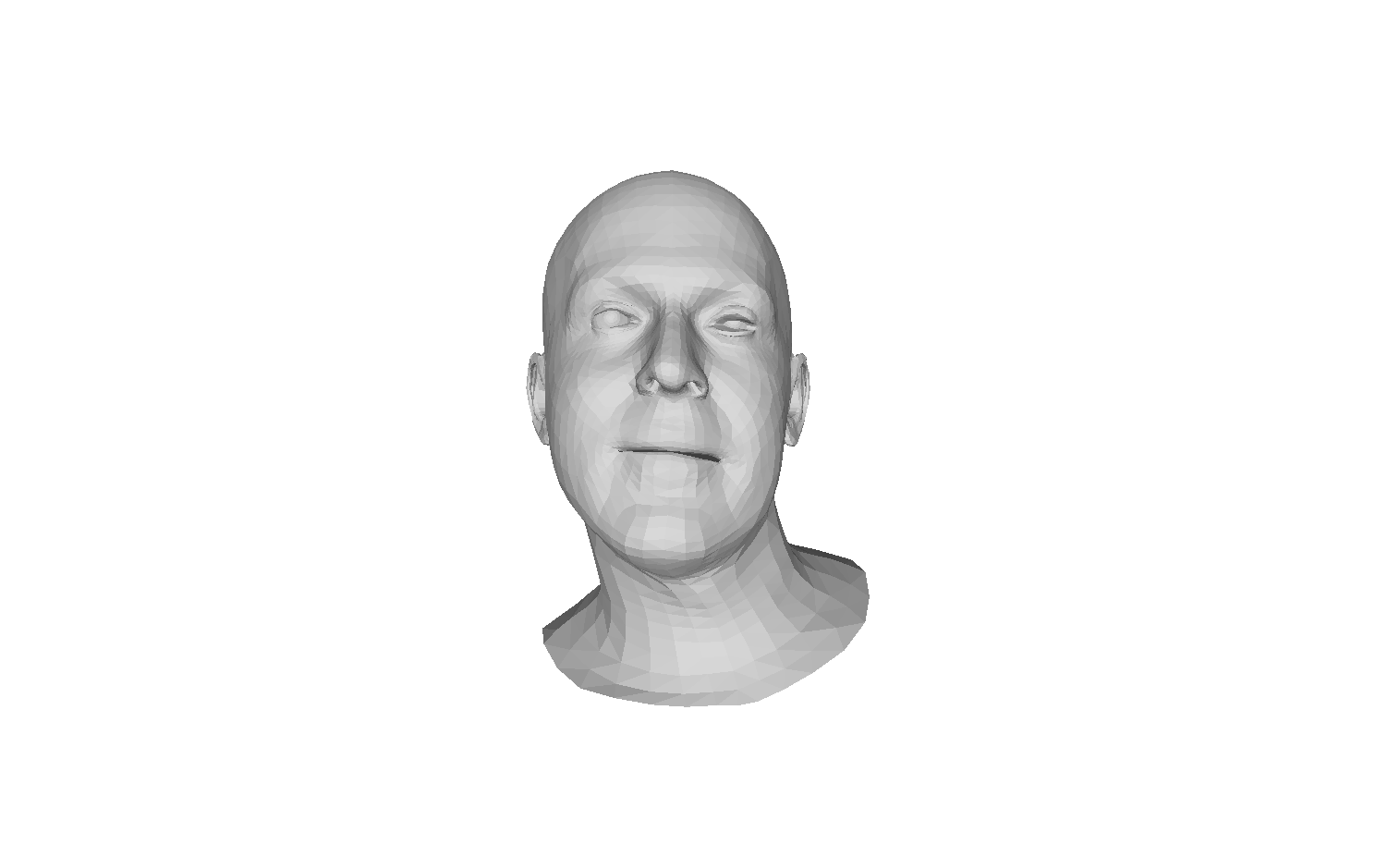}};
    \node[right of=e8, node distance=1.8cm] (e9) {\includegraphics[trim={400 80 400 100},clip,width=0.09\linewidth]{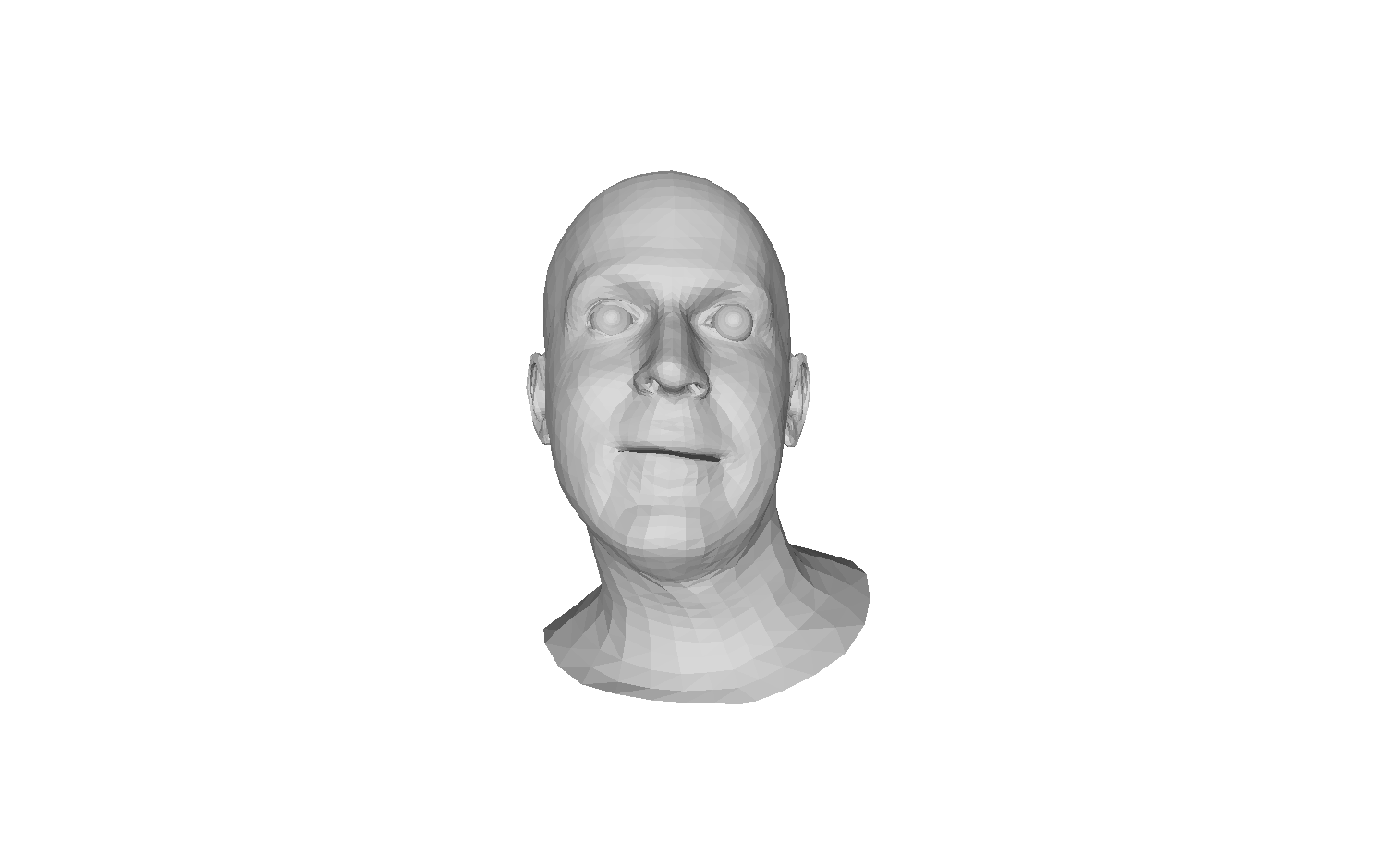}};

    \node[below of=e1, node distance=2.5cm] (f1) {\includegraphics[width=0.11\linewidth]{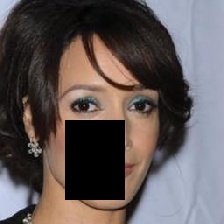}};
    \node[right of=f1, node distance=2.5cm] (f2) {\includegraphics[trim={400 80 400 100},clip,width=0.09\linewidth]{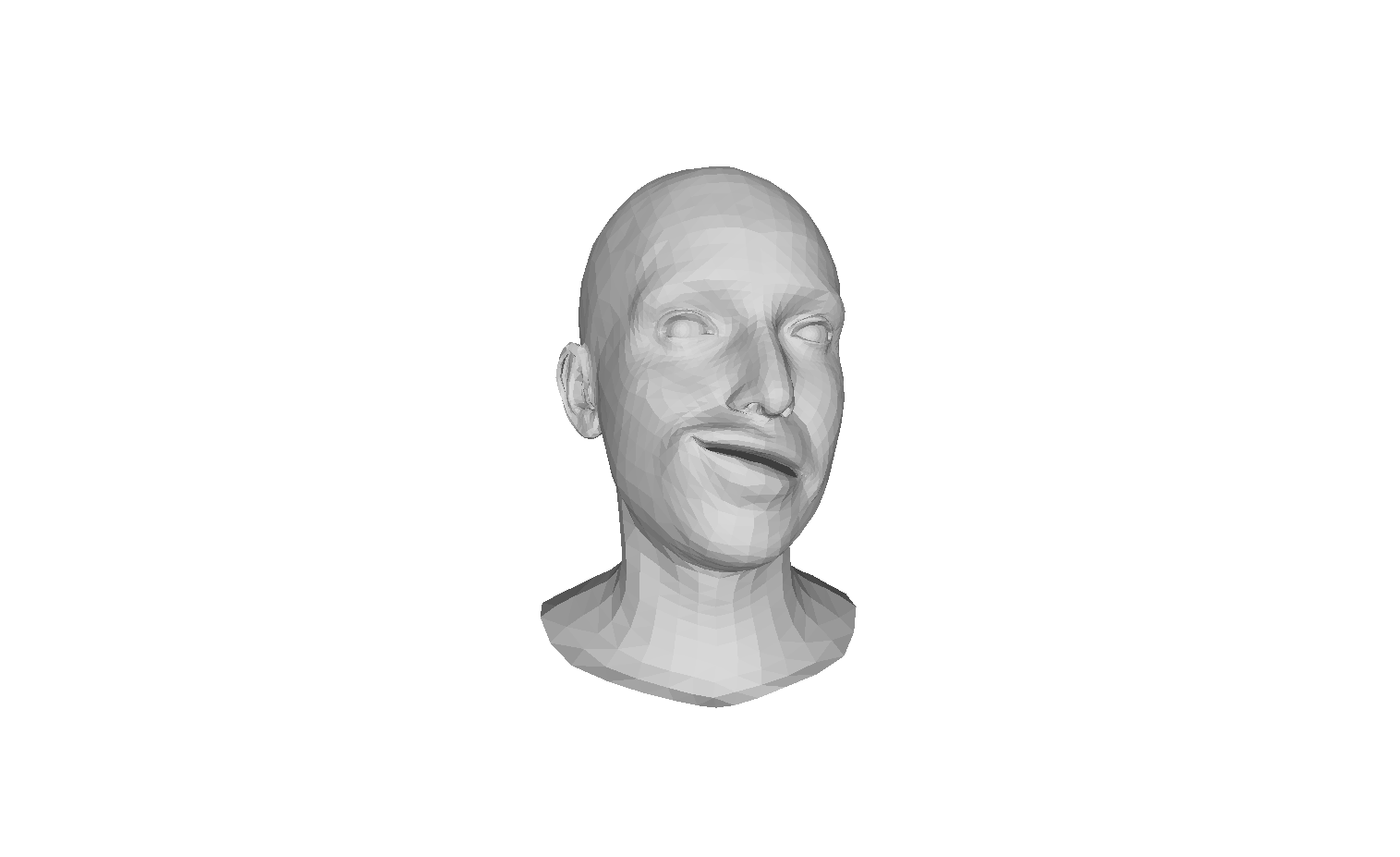}};
    \node[right of=f2, node distance=1.8cm] (f3) {\includegraphics[trim={400 80 400 100},clip,width=0.09\linewidth]{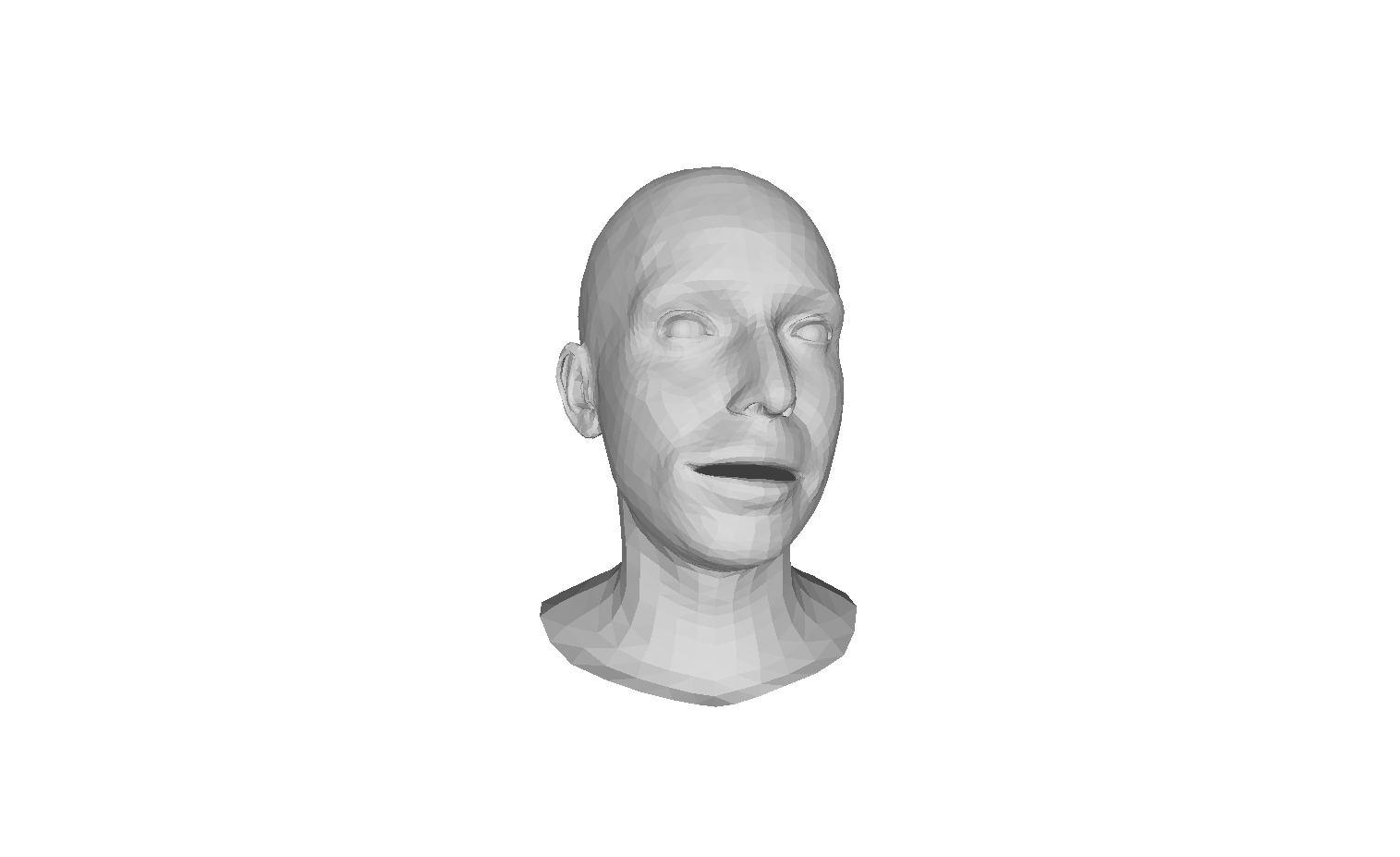}};
    \node[right of=f3, node distance=1.8cm] (f4) {\includegraphics[trim={400 80 400 100},clip,width=0.09\linewidth]{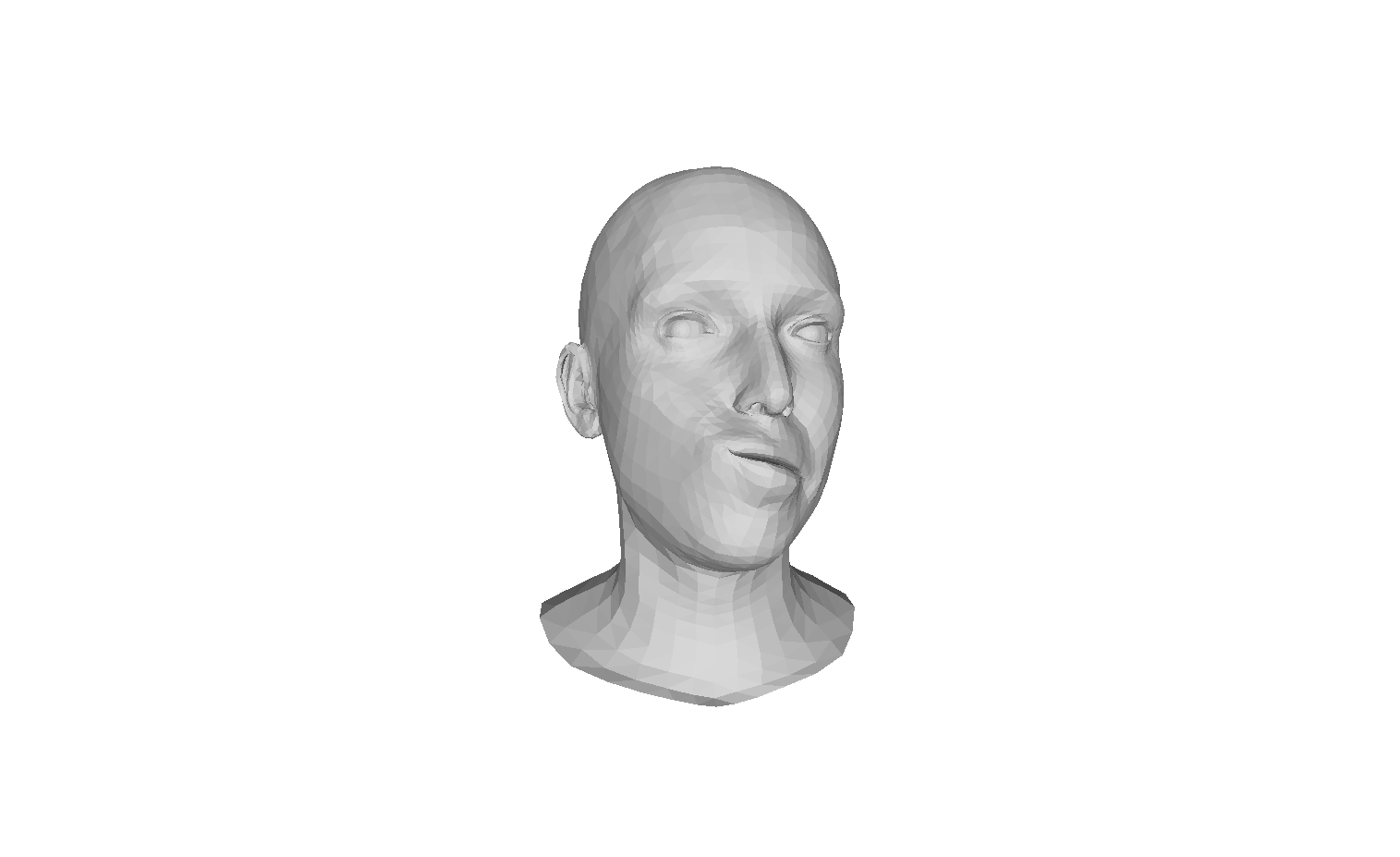}};
    \node[right of=f4, node distance=1.8cm] (f5) {\includegraphics[trim={400 80 400 100},clip,width=0.09\linewidth]{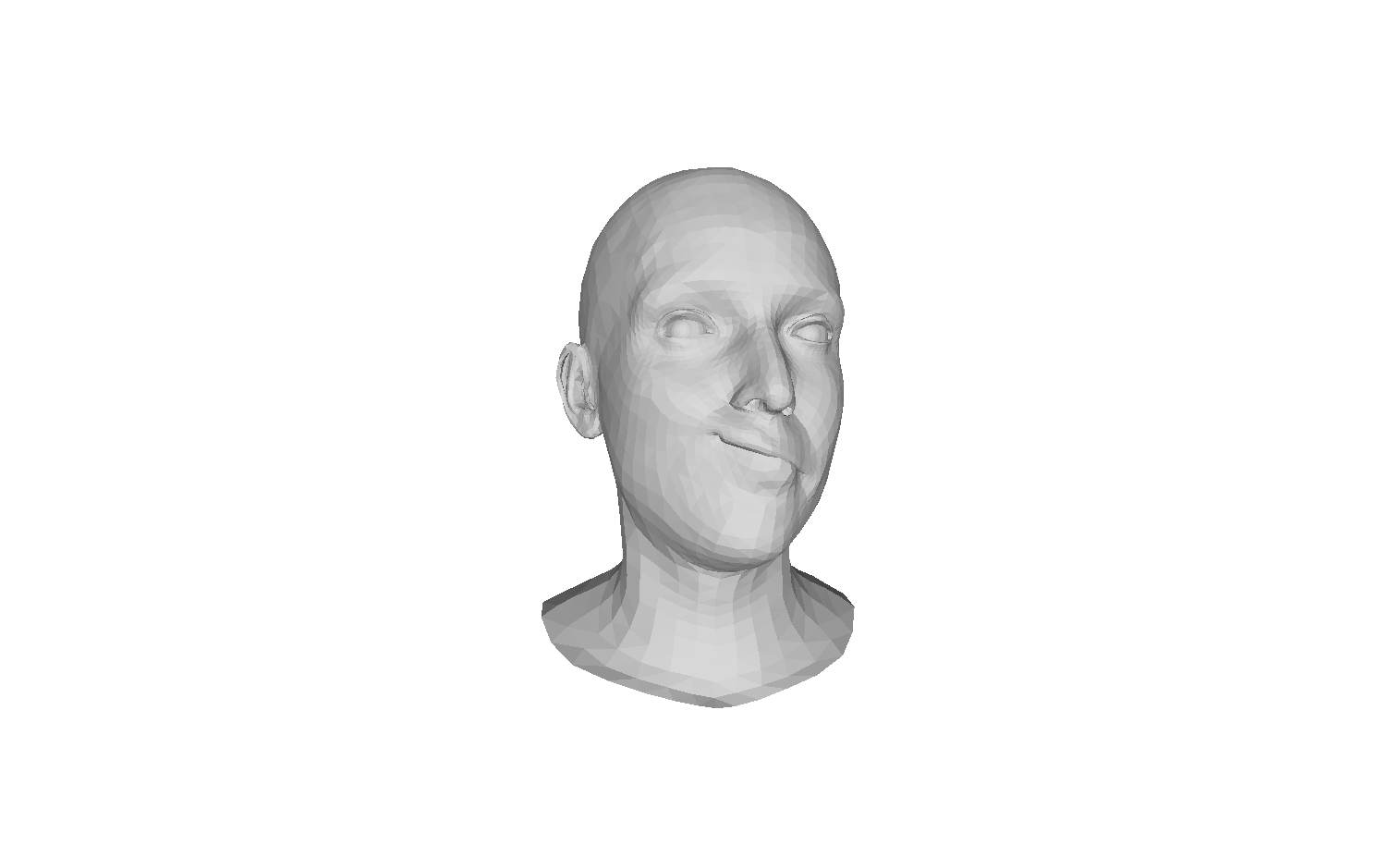}};
    \node[right of=f5, node distance=1.8cm] (f6) {\includegraphics[trim={400 80 400 100},clip,width=0.09\linewidth]{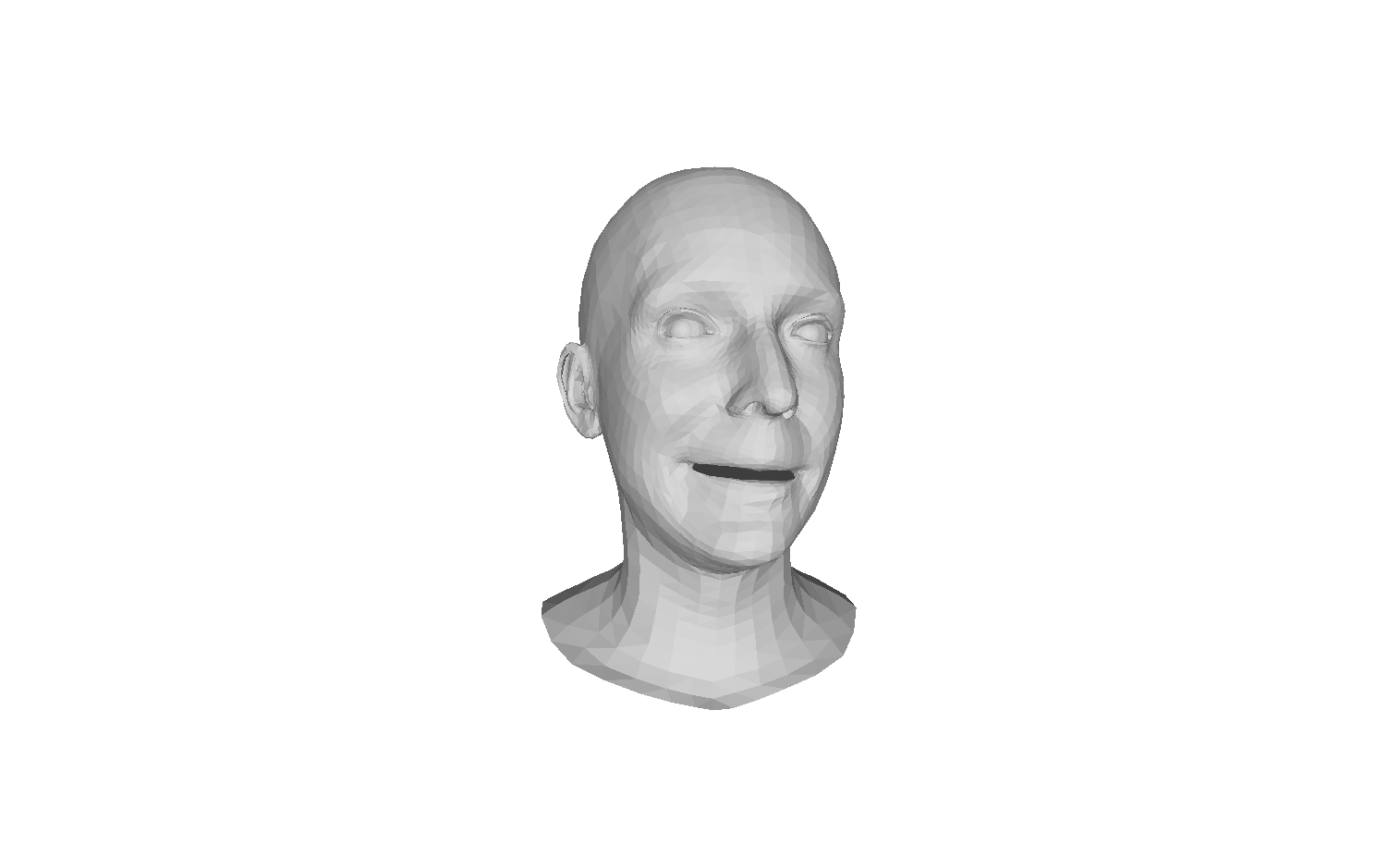}};
    \node[right of=f6, node distance=1.8cm] (f7) {\includegraphics[trim={400 80 400 100},clip,width=0.09\linewidth]{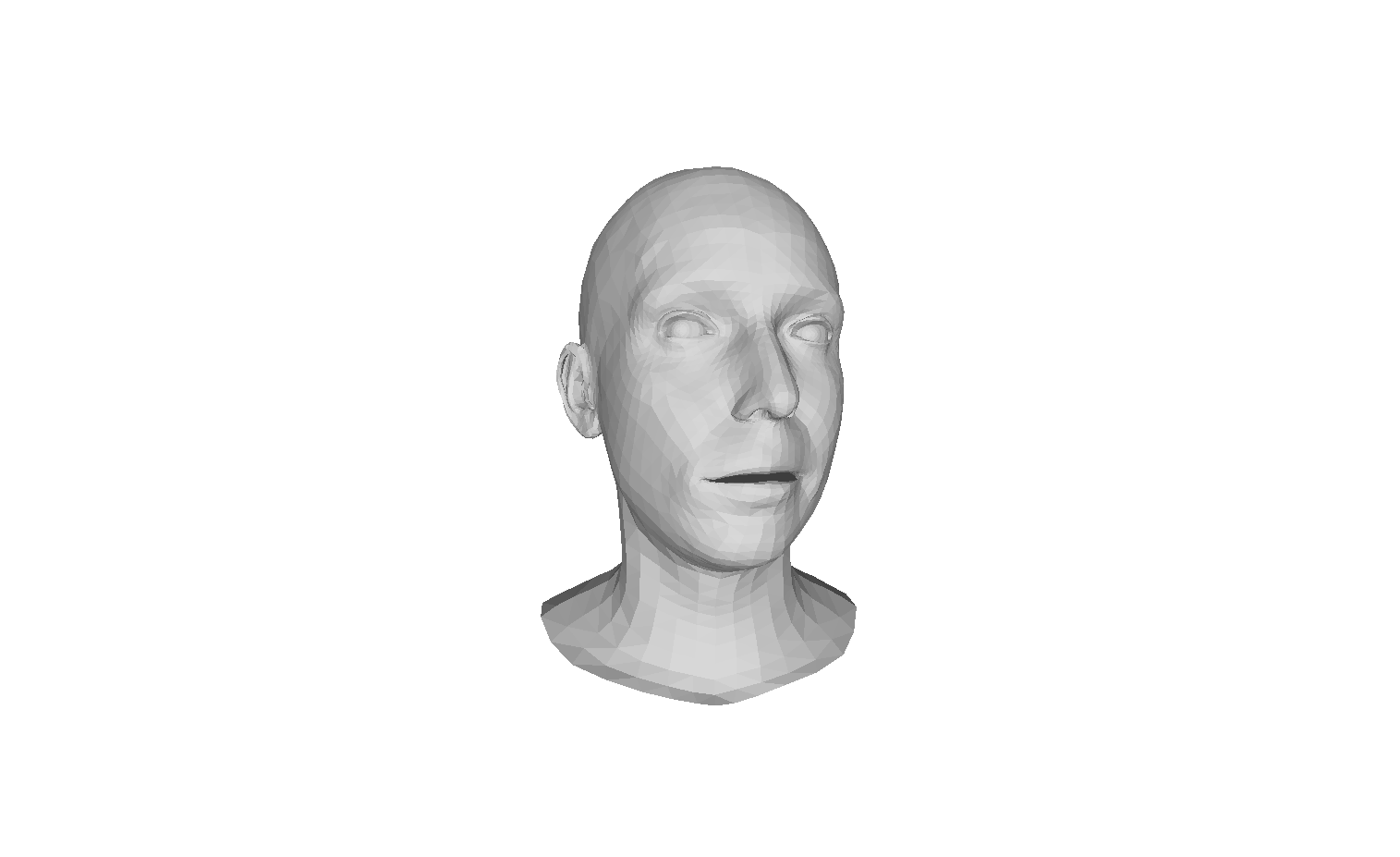}};
    \node[right of=f7, node distance=1.8cm] (f8) {\includegraphics[trim={400 80 400 100},clip,width=0.09\linewidth]{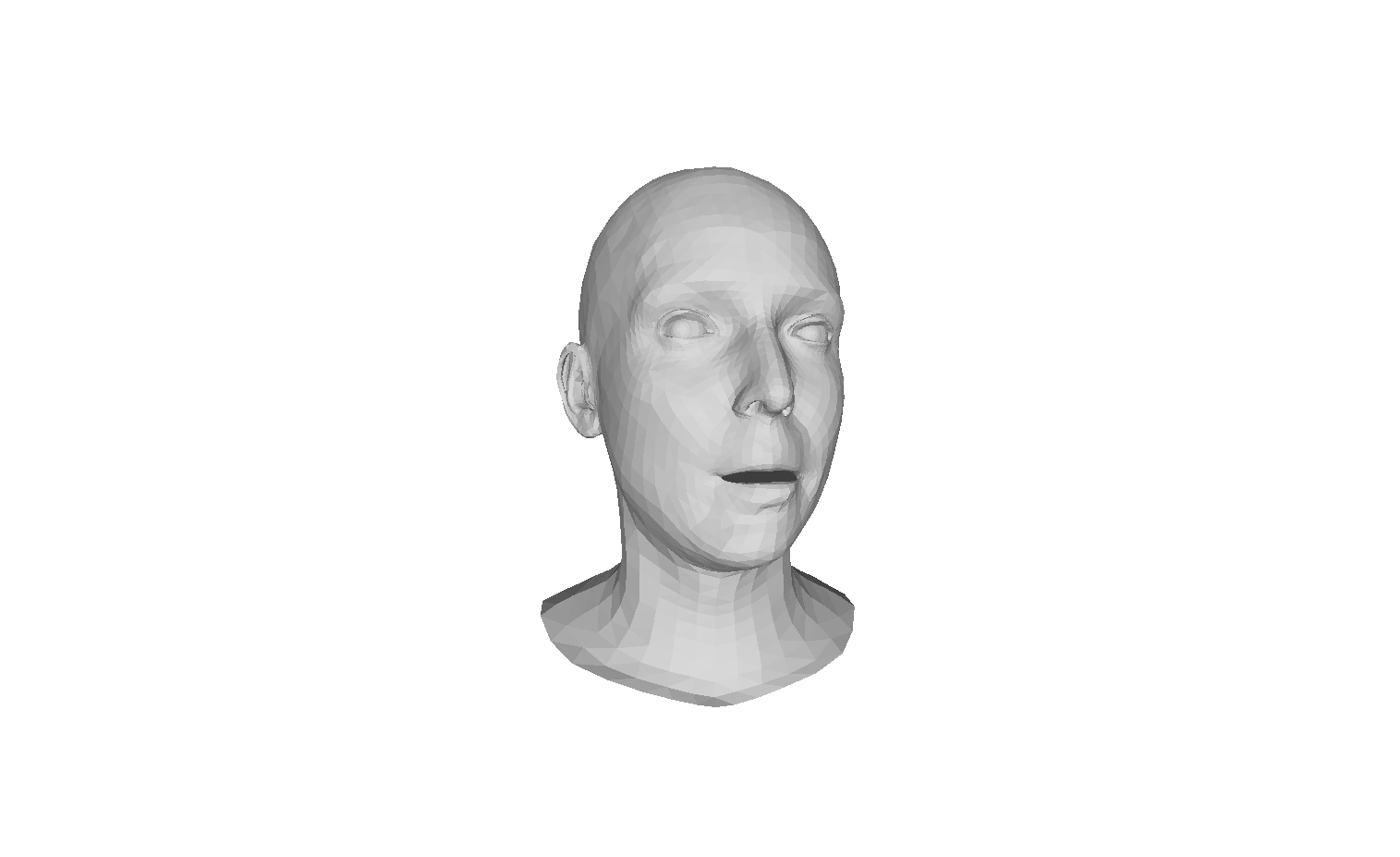}};
    \node[right of=f8, node distance=1.8cm] (f9) {\includegraphics[trim={400 80 400 100},clip,width=0.09\linewidth]{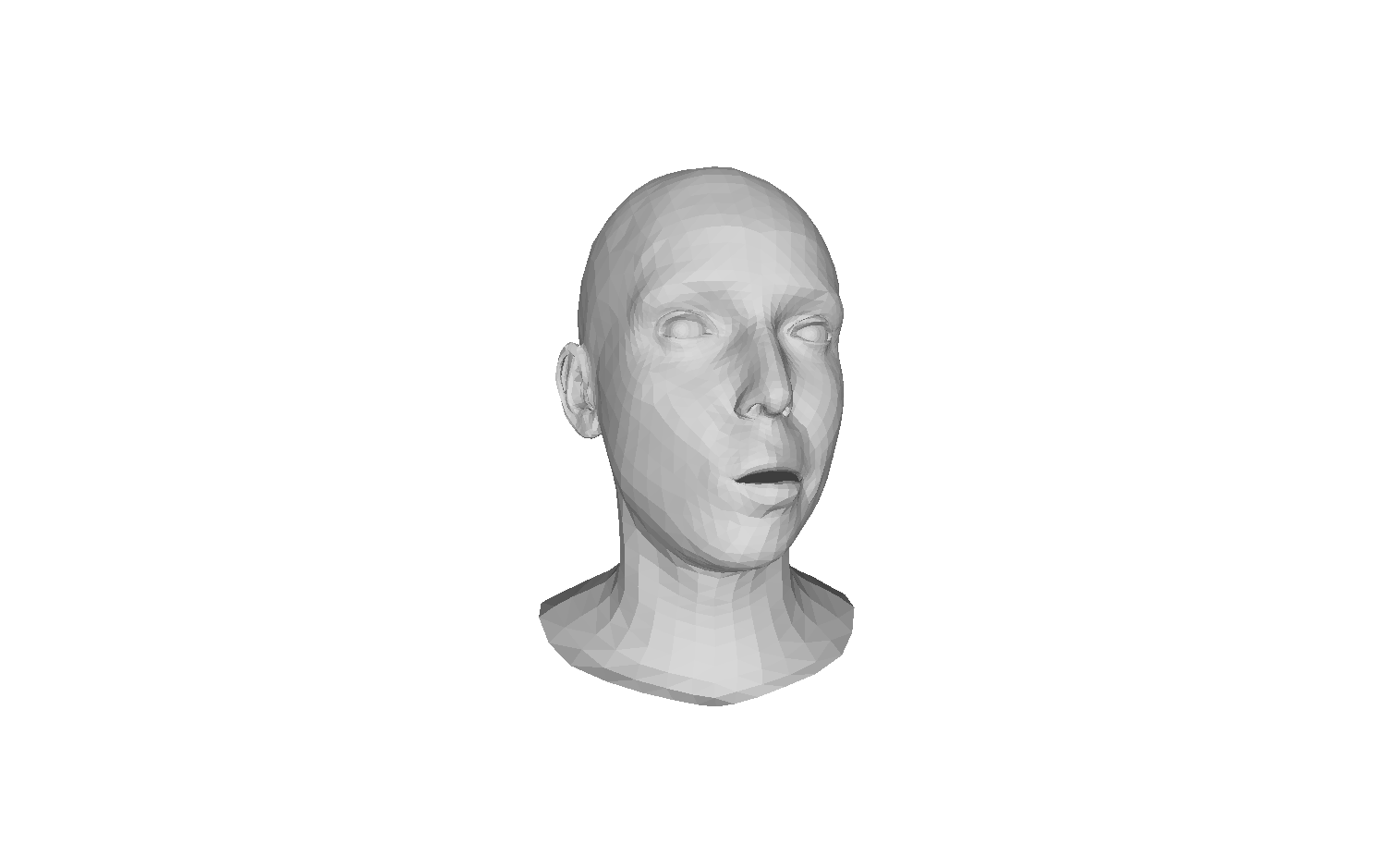}};
    
    \node[below of=f1, node distance=2.1cm] (g1) {\includegraphics[width=0.11\linewidth]{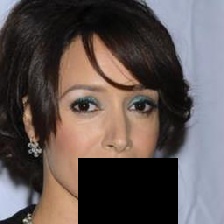}};
    \node[right of=g1, node distance=2.5cm] (g2) {\includegraphics[trim={400 80 400 100},clip,width=0.09\linewidth]{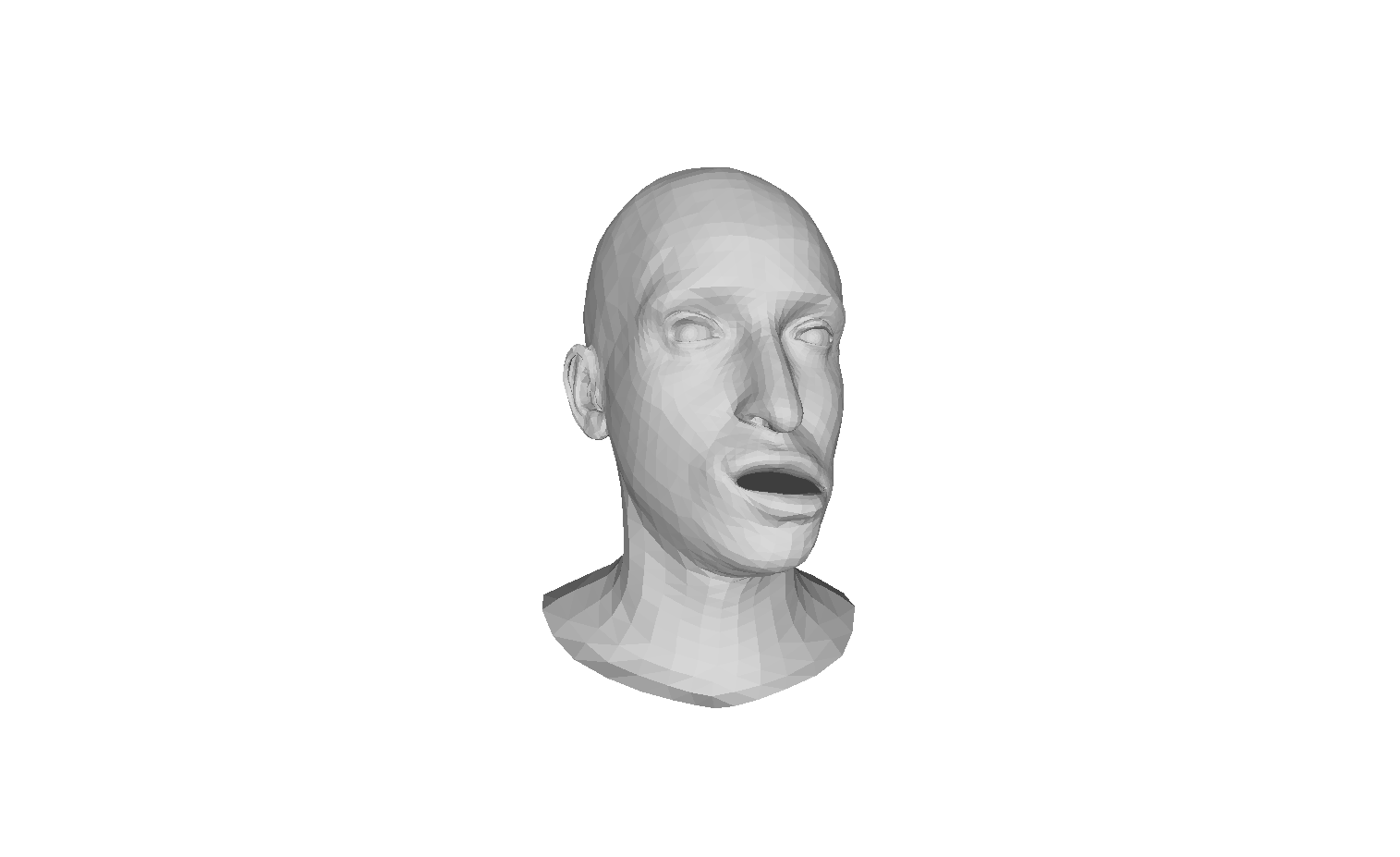}};
    \node[right of=g2, node distance=1.8cm] (g3) {\includegraphics[trim={400 80 400 100},clip,width=0.09\linewidth]{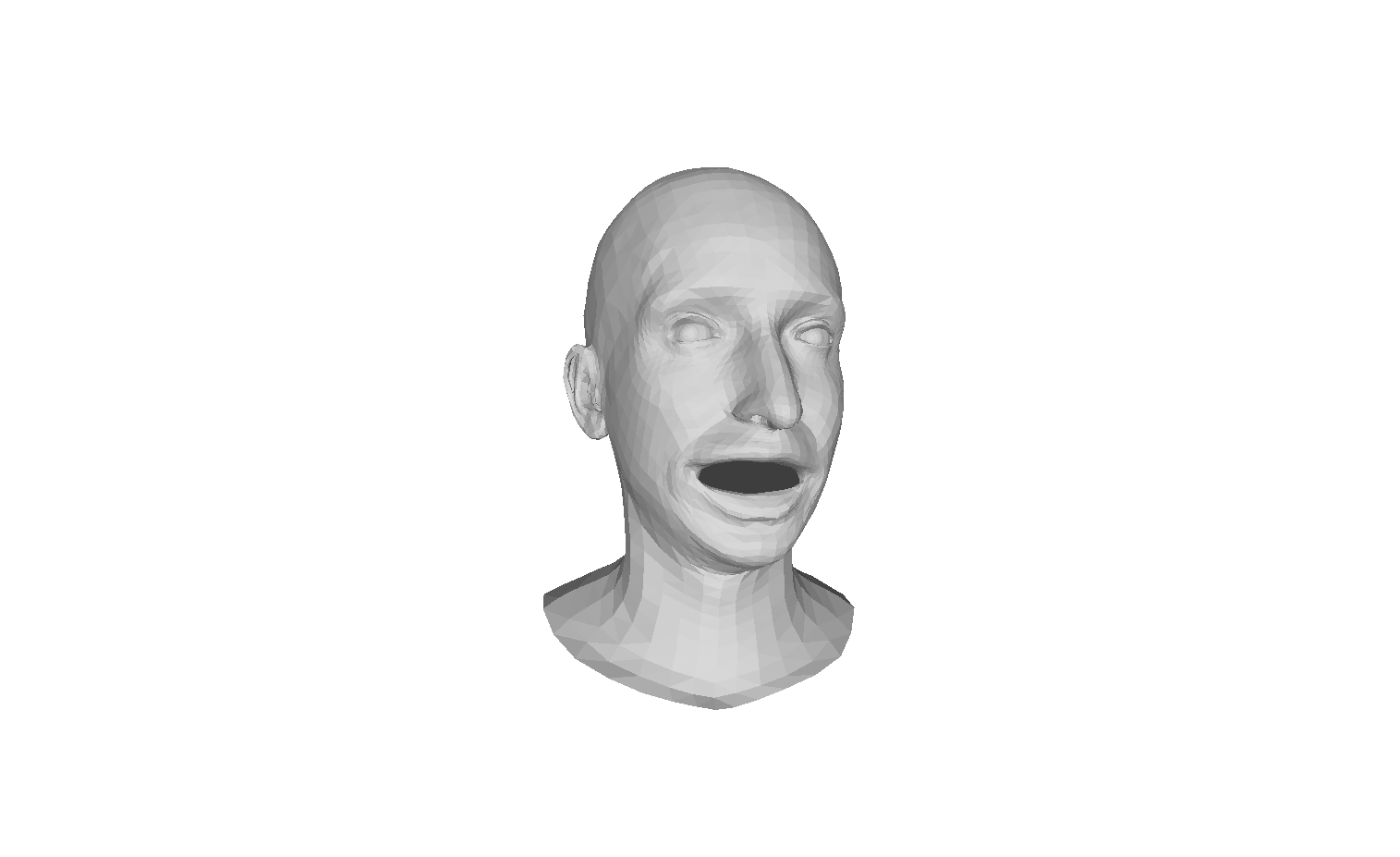}};
    \node[right of=g3, node distance=1.8cm] (g4) {\includegraphics[trim={400 80 400 100},clip,width=0.09\linewidth]{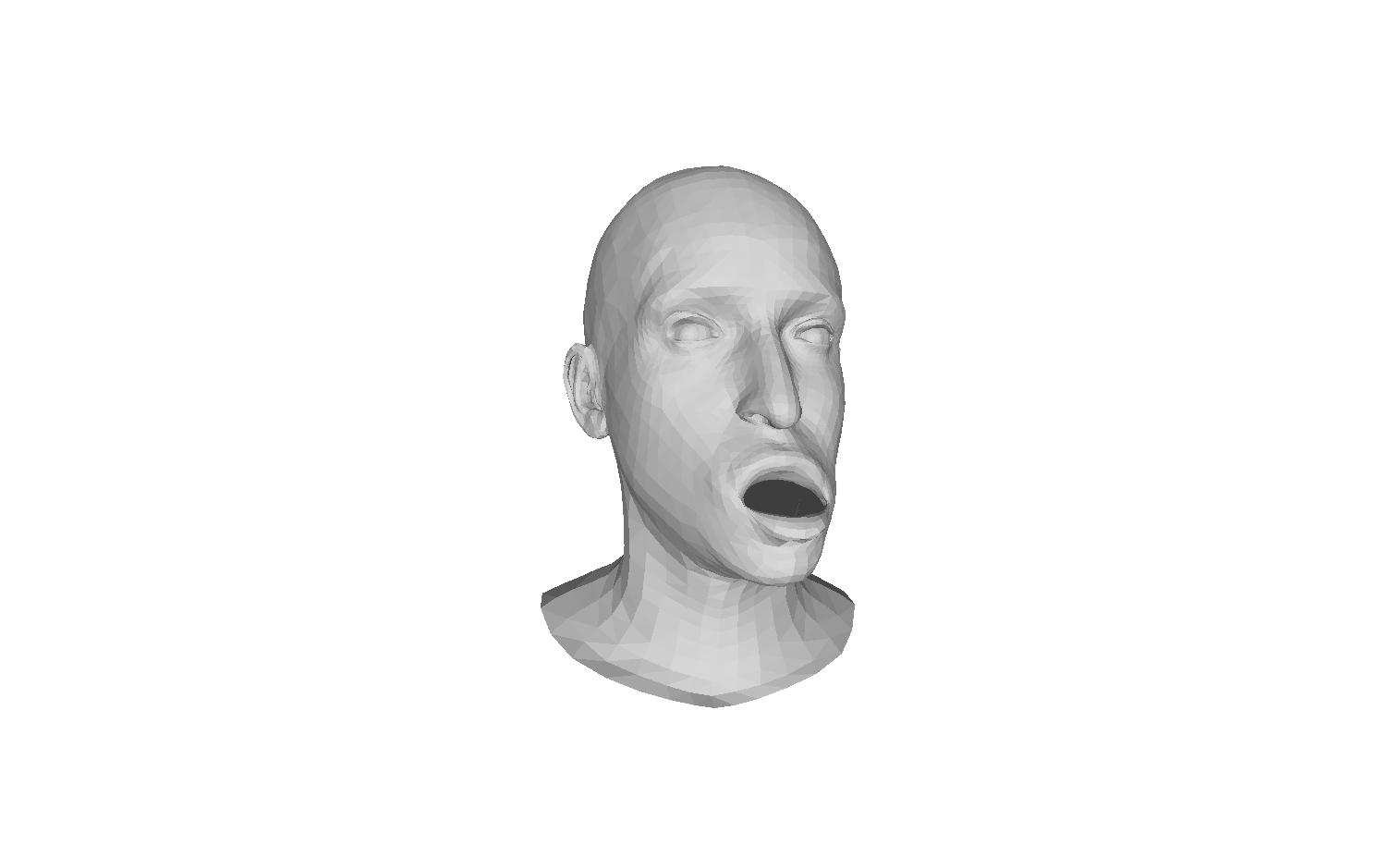}};
    \node[right of=g4, node distance=1.8cm] (g5) {\includegraphics[trim={400 80 400 100},clip,width=0.09\linewidth]{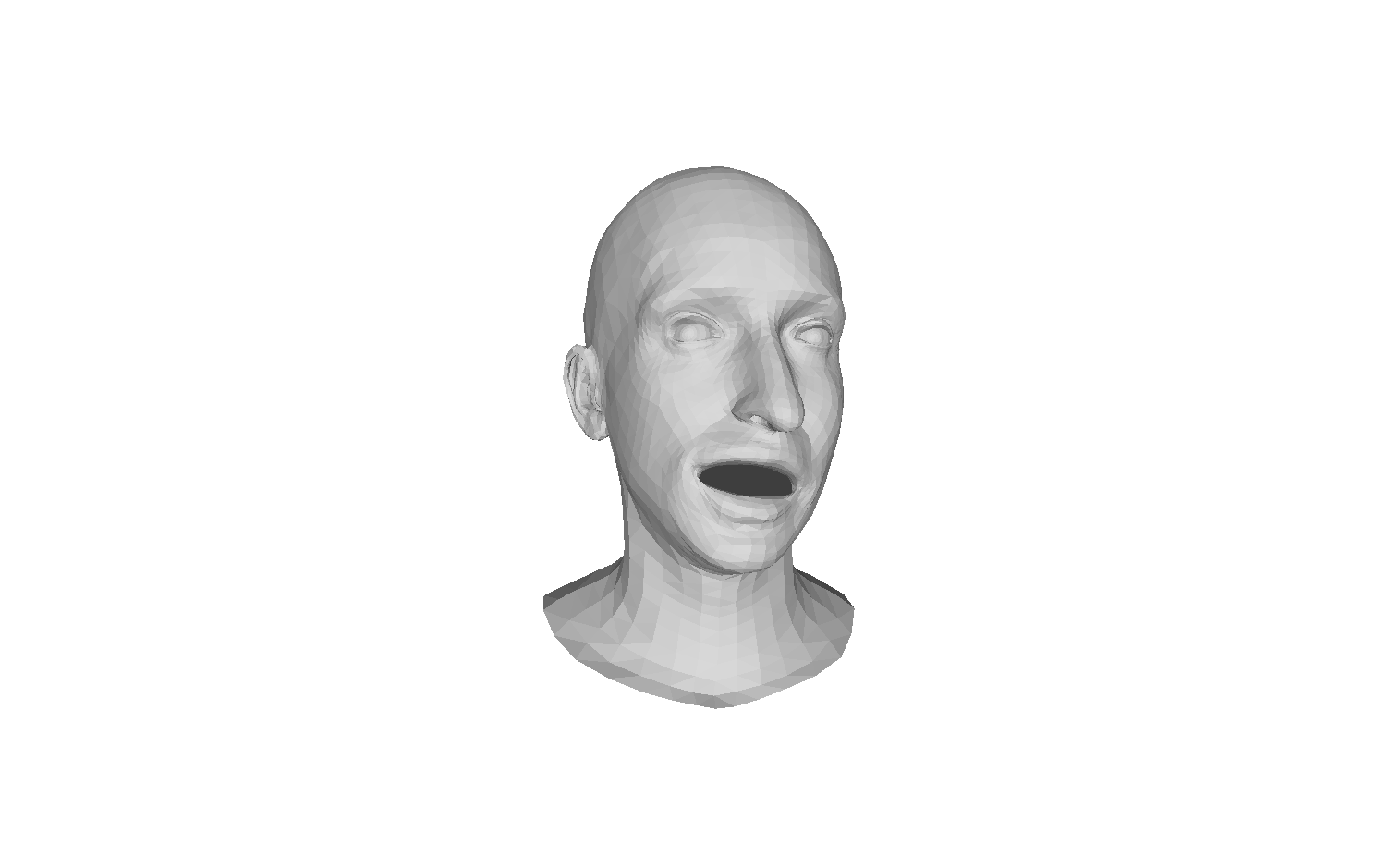}};
    \node[right of=g5, node distance=1.8cm] (g6) {\includegraphics[trim={400 80 400 100},clip,width=0.09\linewidth]{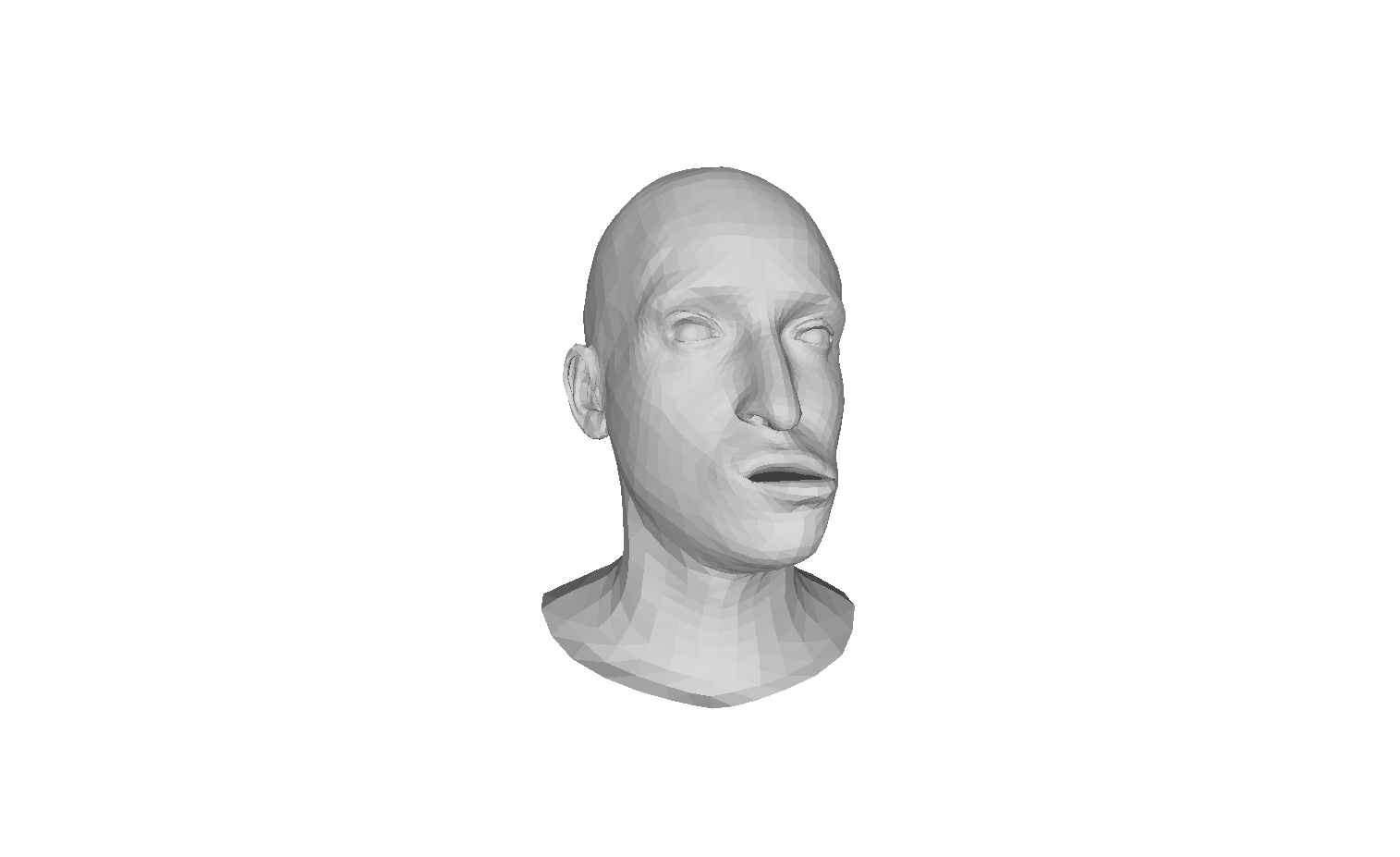}};
    \node[right of=g6, node distance=1.8cm] (g7) {\includegraphics[trim={400 80 400 100},clip,width=0.09\linewidth]{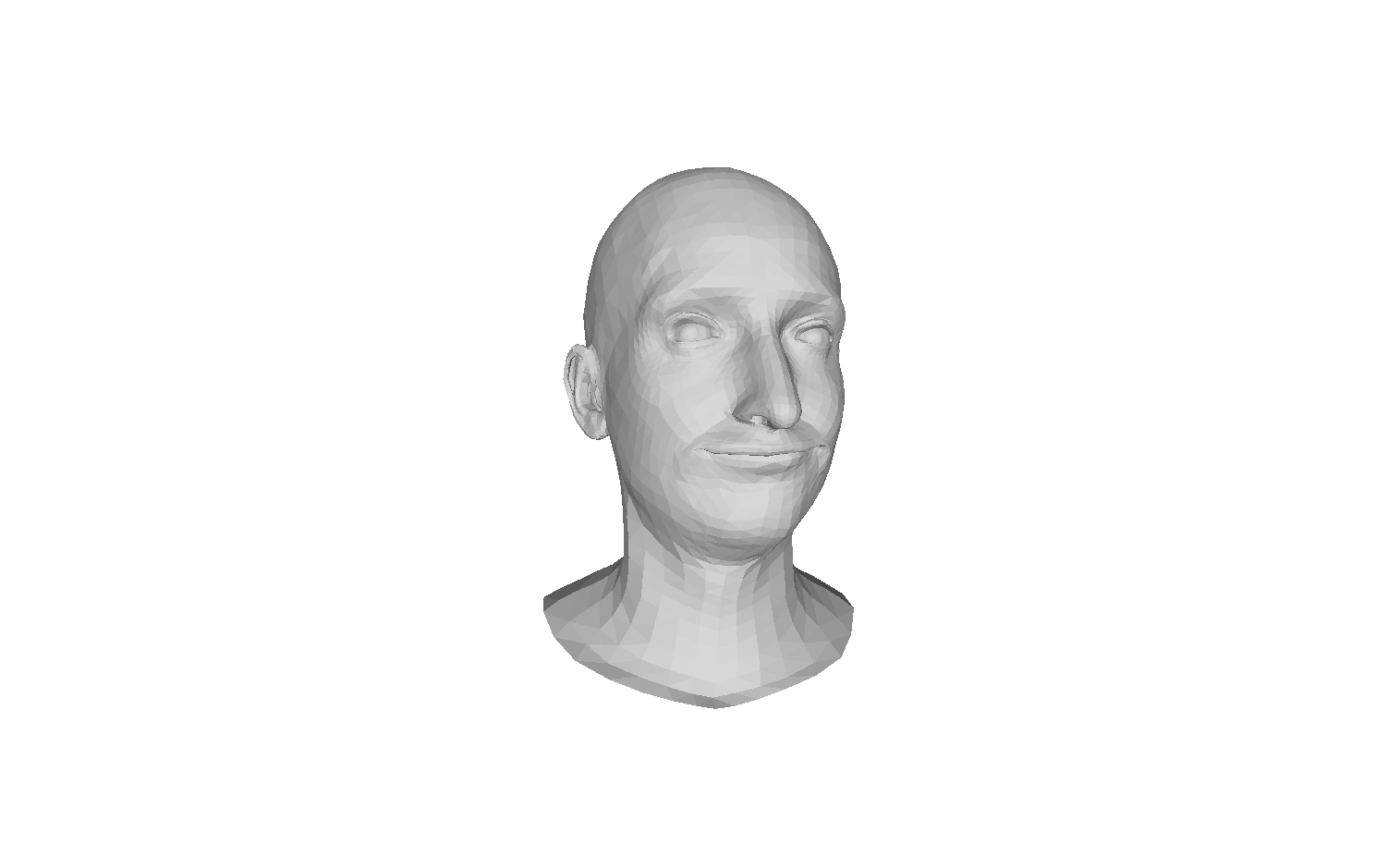}};
    \node[right of=g7, node distance=1.8cm] (g8) {\includegraphics[trim={400 80 400 100},clip,width=0.09\linewidth]{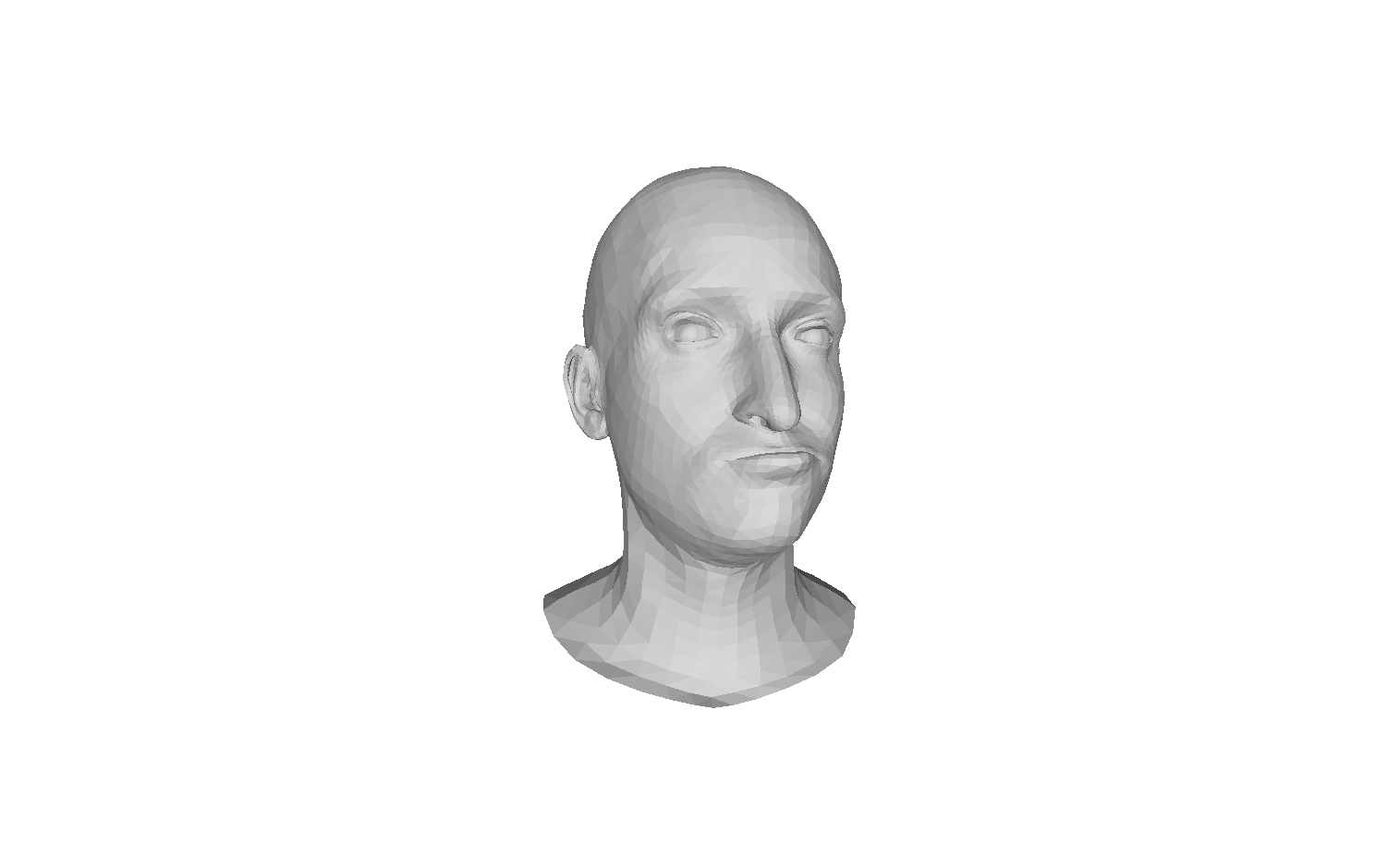}};
    \node[right of=g8, node distance=1.8cm] (g9) {\includegraphics[trim={400 80 400 100},clip,width=0.09\linewidth]{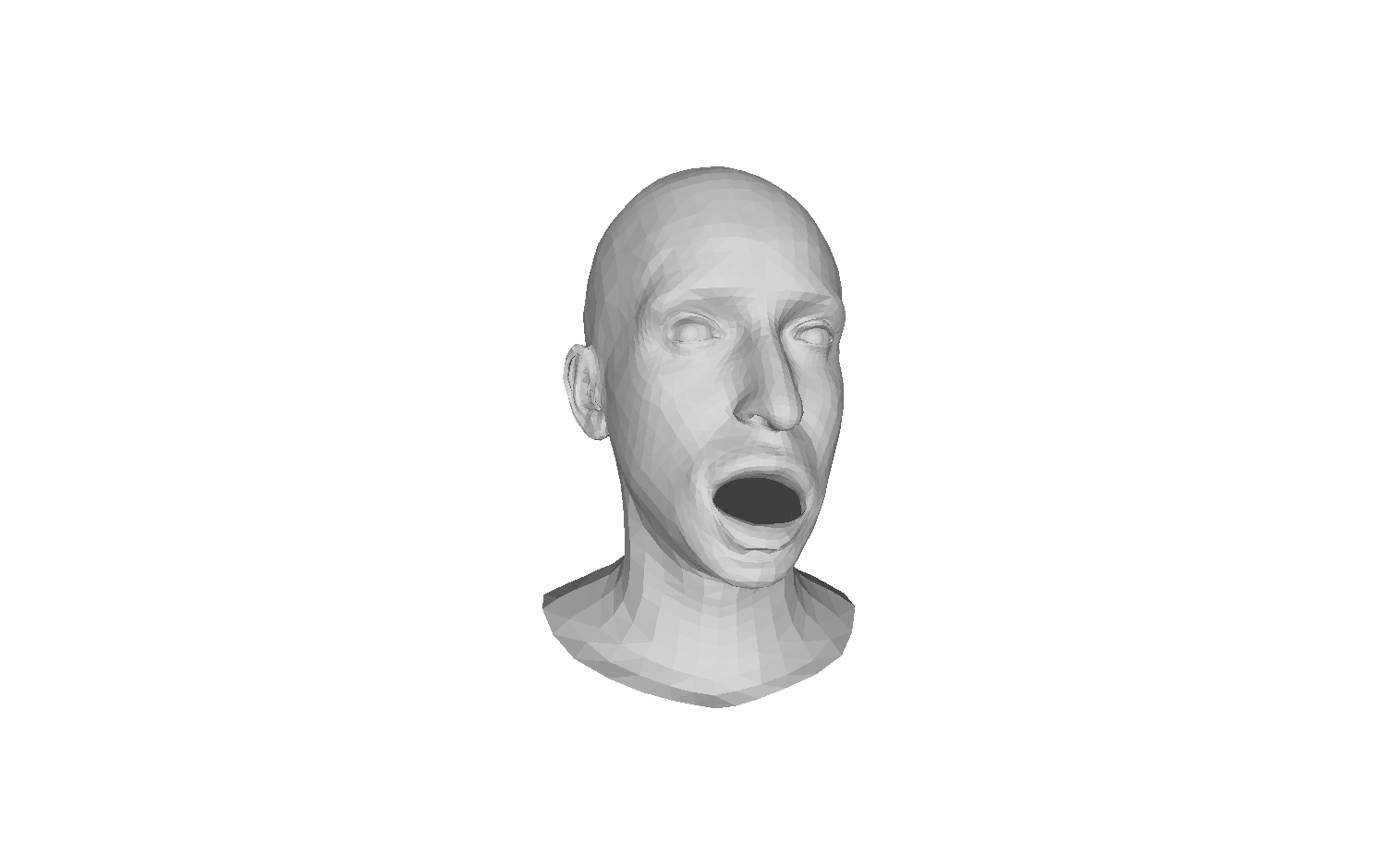}};
    
    \node[below of=g1, node distance=2.1cm] (h1) {\includegraphics[width=0.11\linewidth]{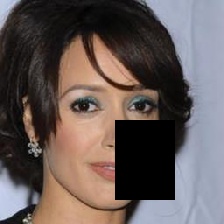}};
    \node[right of=h1, node distance=2.5cm] (h2) {\includegraphics[trim={400 80 400 100},clip,width=0.09\linewidth]{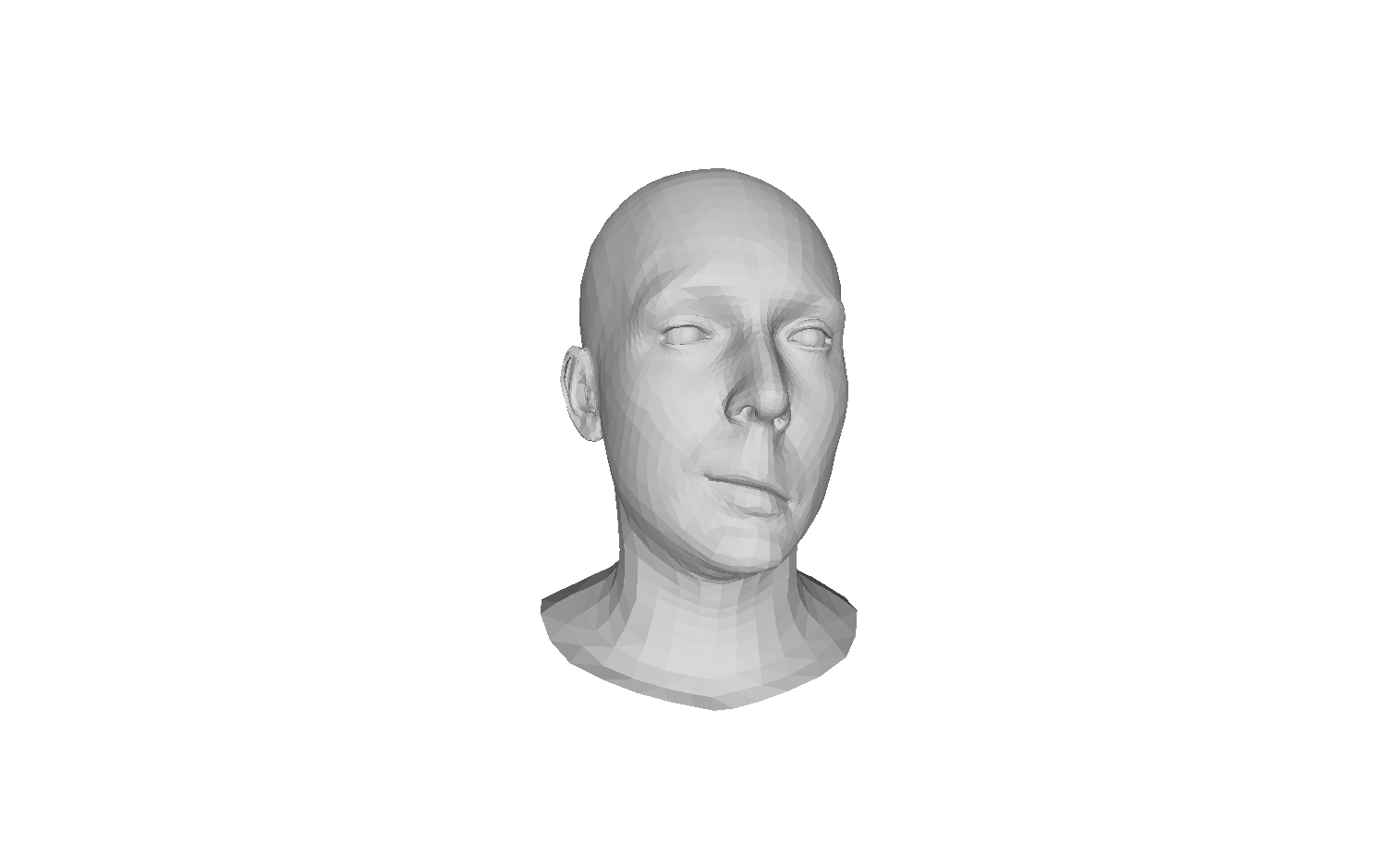}};
    \node[right of=h2, node distance=1.8cm] (h3) {\includegraphics[trim={400 80 400 100},clip,width=0.09\linewidth]{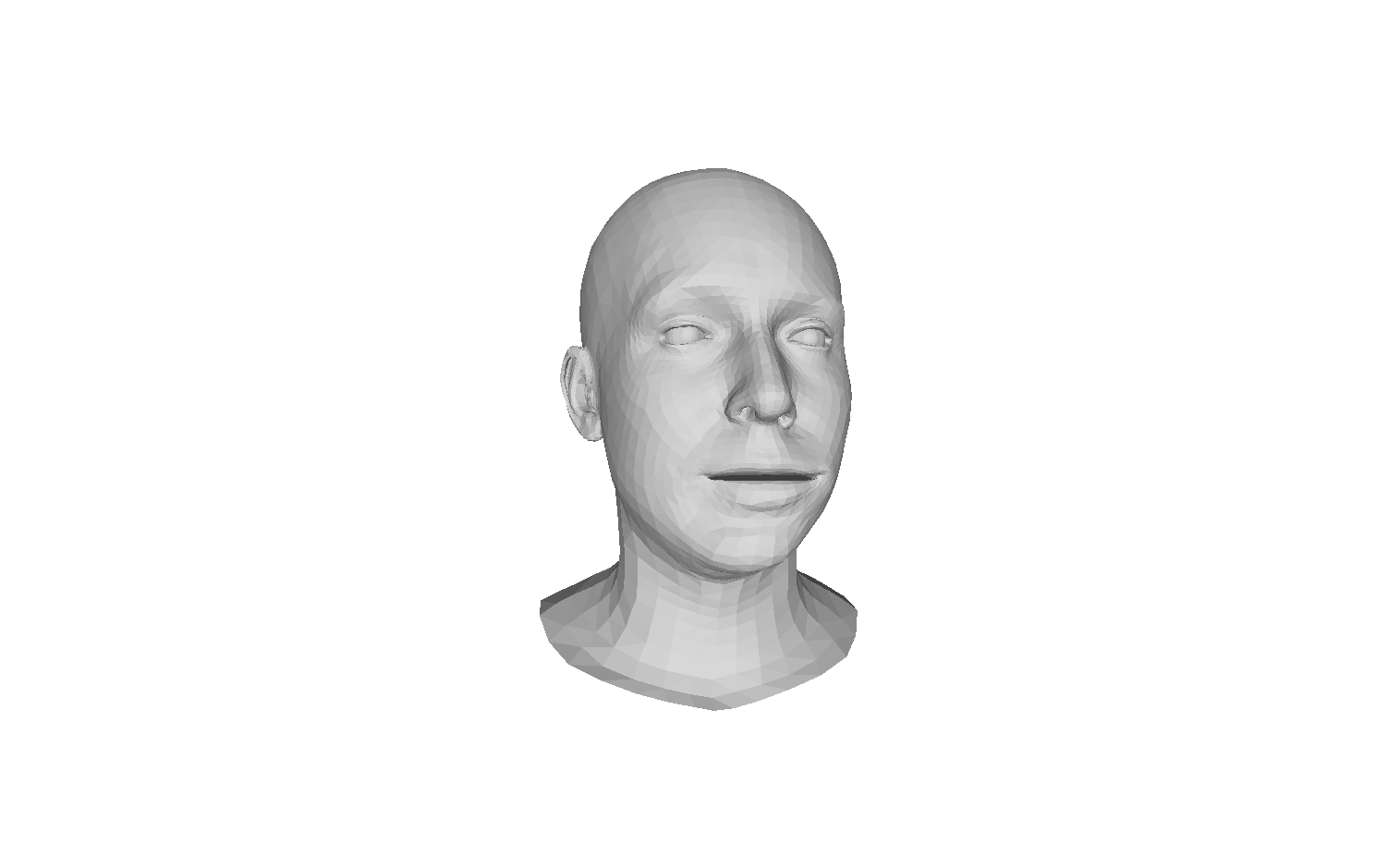}};
    \node[right of=h3, node distance=1.8cm] (h4) {\includegraphics[trim={400 80 400 100},clip,width=0.09\linewidth]{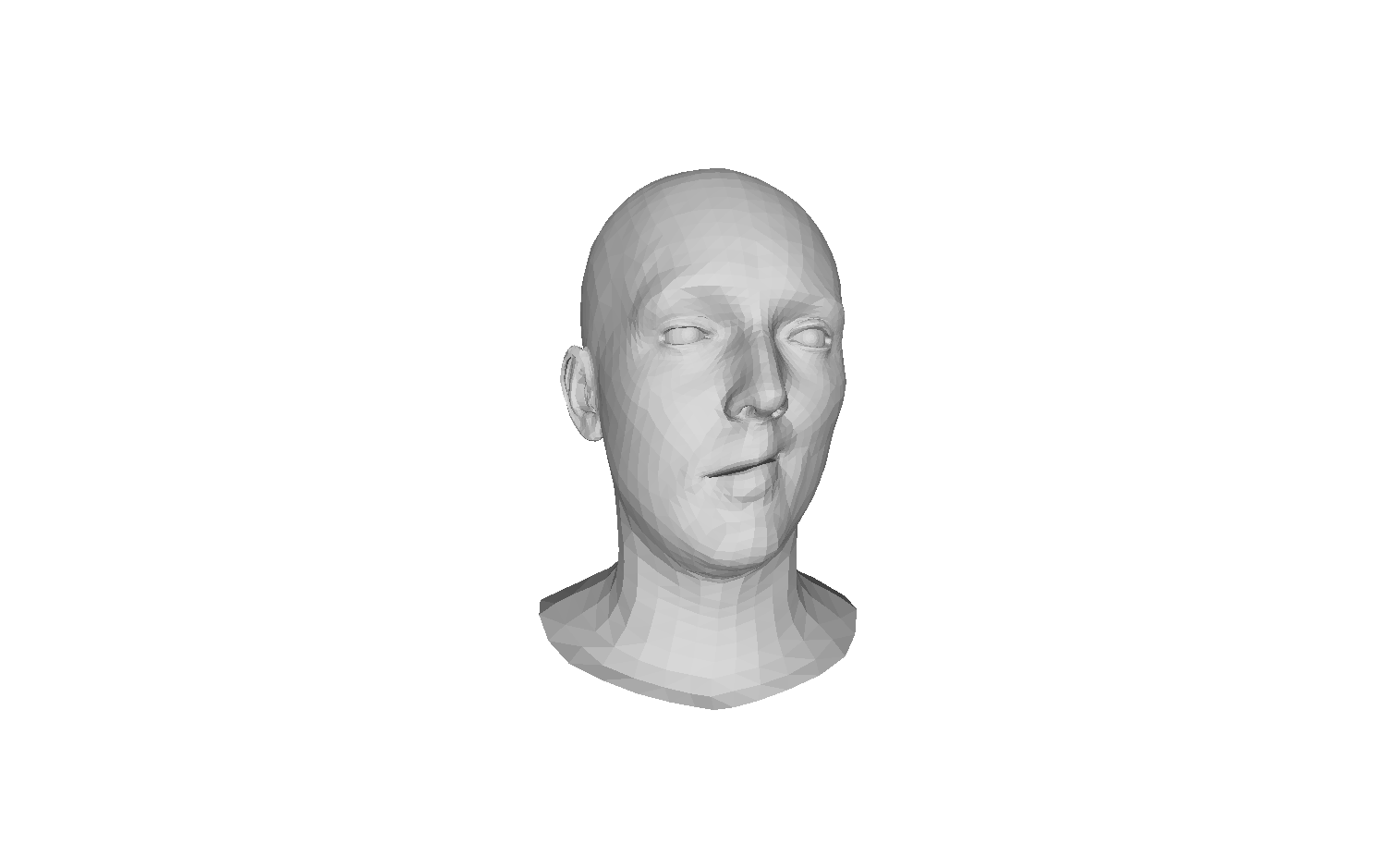}};
    \node[right of=h4, node distance=1.8cm] (h5) {\includegraphics[trim={400 80 400 100},clip,width=0.09\linewidth]{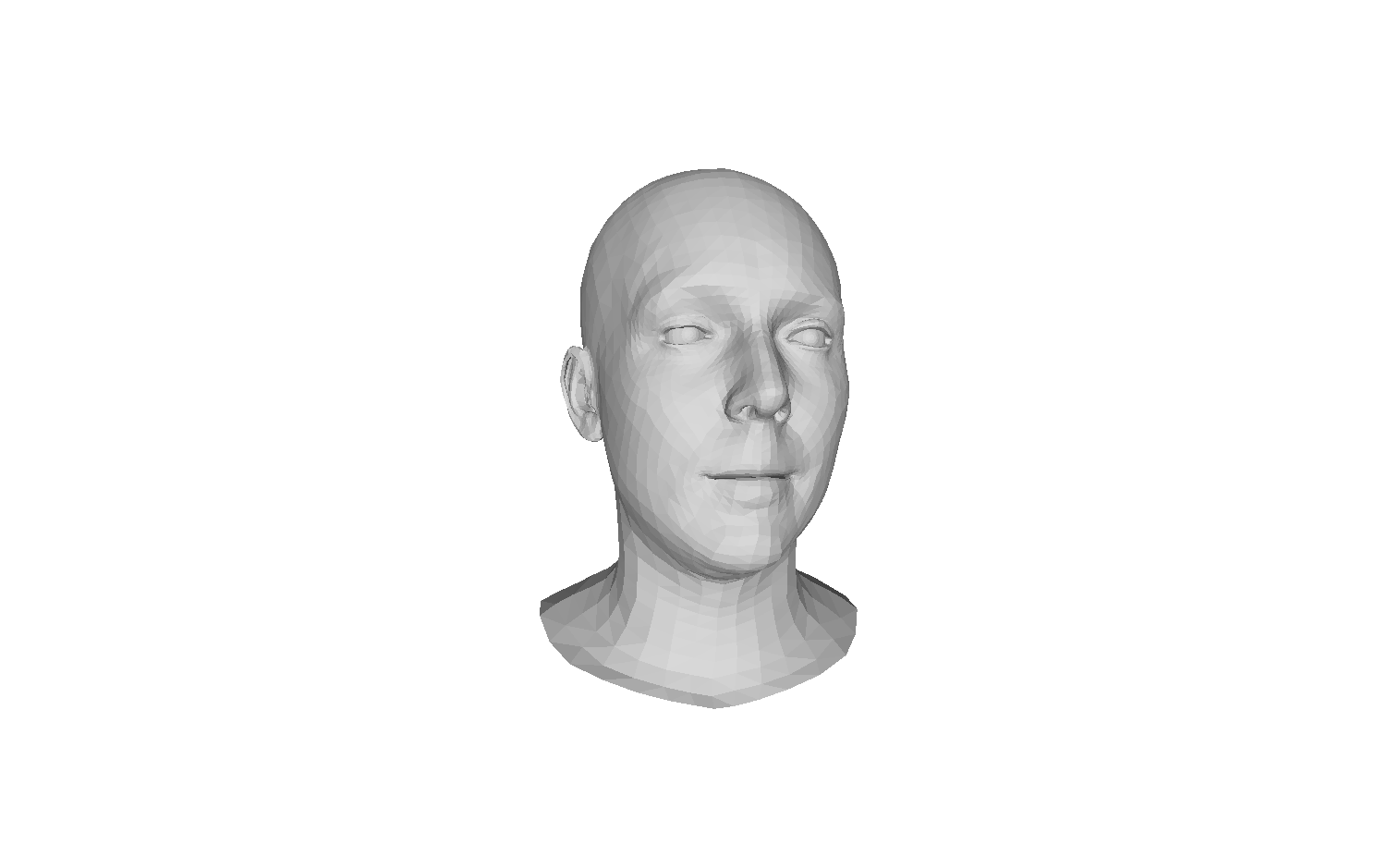}};
    \node[right of=h5, node distance=1.8cm] (h6) {\includegraphics[trim={400 80 400 100},clip,width=0.09\linewidth]{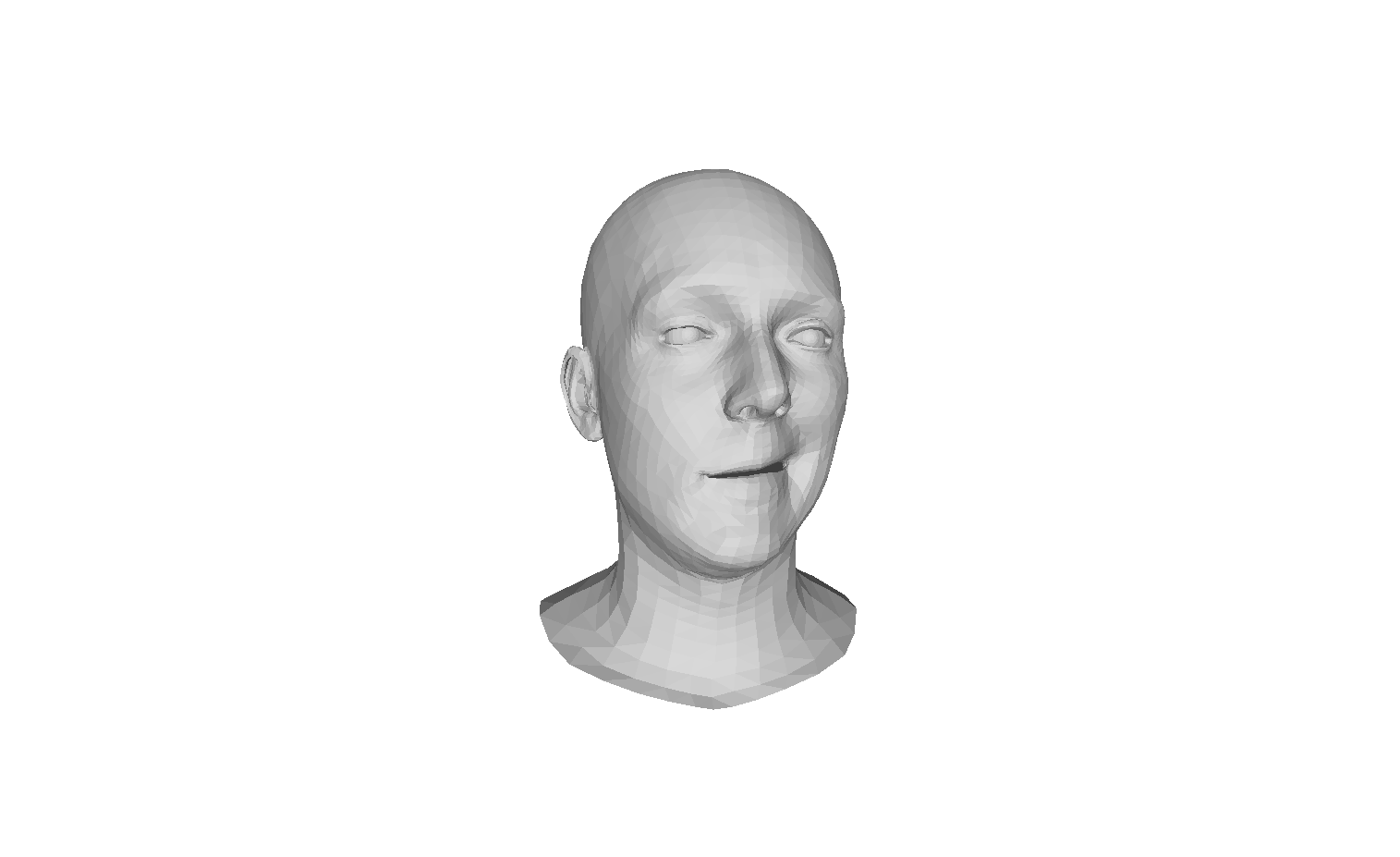}};
    \node[right of=h6, node distance=1.8cm] (h7) {\includegraphics[trim={400 80 400 100},clip,width=0.09\linewidth]{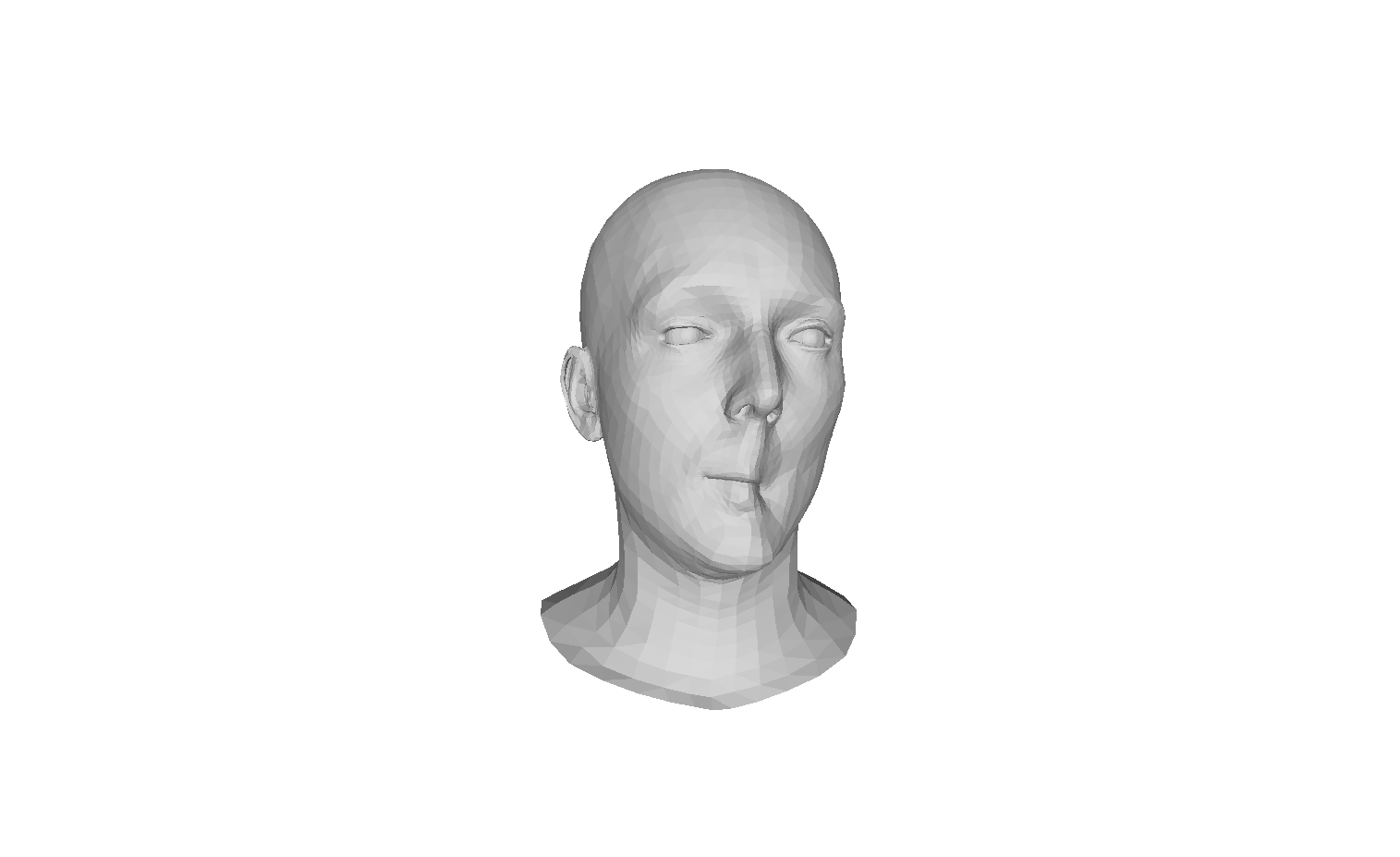}};
    \node[right of=h7, node distance=1.8cm] (h8) {\includegraphics[trim={400 80 400 100},clip,width=0.09\linewidth]{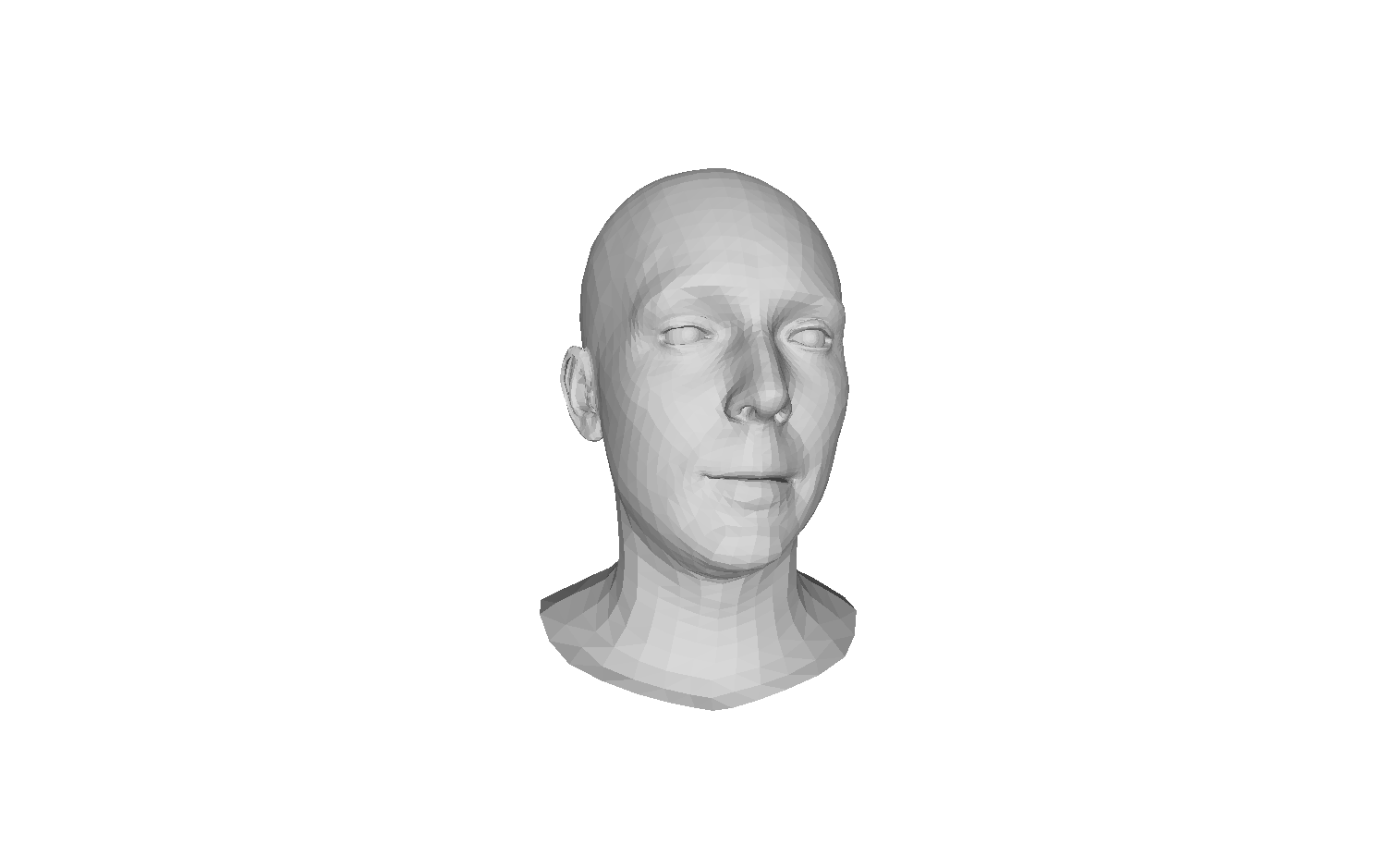}};
    \node[right of=h8, node distance=1.8cm] (h9) {\includegraphics[trim={400 80 400 100},clip,width=0.09\linewidth]{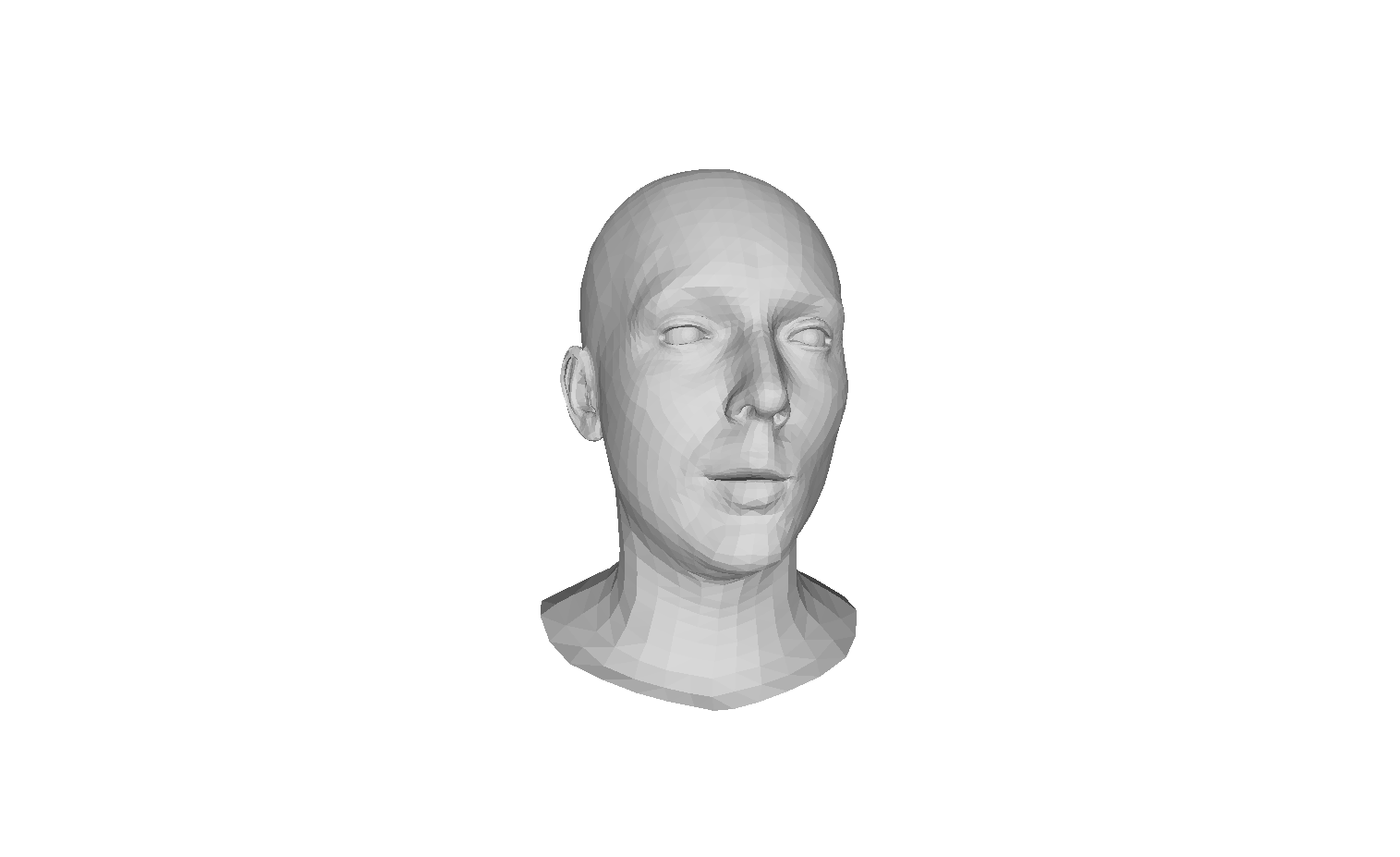}};
    
    \node[below of=h1, node distance=2.1cm] (i1) {\includegraphics[width=0.11\linewidth]{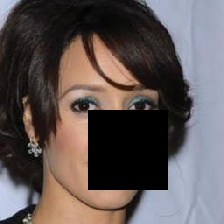}};
    \node[right of=i1, node distance=2.5cm] (i2) {\includegraphics[trim={400 80 400 100},clip,width=0.09\linewidth]{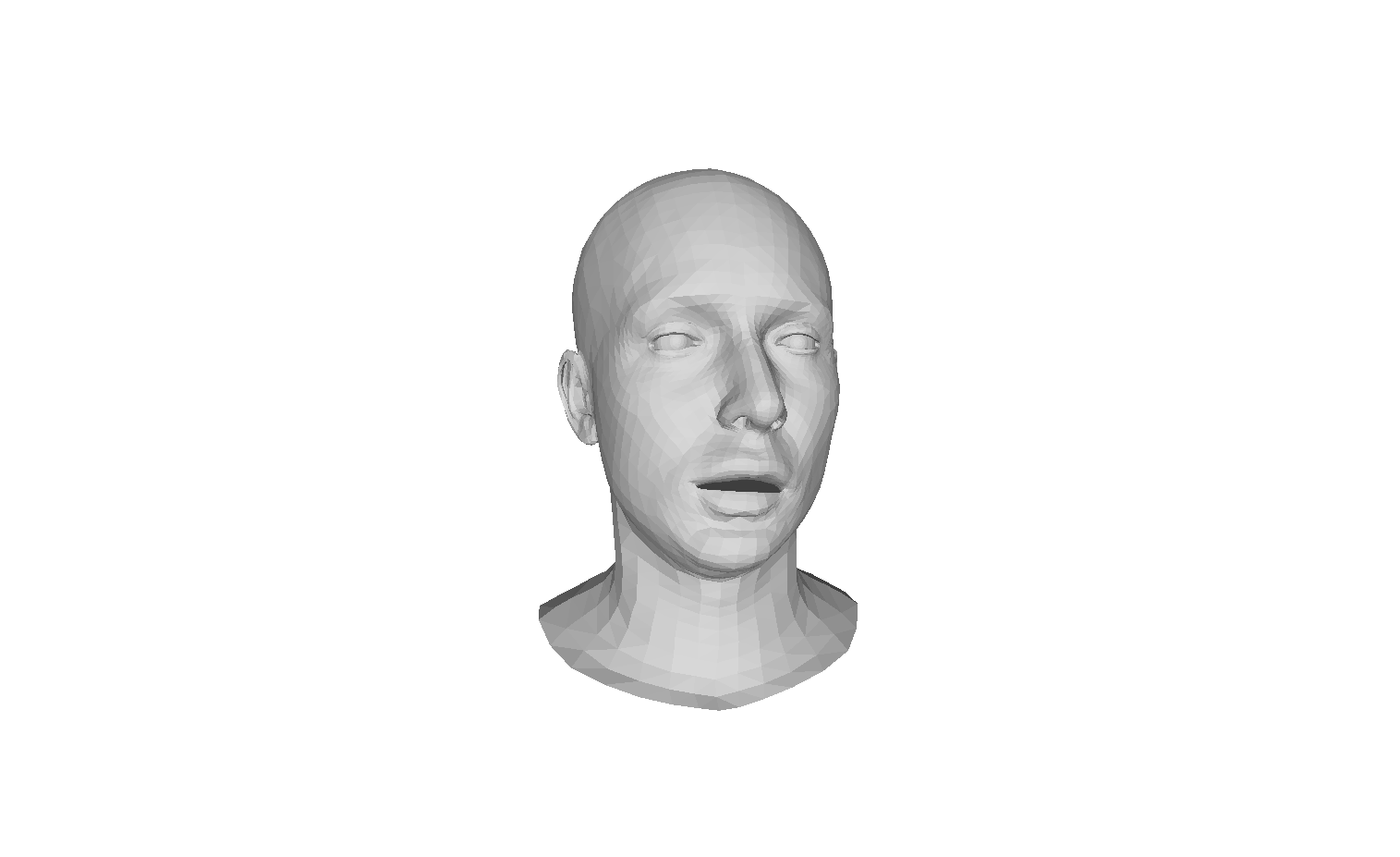}};
    \node[right of=i2, node distance=1.8cm] (i3) {\includegraphics[trim={400 80 400 100},clip,width=0.09\linewidth]{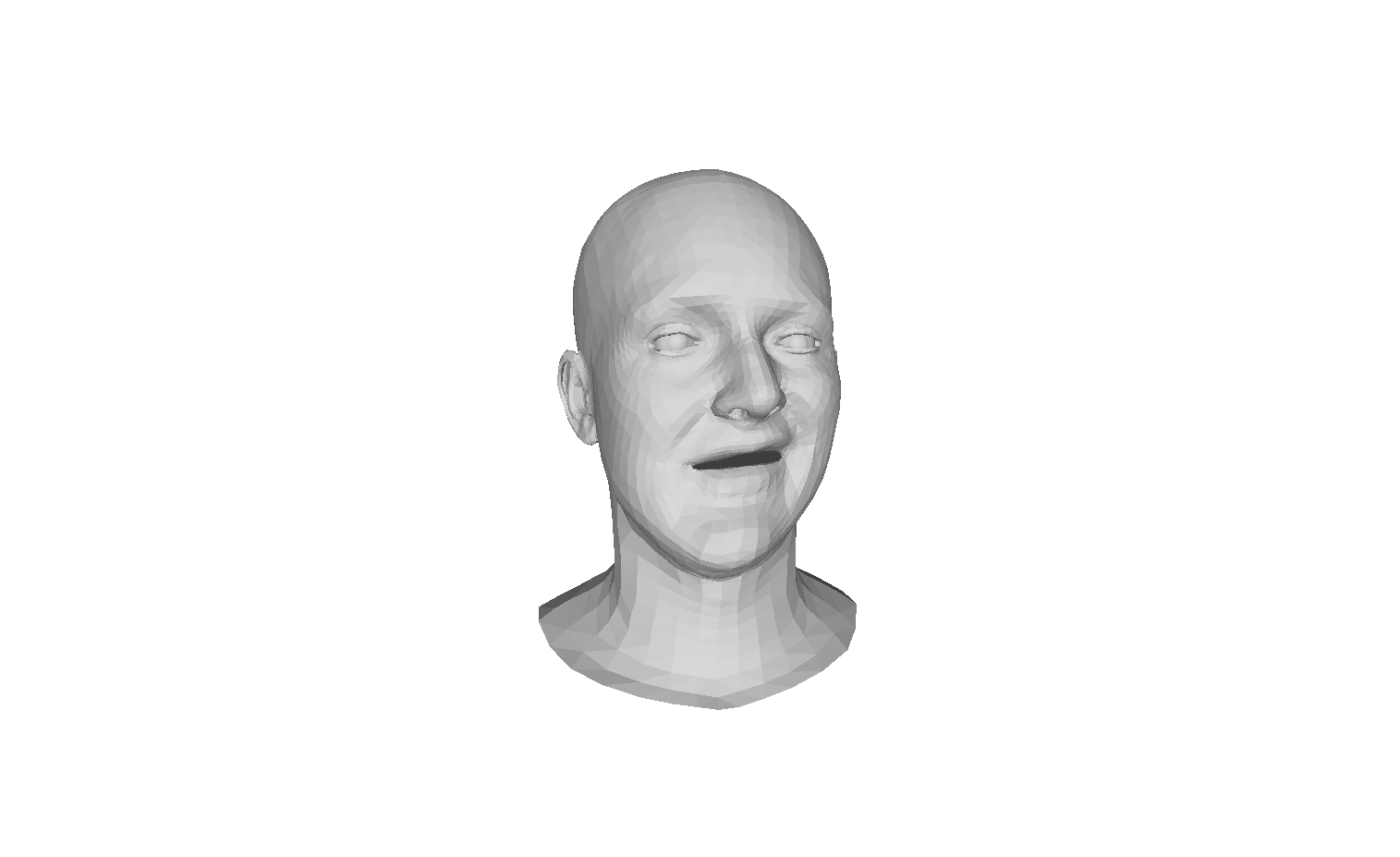}};
    \node[right of=i3, node distance=1.8cm] (i4) {\includegraphics[trim={400 80 400 100},clip,width=0.09\linewidth]{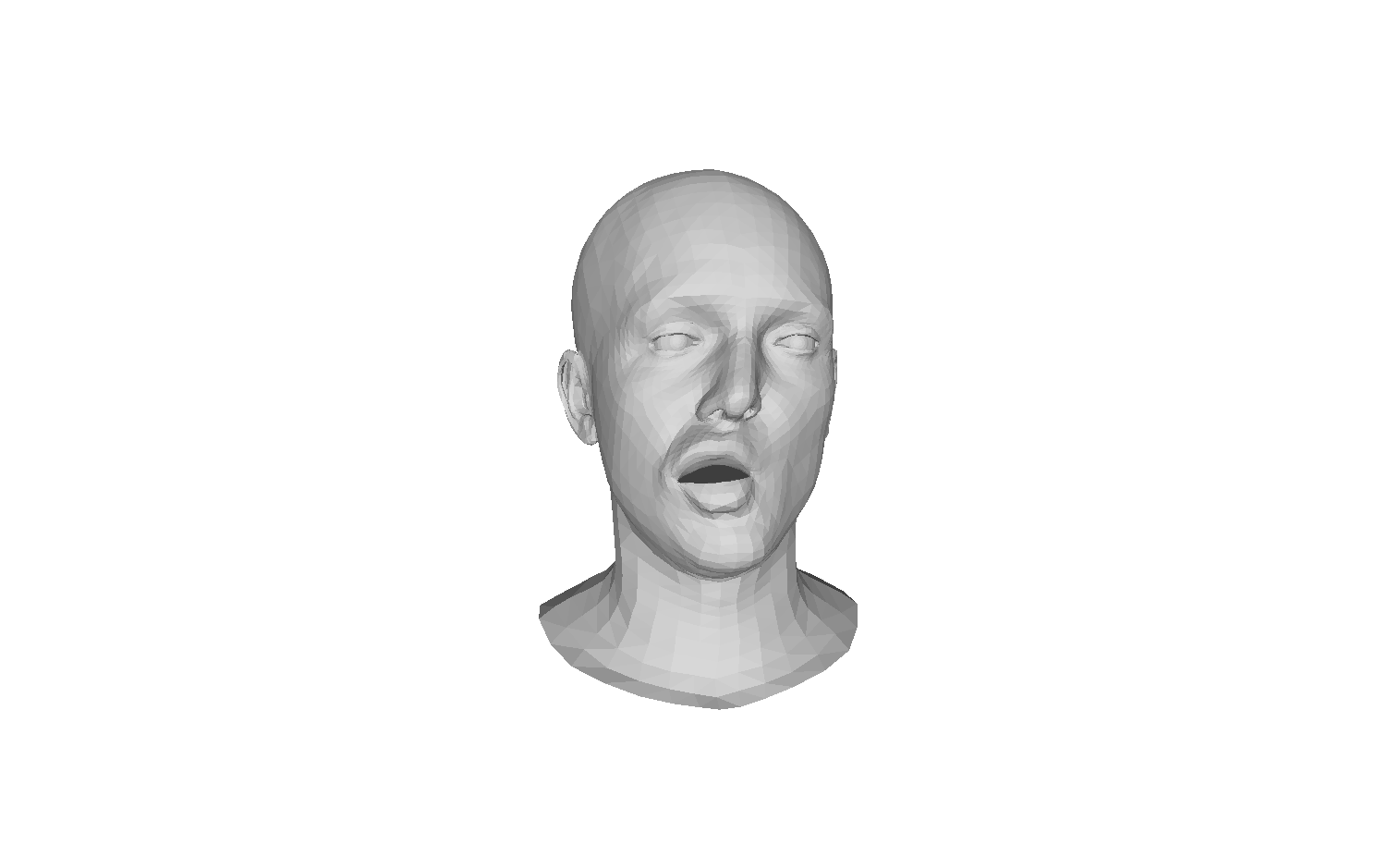}};
    \node[right of=i4, node distance=1.8cm] (i5) {\includegraphics[trim={400 80 400 100},clip,width=0.09\linewidth]{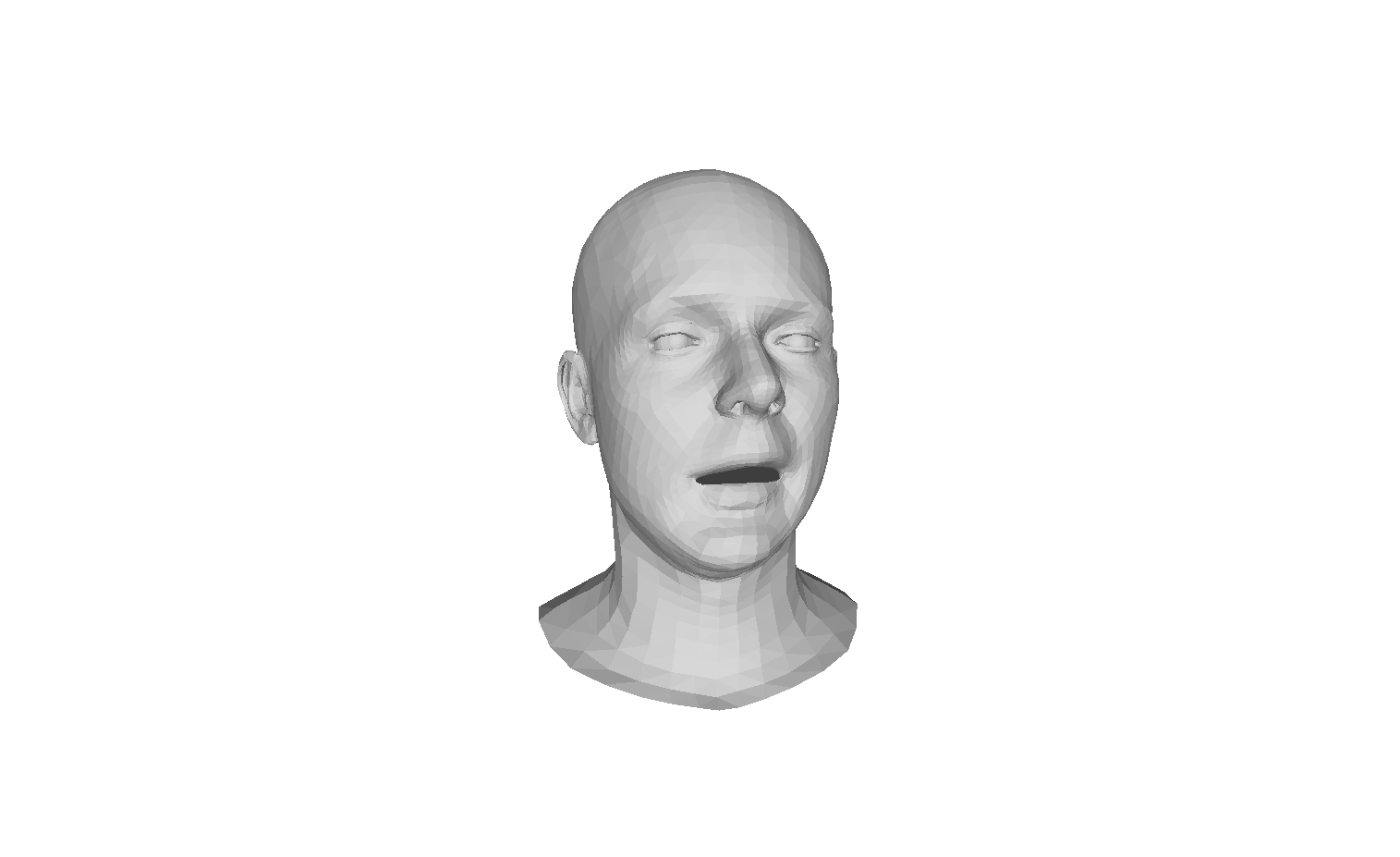}};
    \node[right of=i5, node distance=1.8cm] (i6) {\includegraphics[trim={400 80 400 100},clip,width=0.09\linewidth]{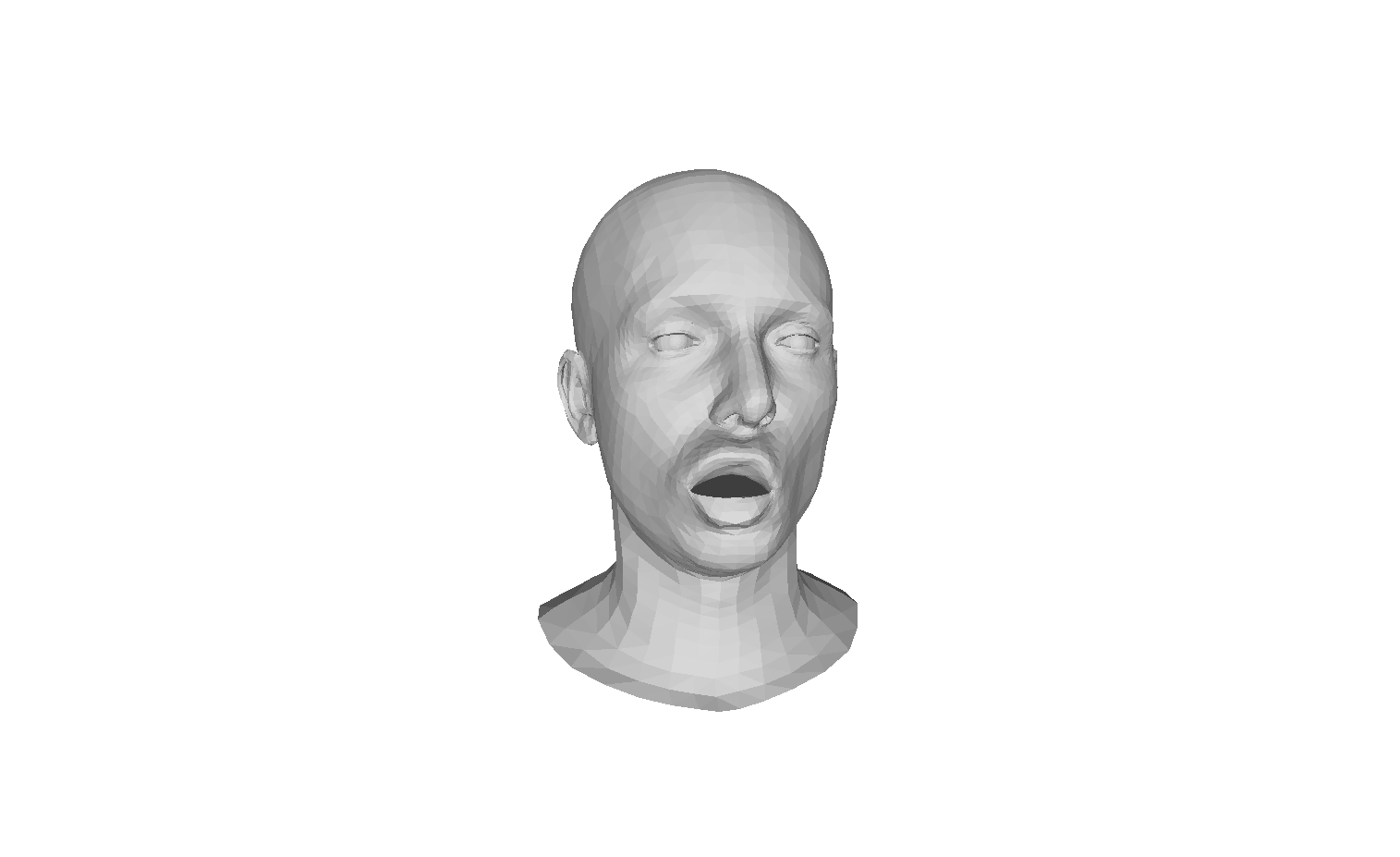}};
    \node[right of=i6, node distance=1.8cm] (i7) {\includegraphics[trim={400 80 400 100},clip,width=0.09\linewidth]{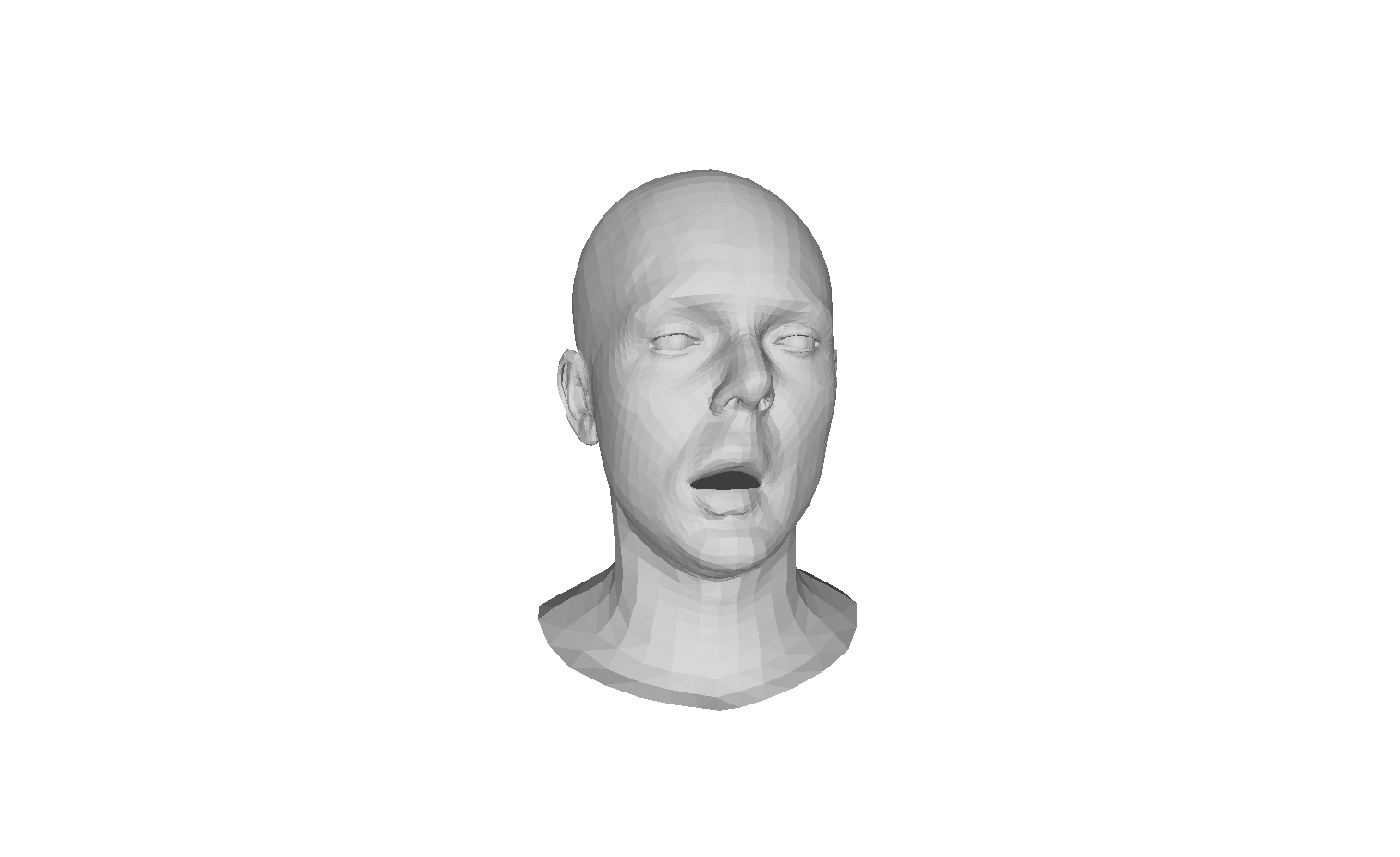}};
    \node[right of=i7, node distance=1.8cm] (i8) {\includegraphics[trim={400 80 400 100},clip,width=0.09\linewidth]{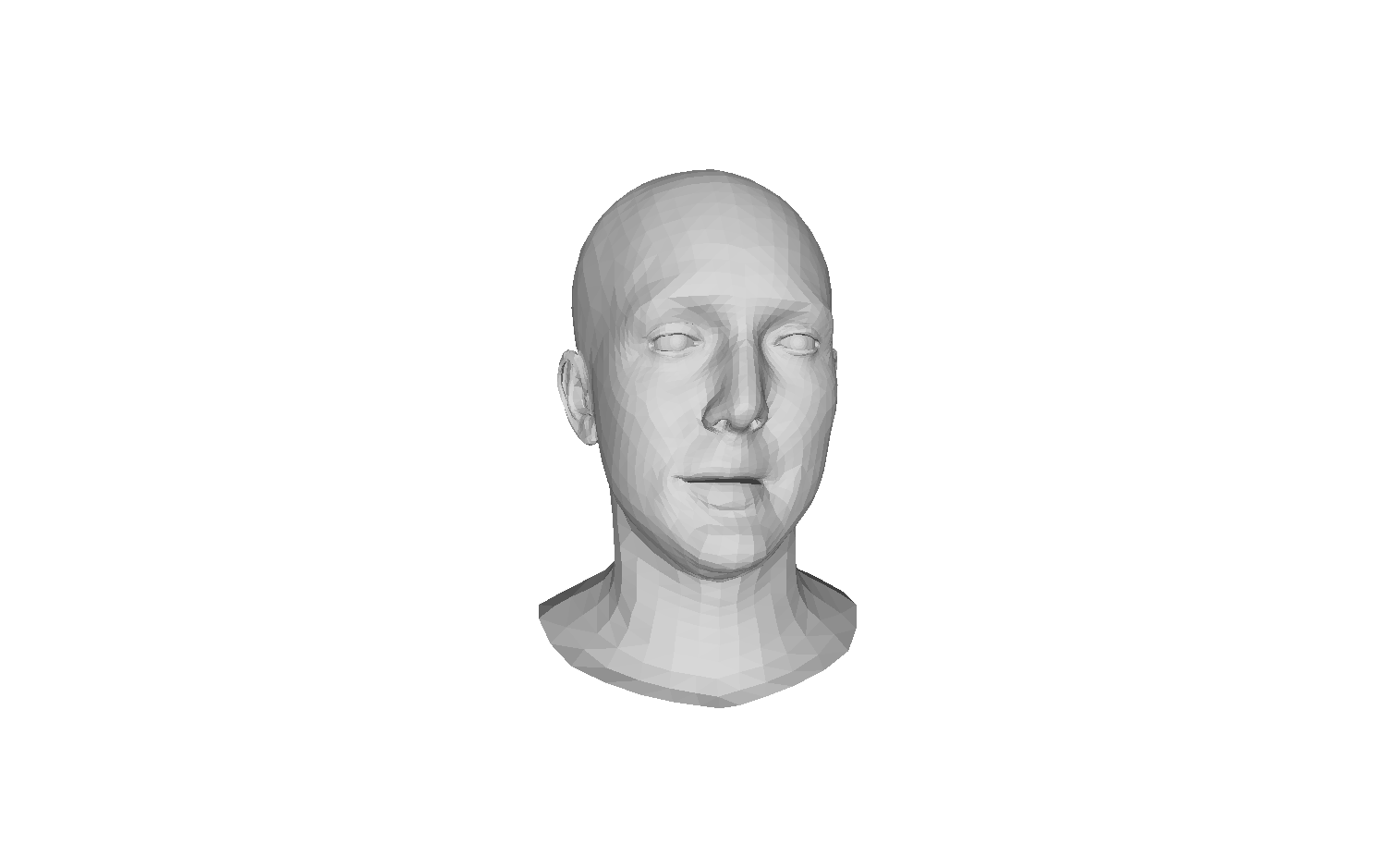}};
    \node[right of=i8, node distance=1.8cm] (i9) {\includegraphics[trim={400 80 400 100},clip,width=0.09\linewidth]{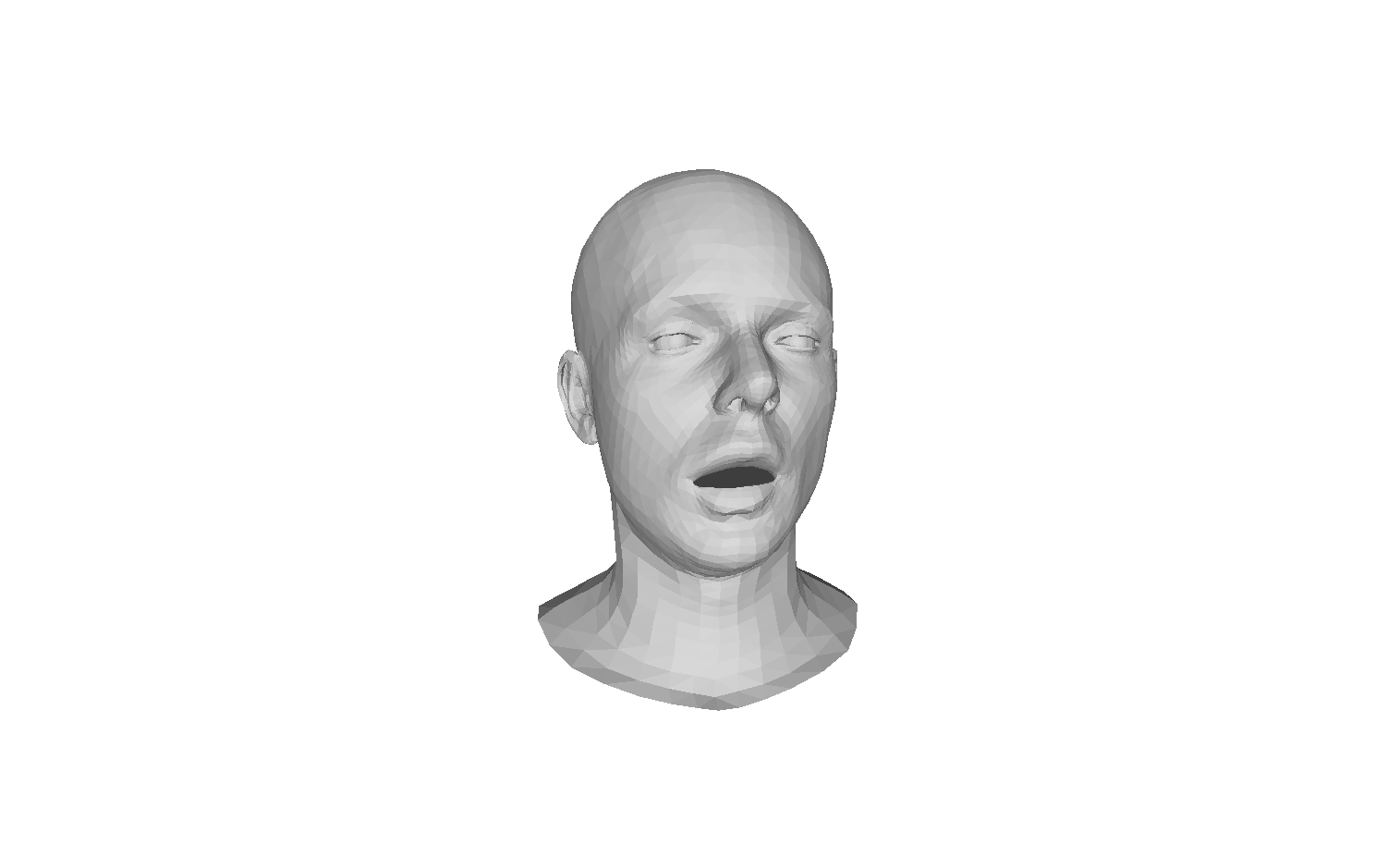}};
    
    \node[below of=i1, node distance=2.1cm] (j1) {\includegraphics[width=0.11\linewidth]{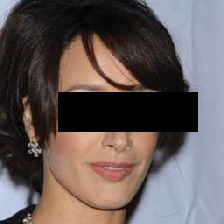}};
    \node[right of=j1, node distance=2.5cm] (j2) {\includegraphics[trim={400 80 400 100},clip,width=0.09\linewidth]{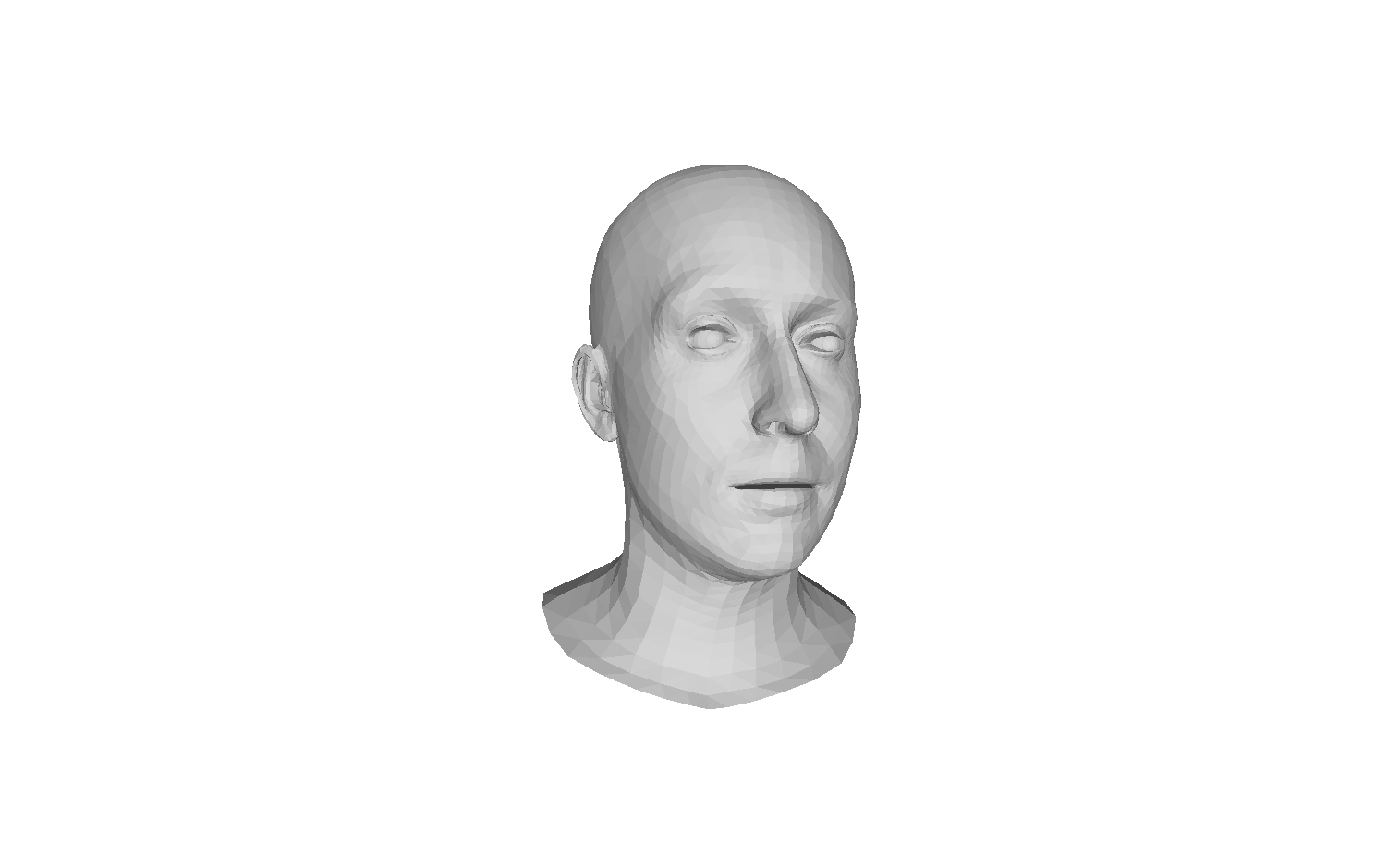}};
    \node[right of=j2, node distance=1.8cm] (j3) {\includegraphics[trim={400 80 400 100},clip,width=0.09\linewidth]{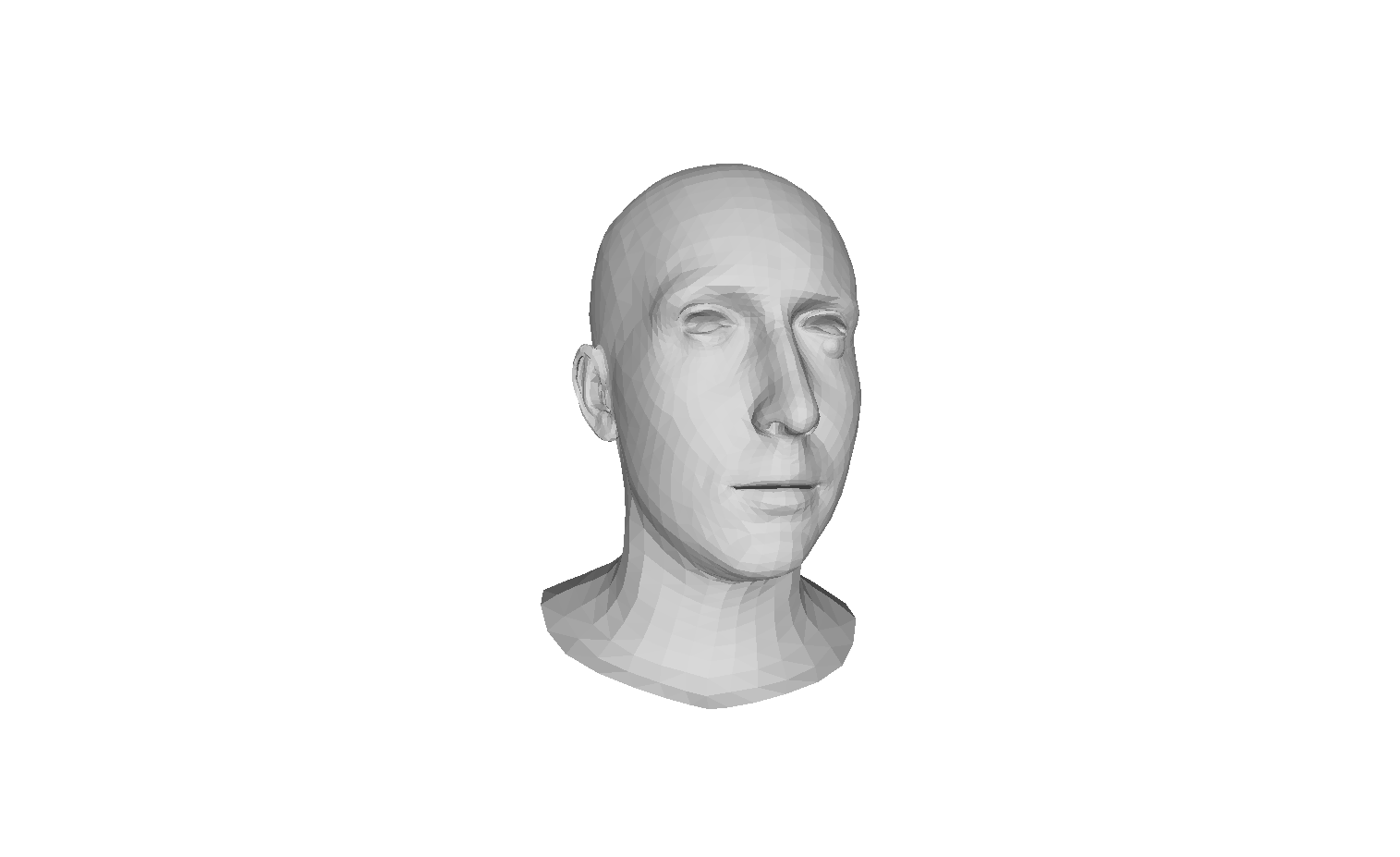}};
    \node[right of=j3, node distance=1.8cm] (j4) {\includegraphics[trim={400 80 400 100},clip,width=0.09\linewidth]{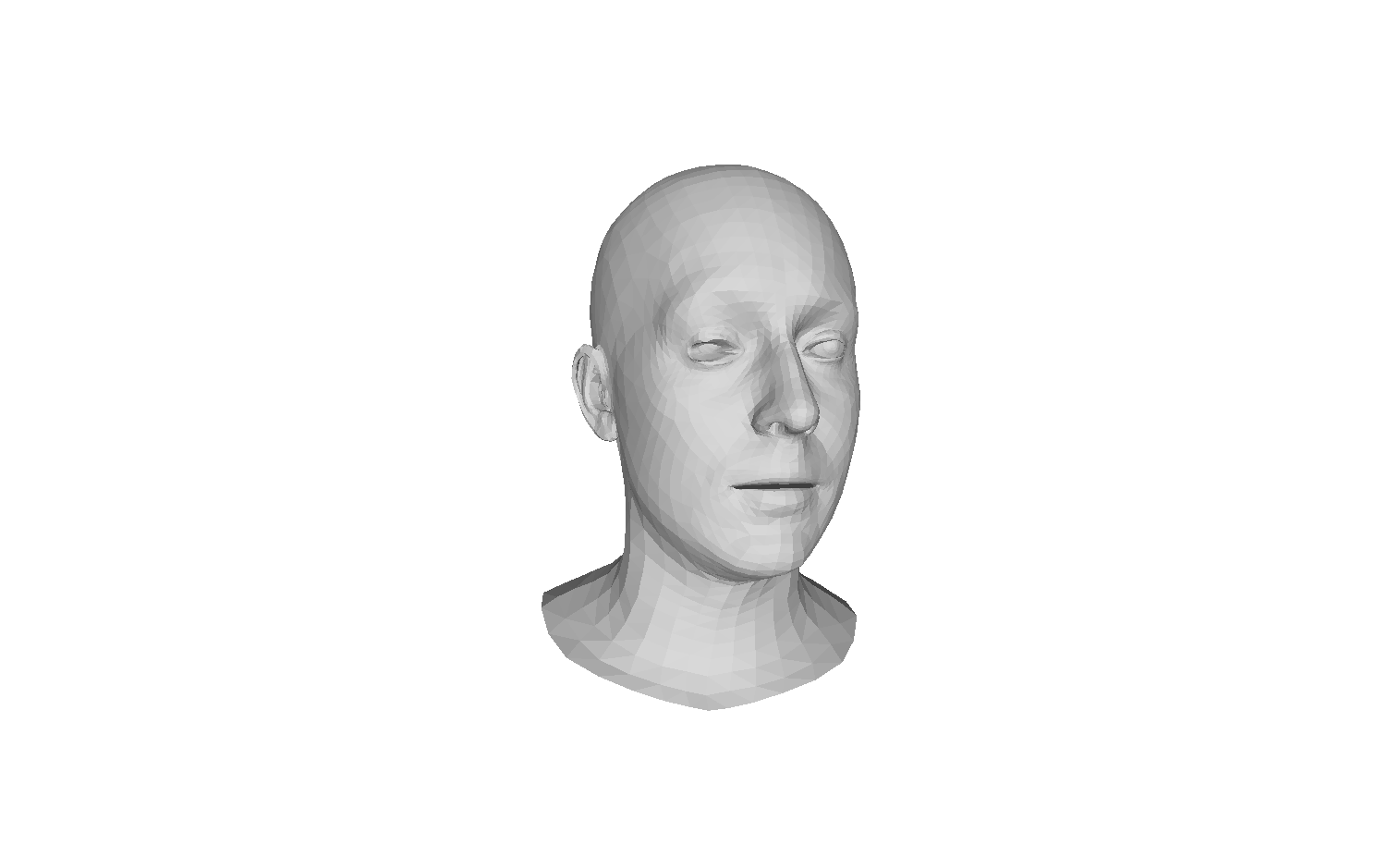}};
    \node[right of=j4, node distance=1.8cm] (j5) {\includegraphics[trim={400 80 400 100},clip,width=0.09\linewidth]{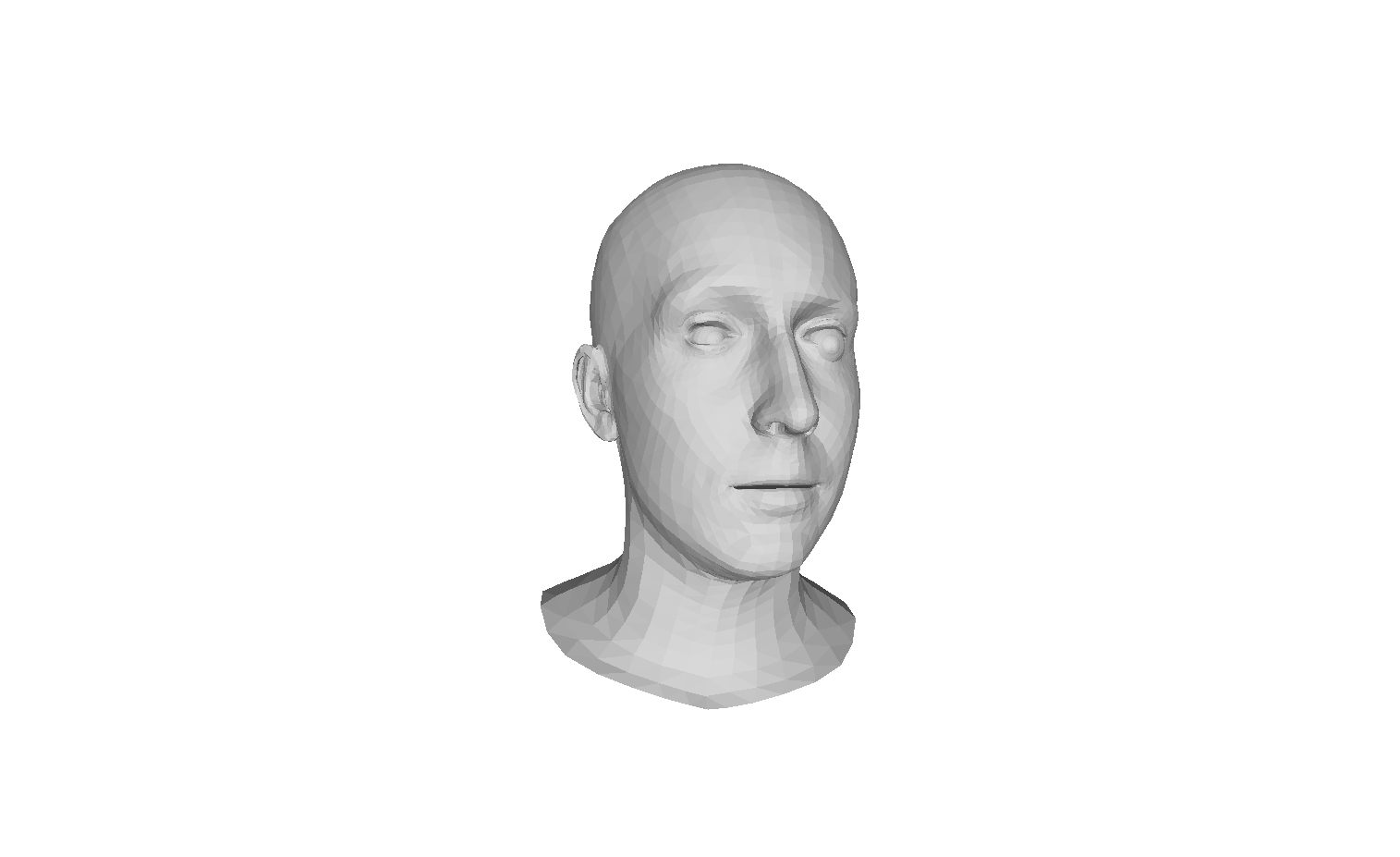}};
    \node[right of=j5, node distance=1.8cm] (j6) {\includegraphics[trim={400 80 400 100},clip,width=0.09\linewidth]{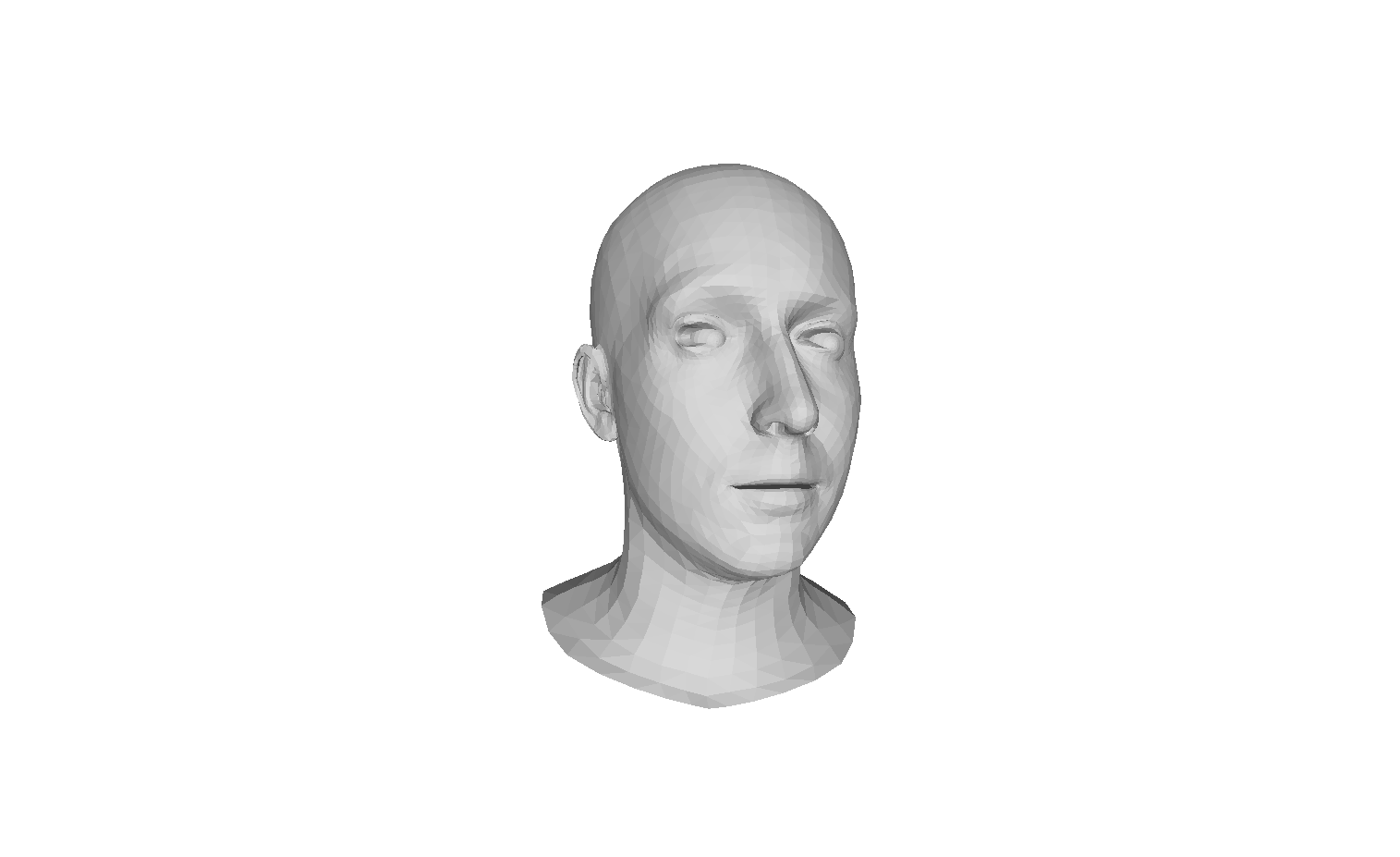}};
    \node[right of=j6, node distance=1.8cm] (j7) {\includegraphics[trim={400 80 400 100},clip,width=0.09\linewidth]{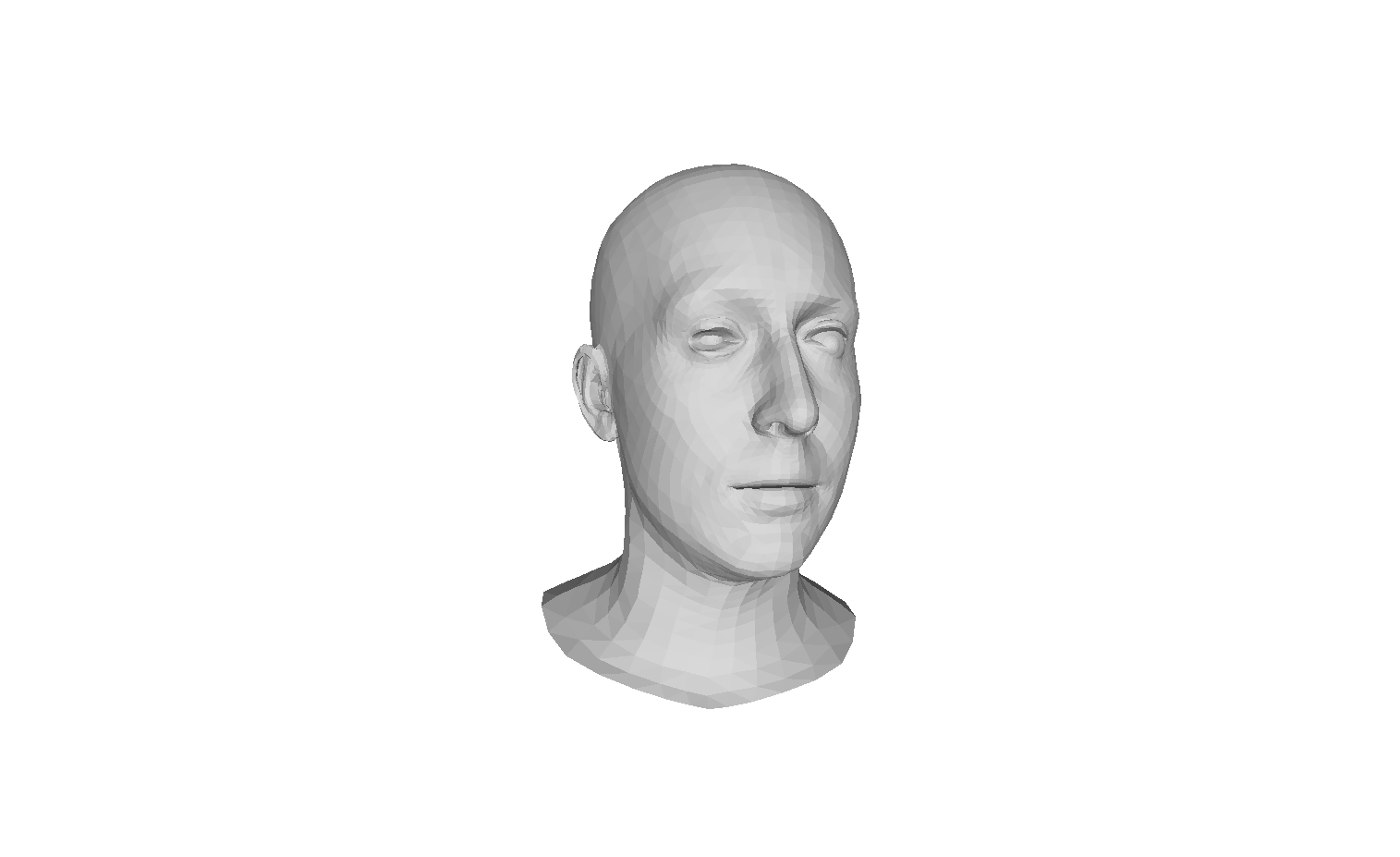}};
    \node[right of=j7, node distance=1.8cm] (j8) {\includegraphics[trim={400 80 400 100},clip,width=0.09\linewidth]{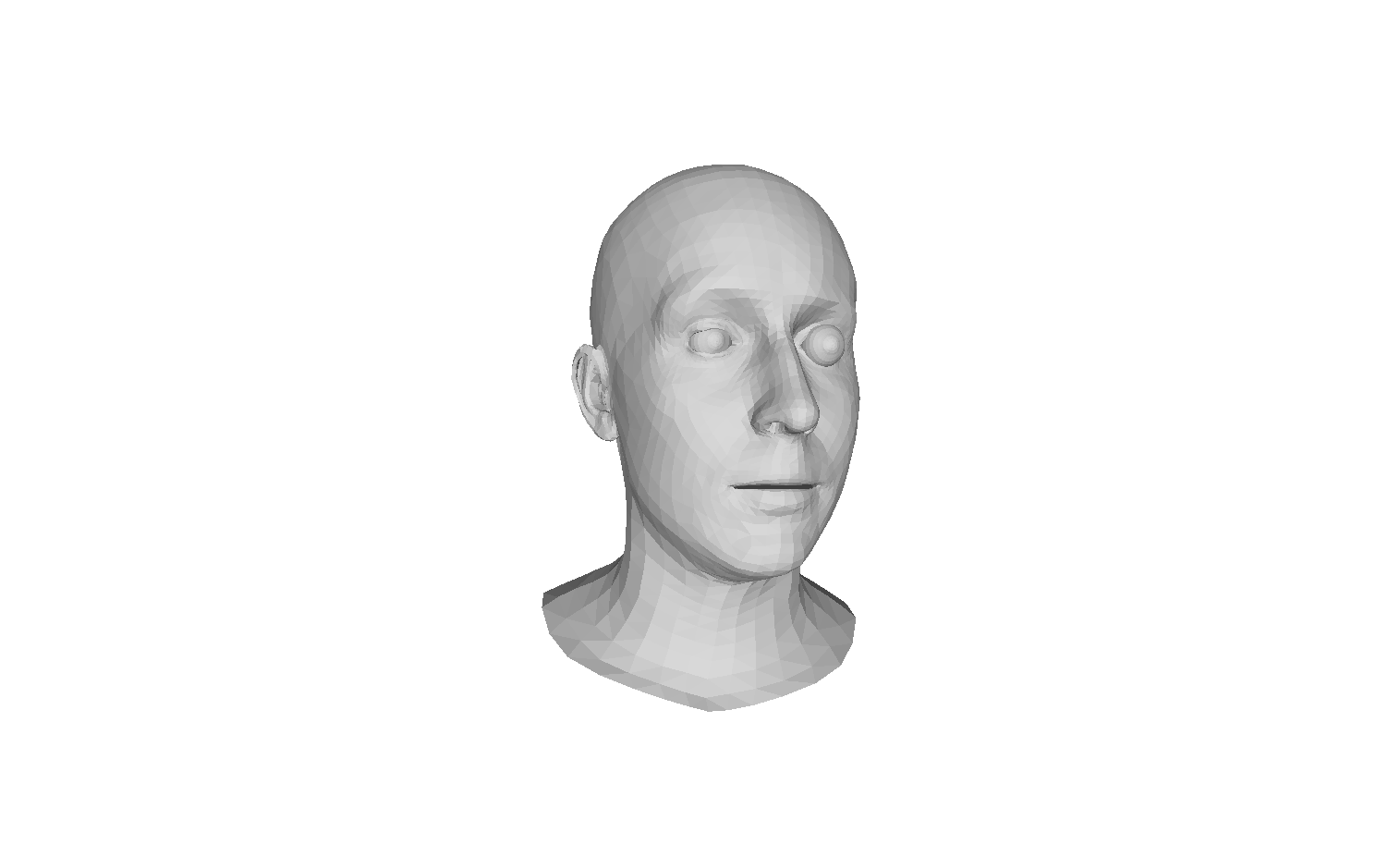}};
    \node[right of=j8, node distance=1.8cm] (j9) {\includegraphics[trim={400 80 400 100},clip,width=0.09\linewidth]{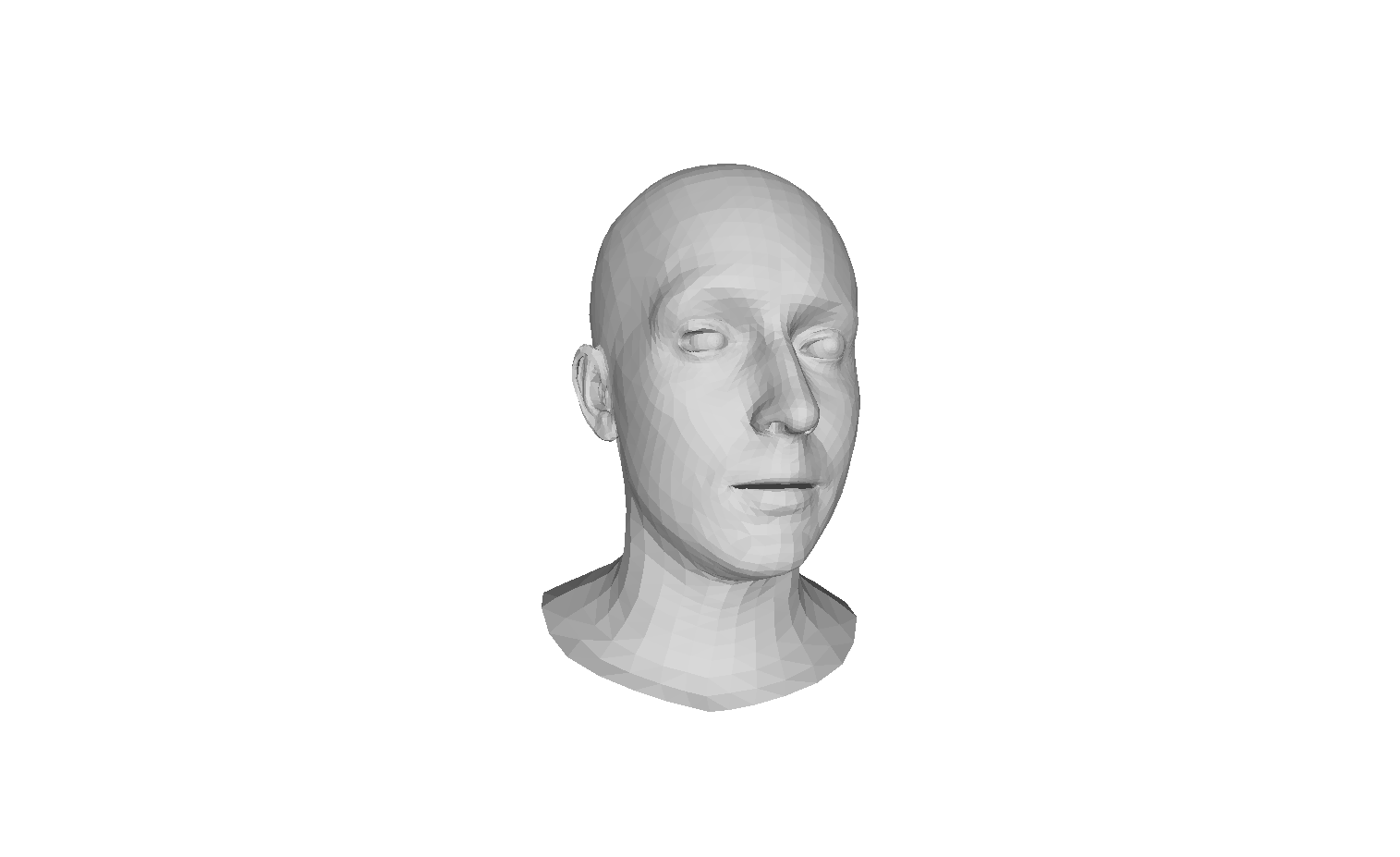}};
    
    \node[below of=j1, node distance=1.5cm] {Target Image};
    \coordinate (j56) at ($(j5)!0.5!(j6)$);
    \node[below of=j56, node distance=1.5cm] {Diverse 3D Reconstructions by \ourmethod{}};
    \end{tikzpicture}
    \caption{Qualitative evaluation of the diversity and robustness performance of \ourmethod{} to occlusions at different facial locations.}
    \label{fig:moveocc}
\end{figure*}

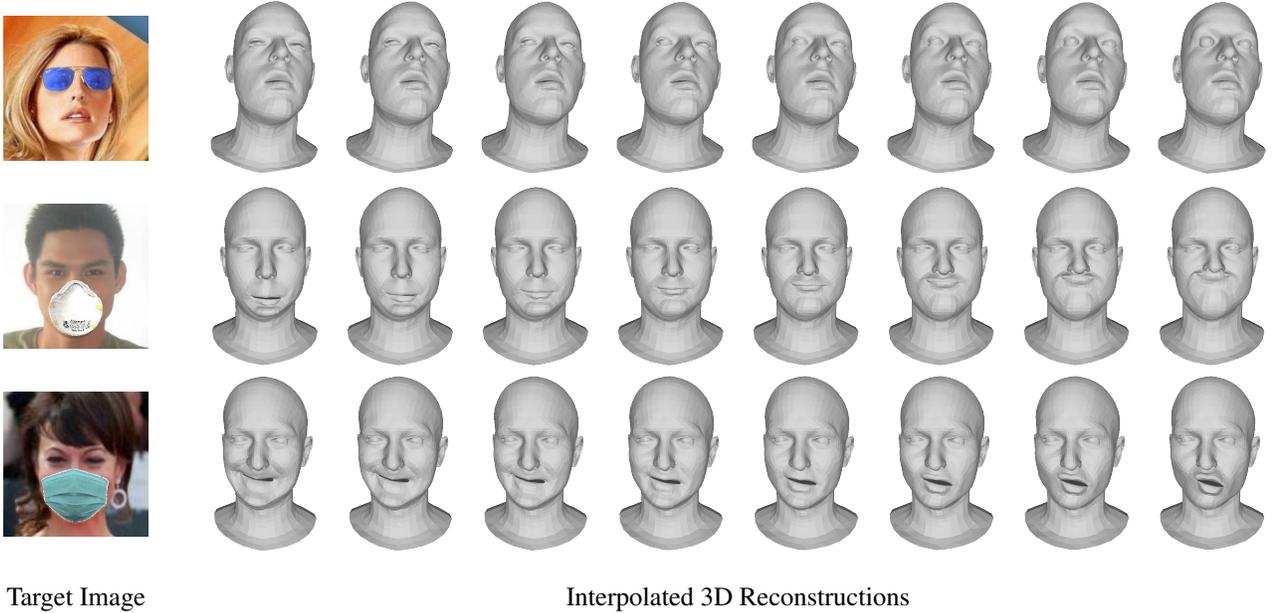
\begin{figure*}
    \centering
    \begin{tikzpicture}
    \node (a1) {\includegraphics[width=0.11\linewidth]{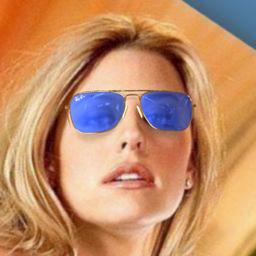}};
    \node[right of=a1, node distance=2.5cm] (a2) {\includegraphics[trim={400 80 400 100},clip,width=0.1\linewidth]{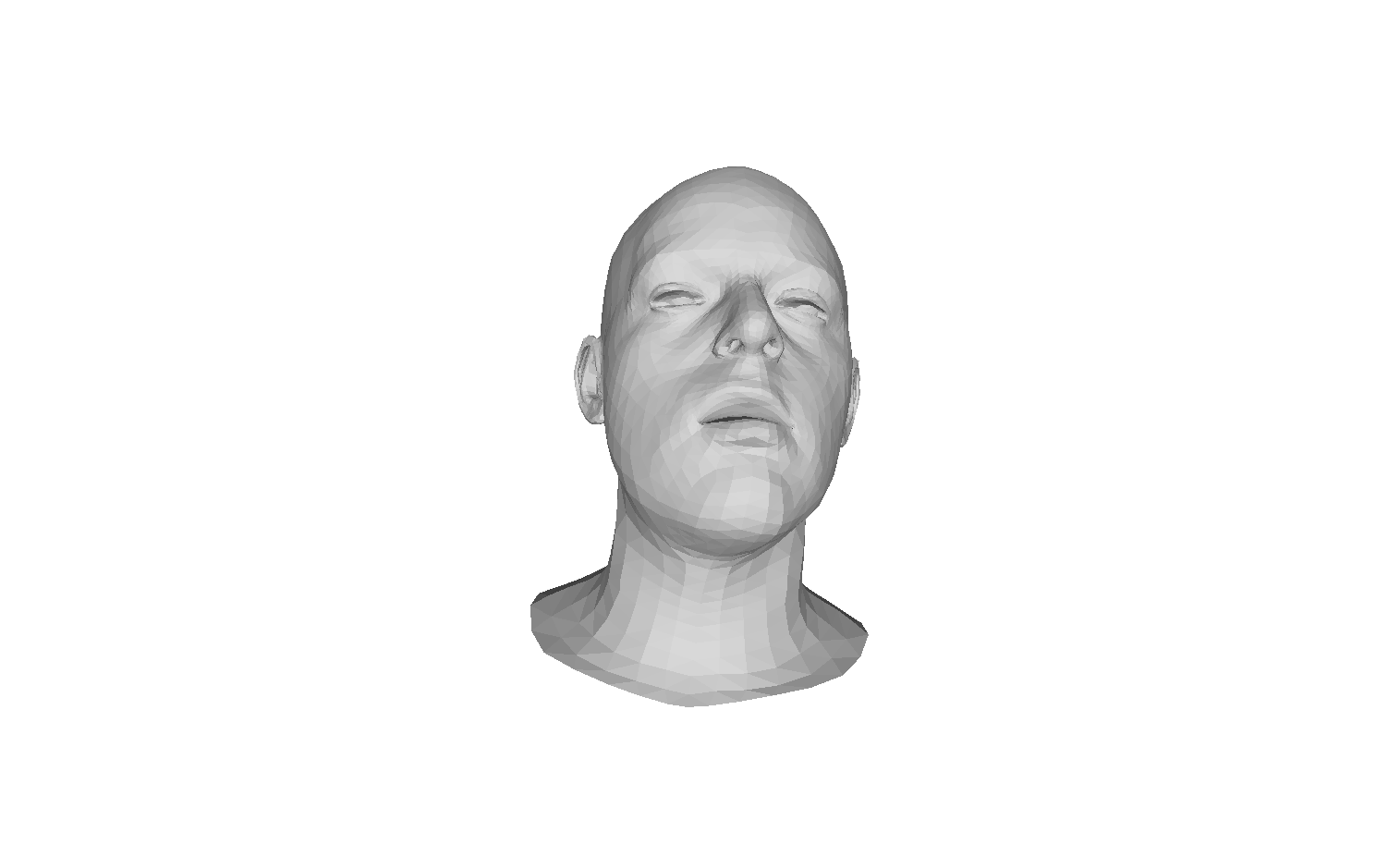}};
    \node[right of=a2, node distance=1.8cm] (a3) {\includegraphics[trim={400 80 400 100},clip,width=0.1\linewidth]{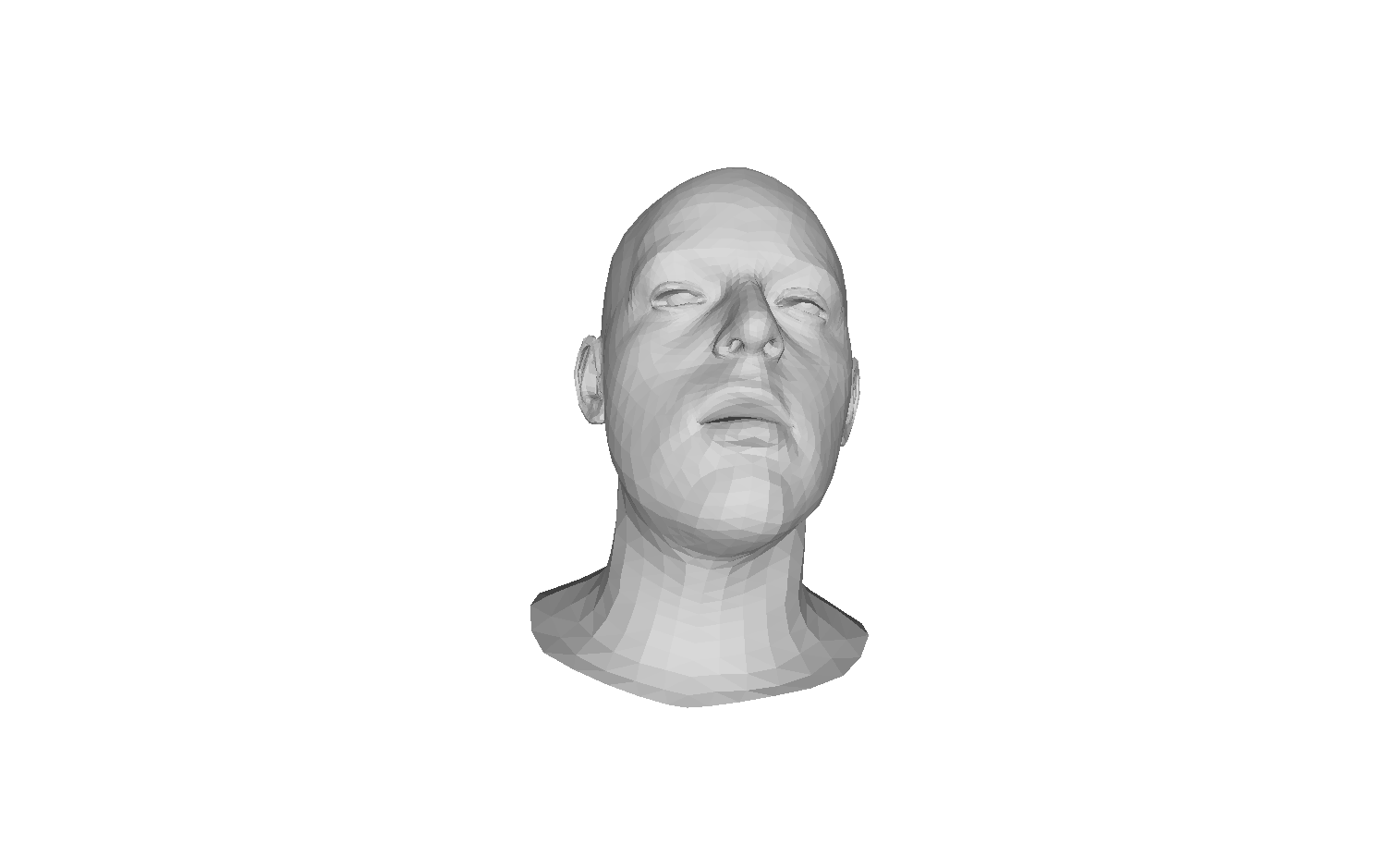}};
    \node[right of=a3, node distance=1.8cm] (a4) {\includegraphics[trim={400 80 400 100},clip,width=0.1\linewidth]{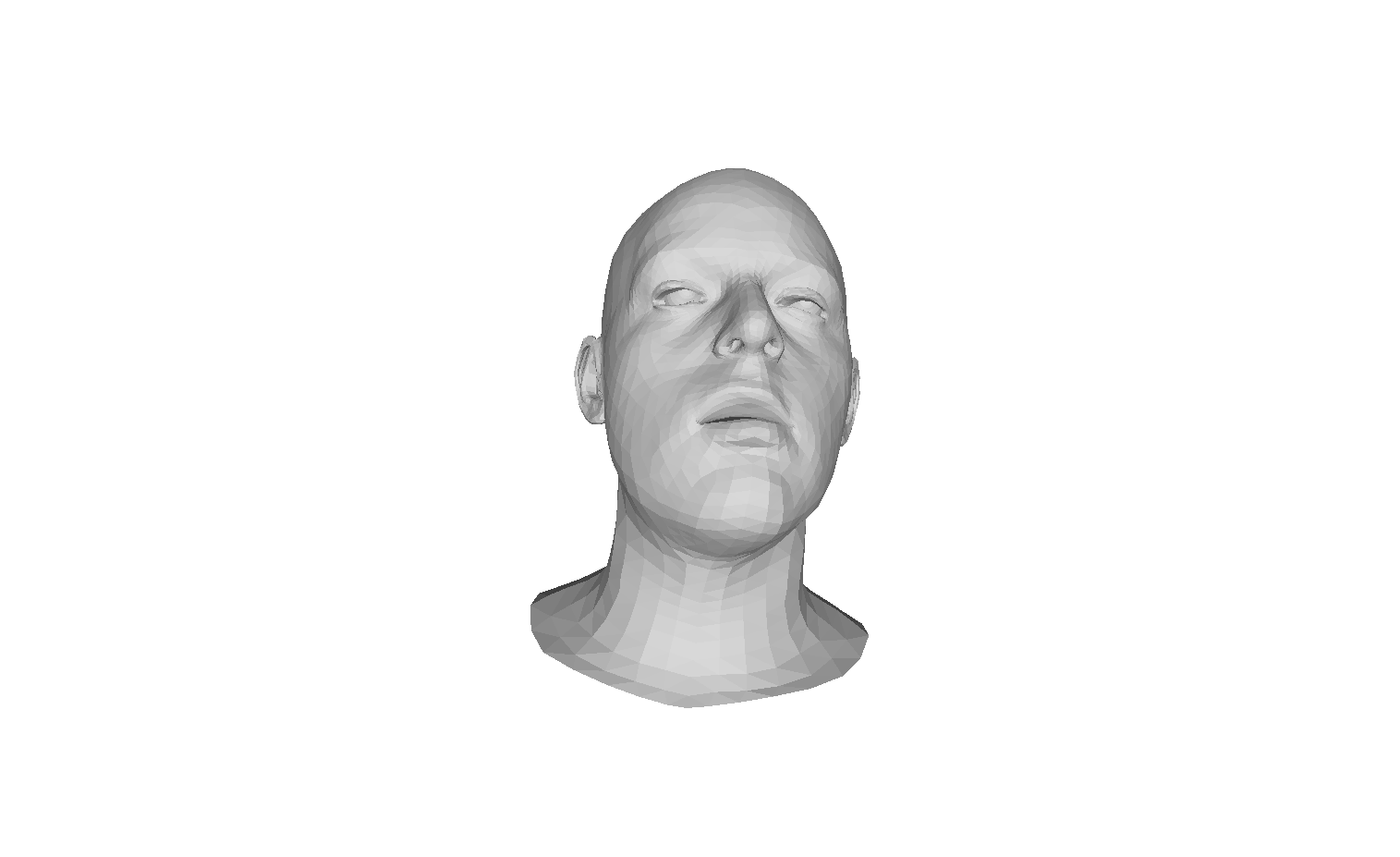}};
    \node[right of=a4, node distance=1.8cm] (a5) {\includegraphics[trim={400 80 400 100},clip,width=0.1\linewidth]{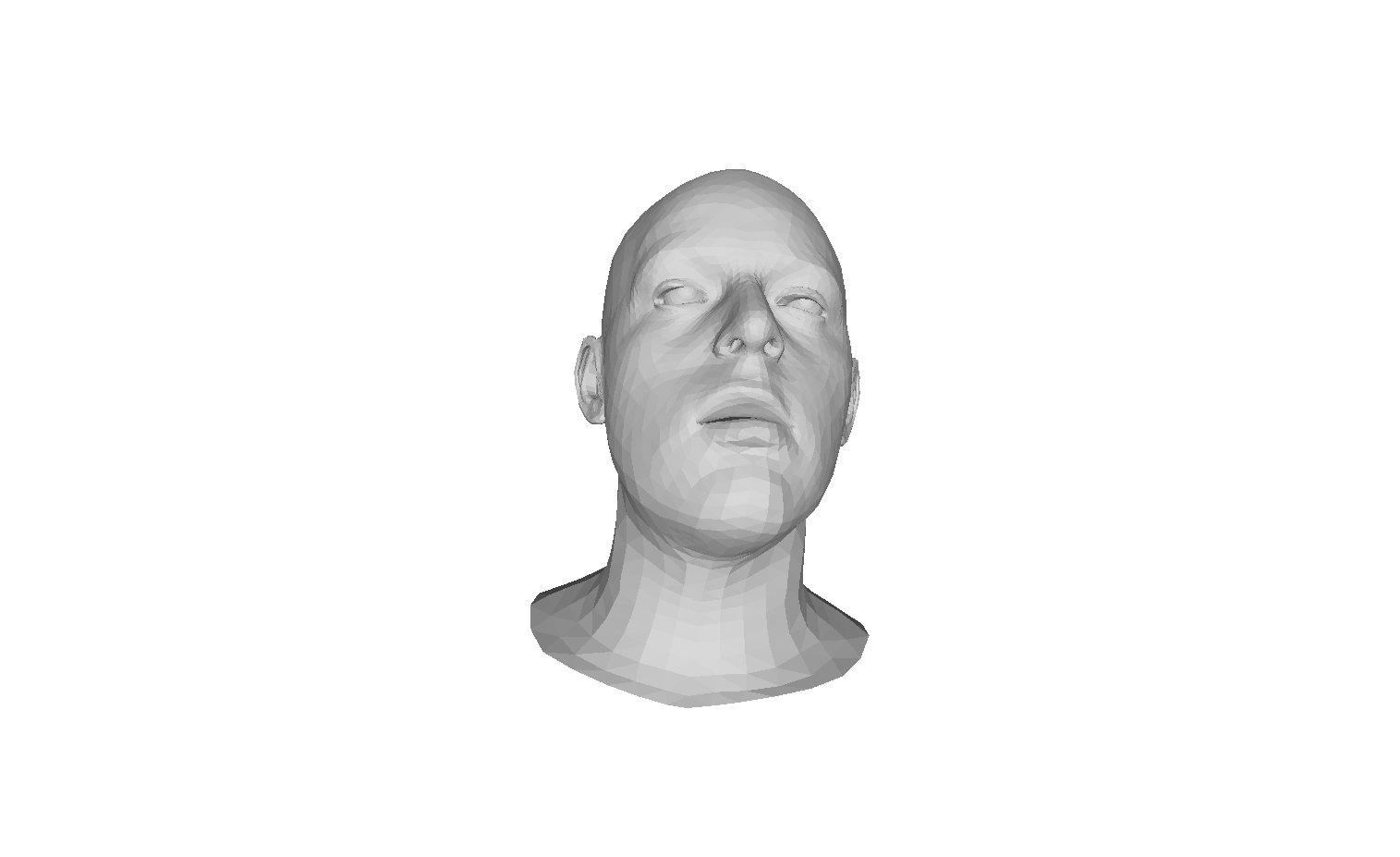}};
    \node[right of=a5, node distance=1.8cm] (a6) {\includegraphics[trim={400 80 400 100},clip,width=0.1\linewidth]{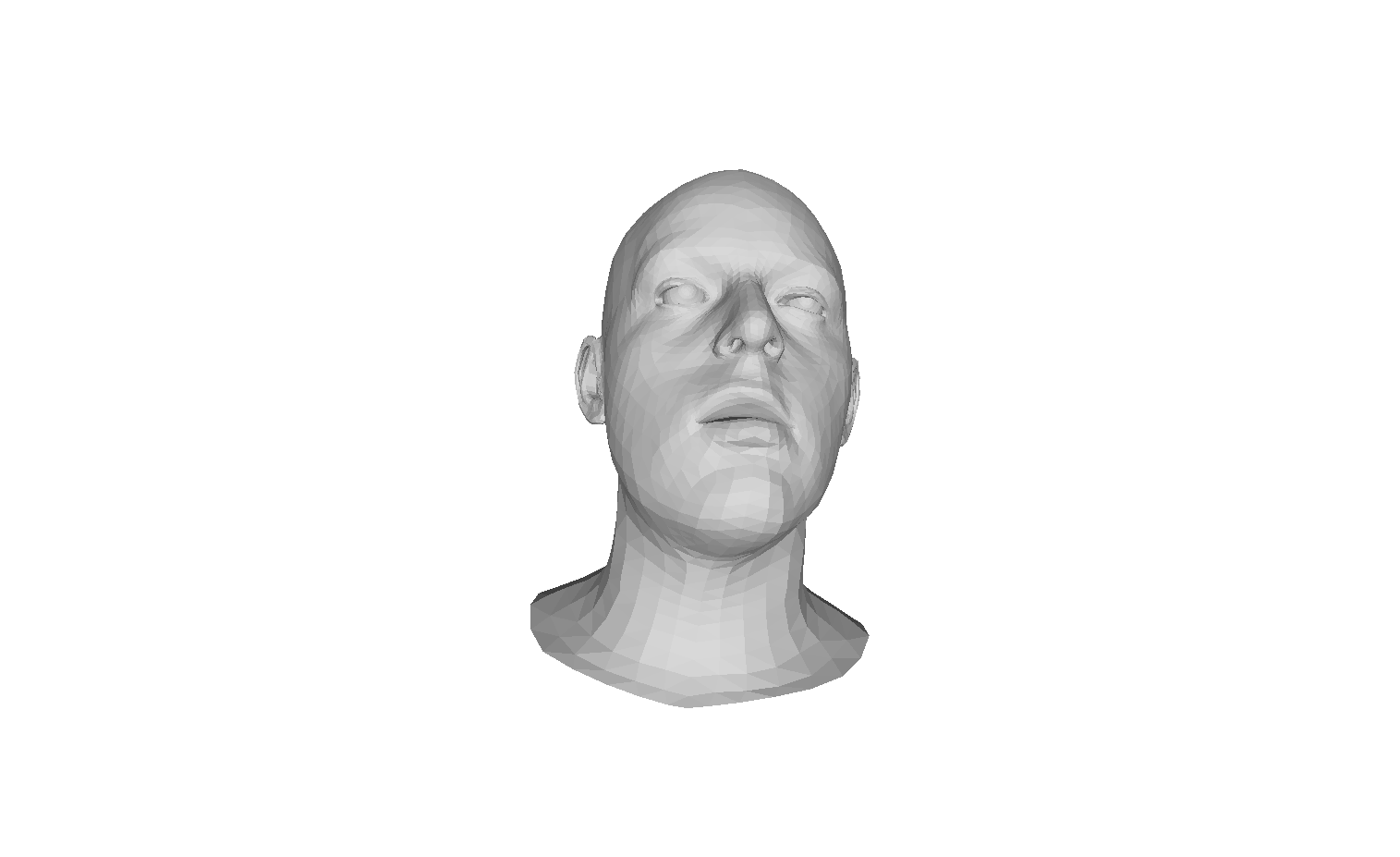}};
    \node[right of=a6, node distance=1.8cm] (a7) {\includegraphics[trim={400 80 400 100},clip,width=0.1\linewidth]{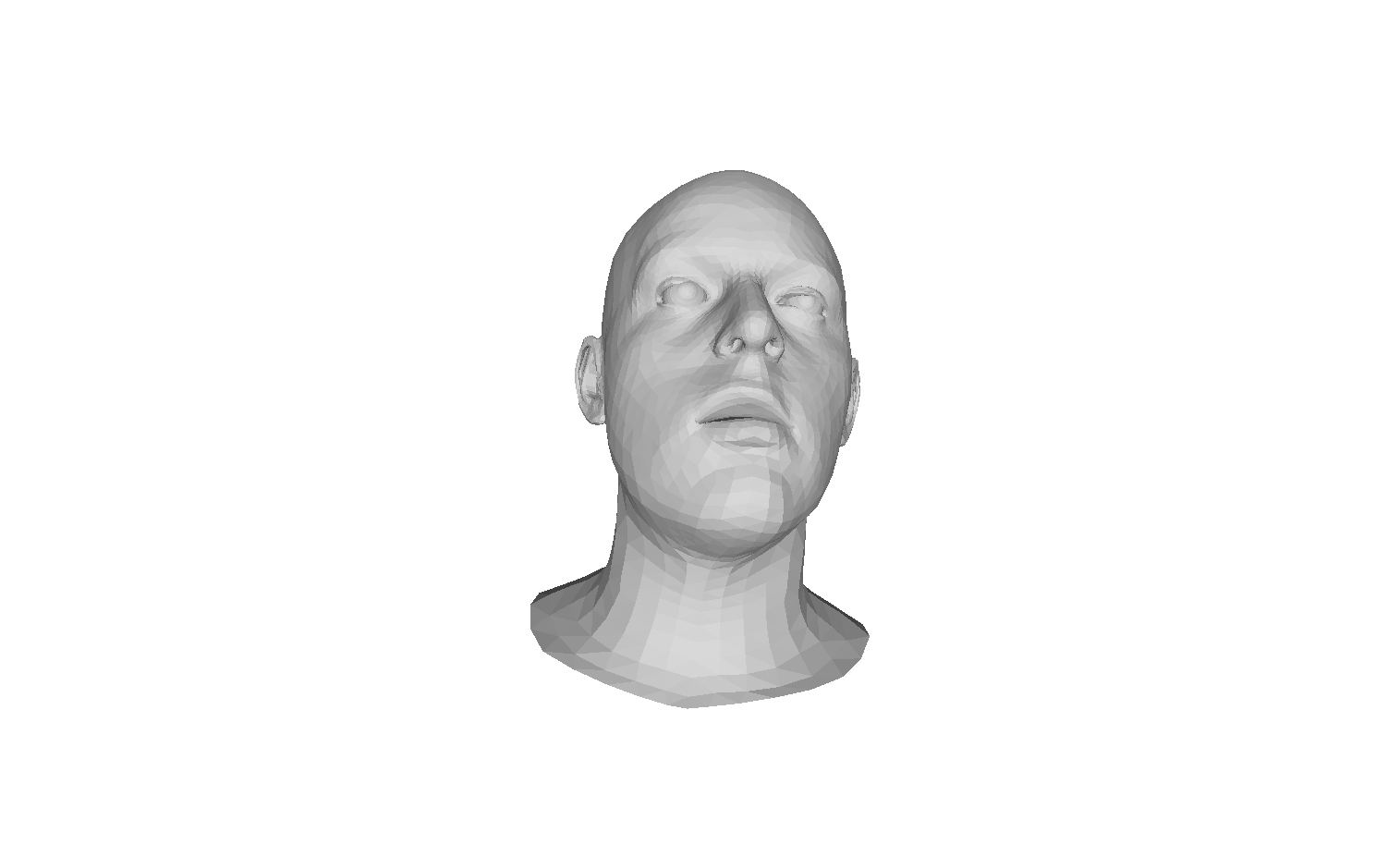}};
    \node[right of=a7, node distance=1.8cm] (a8) {\includegraphics[trim={400 80 400 100},clip,width=0.1\linewidth]{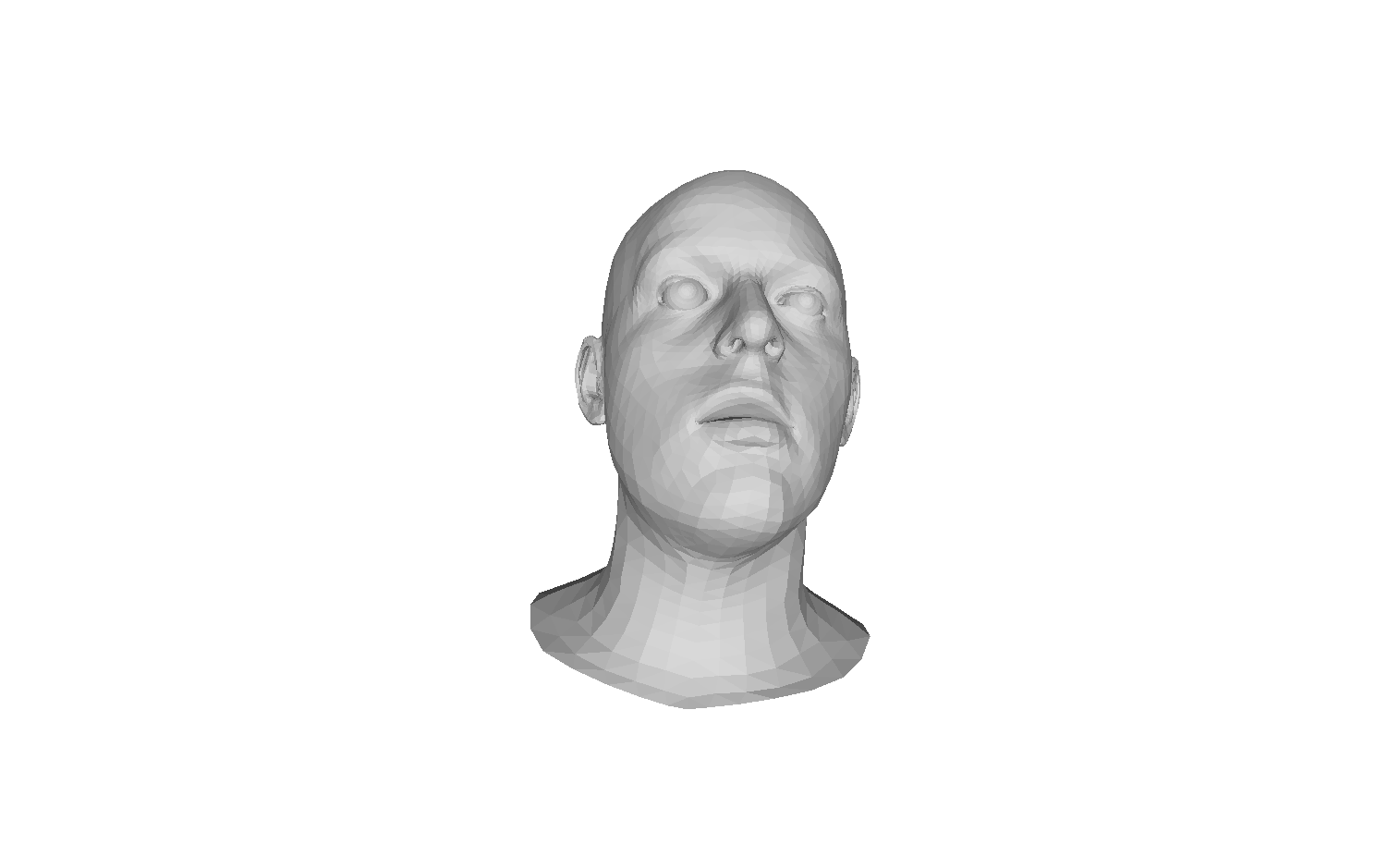}};
    \node[right of=a8, node distance=1.8cm] (a9) {\includegraphics[trim={400 80 400 100},clip,width=0.1\linewidth]{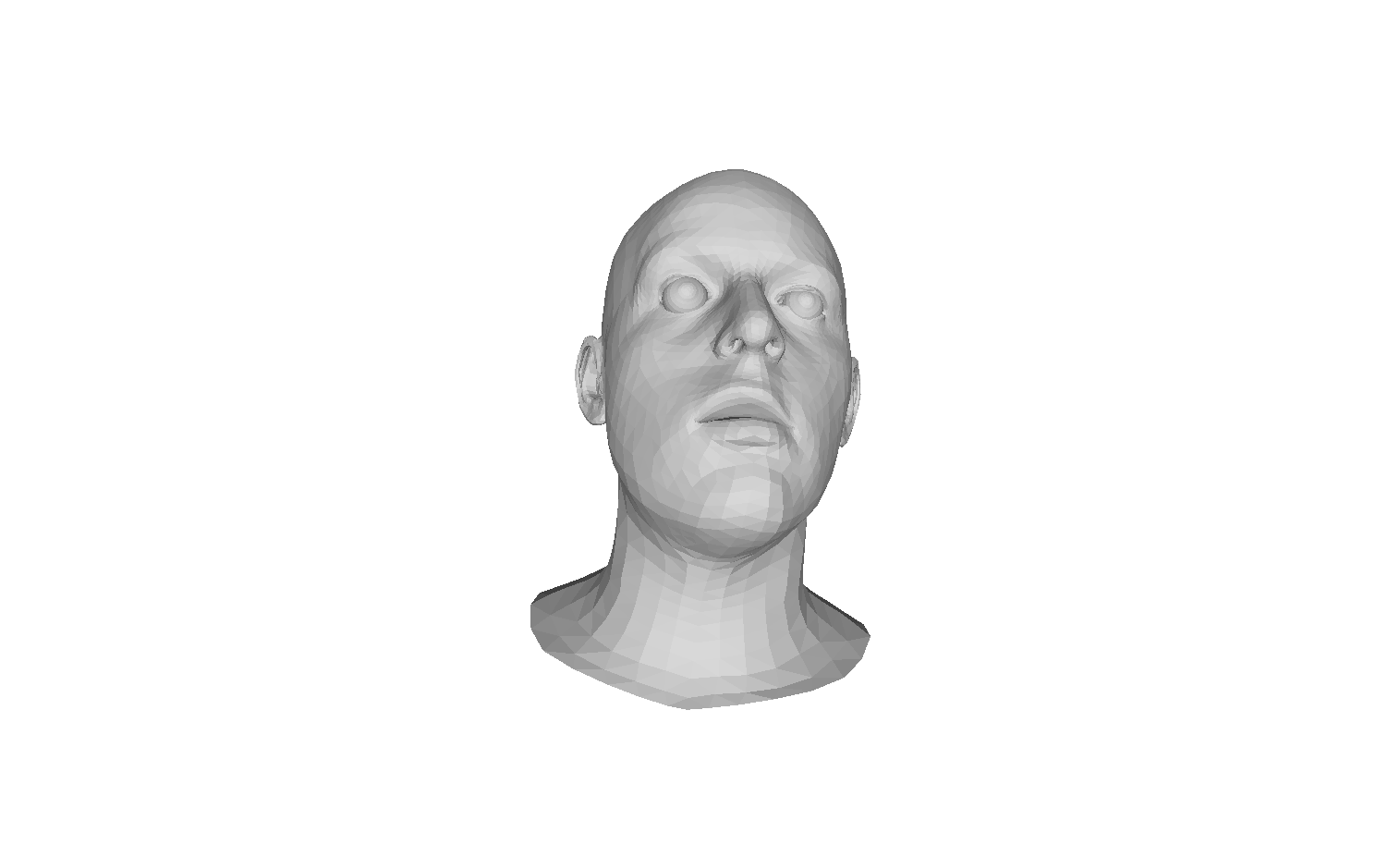}};
    
    \node[below of=a1, node distance=2.5cm] (b1) {\includegraphics[width=0.11\linewidth]{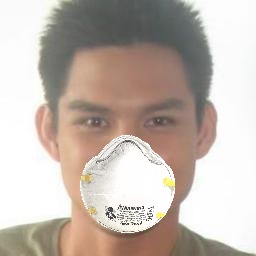}};
    \node[right of=b1, node distance=2.5cm] (b2) {\includegraphics[trim={400 80 400 100},clip,width=0.1\linewidth]{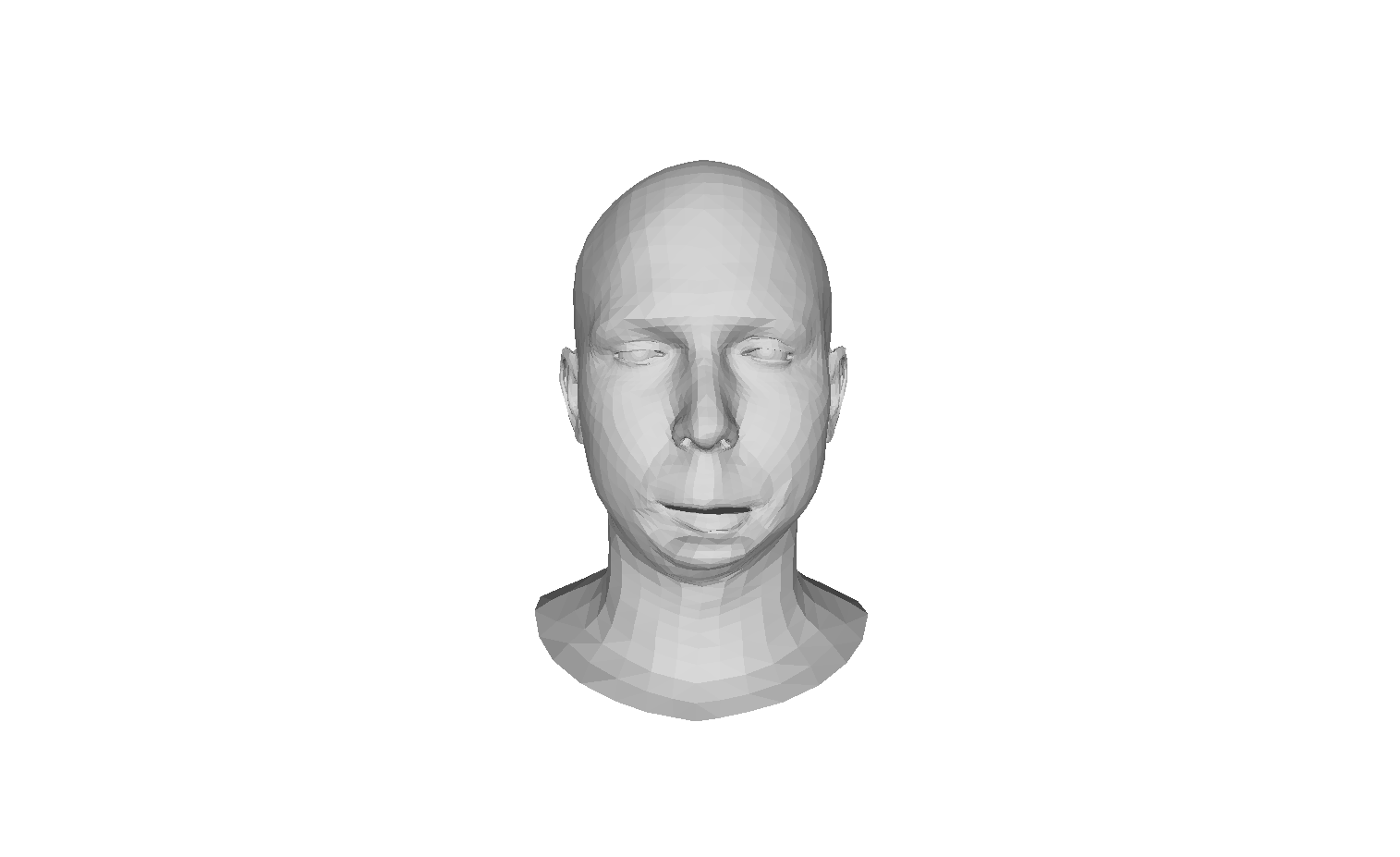}};
    \node[right of=b2, node distance=1.8cm] (b3) {\includegraphics[trim={400 80 400 100},clip,width=0.1\linewidth]{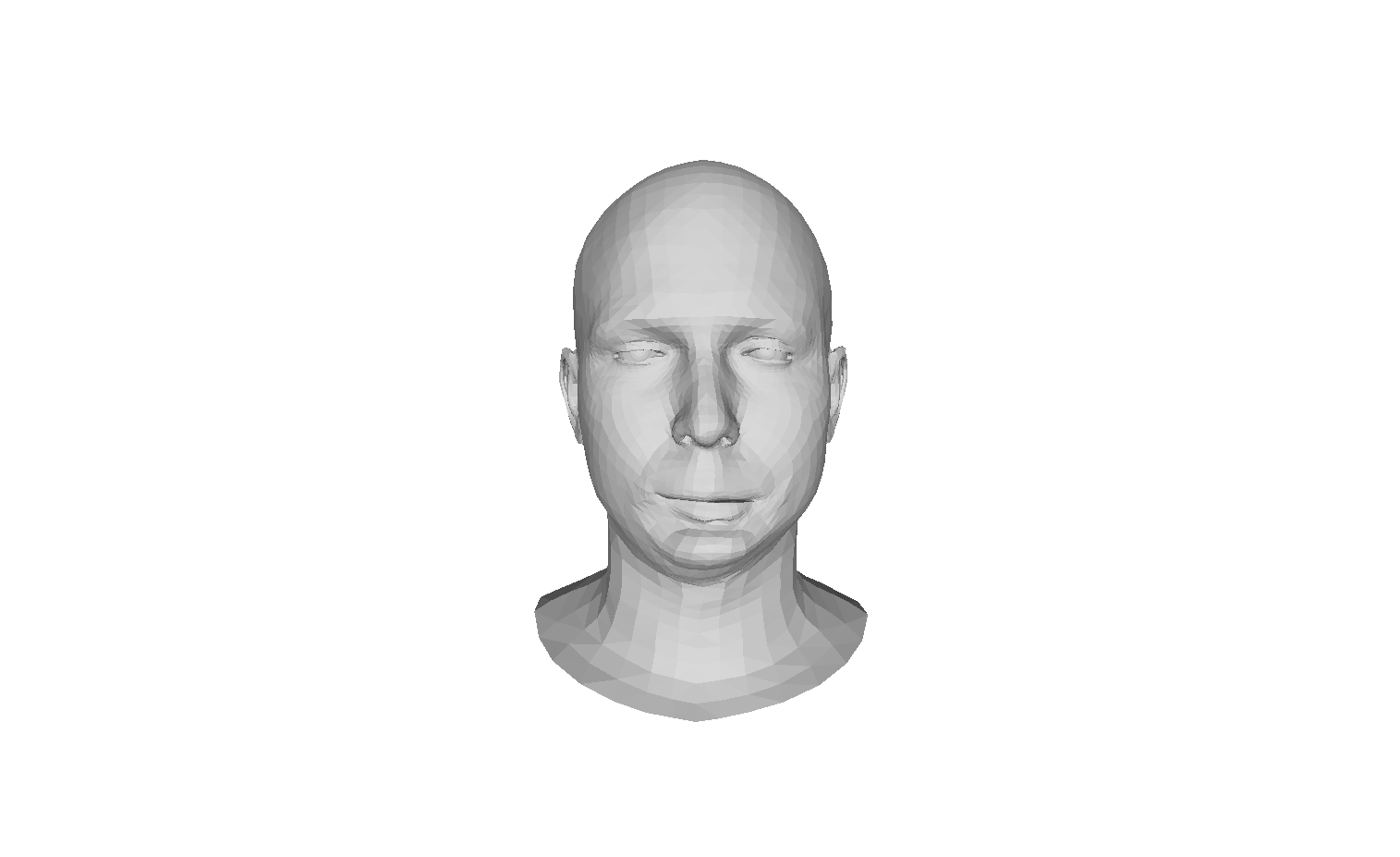}};
    \node[right of=b3, node distance=1.8cm] (b4) {\includegraphics[trim={400 80 400 100},clip,width=0.1\linewidth]{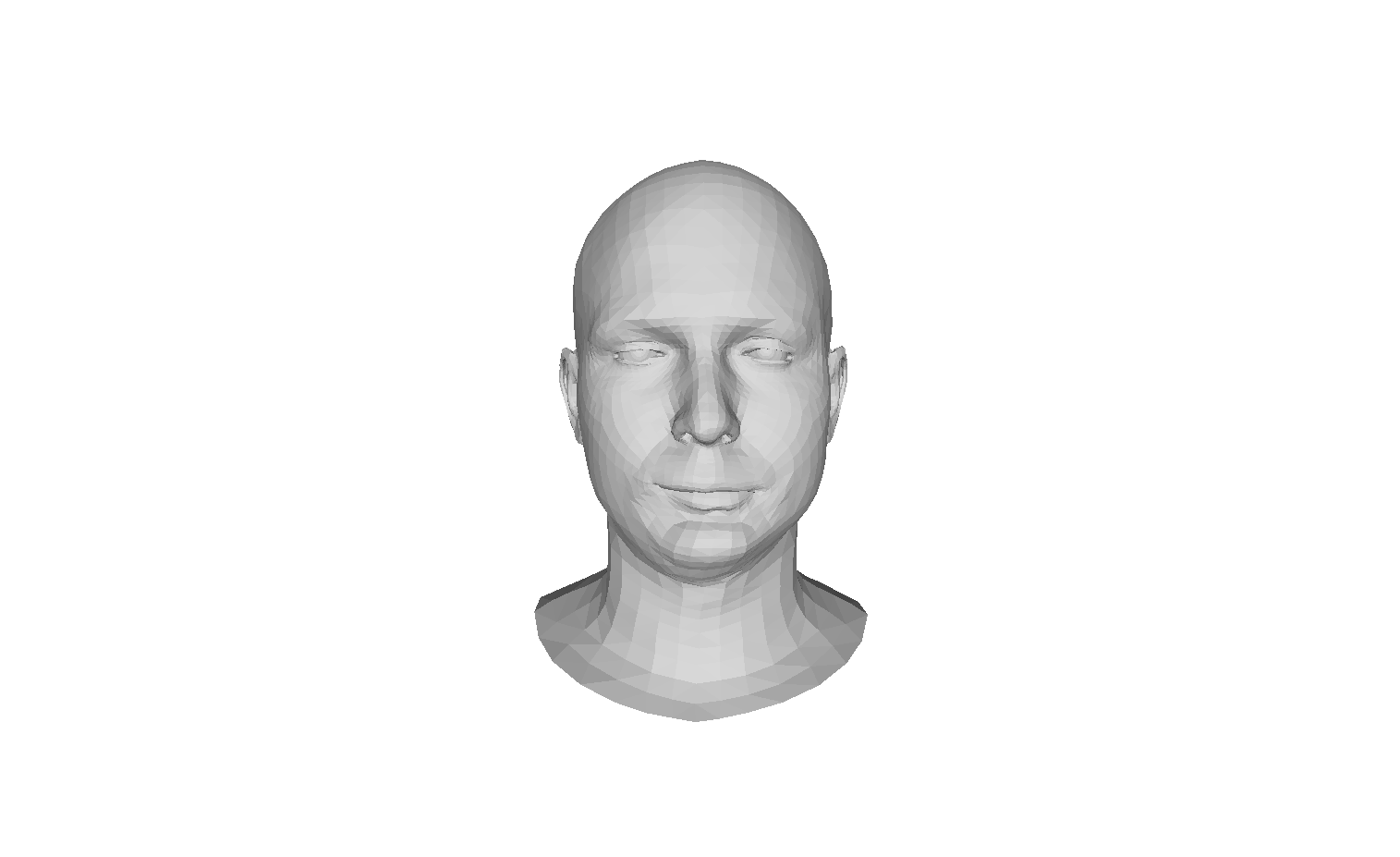}};
    \node[right of=b4, node distance=1.8cm] (b5) {\includegraphics[trim={400 80 400 100},clip,width=0.1\linewidth]{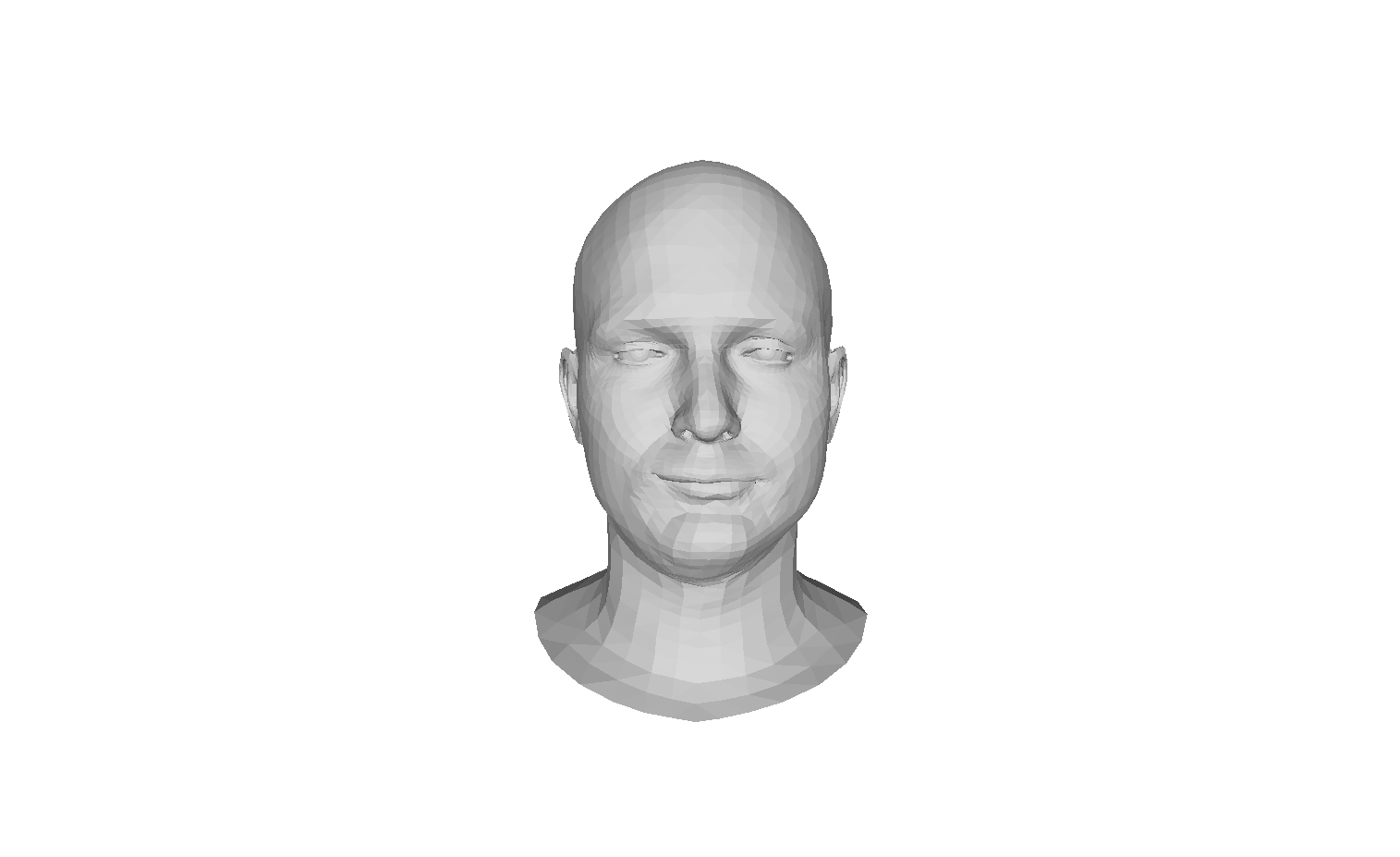}};
    \node[right of=b5, node distance=1.8cm] (b6) {\includegraphics[trim={400 80 400 100},clip,width=0.1\linewidth]{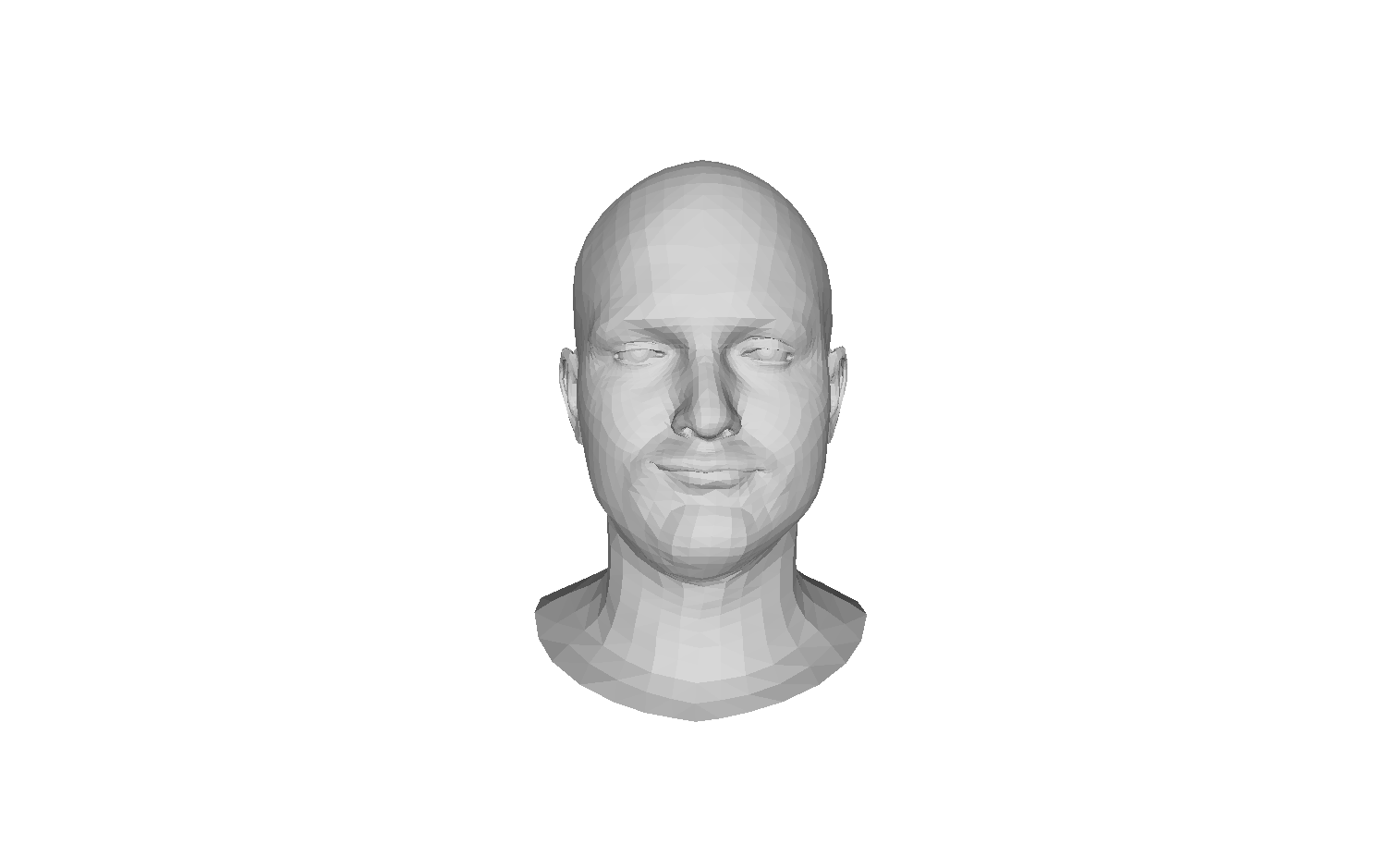}};
    \node[right of=b6, node distance=1.8cm] (b7) {\includegraphics[trim={400 80 400 100},clip,width=0.1\linewidth]{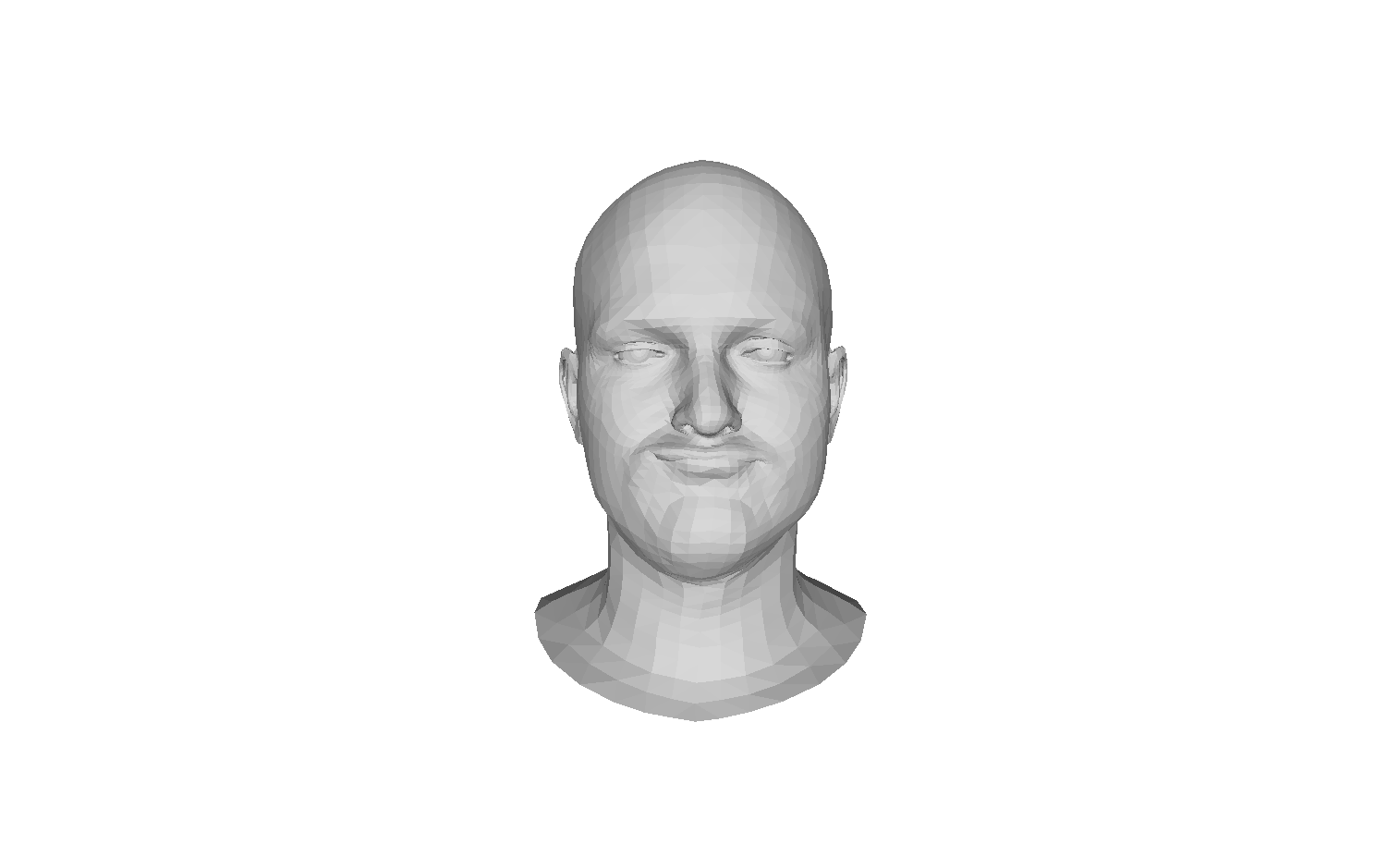}};
    \node[right of=b7, node distance=1.8cm] (b8) {\includegraphics[trim={400 80 400 100},clip,width=0.1\linewidth]{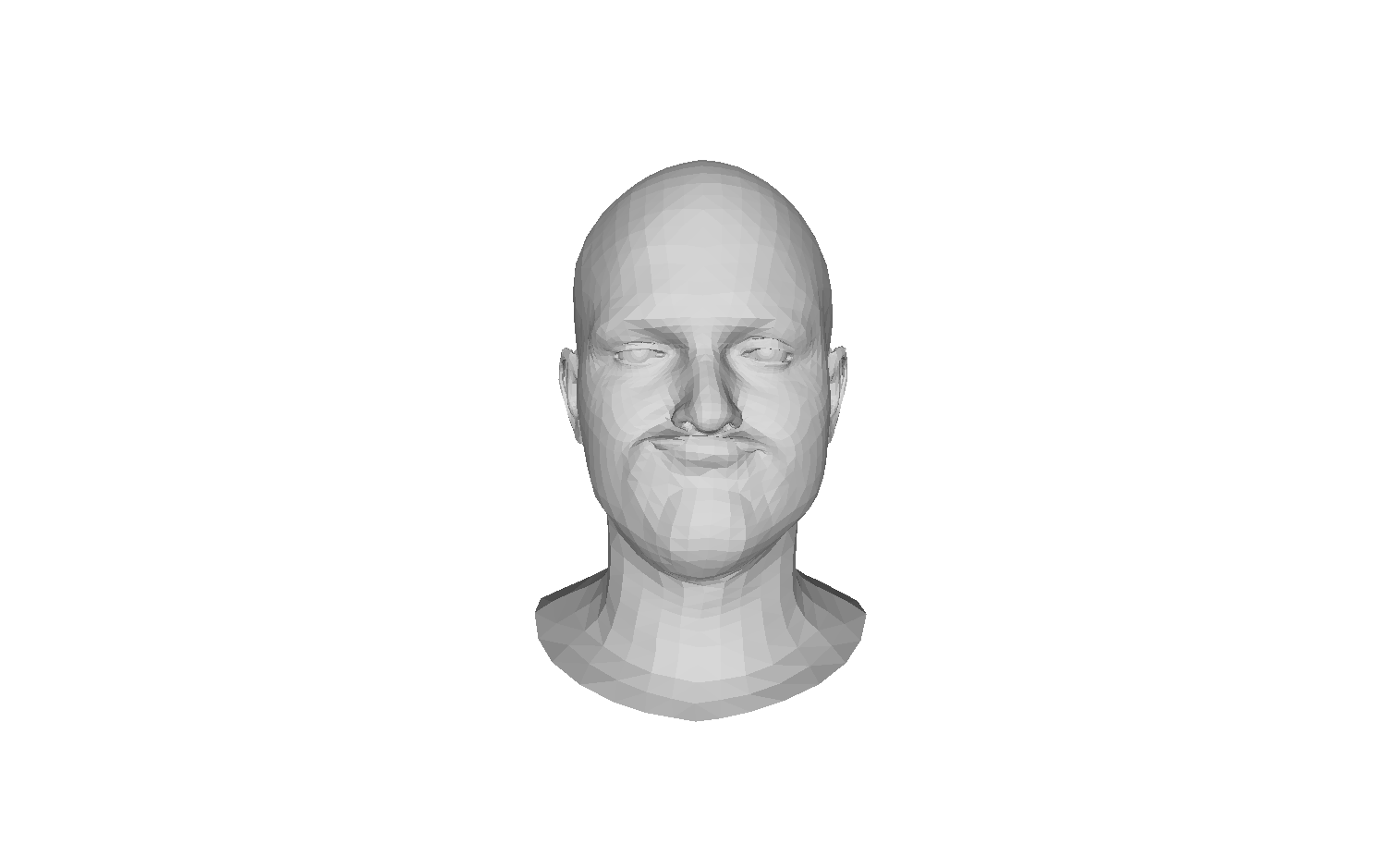}};
    \node[right of=b8, node distance=1.8cm] (b9) {\includegraphics[trim={400 80 400 100},clip,width=0.1\linewidth]{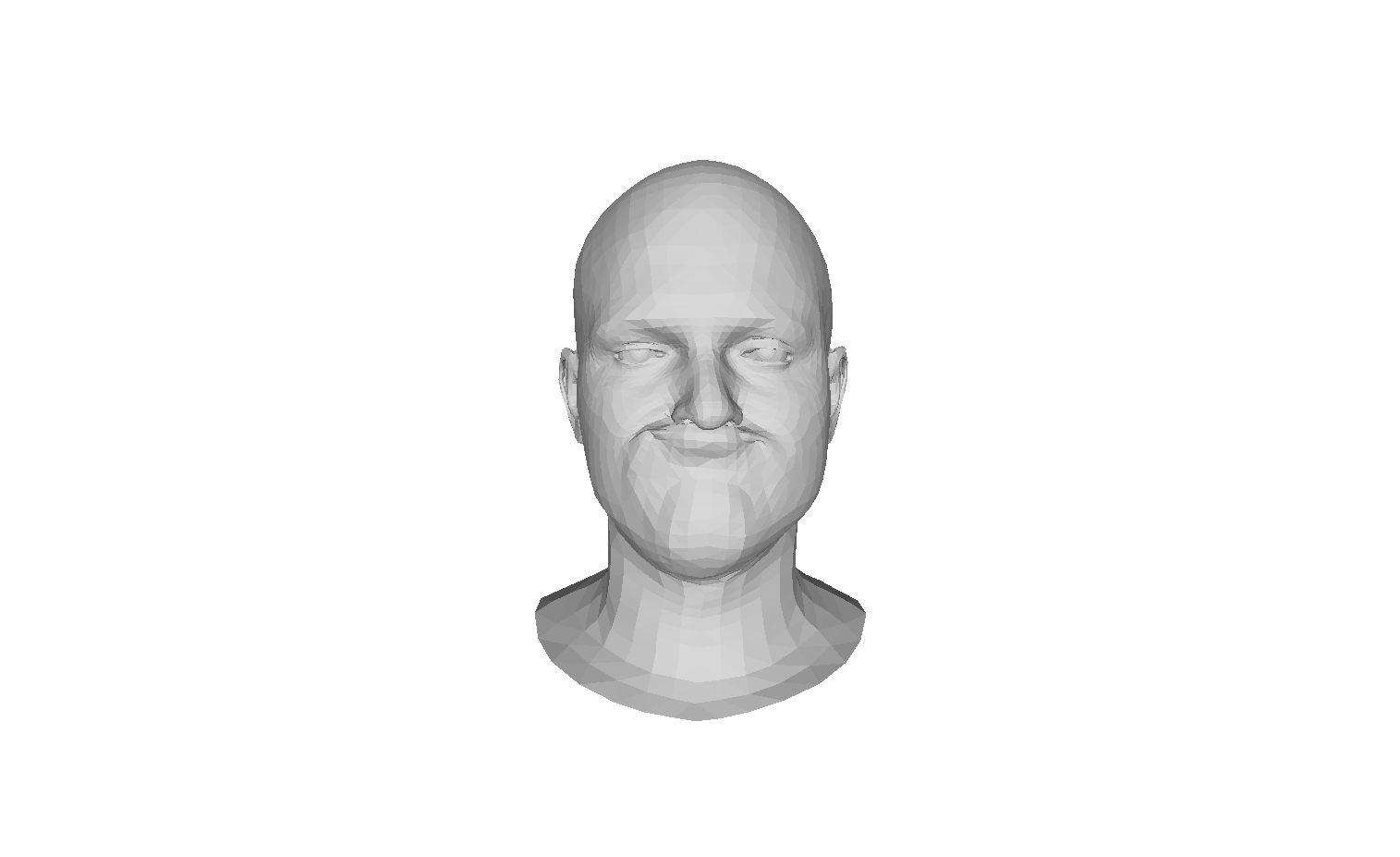}};
    
    \node[below of=b1, node distance=2.5cm] (c1) {\includegraphics[width=0.11\linewidth]{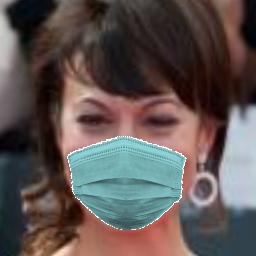}};
    \node[right of=c1, node distance=2.5cm] (c2) {\includegraphics[trim={400 80 400 100},clip,width=0.1\linewidth]{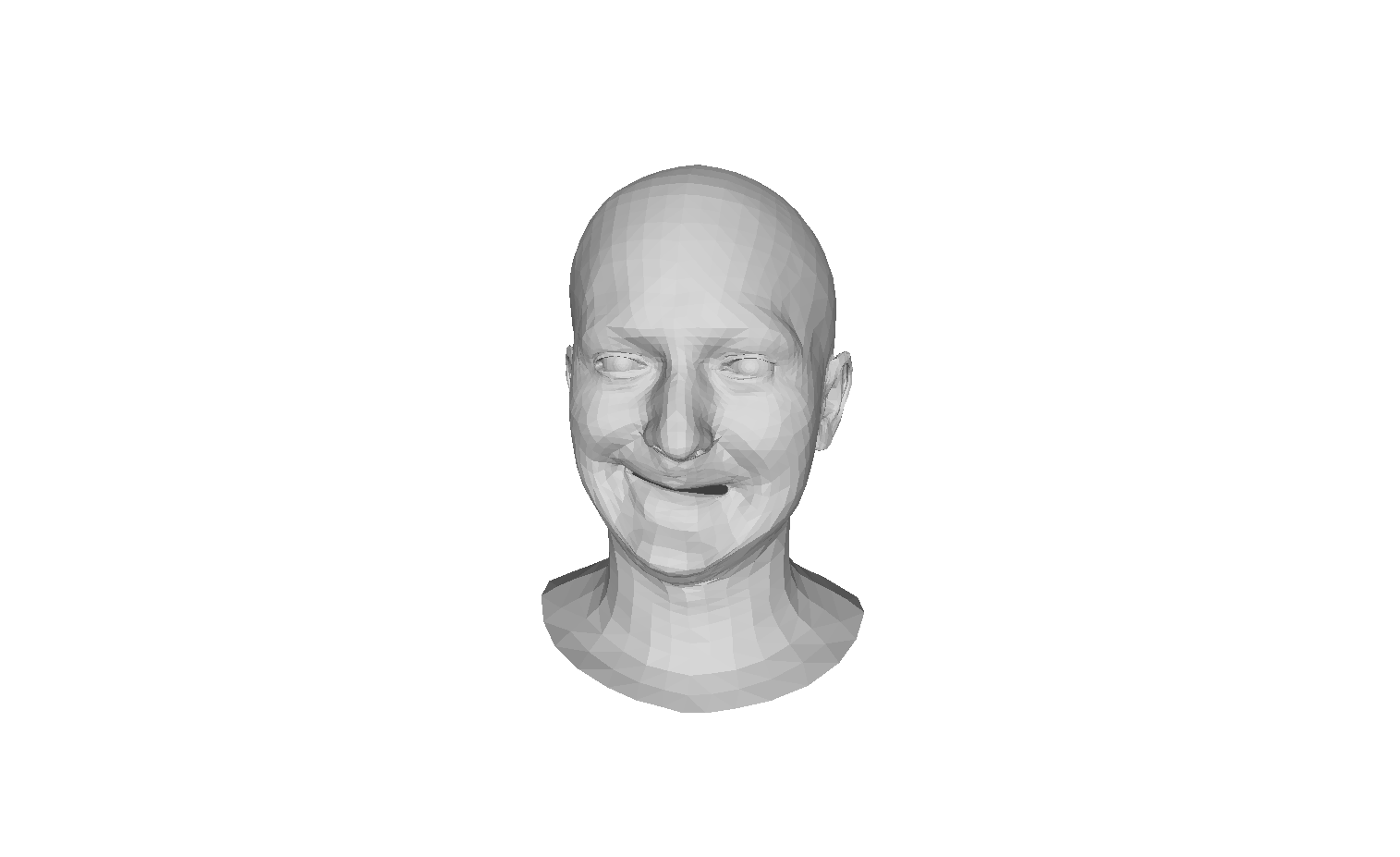}};
    \node[right of=c2, node distance=1.8cm] (c3) {\includegraphics[trim={400 80 400 100},clip,width=0.1\linewidth]{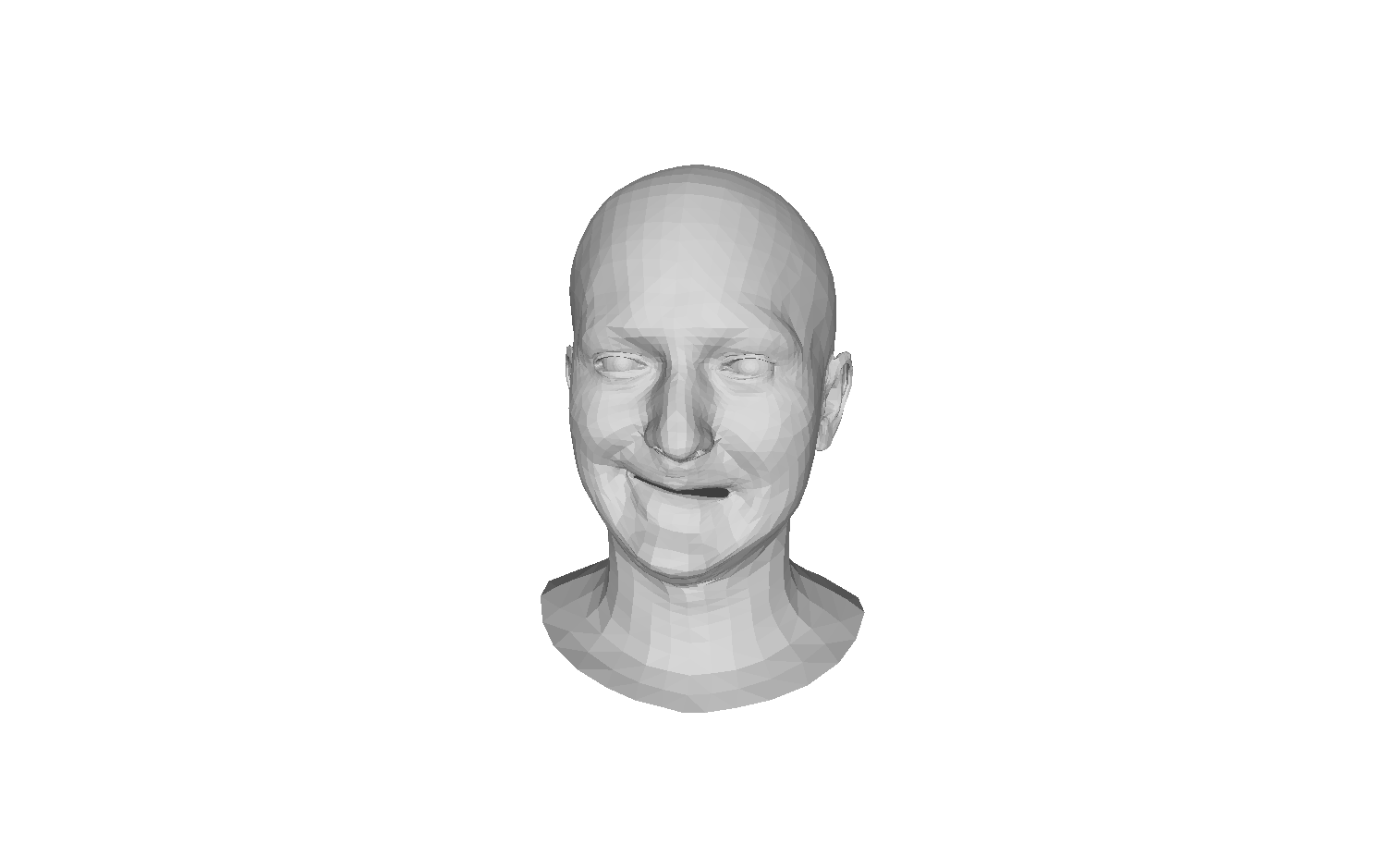}};
    \node[right of=c3, node distance=1.8cm] (c4) {\includegraphics[trim={400 80 400 100},clip,width=0.1\linewidth]{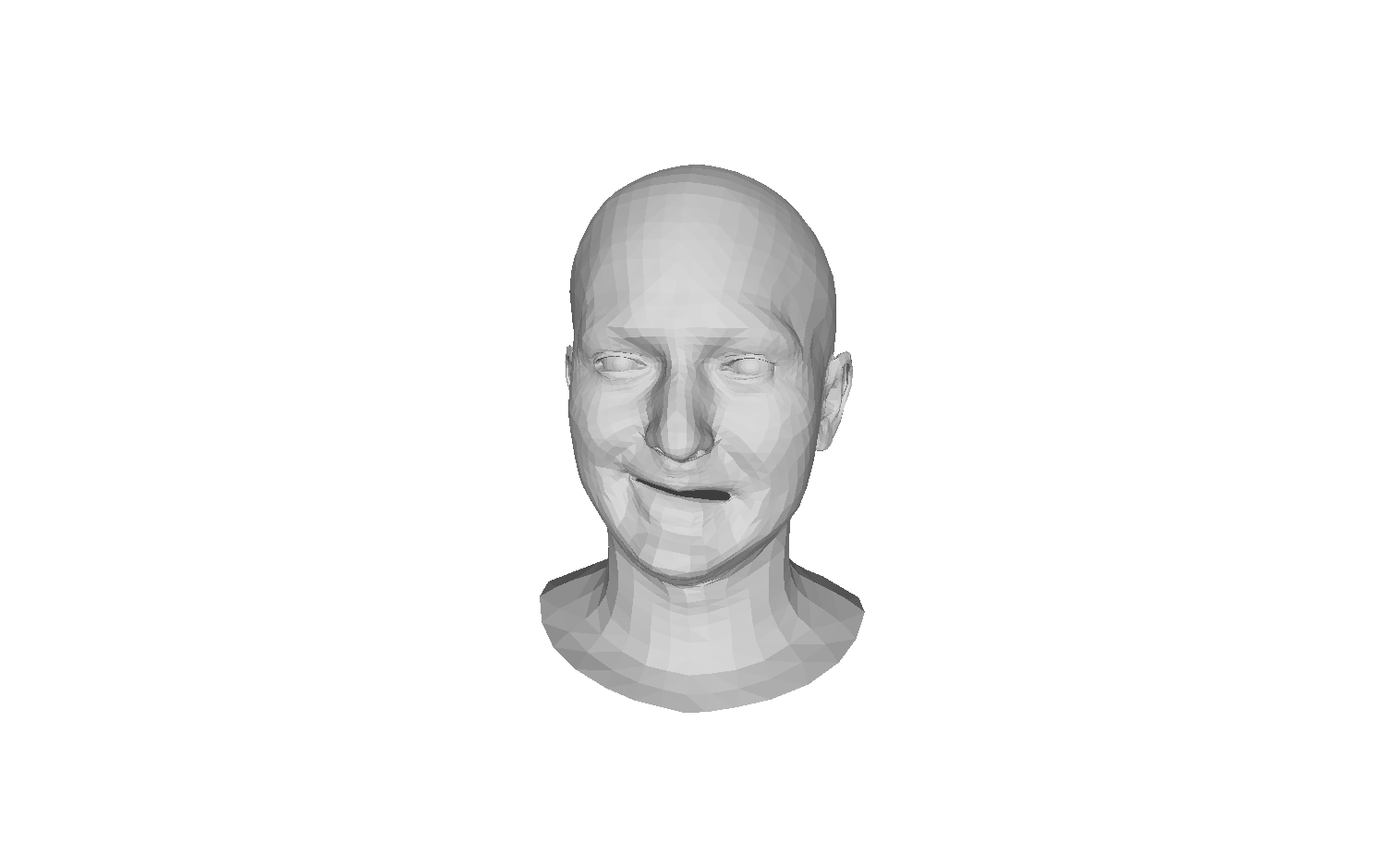}};
    \node[right of=c4, node distance=1.8cm] (c5) {\includegraphics[trim={400 80 400 100},clip,width=0.1\linewidth]{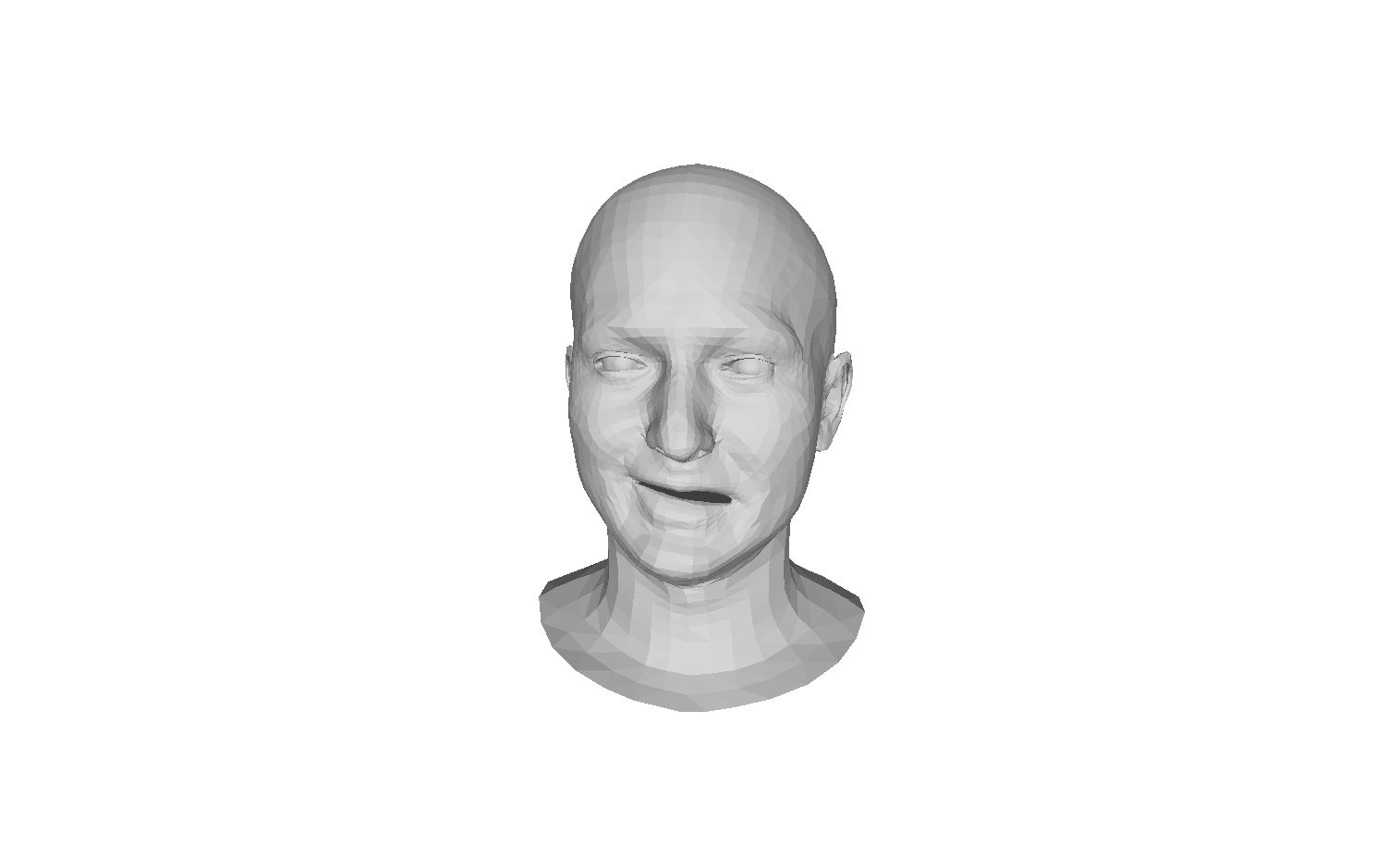}};
    \node[right of=c5, node distance=1.8cm] (c6) {\includegraphics[trim={400 80 400 100},clip,width=0.1\linewidth]{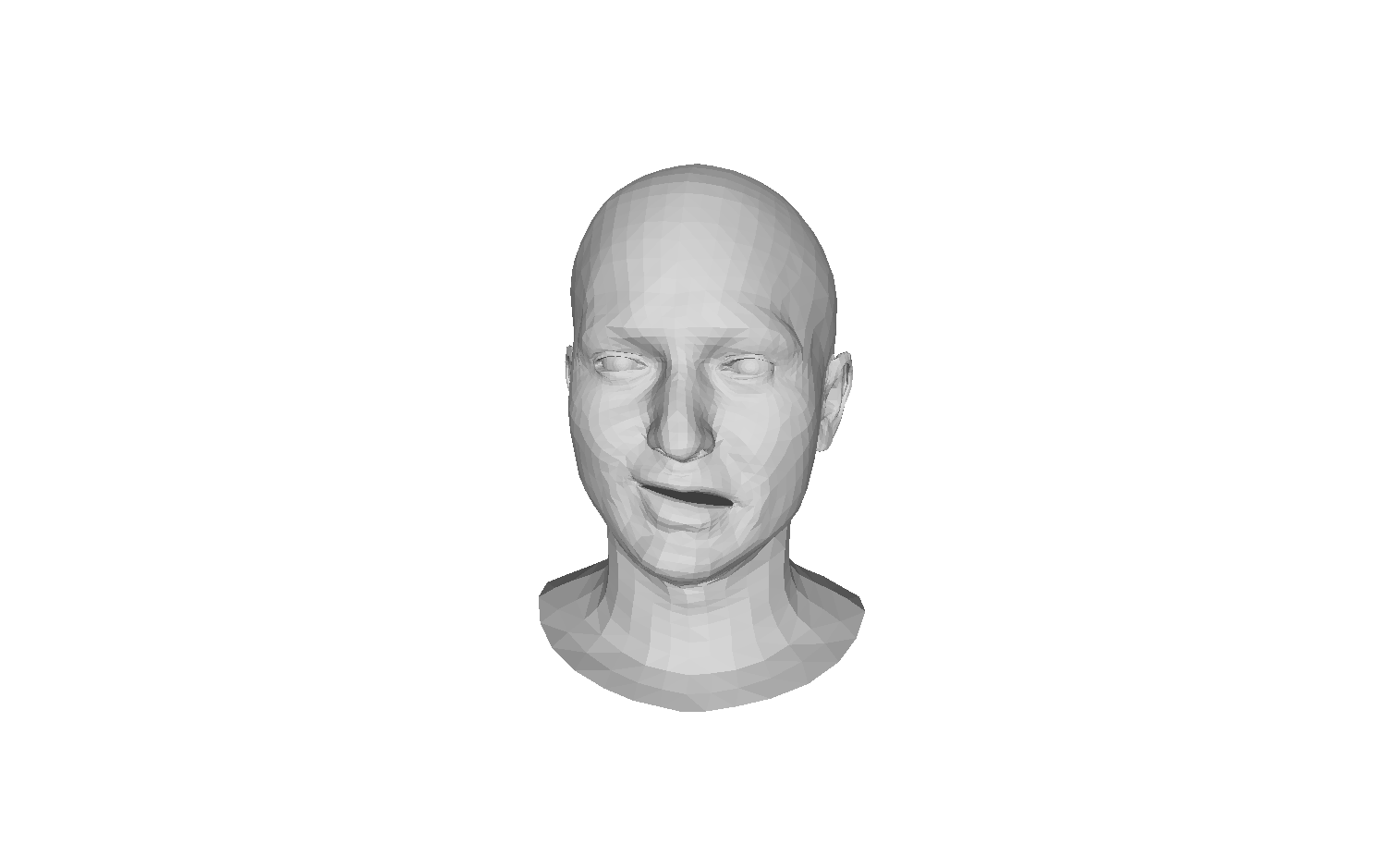}};
    \node[right of=c6, node distance=1.8cm] (c7) {\includegraphics[trim={400 80 400 100},clip,width=0.1\linewidth]{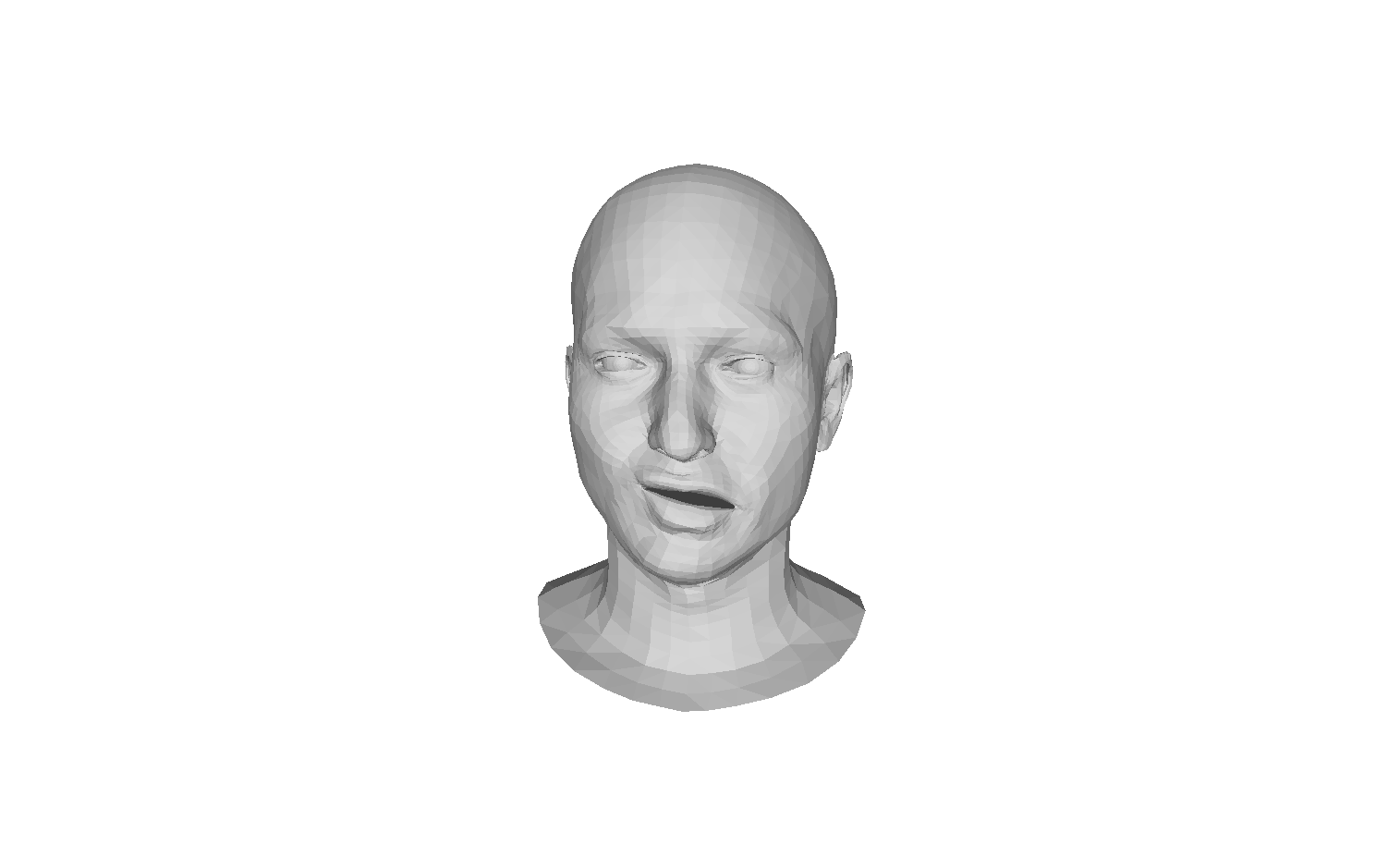}};
    \node[right of=c7, node distance=1.8cm] (c8) {\includegraphics[trim={400 80 400 100},clip,width=0.1\linewidth]{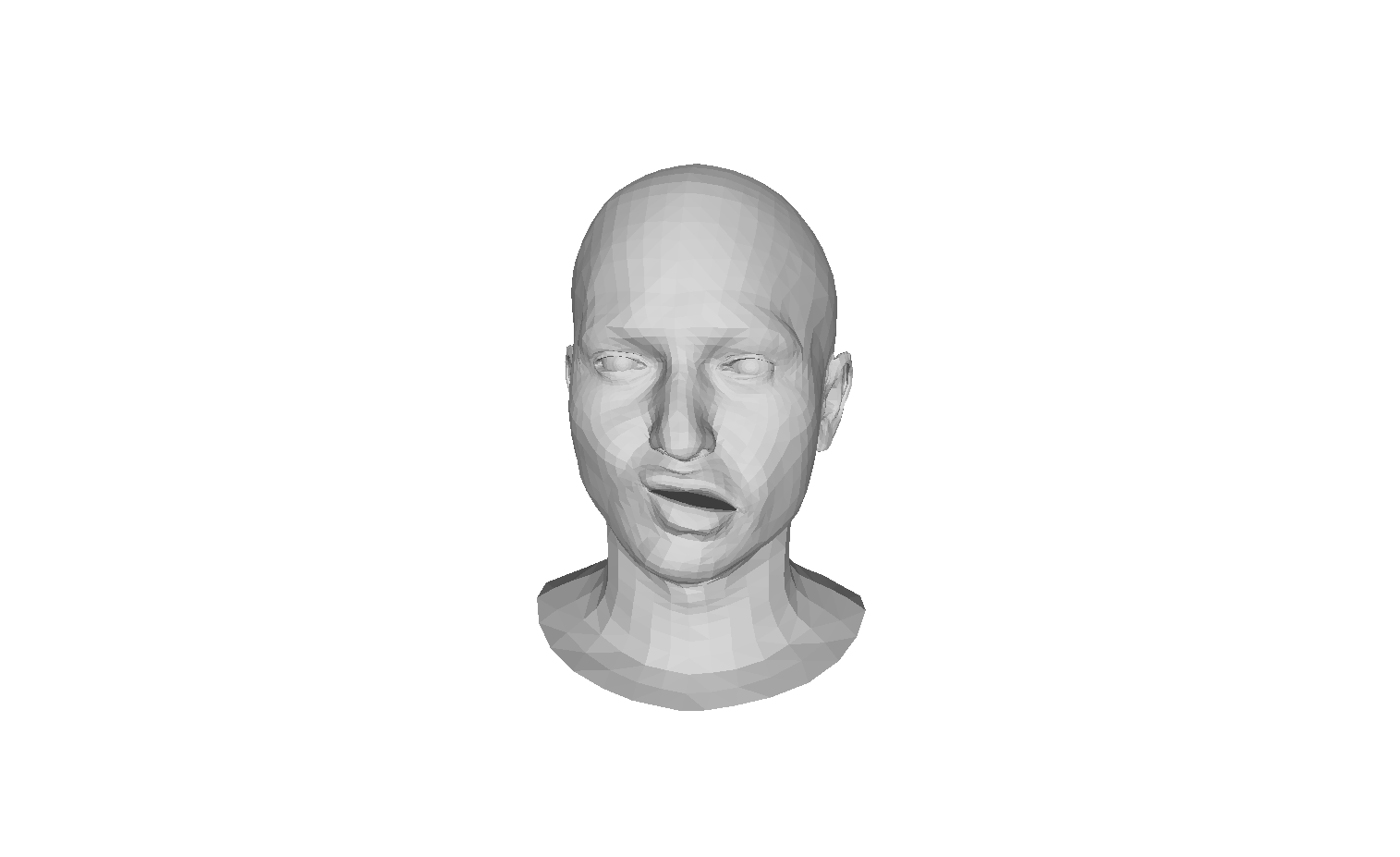}};
    \node[right of=c8, node distance=1.8cm] (c9) {\includegraphics[trim={400 80 400 100},clip,width=0.1\linewidth]{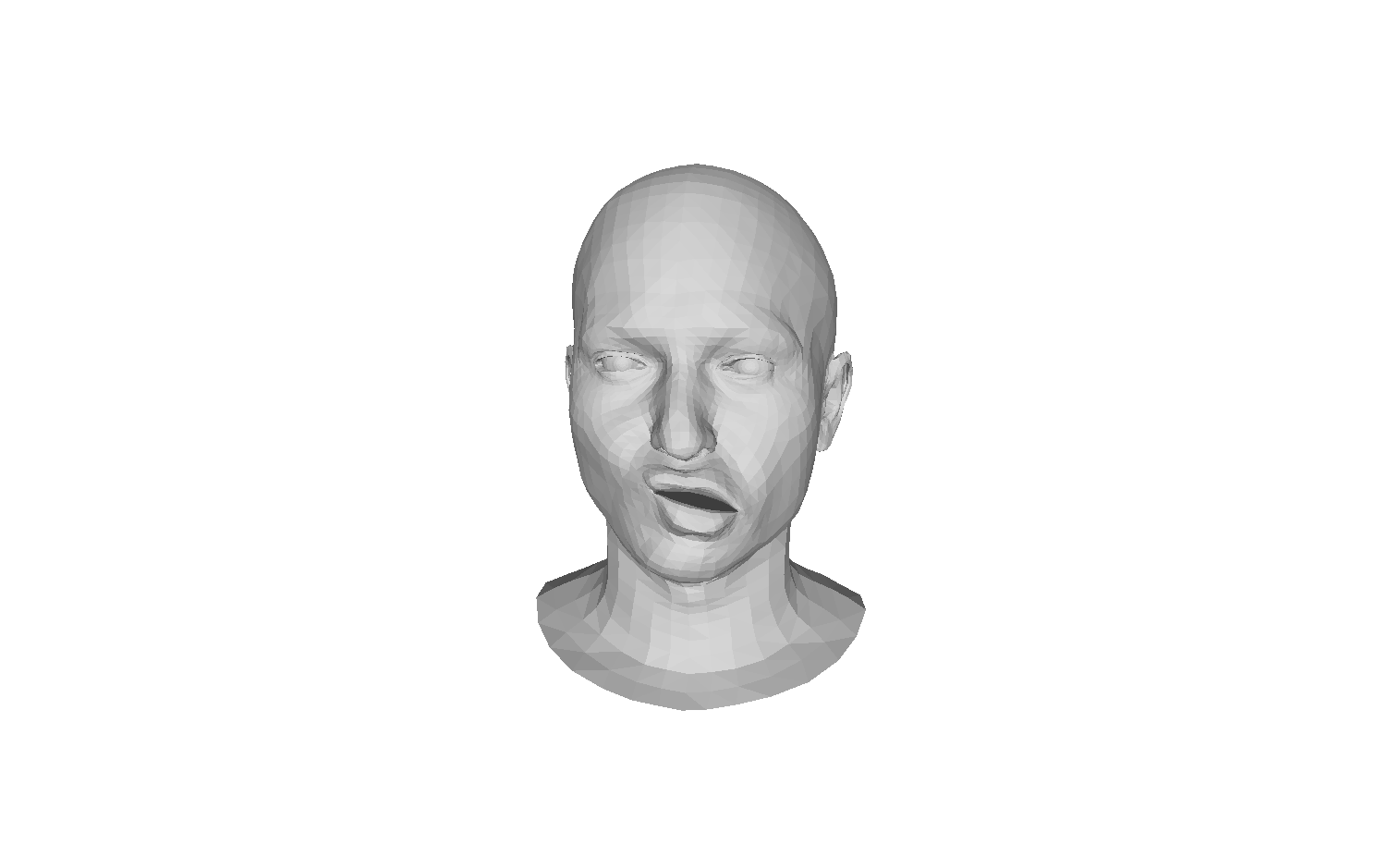}};
    
    \node[below of=c1, node distance=1.8cm] {Target Image};
    \coordinate (c56) at ($(c5)!0.5!(c6)$);
    \node[below of=c56, node distance=1.8cm] {Interpolated 3D Reconstructions};
    
    \end{tikzpicture}
    \caption{\textbf{Controlled generated of diverse 3D reconstructions between two distinct modes.} \ourmethod{} can be used to generate controlled diversity on the occluded regions by performing interpolation between two distinct shapes in the latent space.}
    \label{fig:interpolate}
\end{figure*}

\begin{figure*}
    \centering
    \begin{tikzpicture}
    \footnotesize
    \node (a1) {\includegraphics[width=0.09\linewidth]{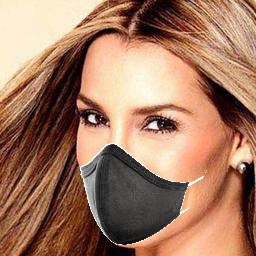}};
    \node[right of=a1, node distance=2.0cm] (a2) {\includegraphics[trim={400 80 400 100},clip,width=0.09\linewidth]{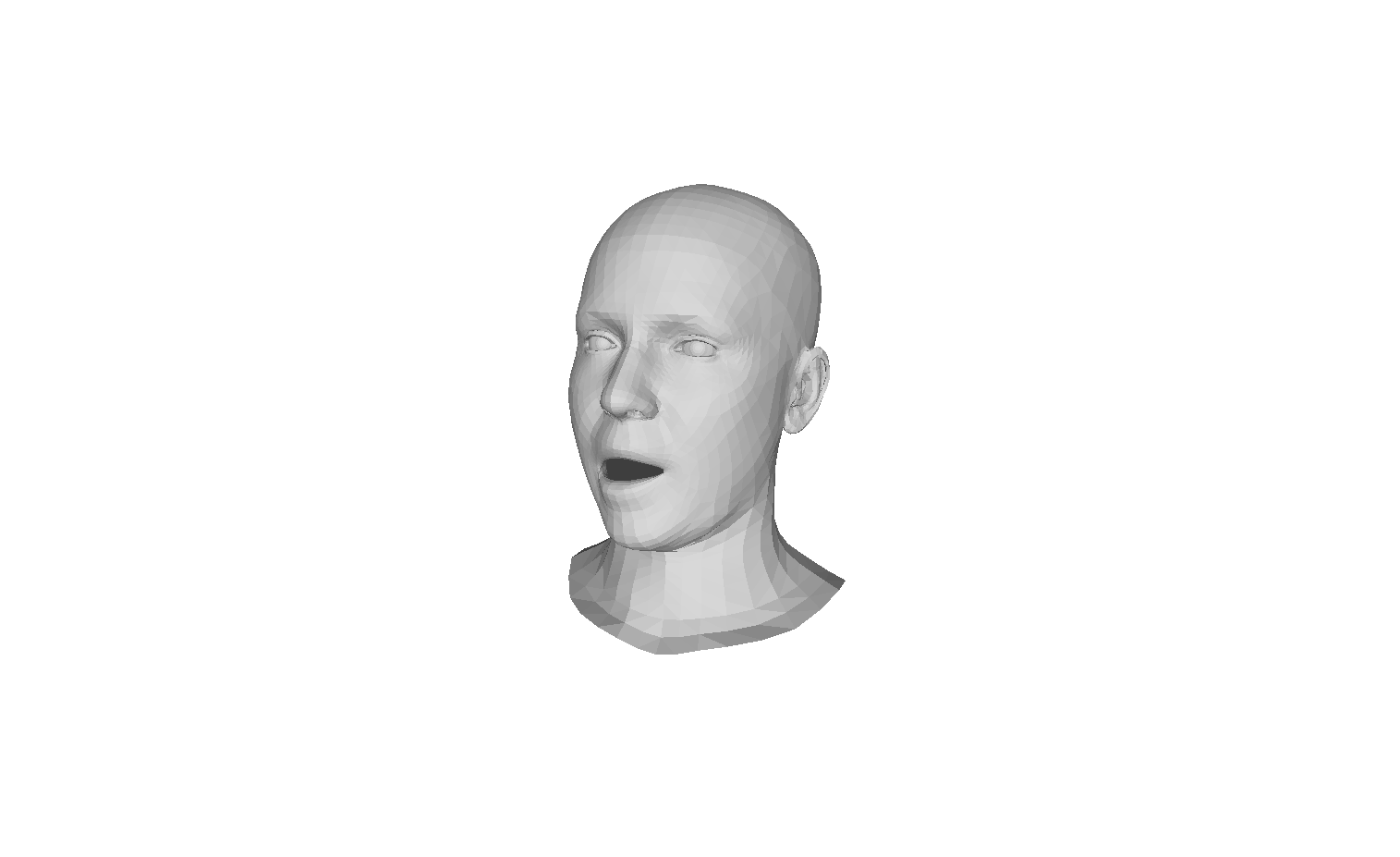}};
    \node[right of=a2, node distance=1.5cm] (a3) {\includegraphics[trim={400 80 400 100},clip,width=0.09\linewidth]{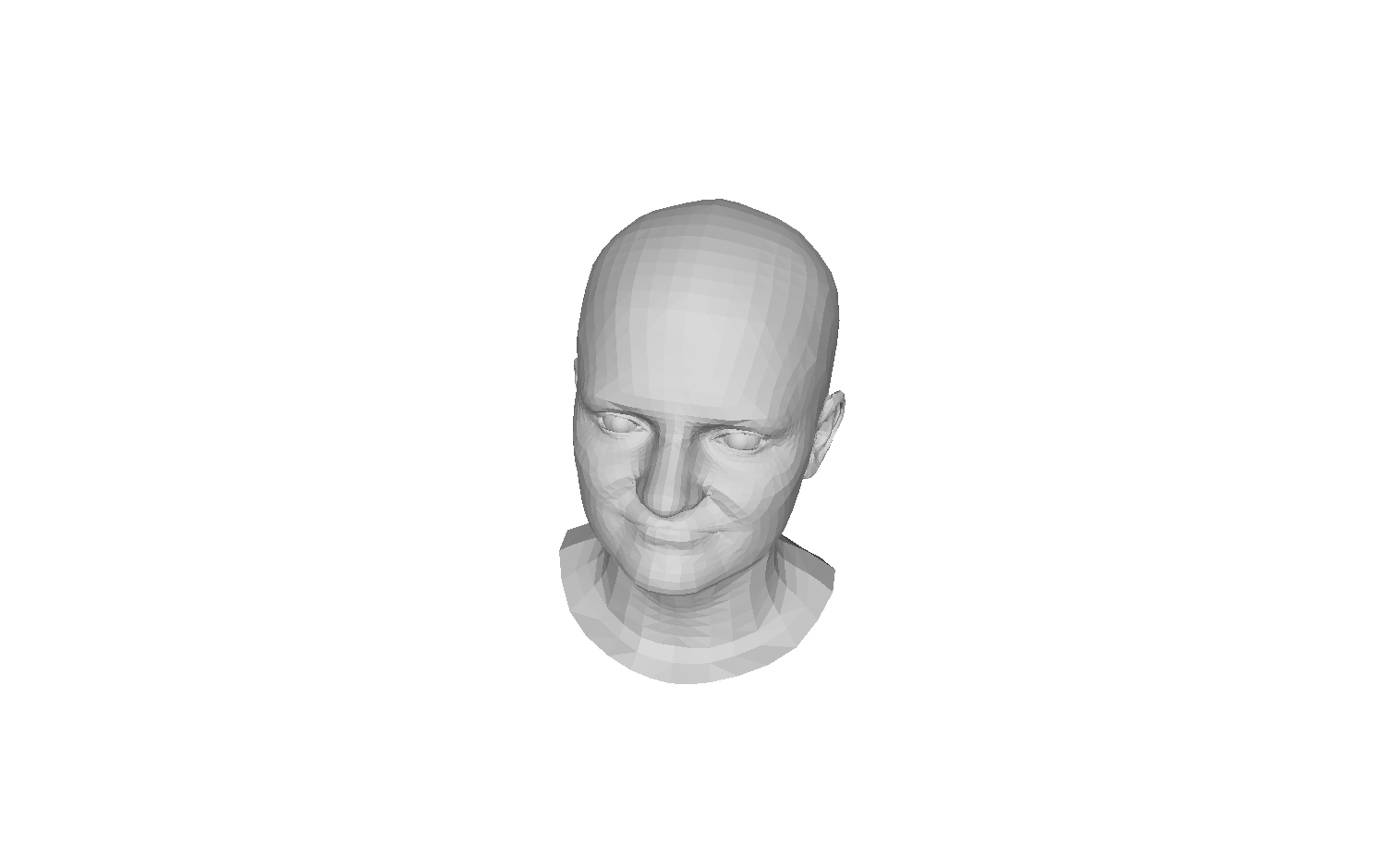}};
    \node[right of=a3, node distance=1.4cm] (a4) {\includegraphics[trim={400 80 400 100},clip,width=0.075\linewidth]{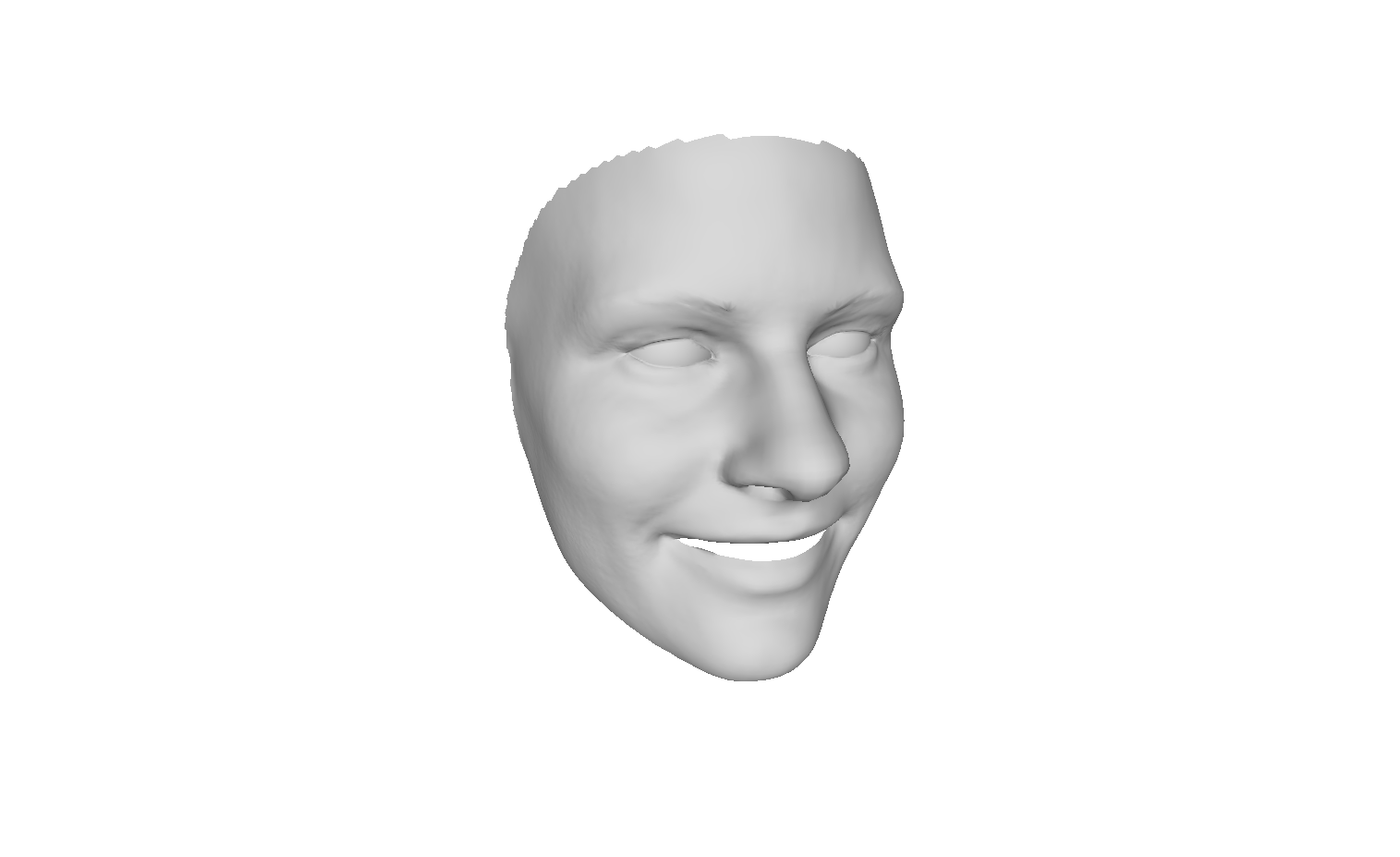}};
    \node[right of=a4, node distance=1.4cm] (a5) {\includegraphics[trim={400 80 400 100},clip,width=0.075\linewidth]{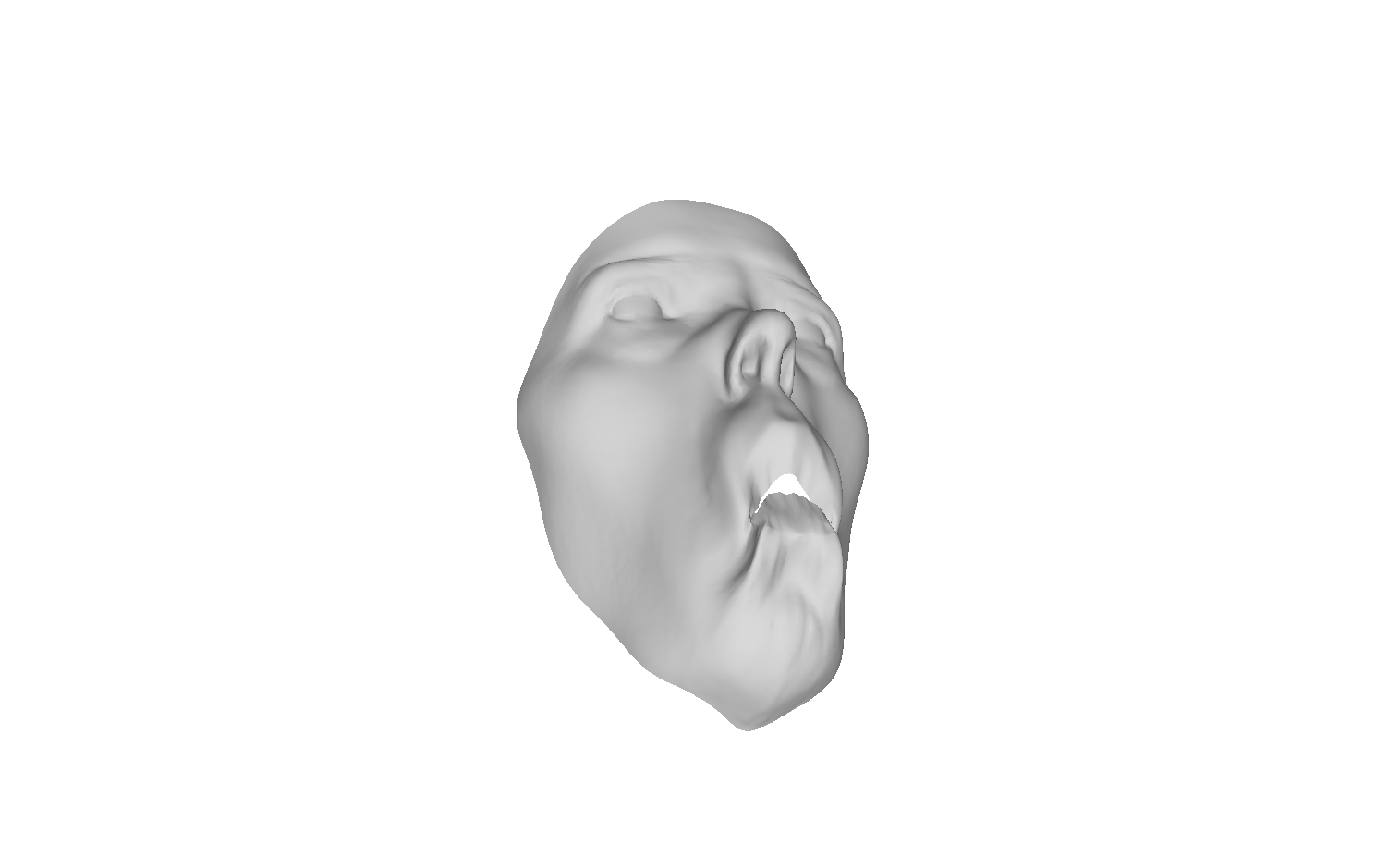}};
    \node[right of=a5, node distance=1.5cm] (a6) {\includegraphics[trim={350 80 400 100},clip,width=0.085\linewidth]{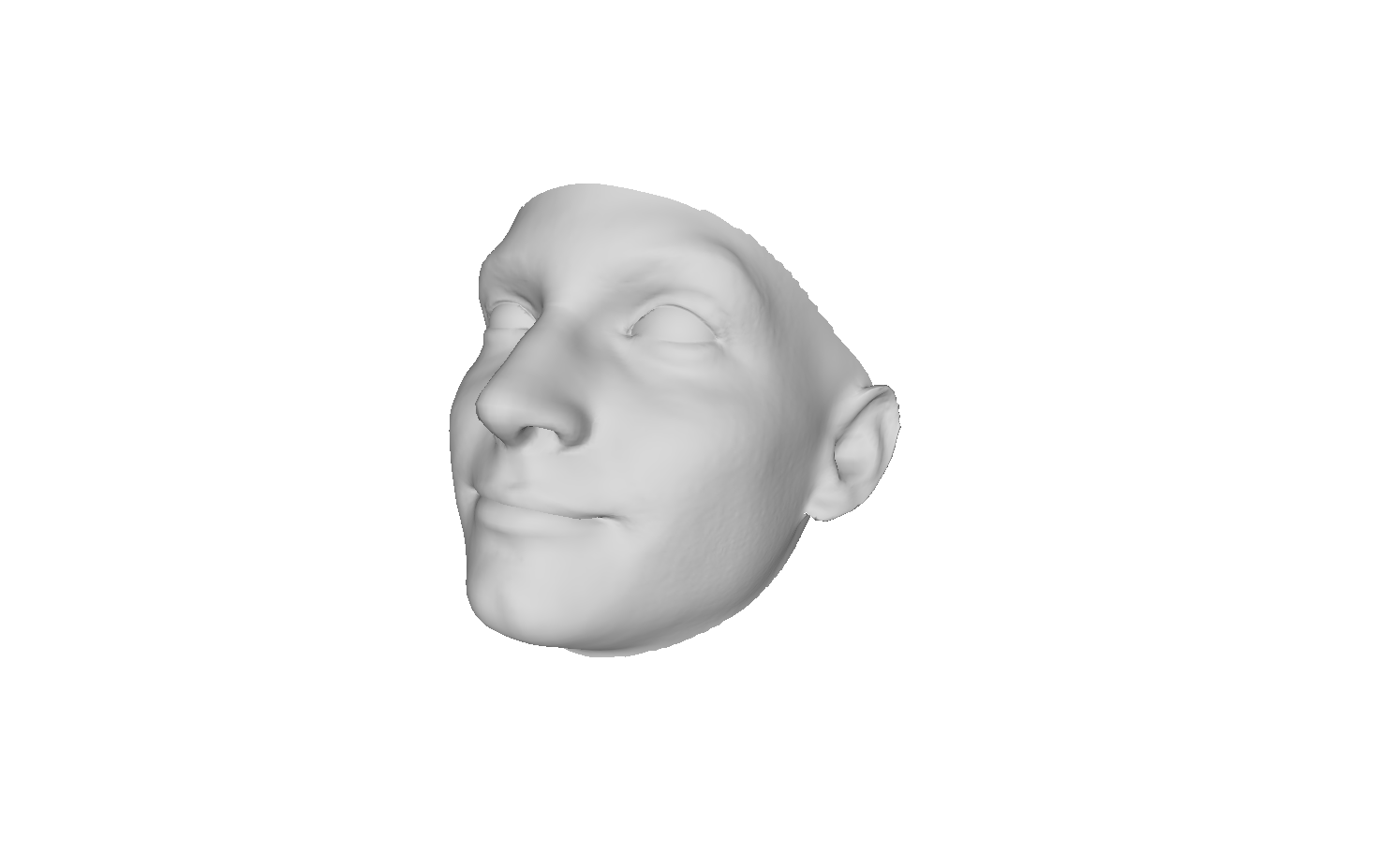}};
    \node[right of=a6, node distance=2.1cm] (a7) {\includegraphics[trim={400 80 400 100},clip,width=0.09\linewidth]{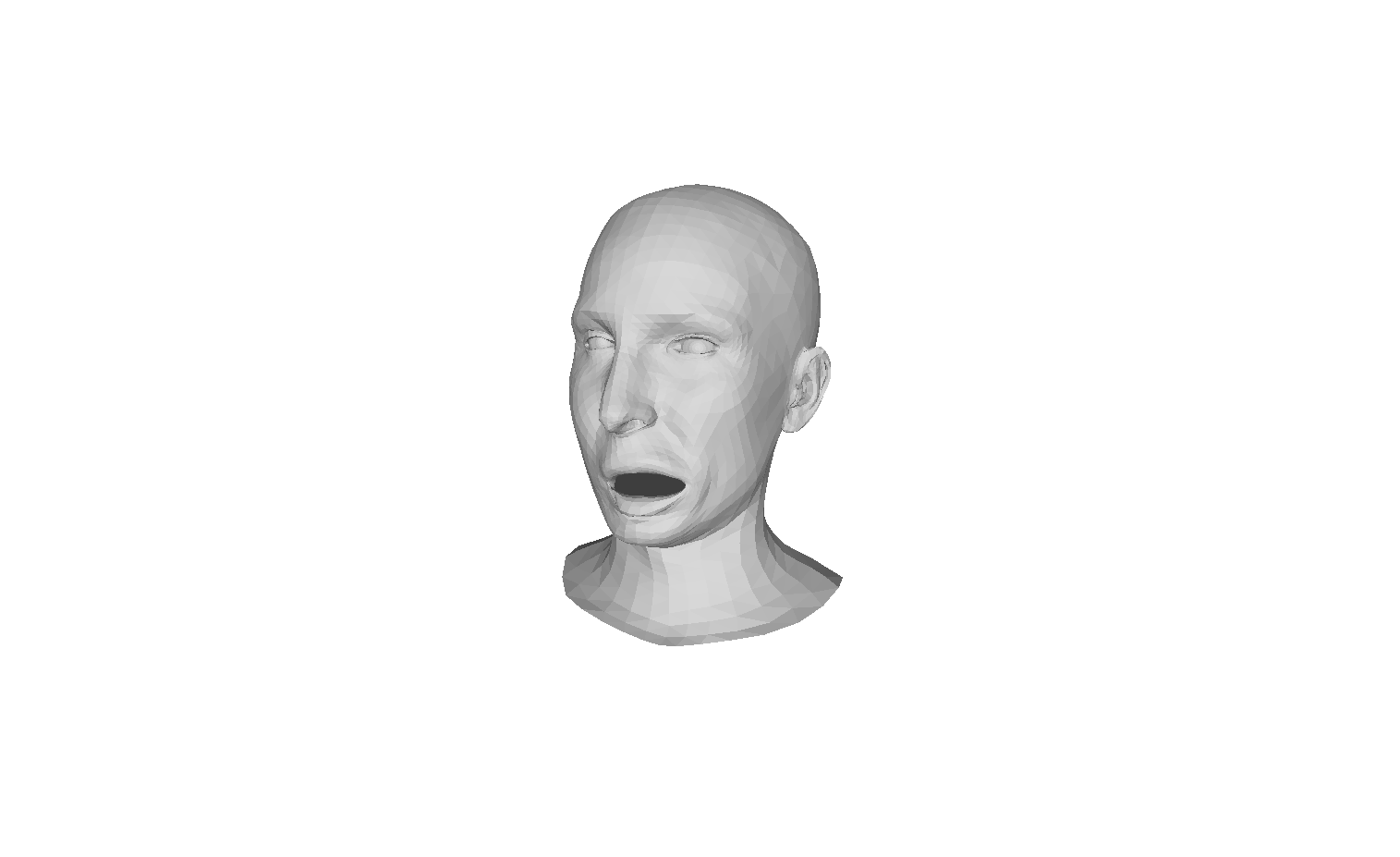}};
    \node[right of=a7, node distance=1.4cm] (a8) {\includegraphics[trim={400 80 400 100},clip,width=0.09\linewidth]{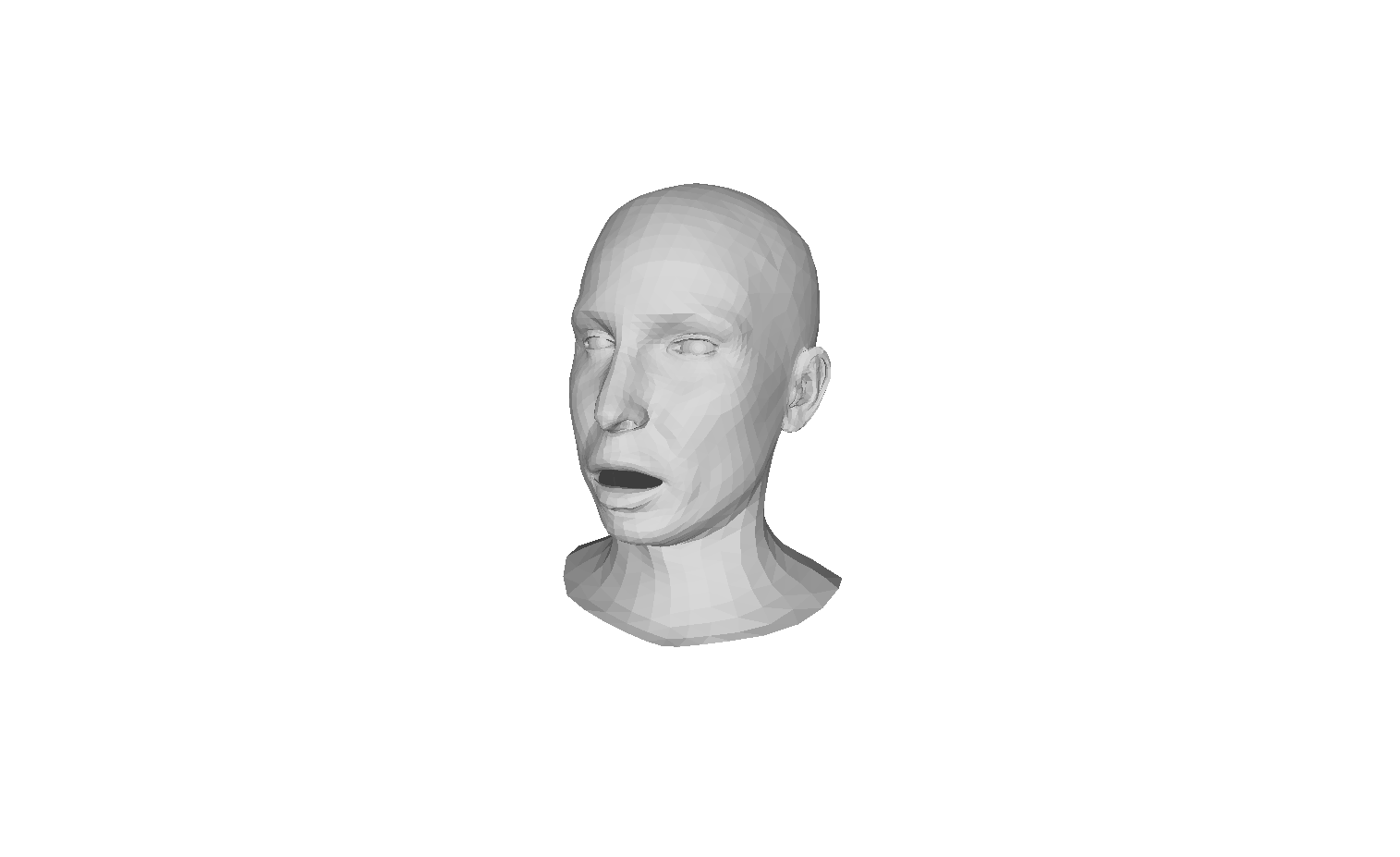}};
    \node[right of=a8, node distance=1.4cm] (a9) {\includegraphics[trim={400 80 400 100},clip,width=0.09\linewidth]{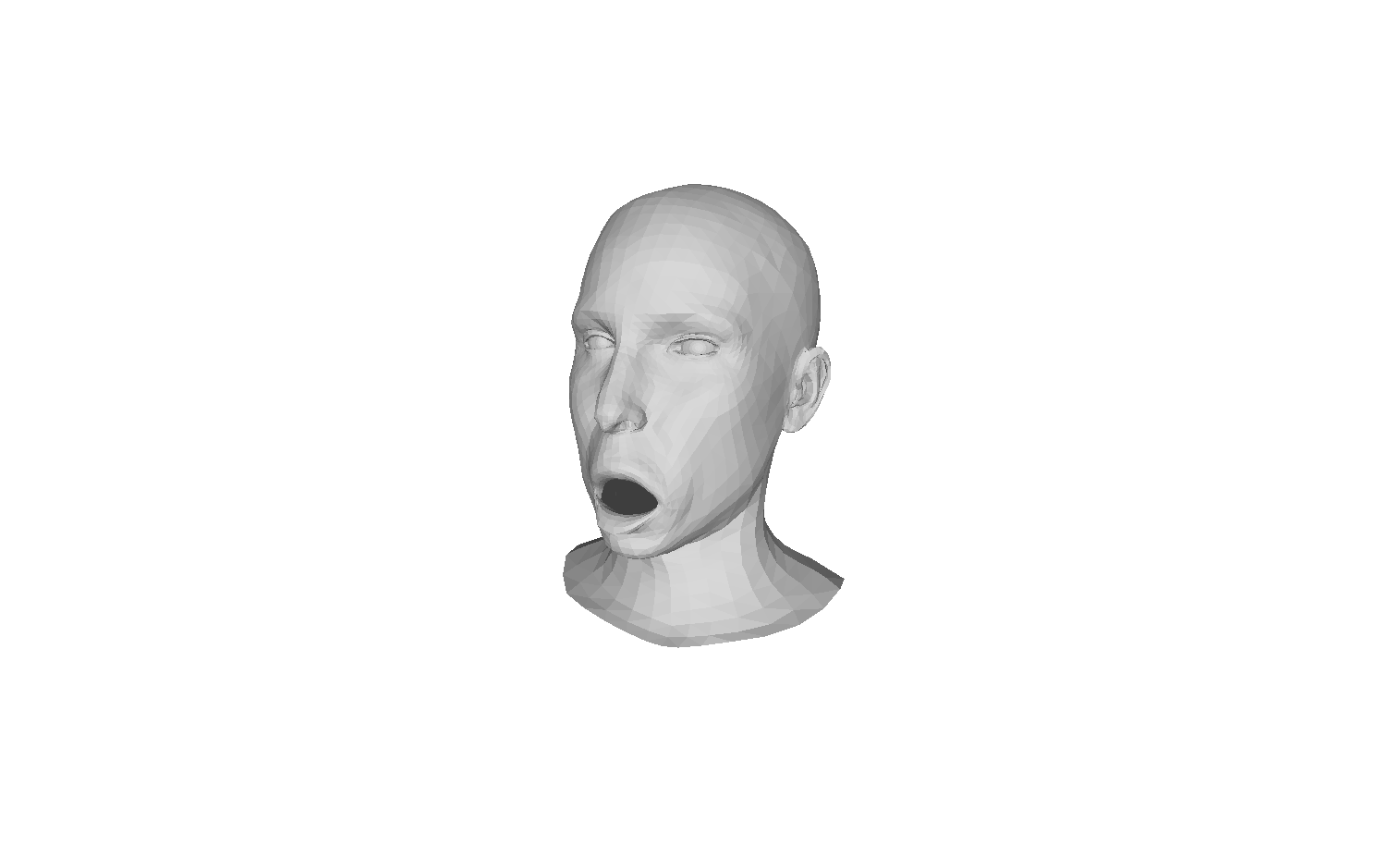}};
    \node[right of=a9, node distance=1.4cm] (a10) {\includegraphics[trim={400 80 400 100},clip,width=0.09\linewidth]{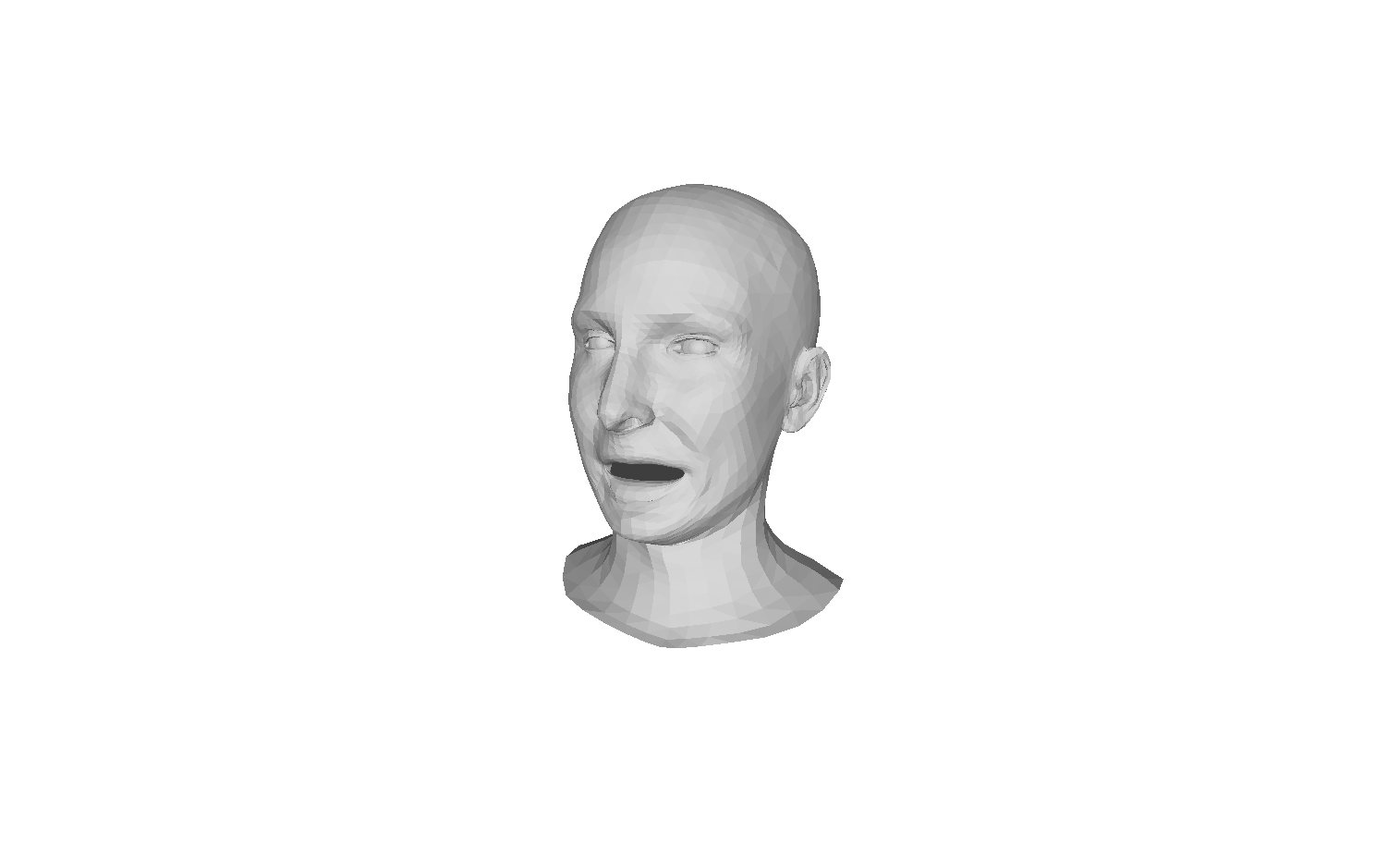}};
    \node[right of=a10, node distance=1.4cm] (a11) {\includegraphics[trim={400 80 400 100},clip,width=0.09\linewidth]{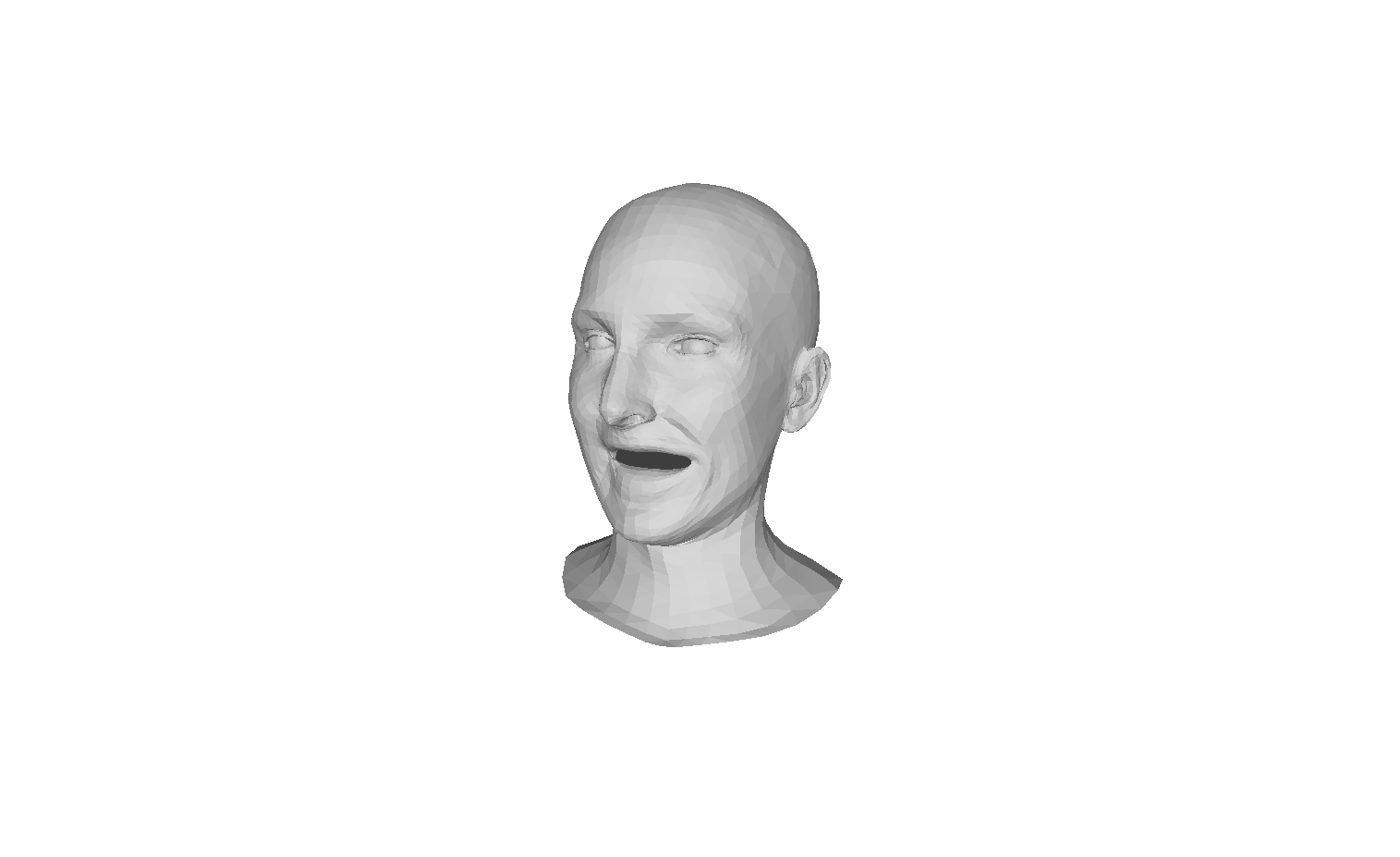}};
    
    \node[below of=a1, node distance=2.2cm] (b1) {\includegraphics[width=0.09\linewidth]{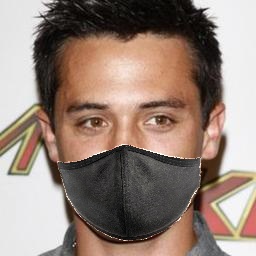}};
    \node[right of=b1, node distance=2.0cm] (b2) {\includegraphics[trim={400 80 400 100},clip,width=0.09\linewidth]{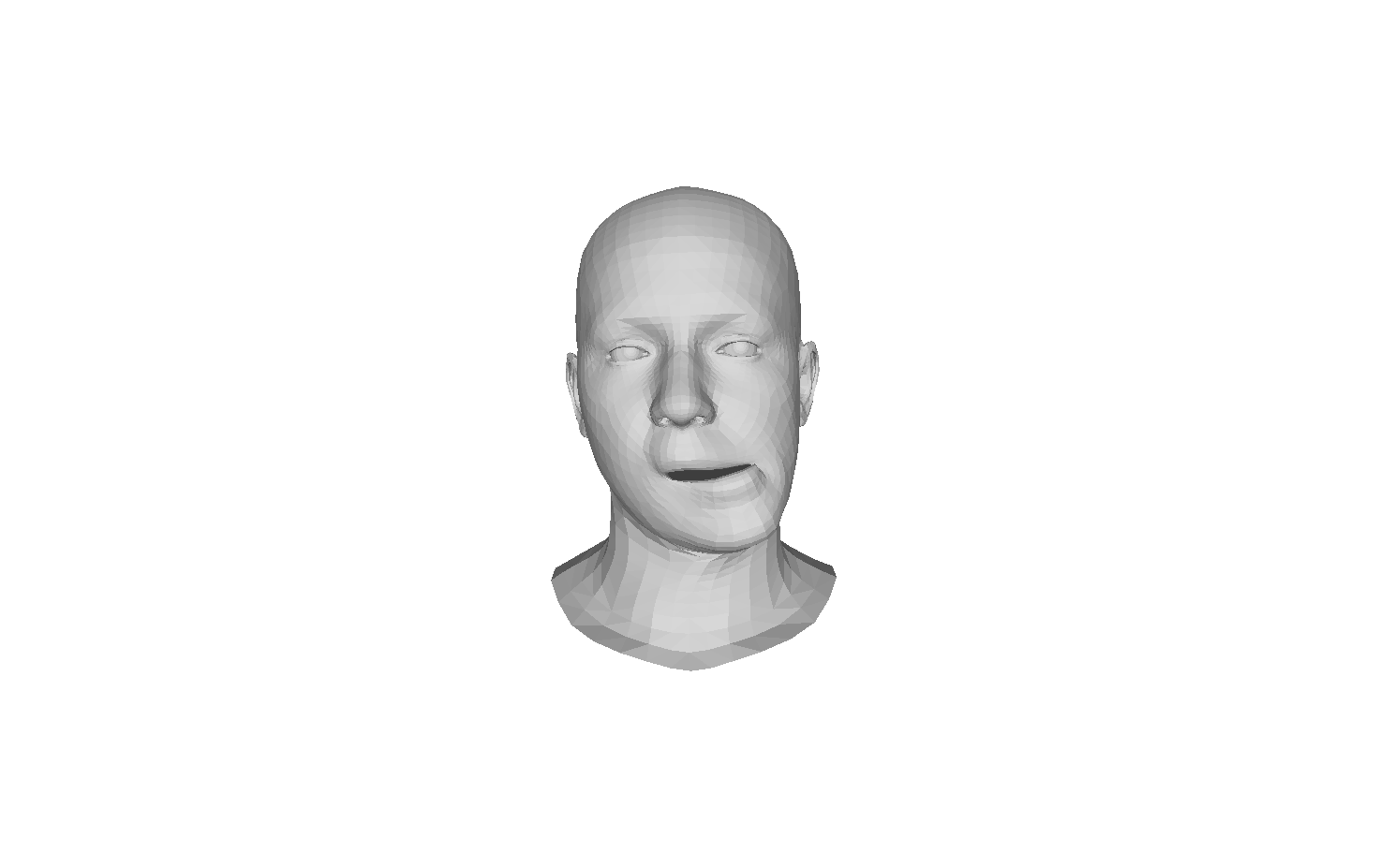}};
    \node[right of=b2, node distance=1.5cm] (b3) {\includegraphics[trim={400 80 400 100},clip,width=0.09\linewidth]{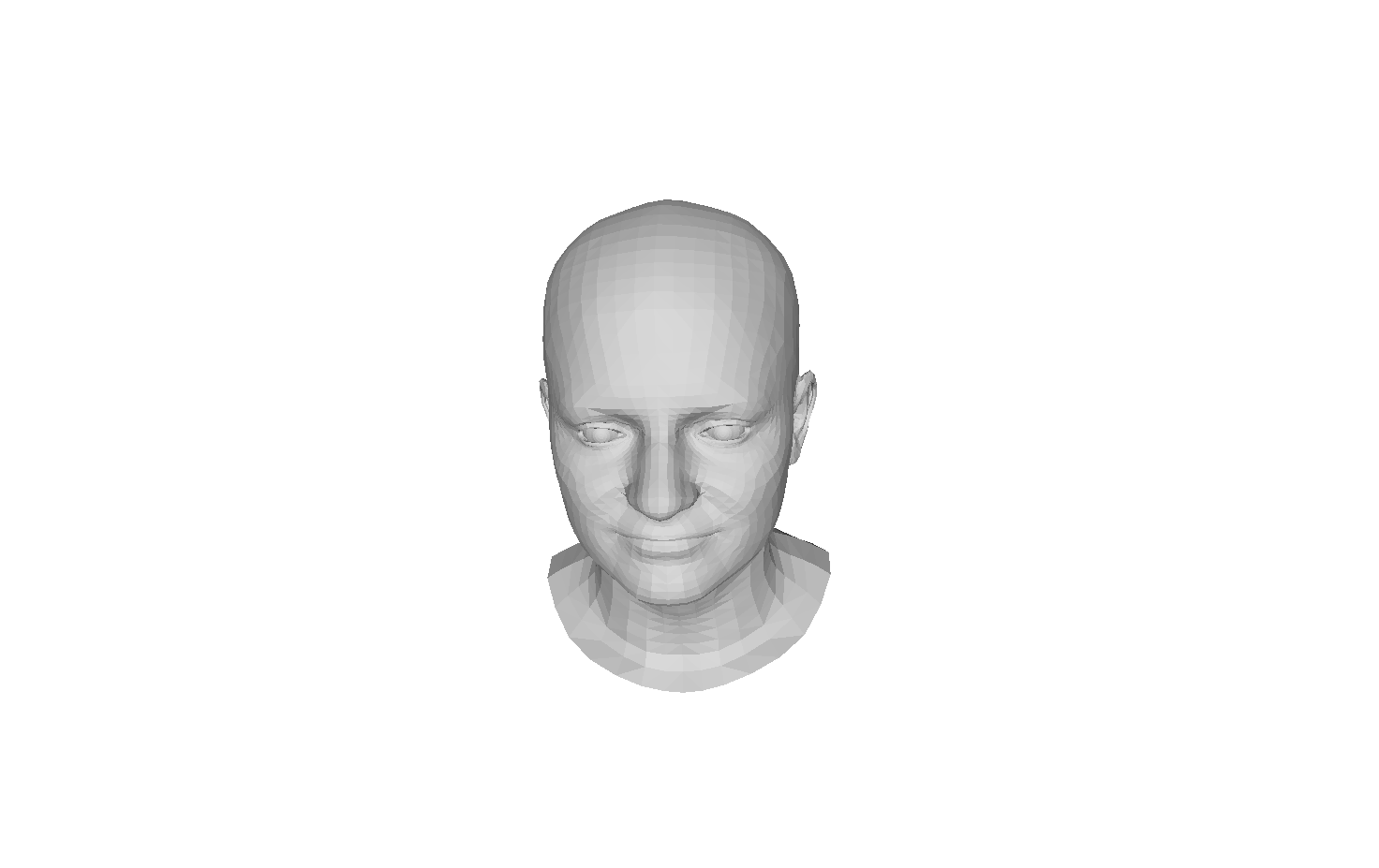}};
    \node[right of=b3, node distance=1.4cm] (b4) {\includegraphics[trim={400 80 400 100},clip,width=0.075\linewidth]{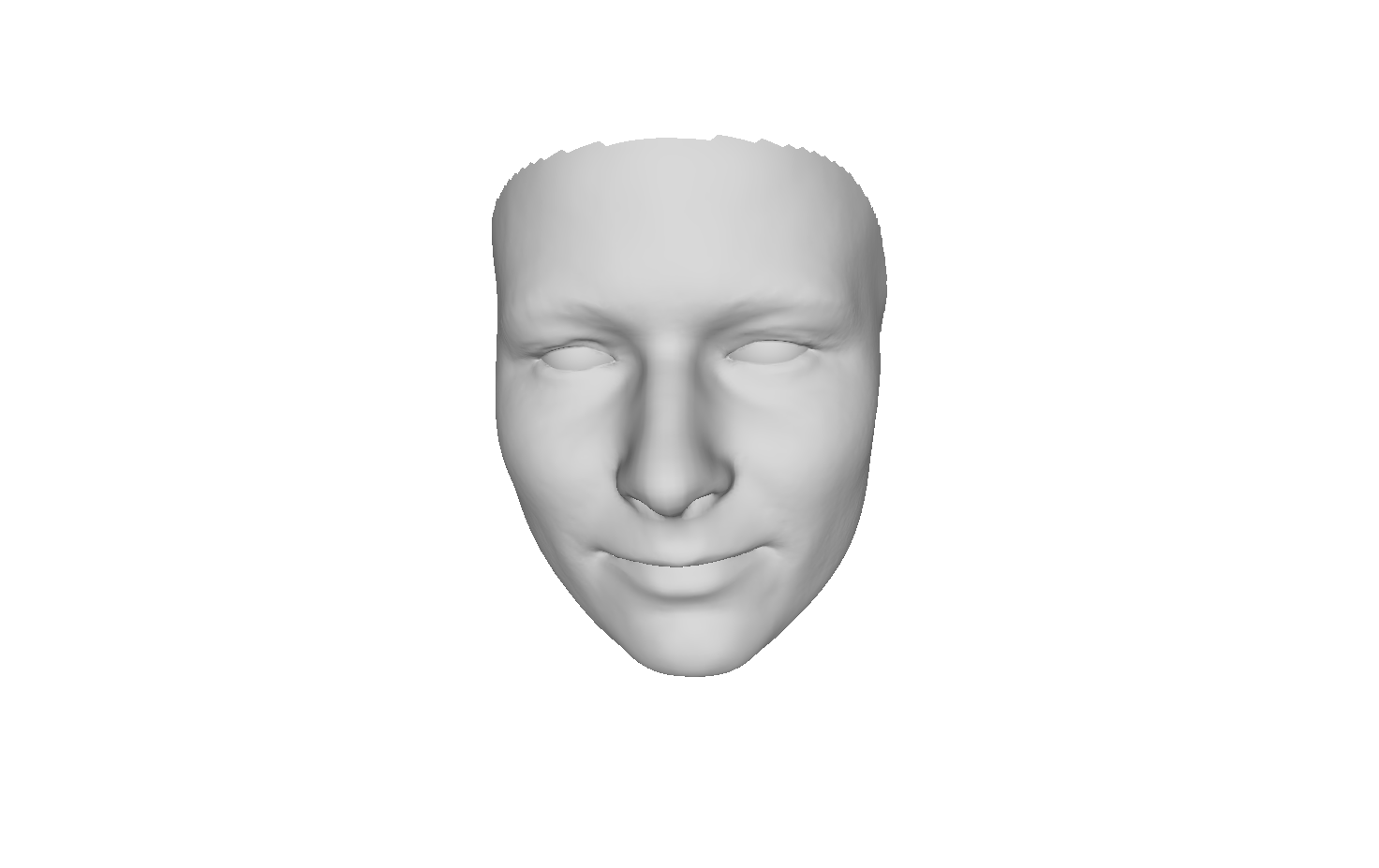}};
    \node[right of=b4, node distance=1.4cm] (b5) {\includegraphics[trim={400 80 400 100},clip,width=0.075\linewidth]{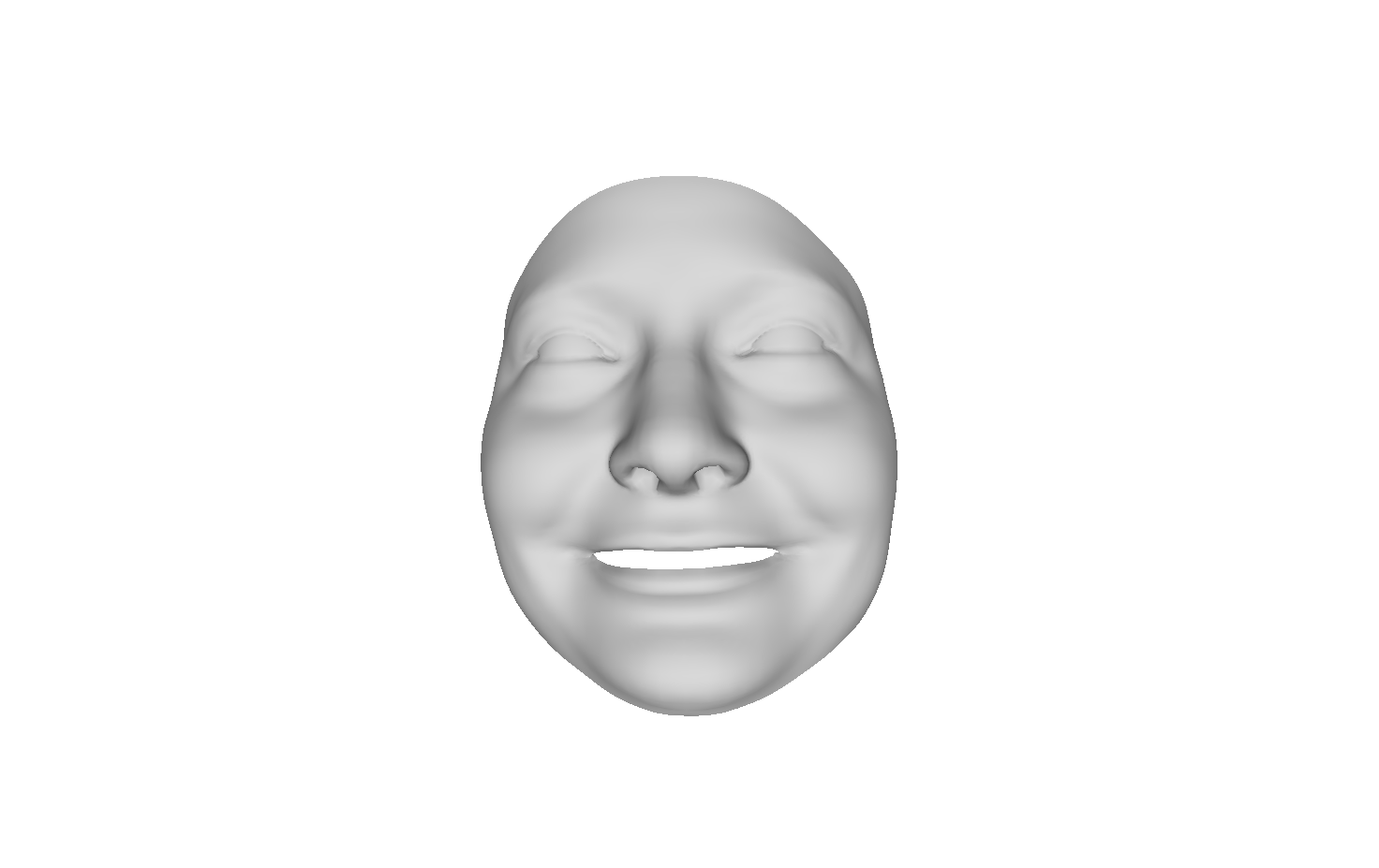}};
    \node[right of=b5, node distance=1.5cm] (b6) {\includegraphics[trim={350 80 400 100},clip,width=0.085\linewidth]{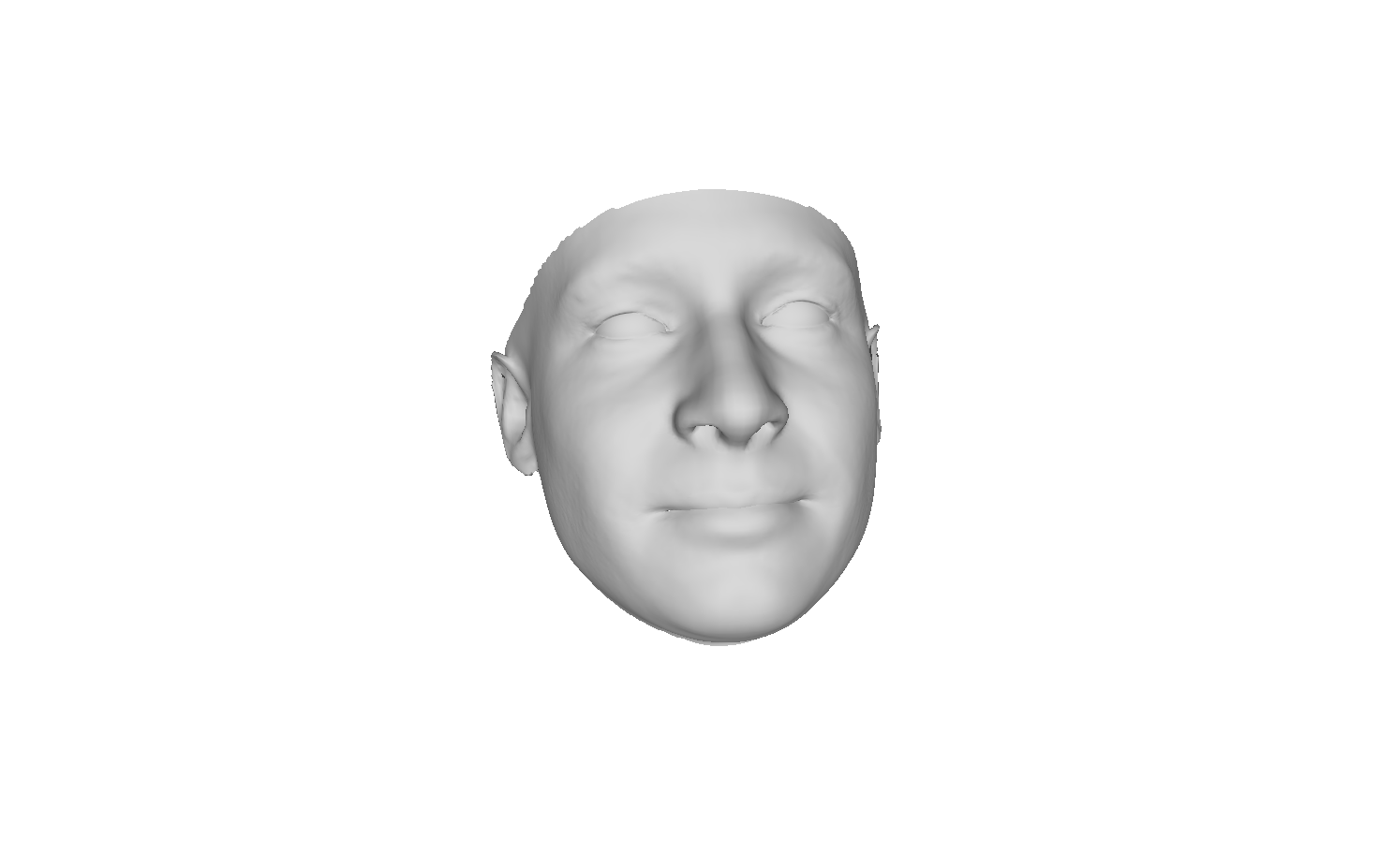}};
    \node[right of=b6, node distance=2.1cm] (b7) {\includegraphics[trim={400 80 400 100},clip,width=0.09\linewidth]{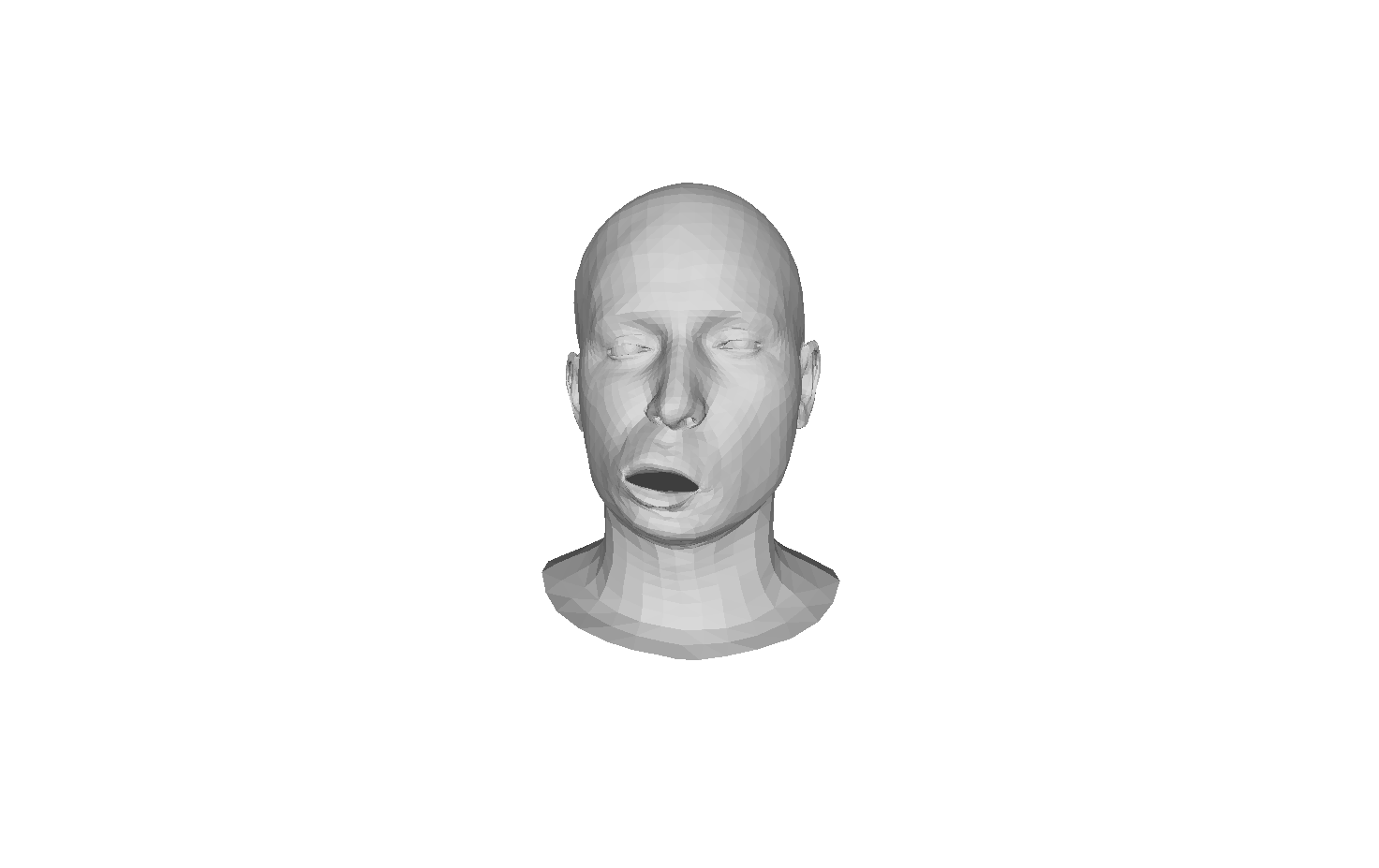}};
    \node[right of=b7, node distance=1.4cm] (b8) {\includegraphics[trim={400 80 400 100},clip,width=0.09\linewidth]{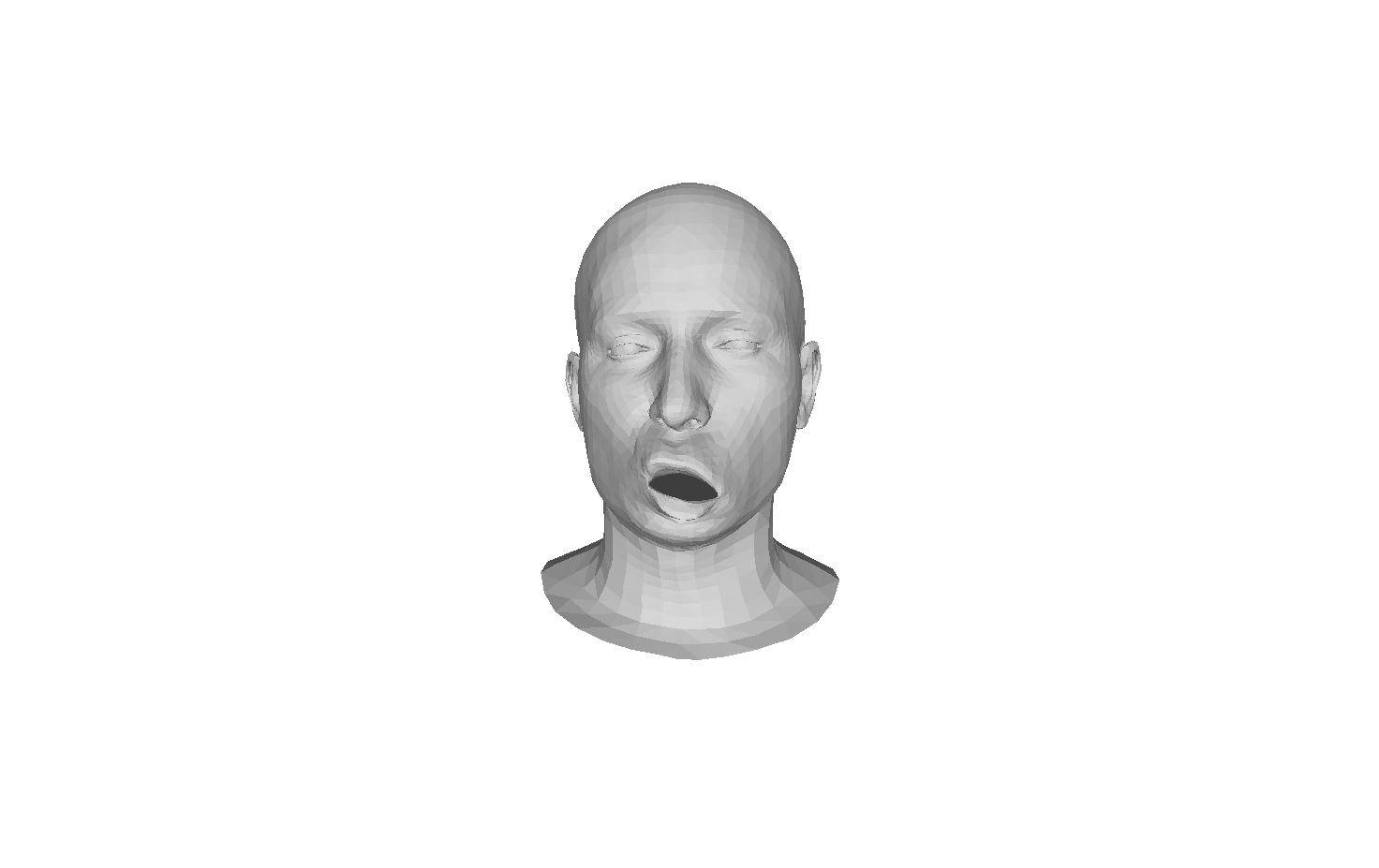}};
    \node[right of=b8, node distance=1.4cm] (b9) {\includegraphics[trim={400 80 400 100},clip,width=0.09\linewidth]{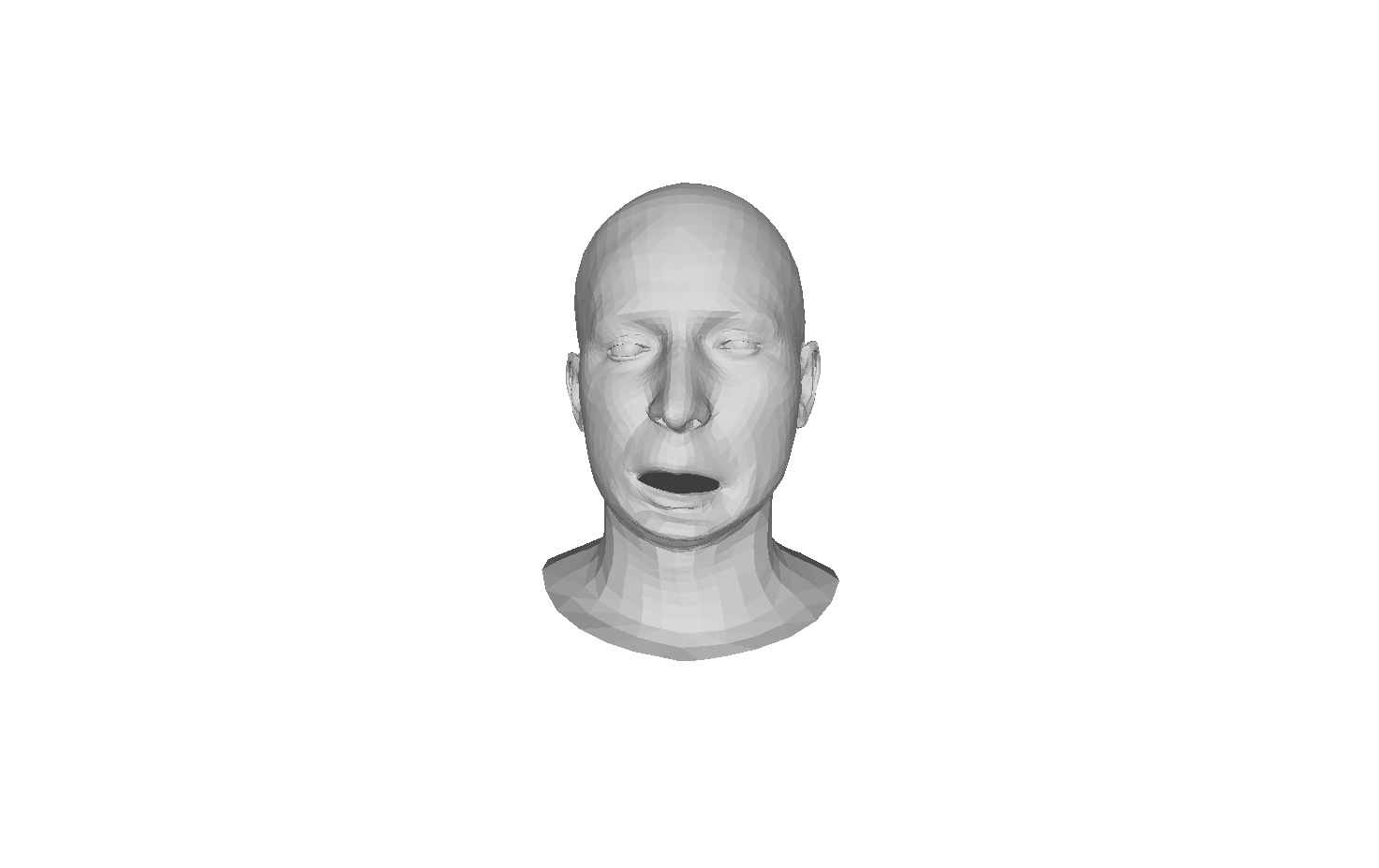}};
    \node[right of=b9, node distance=1.4cm] (b10) {\includegraphics[trim={400 80 400 100},clip,width=0.09\linewidth]{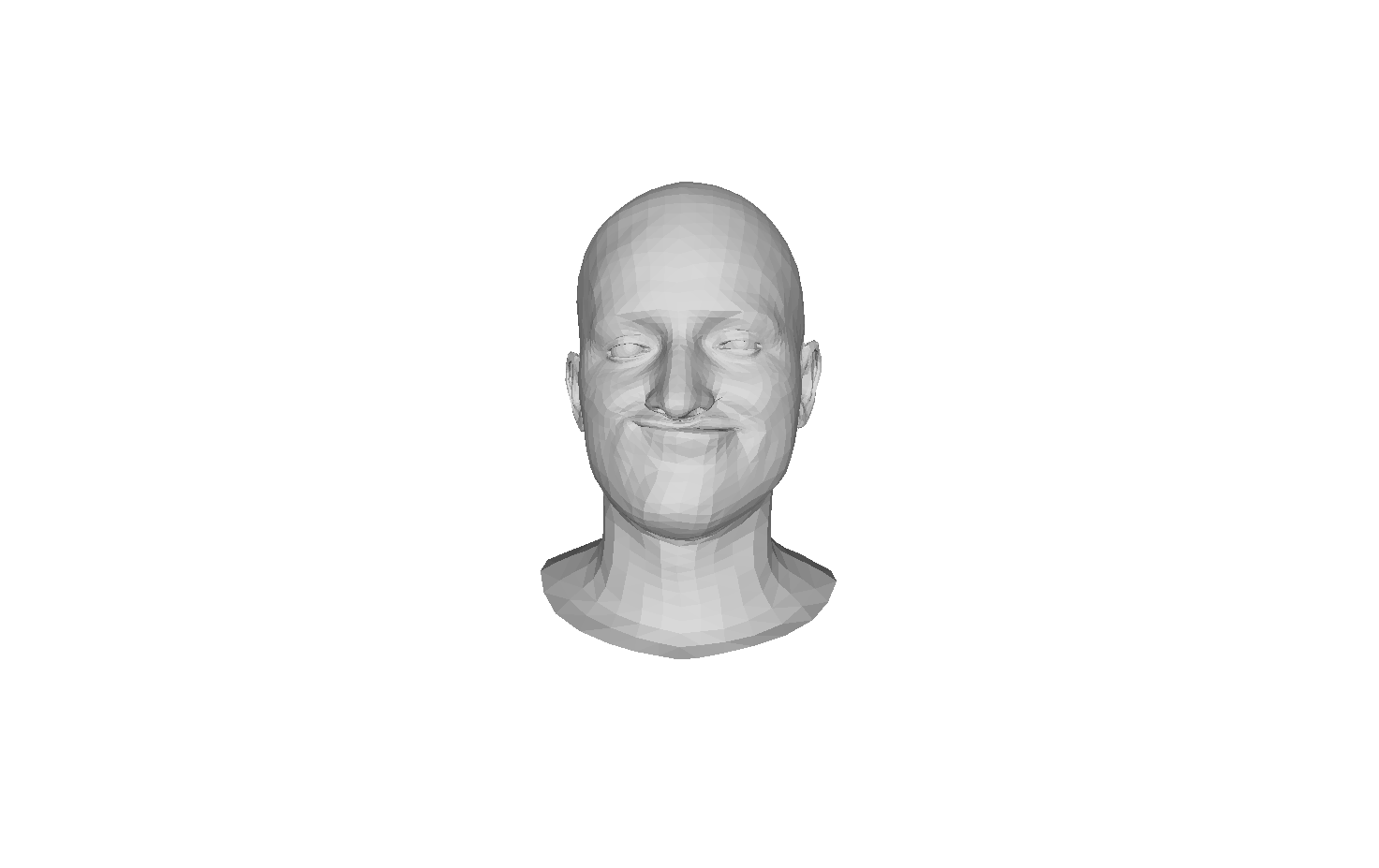}};
    \node[right of=b10, node distance=1.4cm] (b11) {\includegraphics[trim={400 80 400 100},clip,width=0.09\linewidth]{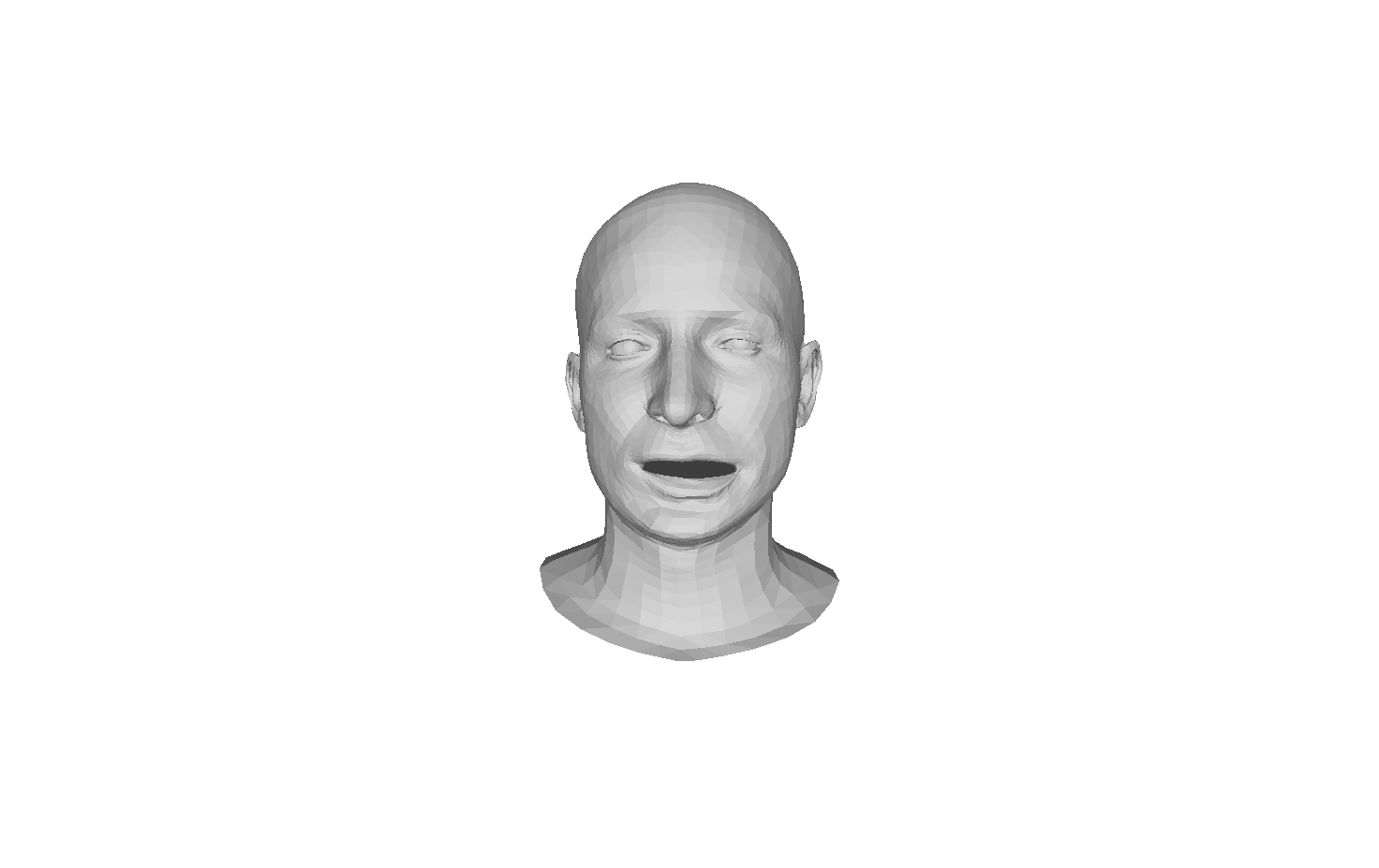}};
    
    \node[below of=b1, node distance=2.2cm] (c1) {\includegraphics[width=0.09\linewidth]{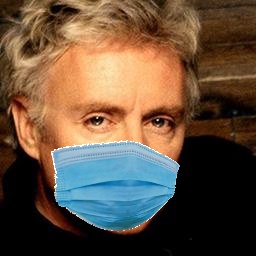}};
    \node[right of=c1, node distance=2.0cm] (c2) {\includegraphics[trim={400 80 400 100},clip,width=0.09\linewidth]{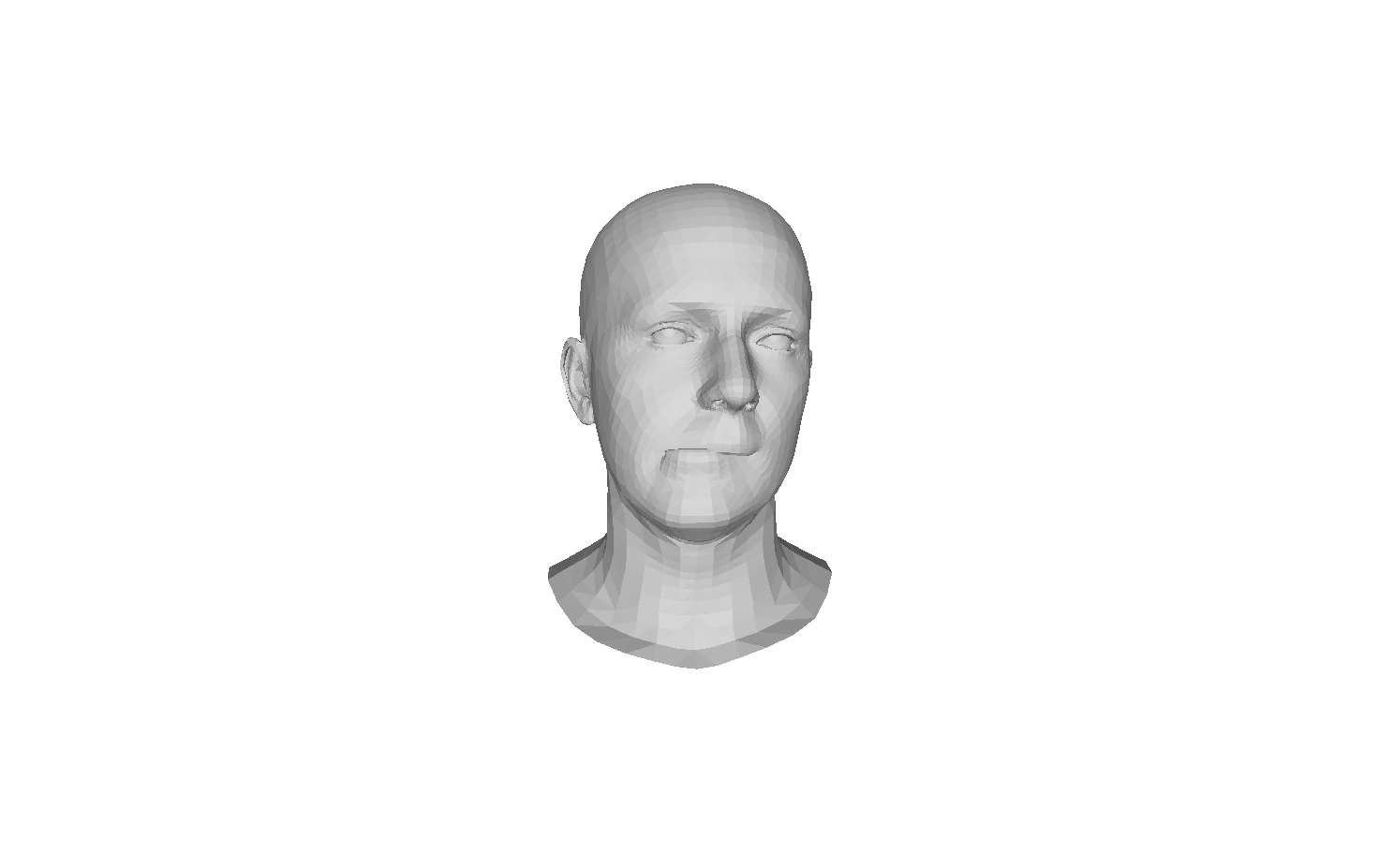}};
    \node[right of=c2, node distance=1.5cm] (c3) {\includegraphics[trim={400 80 400 100},clip,width=0.09\linewidth]{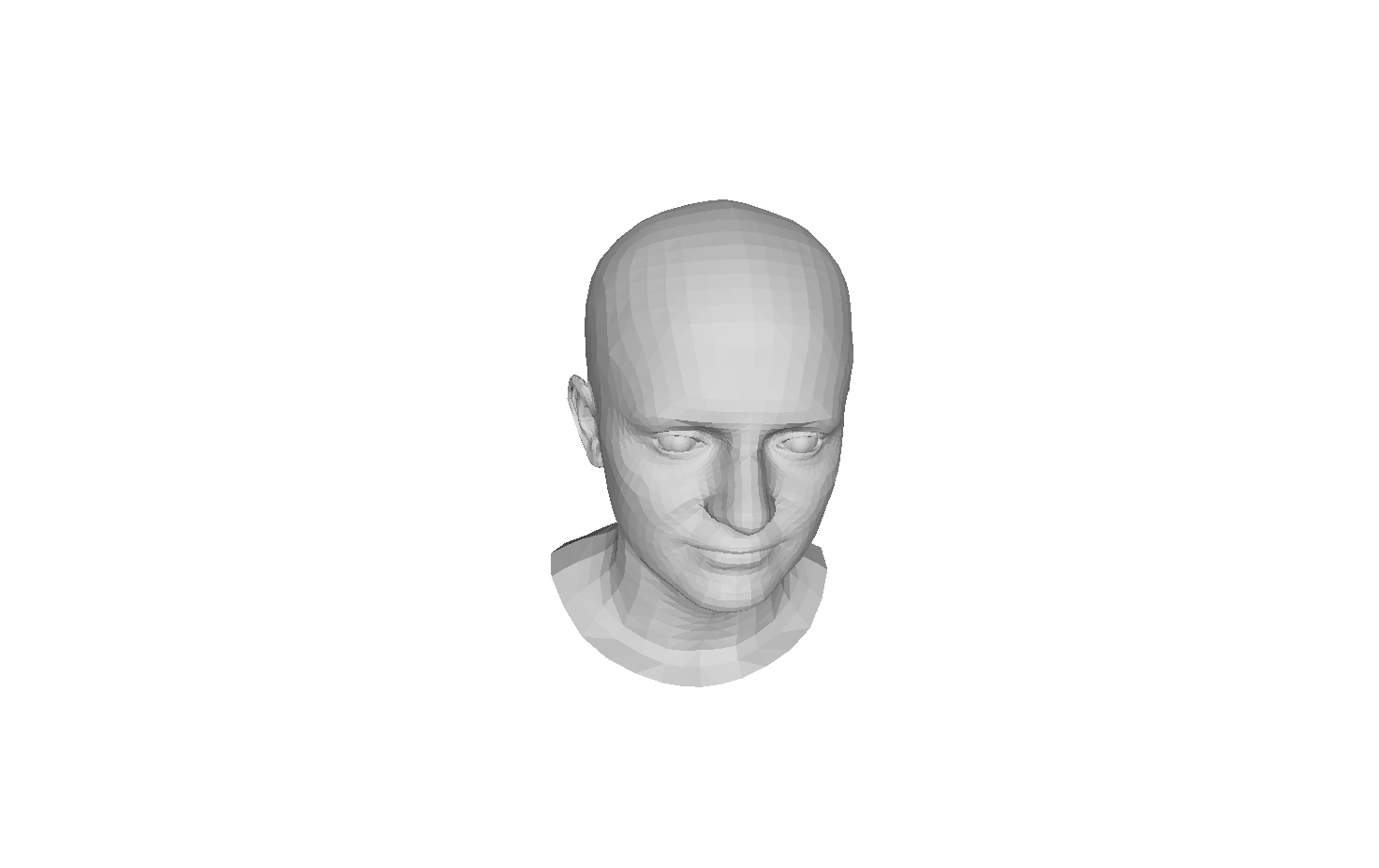}};
    \node[right of=c3, node distance=1.4cm] (c4) {\includegraphics[trim={400 80 400 100},clip,width=0.075\linewidth]{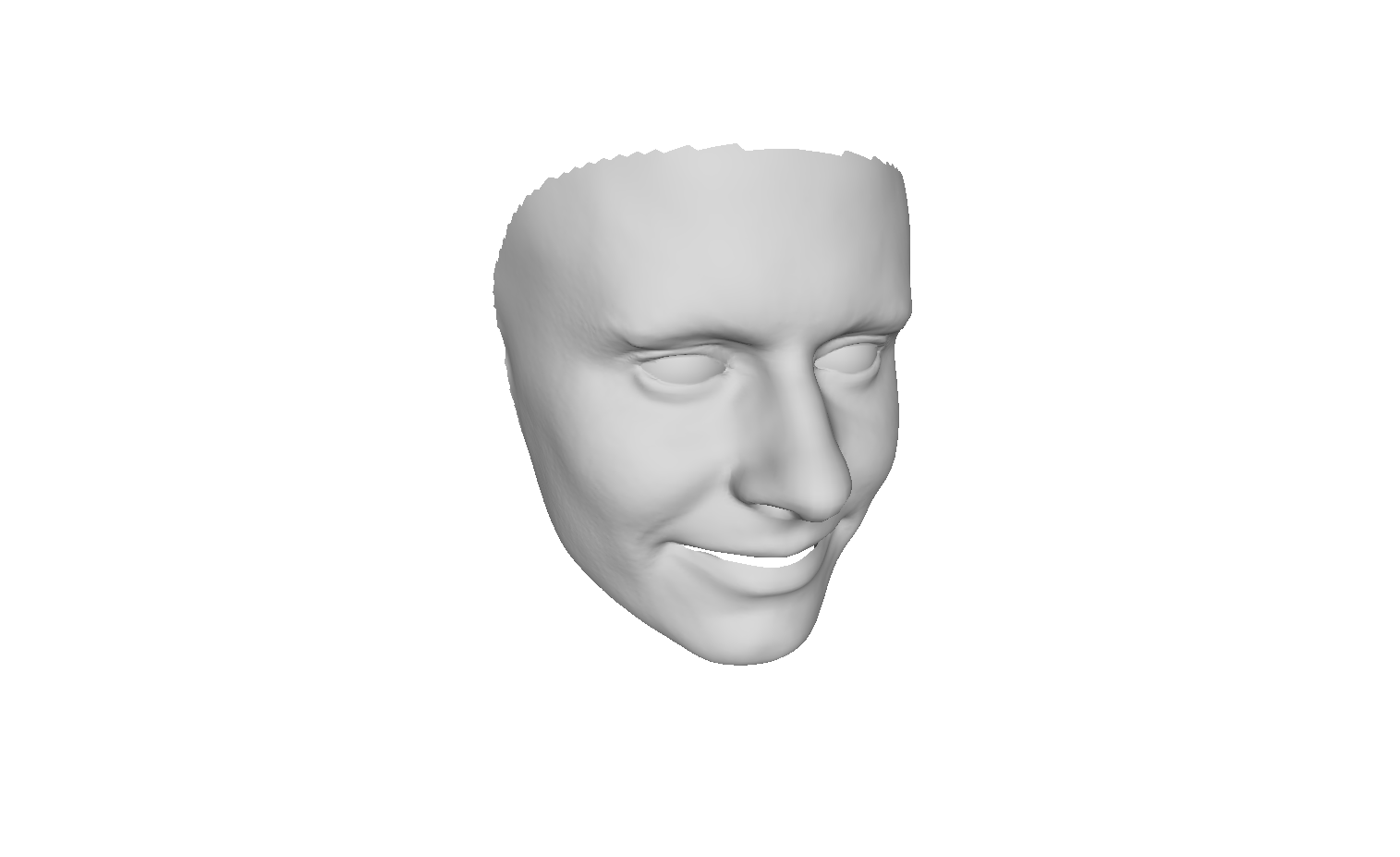}};
    \node[right of=c4, node distance=1.4cm] (c5) {\includegraphics[trim={400 80 400 100},clip,width=0.075\linewidth]{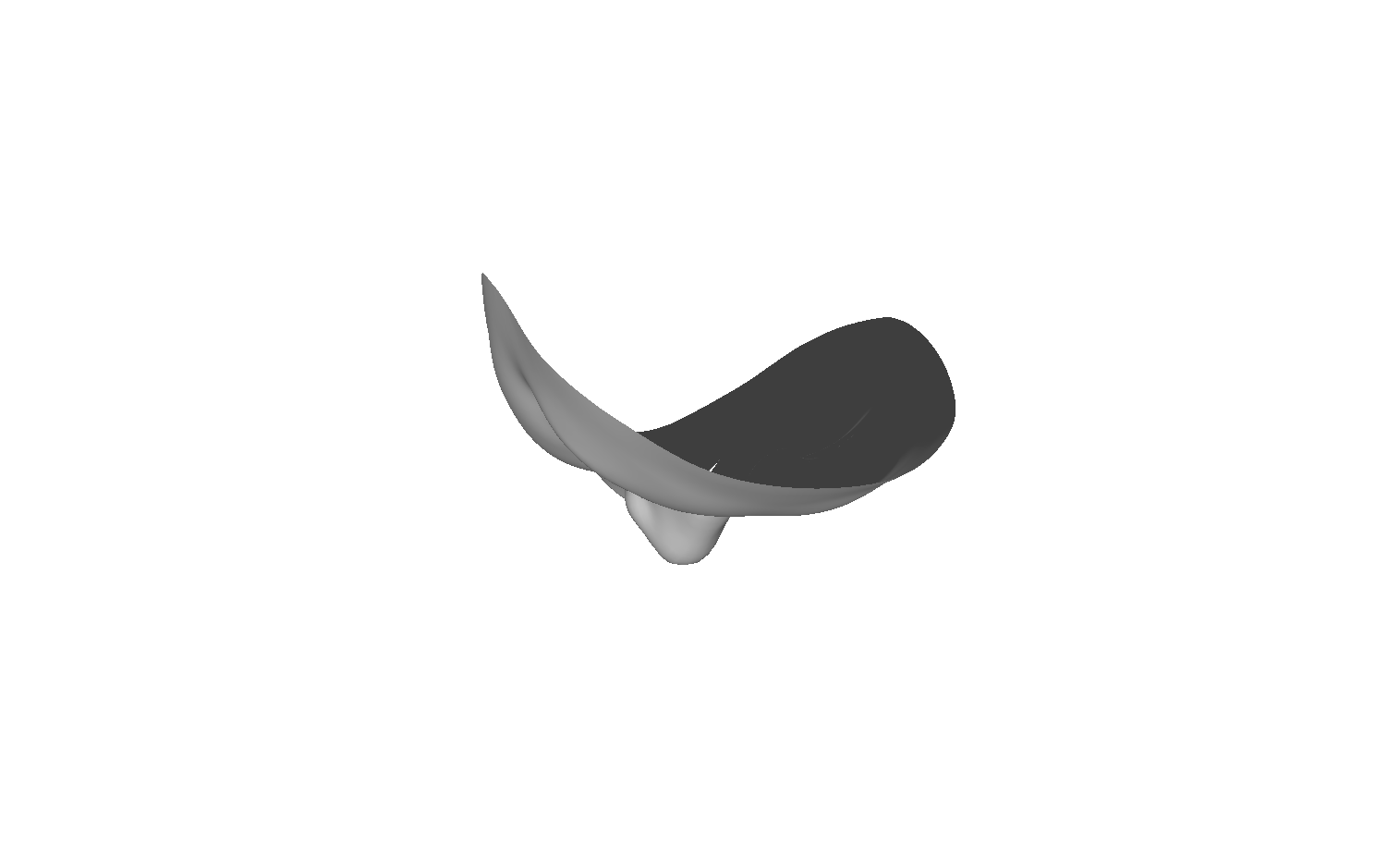}};
    \node[right of=c5, node distance=1.5cm] (c6) {\includegraphics[trim={350 80 400 100},clip,width=0.085\linewidth]{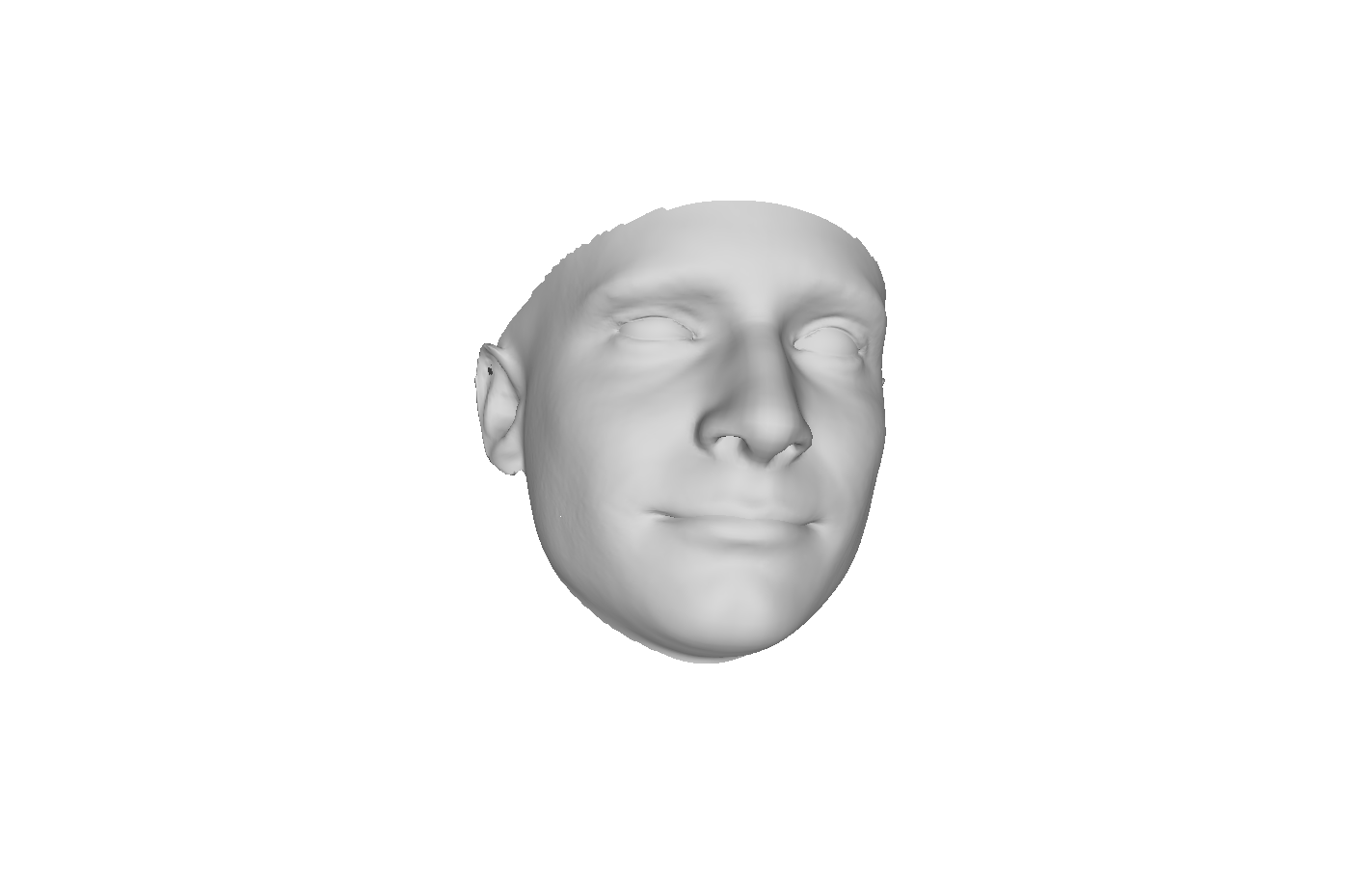}};
    \node[right of=c6, node distance=2.1cm] (c7) {\includegraphics[trim={400 80 400 100},clip,width=0.09\linewidth]{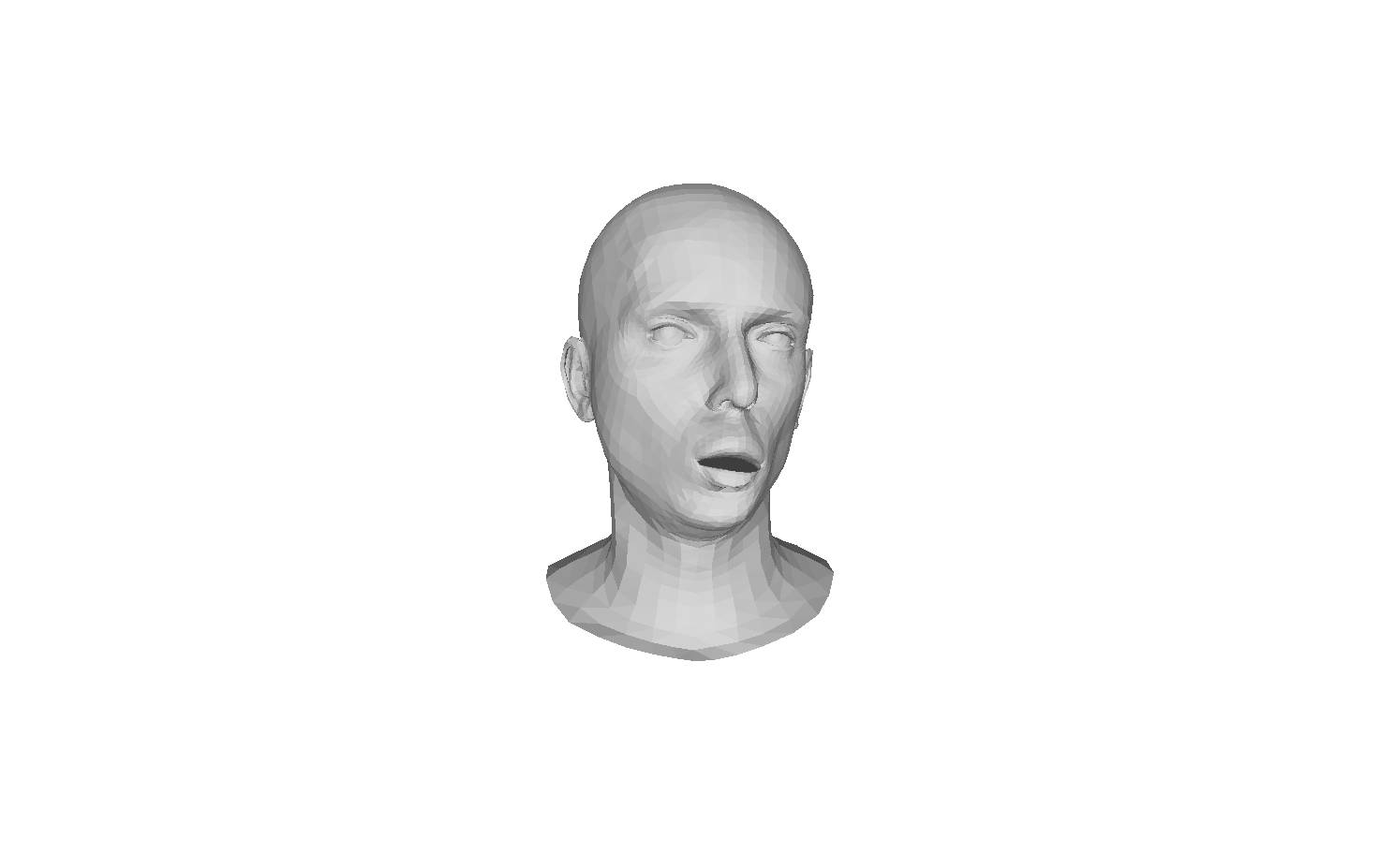}};
    \node[right of=c7, node distance=1.4cm] (c8) {\includegraphics[trim={400 80 400 100},clip,width=0.09\linewidth]{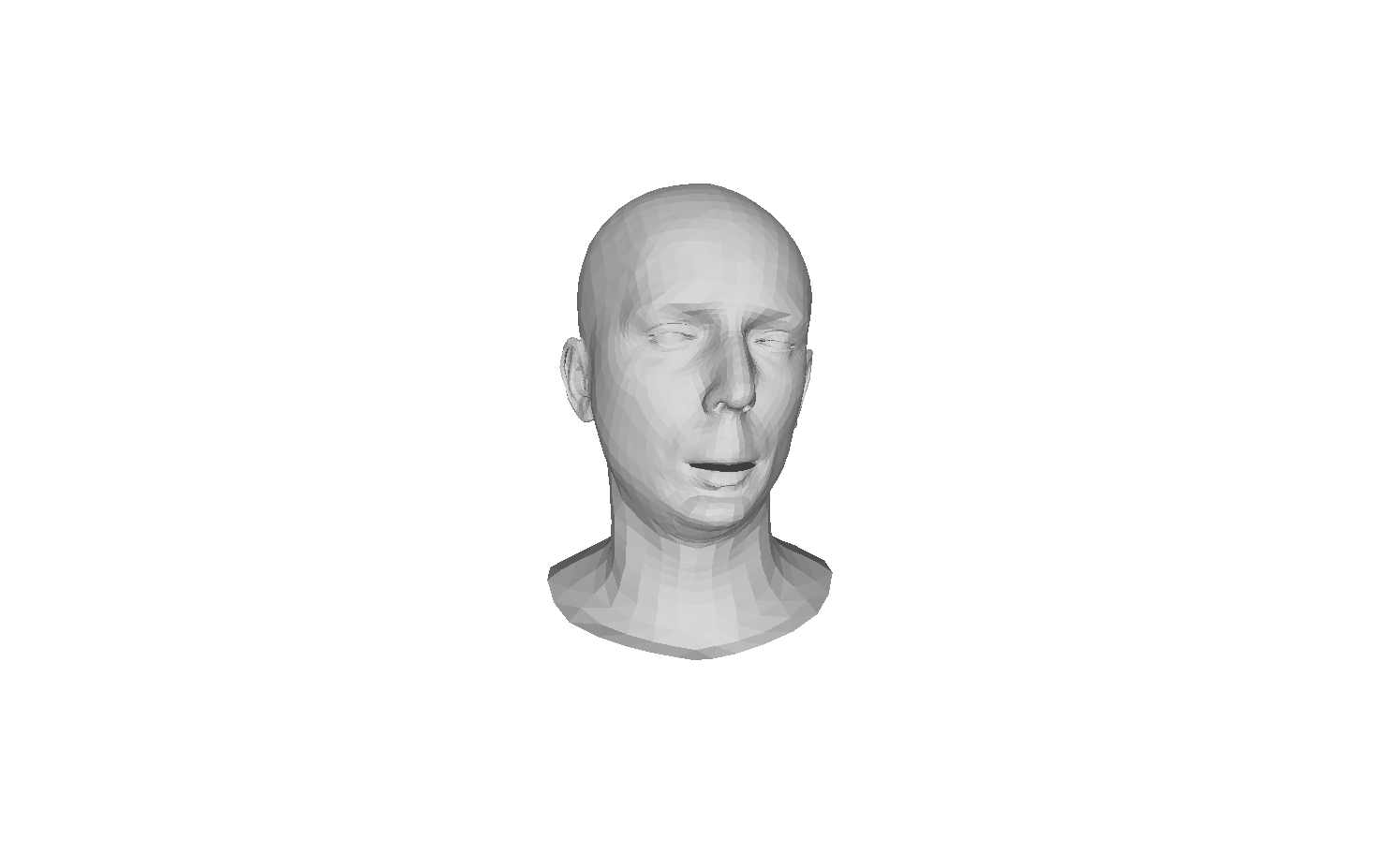}};
    \node[right of=c8, node distance=1.4cm] (c9) {\includegraphics[trim={400 80 400 100},clip,width=0.09\linewidth]{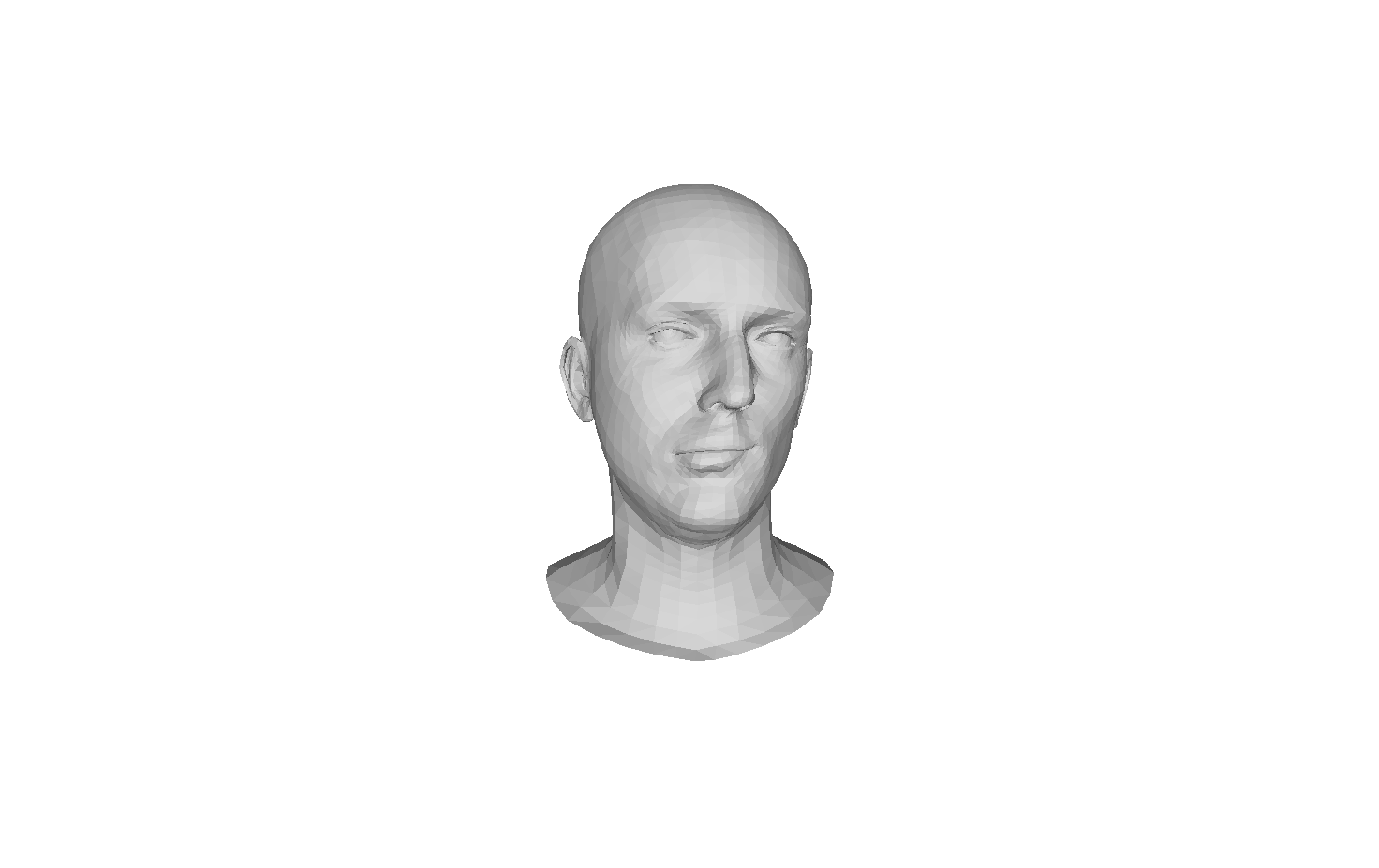}};
    \node[right of=c9, node distance=1.4cm] (c10) {\includegraphics[trim={400 80 400 100},clip,width=0.09\linewidth]{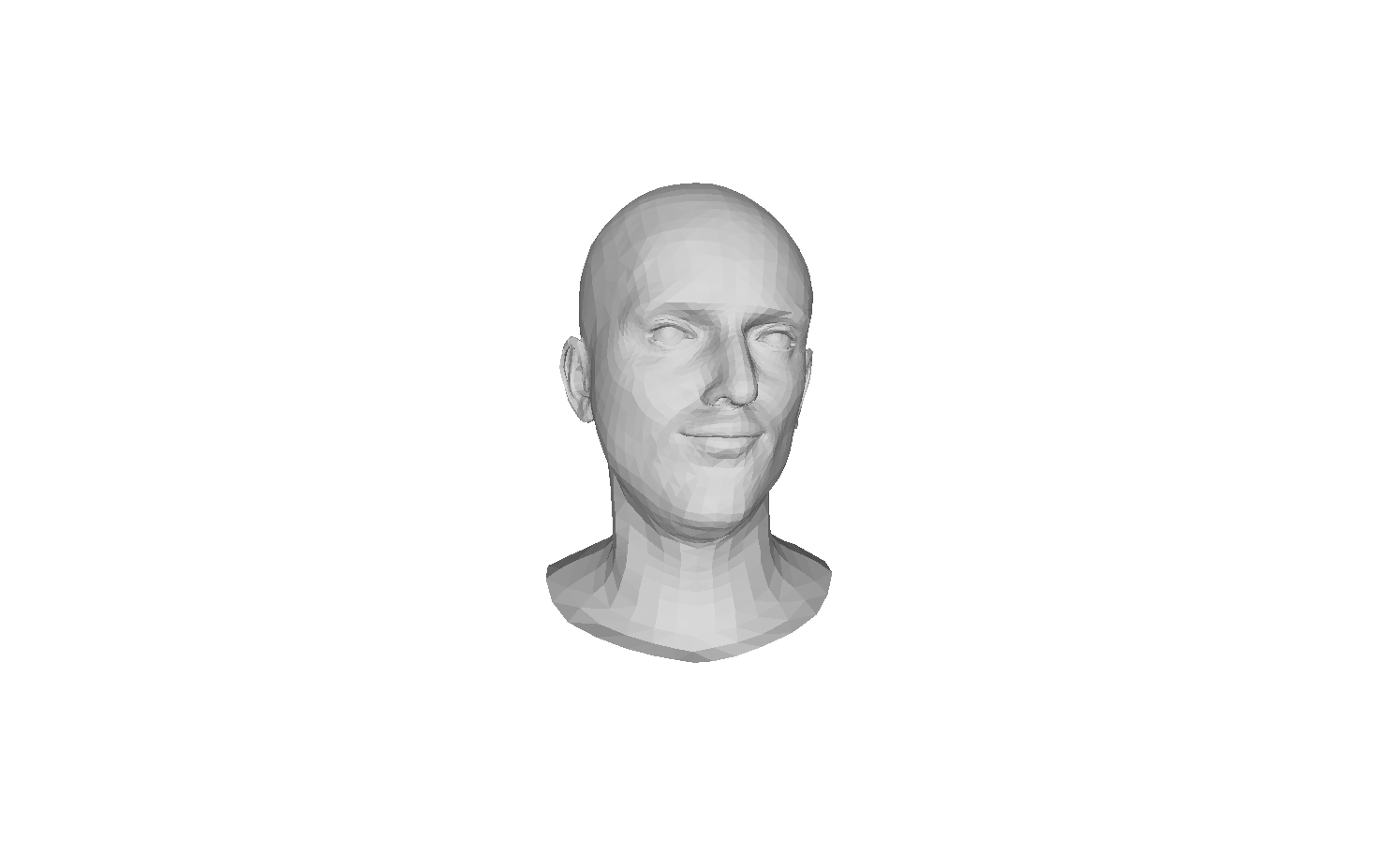}};
    \node[right of=c10, node distance=1.4cm] (c11) {\includegraphics[trim={400 80 400 100},clip,width=0.09\linewidth]{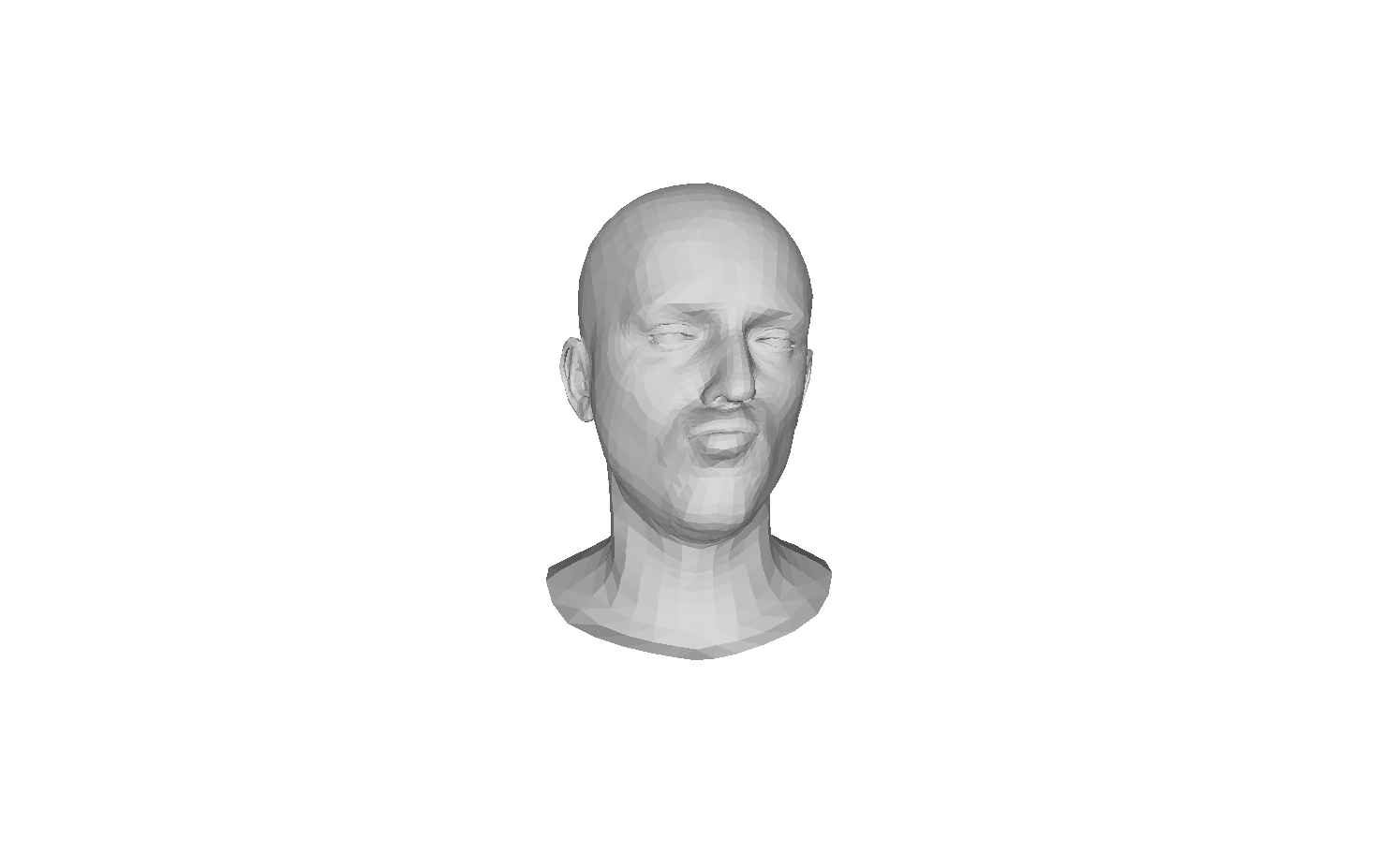}};
    
    \node[below of=c1, node distance=2.2cm] (d1) {\includegraphics[width=0.09\linewidth]{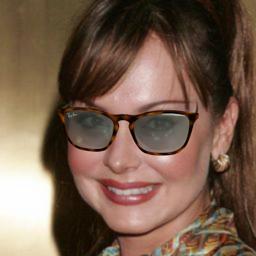}};
    \node[right of=d1, node distance=2.0cm] (d2) {\includegraphics[trim={400 80 400 100},clip,width=0.09\linewidth]{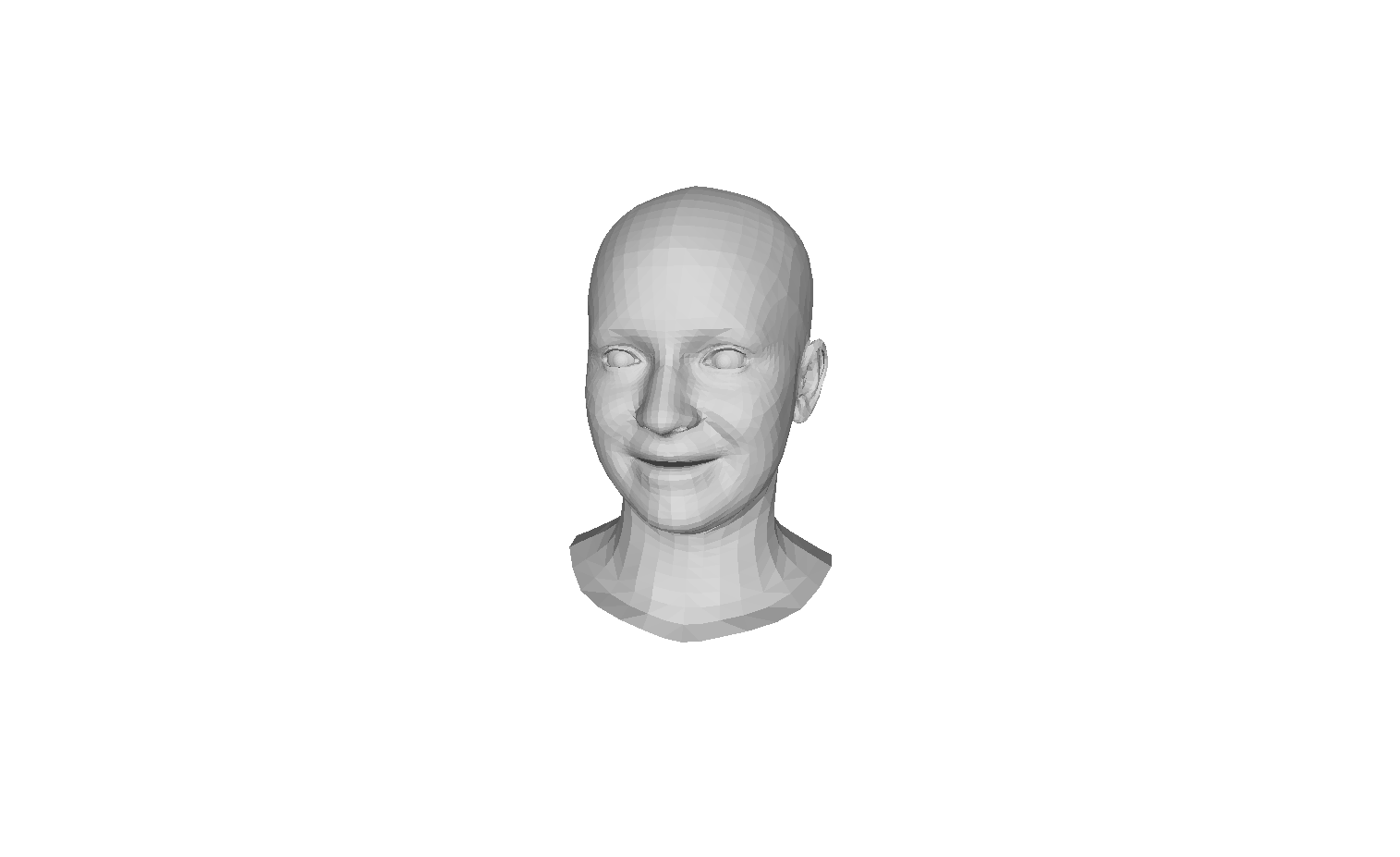}};
    \node[right of=d2, node distance=1.5cm] (d3) {\includegraphics[trim={400 80 400 100},clip,width=0.09\linewidth]{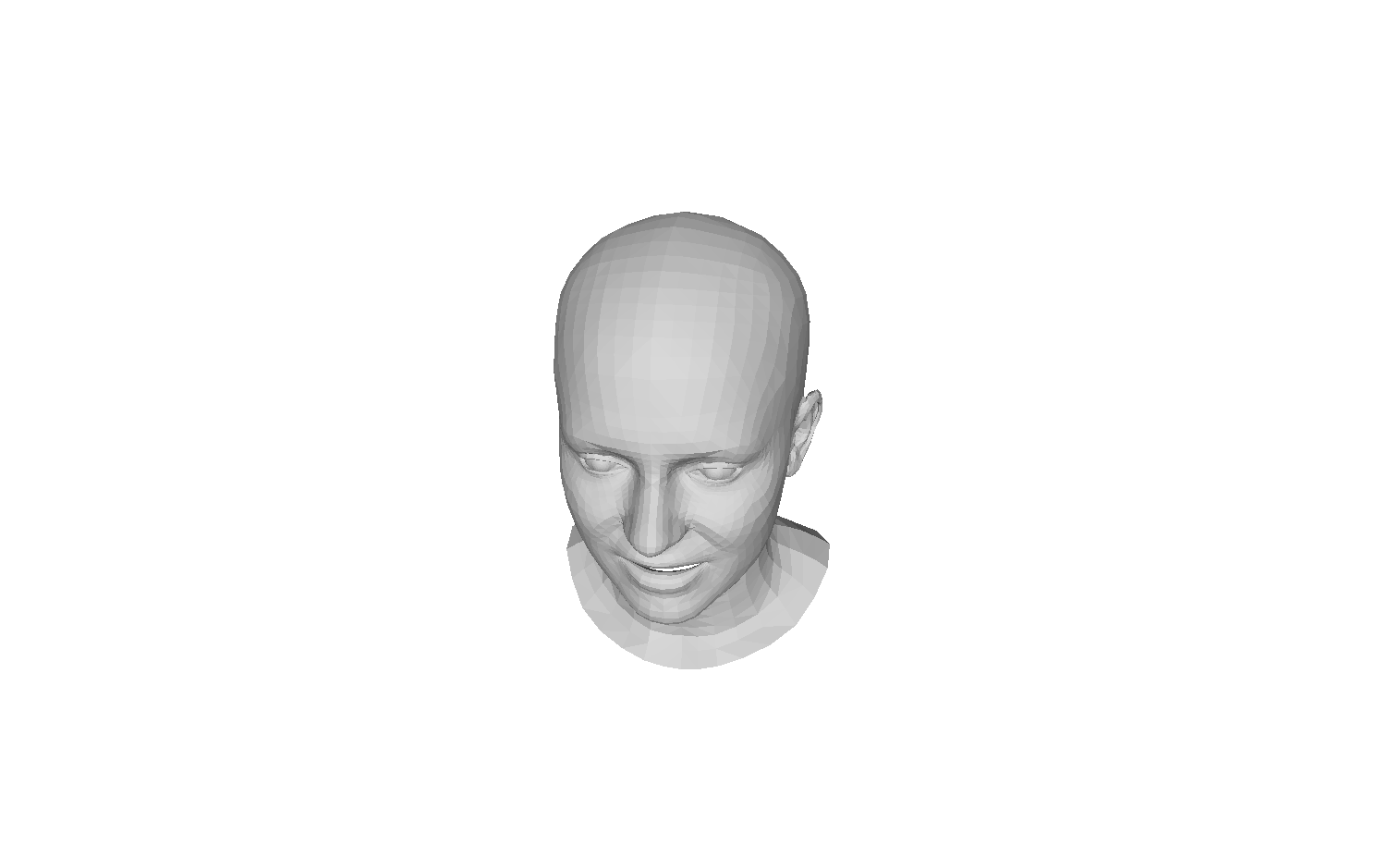}};
    \node[right of=d3, node distance=1.4cm] (d4) {\includegraphics[trim={400 80 400 100},clip,width=0.075\linewidth]{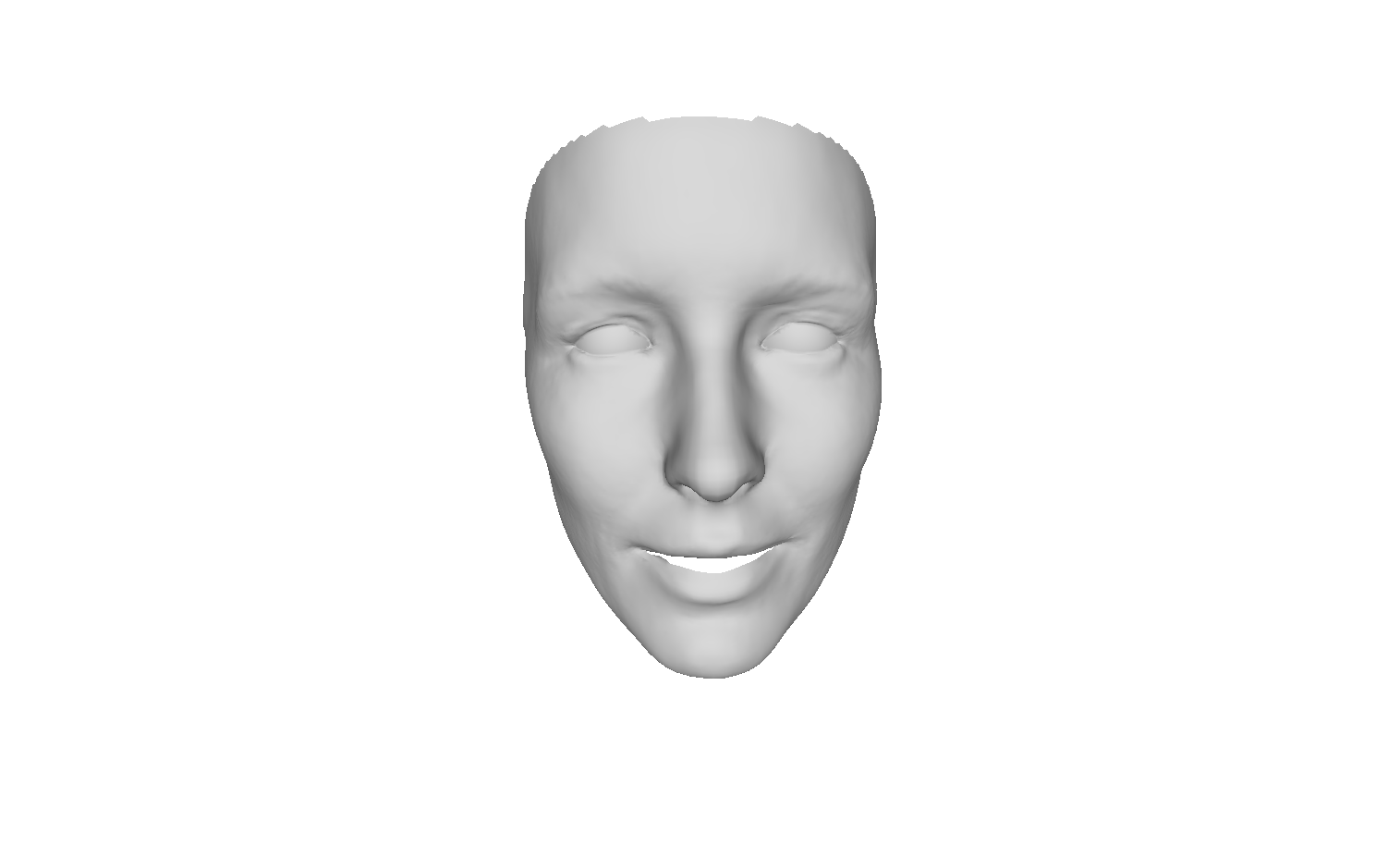}};
    \node[right of=d4, node distance=1.4cm] (d5) {\includegraphics[trim={400 80 400 100},clip,width=0.075\linewidth]{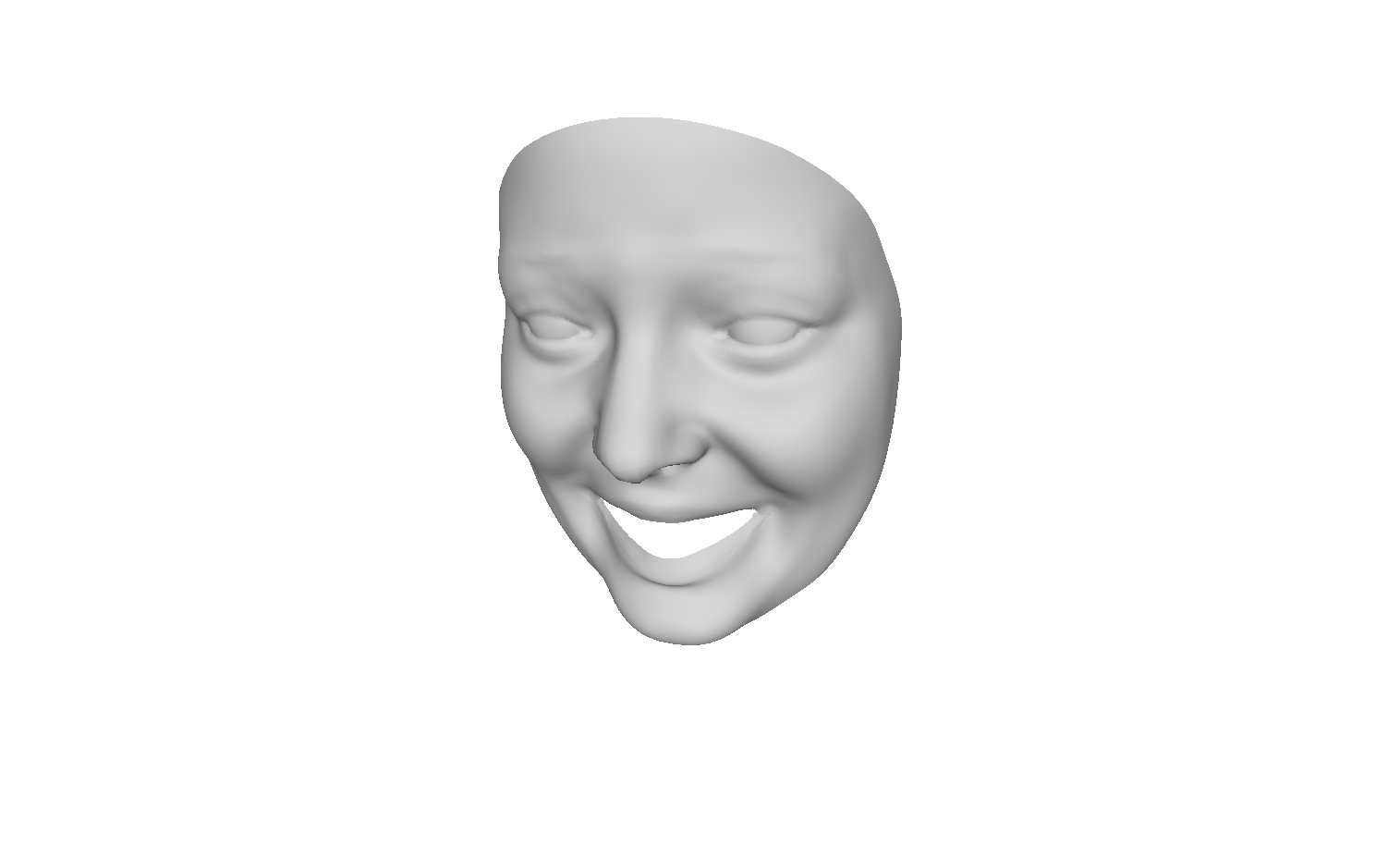}};
    \node[right of=d5, node distance=1.5cm] (d6) {\includegraphics[trim={350 80 400 100},clip,width=0.085\linewidth]{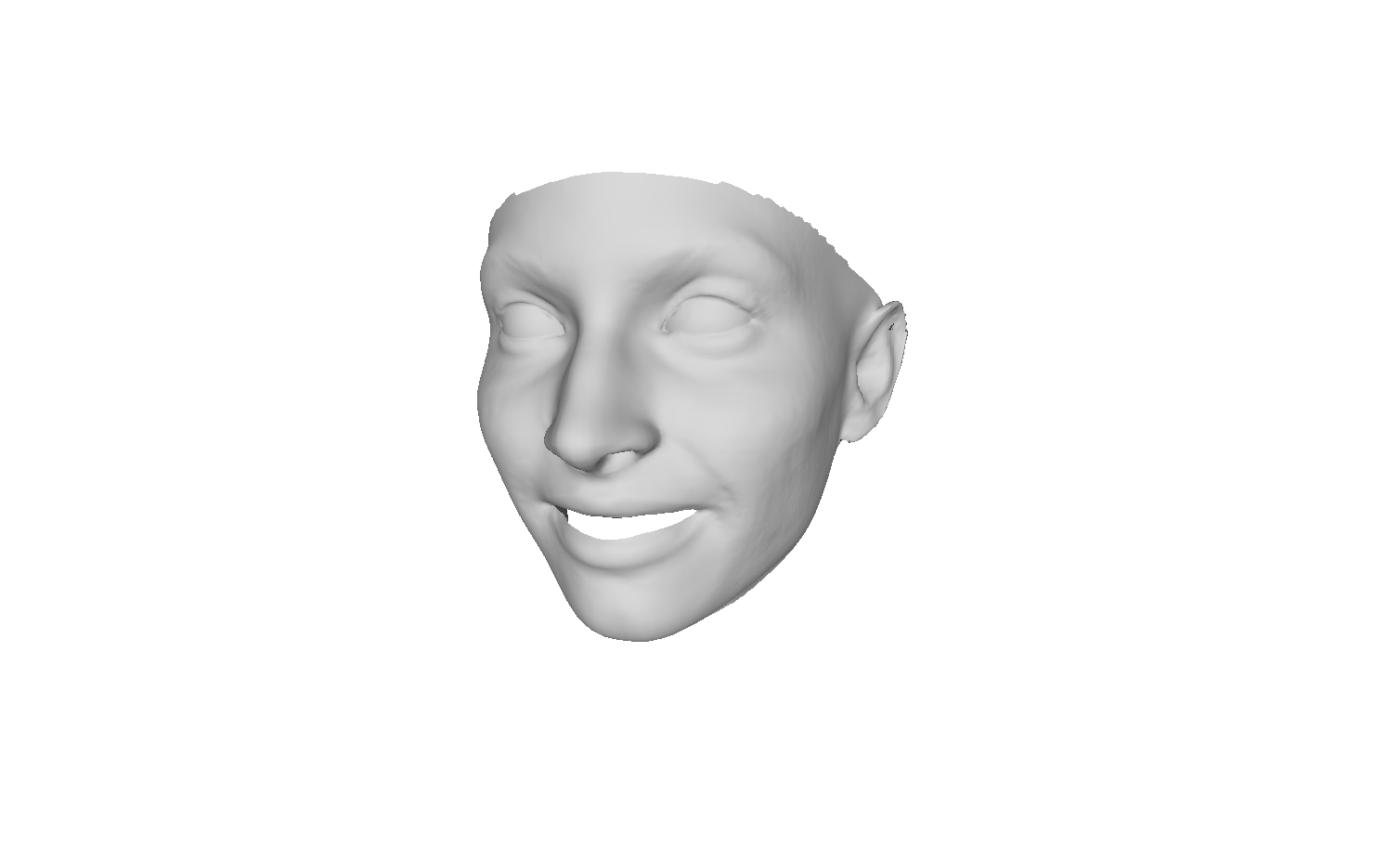}};
    \node[right of=d6, node distance=2.1cm] (d7) {\includegraphics[trim={400 80 400 100},clip,width=0.09\linewidth]{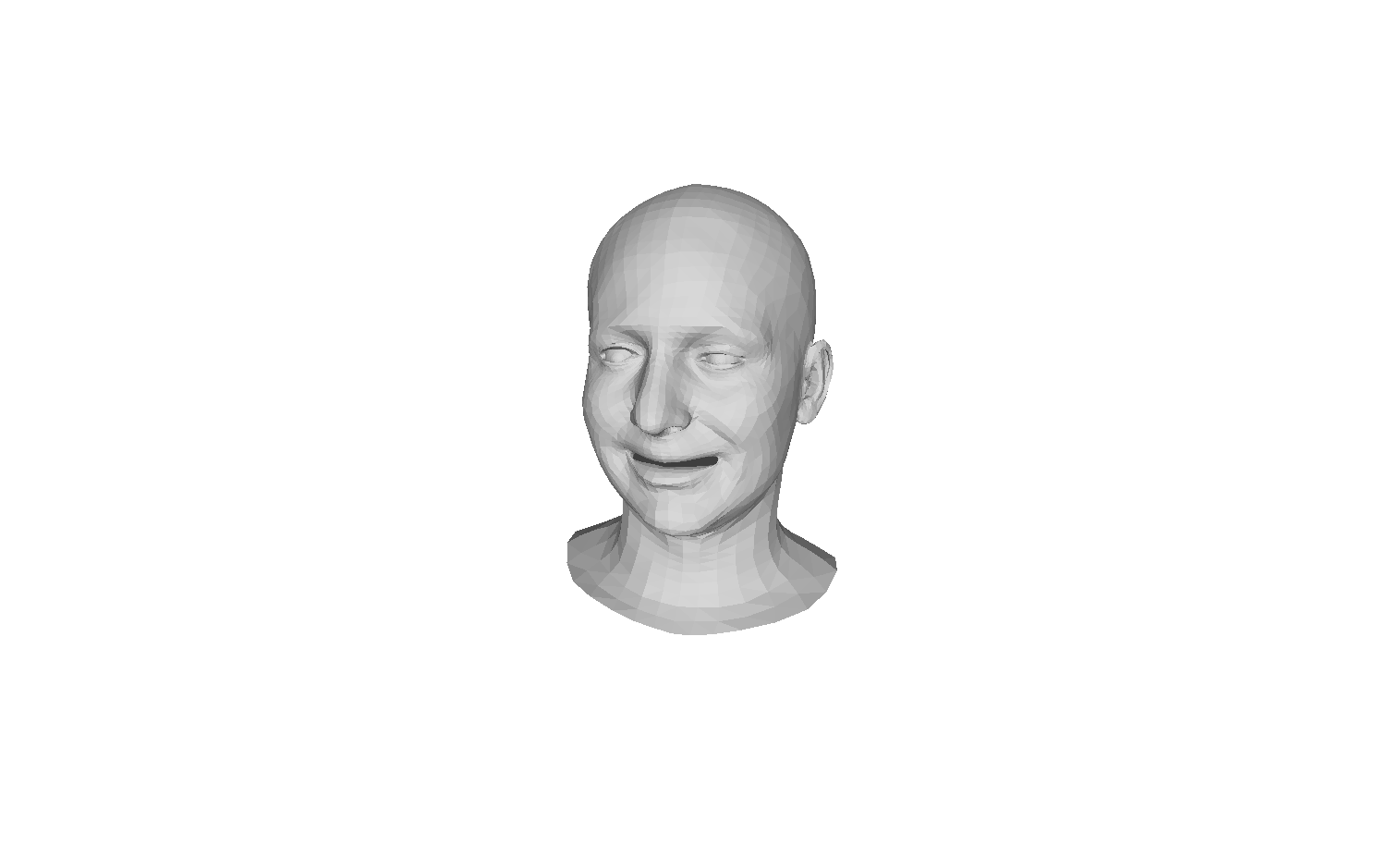}};
    \node[right of=d7, node distance=1.4cm] (d8) {\includegraphics[trim={400 80 400 100},clip,width=0.09\linewidth]{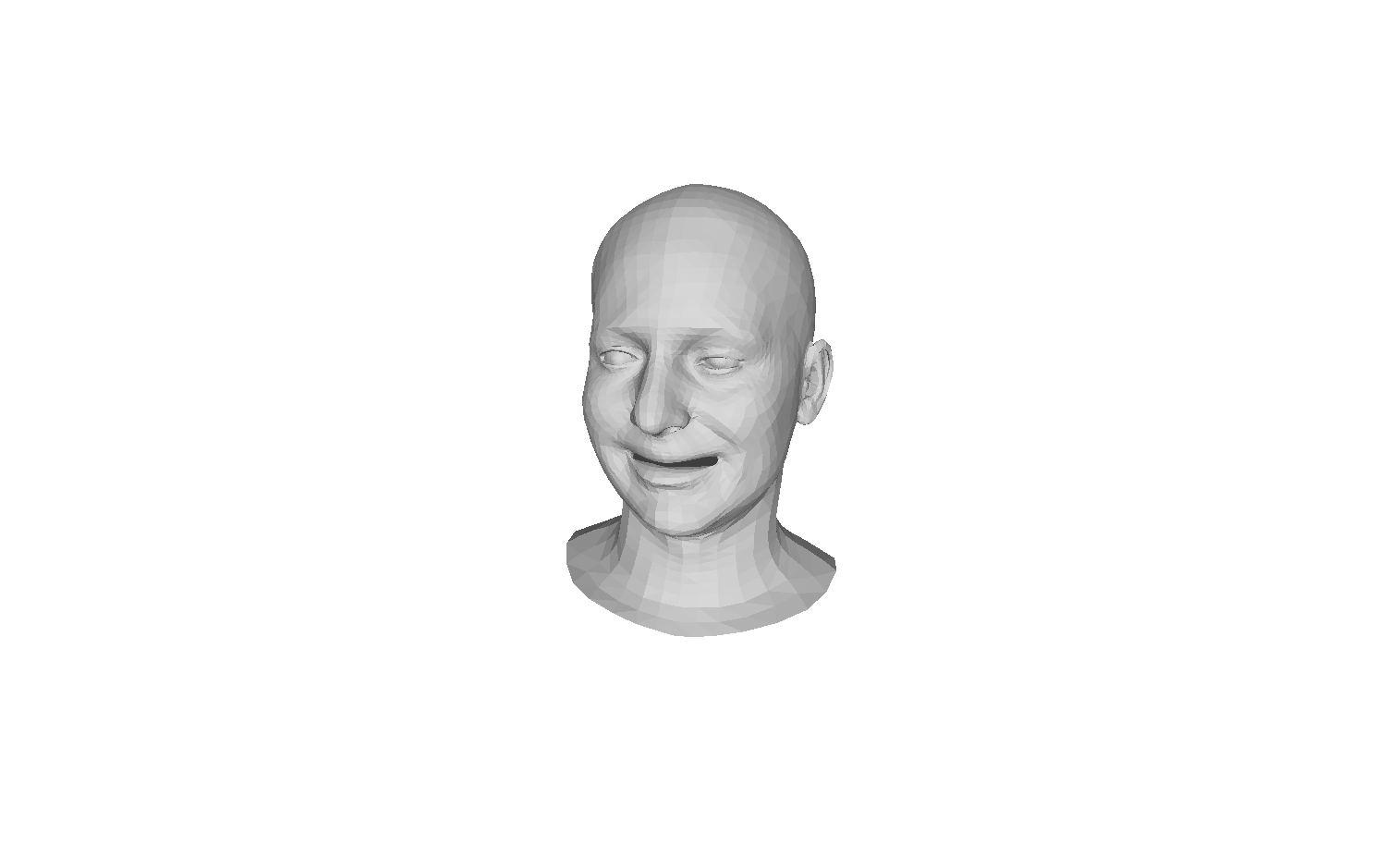}};
    \node[right of=d8, node distance=1.4cm] (d9) {\includegraphics[trim={400 80 400 100},clip,width=0.09\linewidth]{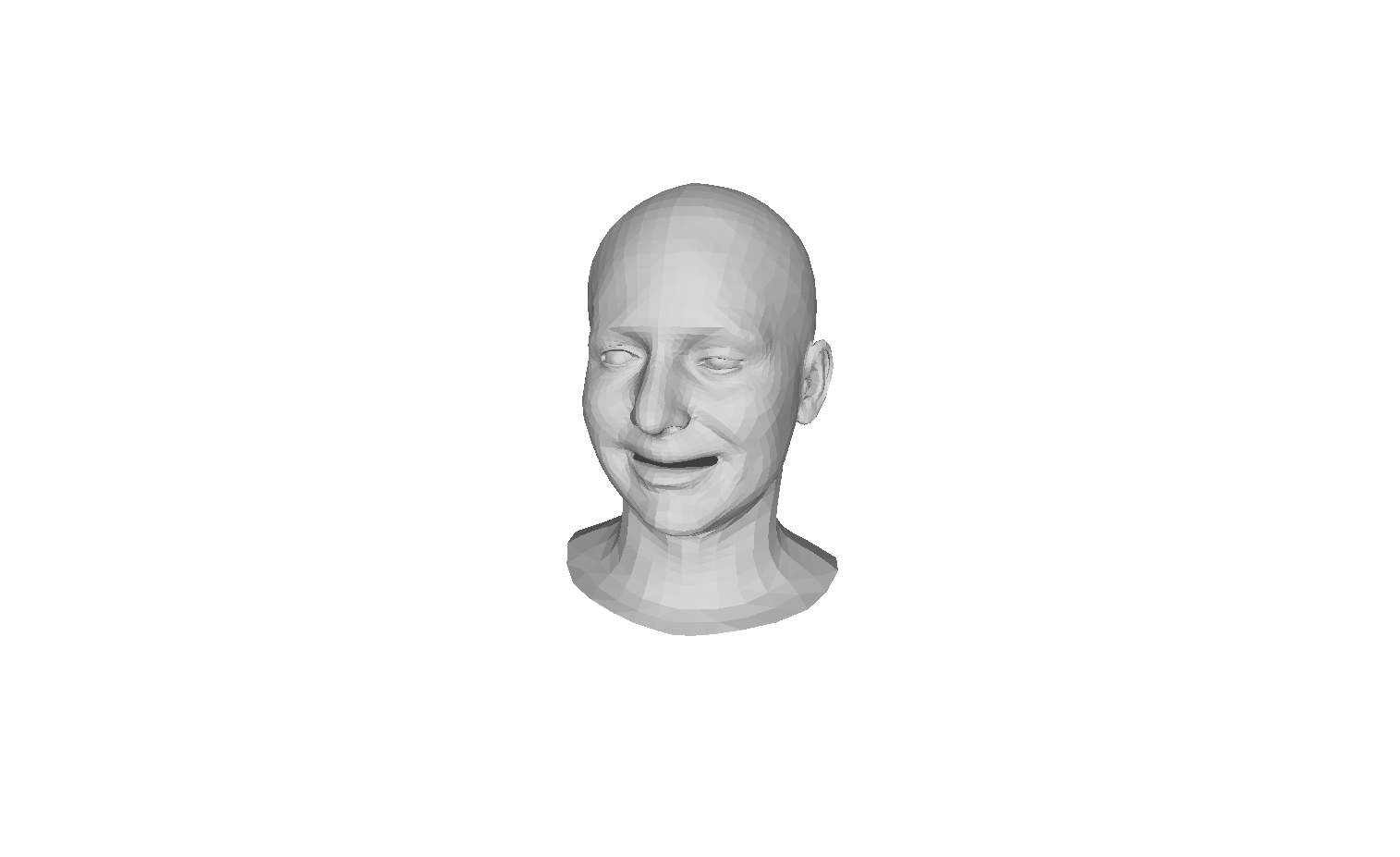}};
    \node[right of=d9, node distance=1.4cm] (d10) {\includegraphics[trim={400 80 400 100},clip,width=0.09\linewidth]{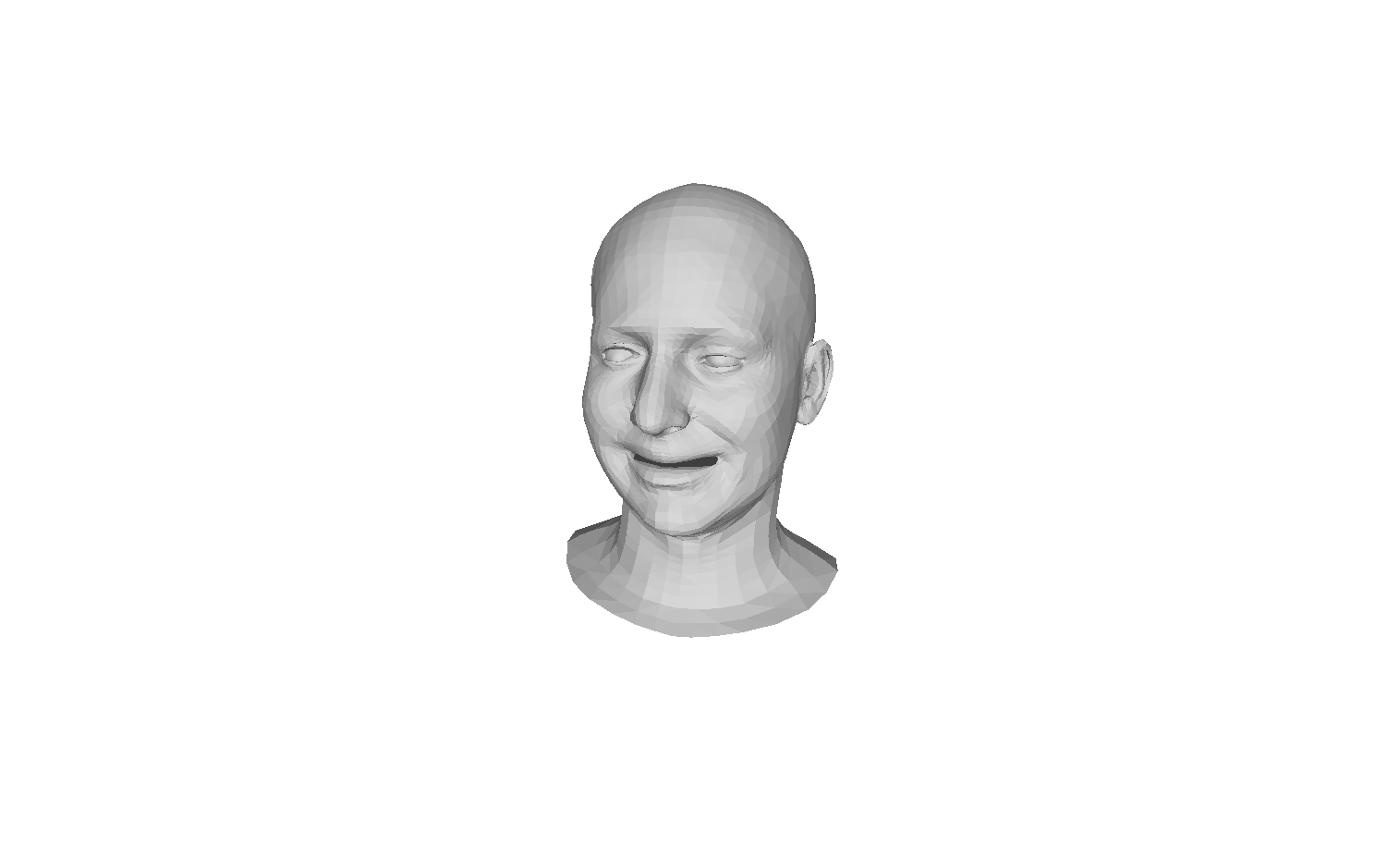}};
    \node[right of=d10, node distance=1.4cm] (d11) {\includegraphics[trim={400 80 400 100},clip,width=0.09\linewidth]{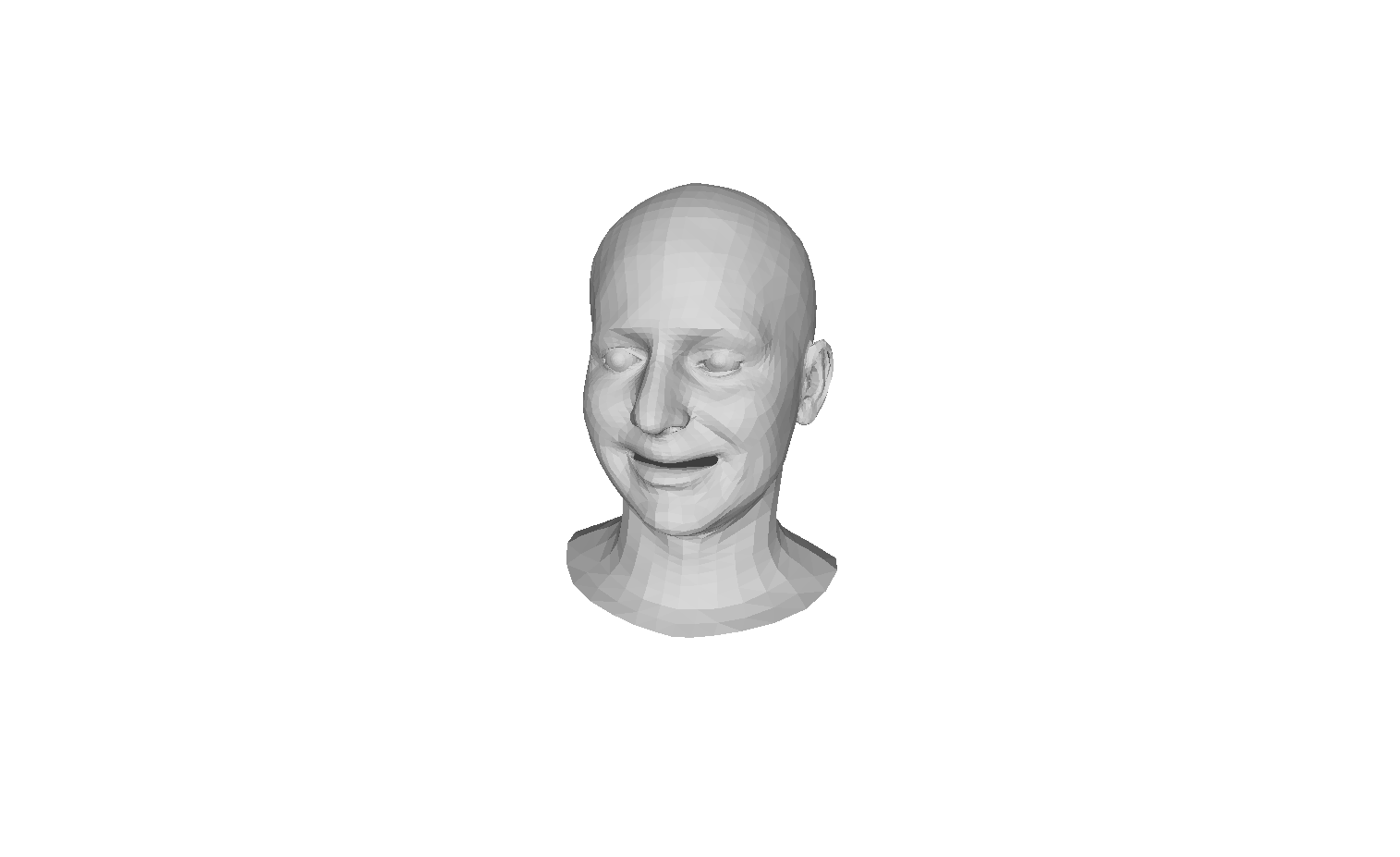}};
    
    \node[below of=d1, node distance=2.2cm] (e1) {\includegraphics[width=0.09\linewidth]{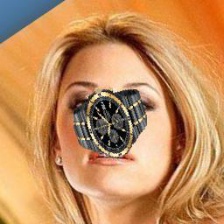}};
    \node[right of=e1, node distance=2.0cm] (e2) {\includegraphics[trim={400 80 400 100},clip,width=0.09\linewidth]{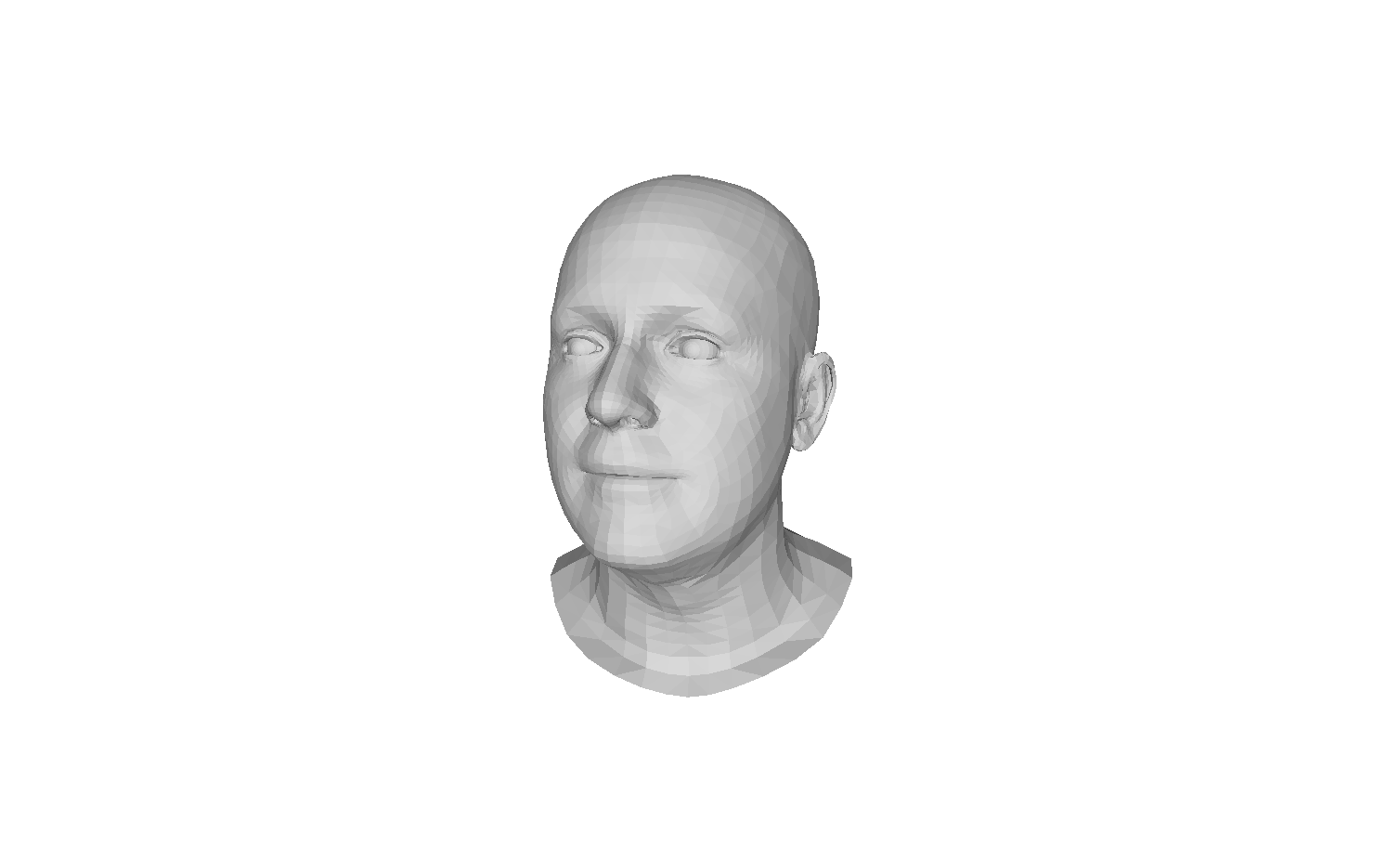}};
    \node[right of=e2, node distance=1.5cm] (e3) {\includegraphics[trim={400 80 400 100},clip,width=0.09\linewidth]{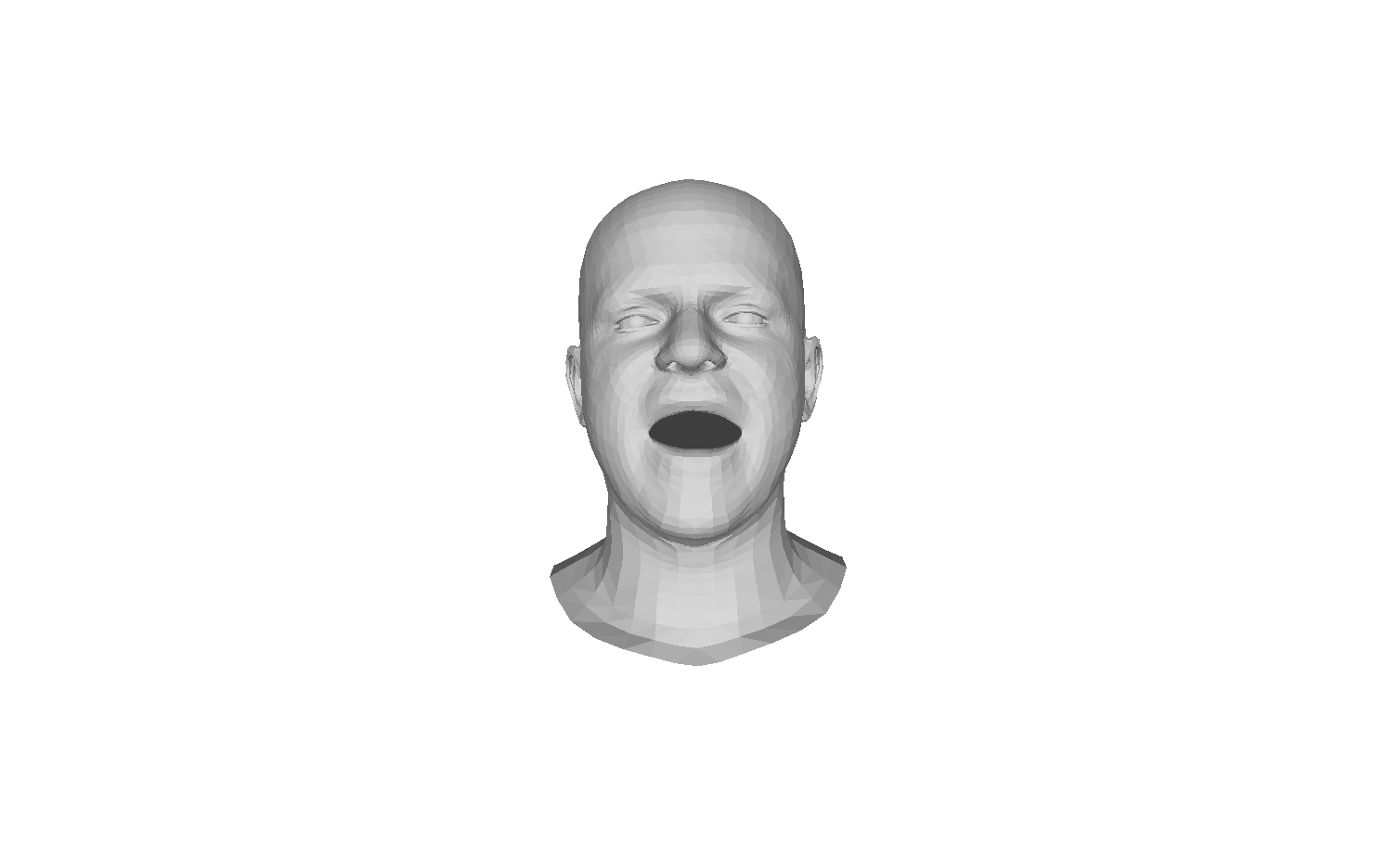}};
    \node[right of=e3, node distance=1.4cm] (e4) {\includegraphics[trim={400 80 400 100},clip,width=0.075\linewidth]{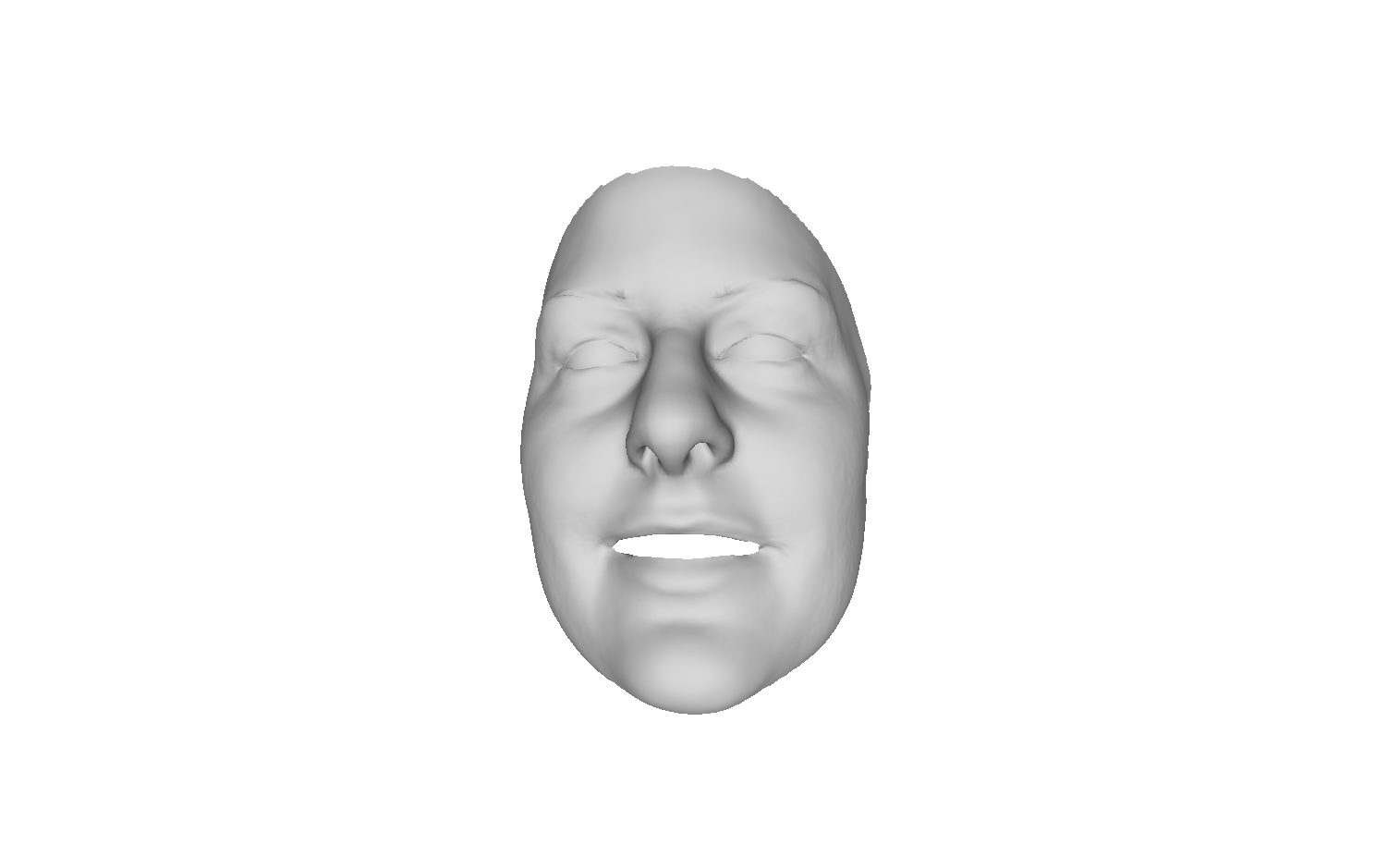}};
    \node[right of=e4, node distance=1.4cm] (e5) {\includegraphics[trim={400 80 400 100},clip,width=0.075\linewidth]{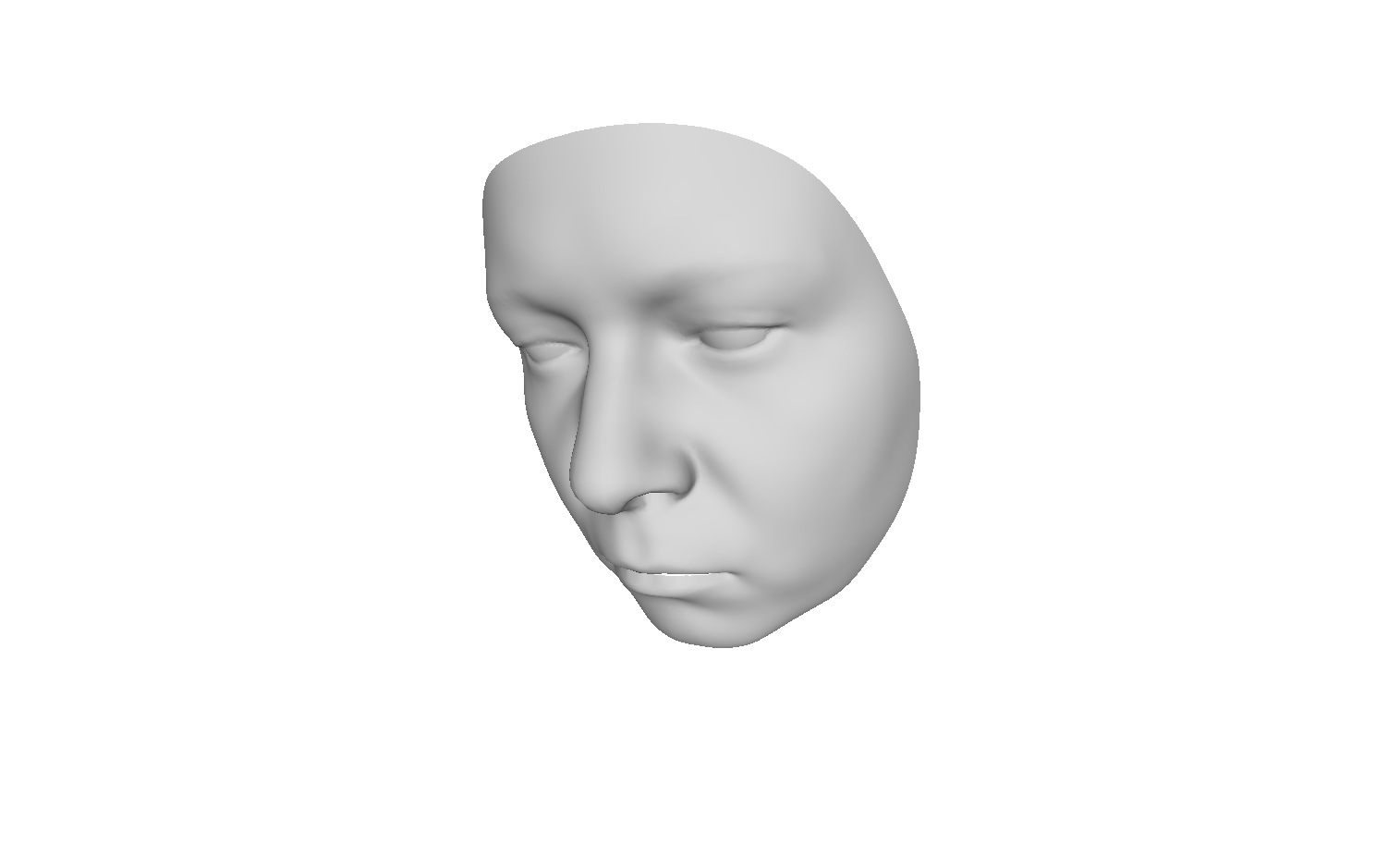}};
    \node[right of=e5, node distance=1.5cm] (e6) {\includegraphics[trim={350 80 400 100},clip,width=0.085\linewidth]{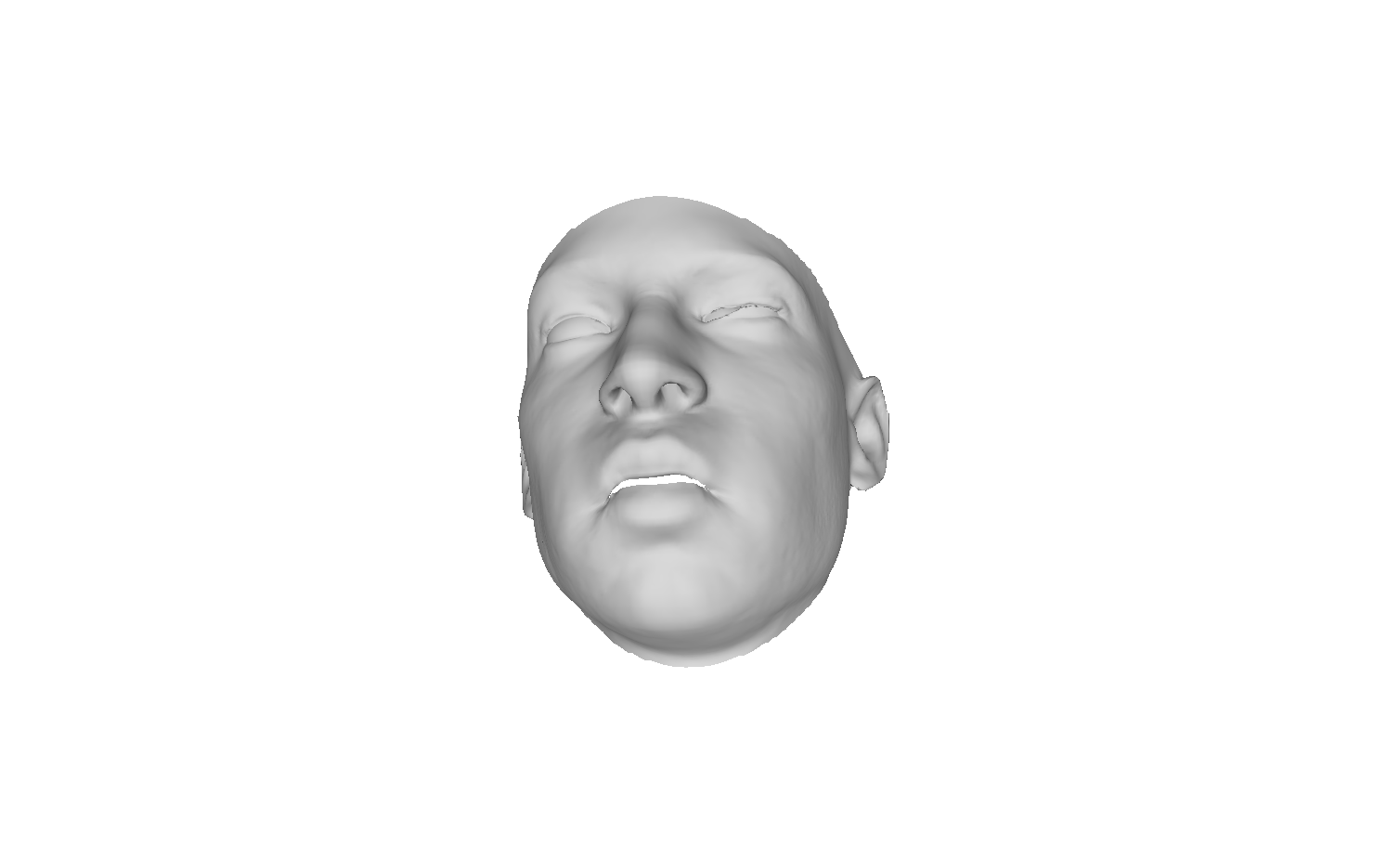}};
    \node[right of=e6, node distance=2.1cm] (e7) {\includegraphics[trim={400 80 400 100},clip,width=0.09\linewidth]{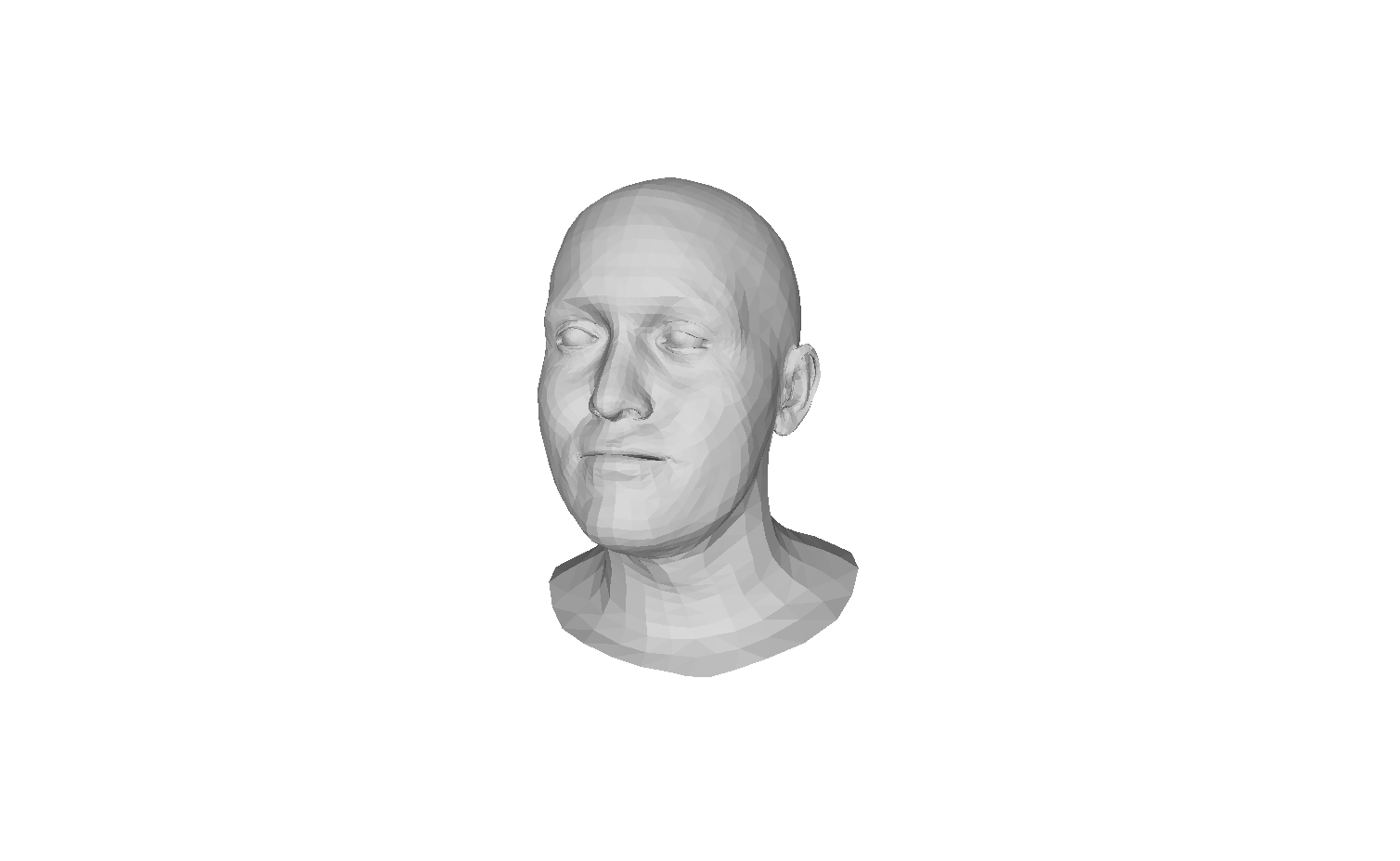}};
    \node[right of=e7, node distance=1.4cm] (e8) {\includegraphics[trim={400 80 400 100},clip,width=0.09\linewidth]{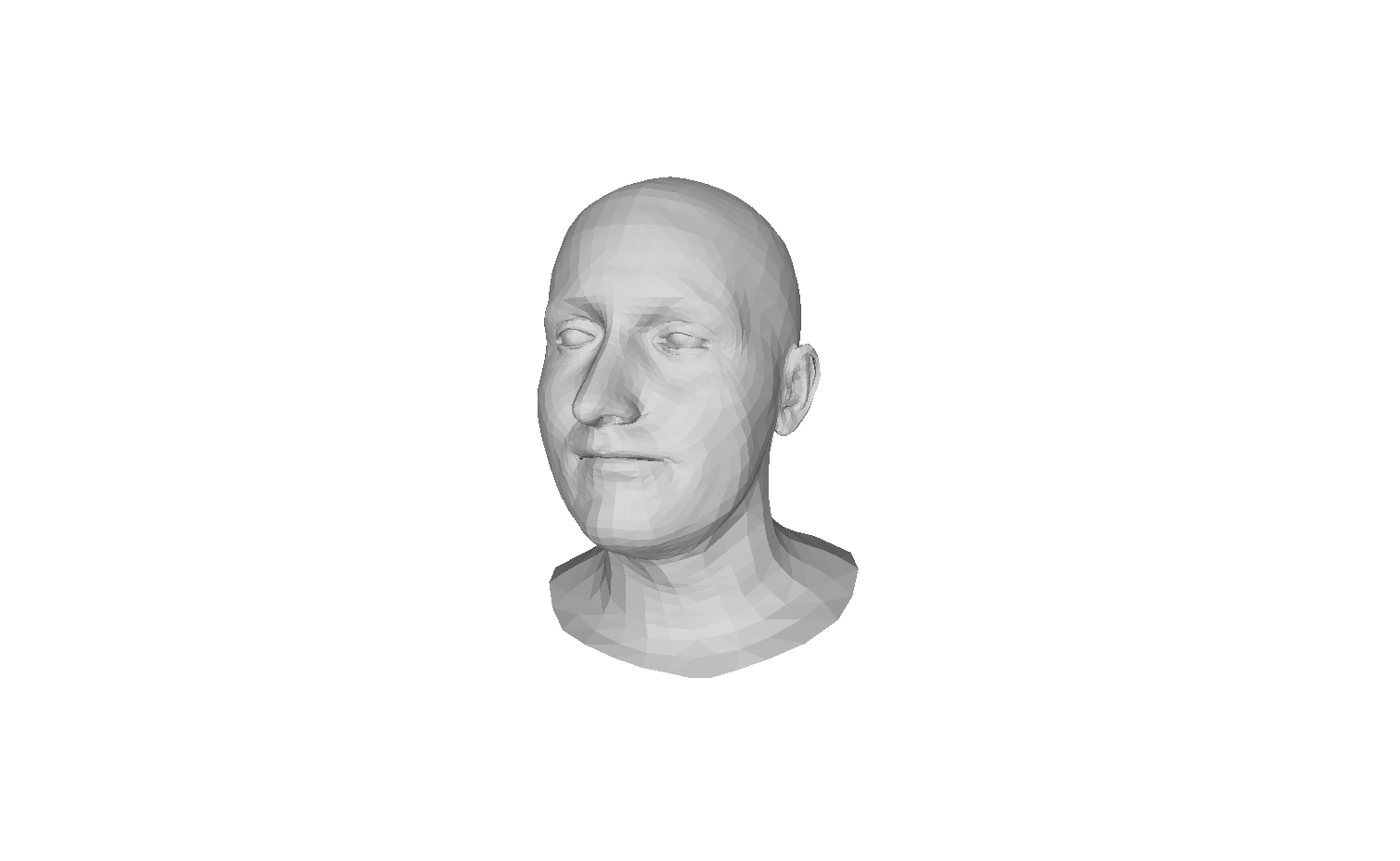}};
    \node[right of=e8, node distance=1.4cm] (e9) {\includegraphics[trim={400 80 400 100},clip,width=0.09\linewidth]{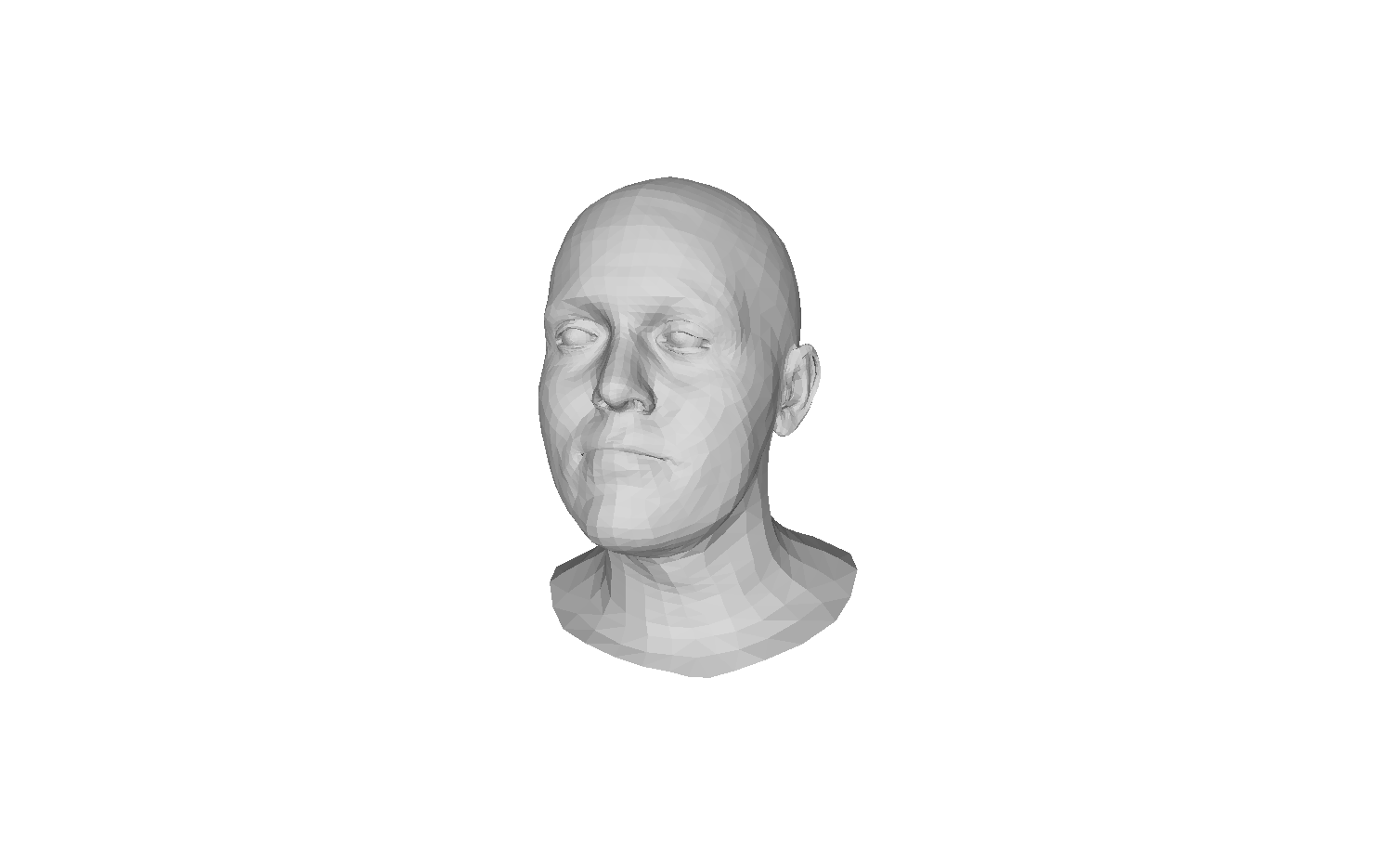}};
    \node[right of=e9, node distance=1.4cm] (e10) {\includegraphics[trim={400 80 400 100},clip,width=0.09\linewidth]{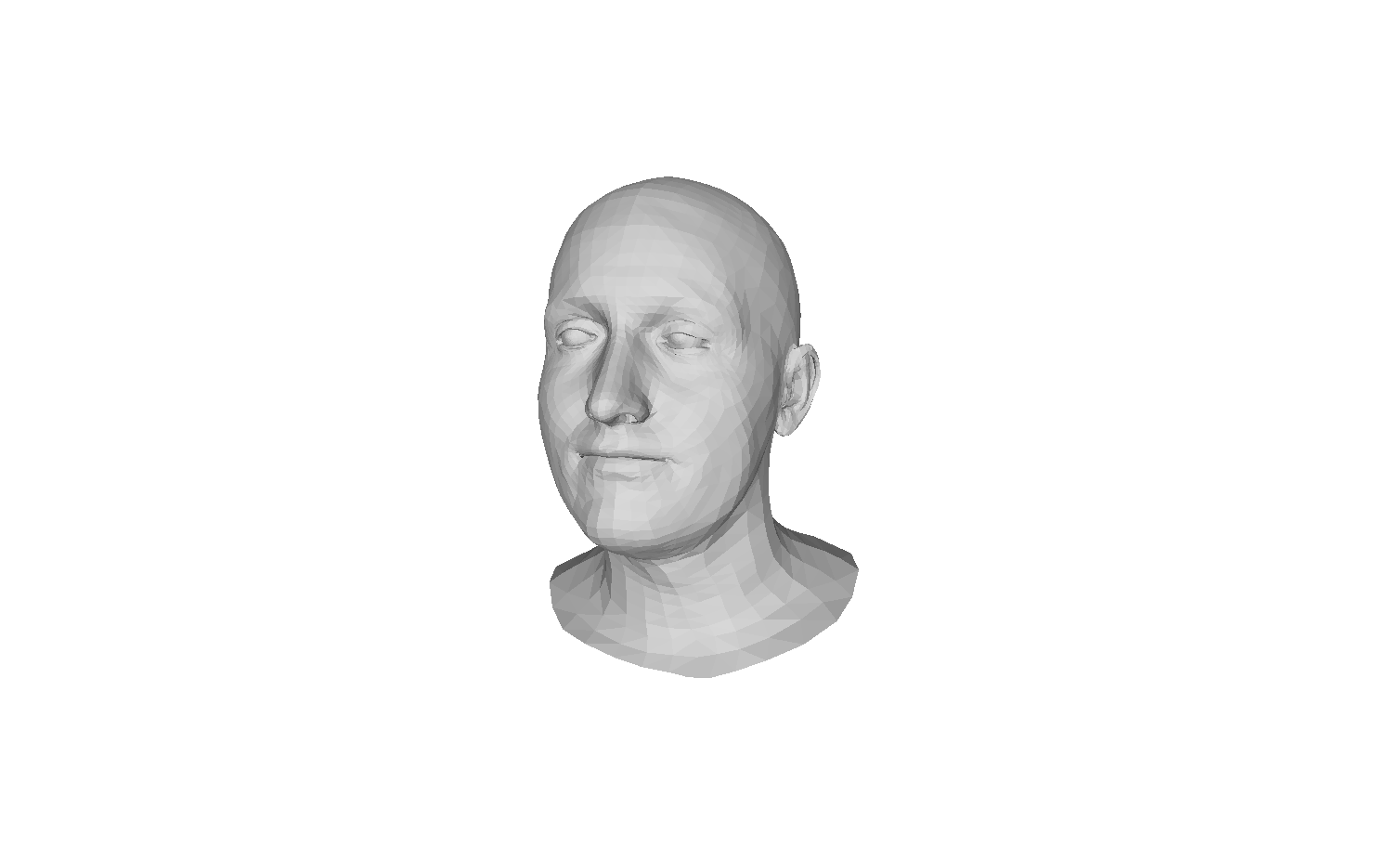}};
    \node[right of=e10, node distance=1.4cm] (e11) {\includegraphics[trim={400 80 400 100},clip,width=0.09\linewidth]{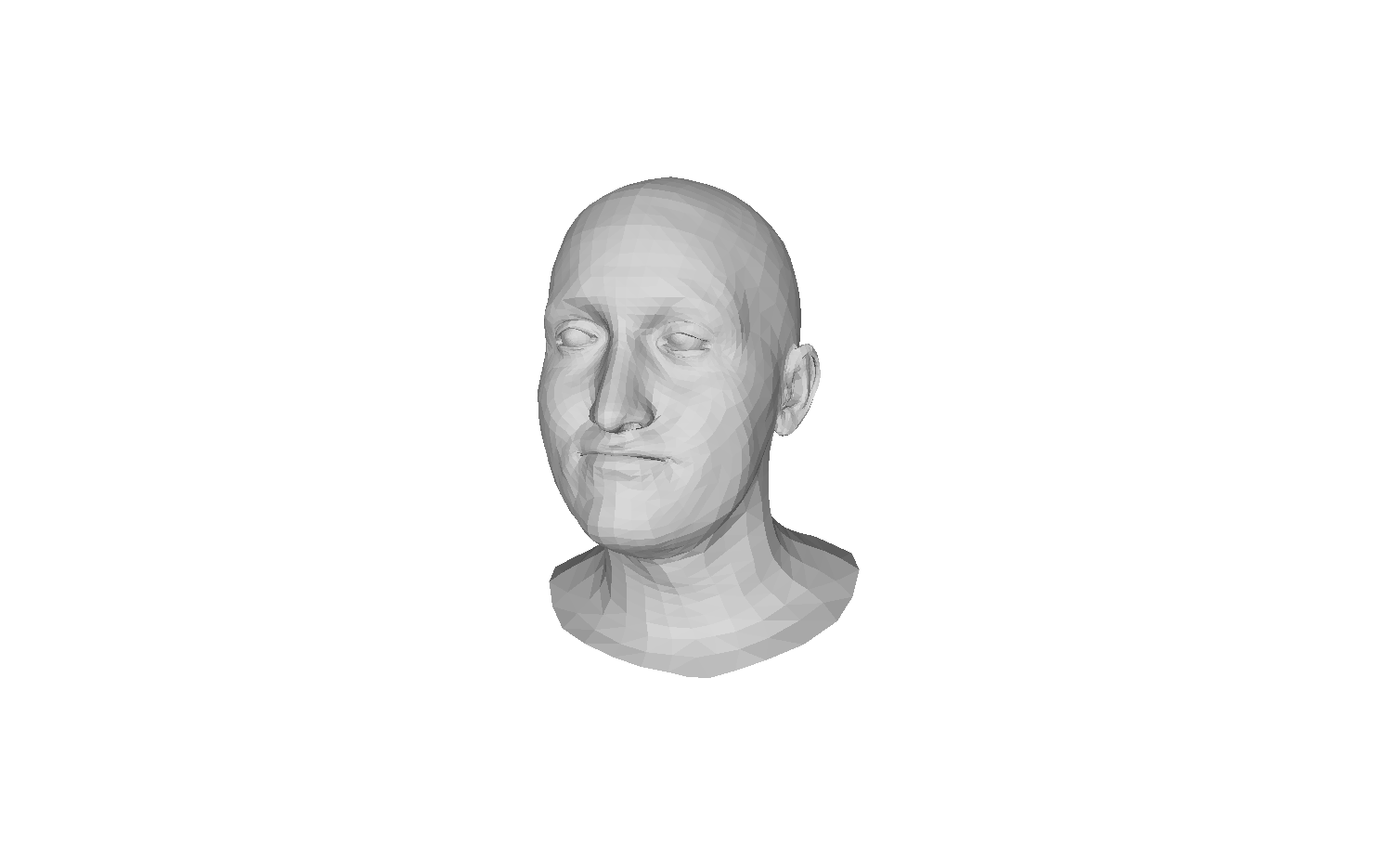}};
    
    \node[below of=e1, node distance=2.2cm] (f1) {\includegraphics[width=0.09\linewidth]{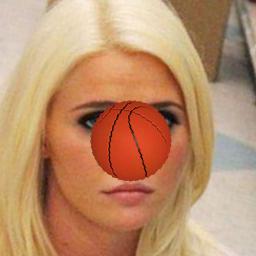}};
    \node[right of=f1, node distance=2.0cm] (f2) {\includegraphics[trim={400 80 400 100},clip,width=0.09\linewidth]{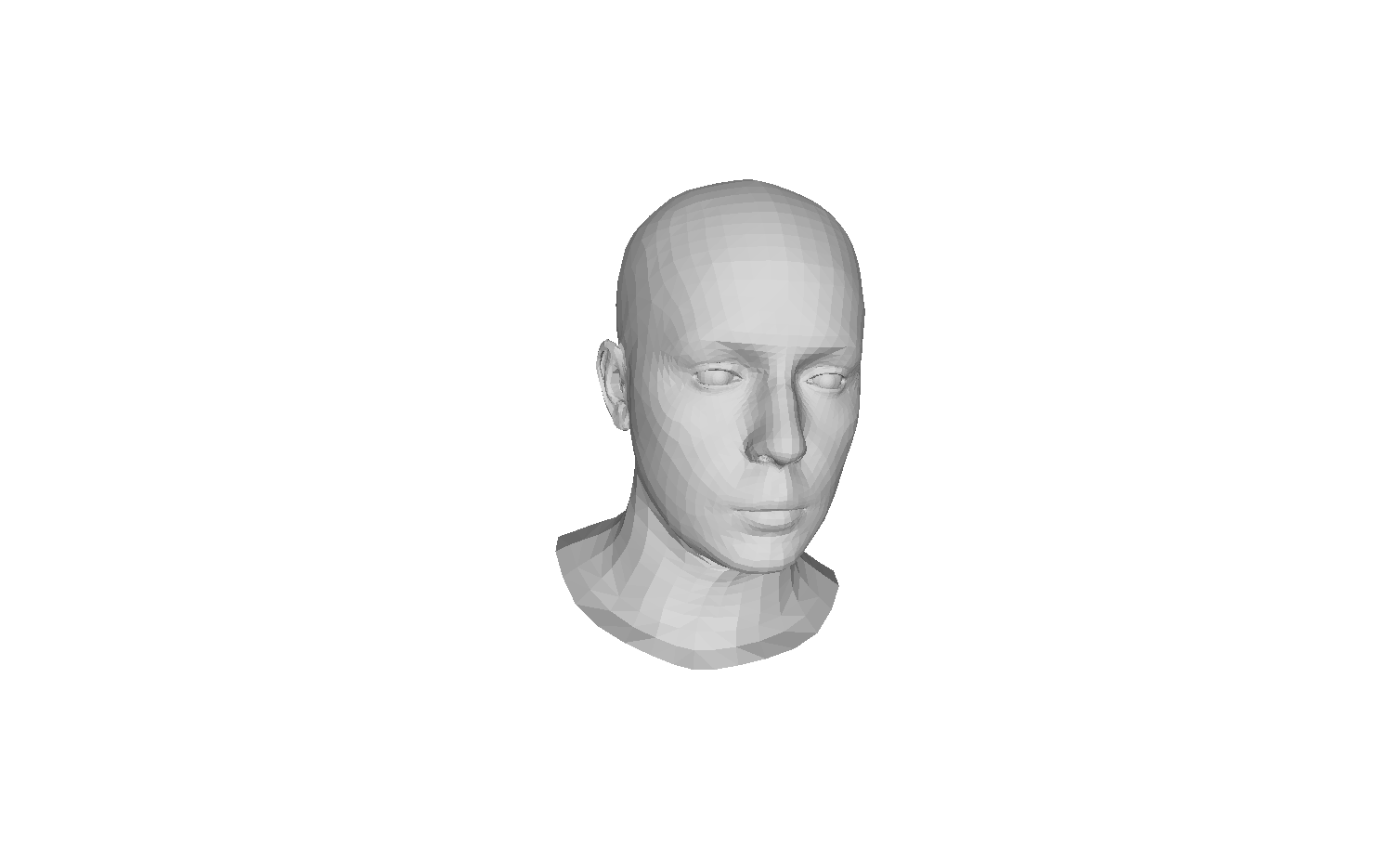}};
    \node[right of=f2, node distance=1.5cm] (f3) {\includegraphics[trim={400 80 400 100},clip,width=0.09\linewidth]{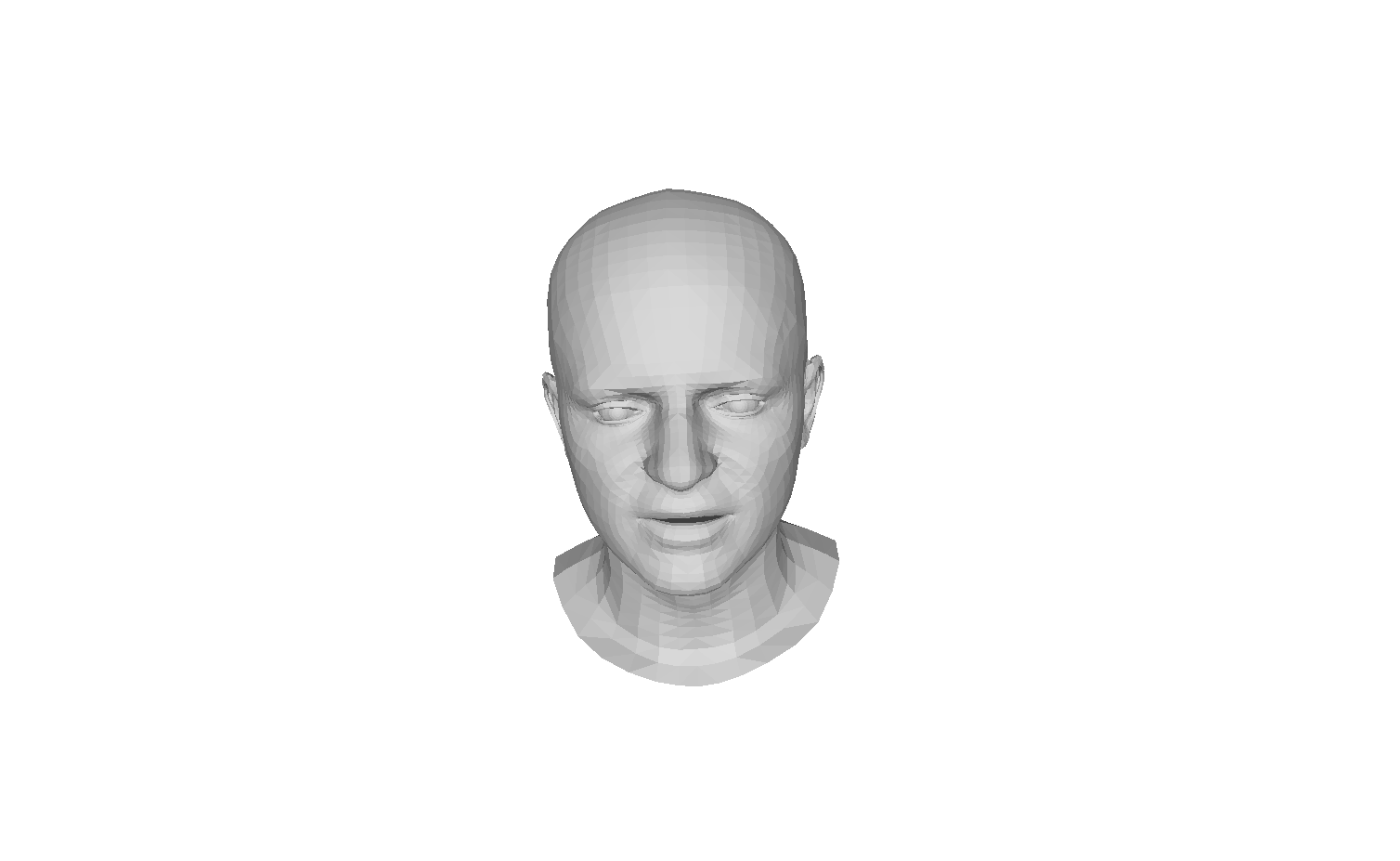}};
    \node[right of=f3, node distance=1.4cm] (f4) {\includegraphics[trim={400 80 400 100},clip,width=0.075\linewidth]{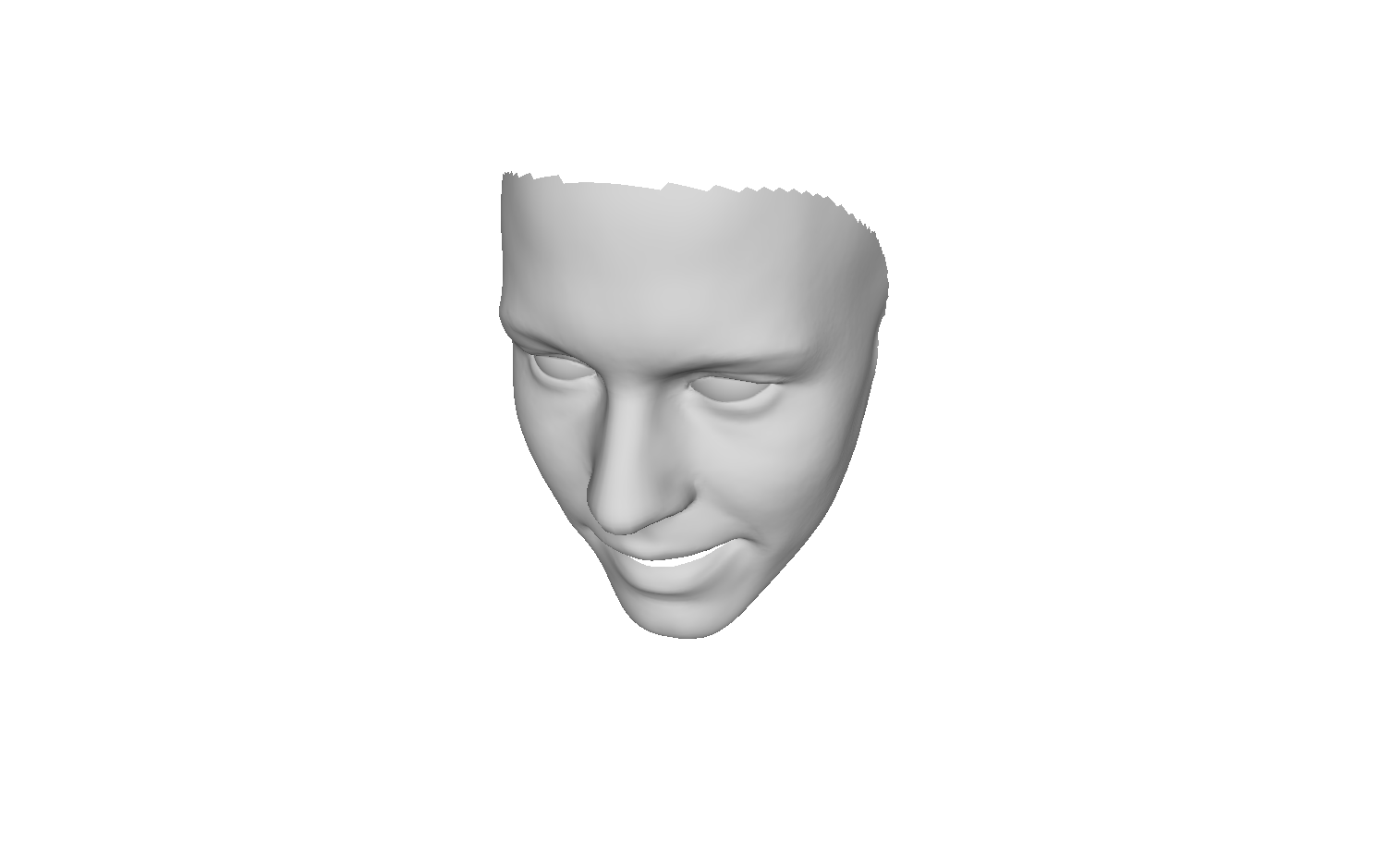}};
    \node[right of=f4, node distance=1.4cm] (f5) {\includegraphics[trim={400 80 400 100},clip,width=0.075\linewidth]{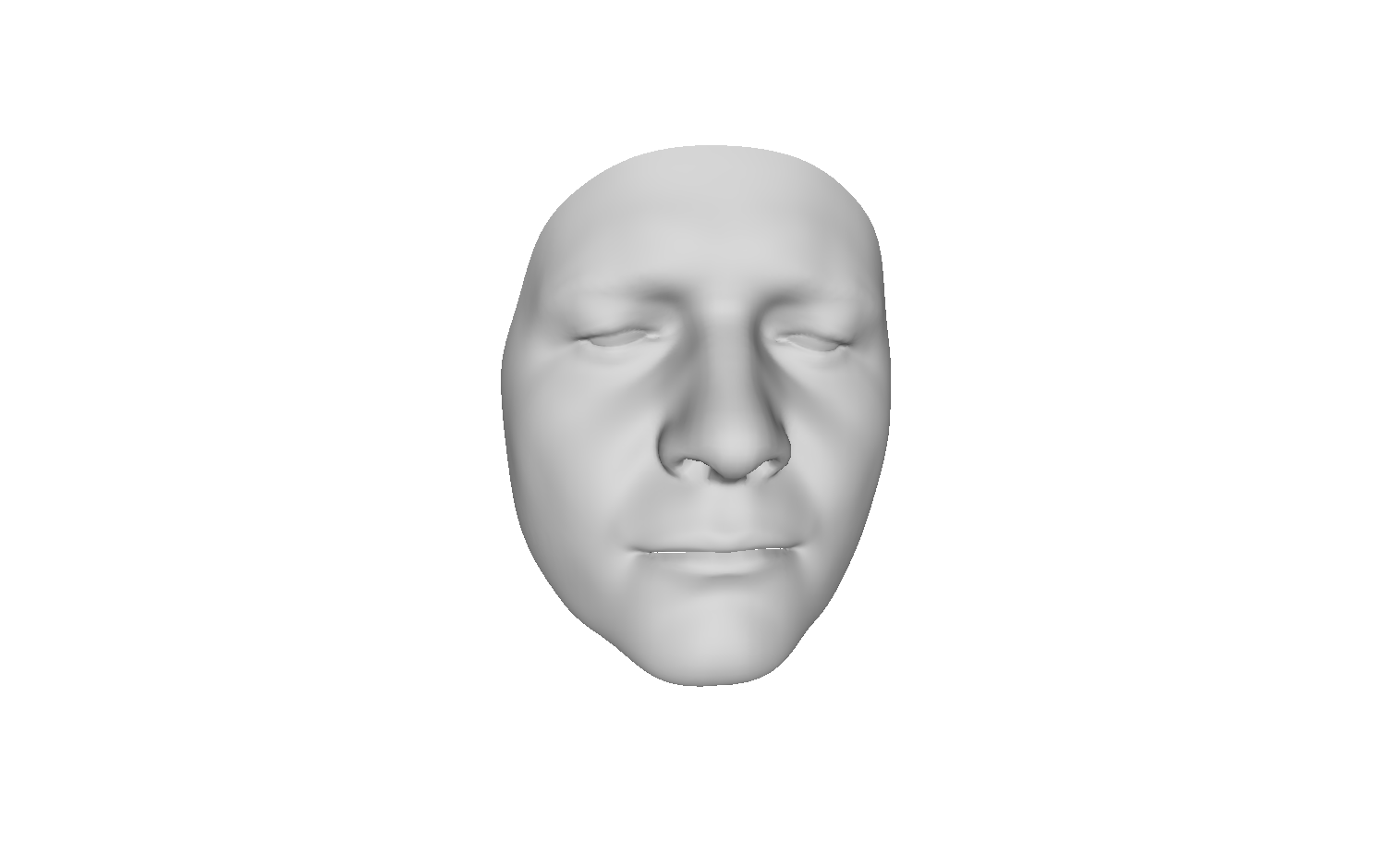}};
    \node[right of=f5, node distance=1.5cm] (f6) {\includegraphics[trim={350 80 400 100},clip,width=0.085\linewidth]{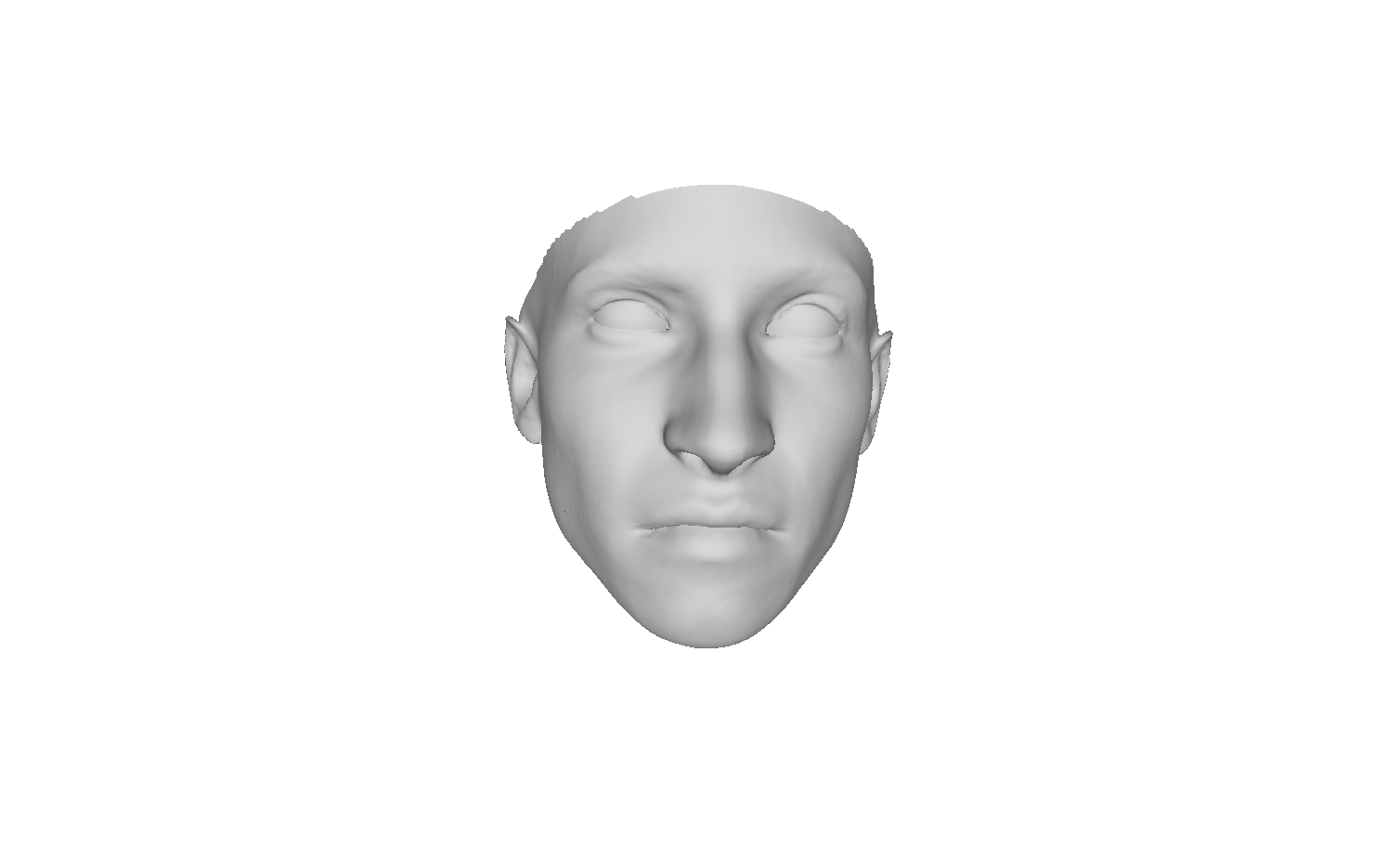}};
    \node[right of=f6, node distance=2.1cm] (f7) {\includegraphics[trim={400 80 400 100},clip,width=0.09\linewidth]{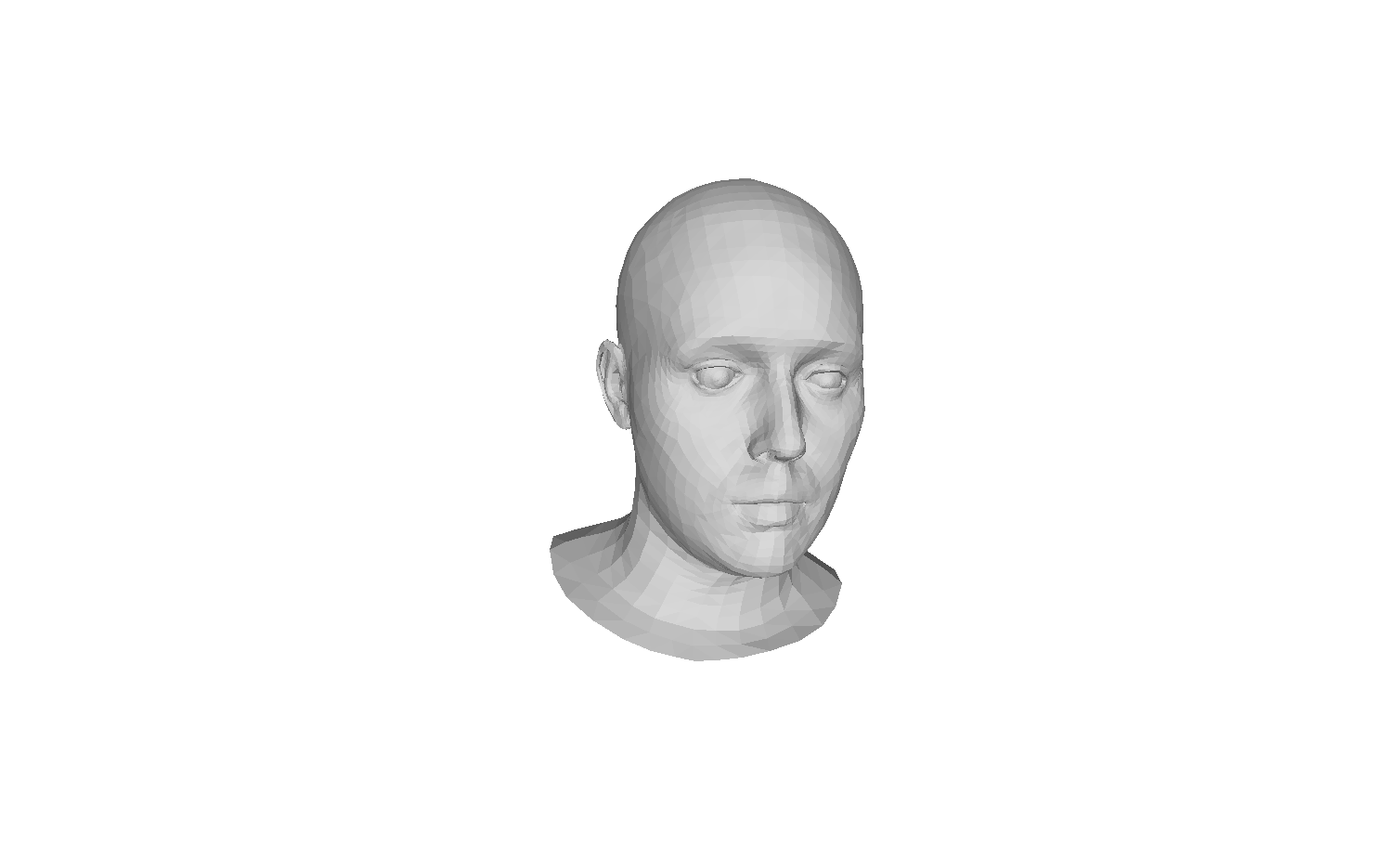}};
    \node[right of=f7, node distance=1.4cm] (f8) {\includegraphics[trim={400 80 400 100},clip,width=0.09\linewidth]{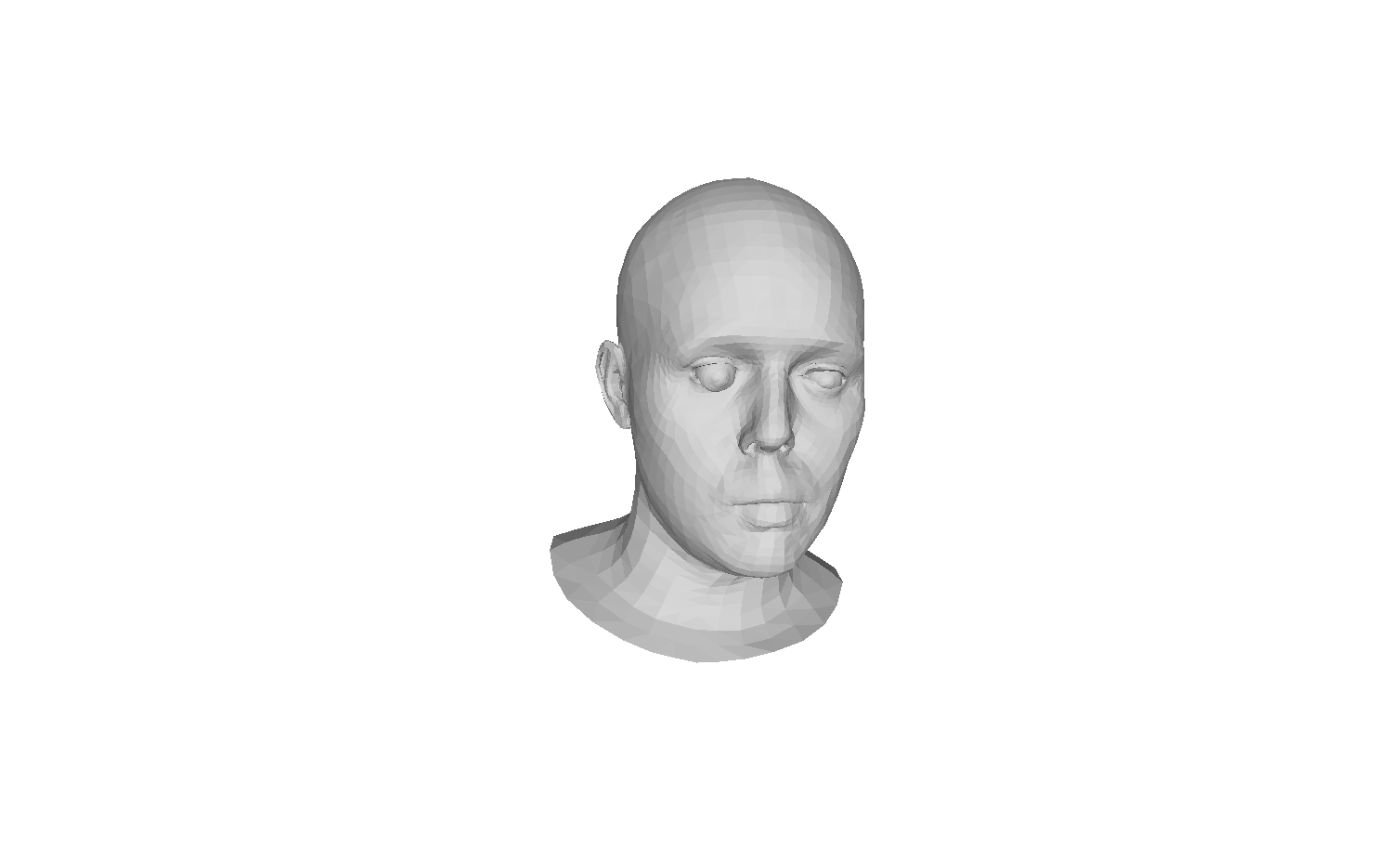}};
    \node[right of=f8, node distance=1.4cm] (f9) {\includegraphics[trim={400 80 400 100},clip,width=0.09\linewidth]{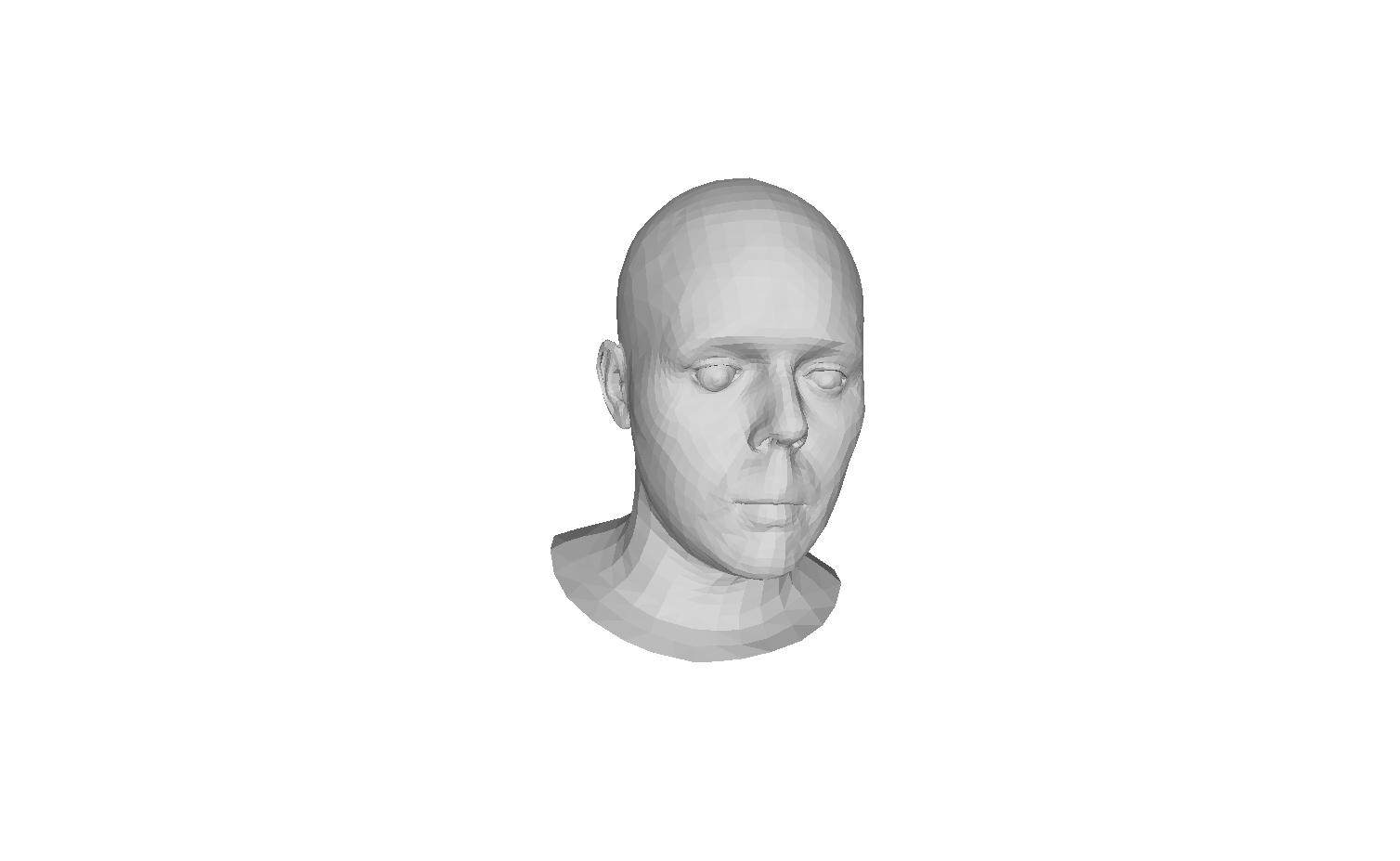}};
    \node[right of=f9, node distance=1.4cm] (f10) {\includegraphics[trim={400 80 400 100},clip,width=0.09\linewidth]{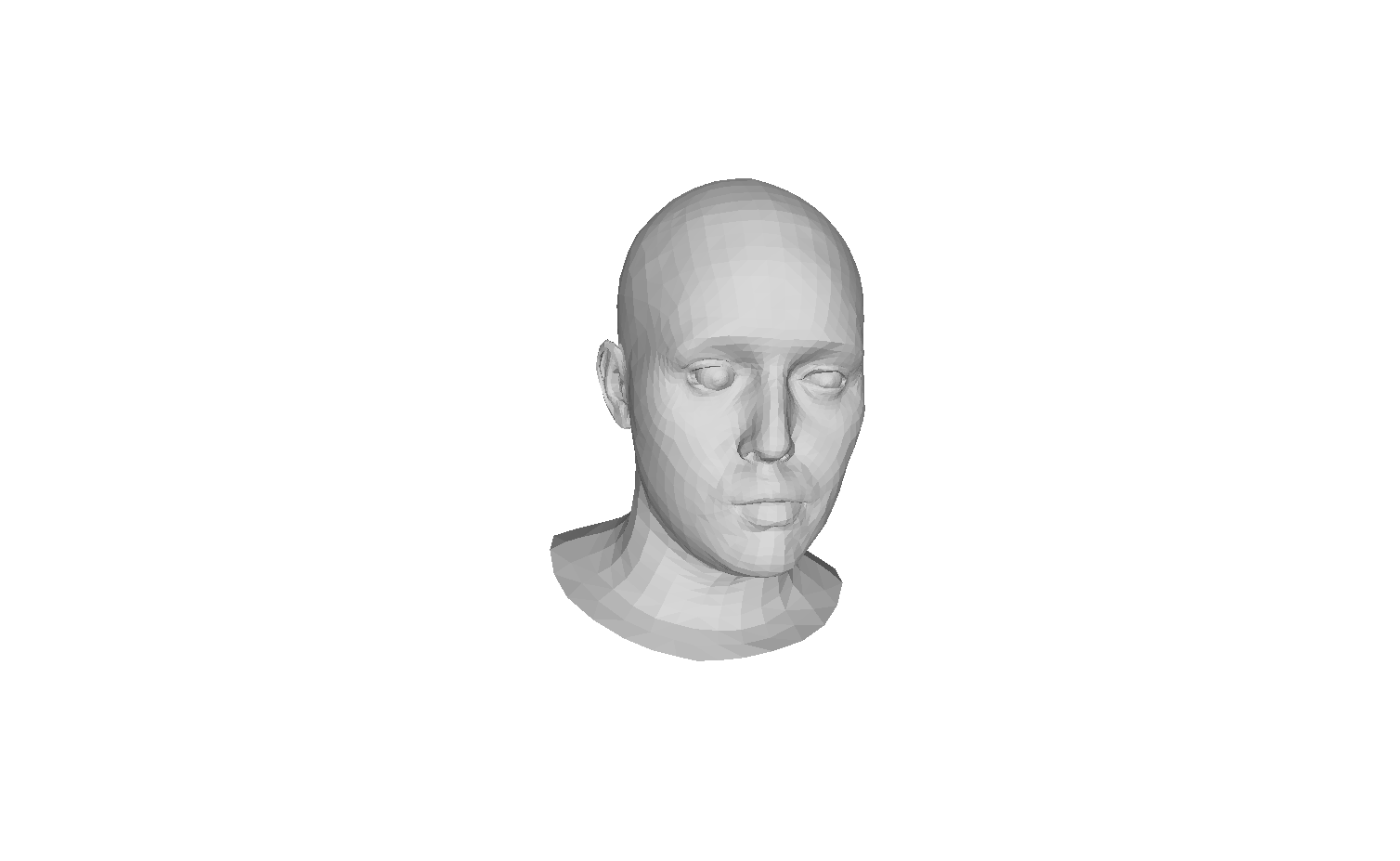}};
    \node[right of=f10, node distance=1.4cm] (f11) {\includegraphics[trim={400 80 400 100},clip,width=0.09\linewidth]{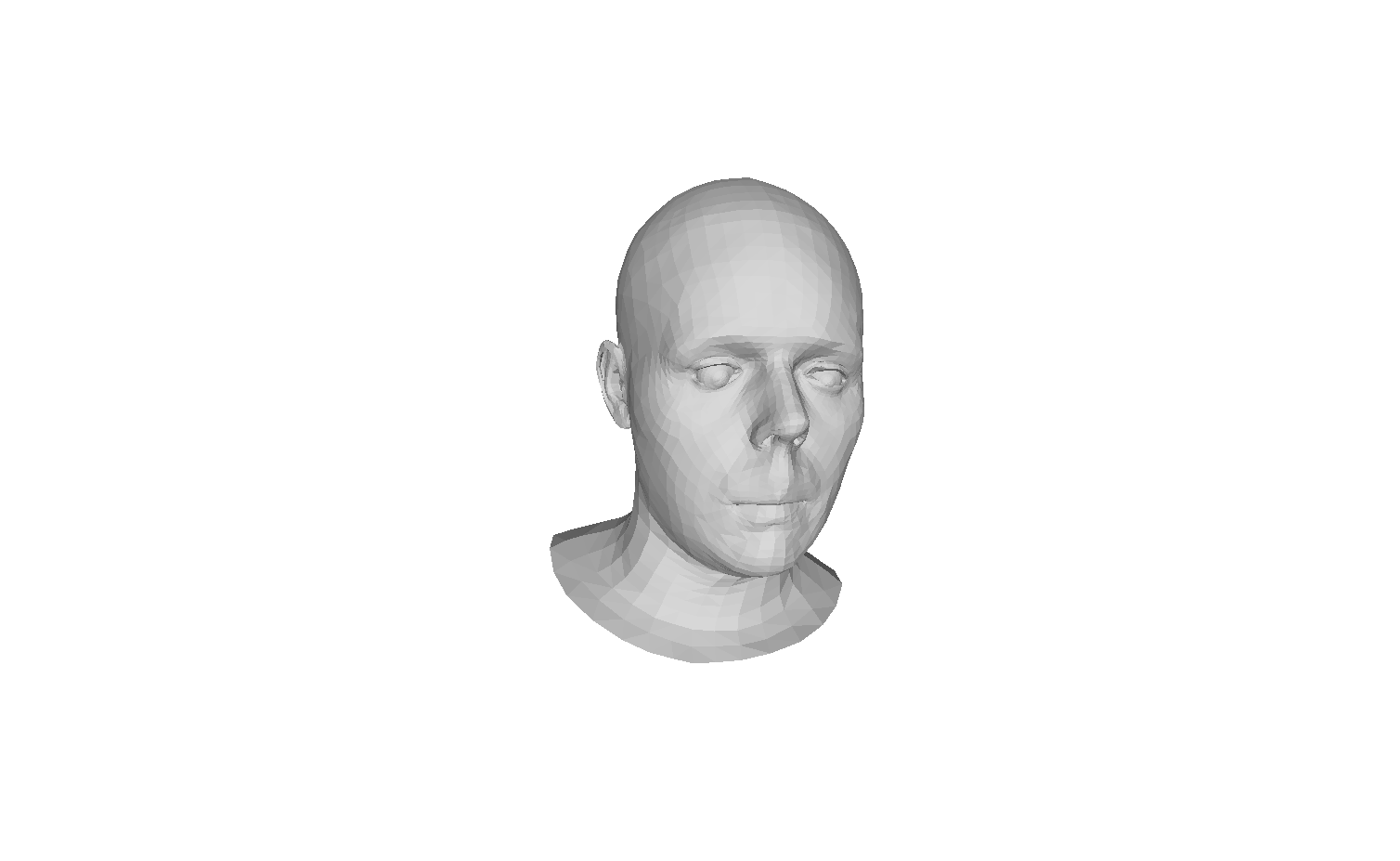}};
    
    \node[below of=f1, node distance=1.4cm] {Target Image};
    \node[below of=f2, node distance=1.4cm] {FLAME \cite{flame}};
    \node[below of=f3, node distance=1.4cm] {DECA \cite{deca}};
    \node[below of=f4, node distance=1.4cm, text width=1.2cm, align=center] {CFR-GAN \cite{occrobustwacv}};
    \node[below of=f5, node distance=1.4cm, text width=1.2cm, align=center] {Occ3DMM \cite{egger2018occlusion}};
    \node[below of=f6, node distance=1.4cm, text width=1.2cm, align=center] {Extreme3D \cite{tran2018extreme}};
    \node[below of=f9, node distance=1.4cm] {Reconstructions by \ourmethod{} (Ours)};
    \end{tikzpicture}
    \caption{\textbf{More Qualitative evaluation on the CelebA dataset \cite{celeba}}: Reconstructed singular 3D meshes from the target image by the baselines \versus{} the diverse reconstructions from \ourmethod{}.}
    \label{fig:celeba}
\end{figure*}

\subsection{Further Quantitative Analysis on Diversity\label{subsec:quantitative}}
We provide further quantitative evaluation of our approach compared to the baselines in terms of diversity performance as measured by the proposed \textit{ASD-O}, \textit{ASD-V} metrics, and the ratio \textit{ASD-O/ASD-V}, on the CelebA dataset \cite{celeba}. Since the CelebA dataset \cite{celeba} is not labeled with groundtruth 3D shape, we do not compute the Closest Sample Distance (\textit{CES}) on this dataset. To re-iterate, lower \textit{ASD-V} indicates better consistency with the visible regions; and higher \textit{ASD-O} indicates higher diversity in the occluded regions. As reported in ~\cref{tab:diversity}, our approach obtains the maximum \textit{ASD-O} across all occlusion types, the lowest \textit{ASD-V} for \textit{Glasses}, as well as the second lowest (compared to Mesh-VAE) \textit{ASD-V} for \textit{Facemasks} and \textit{Random} occlusions. This is further corroborated by the significantly higher \textit{ASD-O/ASD-V} ratios reported by \ourmethod{} compared to the baselines. Compared to this, single-stage diversity fitting baselines \textit{viz.} FLAME+DPP and Global+Local+DPP generate the lowest \textit{ASD-O/ASD-V} ratios, signifying that the 3D reconstructions generated by these approaches are neither diverse on the occluded regions, nor consistent with respect to the visible regions. On the other hand, one-pass samples generated by Global+Local+VAE are consistent with the visible face as reported by low \textit{ASD-V}, but not diverse on the occluded regions (low \textit{ASD-O}).

\subsection{Error Histogram Analysis\label{subsec:histogram}}
In ~\cref{fig:histogram}, we plot the histograms of shape fitting errors (in terms of MSE) when the FLAME \cite{flame} and our global+local model are used to fit to partially occluded face images. One can observe that, while FLAME registers smaller errors (less than 10 MSE) on more number of samples than the global+local model, there are significantly more number of samples $(\sim 15\%)$ where FLAME registers very high MSE errors ($> 50$ MSE) than the global+local model. One can conclude that our global+local model is more robust than the global FLAME model \cite{flame} on samples with challenging occlusions.

\subsection{Diversity Hyperparameters\label{subsec:diversity_hparams}}
The diversity generated by our approach is determined by the DPP loss $L_{dpp} = - tr \left( \mathbf{I} - (\mathbf{L} + \mathbf{I})^{-1} \right)$. Here, the DPP kernel entry for the $i,j$-th element is given by $L_{i,j}=q_iS_{i,j}q_j$, where $q_i$ denotes the quality of element $i$, and $S_{i,j}$ represents the similarity between $i$ and $j$. The DPP optimization tries to maximize the quality of each sample, while minimizing the similarity between distinct samples. As stated in the main paper, we control the similarity term $S_{i,j} = \exp \left(- \frac{k}{\median_{i,j}( dist_{i,j} )} dist_{i,j} \right)$ and the quality term $q_i = \exp(- \max(0, \mathbf{z}_i^T\mathbf{z}_i - n_{\sigma}\sqrt{d}))$ using two parameters $k$ and $n_{\sigma}$, respectively. In ~\cref{tab:diversity_hparams}, we study the effects of the two hyper-parameters $k$ and $n_{\sigma}$ on diversity as measured by the diversity metrics \textit{ASD-V} and \textit{ASD-O}. As shown in ~\cref{tab:diversity_hparams}, we obtain maximum \textit{ASD-V}, as well as, \textit{ASD-O} at $k=0.5$; whereas both metrics increase as $n_{\sigma}$ increases. Thus, we set $k=0.5$ in our experiments while we choose $n_{\sigma}=3$ as a sweet spot between minimizing \textit{ASD-V} and maximizing \textit{ASD-}O. The user can change the value of $n_{\sigma}$ to tweak the diversity-realism trade-off.

\subsection{Real-world Occlusions\label{subsec:realocc}}
We present examples of diverse 3D reconstructions by our approach on real-world occluded face images in ~\cref{fig:realocc}. For these images, we inferred the occlusion mask using the face segmentation model by Nirkin \etal \cite{facesegswappingperception}. These results further demonstrate the efficacy of \ourmethod{} to generate diverse, yet plausible 3D reconstructions on real world occlusions ranging from glasses, scarf, facemasks, \etc.

\subsection{Moving the Occlusion Around the Face\label{subsec:moveocc}}
In this section, we evaluate the diversity and robustness performance of \ourmethod{} to occlusions at different locations on the face. ~\cref{fig:moveocc} shows the set of 3D reconstruction by \ourmethod{} when the occlusion moves around the face occupying the left cheek, mouth, the right cheek, center and the periocular (eye) regions of the face. Our method generates diverse, yet plausible set of 3D reconstructions for all the cases. We particularly note the high degree of diversity in expression that occurs when the mouth region is occluded, as is expected.

\subsection{Diversity Interpolations\label{subsec:interpolation}}
A potential application of \ourmethod{} is to perform controlled diversification around an occluded region during 3D reconstruction. To do this, we can first generate a set of diverse 3D reconstructions for an occluded target image and then allow the user to select two distinct samples to perform interpolation in-between. We perform interpolation in the latent space: $\mathbf{z}(\alpha)=\alpha\mathbf{z}_1 + (1-\alpha)\mathbf{z}_2$. This affords the user control over the extent and type of diversity. We present examples of such interpolations in ~\cref{fig:interpolate}.

\subsection{Further Qualitative Results on CelebA Dataset\label{subsec:qualitative}}
We show further qualitative results of diverse 3D reconstructions on occluded face images from the CelebA dataset \cite{celeba} by \ourmethod{}, compared to the singular reconstruction by FLAME \cite{flame}, DECA \cite{deca}, CFR-GAN \cite{occrobustwacv}, Occ3DMM \cite{egger2018occlusion} and Extreme3D \cite{tran2018extreme} in ~\cref{fig:celeba}. While the baselines often  get the pose, shape or expression wrong, \ourmethod{} generates 3D reconstructions that are consistent with the visible regions, yet plausibly diverse on the occluded regions.
\section{Implementation Details\label{sec:implementation}}

\subsection{Optimization\label{subsec:optimization}}
We use the \textit{PyTorch} library to implement our approach. In our experiments, we found that the SGD optimizer, with a learning rate of $5\times 10^{-3}$ gives the best results as compared to the Adam and RMSprop optimizers. For photometric fitting, we used the texture model provided by \href{https://flame.is.tue.mpg.de/index.html}{FLAME}. We run the fitting stage (Algorithm 1) for $n_{iter}=2000$ iterations and the diversity stage (Algorithm 2) for $n_{comp}=300$ iterations. In Algorithm 1, we set the loss weights as follows: $\lambda_1^f = 5, \lambda_2^f=16, \lambda_3^f=10^{-3}$. During the diversifying shape completion stage (Algorithm 2), we set $\lambda_1 = 1000, \lambda_2 = 500, \lambda_3 = 0.025$. Further, we found that using a slightly smaller learning rate for the eyeball components while fitting the global+local model gives better results. For these components, we set the learning rate to be 0.5 times that of the other components.

\subsection{Mesh-VAE\label{subsec:network}}
The Mesh-VAE model is based on the fully convolutional mesh autoencoder (Meshconv) architecture proposed by Zhou \etal \cite{meshconv}. Meshconv \cite{meshconv} uses spatially varying convolutional kernels for different mesh vertices to account for the irregular structure of a 3D mesh. The spatially varying kernels are sampled from the span of a shared weight basis, using learned per-vertex coefficients. In addition, Meshconv defines pooling and unpooling operations on a 3D mesh by performing feature aggregation Monte Carlo sampling \cite{meshconv}. 

We trained the Mesh-VAE with FLAME \cite{flame} registered groundtruth scans provided in the CoMA \cite{coma} and D3DFACS \cite{d3dfacs} datasets. We perturbed the input meshes with uniformly sampled rectangular masks (in XY) within a range around the mesh center, while gradually increasing the size of the mask per training epoch until it covered $\sim$40\% of the vertices. We detail the network architecture for the Mesh-VAE in ~\cref{tab:encoder,tab:decoder}.

The abbreviated operators used are defined as follows:
\begin{itemize}
    \item vcDownConv($in_c, out_c, s, r, M$) + vcDownRes($s$): Downward residual block (as defined in Meshconv \cite{meshconv}), with $in_c$ input channels, $out_c$ output channels, $s$ stride, $r$ kernel radius and $M$ number of shared weight bases. The output is activated with ELU \cite{elu} activation.
    
    \item vcUpConv($in_c, out_c, s, r, M$) + vcUpRes($s$): Upward residual block (as defined in Meshconv \cite{meshconv}), with $in_c$ input channels, $out_c$ output channels, $s$ stride, $r$ kernel radius and $M$ number of shared weight bases. The output is activated with ELU \cite{elu} activation.
\end{itemize}

\begin{table*}[]
    \caption{Network architecture of the Mesh-VAE Encoder $\mathcal{E}_{mesh}$.}
    \label{tab:encoder}
    \begin{tabular}{clcc}
         \thickhline
         Input & Layer & Output size & Output\\
         \hline
         $5023\times 3$ \textbf{Mesh} & $\rightarrow$ vcDownConv($in_c=3, out_c=32, s=2, r=43, M=17$) + vcDownRes(2) & $1367\times 32 $\\
         & vcDownConv($in_c=32, out_c=64, s=1, r=27, M=17$) + vcDownRes(1) & $1367\times 64 $\\
         & vcDownConv($in_c=64, out_c=128, s=2, r=54, M=17$) + vcDownRes(2) & $270\times 128 $\\
         & vcDownConv($in_c=128, out_c=256, s=1, r=25, M=17$) + vcDownRes(1) & $270\times 256 $\\
         & vcDownConv($in_c=256, out_c=512, s=2, r=81, M=17$) + vcDownRes(2) & $45\times 512 $\\
         & vcDownConv($in_c=512, out_c=1024, s=1, r=27, M=17$) + vcDownRes(1) & $45\times 1024 $ & \textit{feats}\\
         \textit{feats} & vcDownConv($in_c=1024, out_c=64, s=2, r=37, M=17$) + vcDownRes(2) & $10\times 64 $ & $\boldsymbol{\mu}$\\
         \textit{feats} & vcDownConv($in_c=1024, out_c=64, s=2, r=37, M=17$) + vcDownRes(2) & $10\times 64 $ & $\log \boldsymbol{\sigma}^2$\\
         \hline
         \textbf{Model Complexity} & 9M\\
         \thickhline
    \end{tabular}%
\end{table*}

\begin{table*}[]
    \caption{Network architecture of the Mesh-VAE Decoder $\mathcal{D}_{mesh}$.}
    \label{tab:decoder}
    \begin{tabular}{clcc}
         \thickhline
         Input & Layer & Output size & Output\\
         \hline
         $10\times 64$ $\mathbf{z}$ & vcUpConv($in_c=64, out_c=1024, s=2, r=8, M=17$) + vcUpRes(2) & $45\times 1024$\\
         & vcUpConv($in_c=1024, out_c=512, s=1, r=27, M=17$) + vcUpRes(1) & $45\times 512$\\
         & vcUpConv($in_c=512, out_c=256, s=2, r=16, M=17$) + vcUpRes(2) & $270\times 256$\\
         & vcUpConv($in_c=256, out_c=128, s=1, r=25, M=17$) + vcUpRes(1) & $270 \times 128$\\
         & vcUpConv($in_c=128, out_c=64, s=2, r=12, M=17$) + vcUpRes(2) & $1367 \times 64$\\
         & vcUpConv($in_c=64, out_c=32, s=1, r=27, M=17$) + vcUpRes(1) & $1367 \times 32$ &\\
         & vcUpConv($in_c=32, out_c=3, s=2, r=24, M=17$) + vcUpRes(2) & $5023 \times 3$ & \textbf{Output}\\
         \hline
         \textbf{Model Complexity} & 8M\\
         \thickhline
    \end{tabular}%
\end{table*}

{\small
\bibliographystyle{ieee_fullname}
\bibliography{egbib}
}

\end{document}